\documentclass[wcp]{jmlr}

\usepackage{longtable}

\usepackage{booktabs}
\usepackage{kotex}

\makeatletter
\let\Ginclude@graphics\@org@Ginclude@graphics 
\makeatother

\jmlrvolume{222}
\jmlryear{2023}
\jmlrworkshop{ACML 2023}

\title[Edit-A-Video: Single Video Editing with Object-Aware Consistency]{Edit-A-Video: Single Video Editing with Object-Aware Consistency}

\author{\Name{Chaehun Shin$^{*}$} \Email{chaehuny@snu.ac.kr}\\
\Name{Heeseung Kim$^{*}$} \Email{gmltmd789@snu.ac.kr}\\
\Name{Che Hyun Lee} \Email{saga1214@snu.ac.kr}\\
\Name{Sang-gil Lee} \Email{tkdrlf9202@snu.ac.kr}\\
\addr Data Science and AI Lab, ECE, Seoul National University, Seoul 08826, Korea\\
\Name{Sungroh Yoon$^{\dag}$} \Email{sryoon@snu.ac.kr}\\
\addr Data Science and AI Lab, ECE and  Interdisciplinary Program in AI, Seoul National University, Seoul 08826, Korea}

\editors{Berrin Yan{\i}ko\u{g}lu and Wray Buntine}

\begin{document}
\maketitle

\begin{abstract}
With advancements in text-to-image (TTI) models, text-to-video (TTV) models have recently been introduced.
Motivated by approaches on TTV models adapting from diffusion-based TTI models, we suggest the text-guided video editing framework given only a pretrained TTI model and a single $<$text, video$>$ pair, which we term \textbf{Edit-A-Video}.
The framework consists of two stages: (1) inflating the 2D model into the 3D model by appending temporal modules and tuning on the source video (2) inverting the source video into the noise and editing with target text through attention map injection.
Each stage enables the temporal modeling and preservation of semantic attributes of the source video.
One of the key challenges for video editing is a background inconsistency problem, where the regions unrelated to the edit suffer from undesirable and inconsistent temporal alterations.
To mitigate this issue, we also introduce a novel mask blending method, termed as temporal-consistent blending (TC Blending). 
We improve previous mask blending methods to reflect the temporal consistency, ensuring that the area where the editing is applied exhibits smooth transition while also achieving spatio-temporal consistency of the unedited regions.
We present extensive experimental results over various types of text and videos, and demonstrate the superiority of the proposed method compared to baselines in terms of background consistency, text alignment, and video editing quality. Our samples are available on \href{https://editavideo.github.io}{https://editavideo.github.io}.
\end{abstract}

\renewcommand{\thefootnote}{\fnsymbol{footnote}}
\footnotetext[1]{Equal Contribution, $\dag$. Corresponding Author.}
\renewcommand{\thefootnote}{\arabic{footnote}}

\section{Introduction}

Recently, generative models have made remarkable progress across various domains. 
Diffusion models \citep{thermodynamics, ddpm}, in particular, have shown state-of-the-art generation performance across multiple tasks, including text-to-image (TTI) generation \citep{rombach2022high, saharia2022photorealistic, ramesh2022hierarchical}. 
In addition to generating images from text prompts, several works have used a pretrained TTI model for extended applications, such as personalized text-to-image \citep{ruiz2022dreambooth, gal2022image, kumari2022multi} or text-guided image editing \citep{hertz2022prompt, mokady2022null, kawar2022imagic}.

\begin{figure}[t]
\begin{center}
\makebox[0.12\textwidth]{\colorbox{pink}{\textbf{Training video}} A man is doing a pushup.}\\
\includegraphics[width=0.11\textwidth]{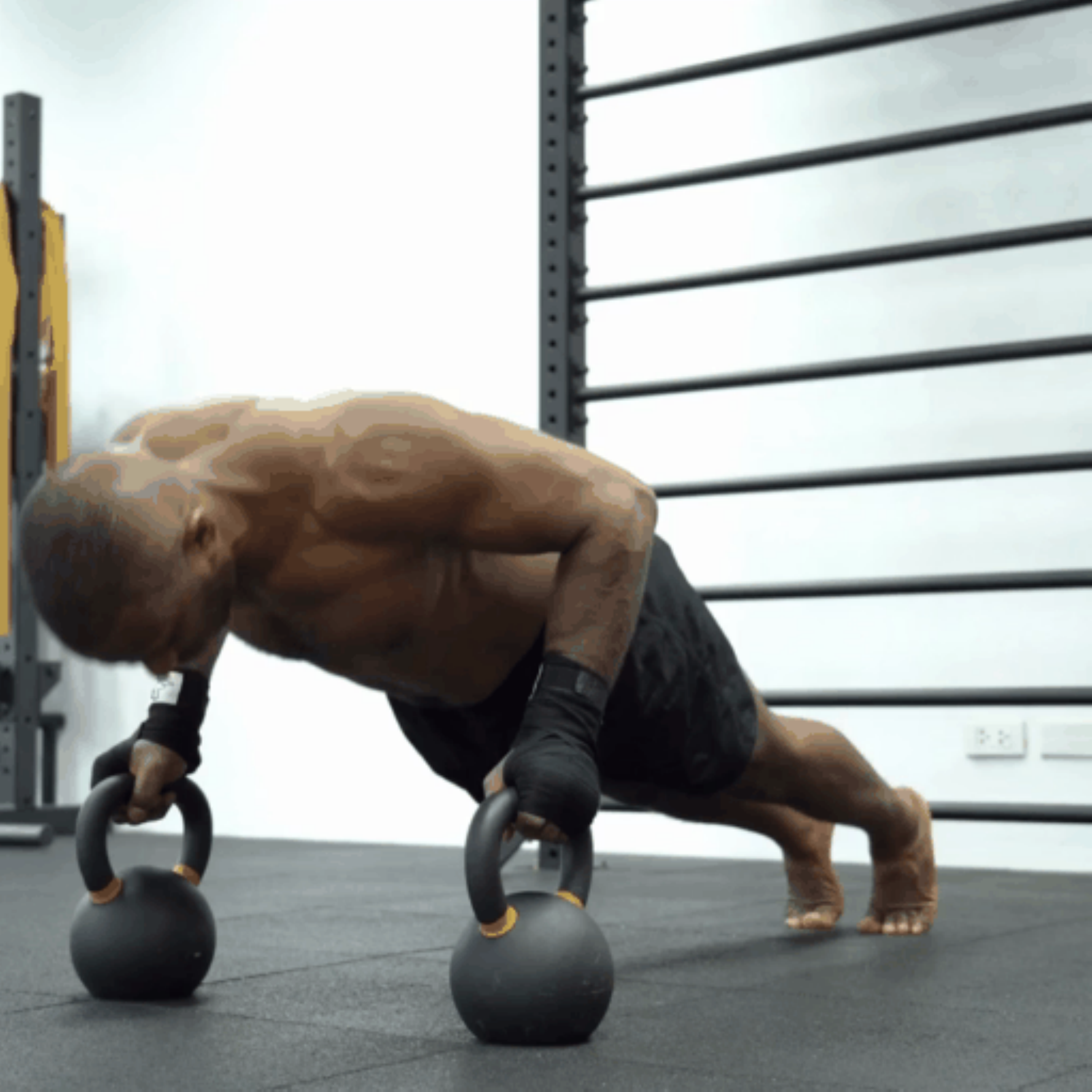}
\includegraphics[width=0.11\textwidth]{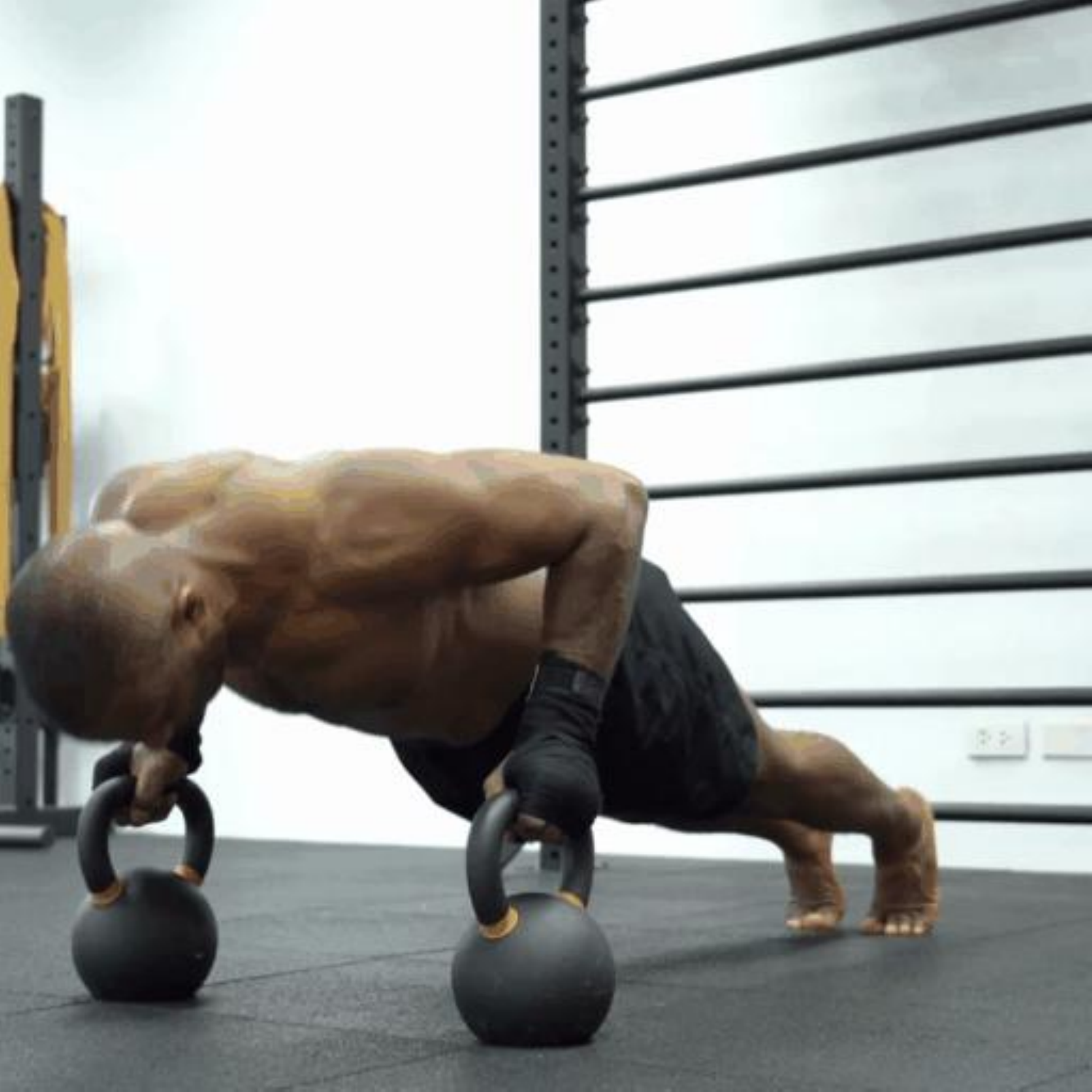}
\includegraphics[width=0.11\textwidth]{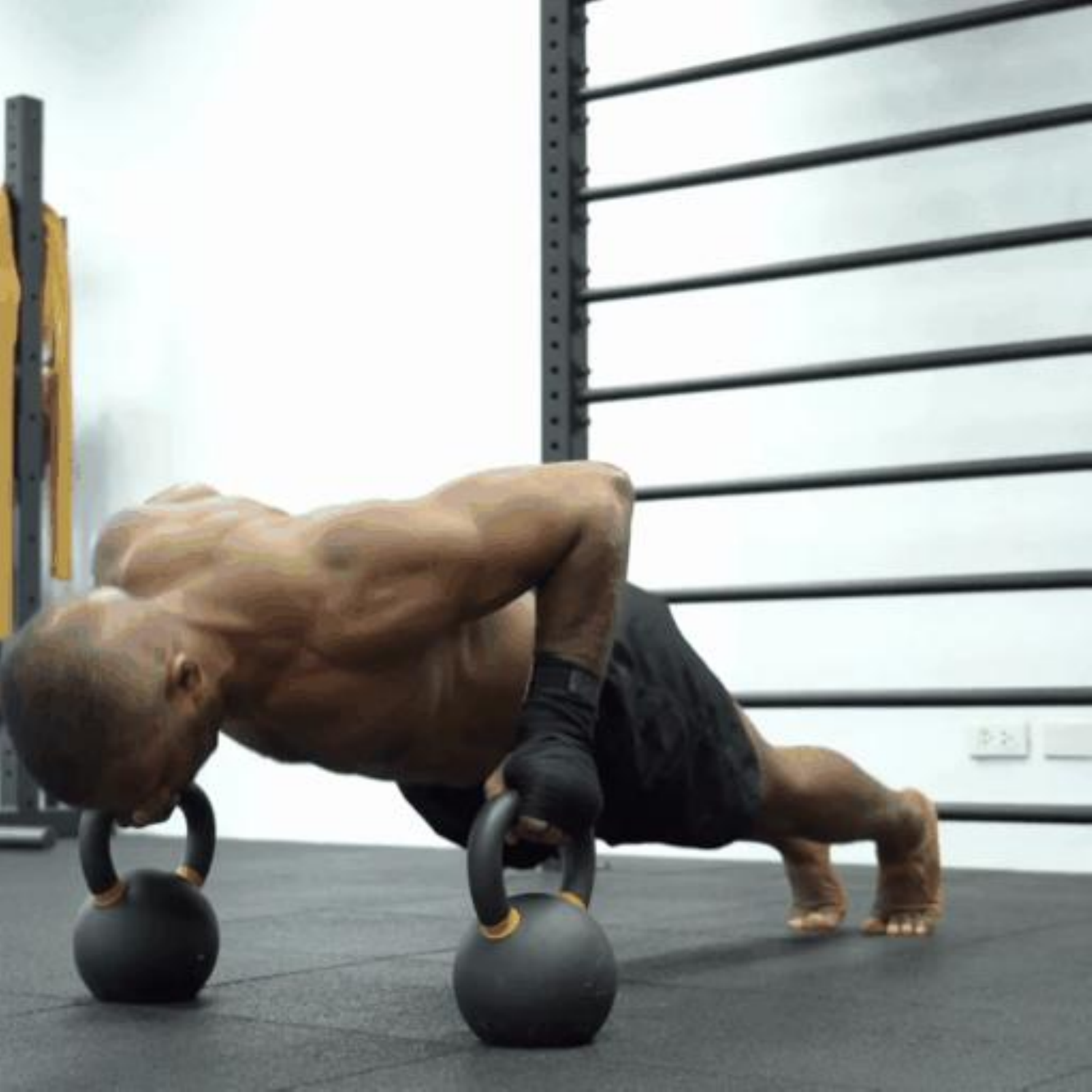}
\includegraphics[width=0.11\textwidth]{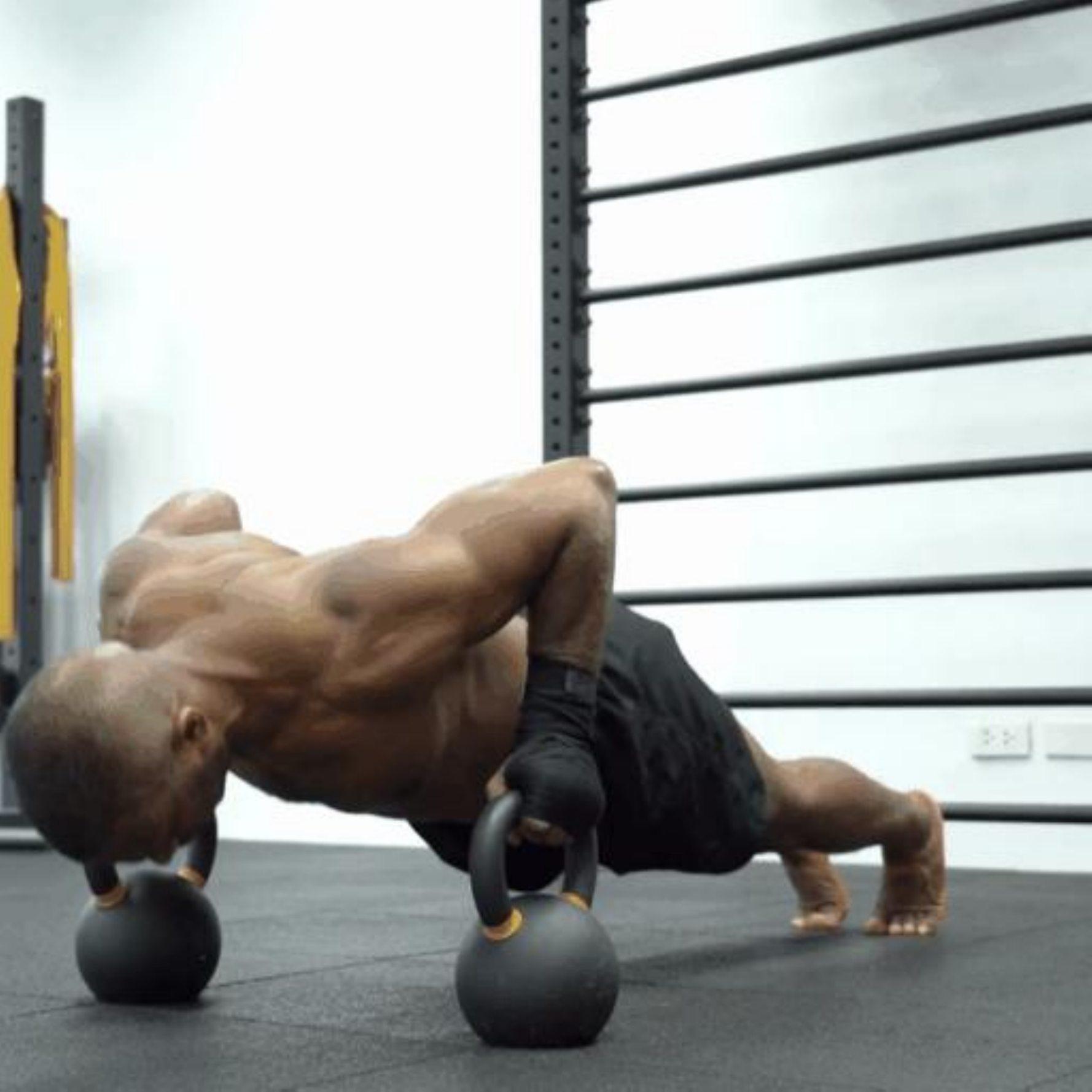}
\includegraphics[width=0.11\textwidth]{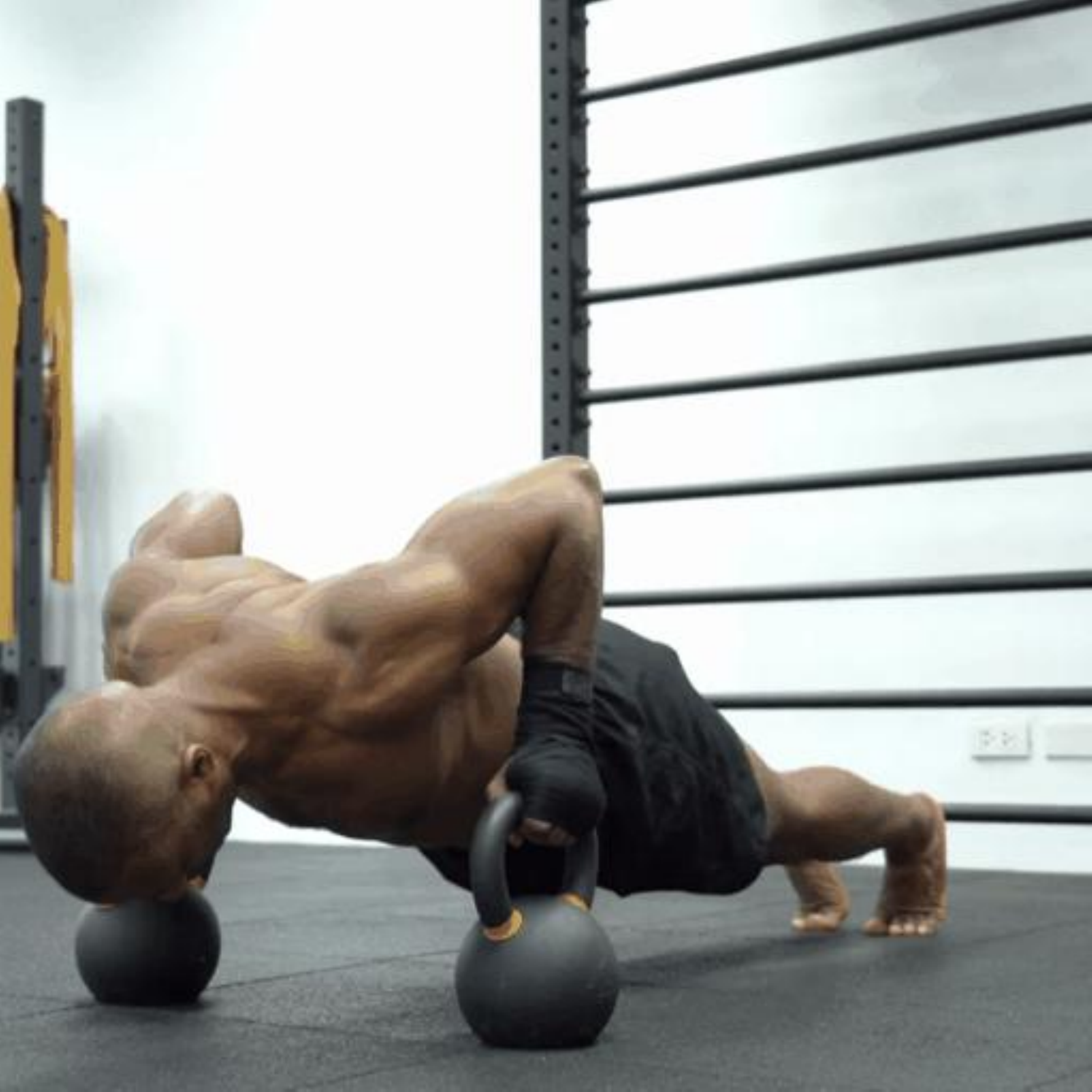}
\includegraphics[width=0.11\textwidth]{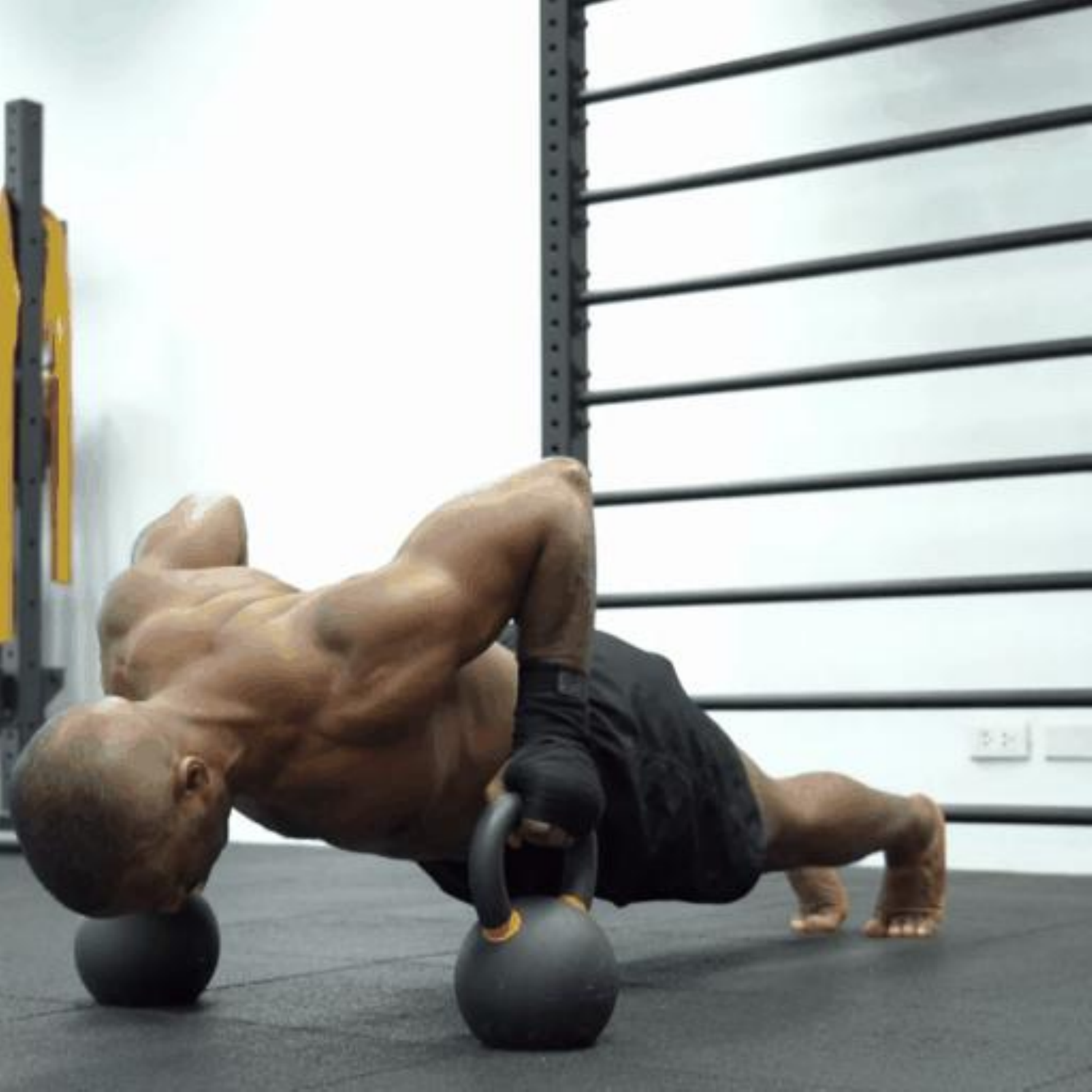}
\includegraphics[width=0.11\textwidth]{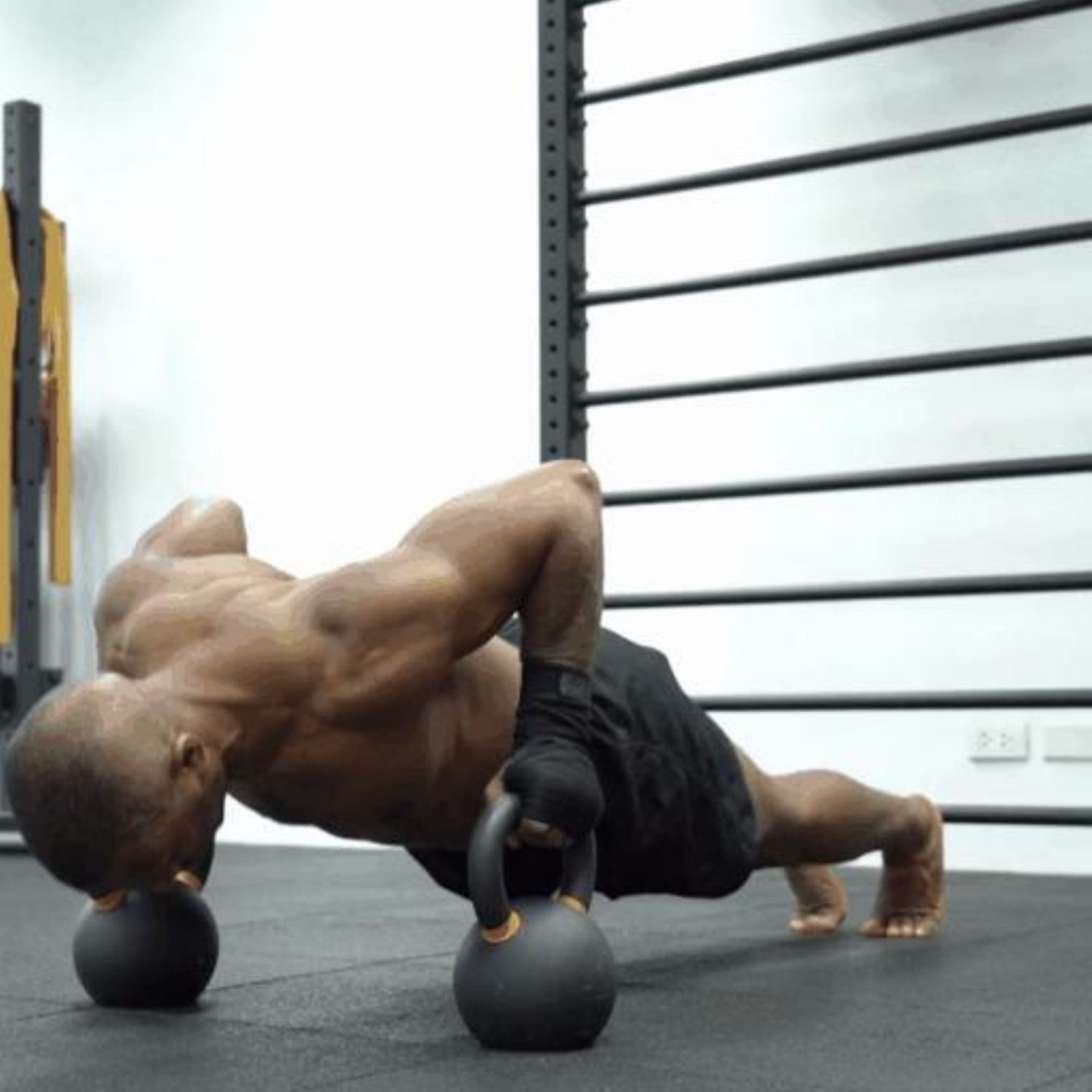}
\includegraphics[width=0.11\textwidth]{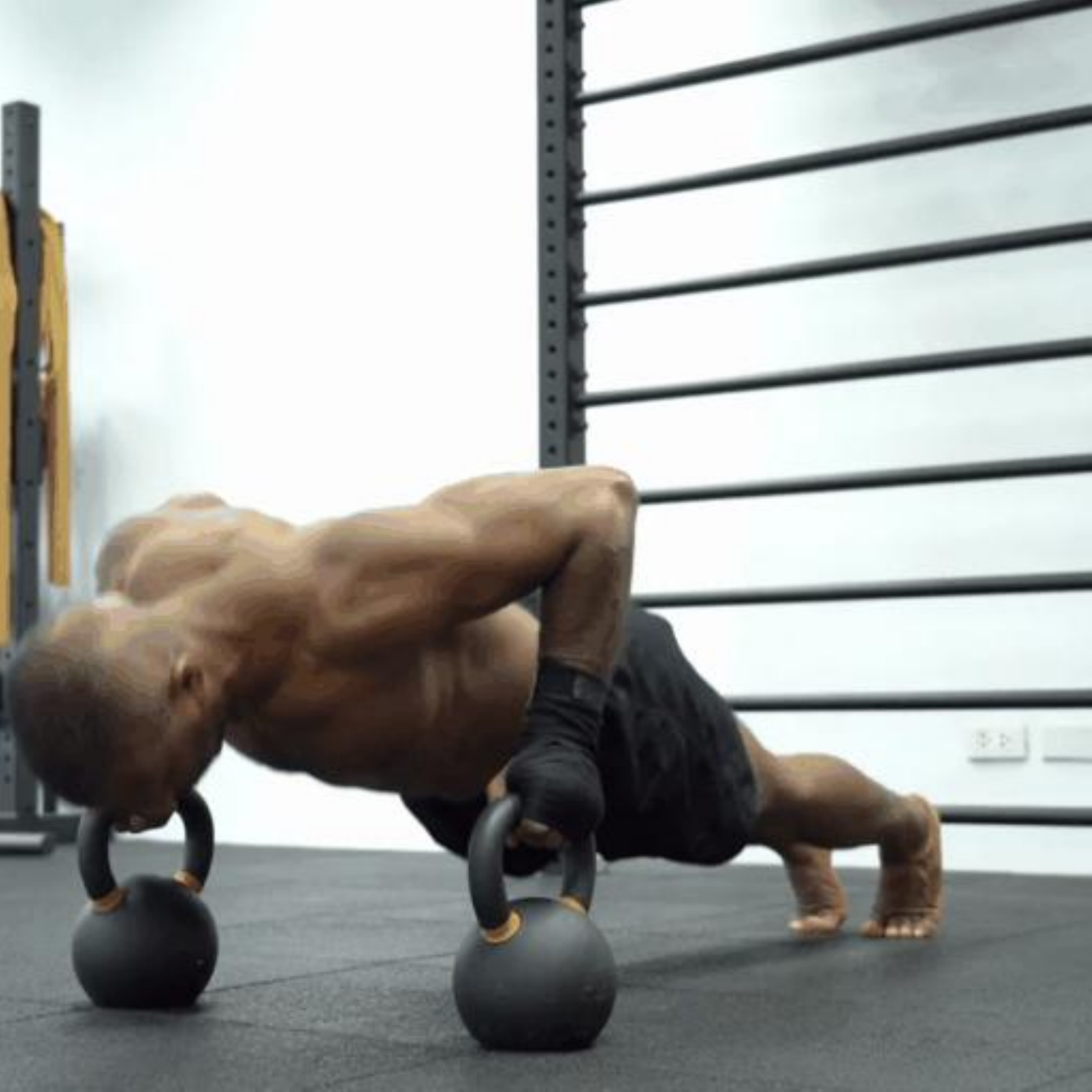}

\makebox[0.12\textwidth]{A \textcolor{blue}{\textbf{Iron Man}} is doing a pushup.}\\
\includegraphics[width=0.11\textwidth]{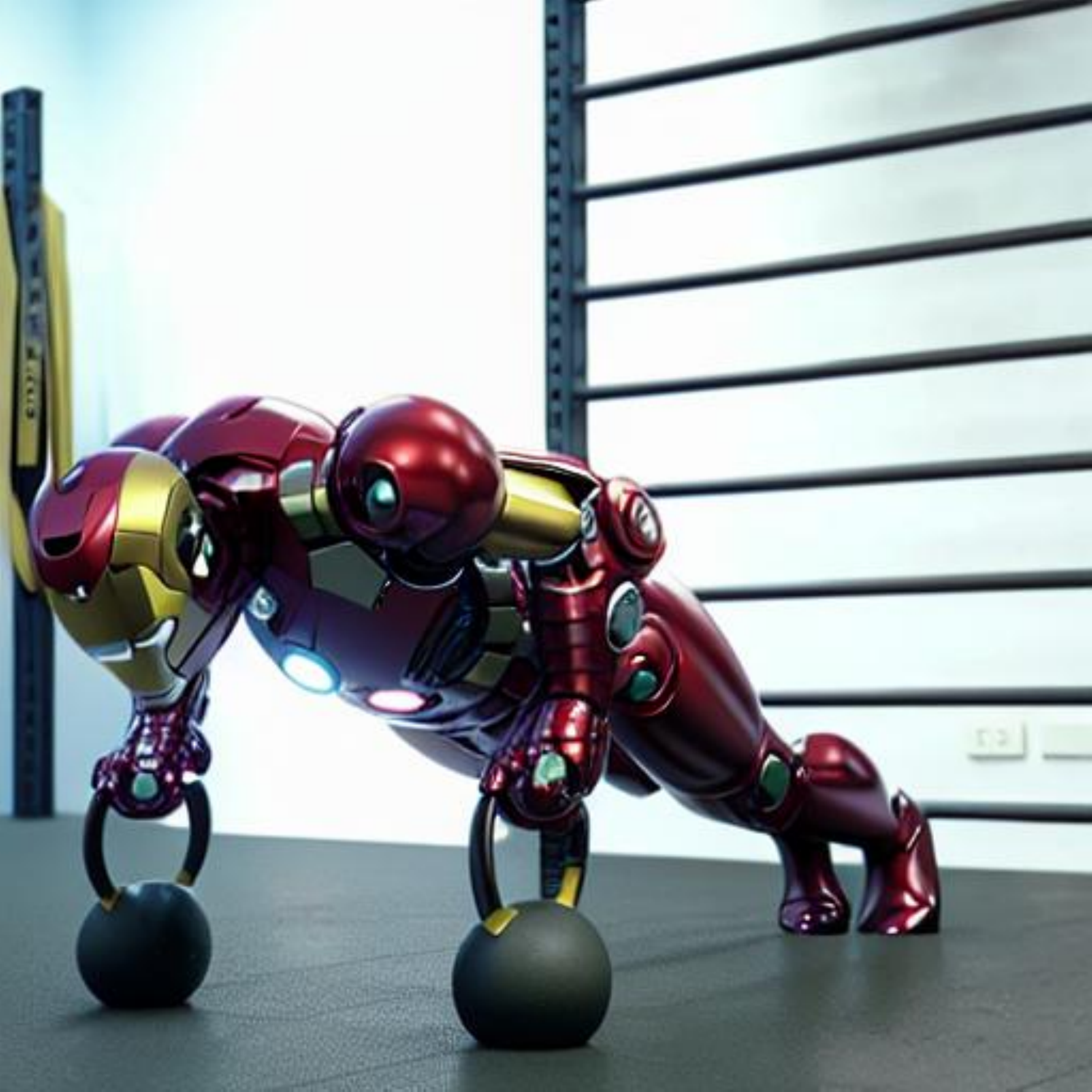}
\includegraphics[width=0.11\textwidth]{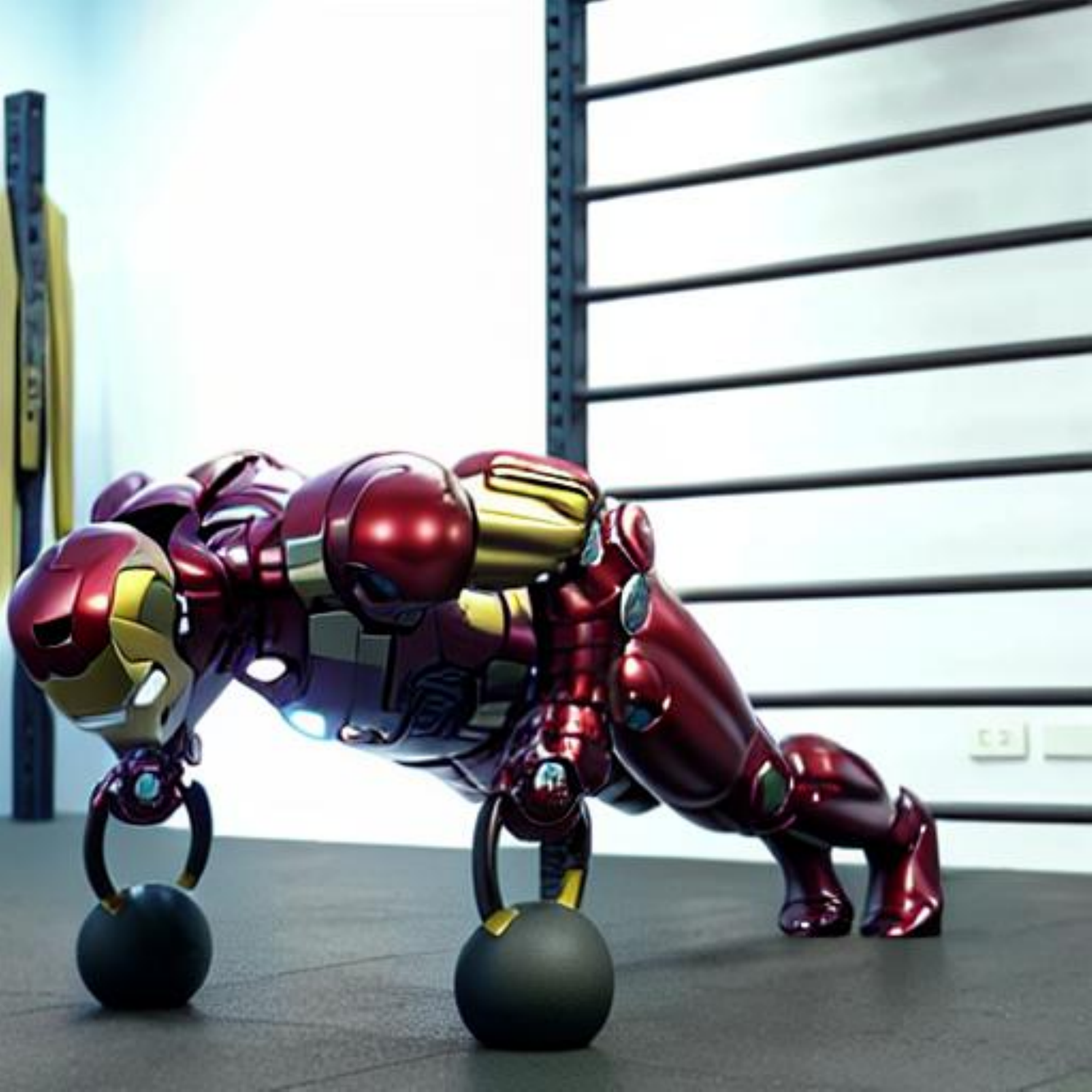}
\includegraphics[width=0.11\textwidth]{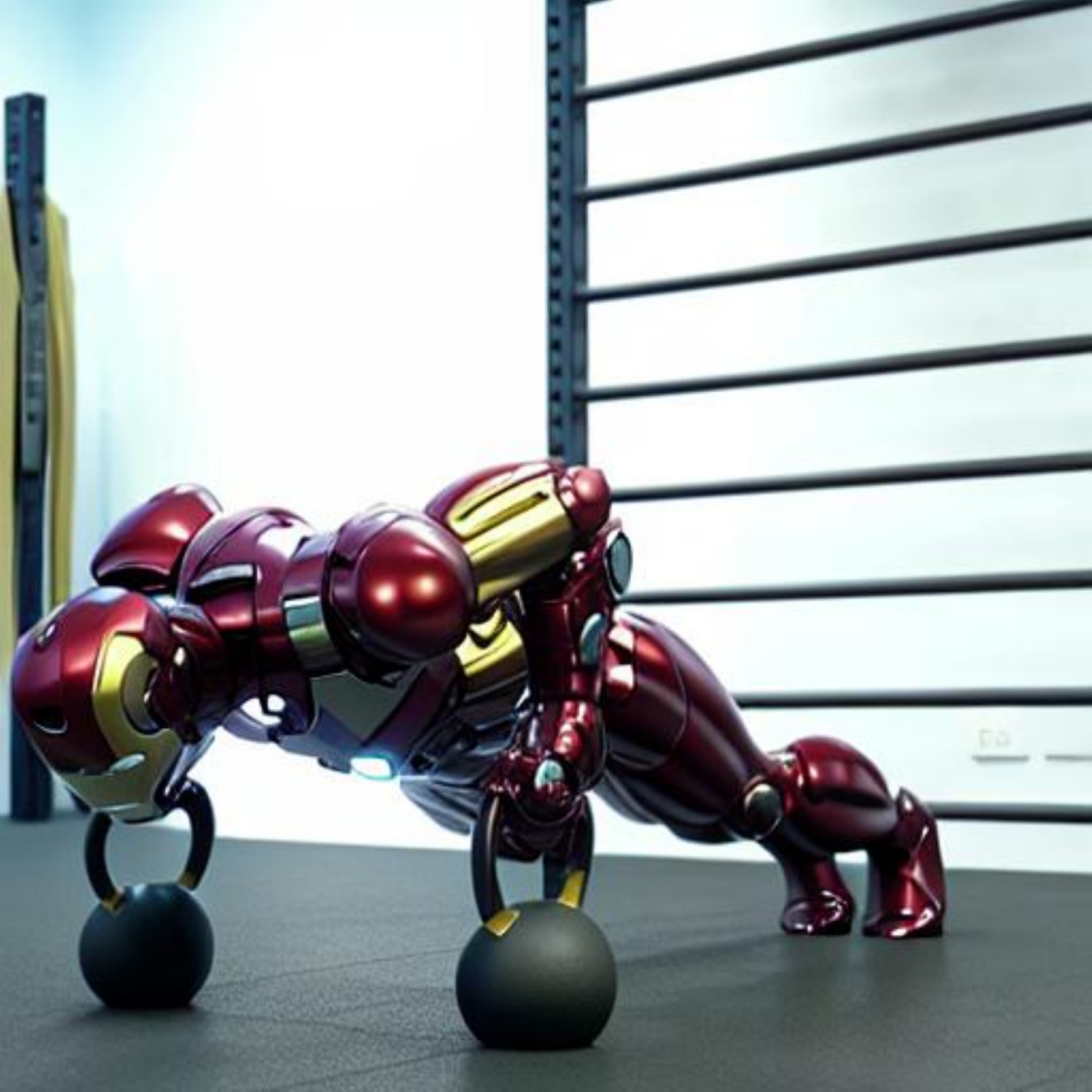}
\includegraphics[width=0.11\textwidth]{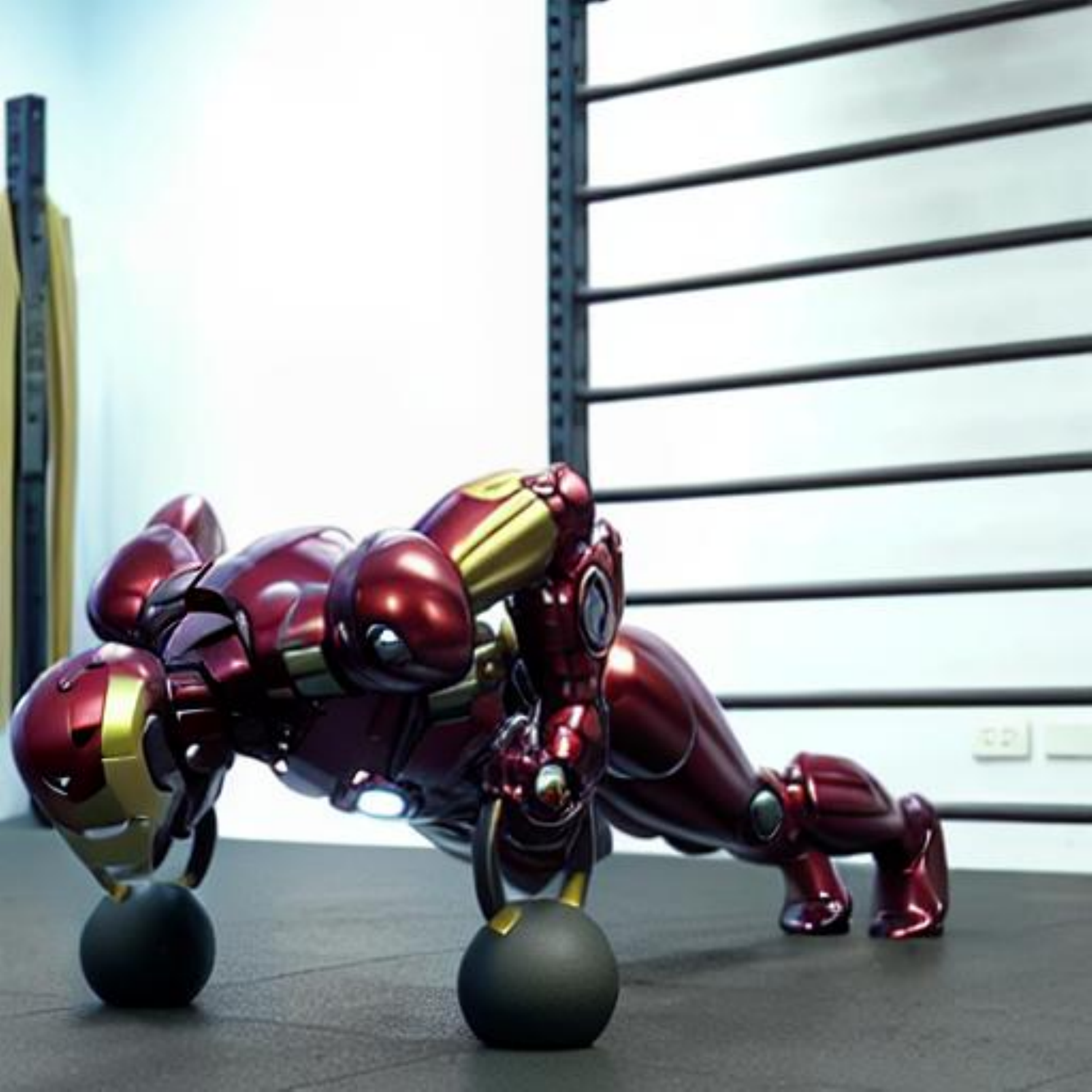}
\includegraphics[width=0.11\textwidth]{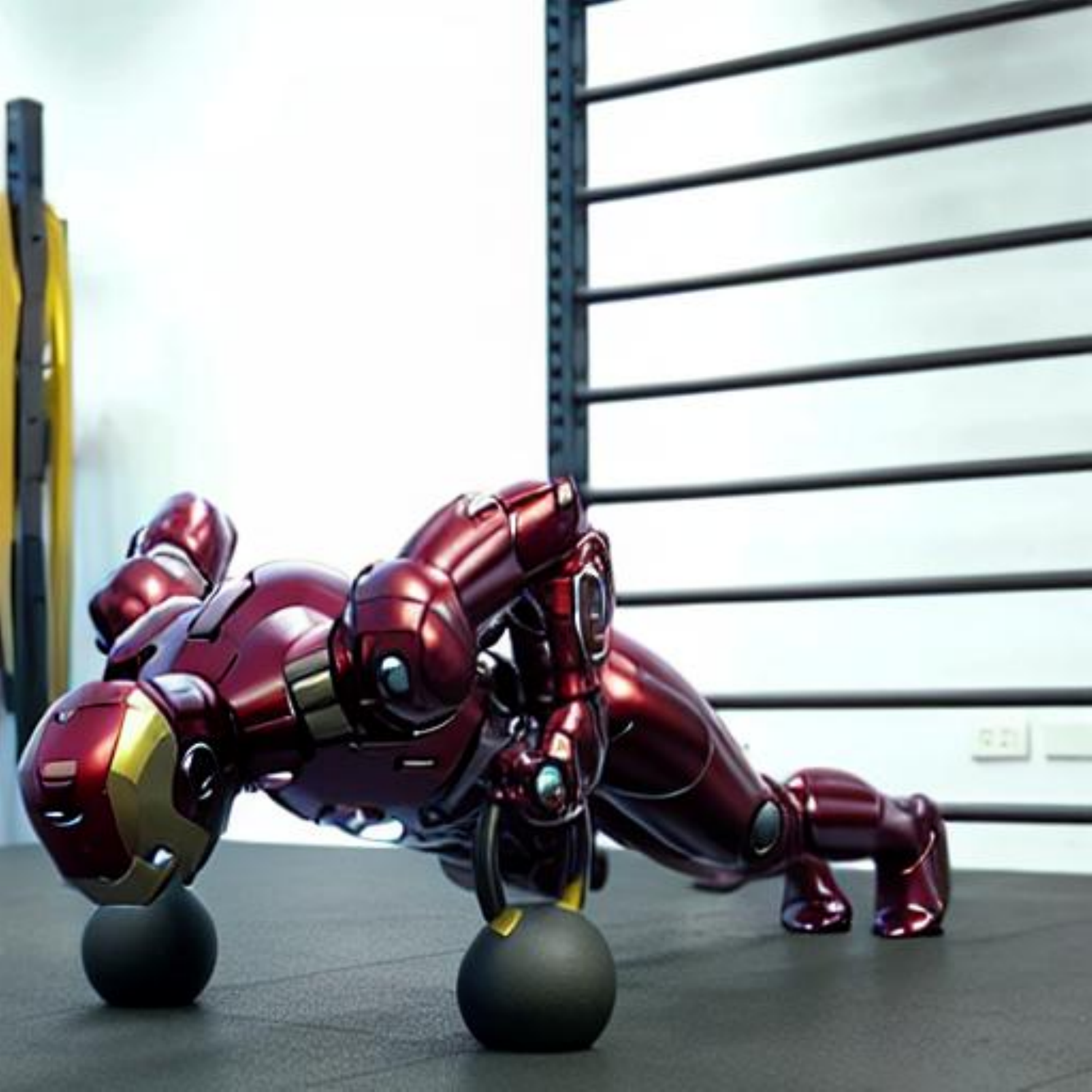}
\includegraphics[width=0.11\textwidth]{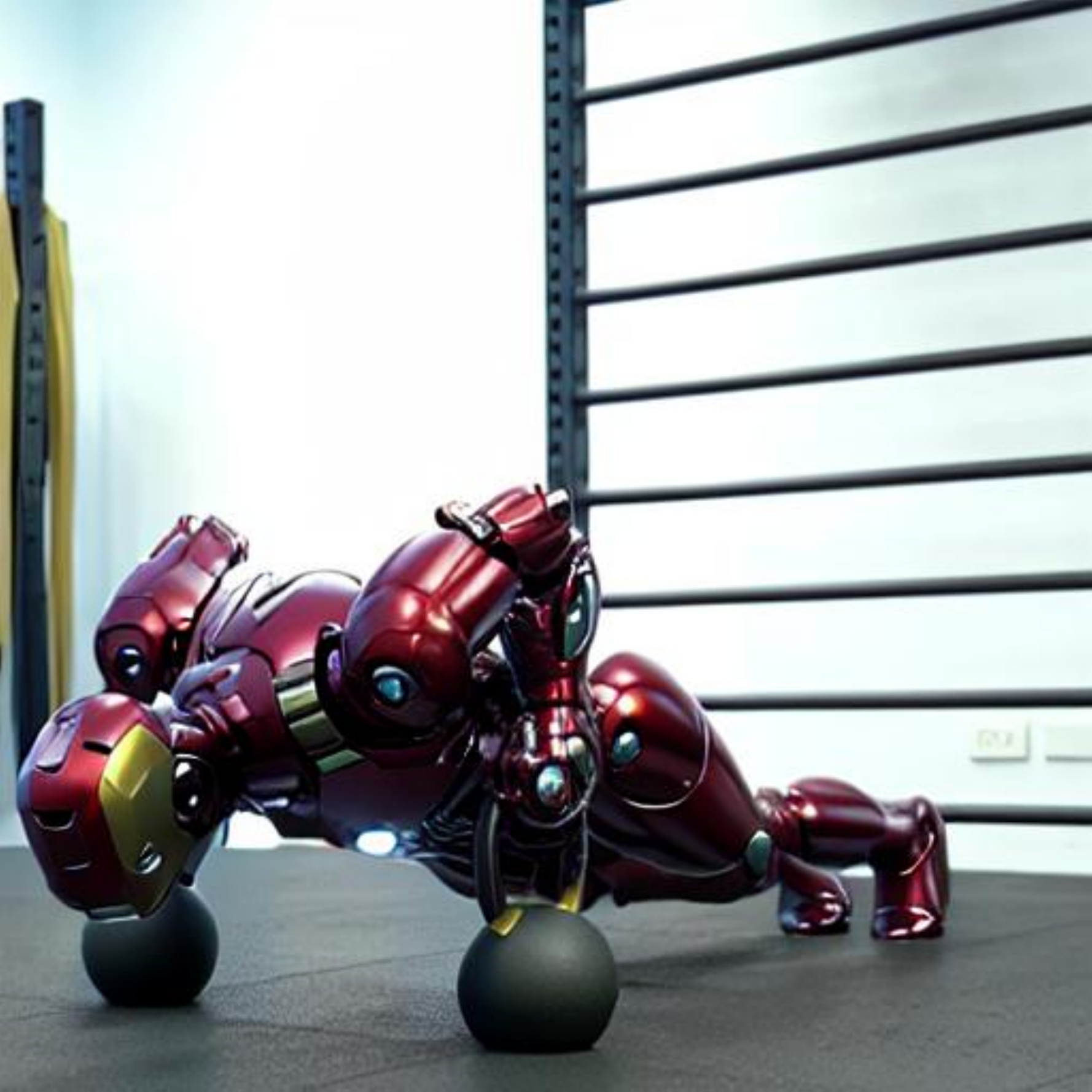}
\includegraphics[width=0.11\textwidth]{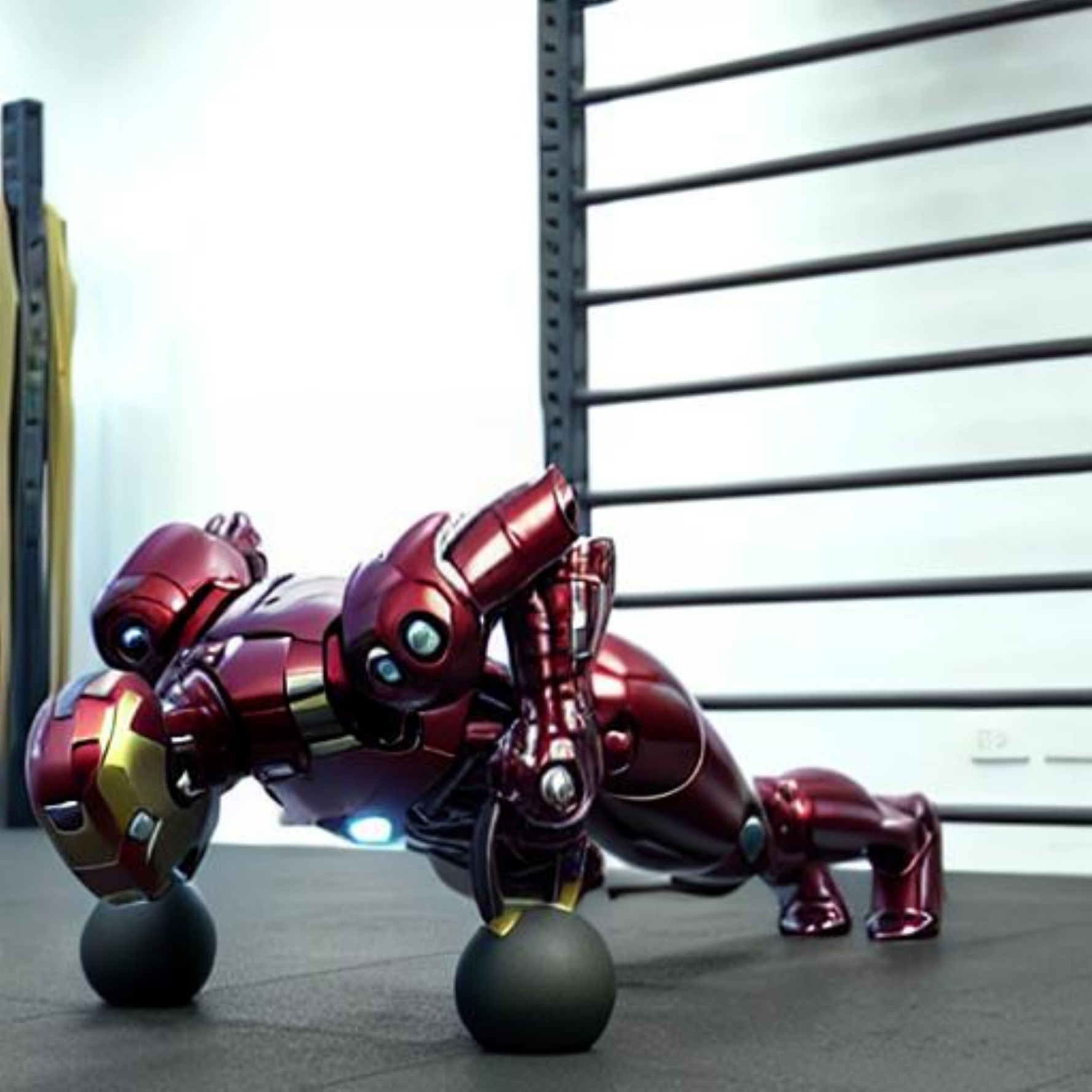}
\includegraphics[width=0.11\textwidth]{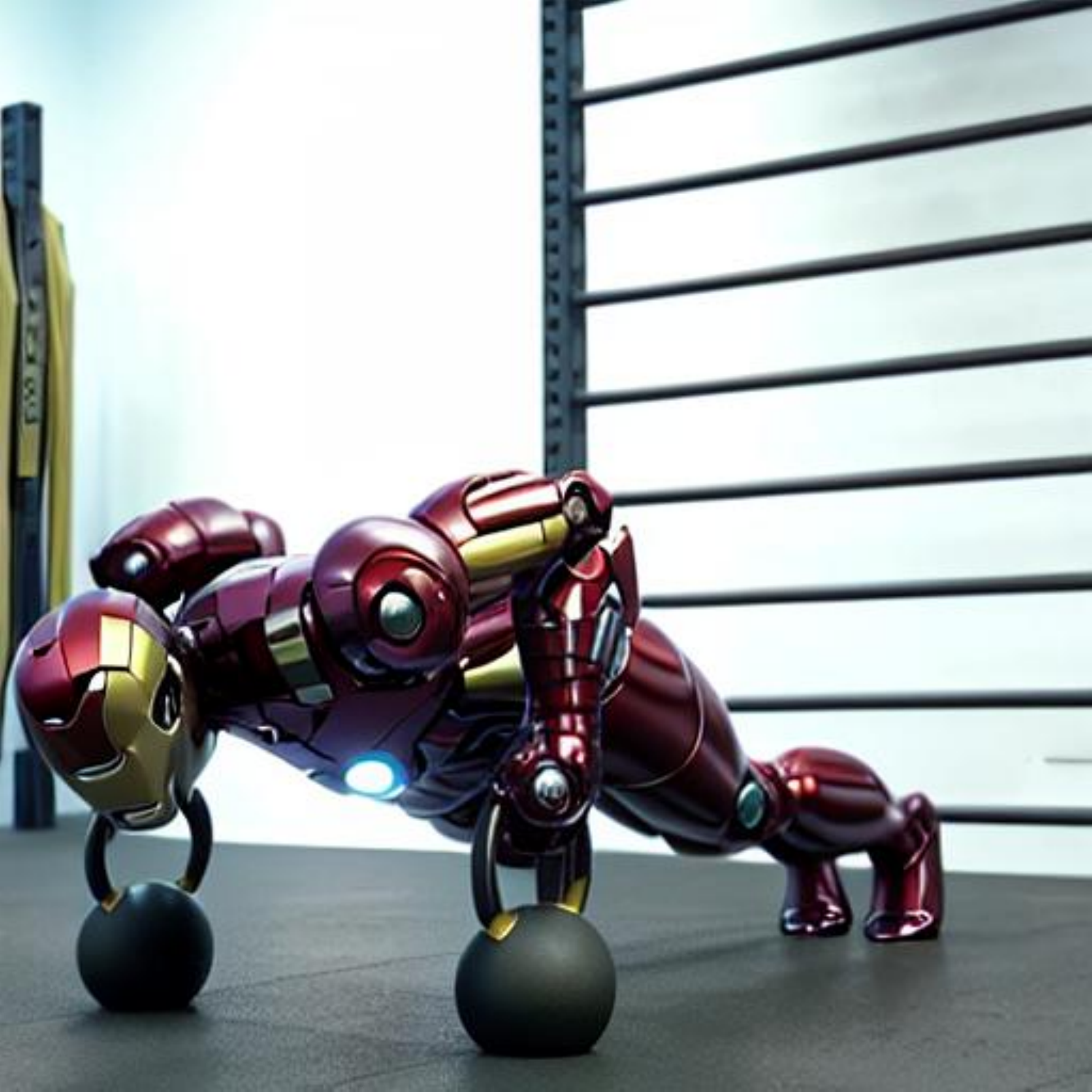}

\makebox[0.12\textwidth]{A \textcolor{blue}{\textbf{digital illustration}} that a man is doing a pushup.}\\
\includegraphics[width=0.11\textwidth]{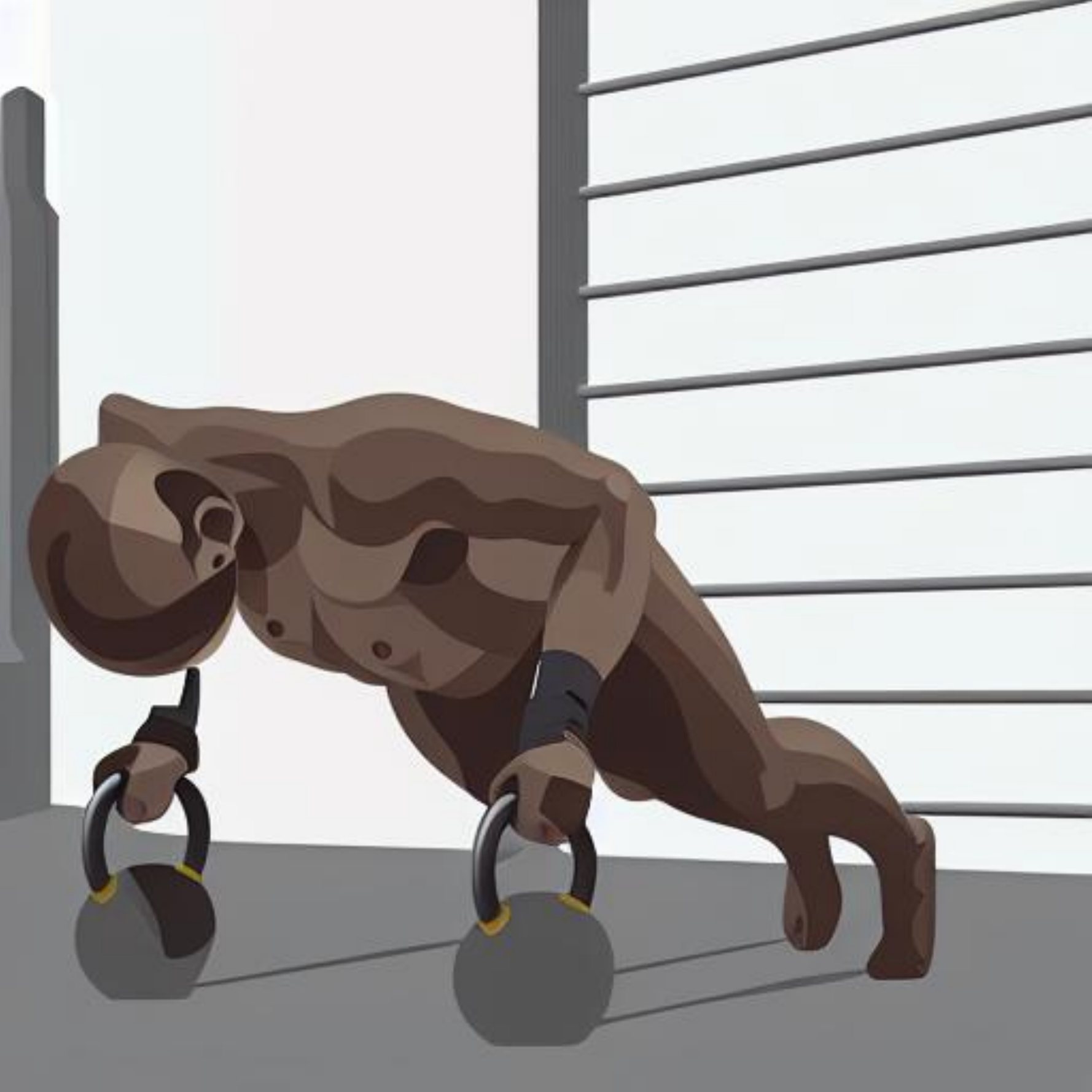}
\includegraphics[width=0.11\textwidth]{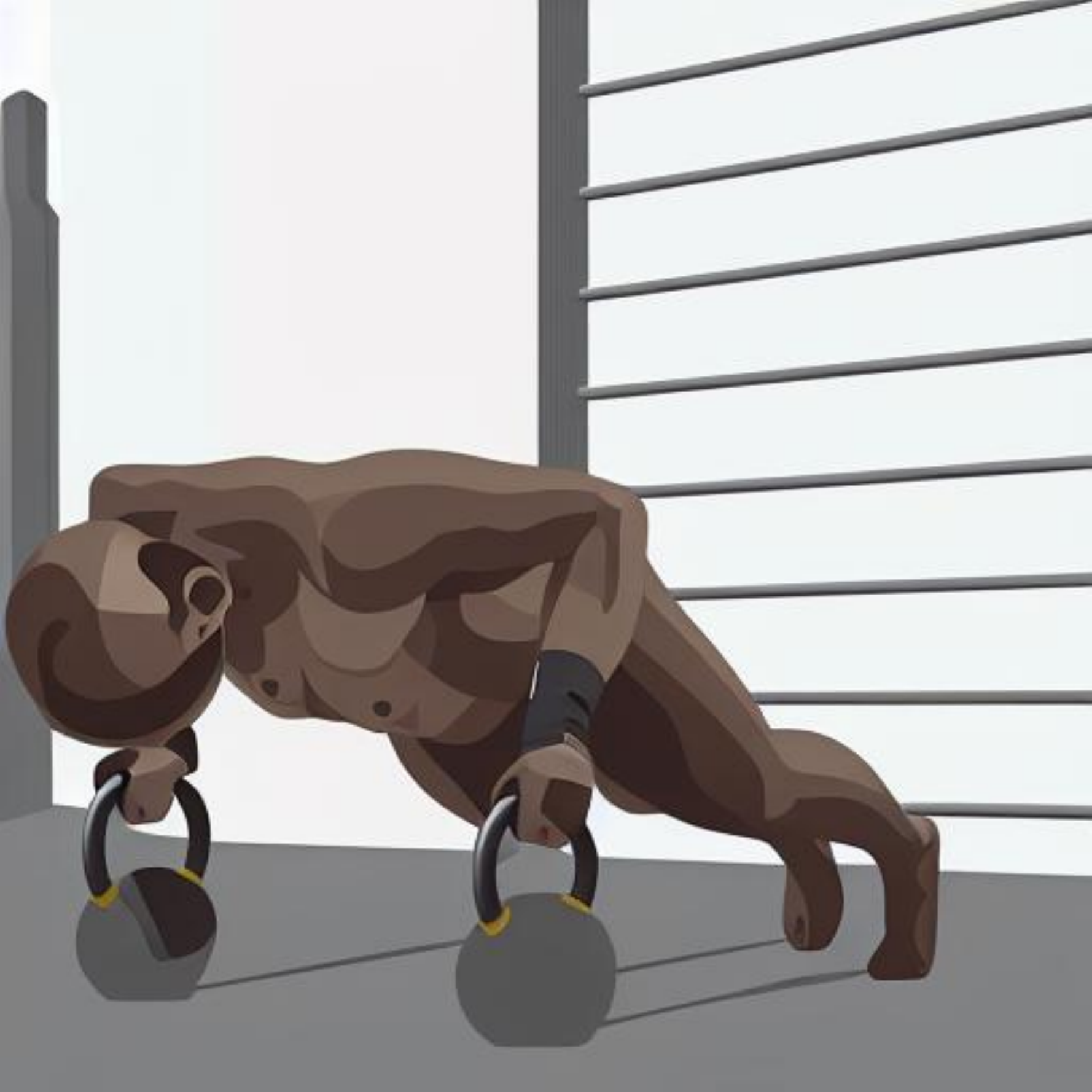}
\includegraphics[width=0.11\textwidth]{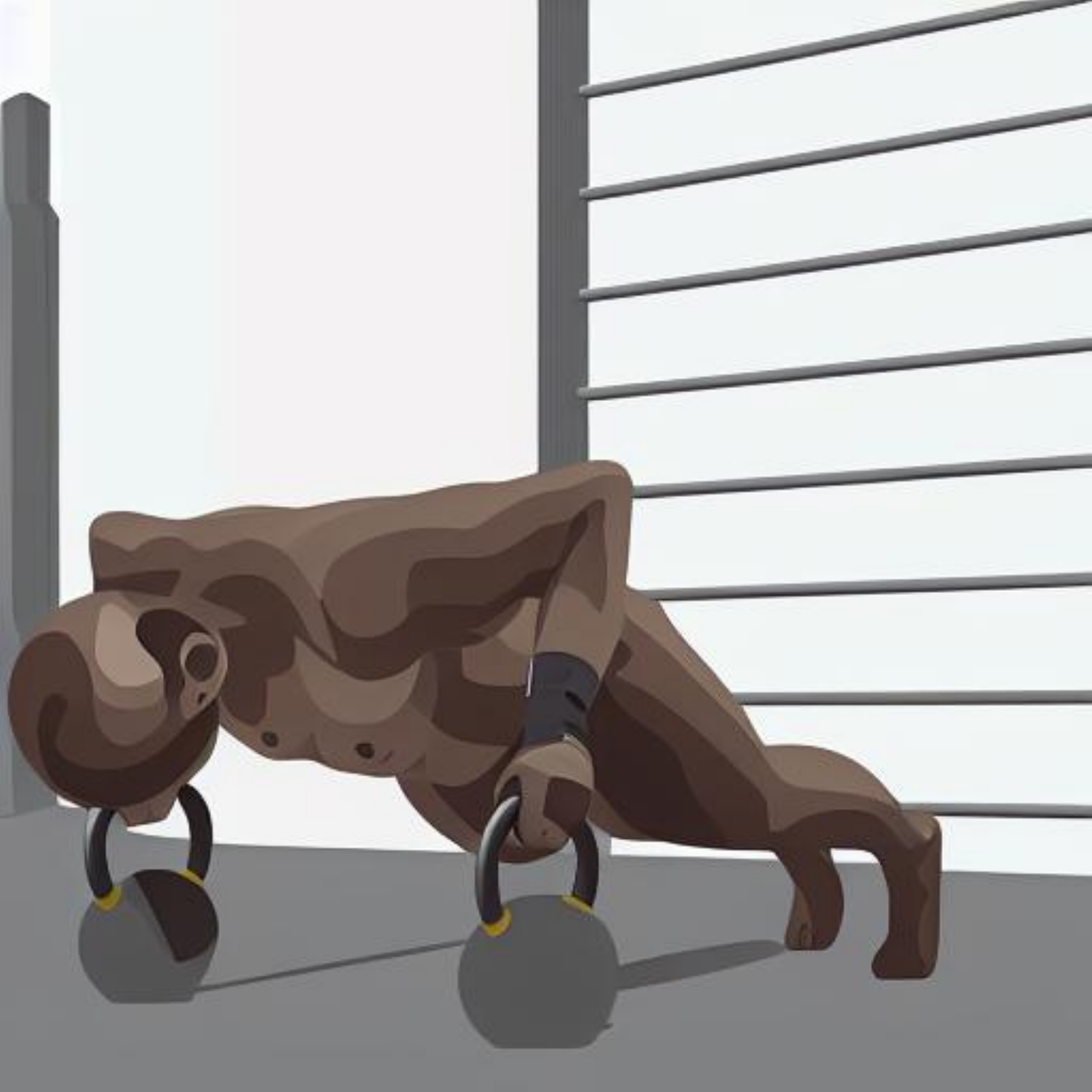}
\includegraphics[width=0.11\textwidth]{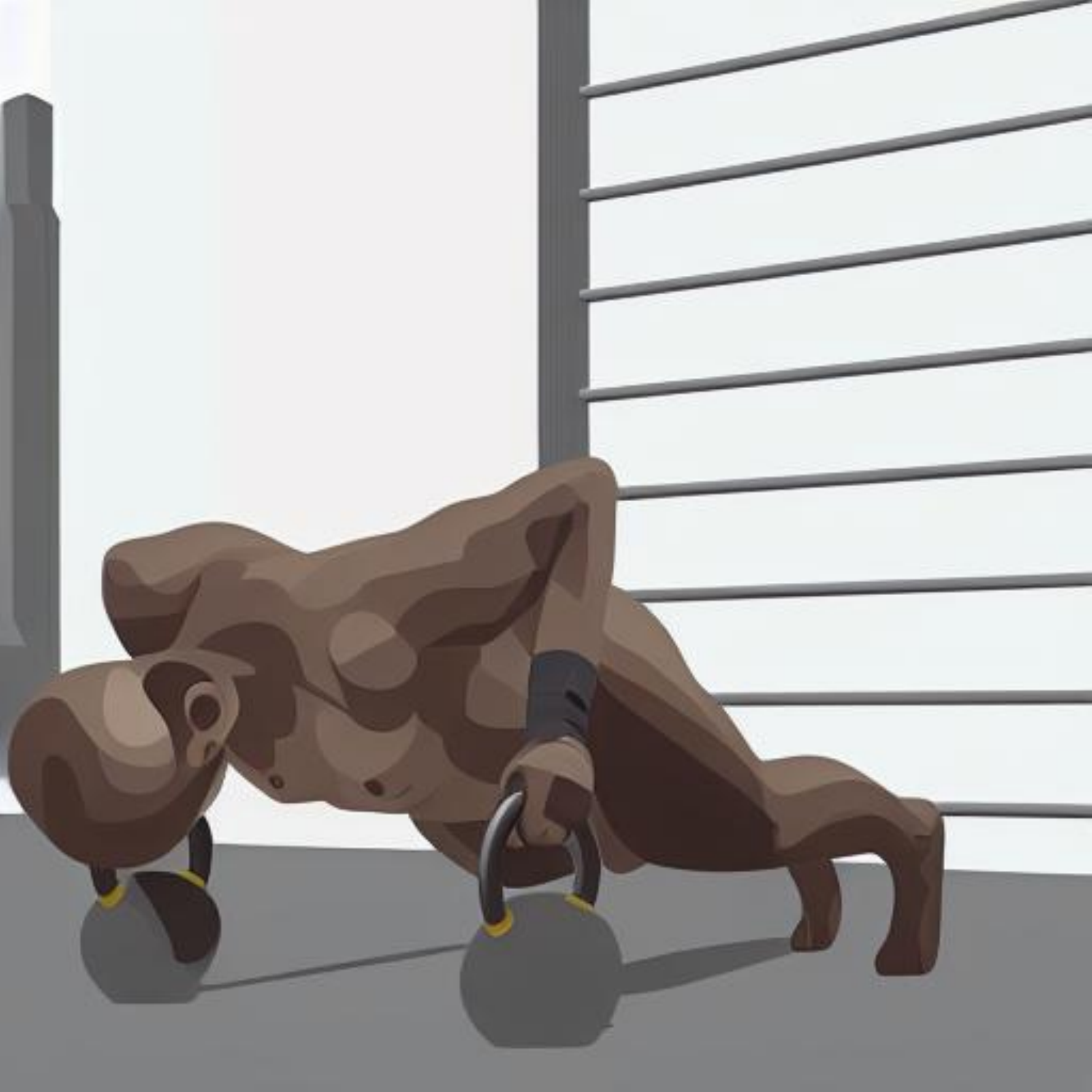}
\includegraphics[width=0.11\textwidth]{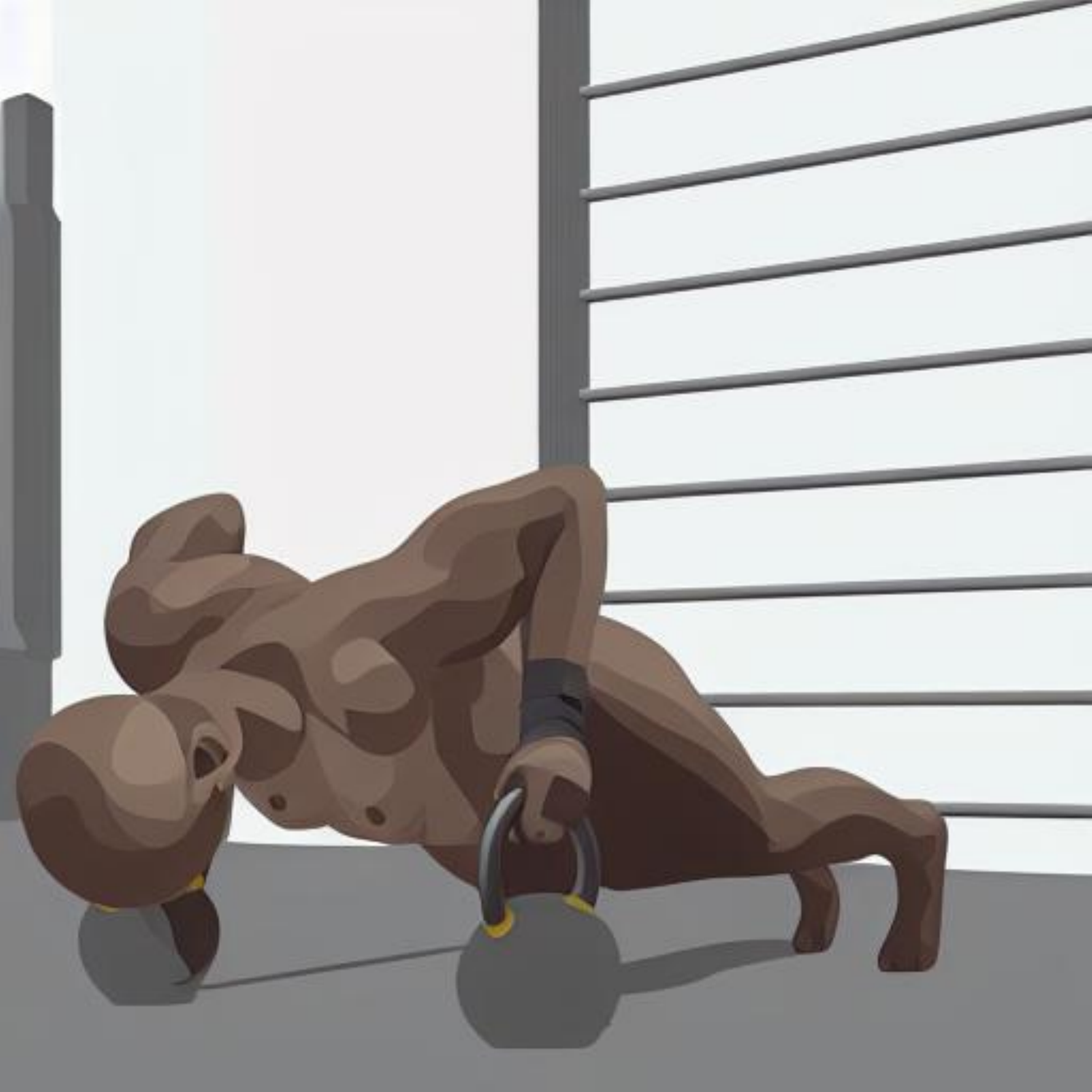}
\includegraphics[width=0.11\textwidth]{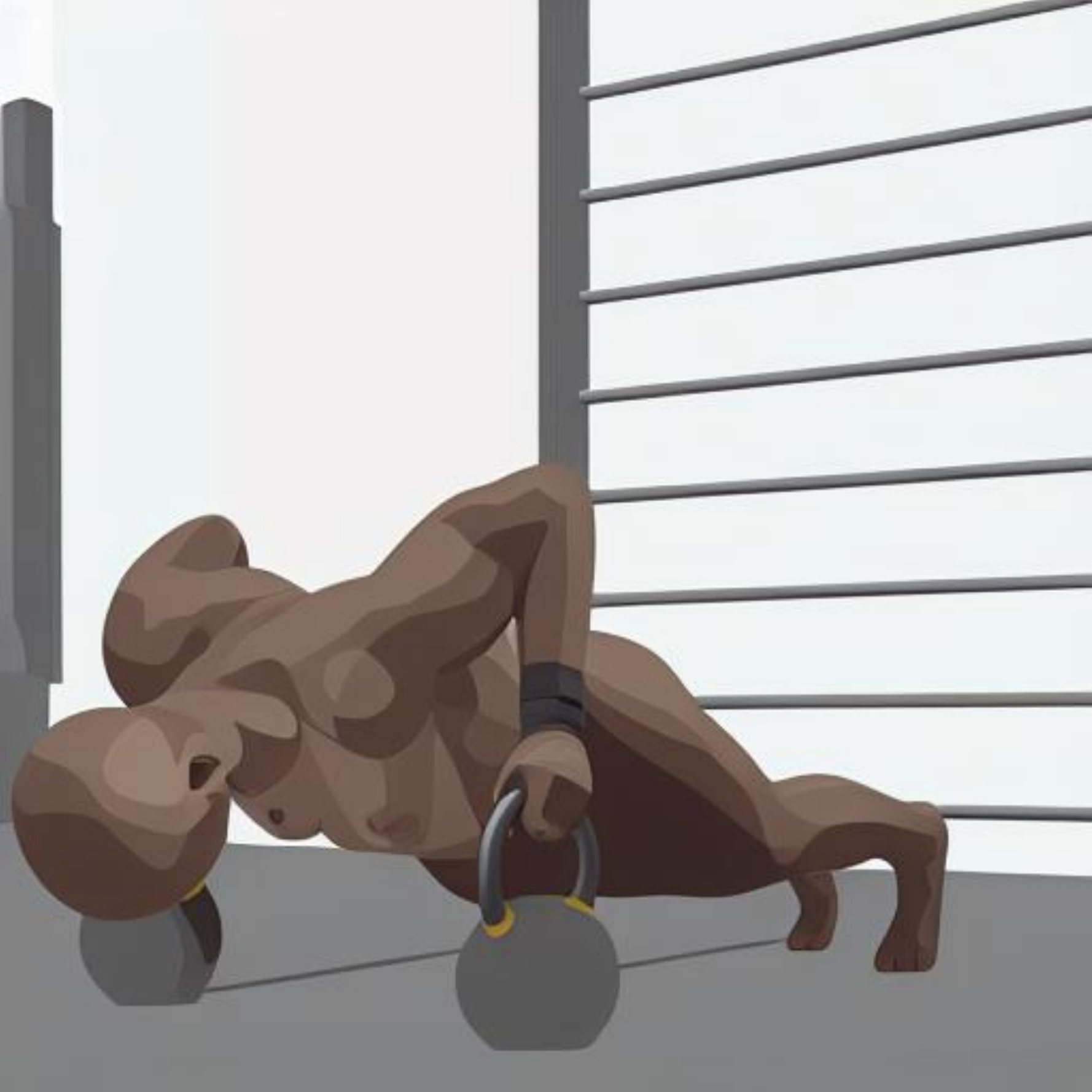}
\includegraphics[width=0.11\textwidth]{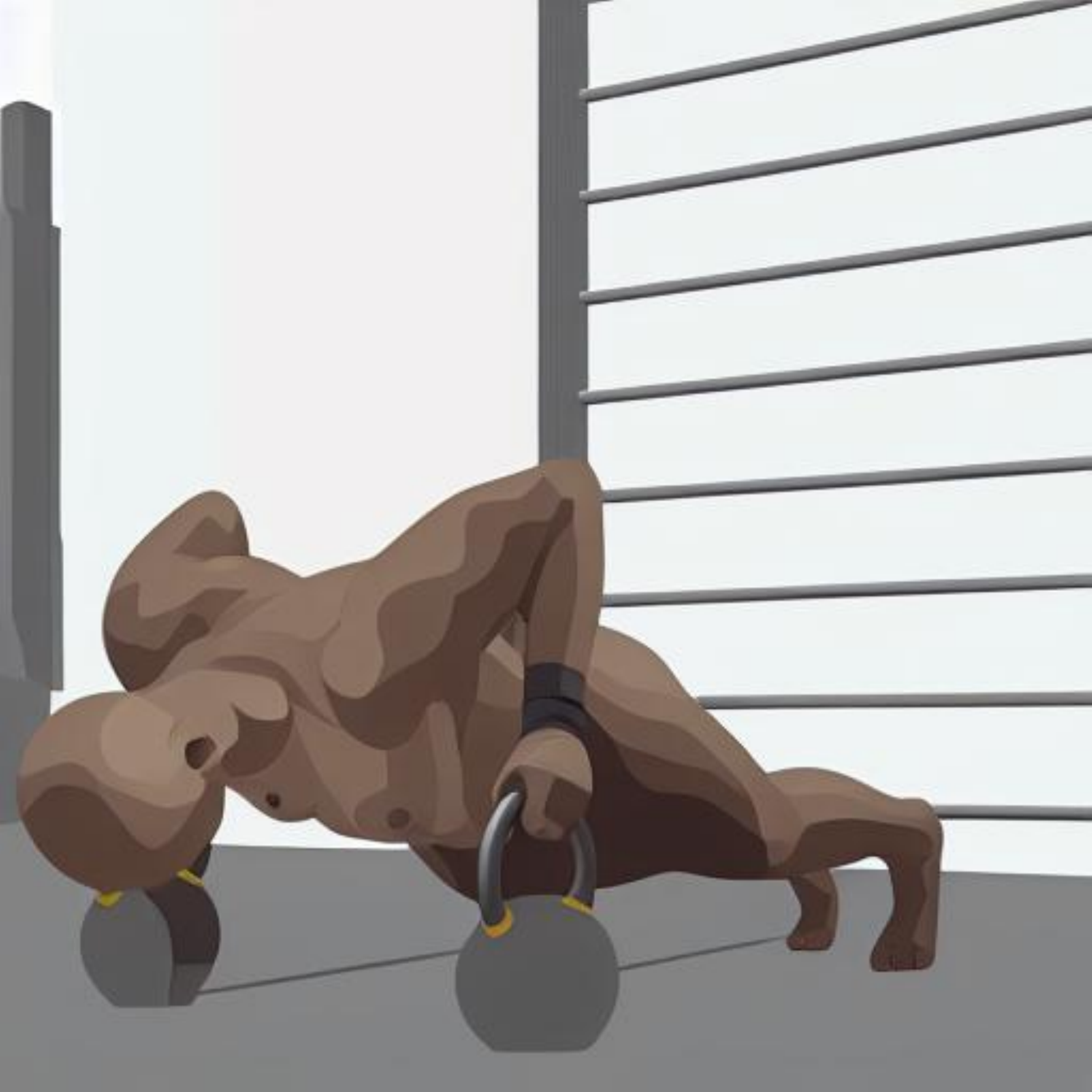}
\includegraphics[width=0.11\textwidth]{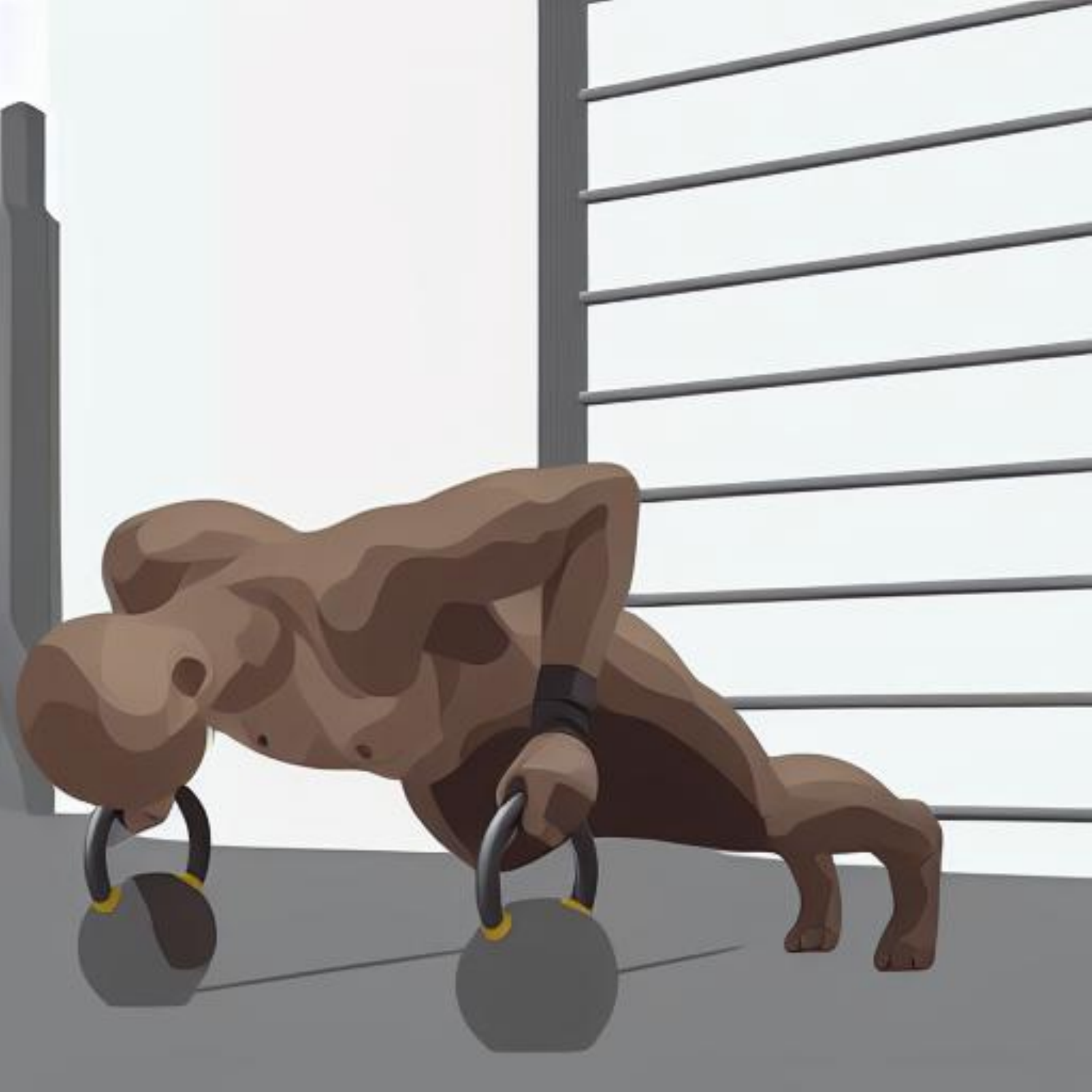}

\makebox[0.12\textwidth]{A \textcolor{blue}{\textbf{gorilla}} is doing a pushup, \textcolor{blue}{\textbf{cartoon style}}.}\\
\includegraphics[width=0.11\textwidth]{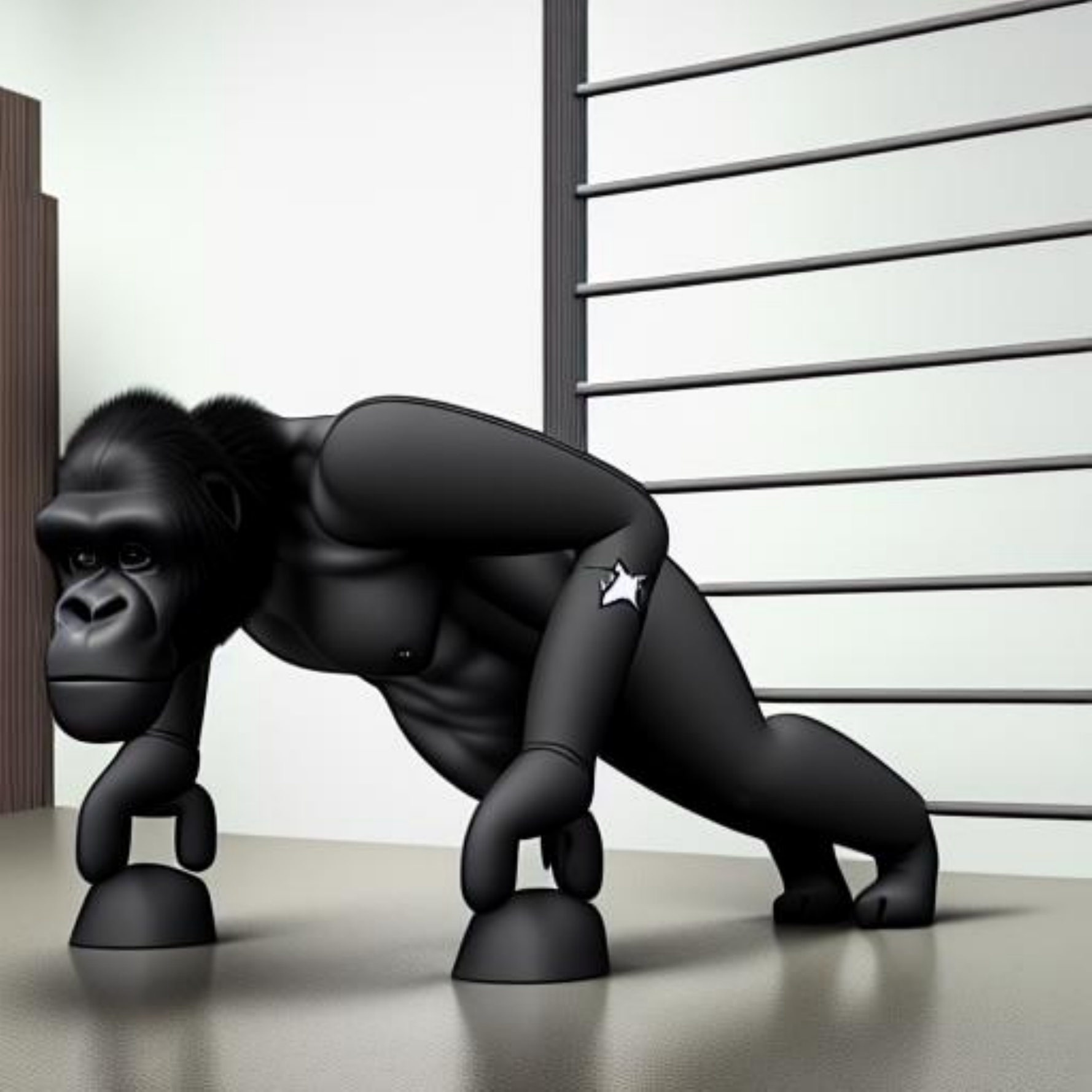}
\includegraphics[width=0.11\textwidth]{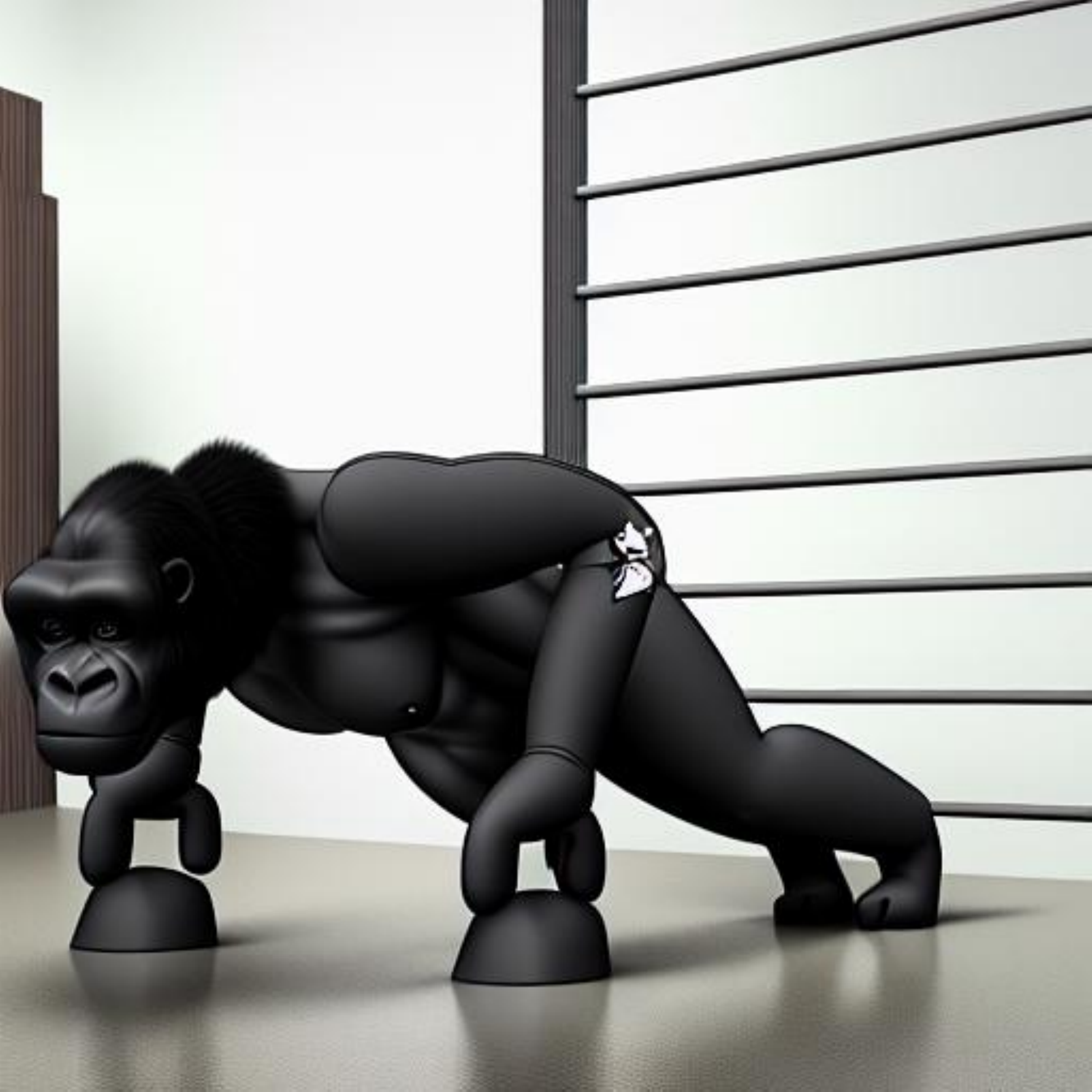}
\includegraphics[width=0.11\textwidth]{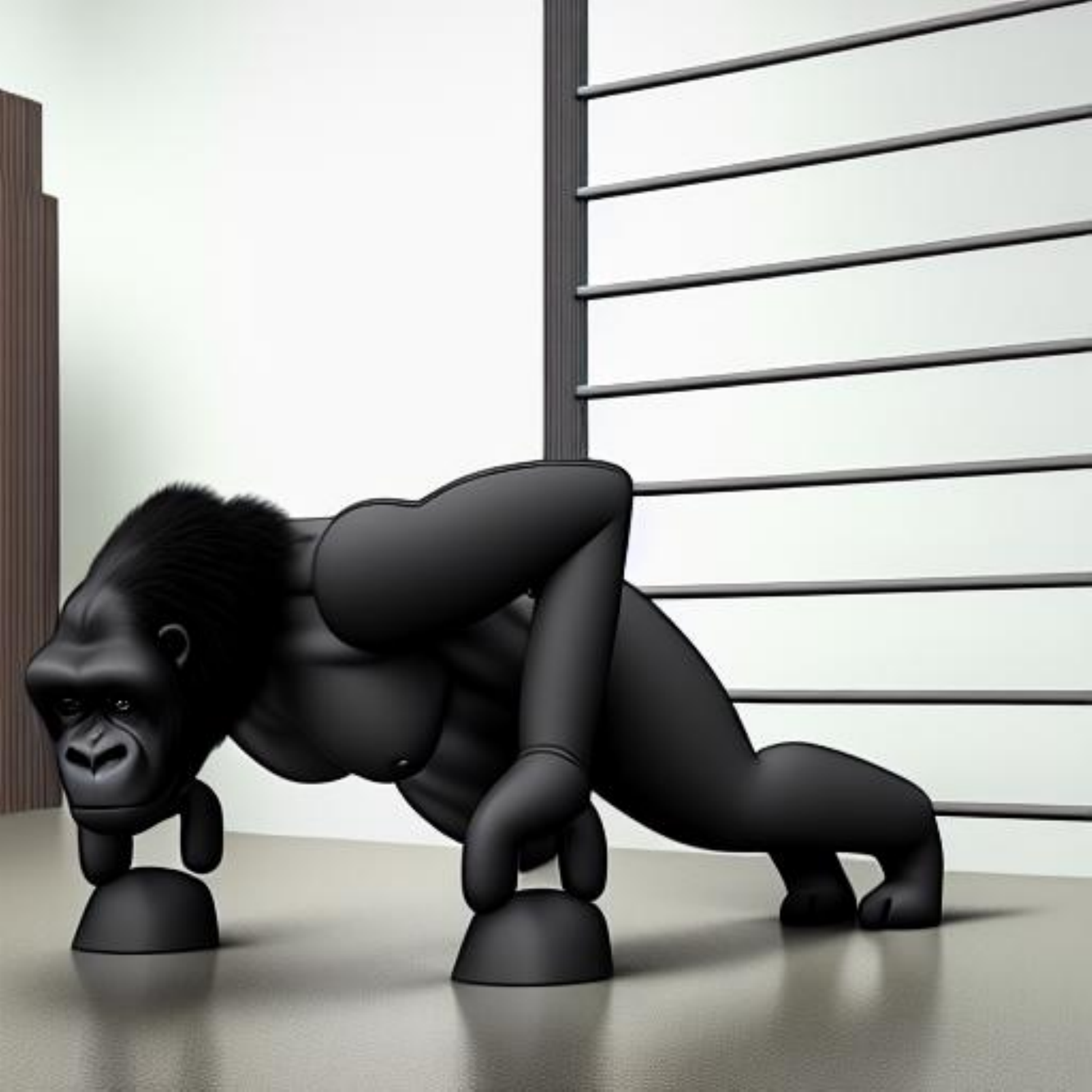}
\includegraphics[width=0.11\textwidth]{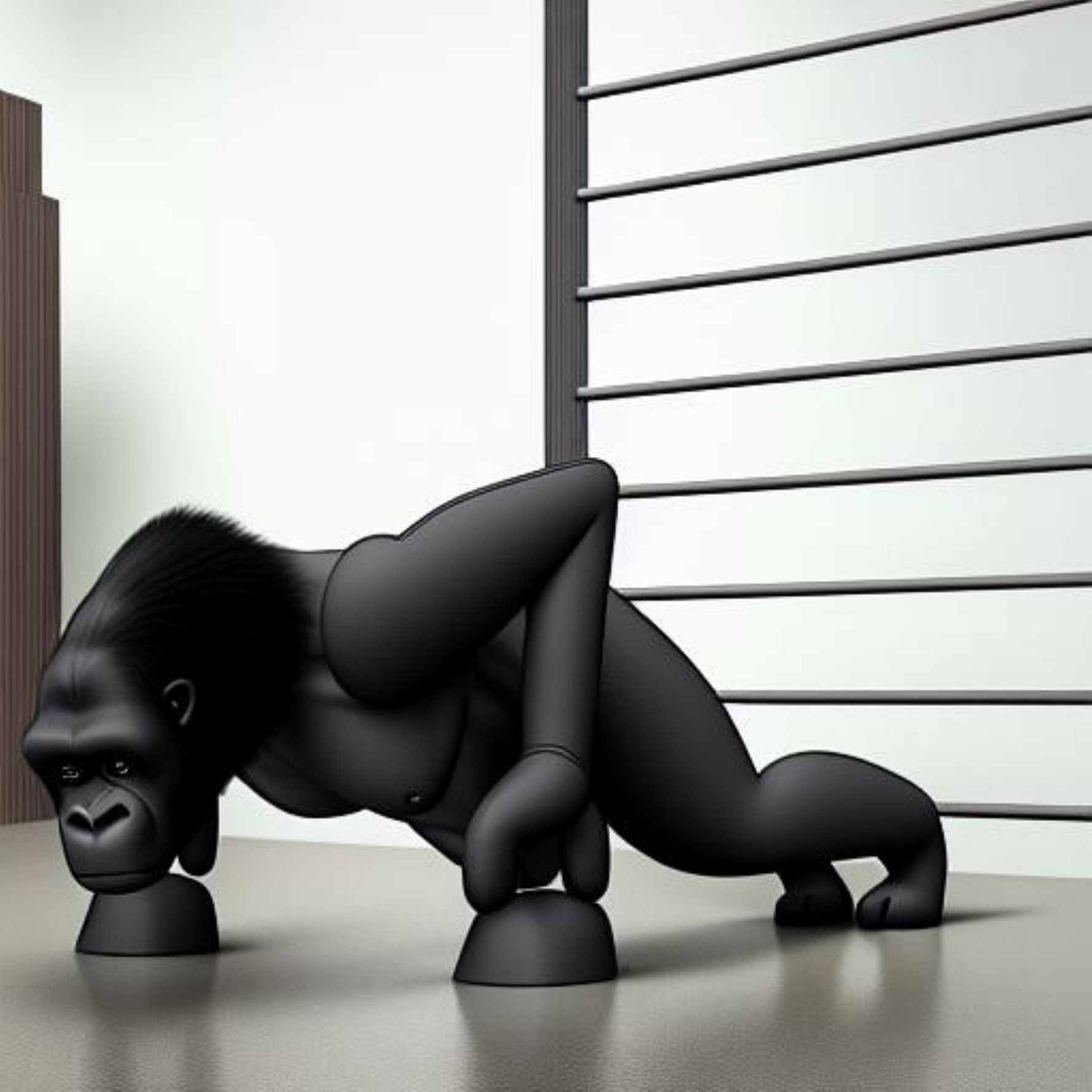}
\includegraphics[width=0.11\textwidth]{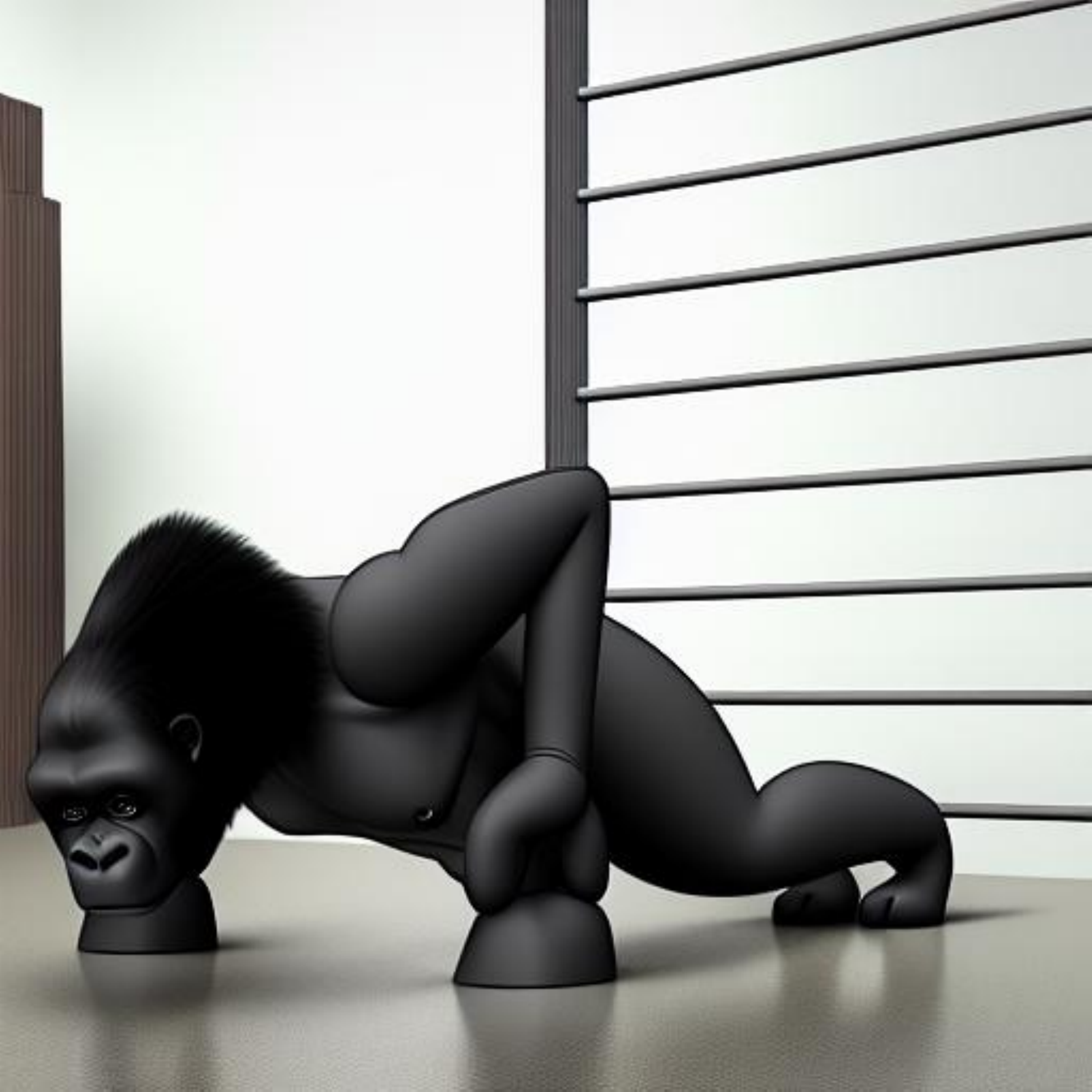}
\includegraphics[width=0.11\textwidth]{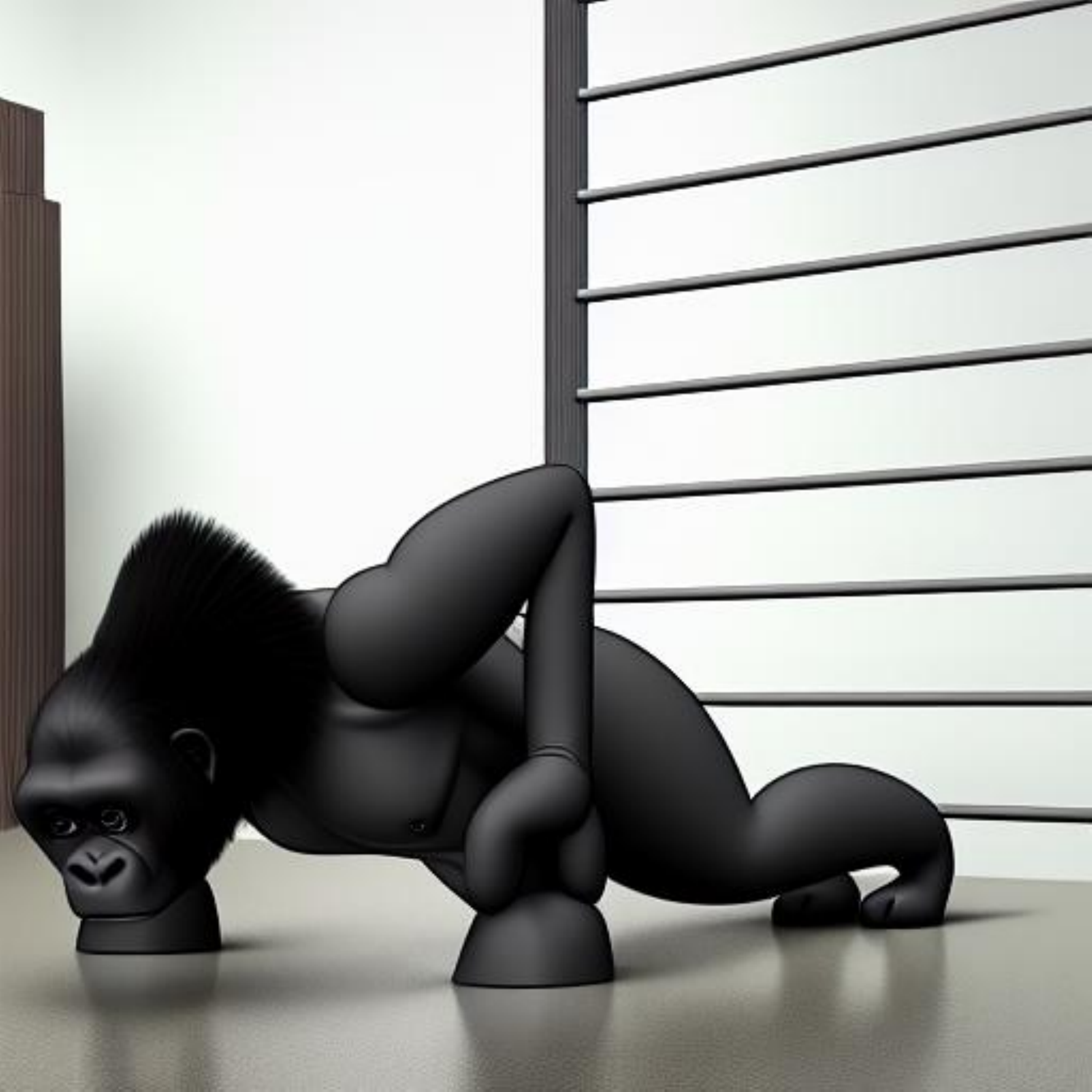}
\includegraphics[width=0.11\textwidth]{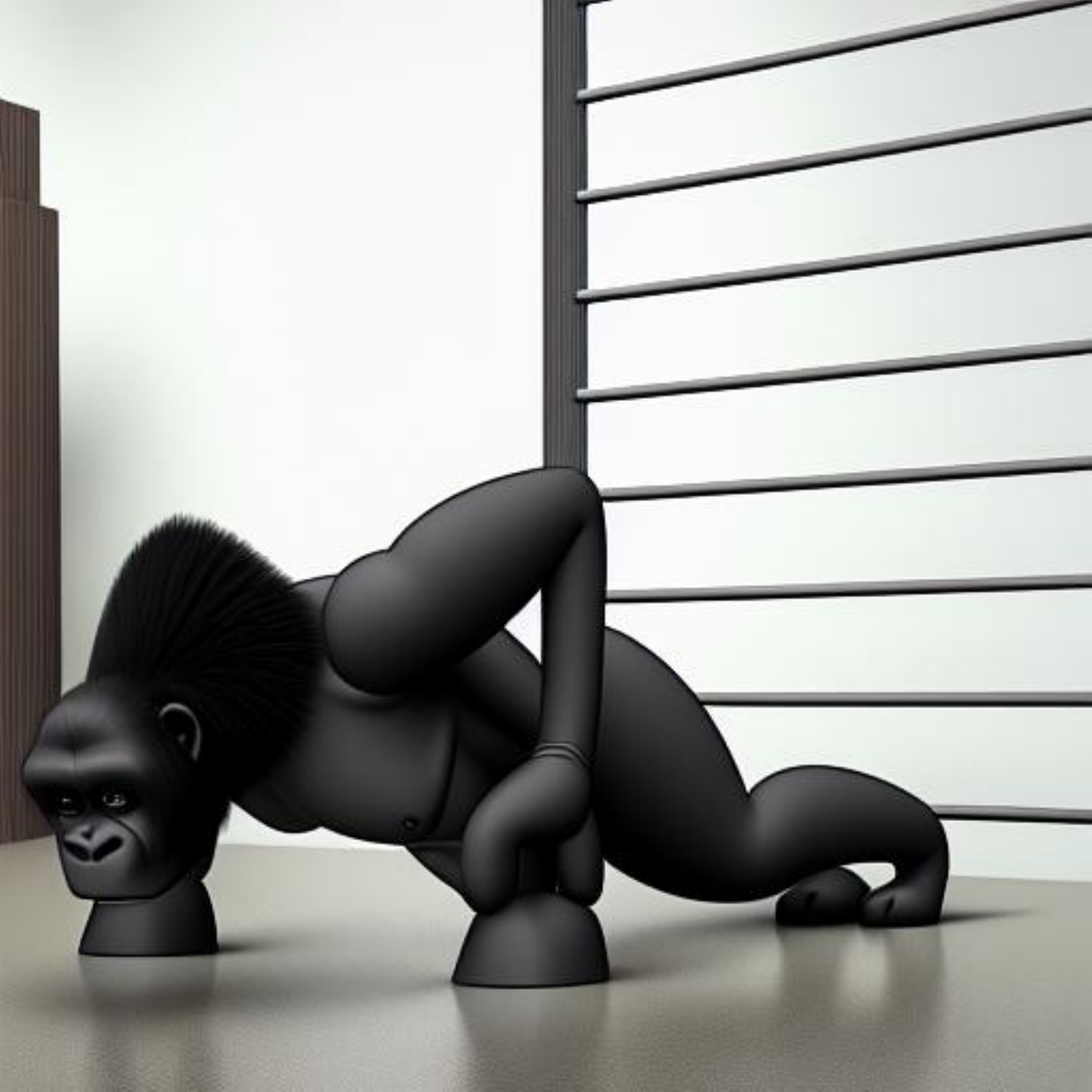}
\includegraphics[width=0.11\textwidth]{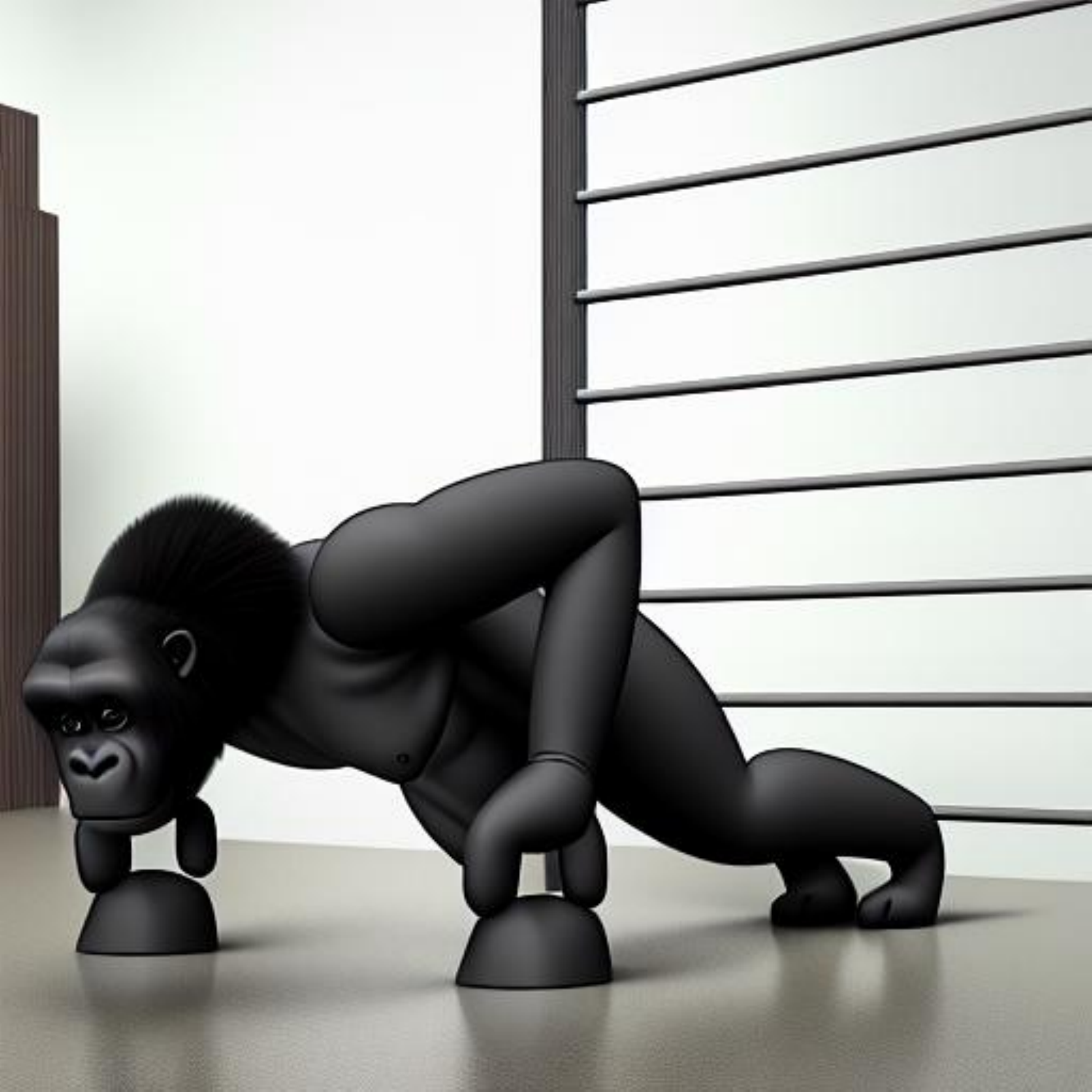}
\caption{Edit-A-Video performs text-guided video editing from a single $<$text, video$>$ pair and a text-to-image model.}
\label{fig:teaser}
\vspace{-5mm}
\end{center}
\end{figure}

Inspired by the success of the diffusion-based TTI models, various works have extended their output modality to videos, including text-to-video (TTV) generation and text-guided video editing.
Among them, text-guided video editing models perform editing by leveraging the prior knowledge of the TTV model \citep{molad2023dreamix, esser2023structure} or the inflated TTI model that captures temporal information \citep{wu2022tune}. 
Several works \citep{wu2022tune, ceylan2023pix2video} generate the whole video while editing, resulting in the editing of unwanted areas. 
Other methods \citep{liu2023videop2p, qi2023fatezero} have proposed using image editing techniques to specify areas to be edited within each frame. 
However, these approaches have limitations in that they do not consider temporal information when calculating the editing regions for each frame.

In this paper, we propose Edit-A-Video, a framework designed for achieving temporal consistency in text-guided video editing.
Edit-A-Video is a two-stage framework (see Fig. \ref{fig1} for illustration). 
In the first stage, a pretrained 2D TTI model is inflated to a 3D TTV model with attention modules for spatio-temporal modeling, which is finetuned using a single source video.
In the second stage, the source video is edited to match the target text description by inverting the source video to the Gaussian noise and injecting attention maps from the source to the target along the denoising process.

As previously mentioned, one of the main challenges in diffusion-based video editing is the accurate handling of the target object and the background. 
We observe that the edited video suffers from a \textit{background inconsistency problem}, where the edited video contains abrupt temporal changes in the background, which significantly degrades the quality. 
To tackle this issue, we propose a novel blending method tailored to the task, called \textit{temporal-consistent blending (TC Blending)}. TC Blending extends a spatial local blending technique originally proposed for a still image \citep{hertz2022prompt} and automatically generates a sharp, spatio-temporally consistent blending mask that closely approximates the region to be edited in the source video.

Together with the proposed TC Blending method, Edit-A-Video can effectively generate a realistic video that matches the target text description and captures the dynamic actions of the source video ensuring a smooth transition, while also maintaining the spatio-temporal consistency of the background.
We carry out extensive experiments on one-shot editing over various videos and prompts, and demonstrate that Edit-A-Video is the most favorable method compared to baselines through subjective human preference study along with qualitative analysis.
We further conduct an in-depth analysis by comparing to baselines based on automatic evaluation of numerical metrics, which shows the improvement of consistency and text alignment.
We also demonstrate the effectiveness of the proposed TC Blending in terms of consistency and overall editing quality through ablation studies.

In summary, our key contributions are as follows:
\begin{itemize}
    \item This study presents Edit-A-Video, a one-shot video editing framework that effectively combines a pretrained text-to-image model and editing techniques suitable for video editing.
    \item Along with the extension of image editing techniques to video, we propose a novel blending method called temporal-consistent blending (TC Blending), alleviating the background inconsistency problem. 
    \item We analyze the effect of injecting each attention map for spatio-temporal consistency in our framework.
\end{itemize}

\section{Background}
\subsection{Denoising Diffusion Probabilistic Models (DDPM)}
Denoising Diffusion Probabilistic Models (DDPM) \citep{thermodynamics, ddpm} is a type of probabilistic generative model that consists of a forward process that gradually transforms data into $z\sim N(0,I)$ and a reverse process, the opposite trajectory. Following notation in \cite{ddpm}, the forward process of diffusion is defined as follows:
\begin{equation}
  \begin{aligned}
    \label{forward diffusion}
    q(x_t|x_{t-1})&:=N(x_t;\sqrt{1-\beta_t}x_{t-1}, \beta_tI),\\
    x_t&=\sqrt{\bar{\alpha}_t}x_0 + \sqrt{1 - \bar{\alpha}_t}\epsilon_t,
  \end{aligned}
\end{equation}
where $z_t\sim N(0,I)$, $\beta_t$ is a user-defined noise schedule, and $\bar{\alpha}_t:=\prod_{i=1}^{t}(1-\beta_i)$. Using Eq. \ref{forward diffusion}, we can obtain the posterior $q(x_{t-1}|x_t,x_0)=N(x_{t-1};\frac{1}{\sqrt{\bar{\alpha}_t}}(x_t-\frac{\beta_t}{\sqrt{1-\bar{\alpha}_t}}\epsilon_t),\frac{1-\bar{\alpha}_{t-1}}{1-\bar{\alpha}_{t}}\beta_t)$.

For sampling data from noise, we define a reverse process that gradually transforms noise into data for sampling. Since $q(x_{t-1}|x_t)$ is intractable, \cite{ddpm} approximate this term to the transition network $p_\theta(x_{t-1}|x_t)$, which is defined in a Gaussian form as below.
\begin{equation}
    \begin{aligned}
    \label{reverse diffusion}
    p_\theta(x_{t-1}|x_t)&:=N(x_{t-1};\mu_\theta(x_t,t), \Sigma_\theta(x_t,t)), \\   
    \mu_\theta(x_t,t)&=\frac{1}{\sqrt{\bar{\alpha}_t}}(x_t-\frac{\beta_t}{\sqrt{1-\bar{\alpha}_t}}\epsilon_\theta(x_t,t)),
    \end{aligned}
\end{equation}
where $\Sigma_\theta(x_t,t)$ can be a fixed value \citep{ddpm} or trainable parameters \citep{pmlr-v139-nichol21a}, and $\epsilon_\theta(x_t,t)$ can be trained by minimizing $\mathop{\mathbb{E}}_{t,x_0,\epsilon}[||\epsilon-\epsilon_\theta(x_t,t)||^2]$ \citep{ddpm}. On the contrary, DDIM \citep{ddim} proposes to set $\Sigma_\theta(x_t,t)$ to 0 for a subset of the reverse diffusion process, which enables a deterministic and fast sampling. In this paper, we follow the formulation of DDIM \citep{ddim}.

\subsection{Tune-A-Video}
Compared to text-to-image generation, text-to-video generation~\citep{singer2022make,ho2022imagen,esser2023structure} is a data-hungry task since the training of TTV model requires a large-scale $<$text, video$>$ paired dataset. 
Recently, Tune-A-Video \citep{wu2022tune} successfully trained a TTV model with a single video leveraging prior knowledge of a pretrained TTI model, while maintaining consistency between frames. To generate a temporally-coherent video, Tune-A-Video inflated 3x3 2D convolution layers from TTI models to 1x3x3 3D convolution layers and appended additional temporal attention (T-Attn) modules.
To enhance the temporal consistency more, Tune-A-Video replaces spatial self-attention with a novel attention layer named as sparse spatio-temporal attention (ST-Attn).
Instead of full frame attention which computes the attention for every other frame, ST-Attn computes the attention of current frames only on the first and the previous frame.
It reduces the computational cost of full frame attention which is quadratic with respect to the number of frames to linear cost, while maintaining the temporal consistency in video generation.
By tuning three types of attention layers (T-Attn, ST-Attn, and cross-attention between text and video) from the inflated model, Tune-A-Video is capable of generating temporally-coherent videos.

Despite its ability to synthesize video with temporal consistency, Tune-A-Video fails to maintain the contents that should remain unedited since it generates entire video frames using a modified text prompt.
For example, the background regions of the video generated by Tune-A-Video are modified and are not preserved with respect to the source video.
In contrast, Edit-A-Video edits a video with given text prompts while maintaining the attributes unrelated to the modified text based on the novel blending method.

\subsection{Real Image Editing with Null-Text Inversion} \label{null-text}

Text-guided image editing is a task to edit the content of a given image only specified by the text prompts.
While TTI models can produce high-quality images from the given text prompts, the model tends to generate completely different images as the source text is modified to serve as the editing target.
\cite{hertz2022prompt} observe that the cross-attention maps between the image feature and the source text reflect the spatial layout of the source image. 
Based on the remarkable property of cross-attention maps, \cite{hertz2022prompt} propose a novel framework, Prompt-to-Prompt (PTP), for image editing which injects the attention maps of the source image into the generation process conditioned on the target text.
PTP further enhances the degree of preservation of the source image by approximating the desired editing target region from the cross-attention maps and blending the target region of the edited image with the source image.

Although PTP is a strong framework for synthetic image editing, a latent vector corresponding to the real image is required for real image editing. 
DDIM inversion is one of the methods to invert the real image to a latent vector.
However, DDIM trajectory is severely distorted when a classifier-free guidance~\citep{ho2021classifierfree} is applied through the reverse process, which results in the poor reconstruction quality of the real image. 
As an effective approach for the limitation, Null-Text Inversion (NTI) \citep{mokady2022null} optimizes the null-text embedding which is used in classifier-free guidance so that the reverse process trajectory with classifier-free guidance does not deviate from the DDIM inversion trajectory.
NTI results in the near-perfect reconstruction of the image even under classifier-free guidance. 
Together with the PTP framework, NTI shows that text-guided image editing can also be applied to real image.

Despite the success of text-guided image editing, image editing methods cannot be directly applied to video editing.
Compared to image editing, maintaining temporal consistency is crucial when it comes to video editing.
We hence append sparse spatio-temporal attention (ST-Attn) and temporal attention (T-Attn) on top of NTI and PTP framework and propose a novel blending method called temporal-consistent blending (TC Blending) for successful video editing.

\subsection{Text-Guided Video Editing}
Recently, text-based image creations expand their domain from the image to the video. Text-to-Video models synthesize high-quality videos corresponding to a given text.
Likewise, following the text-guided image editing methods, various approaches for text-guided video editing have also been proposed. 
Text-guided video editing modifies the given source video to reflect the target text while preserving some content from the source video.
Earlier works~\citep{esser2023structure, molad2023dreamix} edit the video by exploiting the TTV model trained on a large-scale $<$text, video$>$ paired dataset, which is computationally challenging to most practitioners.
Other approaches~\citep{bar2022text2live, Lee_2023_CVPR} utilize the neural layered atlas (NLA) of the source video for editing. 
While they enable video editing by applying image editing methods to the NLA of the video, acquiring NLA is time-consuming and lacks efficiency.

In contrast to prior approaches, there are several works that leverage the TTI model to perform text-guided video editing similar to ours. Some previous works~\citep{wu2022tune, ceylan2023pix2video, vid2vid-zero} generate the entire video frames including the area not related to the target object during editing, which makes it difficult to maintain the content of the source video. Similar to Edit-A-Video, some concurrent works~\citep{qi2023fatezero, liu2023videop2p} perform editing through attention map control. 
These methods improve editing performance by blending the source video and target region specified by the target text using a mask extracted from the cross-attention map. 
However, the cross-attention map is calculated between text and each frame and does not consider temporal information between frames.
Thus, the frame-wise blending mask extracted from cross-attention map can be temporally inconsistent, resulting in undesirable artifacts including abrupt changes in the background.
To mitigate the aforementioned issue, Edit-A-Video proposes TC blending, a blending method that is capable of extracting sharp and frame-consistent masks by considering temporal information from sparse spatio-temporal attention.
\section{Method}
\begin{figure}[t]
\begin{center}
    \includegraphics[width=\linewidth]{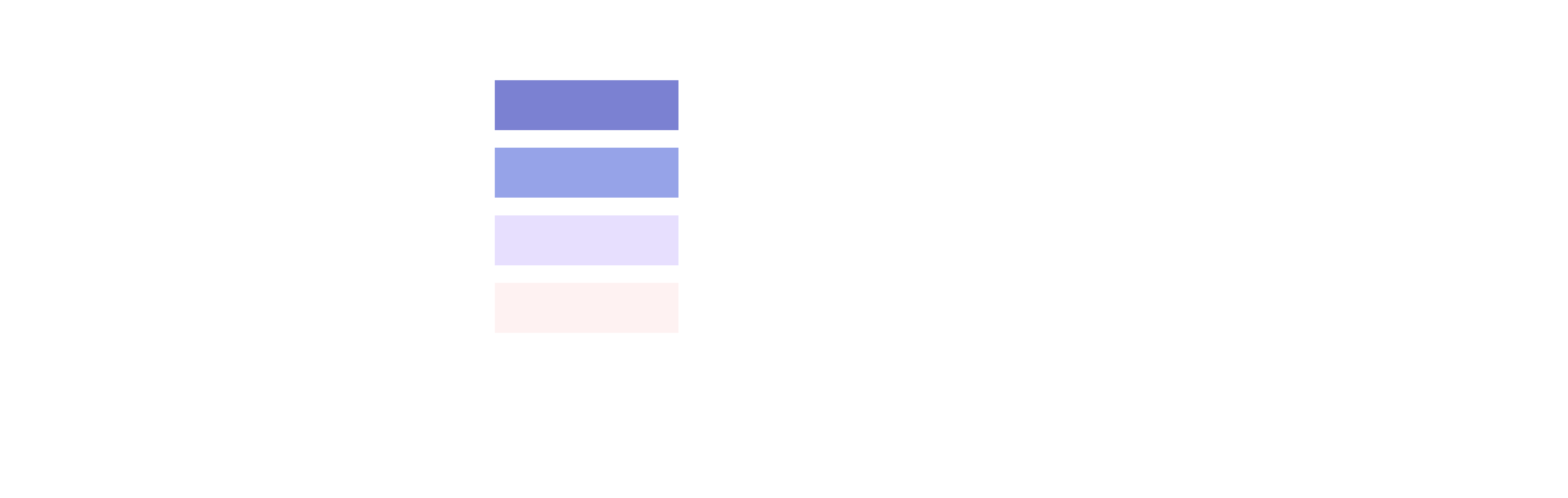}
\caption{\textbf{Overall Editing Procedure of Edit-A-Video} In stage 1, Edit-A-Video inflates the 2D UNet into 3D model by appending the temporal module and evolving the 2D conv into 3D conv and self-attention into sparse spatio-temporal attention. Then, the source video is inverted to a specific Gaussian noise by DDIM inversion, and null-text embedding is optimized so that the source video can be reconstructed along with the null-text guidance in stage 2(a). Finally, in stage 2(b), the edited video is generated from the inverted noise and target text through attention map injection, TC Blending, and optimized null-text embedding.} 
    \label{fig1}
    \vskip -0.1in
    \vspace{-5mm}
\end{center}
\end{figure}
In this section, we introduce Edit-A-Video, a framework designed to consistently edit a given video into a desired object or style using a diffusion-based TTI model.
In Sec. \ref{framework}, we formulate our 2-stage editing procedure, extending TTI model to learn the temporal relationship and editing the contents of the source video according to the target prompt \\textcolor{red}{via} attention map injection.
To achieve more consistent editing frame by frame, we propose a temporal-consistent blending (TC Blending) method in Sec. \ref{sc}. 
Finally, we discuss the role and effect of three types of attention modules in our methodology in Sec. \ref{hparams}.

\subsection{Framework} \label{framework}

As shown in Fig. \ref{fig1}, Edit-A-Video follows a two-step process to edit the given video corresponding to a target prompt. 
In the first stage, we inflate the TTI model to TTV model using the method of Tune-A-Video \citep{wu2022tune}.
Unlike the TTI model, which has two types of attention (self-attention in images and cross-attention between text and image), our inflated TTV model has three types of attention: cross-attention, temporal attention, and sparse spatio-temporal attention. 
We only train these attention modules in the inflated 3D TTV model using a single video, ensuring frame-by-frame consistency during video editing.

In the second stage, we extract the inversion trajectory $\{z_{t}\}_{t=1}^{T}$ starting from the source video to the Gaussian noise ${z_{T}}$ and train the null-text embeddings so that the generation trajectory from ${z_{T}}$ is still close to the inversion trajectory under the classifier-free guidance.
Starting from the latent variable ${z_{T}}$ of the source video, we edit the video by injecting three types of attention maps of the source video into the generation process of edited video, extending the editing methods in the image domain \citep{hertz2022prompt, mokady2022null}.
Since the inflated model has newly added attentions previously absent in the TTI model, we analyze the role and effect of each attention module and describe in Sec. \ref{hparams}.

However, when we edit the video as described above, we observe that unwanted region is edited inconsistently along the frames, which results in undesirable and abrupt artifacts. 
We call this issue the \textit{background inconsistency problem}.
To mitigate this issue, we analyze the cause and propose a \textit{temporal-consistent blending}, which we call \textit{TC Blending} for short, in the following section.

\begin{figure*}[t]
\begin{center}
\makebox[0.12\textwidth]{A man is on the surfing $\rightarrow$ A \textcolor{blue}{\textbf{wooden man sculpture}} is surfing}\\

\includegraphics[width=0.11\textwidth]{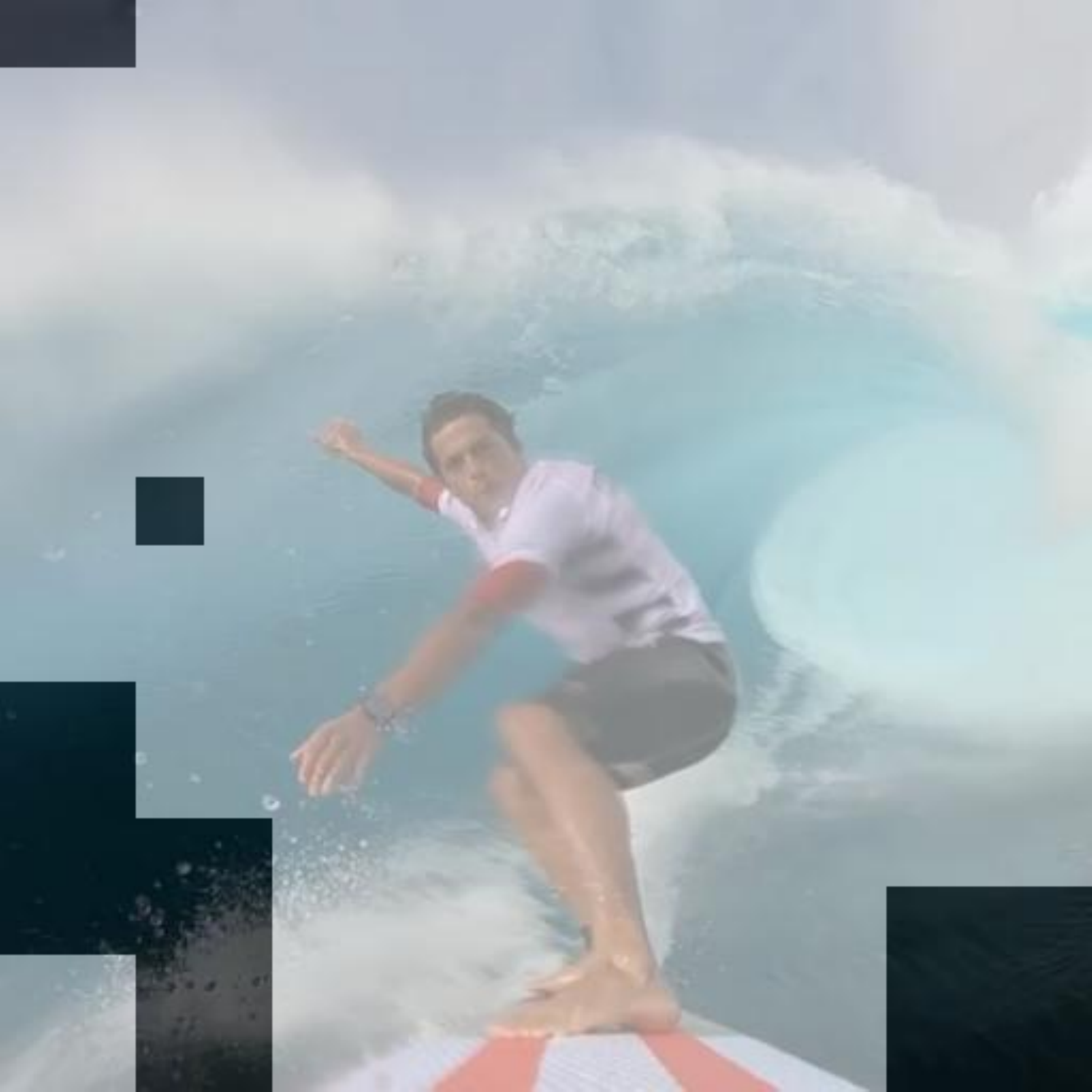}
\includegraphics[width=0.11\textwidth]{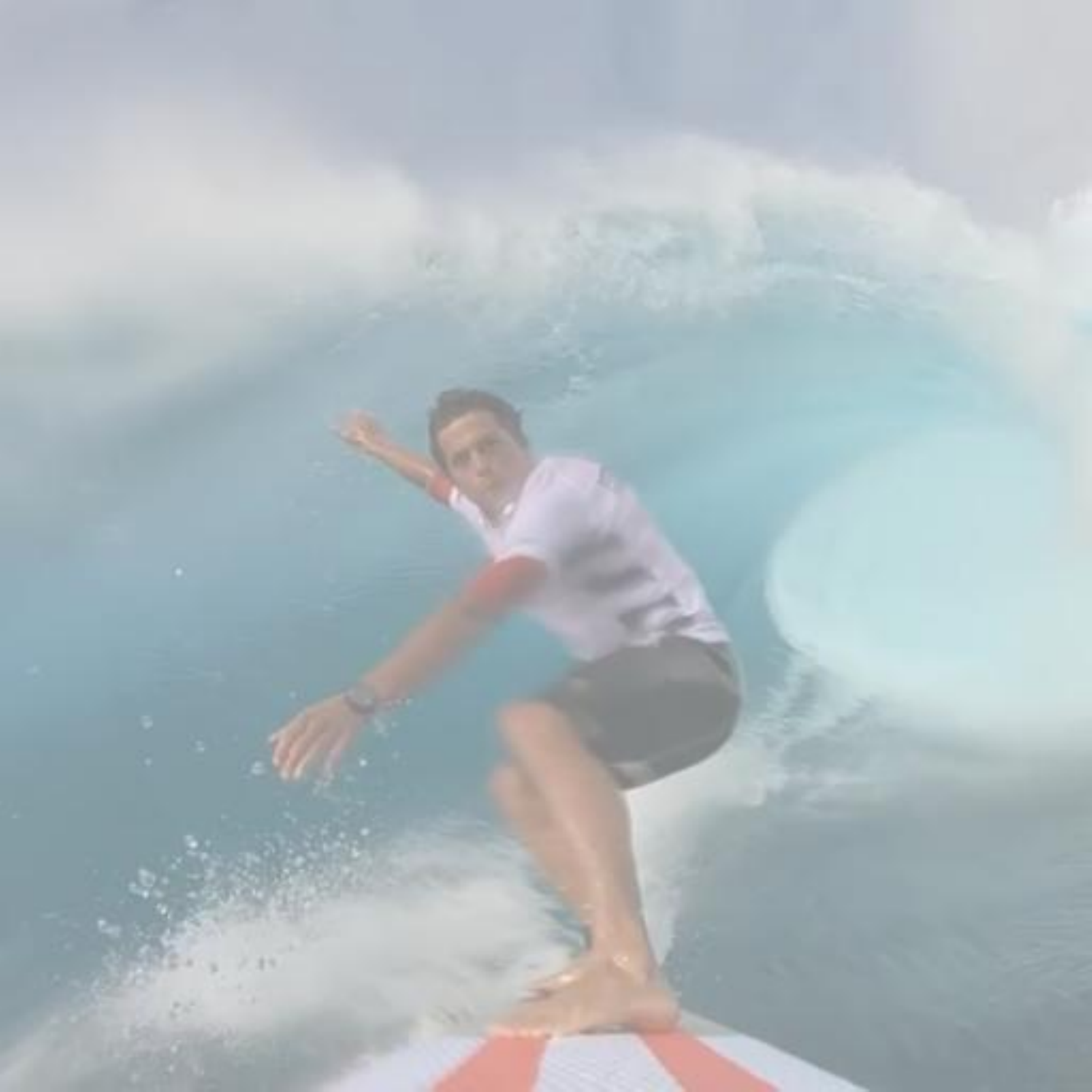}
\includegraphics[width=0.11\textwidth]{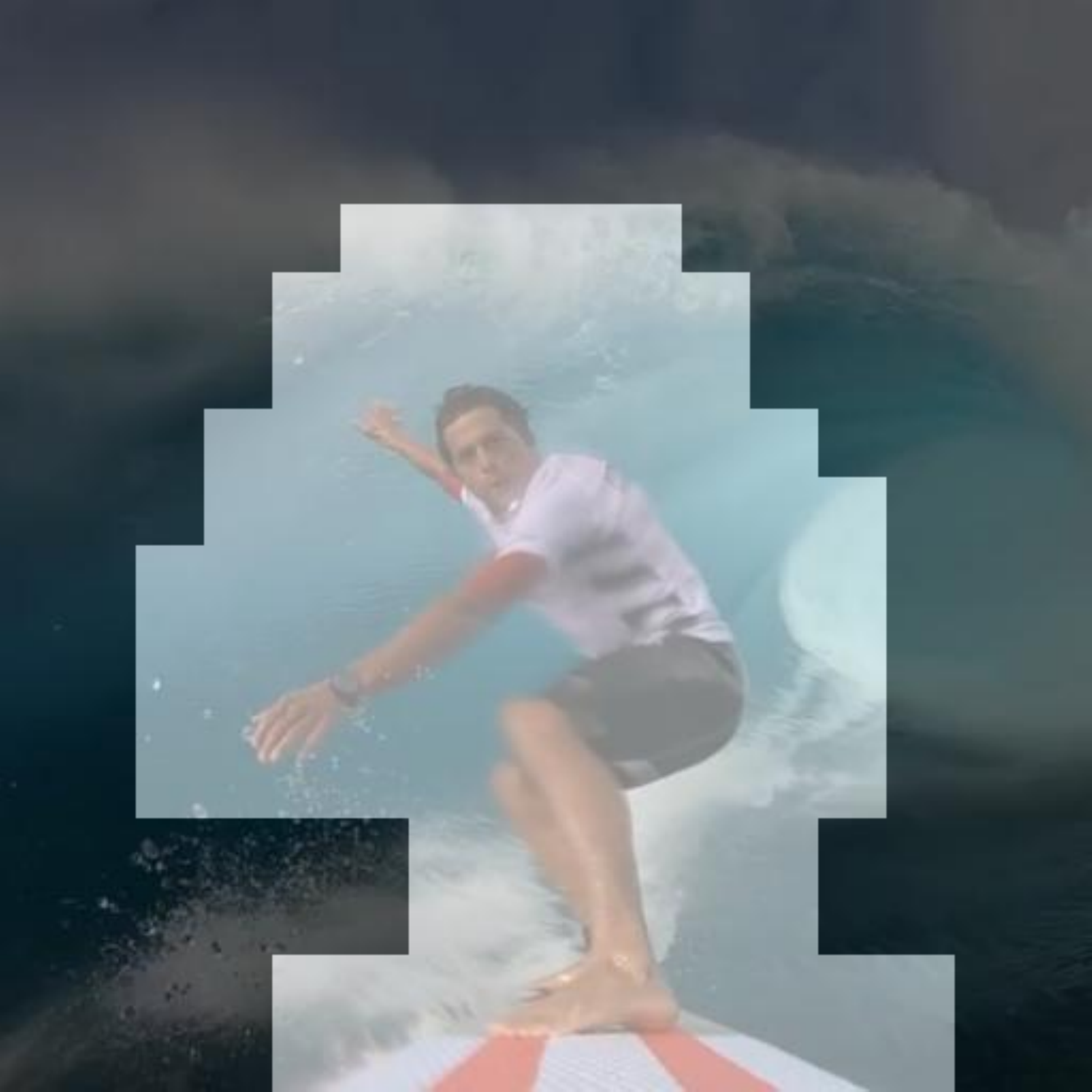}
\includegraphics[width=0.11\textwidth]{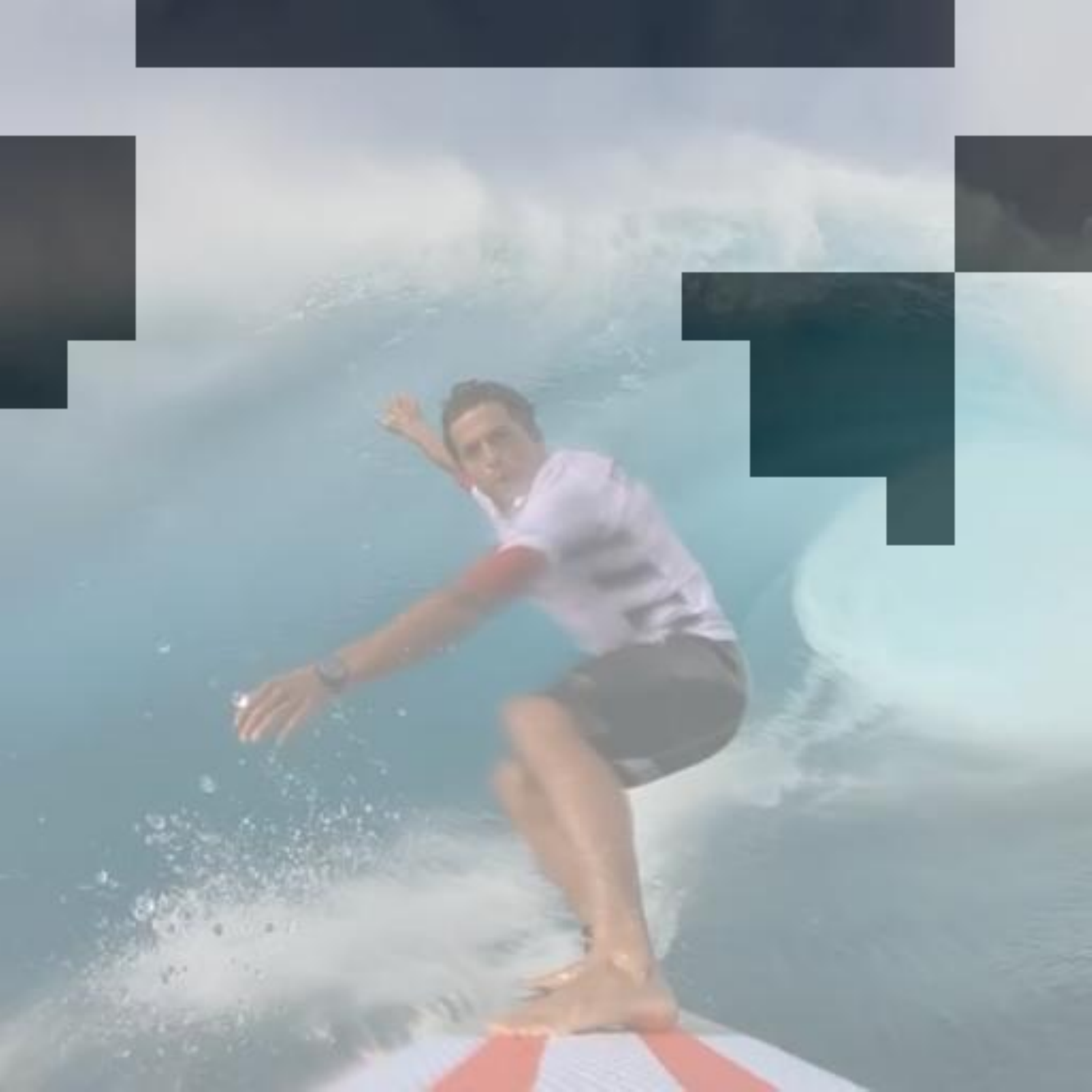}
\includegraphics[width=0.11\textwidth]{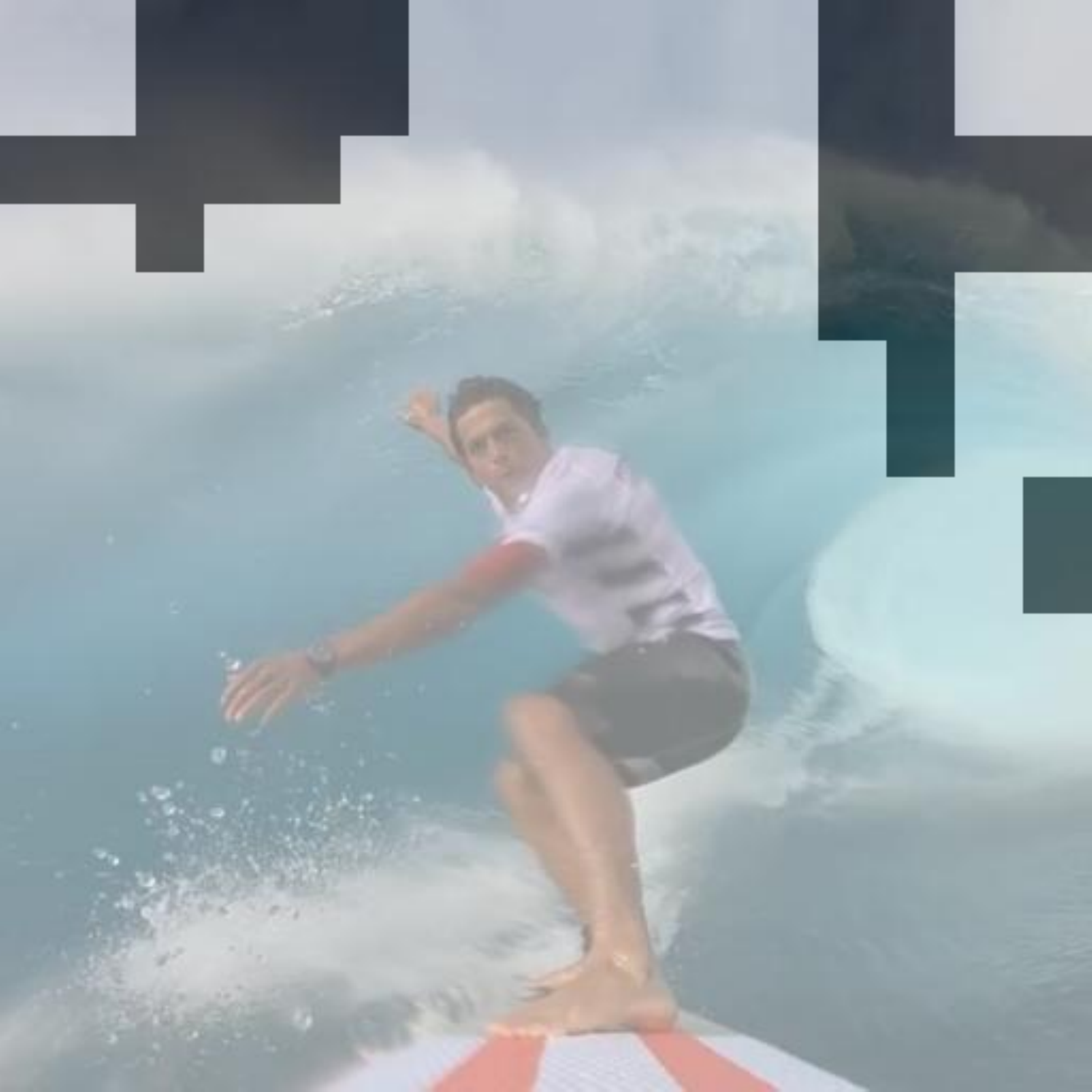}
\includegraphics[width=0.11\textwidth]{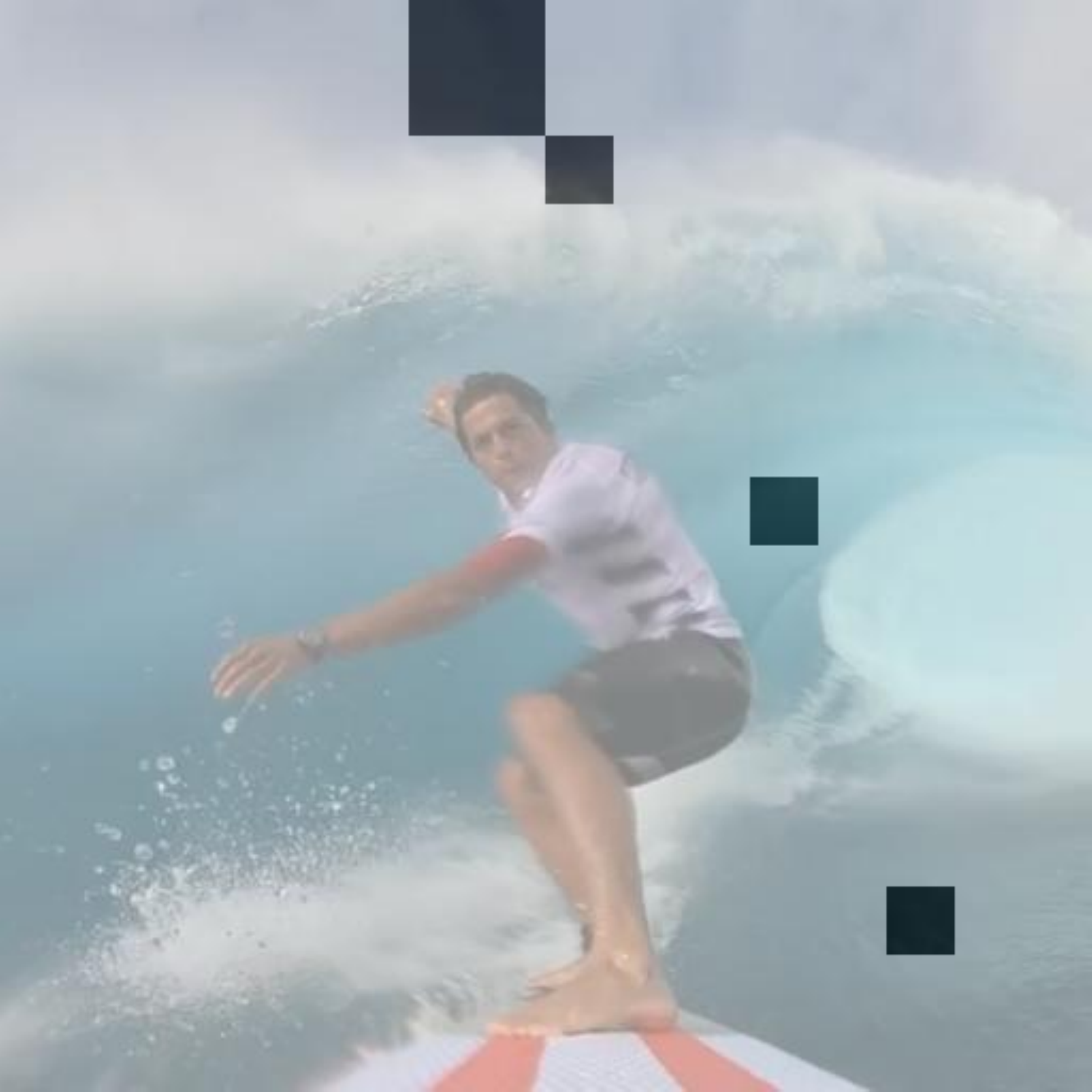}
\includegraphics[width=0.11\textwidth]{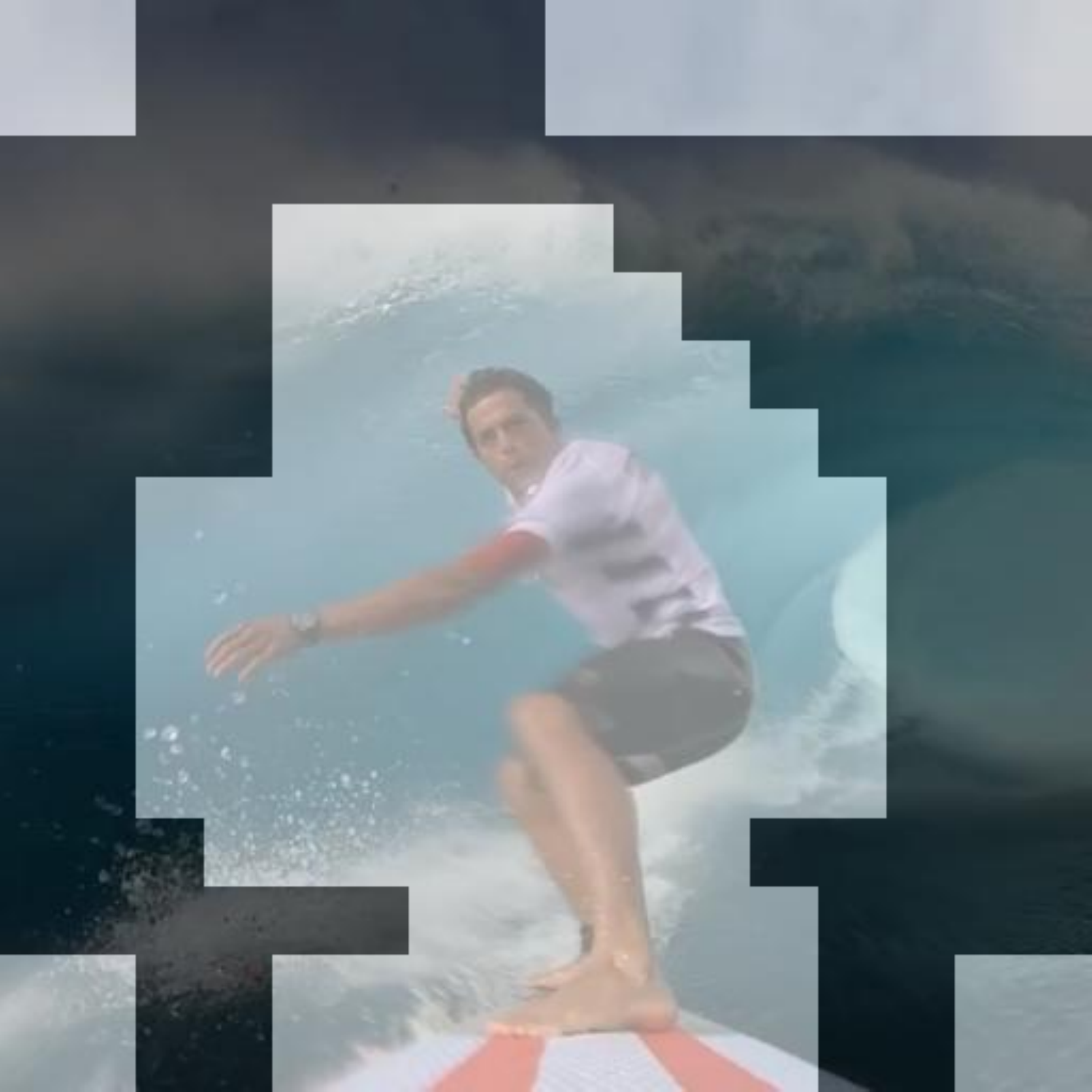}
\includegraphics[width=0.11\textwidth]{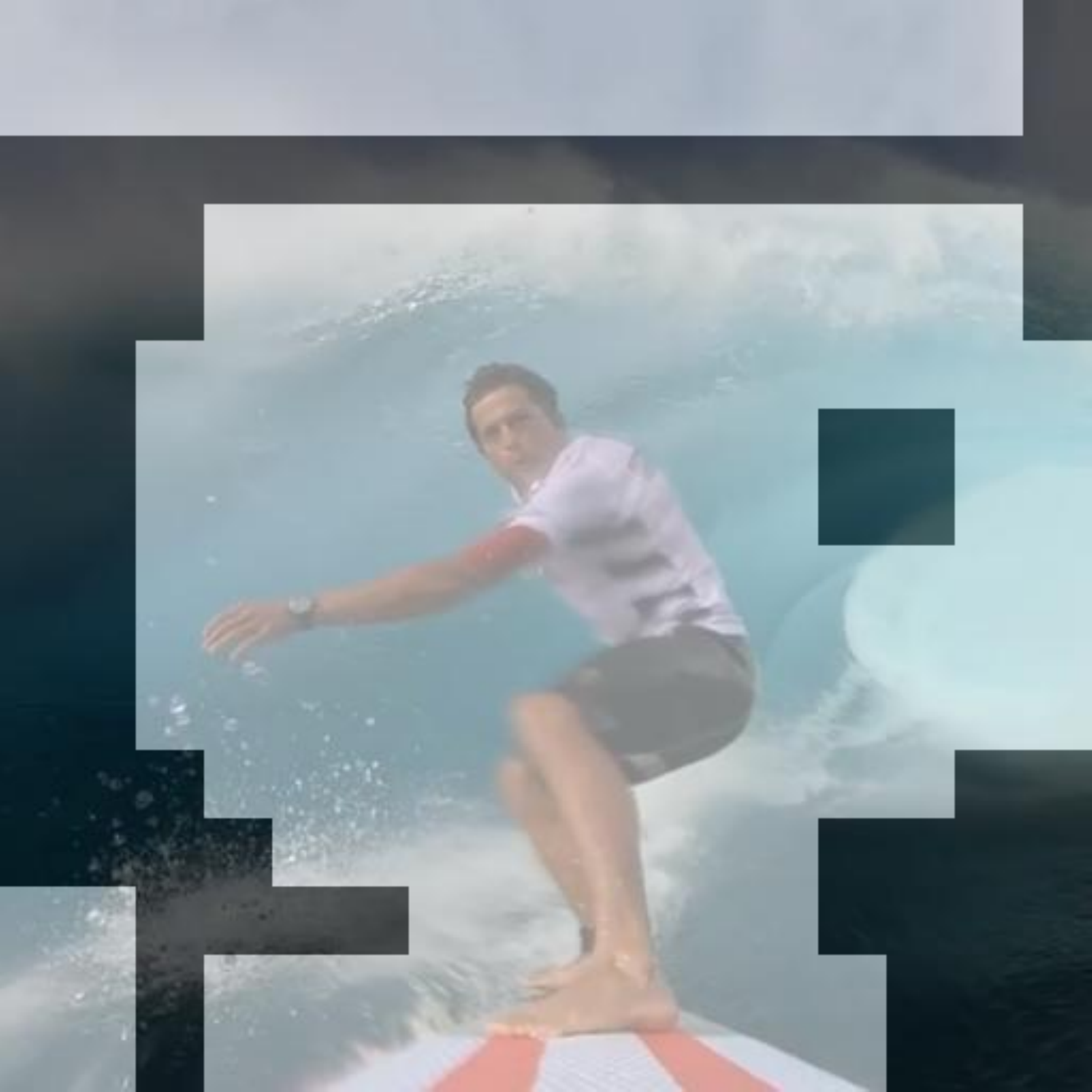}

\includegraphics[width=0.11\textwidth]{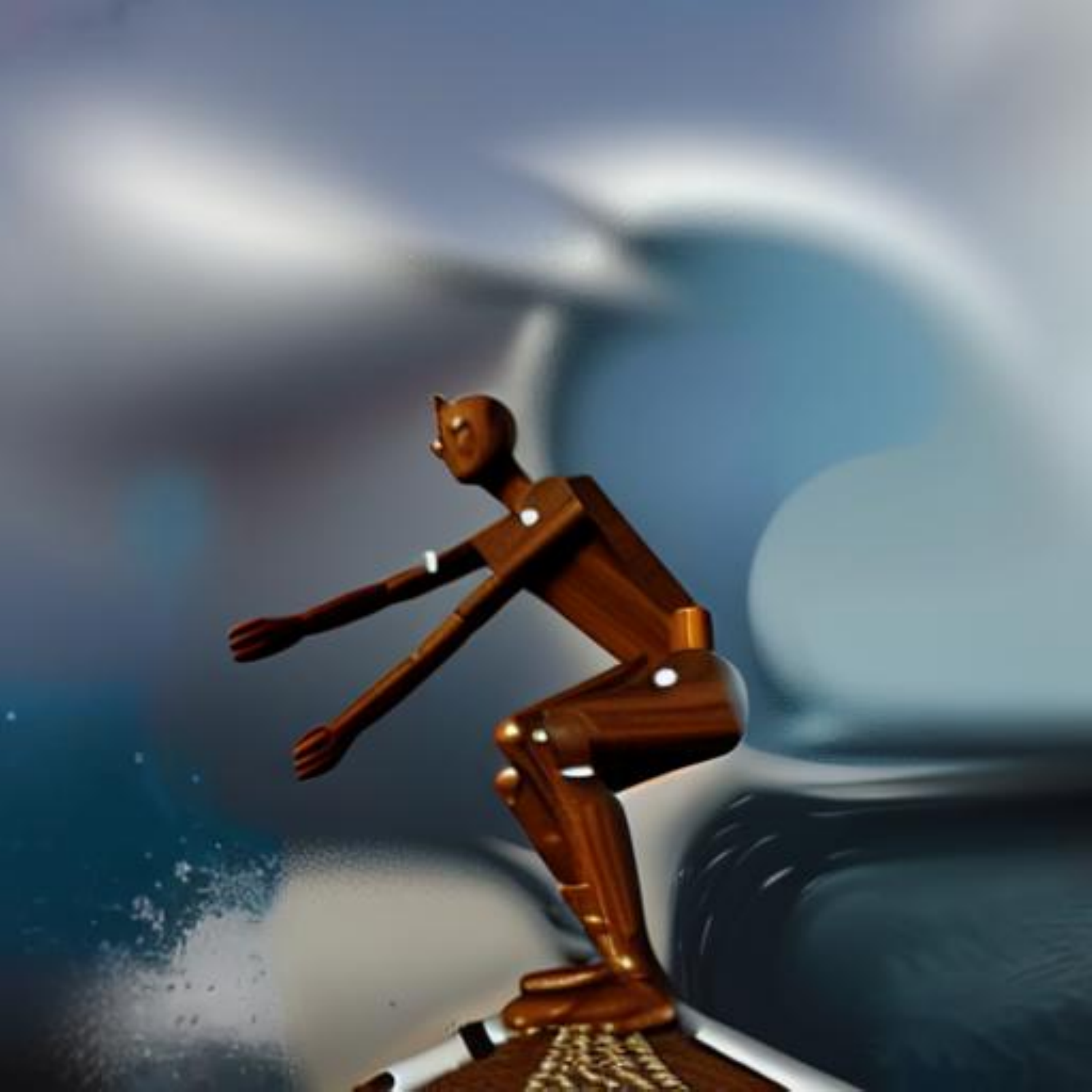}
\includegraphics[width=0.11\textwidth]{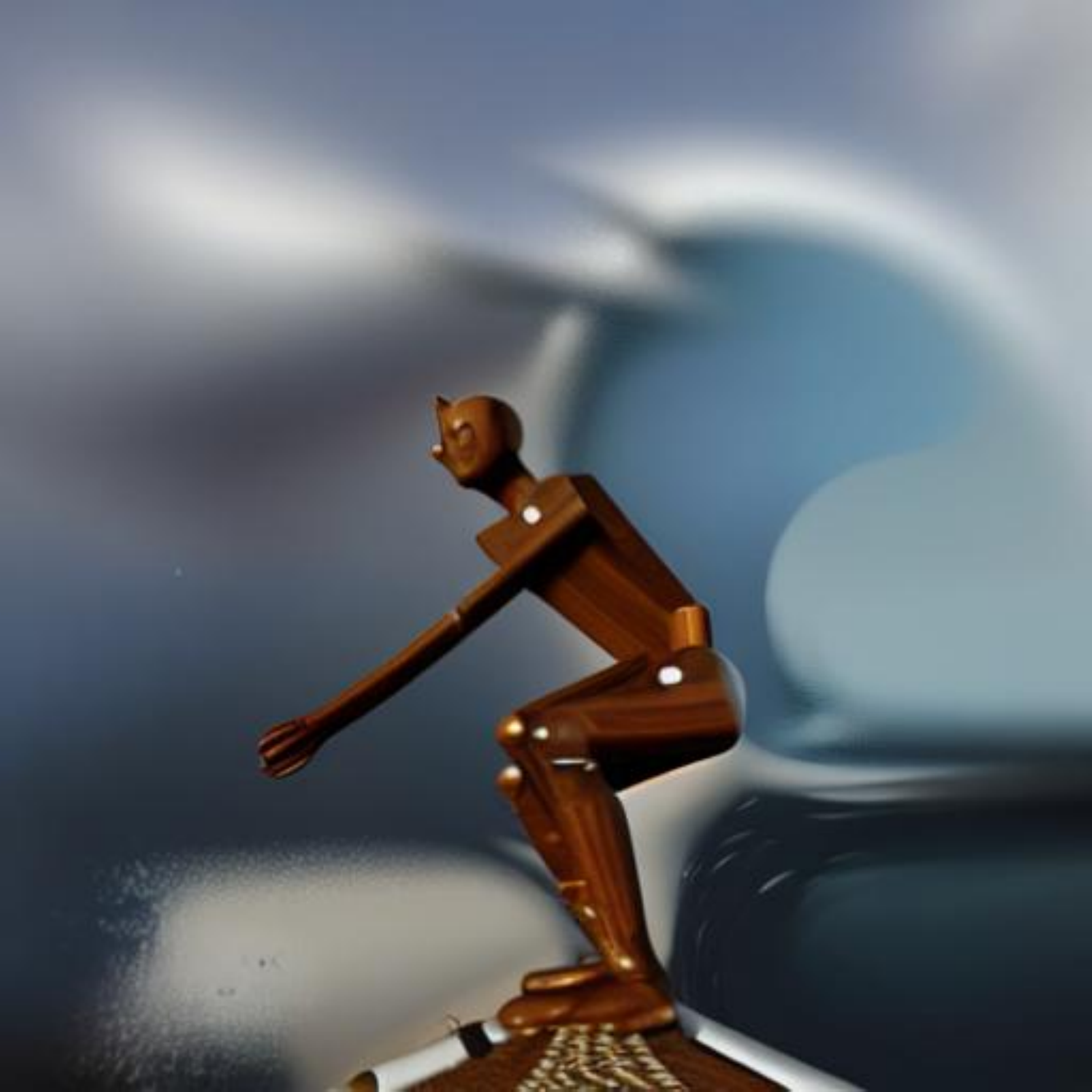}
\includegraphics[width=0.11\textwidth]{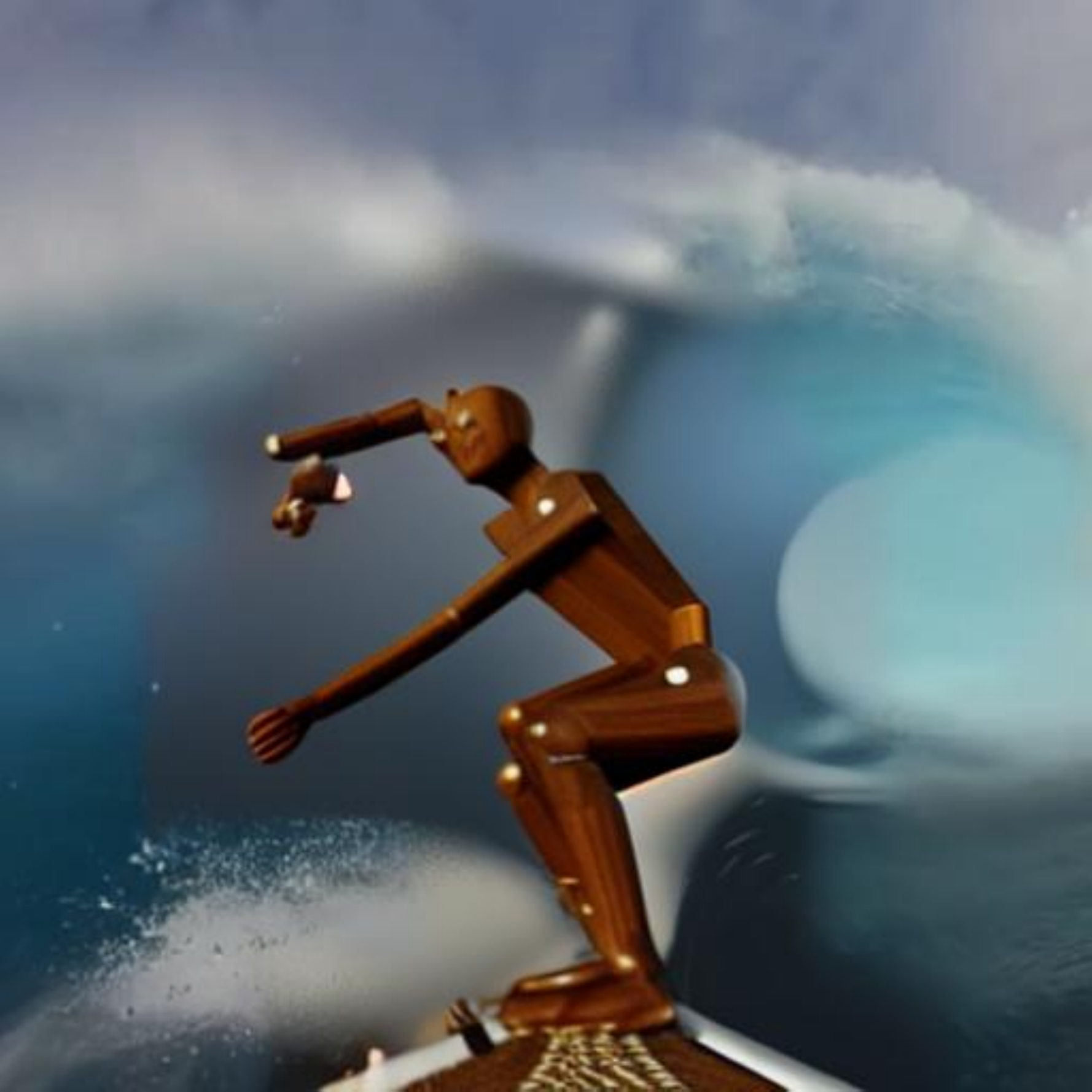}
\includegraphics[width=0.11\textwidth]{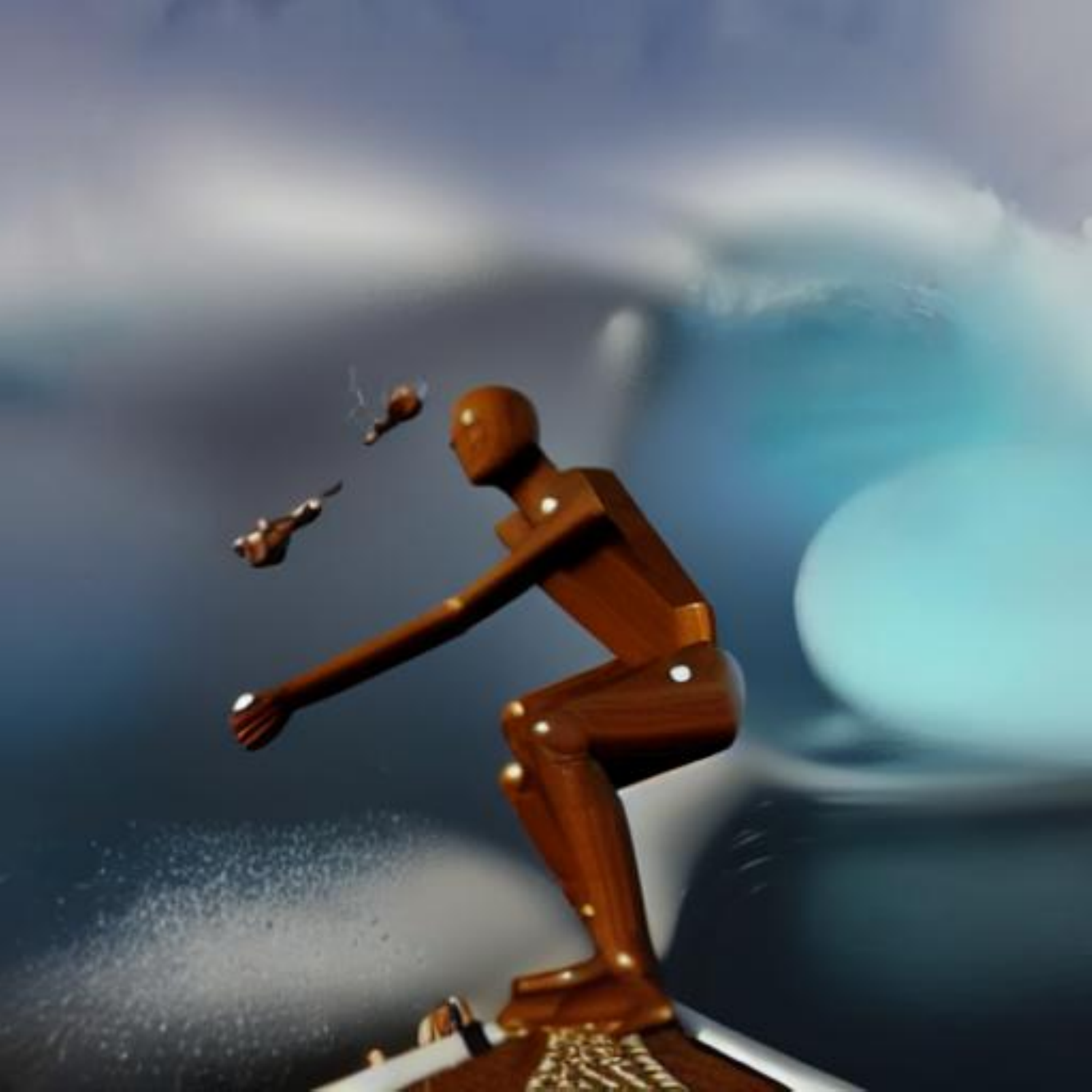}
\includegraphics[width=0.11\textwidth]{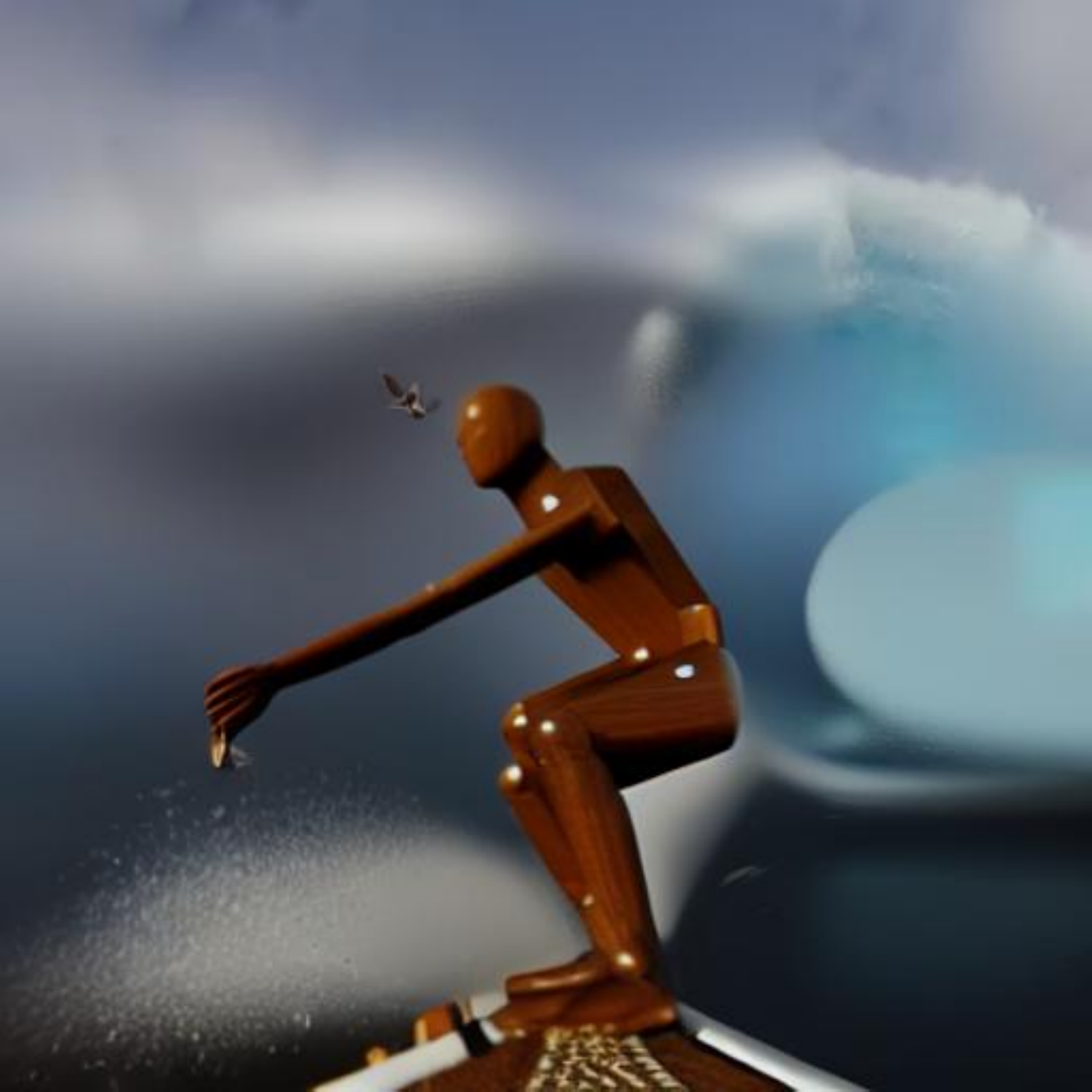}
\includegraphics[width=0.11\textwidth]{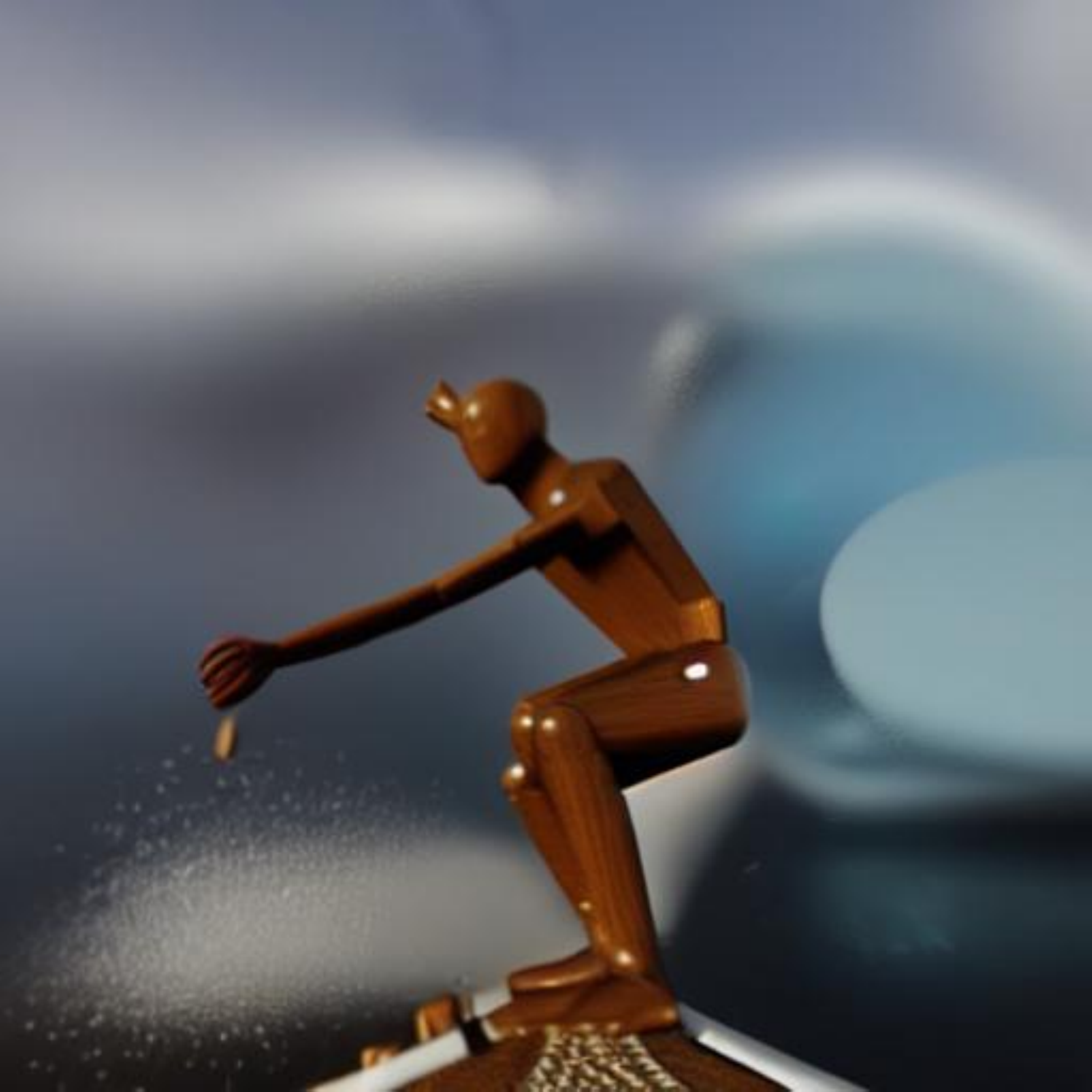}
\includegraphics[width=0.11\textwidth]{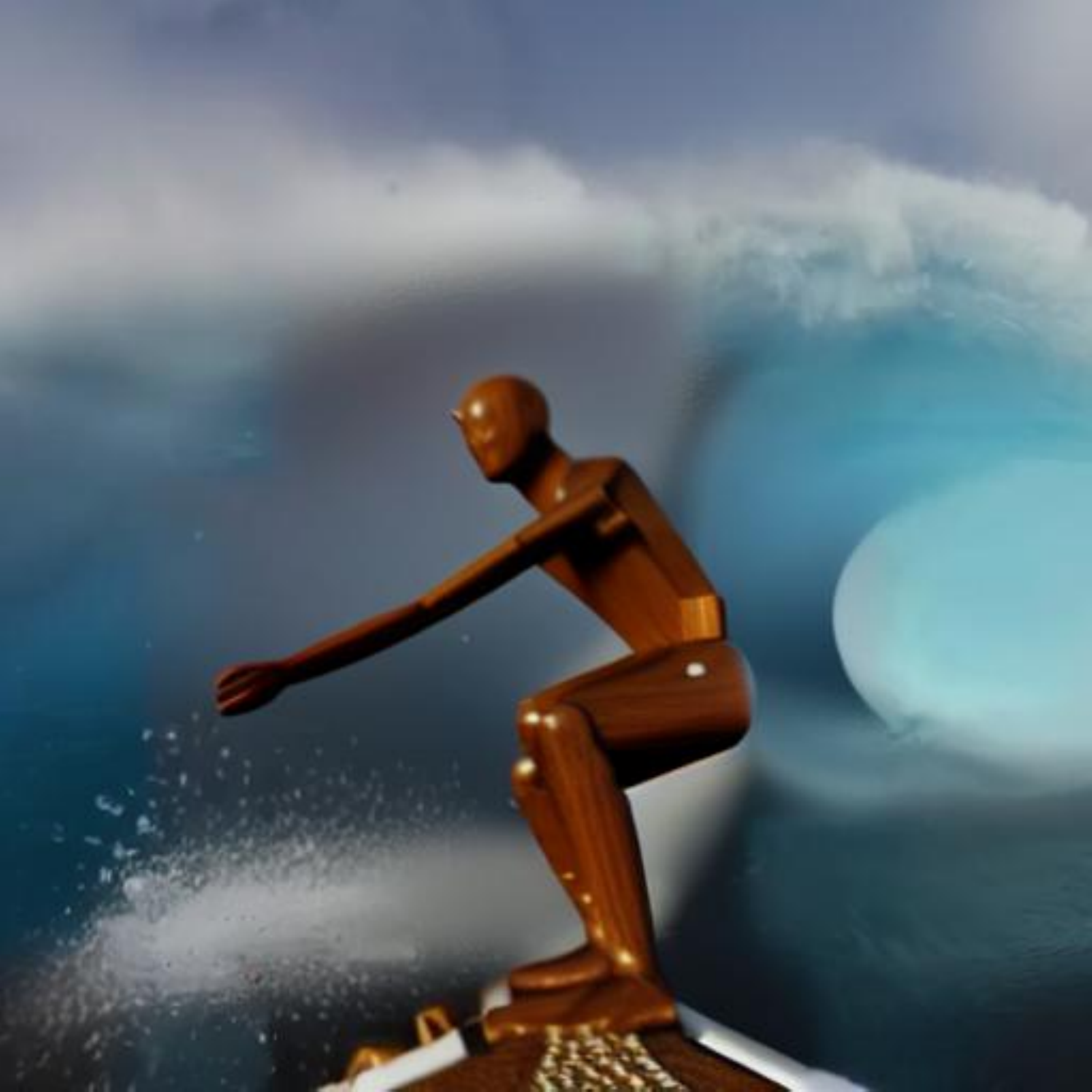}
\includegraphics[width=0.11\textwidth]{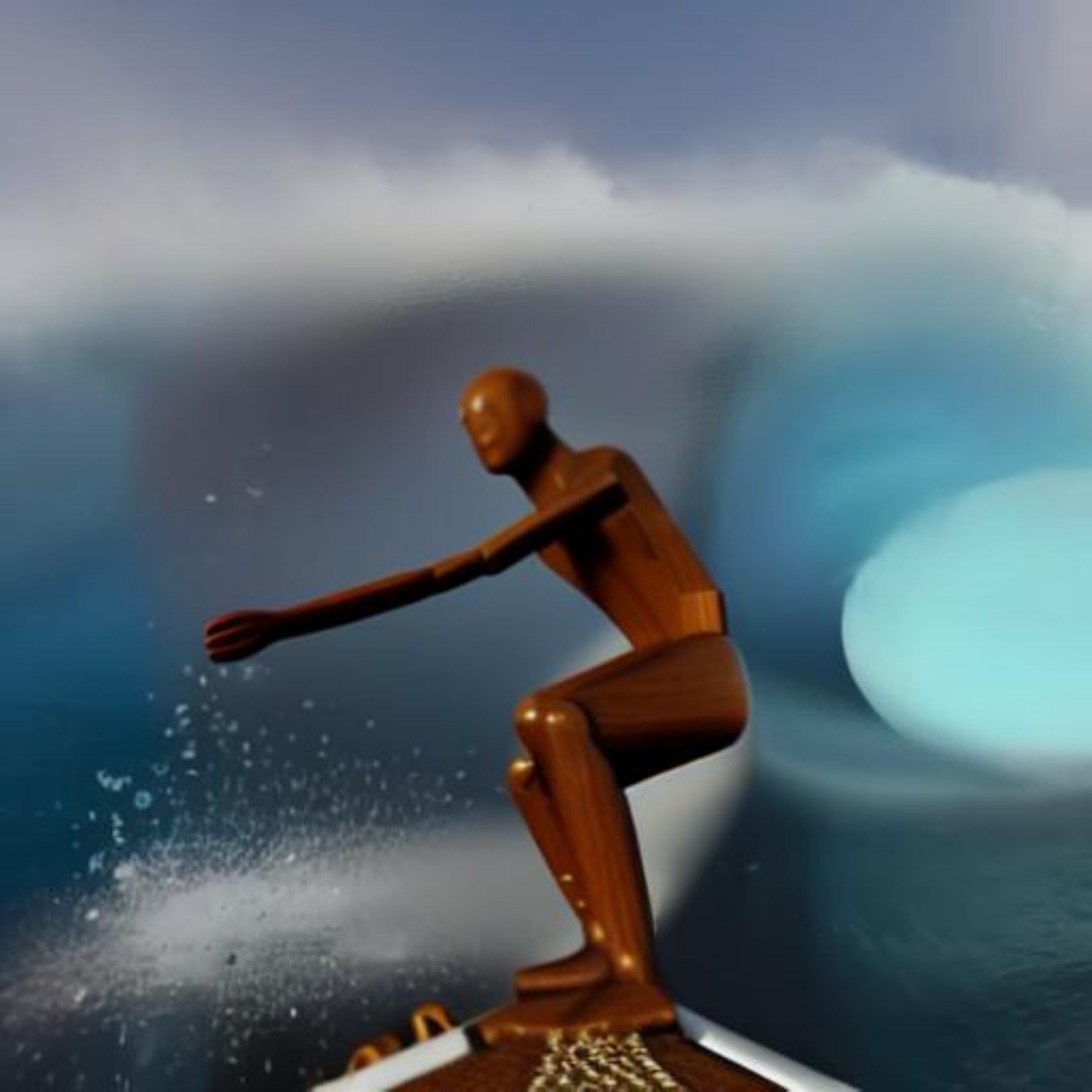}

\caption{\textbf{Background Inconsistency Problem} The spatial local blending mask originally proposed in \cite{hertz2022prompt} cannot consider the temporal consistency, resulting in temporally variant background after editing, which can be identified in the waves close to the wooden sculpture.}
\label{fig:bg}
\end{center}
\vspace{-1.3em}
\end{figure*}

\subsection{Temporal-Consistent Blending} \label{sc}
When Edit-A-Video edits an object, only the specific region that correlates to the target object should be modified, while the rest of the video should be preserved.
In previous work of image editing, PTP \citep{hertz2022prompt} proposed the local blending method, which approximates a mask of the object region from the cross-attention map and performs editing only in the masked region.
However, since the cross-attention of the inflated 3D model is computed frame-by-frame, the local blending mask from the cross-attention map considers only the spatial dimension and lacks modeling of temporal dependency.
This indicates that the smoothness of blending masks across frames cannot be ensured, resulting in the potential occurrence of color-variant artifacts when a specific region is included in the blending mask of one frame but not in another.
This background inconsistency problem is demonstrated clearly in Fig. \ref{fig:bg}, where the background region close to the target object of editing is severely distorted, and the frames are highly inconsistent across the temporal axis.

To address this problem, we propose a novel blending method called temporal-consistent blending (TC Blending), which acquires a spatio-temporally consistent blending mask.
For efficient temporal modeling, we use the sparse ST attention map in our 3D inflated model as a proxy for the interaction between the current frame mask and the first and previous frame masks.
The sparse ST attention map is calculated by taking the current frame feature $z_{f}$ as query, the first frame feature $z_{1}$ and the previous frame feature $z_{f-1}$ as key:
\begin{equation}
    \mathrm{ST}\mbox{-}\mathrm{Attn}_{f} = \mathrm{Softmax}(QK^T/\sqrt{d}), 
    Q = W^{Q}z_{f}, K = W^{K}[z_{1}; z_{f-1}],
\end{equation}
where $f$ is the index of current frame, [;] denotes the concatenation, and $d$ is the feature dimension of the projection layer.  

Intuitively, the sparse ST attention map of the current frame with the first frame ensures that the target object is maintained in every mask, while that with the previous frame enforces a smooth transition in the blending mask sequence.
With TC Blending, it is possible to achieve a sharp blending mask that accurately detects the target object in the frame and ensures a smooth transition in the blending masks, reducing background inconsistencies.

\begin{figure}[t]
    \centering
    \includegraphics[width=0.8\linewidth]{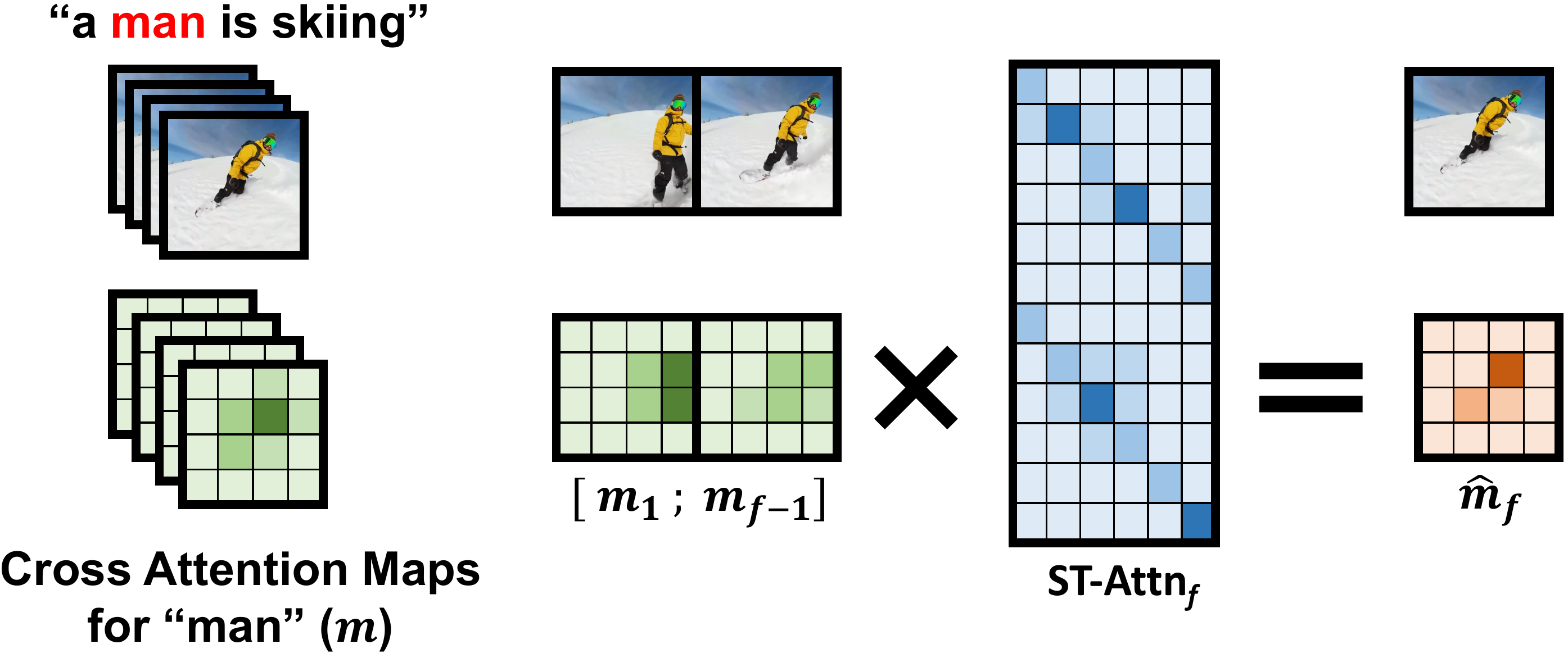}
    \caption{\textbf{TC Blending Mask Computation} The first and previous frames interact with the corresponding sparse spatio-temporal attention map and yield the new temporal-consistent blending mask.}
    \label{fig2}
    \vskip -0.1in
\end{figure} 
Following \cite{hertz2022prompt}, we first acquire the cross-attention maps $m \in {\rm I\!R^{\mathit{F} \times \mathit{H} \times \mathit{W}}}$ according to the original word and new word as initial blending maps, where the $F,H,W$ are the temporal and spatial dimension of the feature.
Then, for each attention map $m$, we normalize it frame-wise to balance the scale between each frame and flattening:
\begin{equation}
    \tilde{m} = \mathrm{Flatten}(m/ \sum_{H,W} (m)), \text{where } \tilde{m} \in {\rm I\!R^{\mathit{F}x\mathit{HW}}}.
\end{equation}

Afterward, we encourage the interaction between the frame-wise maps by computing the weighted average of the first and previous map $[\tilde{m}_{1}; \tilde{m}_{f-1}] \in {\rm I\!R^{(\mathit{HW}\times 2)}}$ with sparse ST attention map $\mathrm{ST}\mbox{-}\mathrm{Attn}_{f} \in {\rm I\!R^{(\mathit{HW}\times2)\times(\mathit{HW})}}$ of the current frame, which is shown in Fig.~\ref{fig2}.
Then, we binarize it by thresholding:
\begin{equation}
    \alpha_{f} = B(\hat{m}_{f}, \tau), \hat{m}_{f} = [\tilde{m}_{1}; \tilde{m}_{f-1}] \times \mathrm{ST}\mbox{-}\mathrm{Attn}_{f},
\end{equation}
where $B$ is the binarizing function with threshold $\tau$.

Similar to PTP, we utilize the union of binary blending masks for the original word and new word, denoted as $\alpha$, as the final blending mask. This allows us to edit both the areas of the original object and the target object.
We use this binary blending mask $\alpha$ to perform local editing, where we preserve the background outside the mask and only generate the contents inside the mask:
\begin{equation}
    \hat{z}_{t} = \bar{z}_{t} \odot (1 - \alpha) + z_{t}^{*} \odot \alpha,
\end{equation}
where $\bar{z}_{t}$ is the reconstruction of the source video, $z_{t}^{*}$ is the edited video, and $\odot$ is element-wise multiplication.

\subsection{Hyperparameters for Editing}  \label{hparams}
The core idea of Prompt-to-Prompt (PTP) to preserve the spatial layout of input is by injecting 2D cross-attention maps and self-attention maps from pretrained TTI models.
The duration of the injection process, a portion of timestep over the entire sampling step until which the attention map is injected, is a key factor in controlling the reflection ratio of the target prompt. 
Edit-A-Video, on the other hand, uses sparse spatio-temporal attention and temporal attention instead of self-attention.
As a result, the effect of the duration of attention injection may differ from that of PTP.
In this section, we describe the effect of injection for each type of attention, and the corresponding samples are visualized in the Supplementary Materials.

\textbf{Cross-Attention}
The cross-attention layer performs an attention operation between text tokens and frames, taking into account the spatial layout of each text token in the frames. There is a trade-off depending on the duration of the injection phase. If injection occurs only at the beginning of generation, the generated frames are strongly conditioned on the target text prompt, but hard to maintain the spatial layout. 
On the other hand, if injection occurs throughout the entire generation process, 
the spatial layout of the source video is well preserved, yet it does not include the concepts from the text prompt. 
Empirically, we found that the duration of $0.2$ is sufficient to retain the spatial layout of the source video, while the generated video represents the semantics of the target text.

\textbf{Temporal Attention}
The temporal attention (T-Attn) layer is an additional attention module computed along the time axis of a video to model the temporal relationship between frames, yet it does not model spatial dependency. 
We found that T-Attn maps are distributed uniformly and that the duration of injection does not have a significant impact. We set a duration value of 0.8 for temporal attention. However, varying this value does not have a notable impact on the editing quality, and the corresponding qualitative results are in Supplementary Materials.

\textbf{Sparse Spatio-Temporal Attention}
The sparse spatio-temporal attention (ST-Attn) layer is an attention method designed for video, where the attention matrix of the current frame is calculated only on the first and previous frames.
ST-Attn replaces the self-attention layer from the pretrained TTI models, where the model only requires the dependencies between pixels in a single image. 
In addition to temporal attention, ST-Attn improves temporal consistency by attending to other frames while maintaining efficiency by only visiting two frames. 
We observed that insufficient duration of ST-Attn results in the inability to adequately represent the dynamic action of the source video. On the contrary, an excessively long duration tends to capture not only the actions of the source video but also the objects within it, which hinders the editing process toward the target object. 
Therefore, to maintain the dynamic of the source video and facilitate editing toward the target object, we set the duration of sparse spatio-temporal attention to 0.5.
\section{Experiments}

\begin{figure*}
\vspace{0.8em}
\begin{center}
\makebox[0.12\textwidth]{\colorbox{pink}{\textbf{Training video}} A woman is on the swing.}\\

\includegraphics[width=0.11\textwidth]{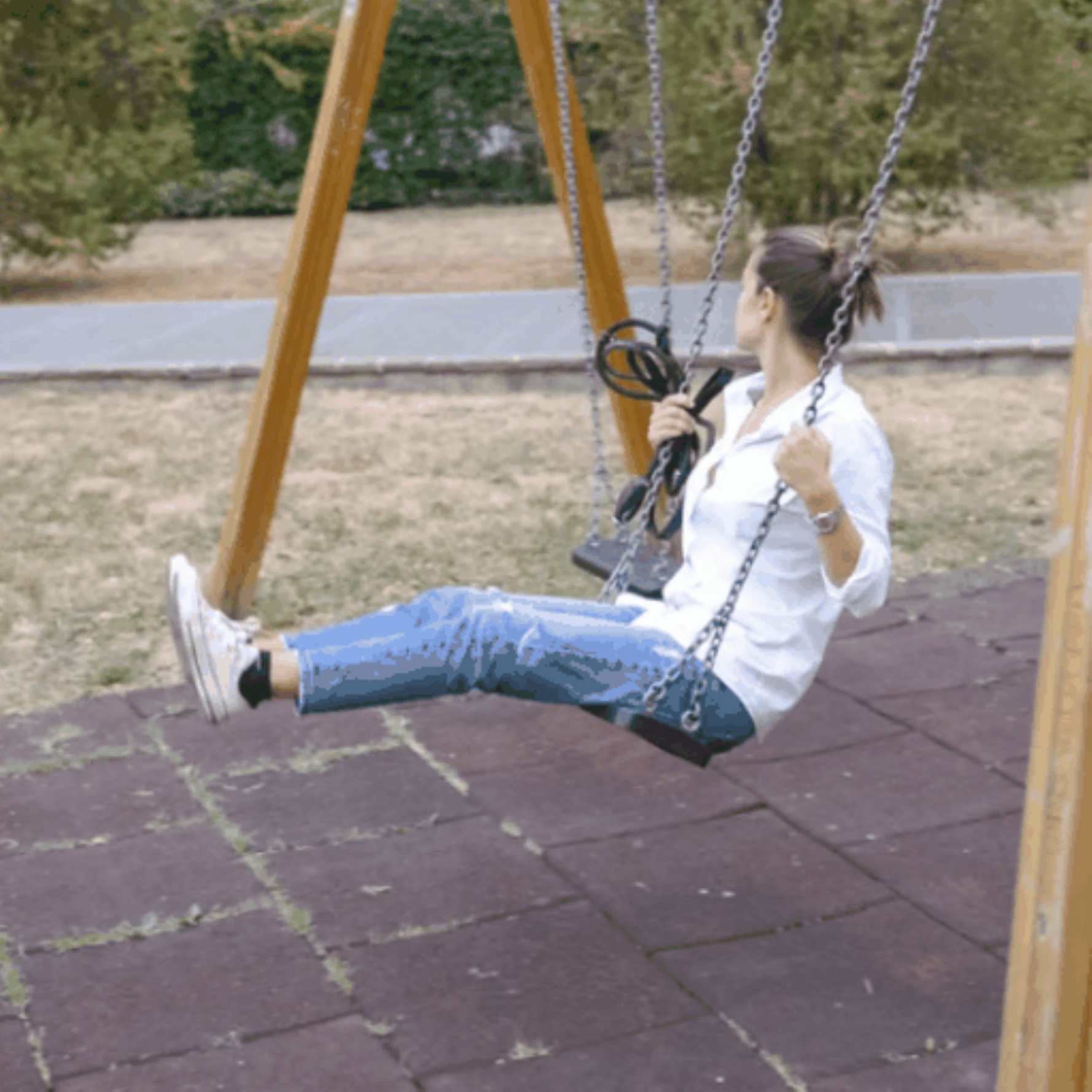}
\includegraphics[width=0.11\textwidth]{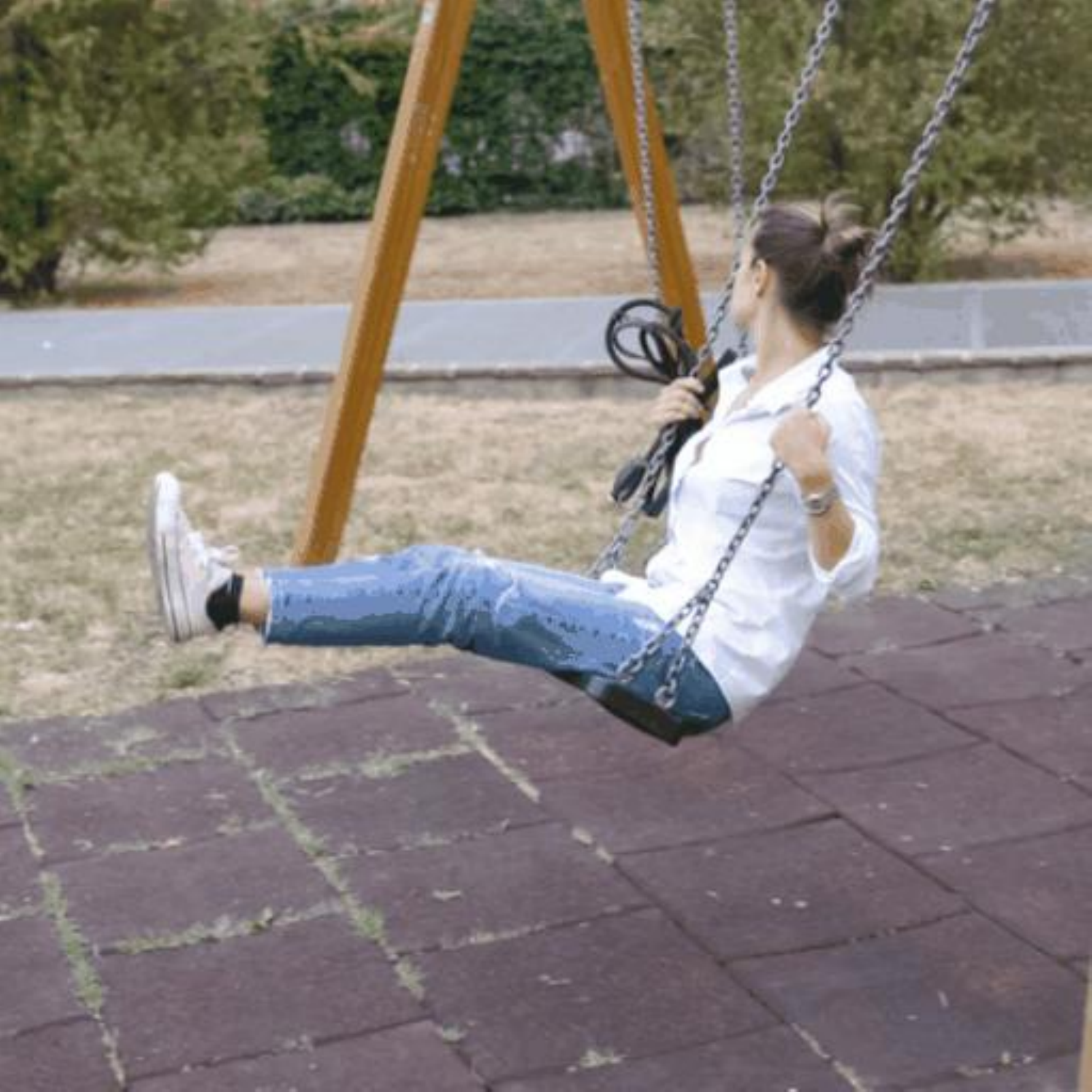}
\includegraphics[width=0.11\textwidth]{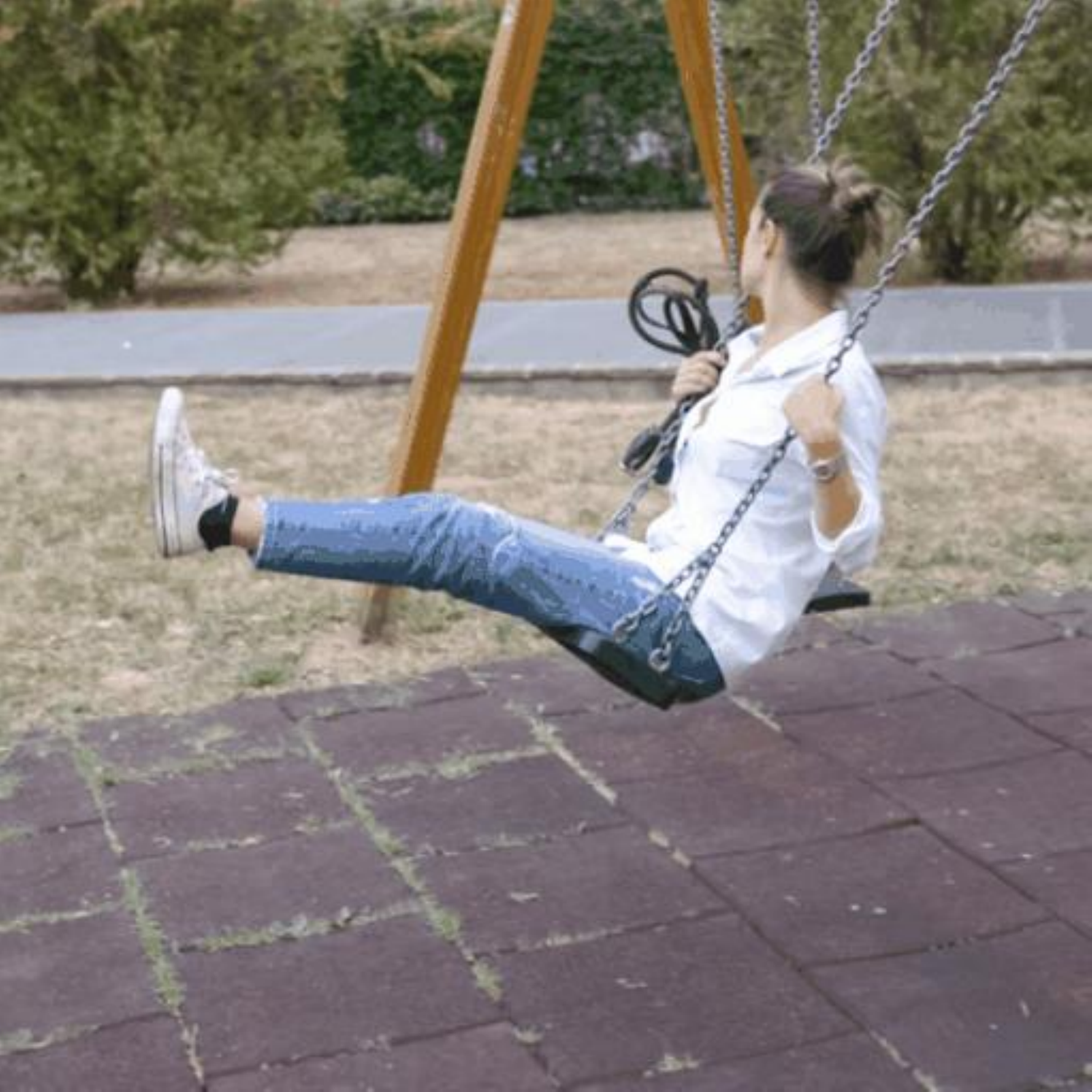}
\includegraphics[width=0.11\textwidth]{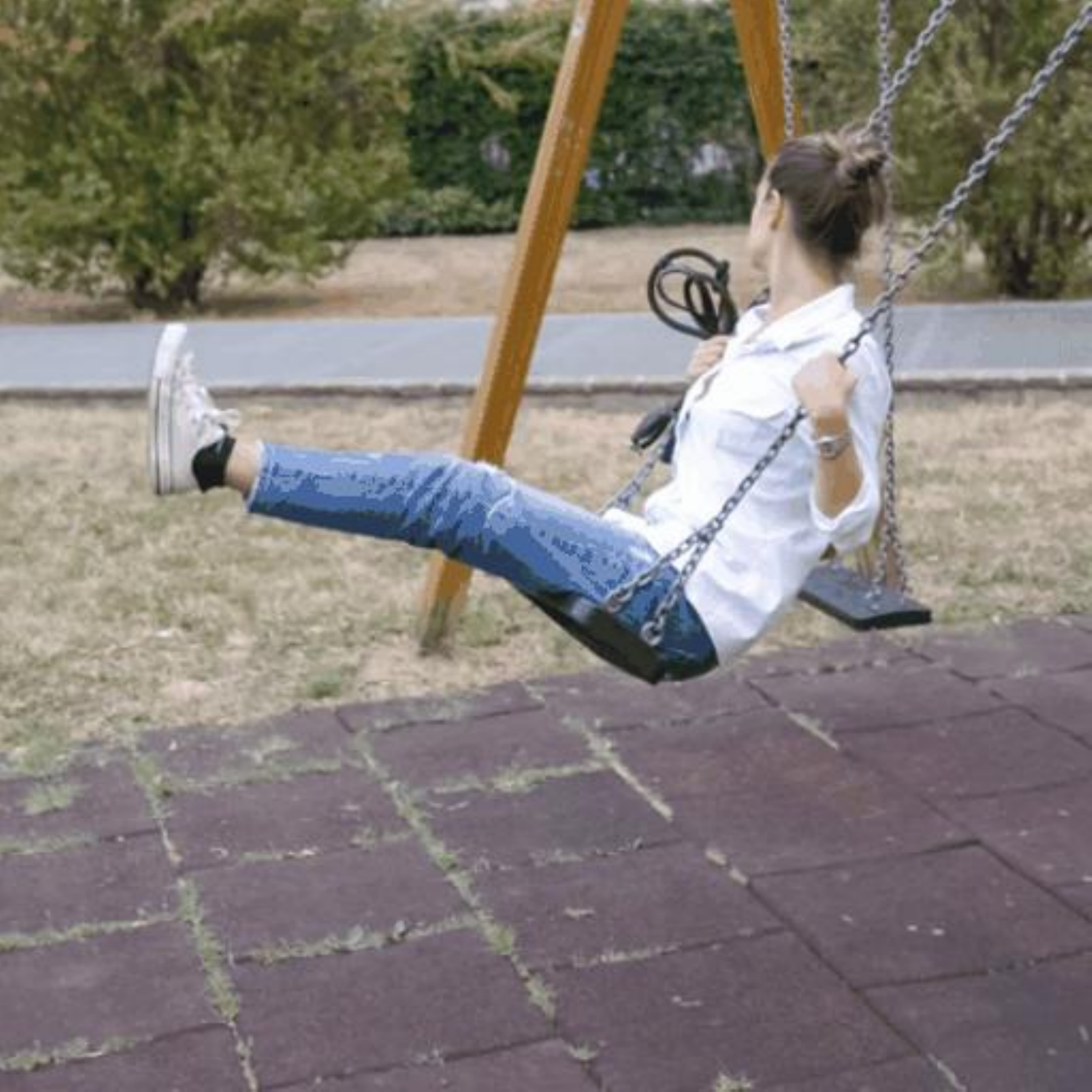}
\includegraphics[width=0.11\textwidth]{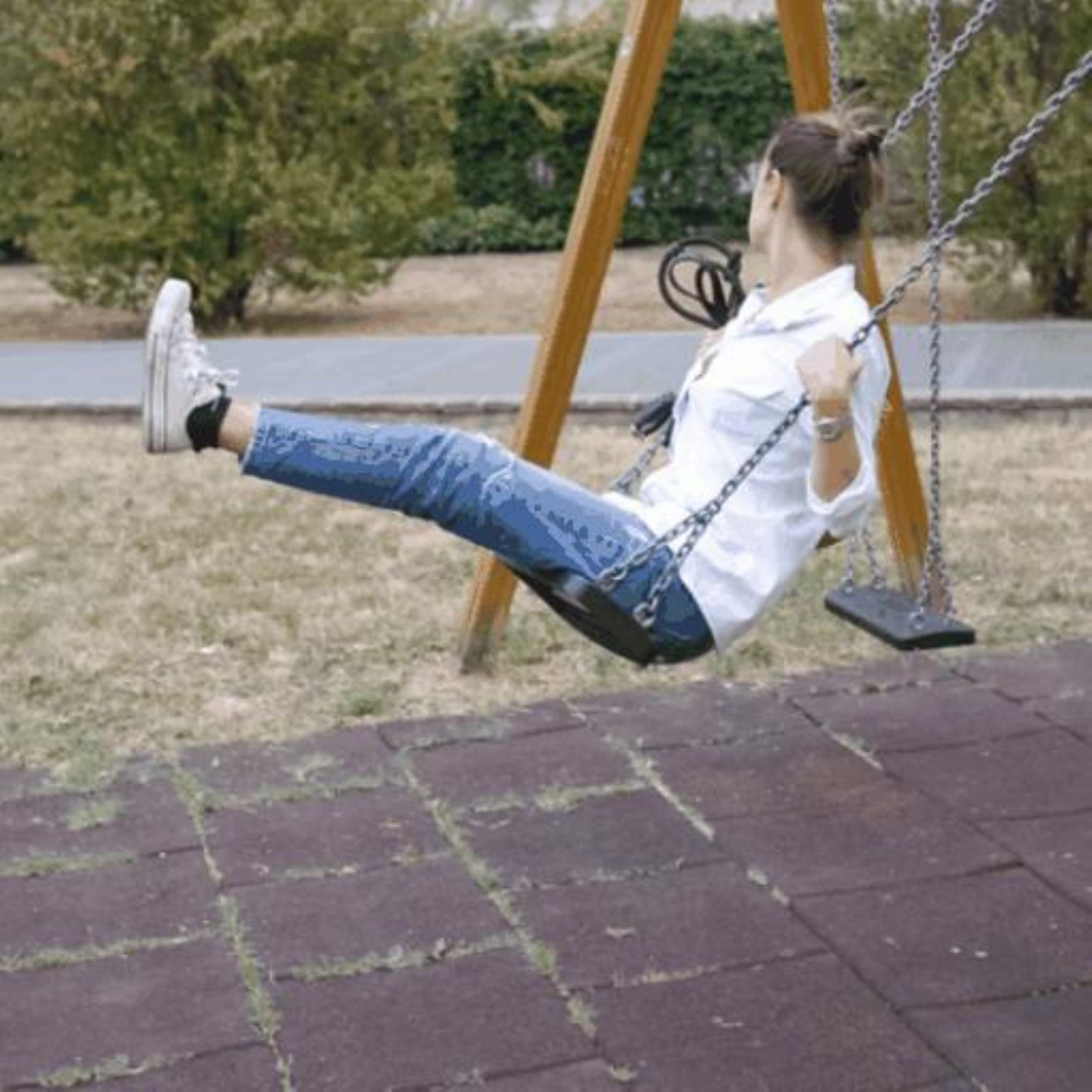}
\includegraphics[width=0.11\textwidth]{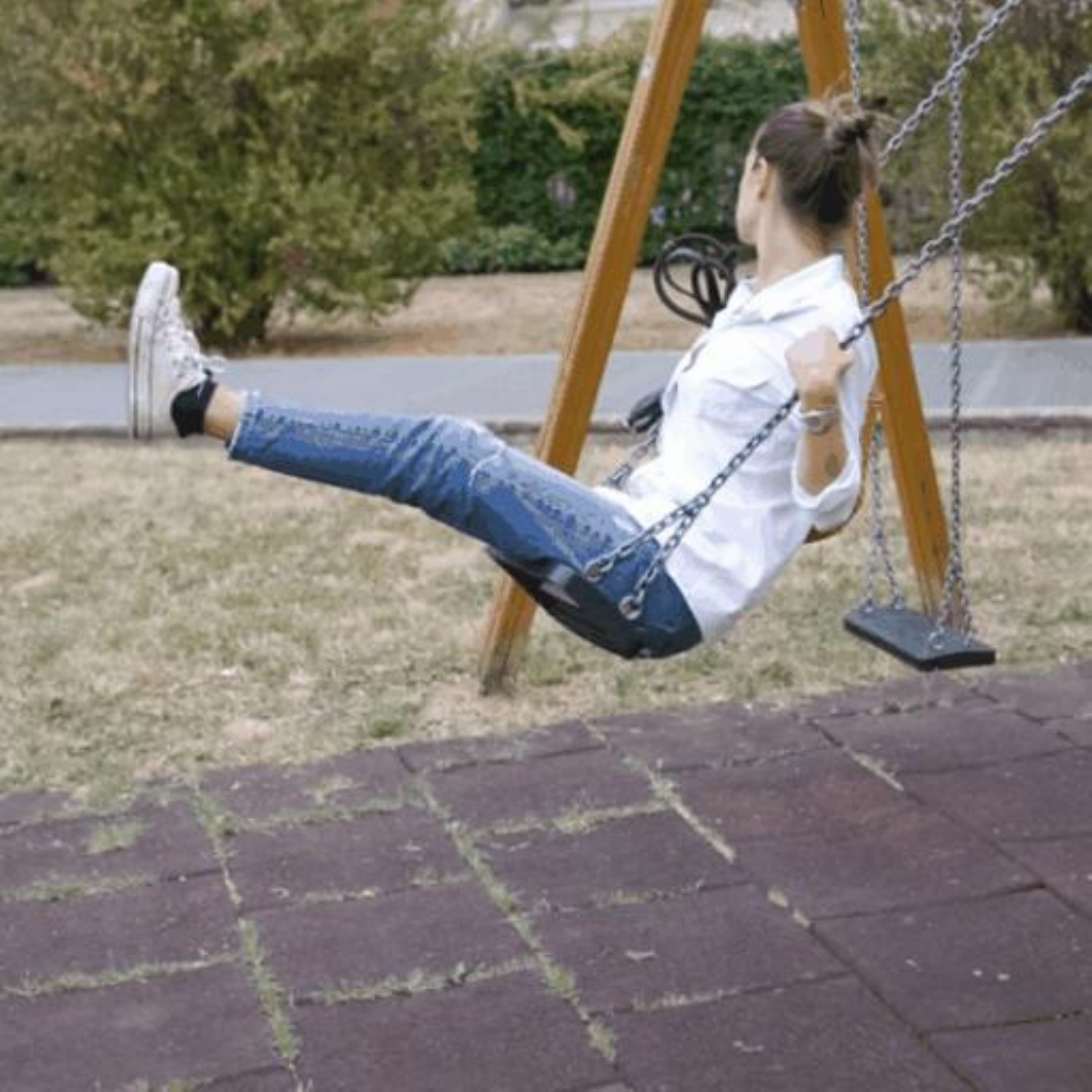}
\includegraphics[width=0.11\textwidth]{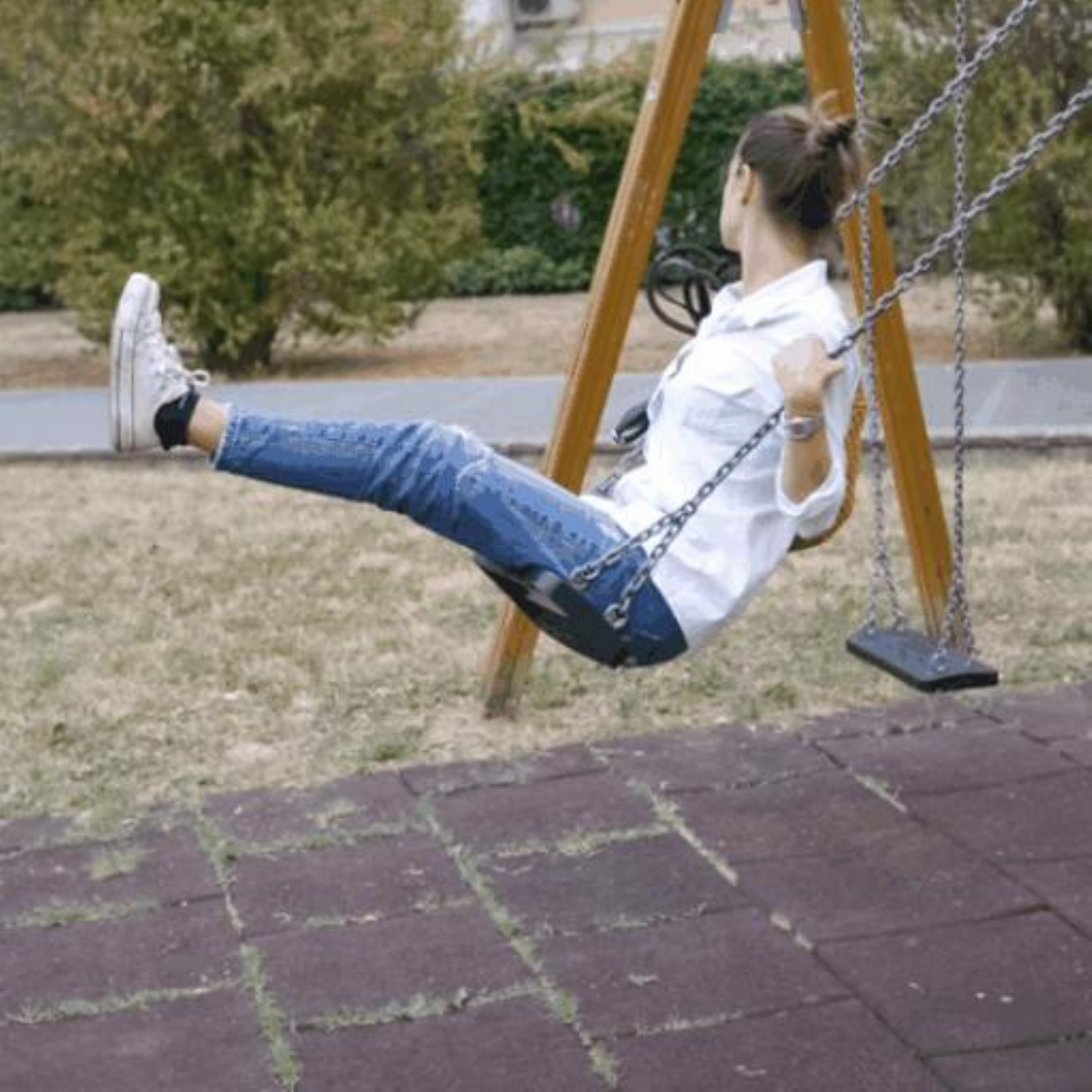}
\includegraphics[width=0.11\textwidth]{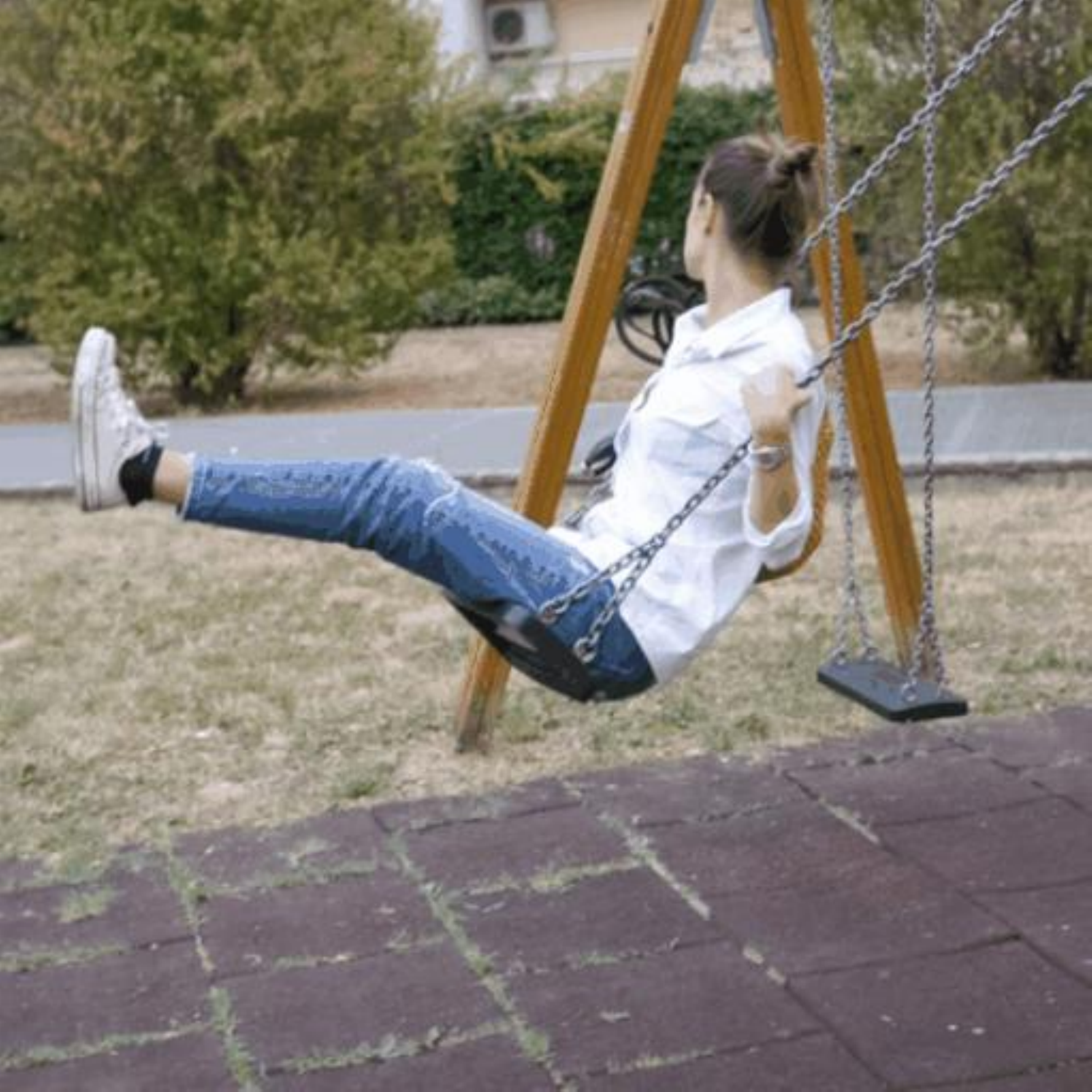}

\makebox[0.12\textwidth]{\colorbox{yellow}{\textbf{Edit-A-Video (Ours)}} A \textcolor{blue}{\textbf{watercolor}} painting that a woman is on the swing.}\\

\includegraphics[width=0.11\textwidth]{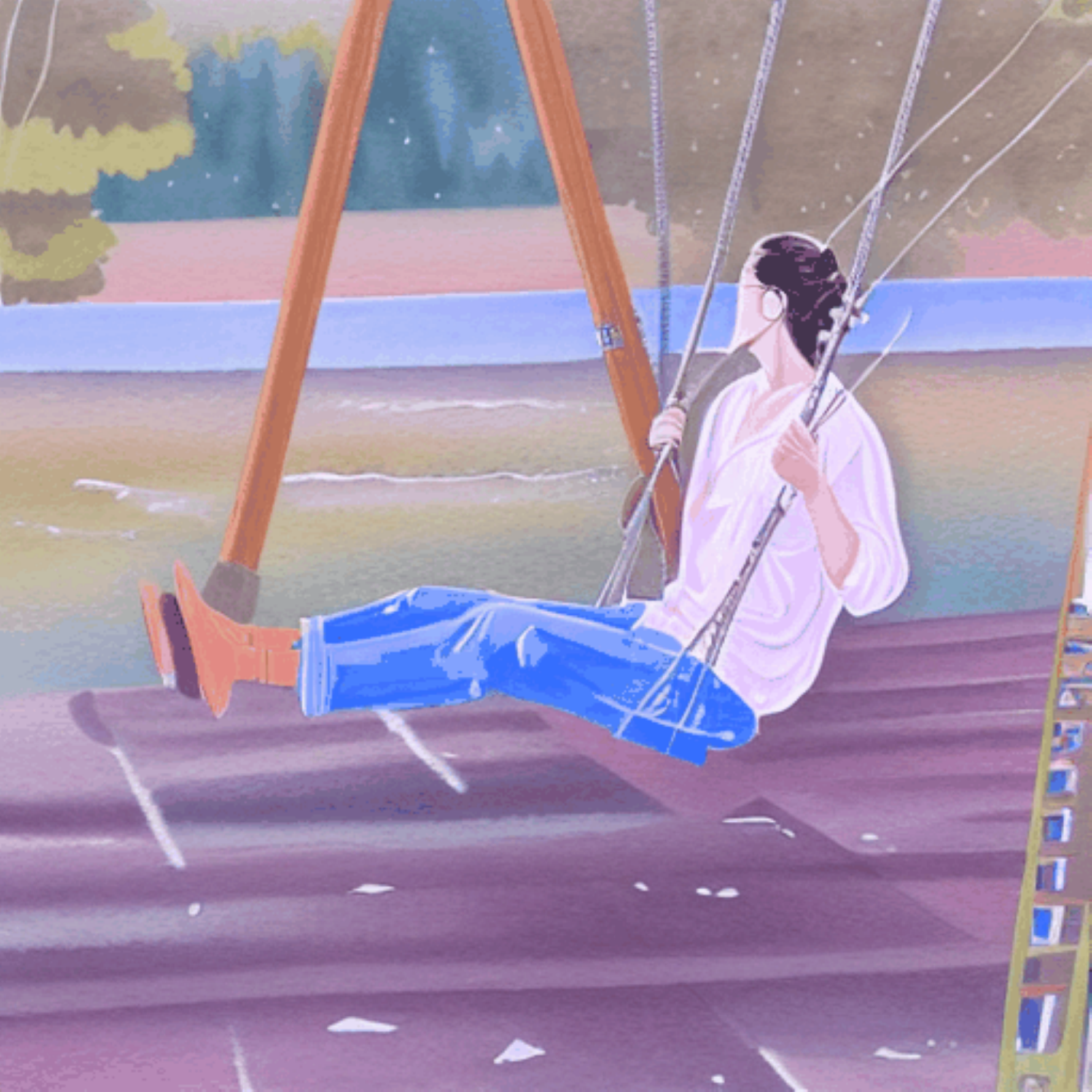}
\includegraphics[width=0.11\textwidth]{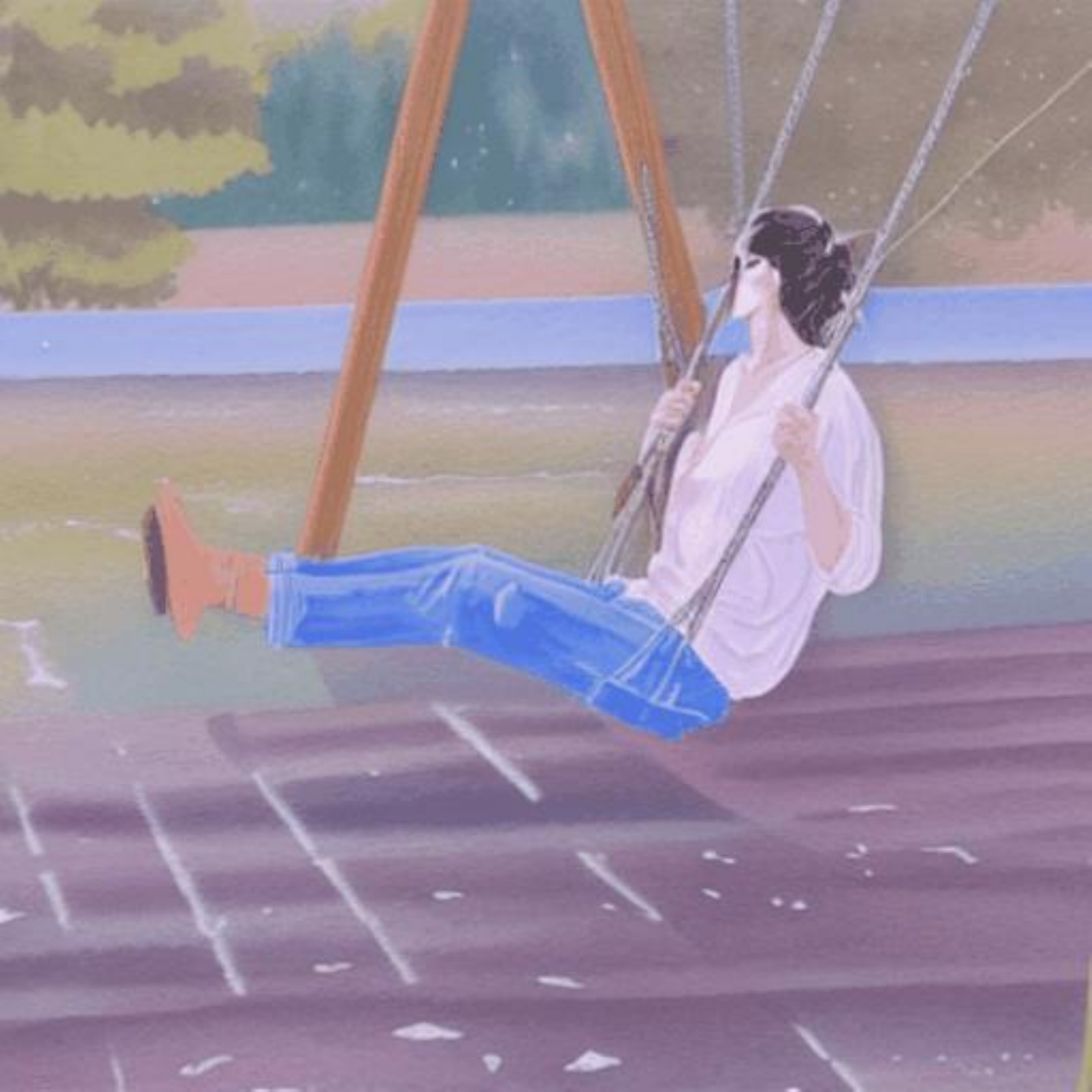}
\includegraphics[width=0.11\textwidth]{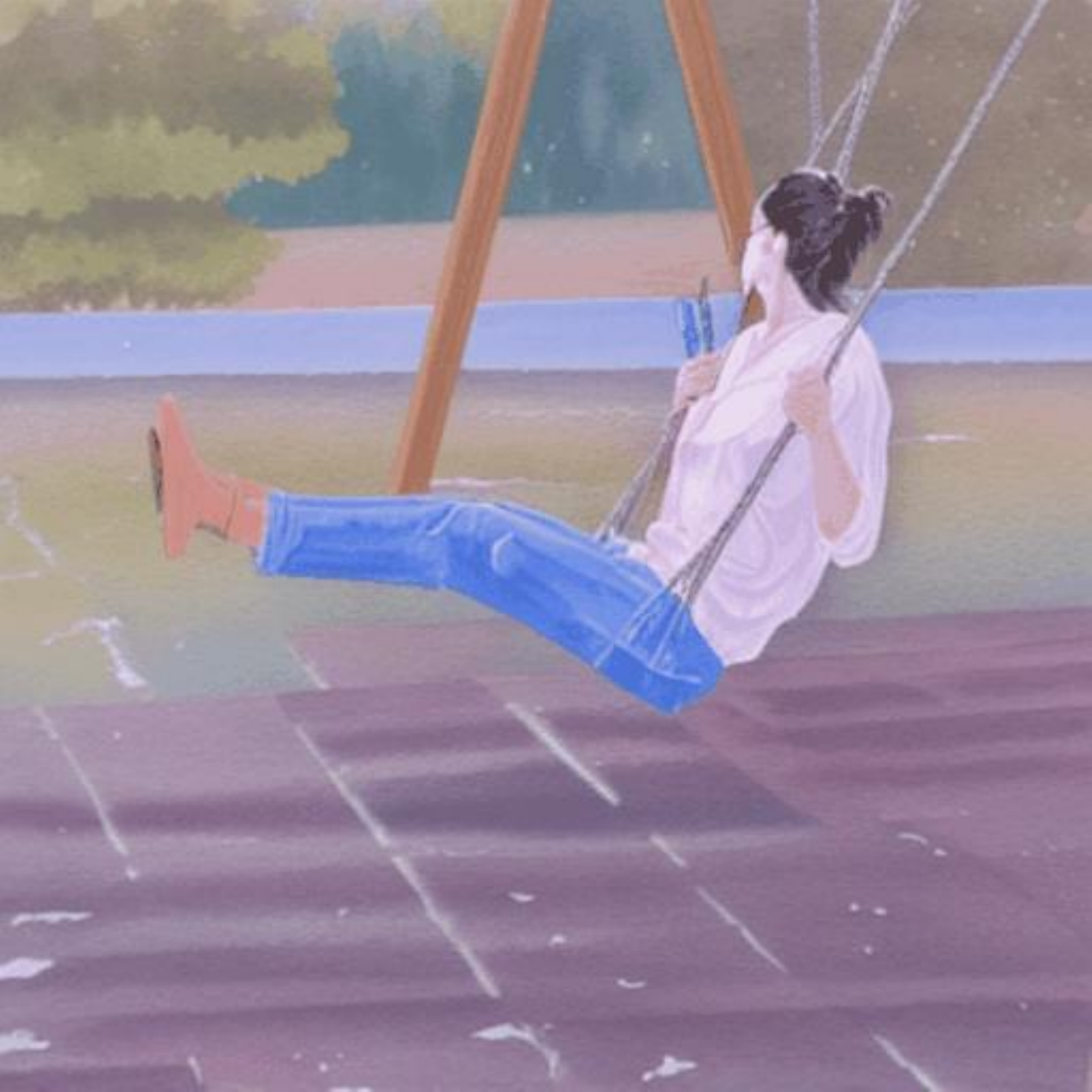}
\includegraphics[width=0.11\textwidth]{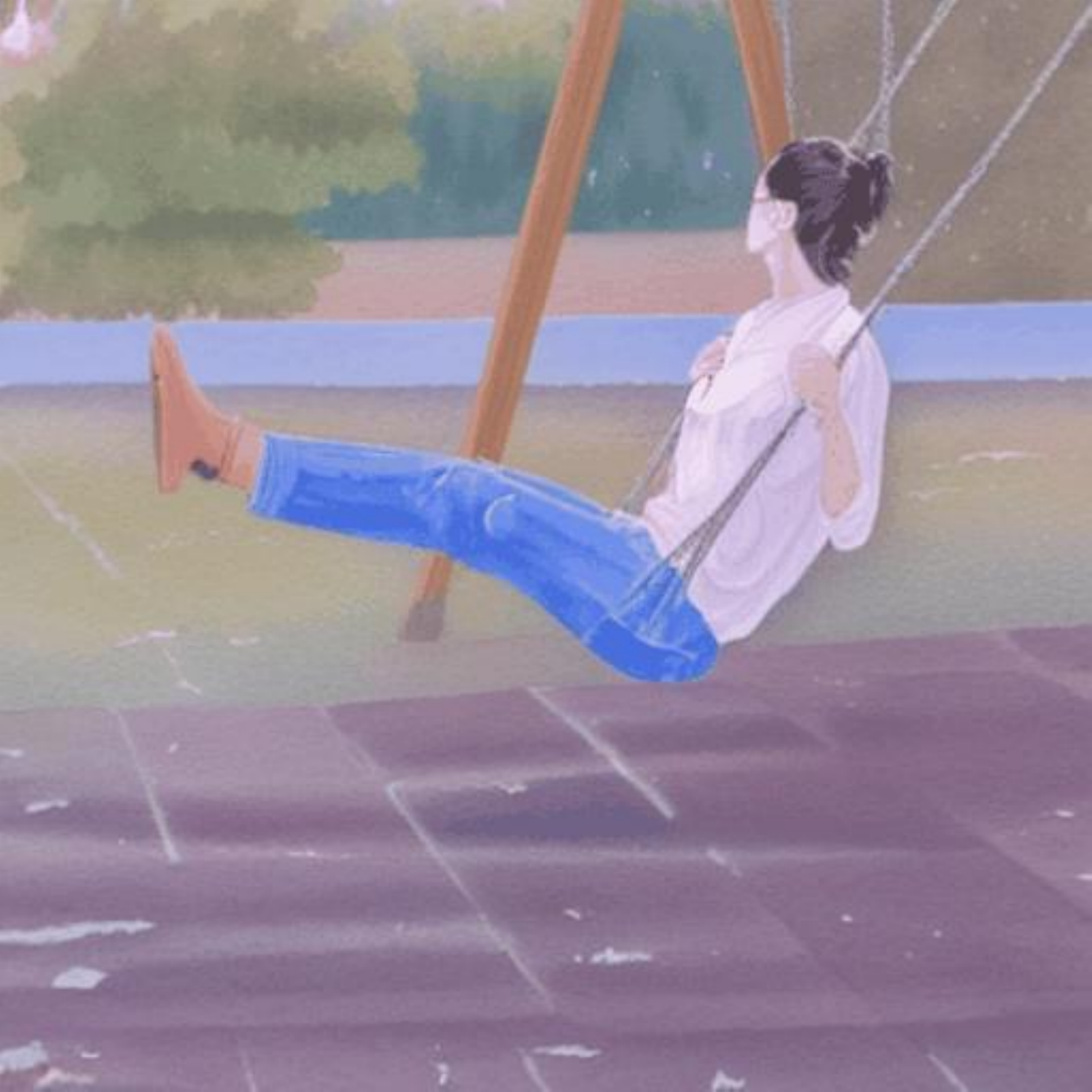}
\includegraphics[width=0.11\textwidth]{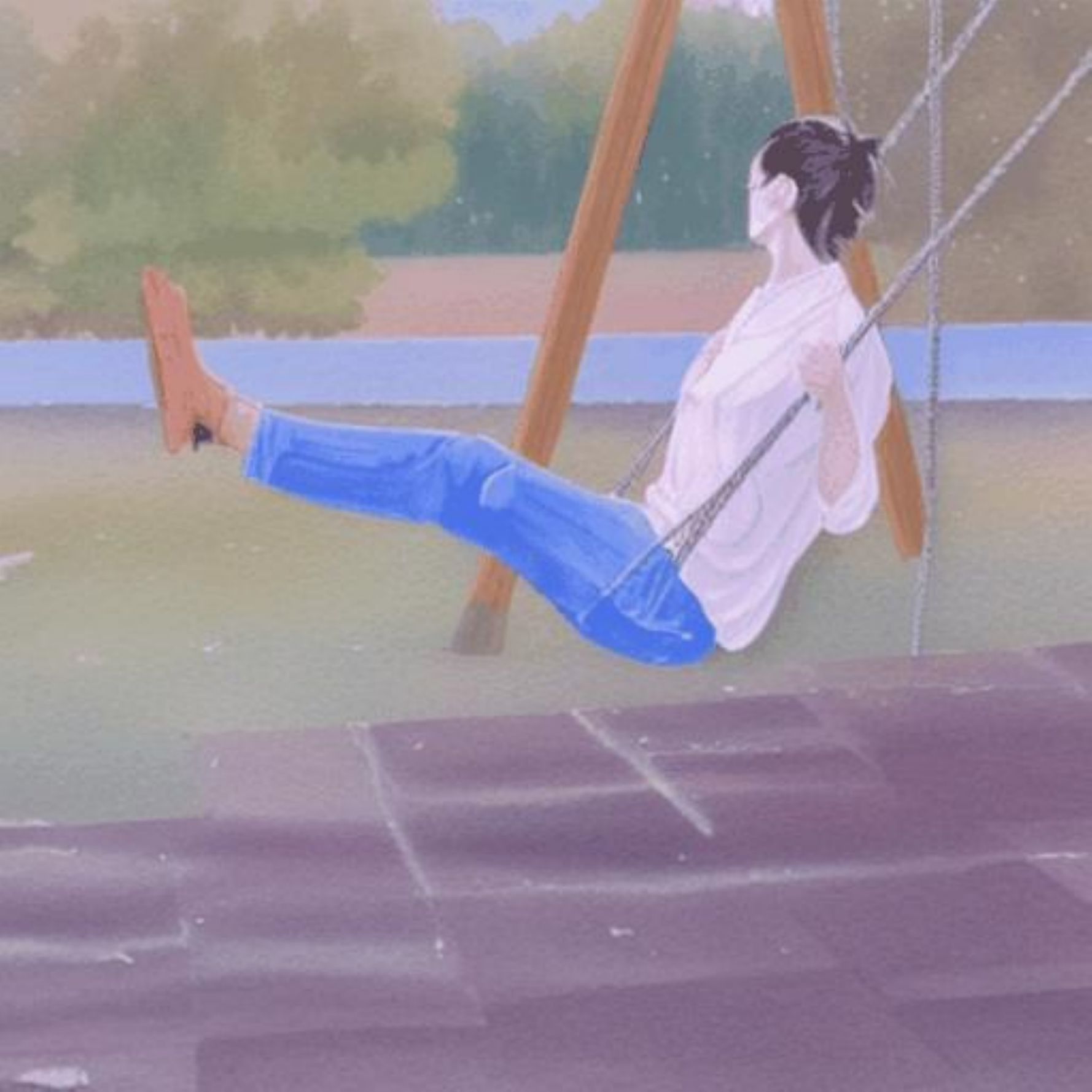}
\includegraphics[width=0.11\textwidth]{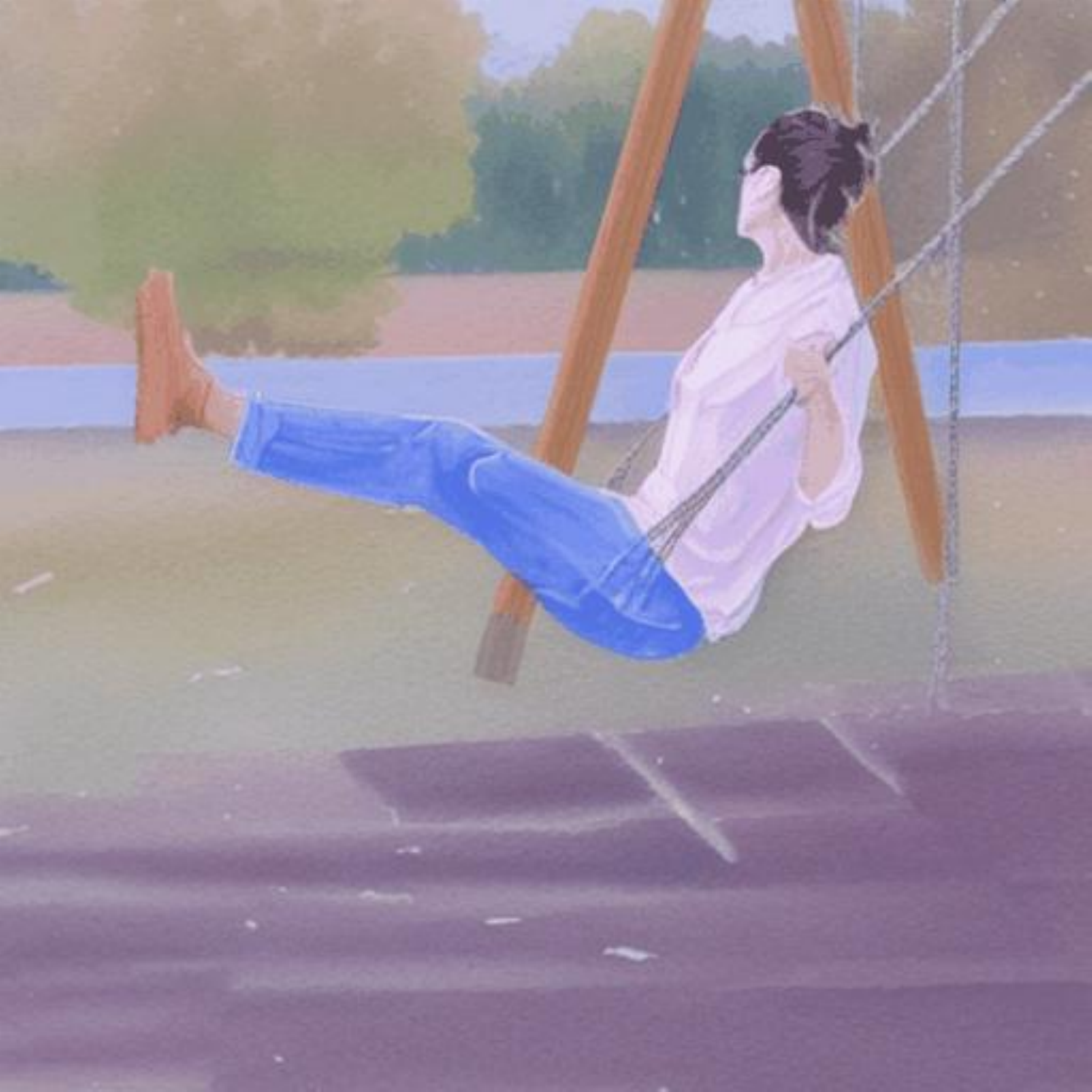}
\includegraphics[width=0.11\textwidth]{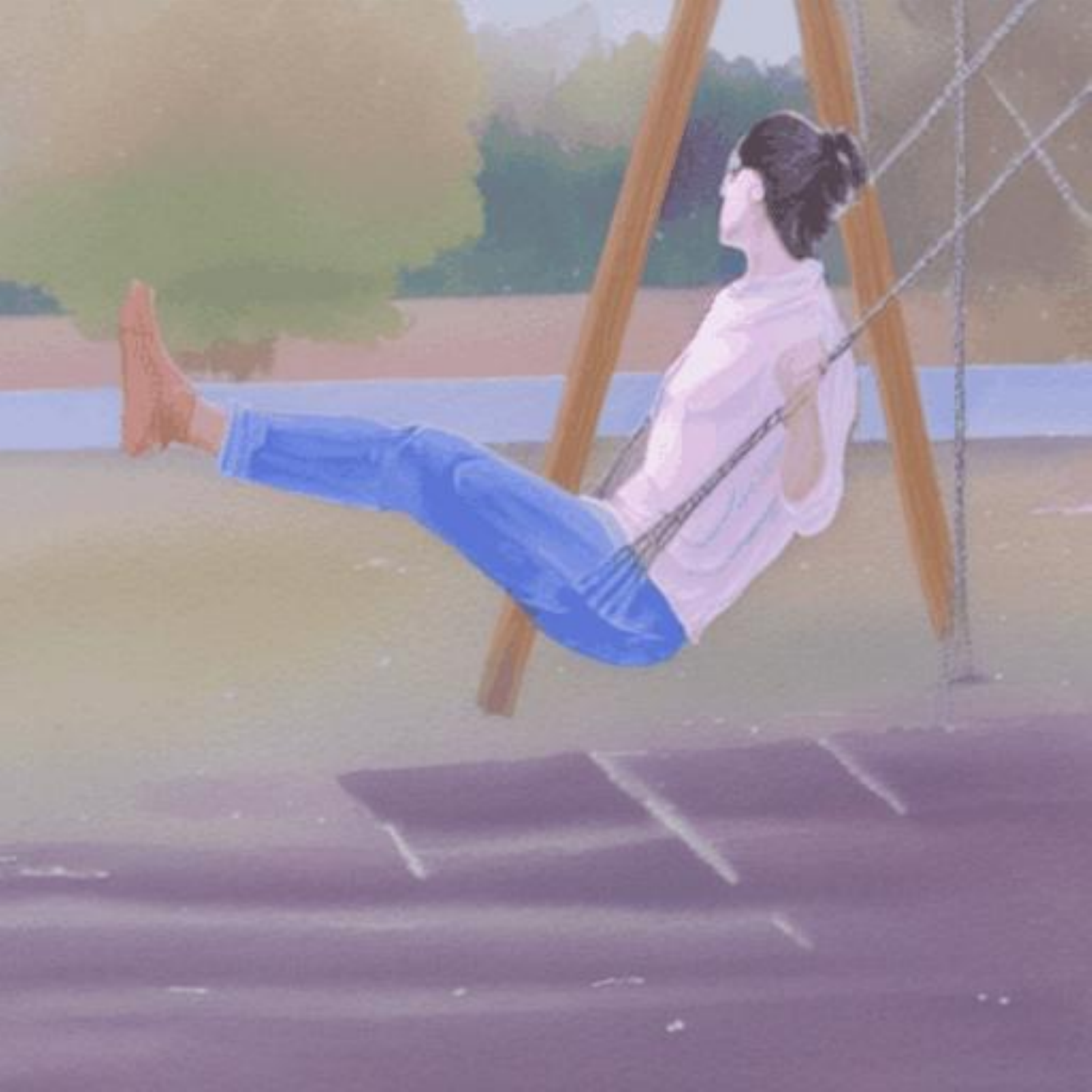}
\includegraphics[width=0.11\textwidth]{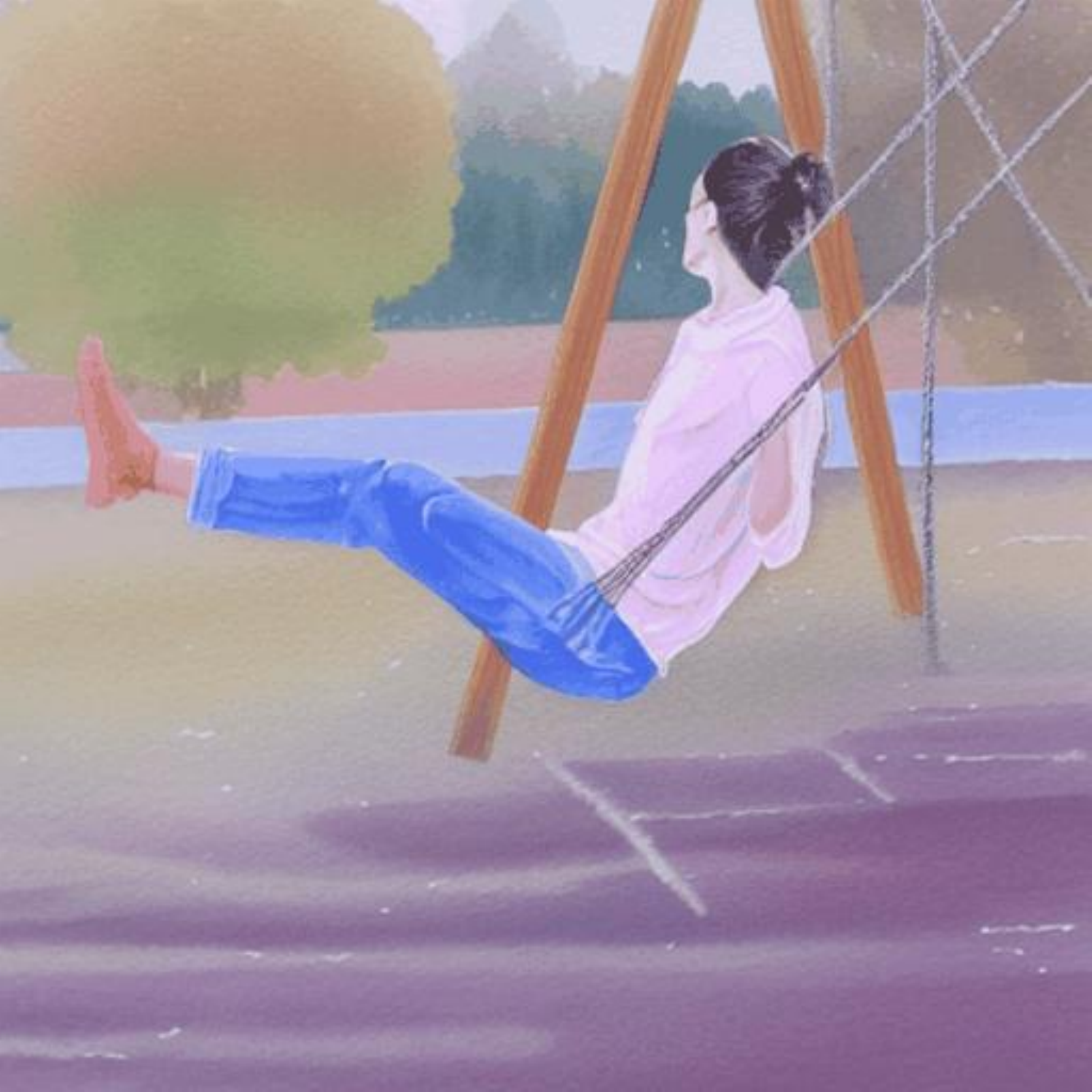}

\makebox[0.12\textwidth]{\colorbox{yellow}{\textbf{Edit-A-Video (Ours)}} A woman is on the swing, \textcolor{blue}{\textbf{Van Gogh}} style.}\\

\includegraphics[width=0.11\textwidth]{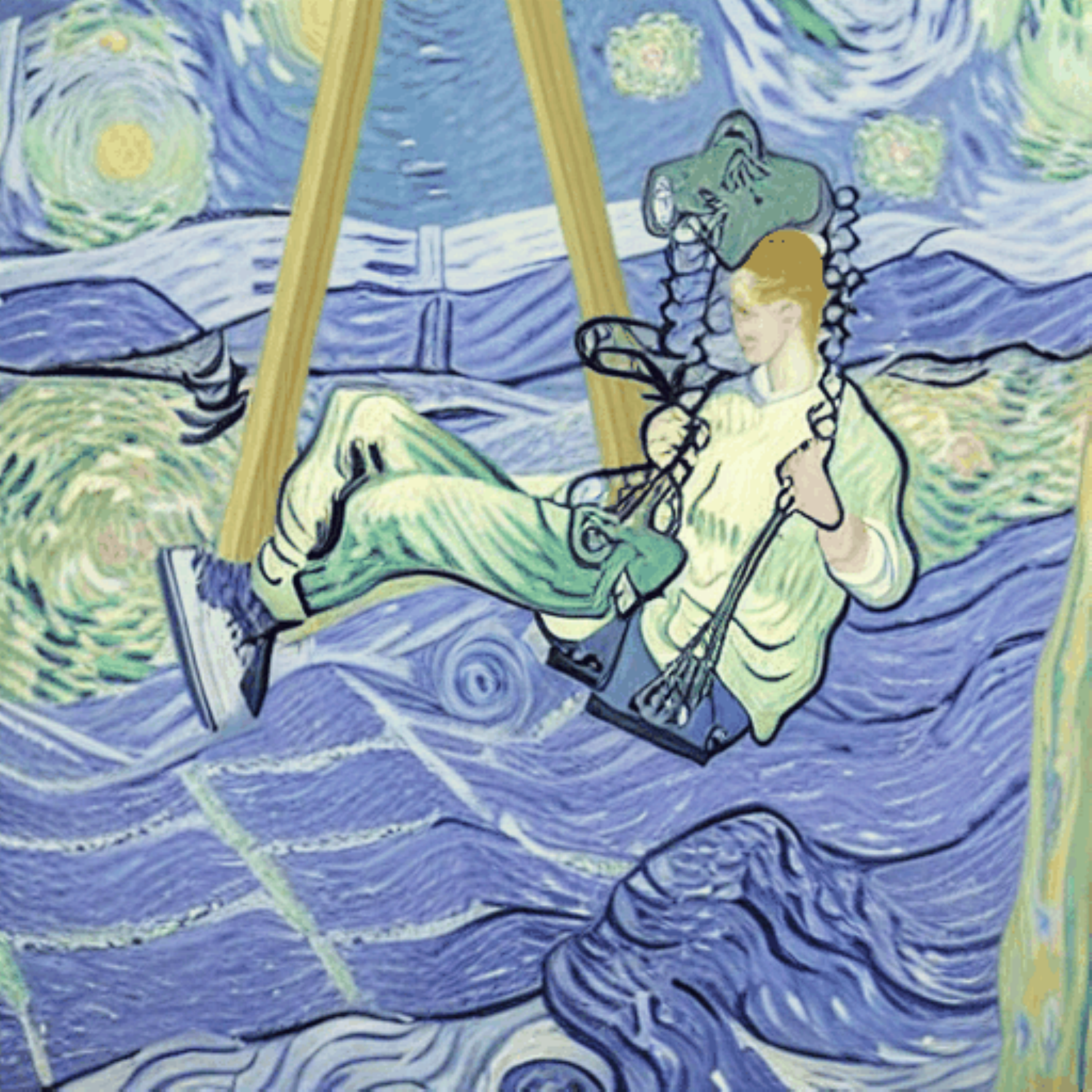}
\includegraphics[width=0.11\textwidth]{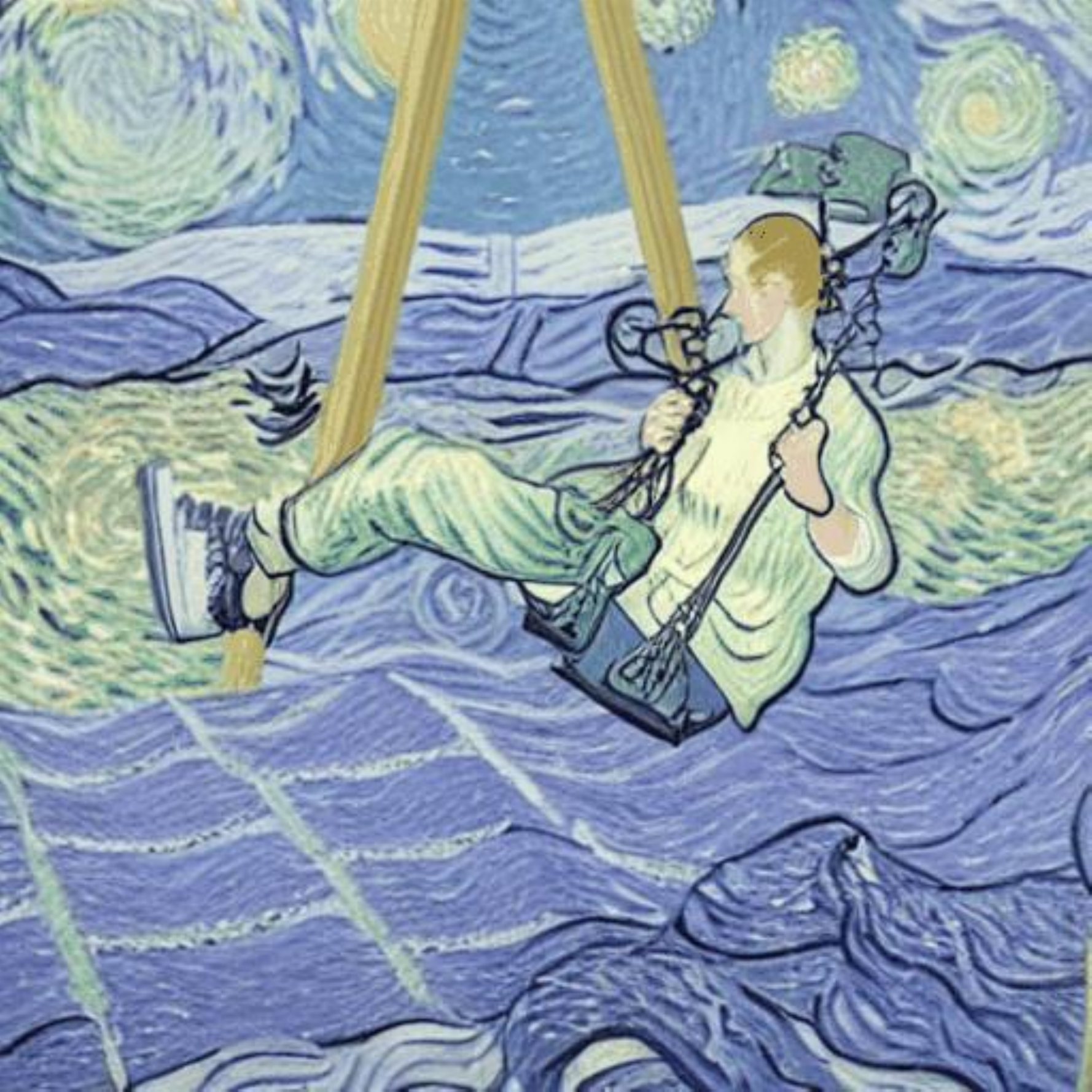}
\includegraphics[width=0.11\textwidth]{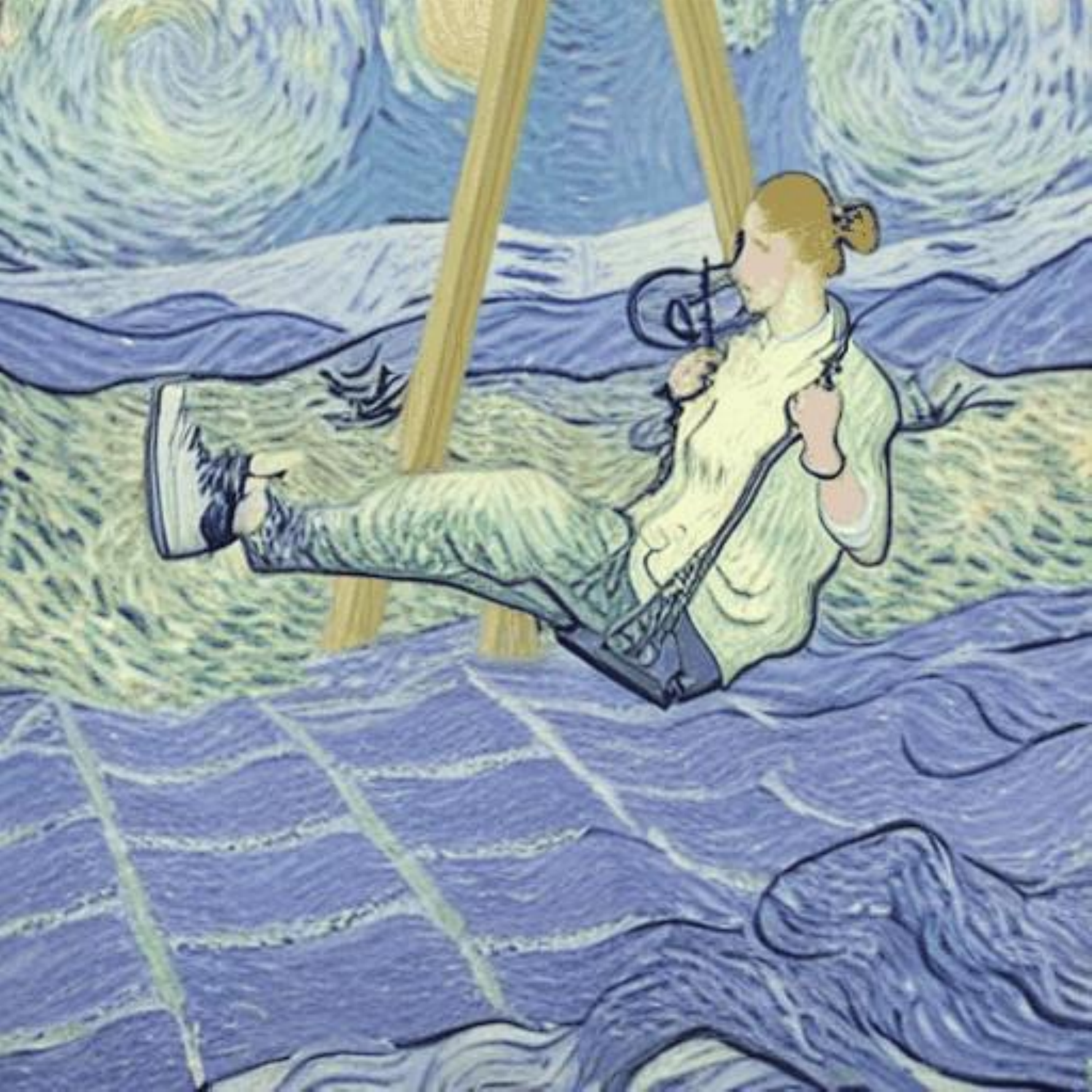}
\includegraphics[width=0.11\textwidth]{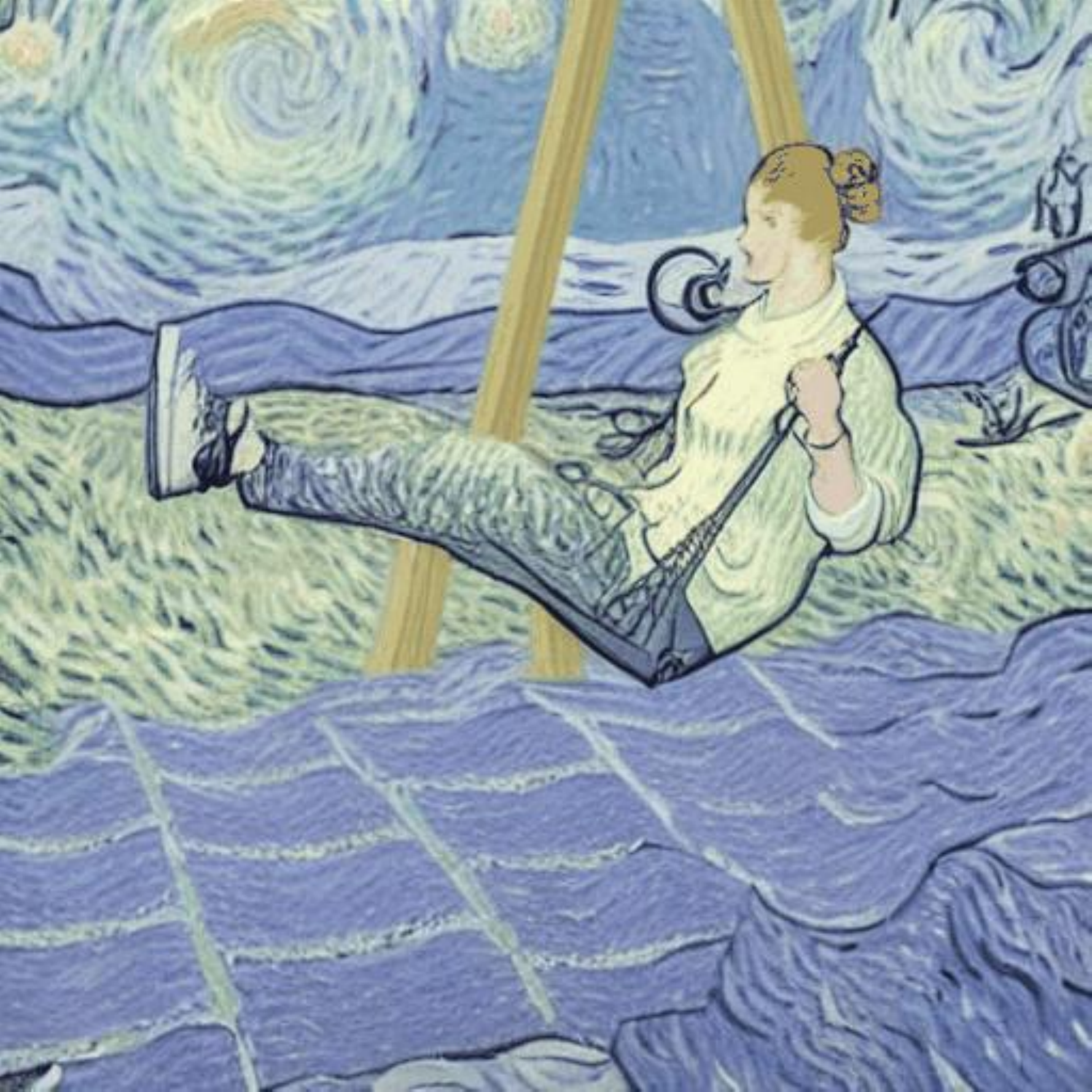}
\includegraphics[width=0.11\textwidth]{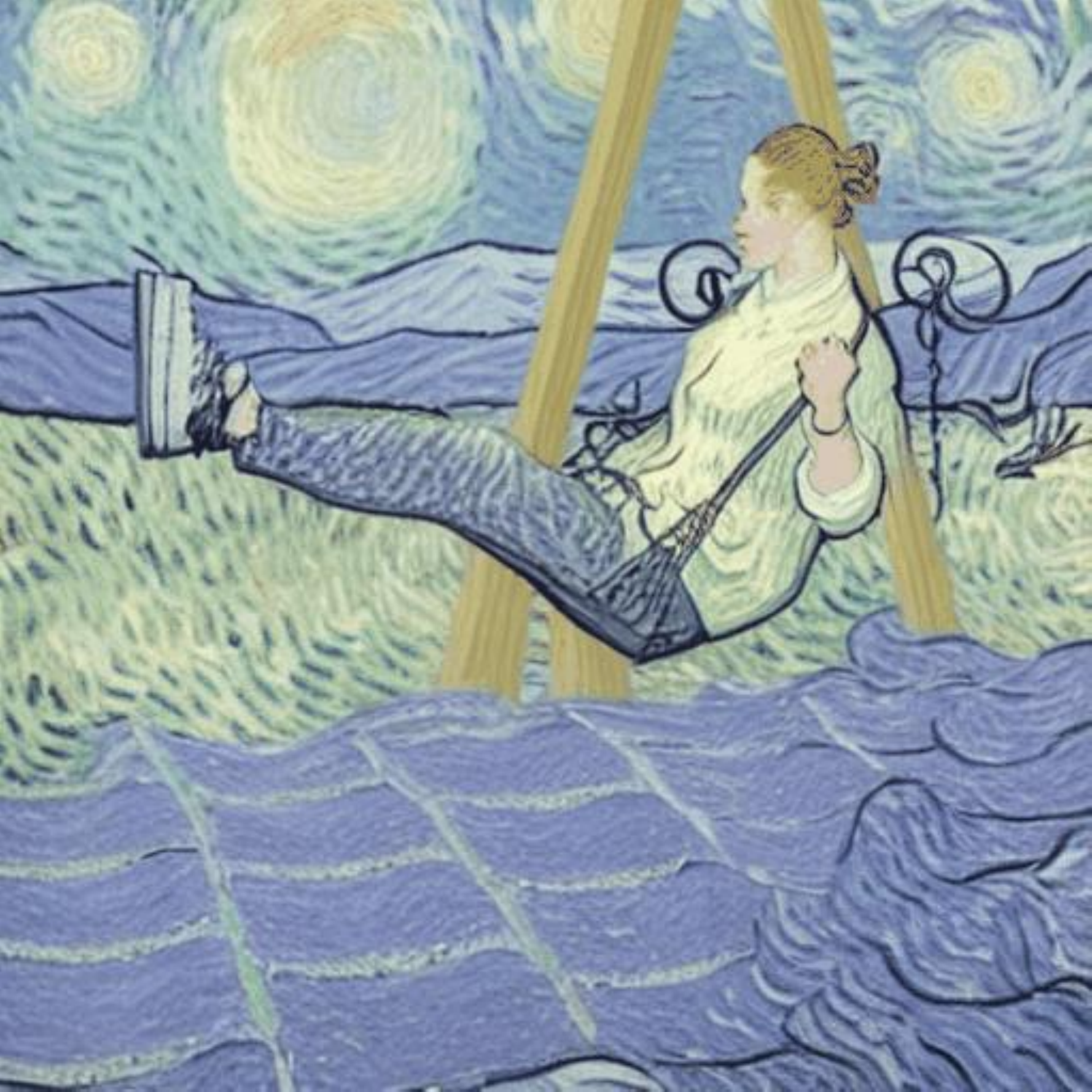}
\includegraphics[width=0.11\textwidth]{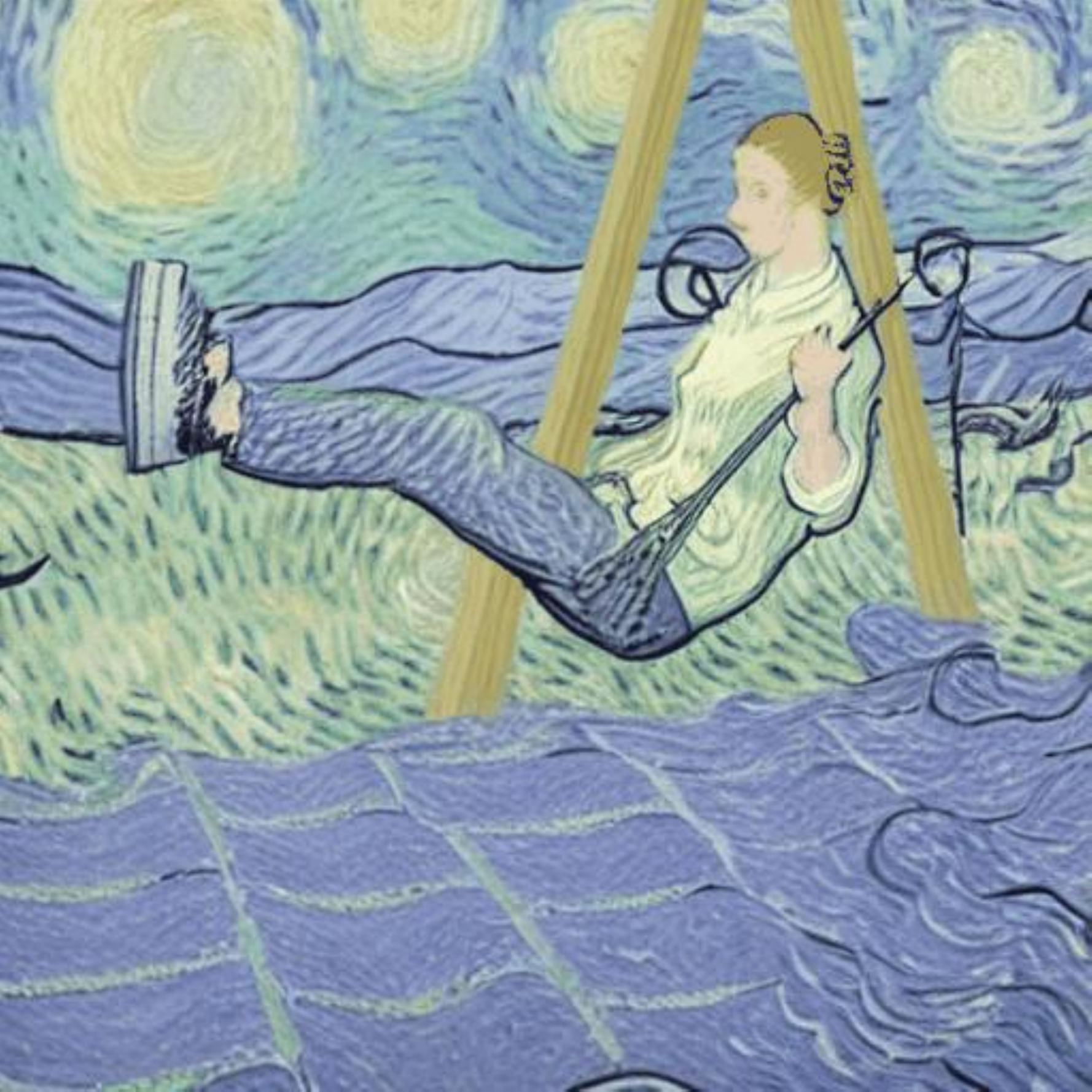}
\includegraphics[width=0.11\textwidth]{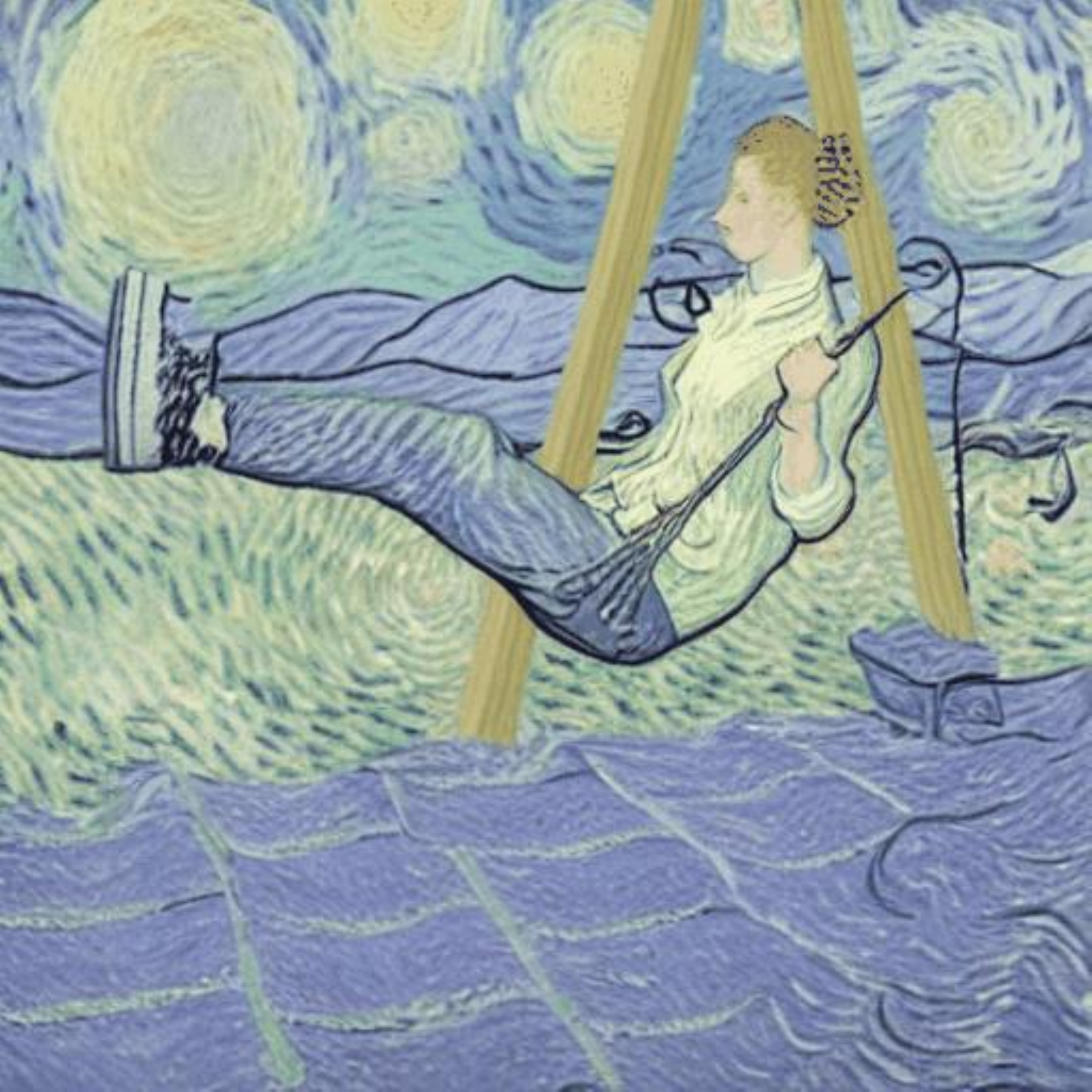}
\includegraphics[width=0.11\textwidth]{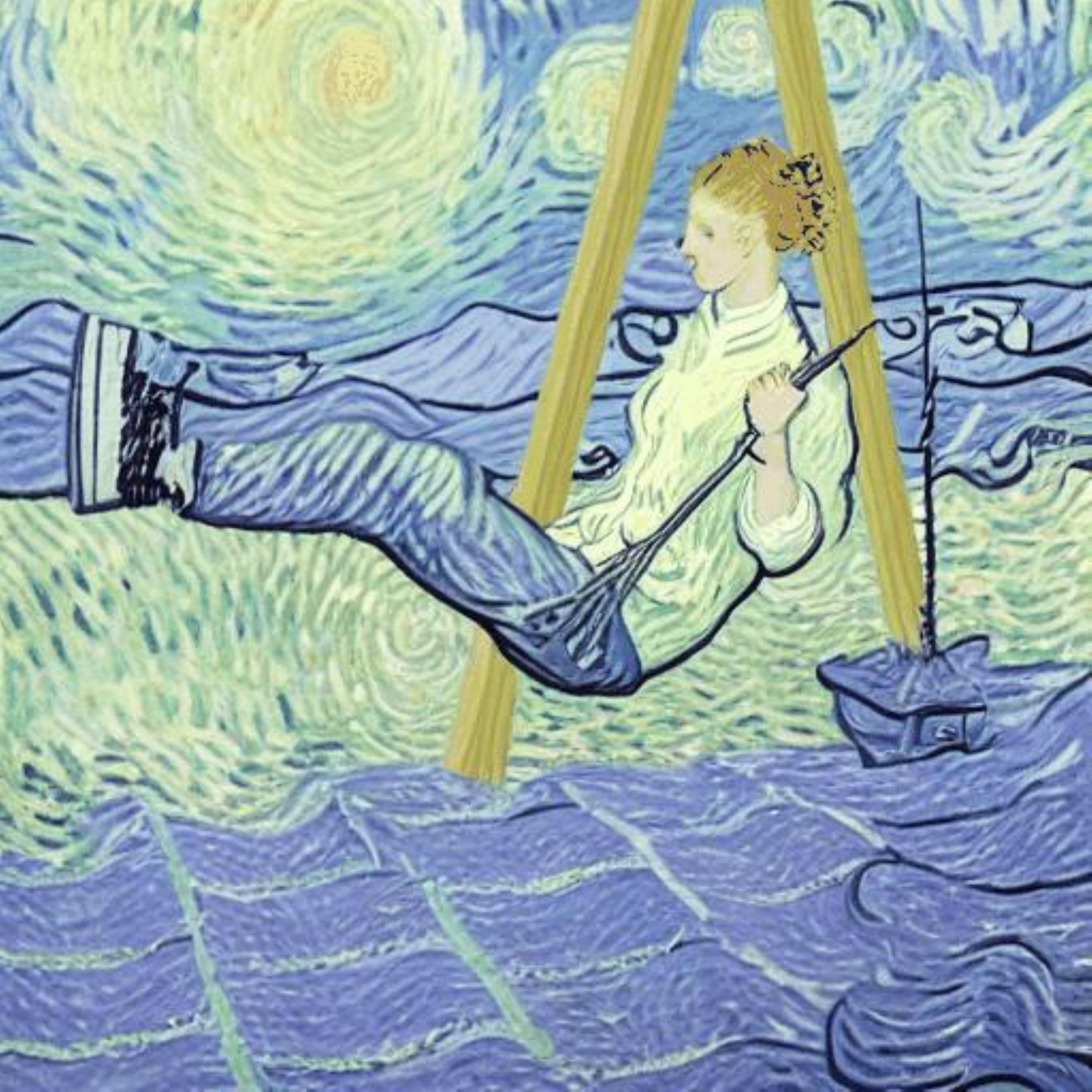}

\makebox[0.12\textwidth]{\colorbox{yellow}{\textbf{Edit-A-Video (Ours)}} A \textcolor{blue}{\textbf{man}} is on the swing}\\

\includegraphics[width=0.11\textwidth]{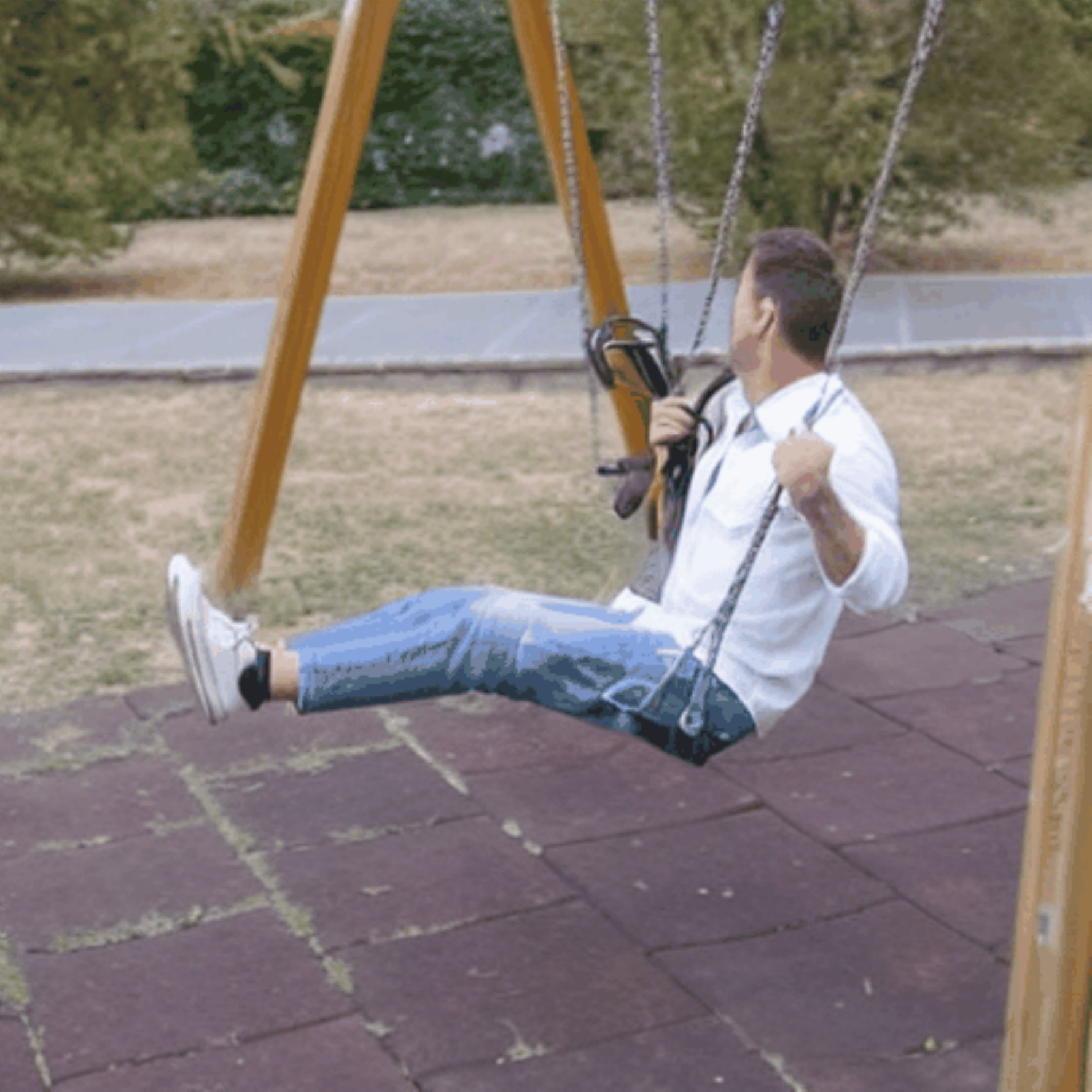}
\includegraphics[width=0.11\textwidth]{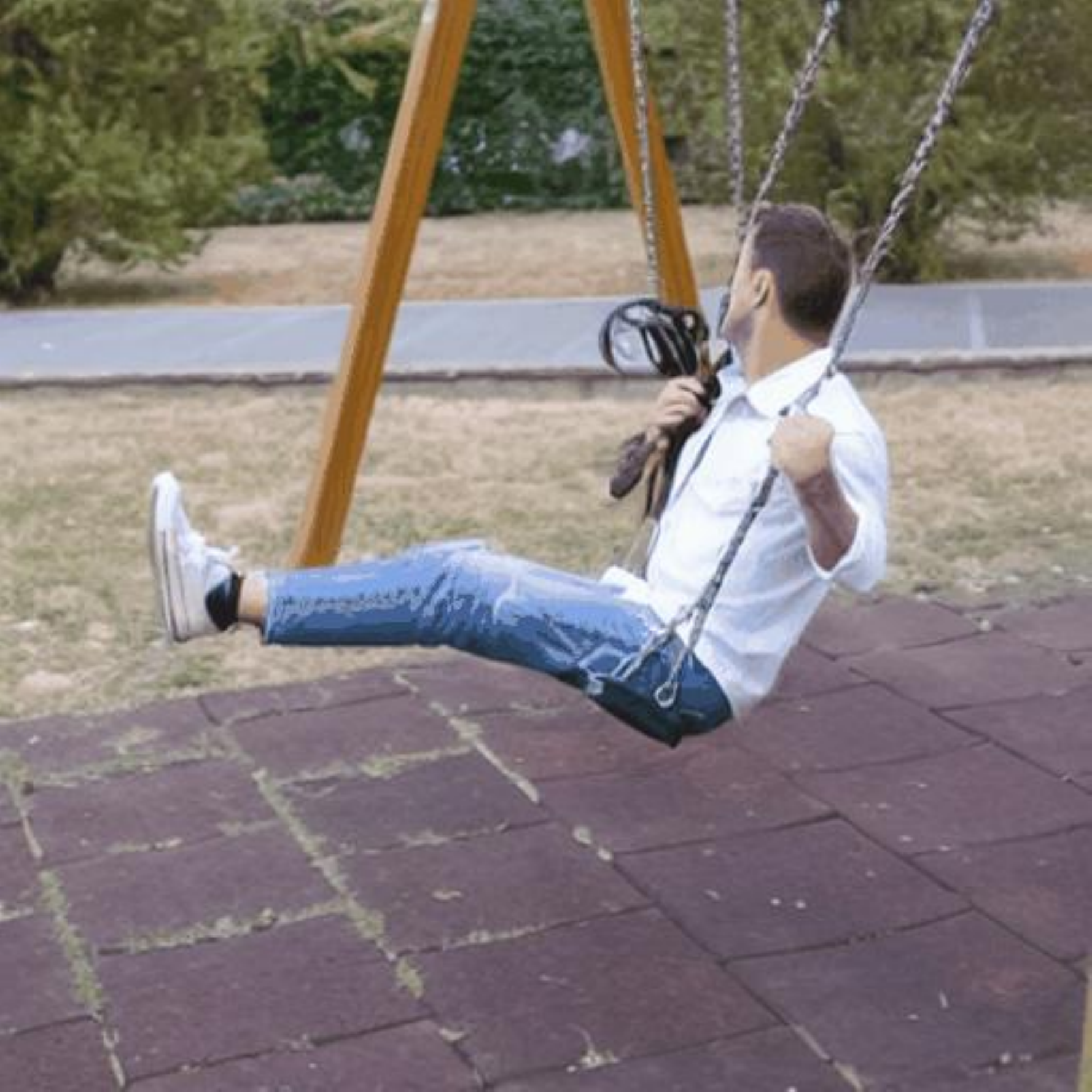}
\includegraphics[width=0.11\textwidth]{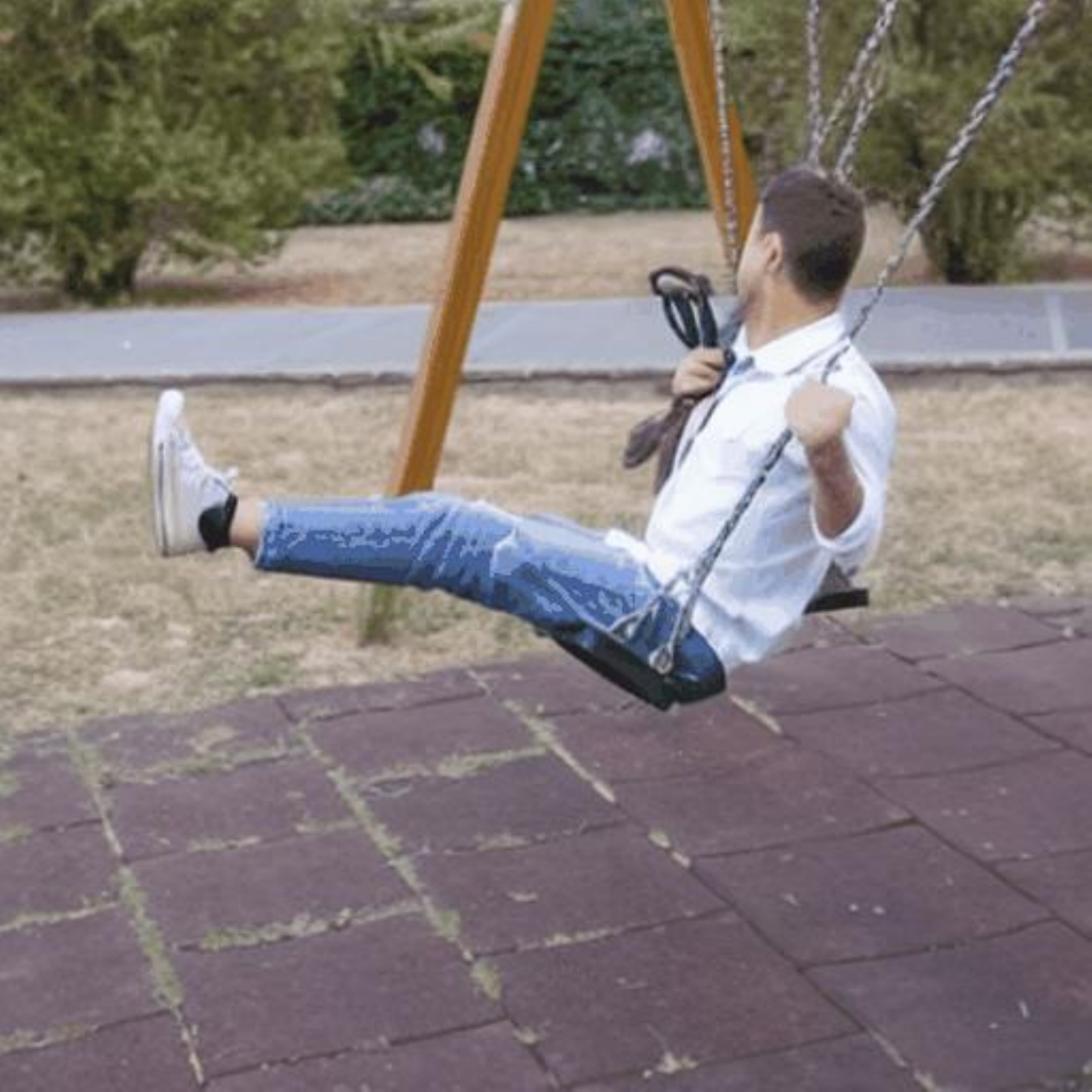}
\includegraphics[width=0.11\textwidth]{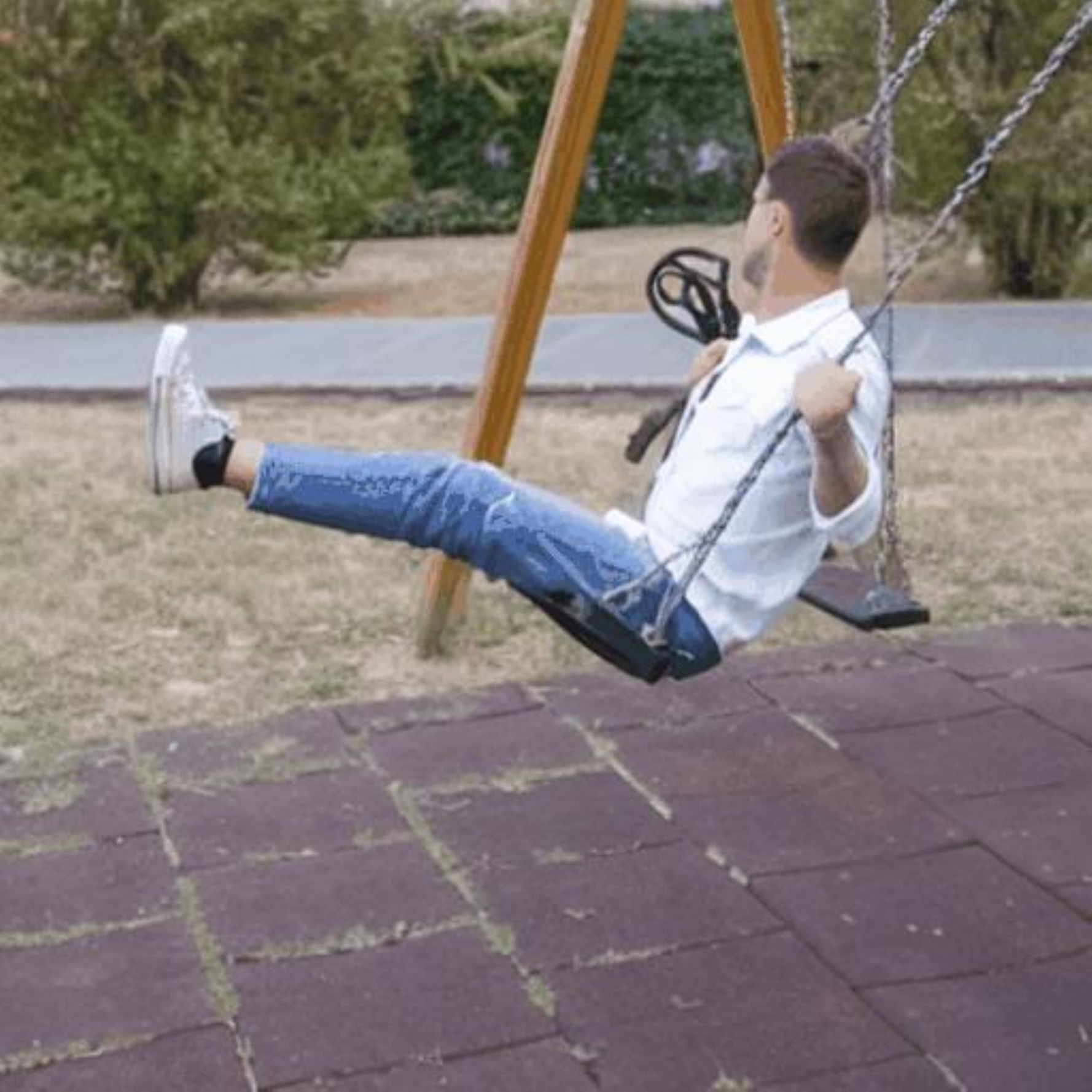}
\includegraphics[width=0.11\textwidth]{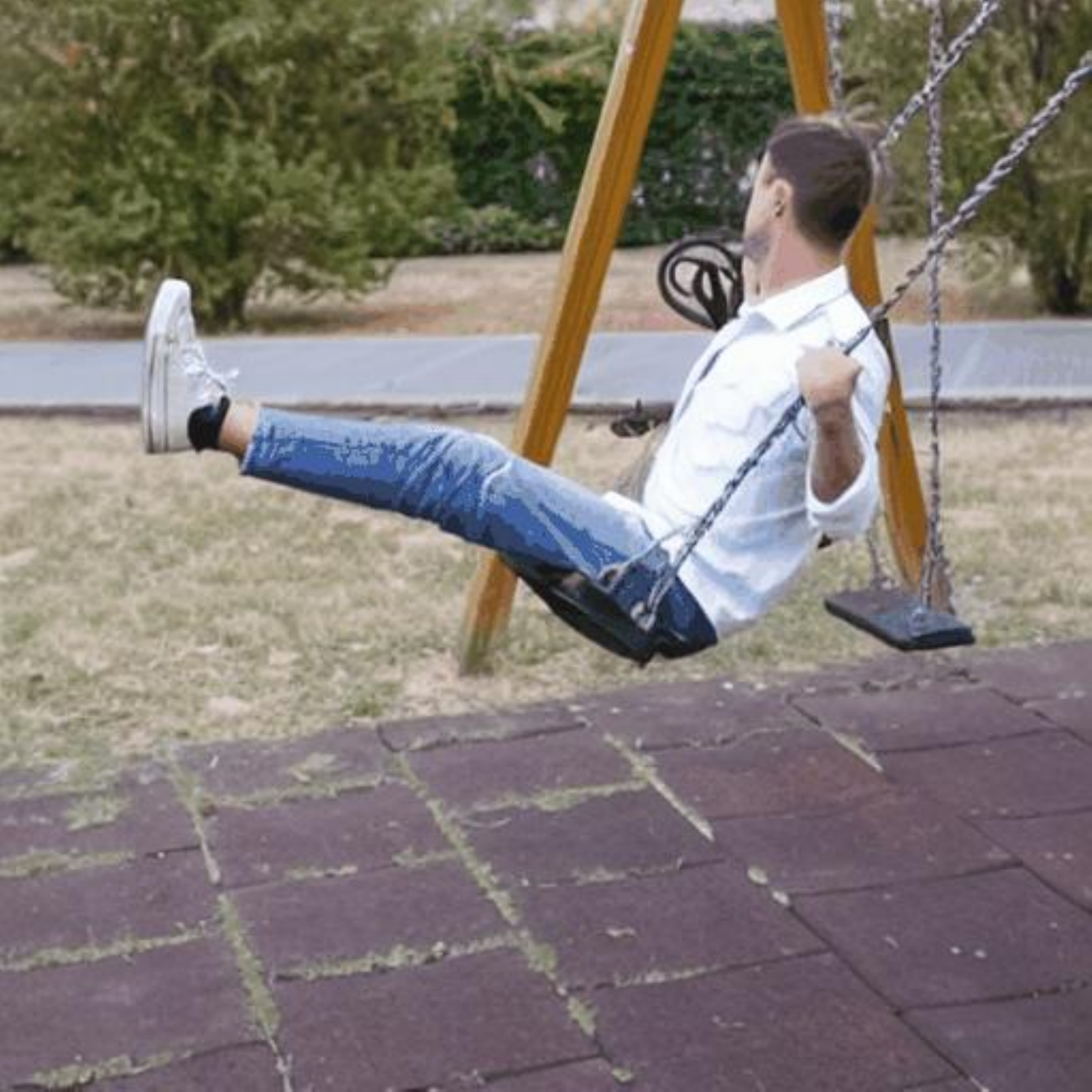}
\includegraphics[width=0.11\textwidth]{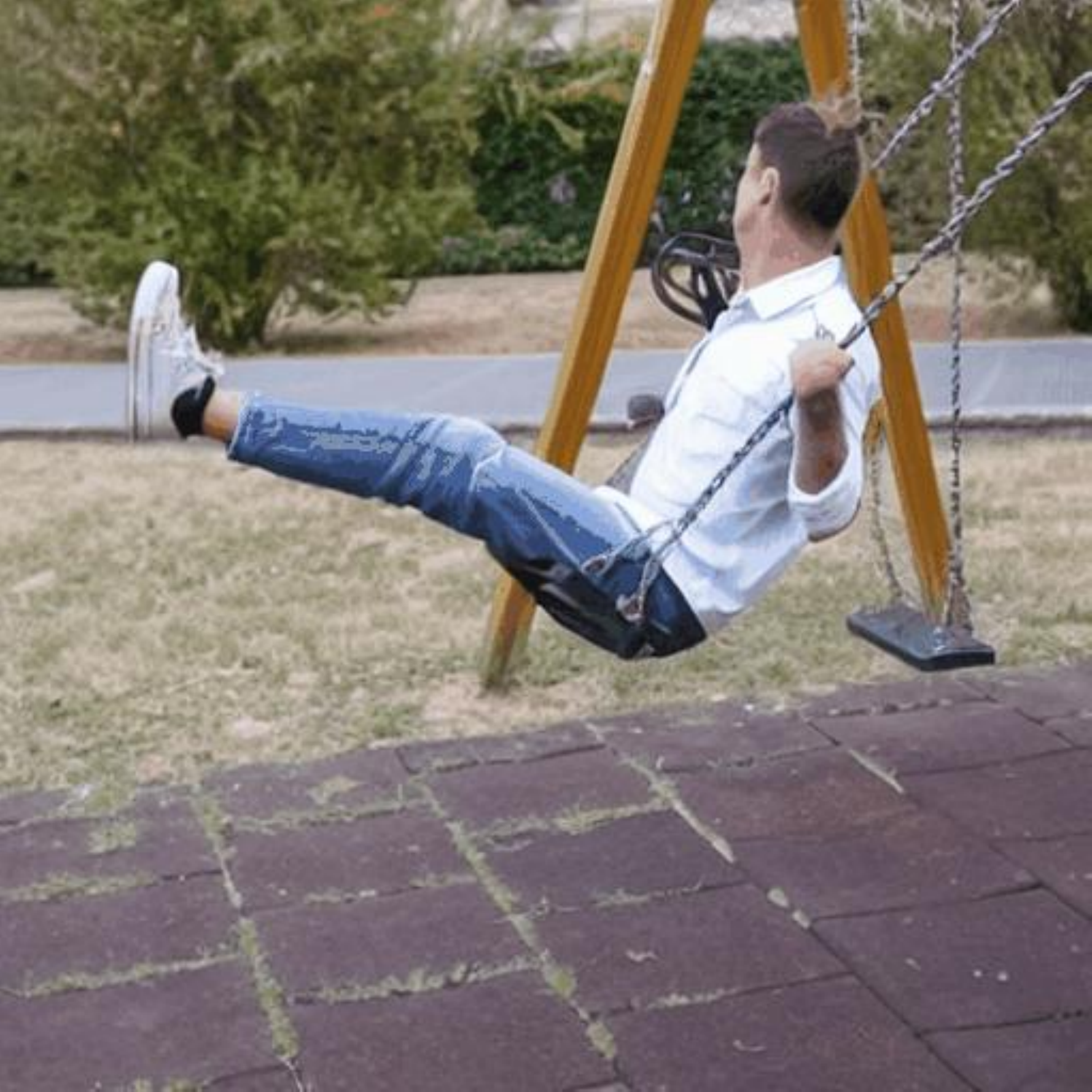}
\includegraphics[width=0.11\textwidth]{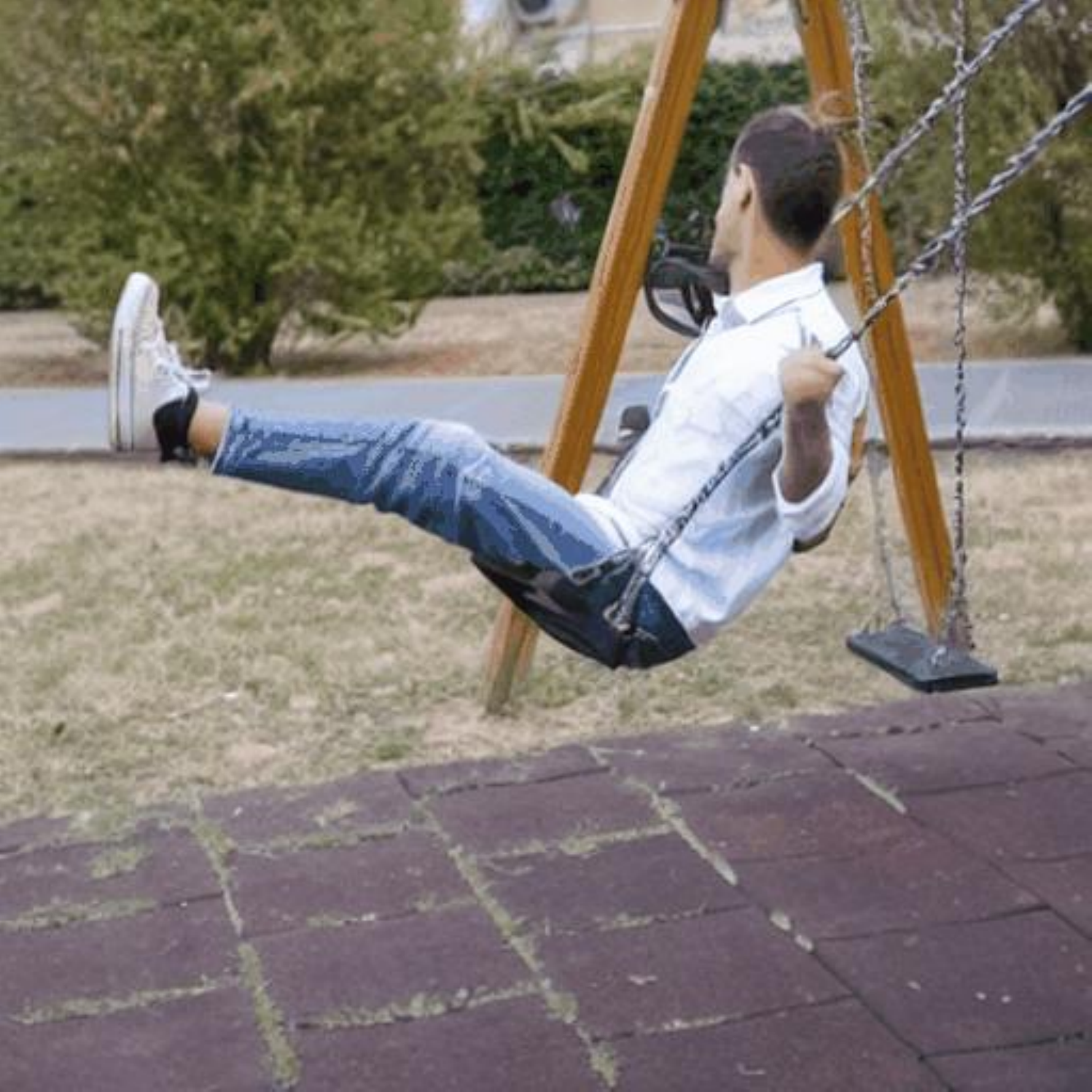}
\includegraphics[width=0.11\textwidth]{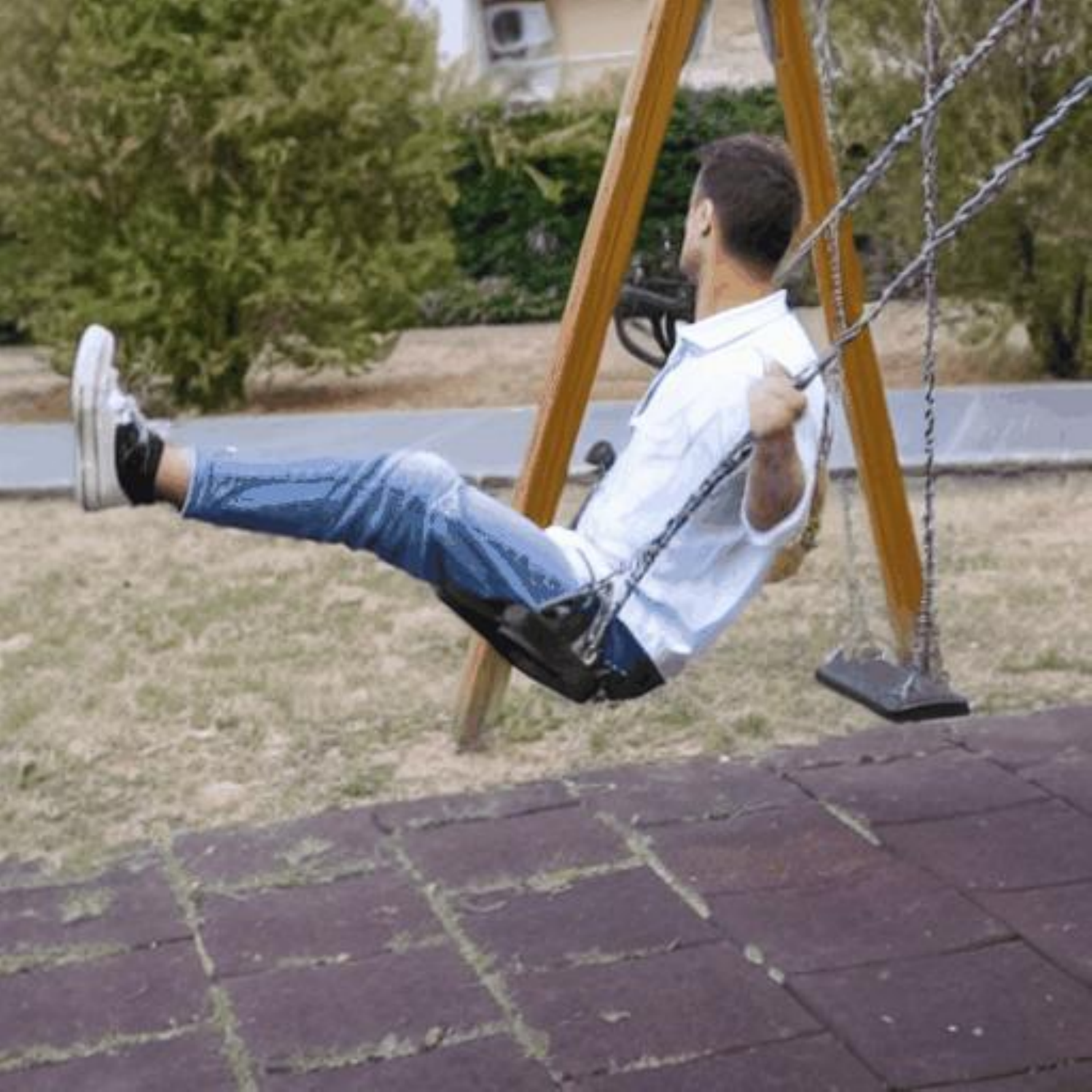}

\makebox[0.12\textwidth]{\colorbox{yellow}{\textbf{Edit-A-Video (Ours)}} A \textcolor{blue}{\textbf{Iron Man}} is on the swing}\\

\includegraphics[width=0.11\textwidth]{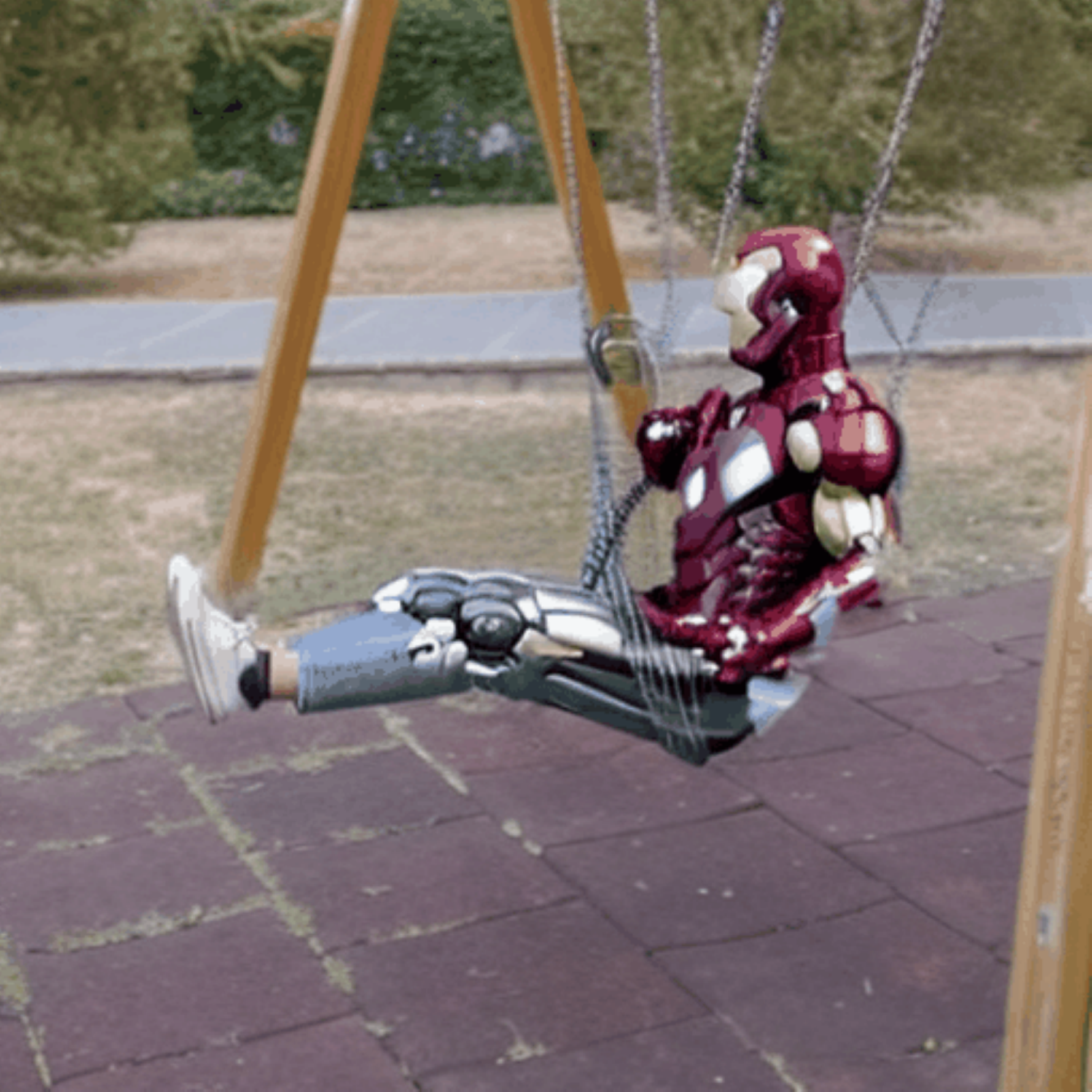}
\includegraphics[width=0.11\textwidth]{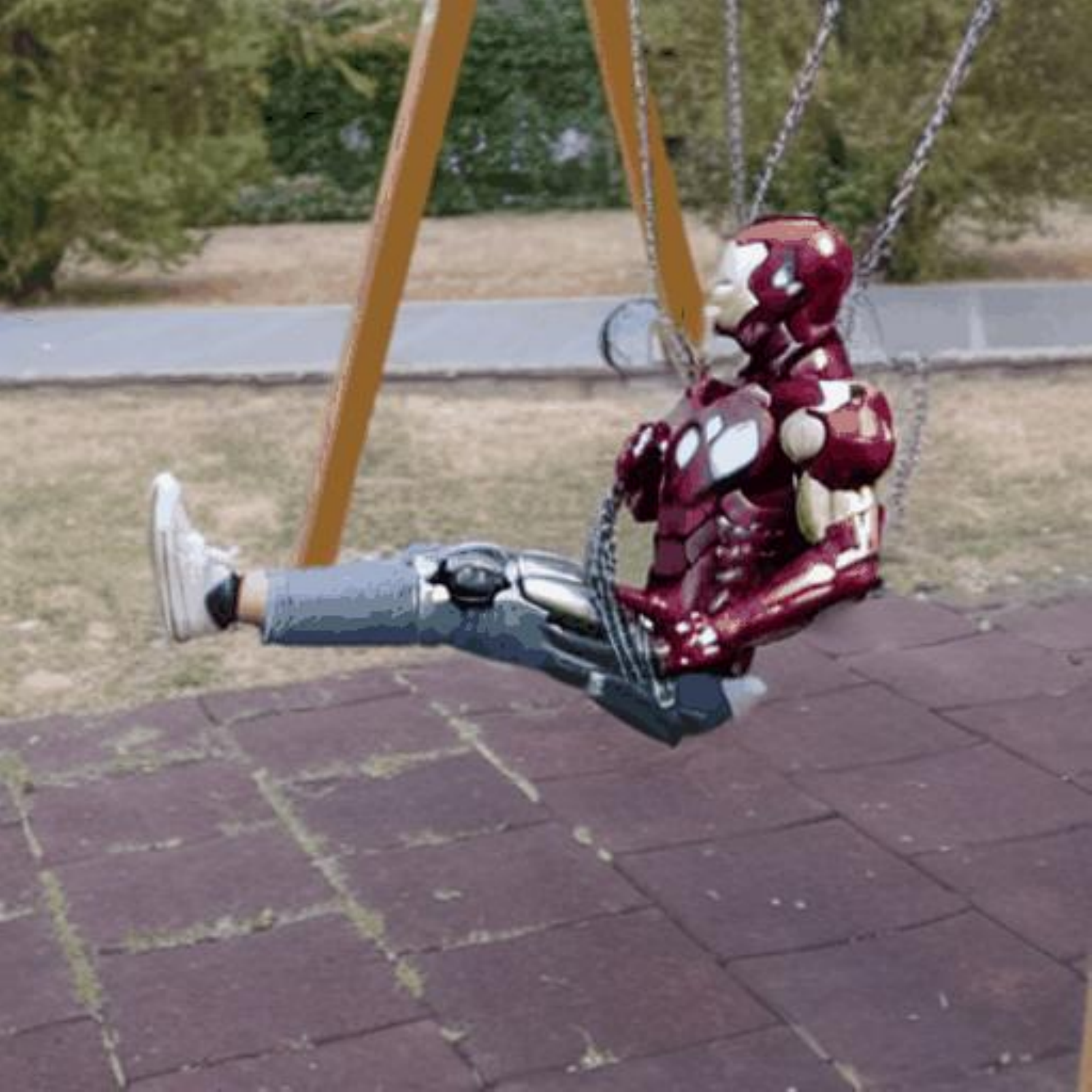}
\includegraphics[width=0.11\textwidth]{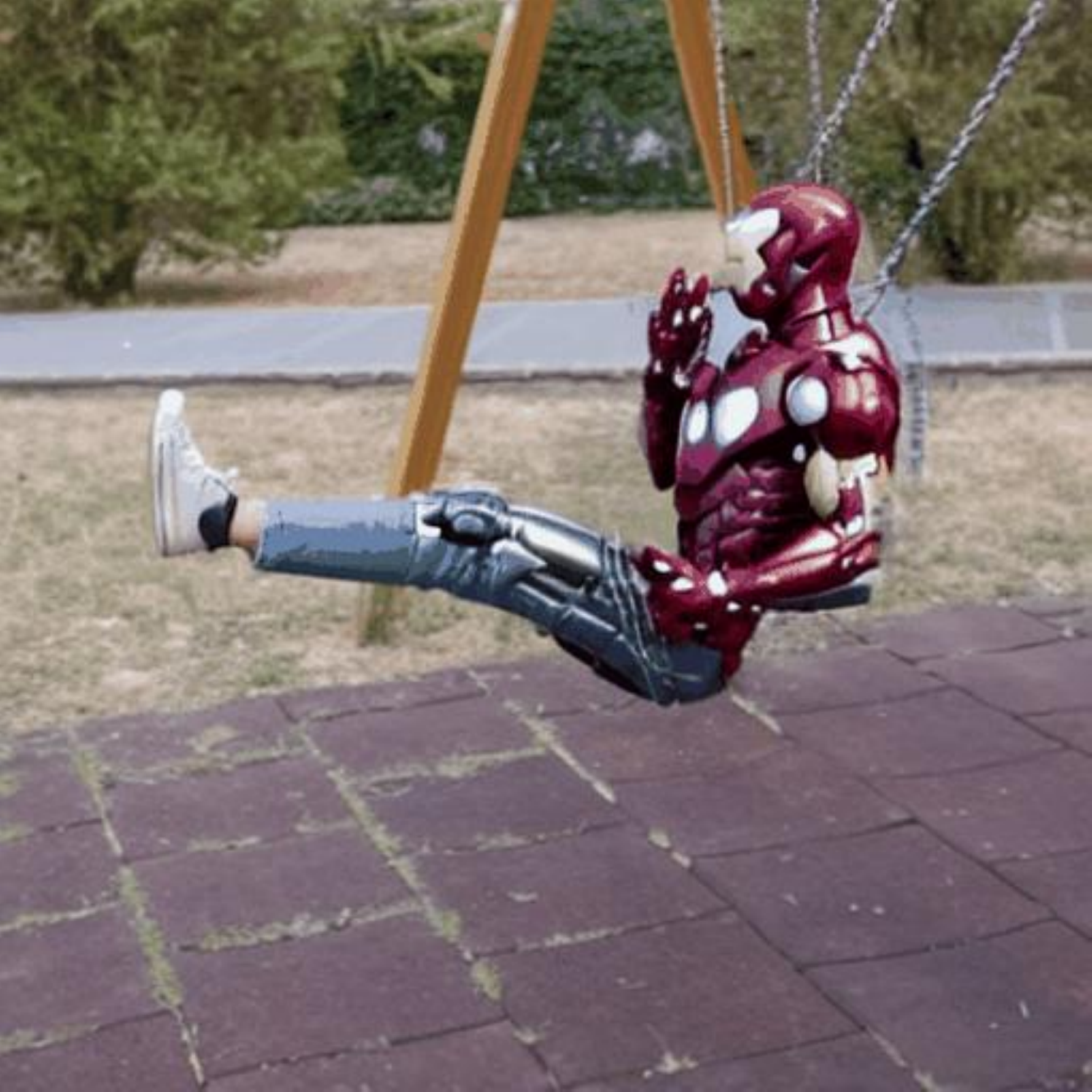}
\includegraphics[width=0.11\textwidth]{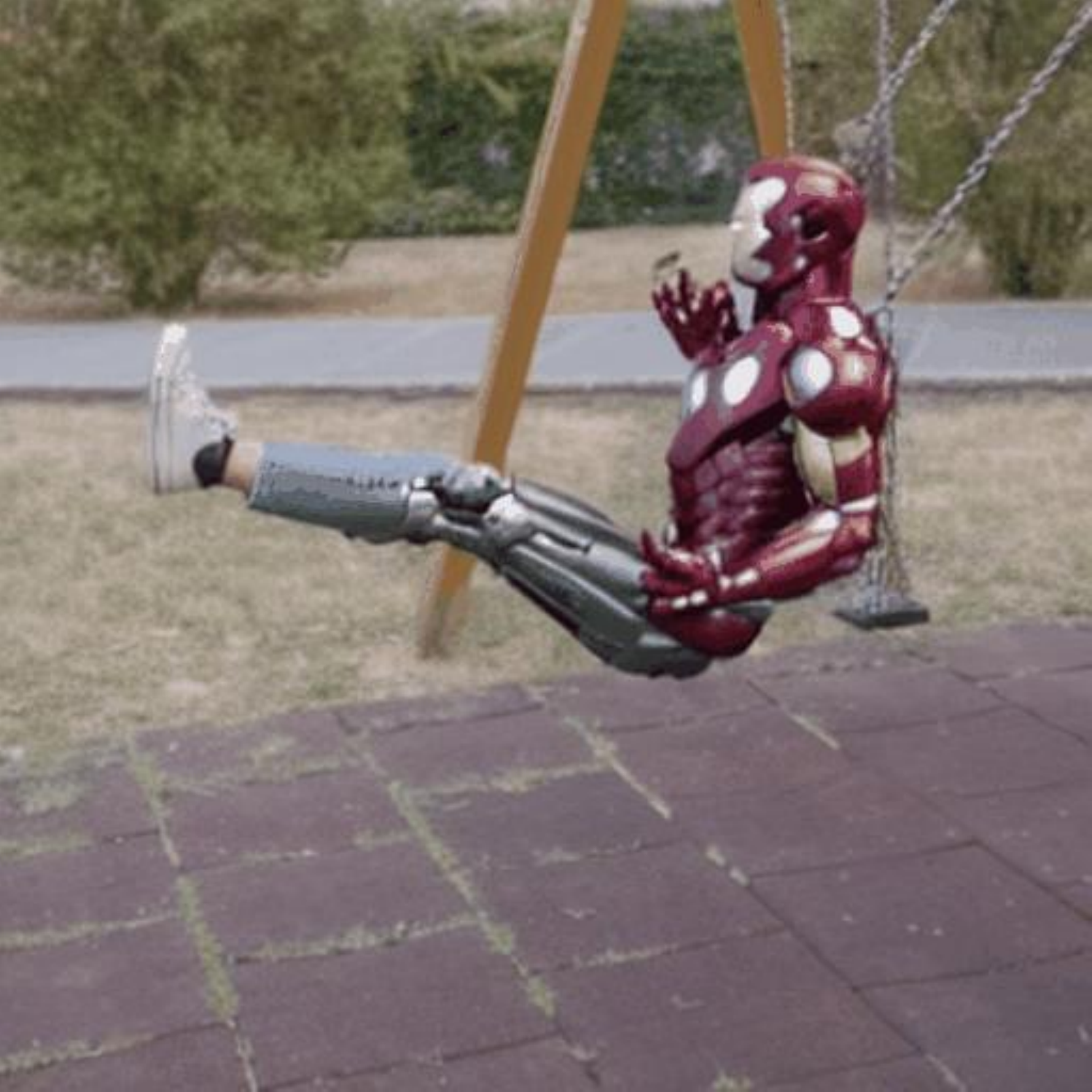}
\includegraphics[width=0.11\textwidth]{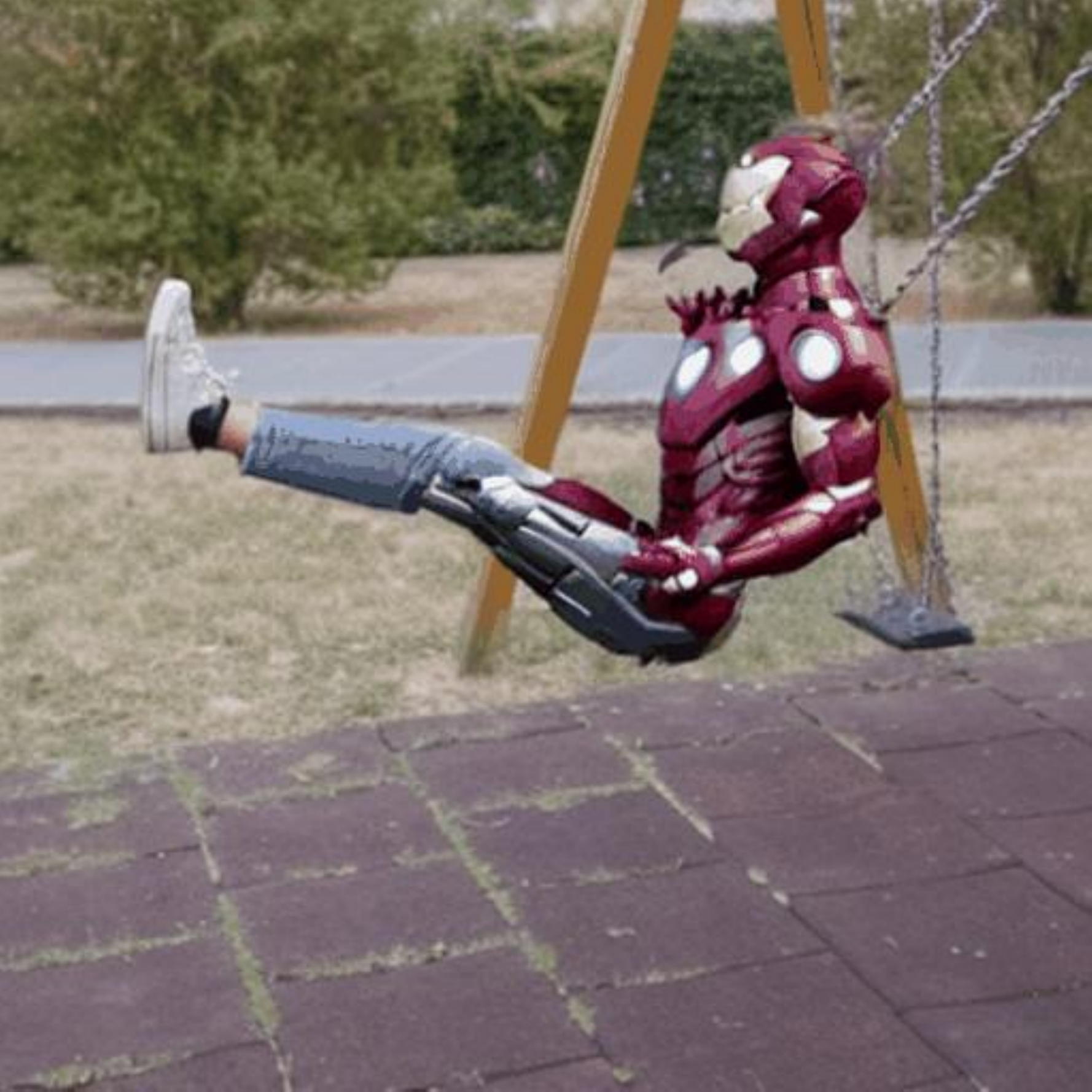}
\includegraphics[width=0.11\textwidth]{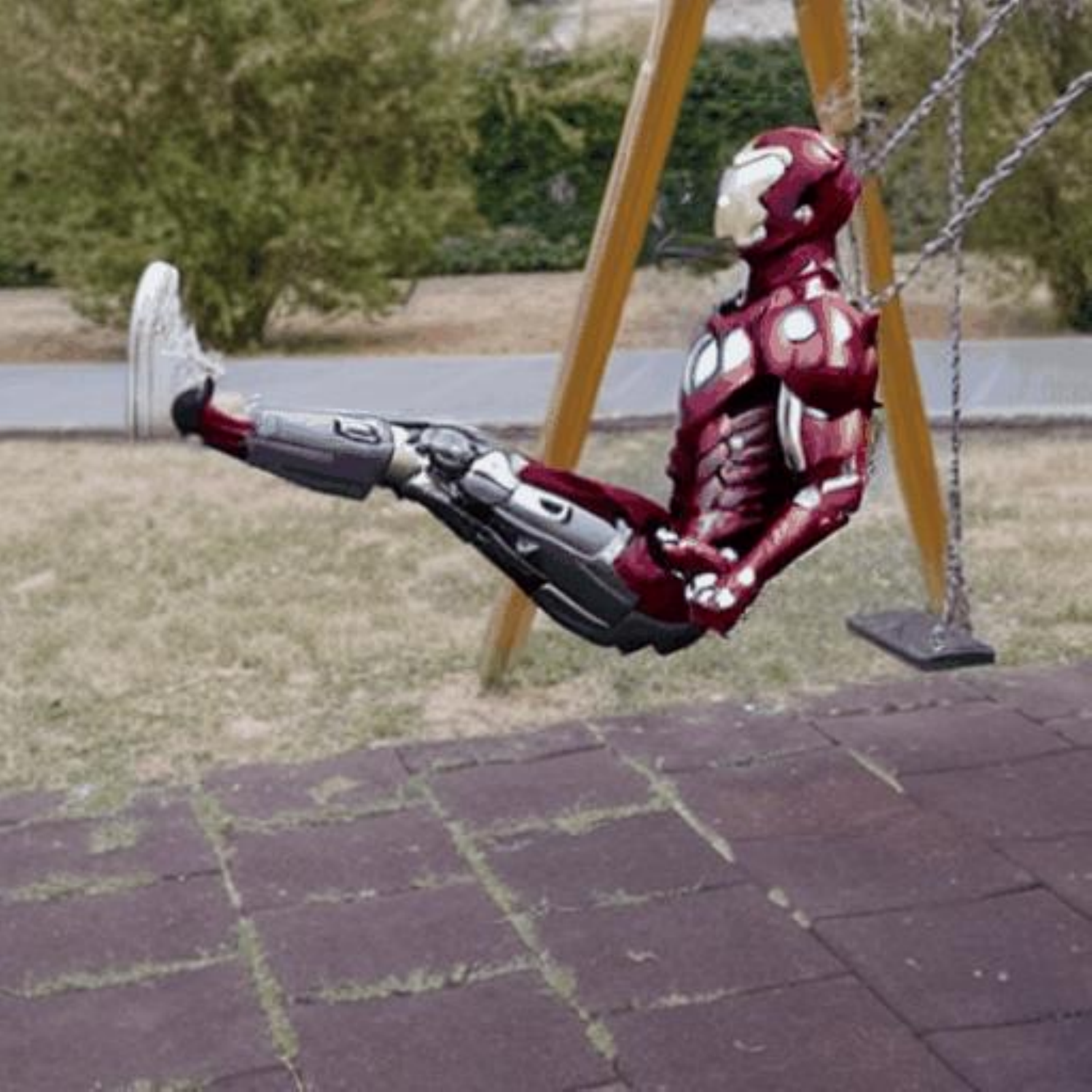}
\includegraphics[width=0.11\textwidth]{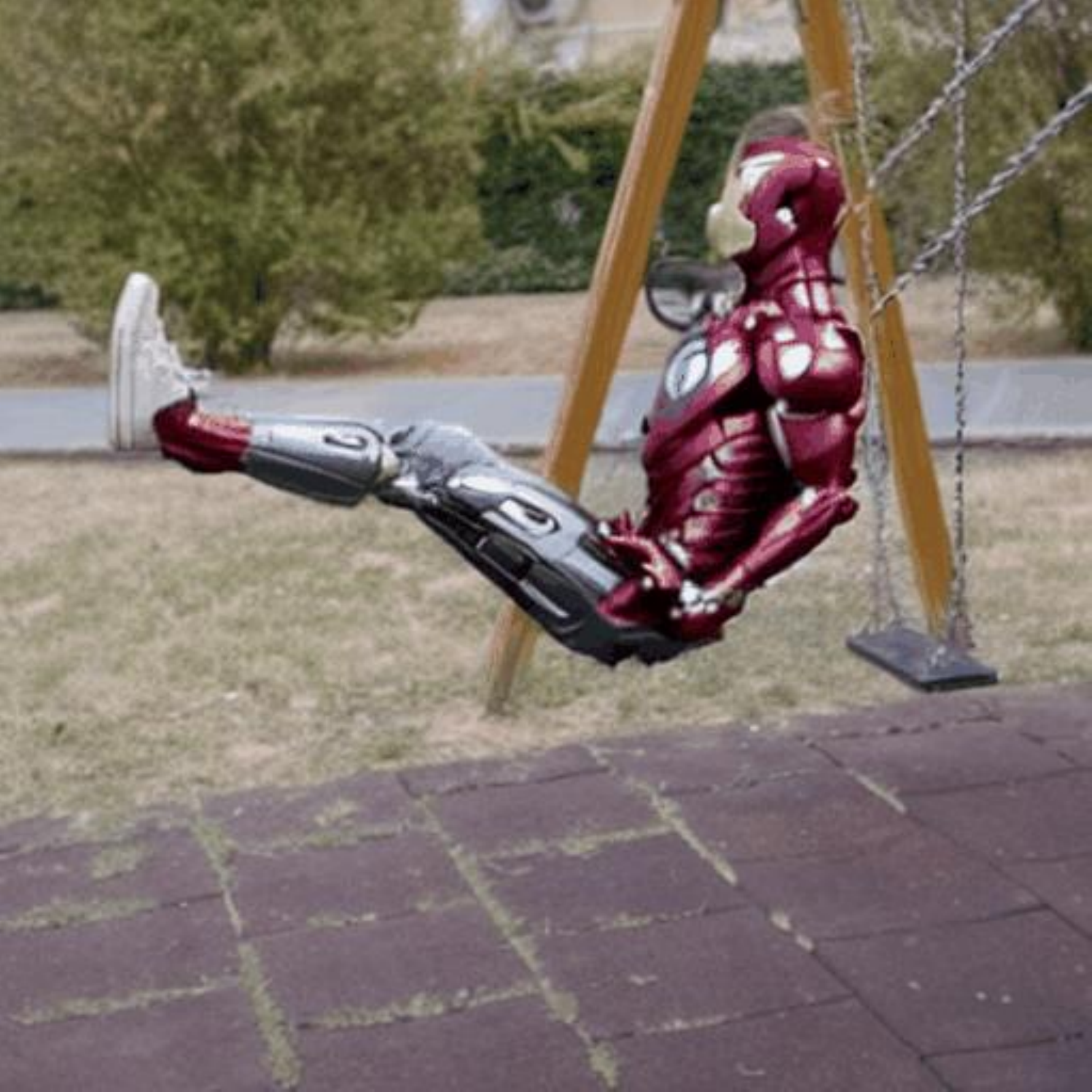}
\includegraphics[width=0.11\textwidth]{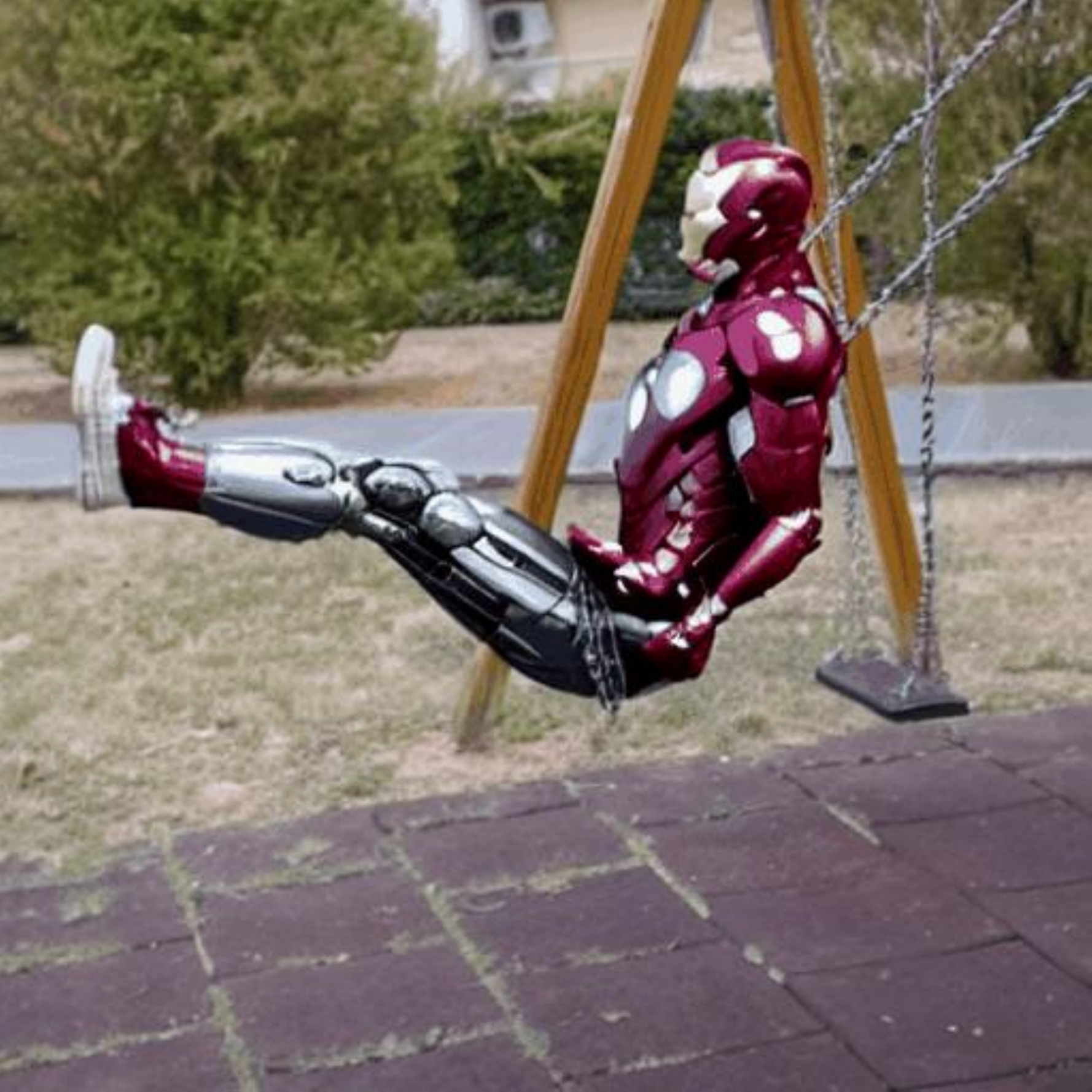}

\makebox[0.12\textwidth]{\colorbox{green}{\textbf{Tune-A-Video}} A \textcolor{blue}{\textbf{Iron Man}} is on the swing}\\

\includegraphics[width=0.11\textwidth]{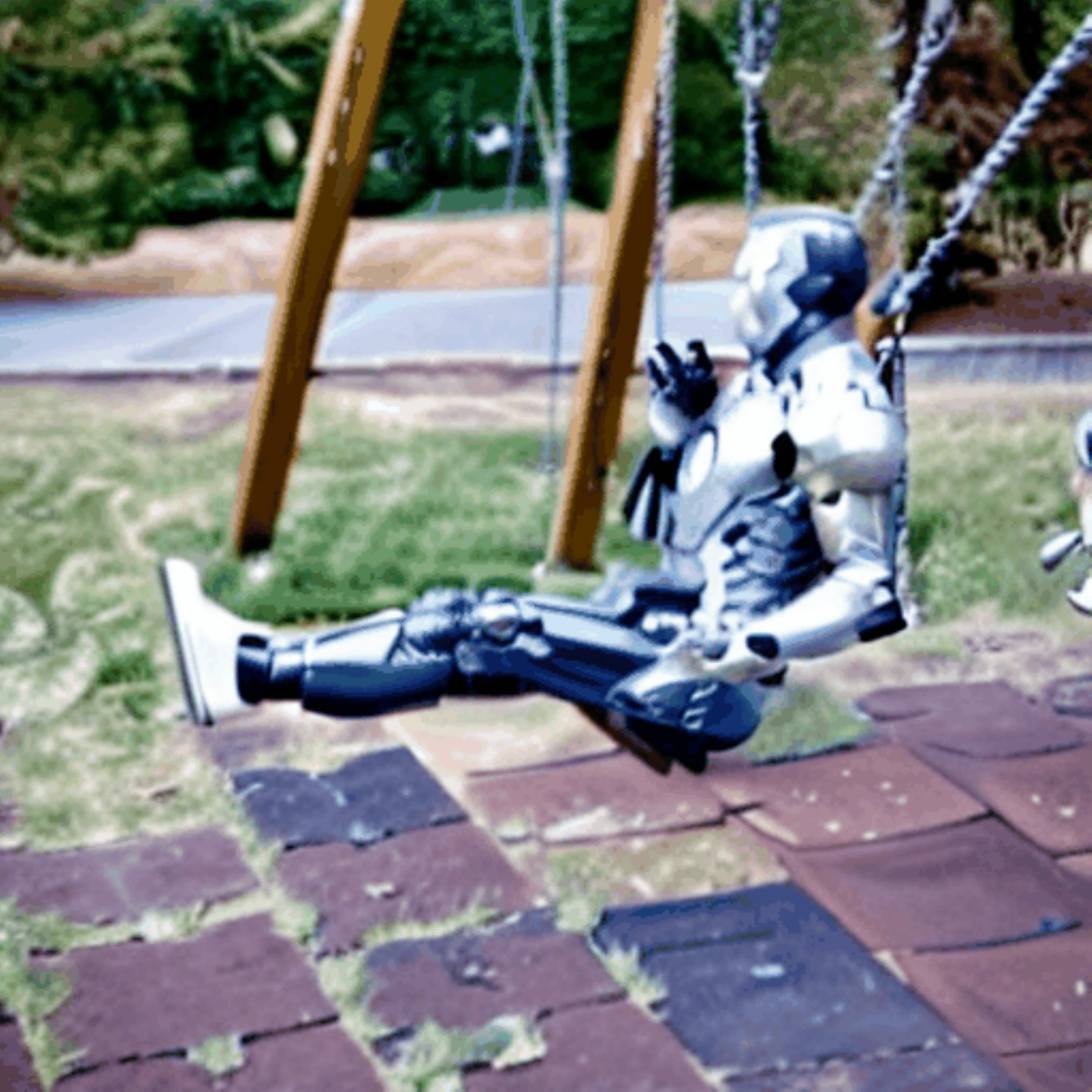}
\includegraphics[width=0.11\textwidth]{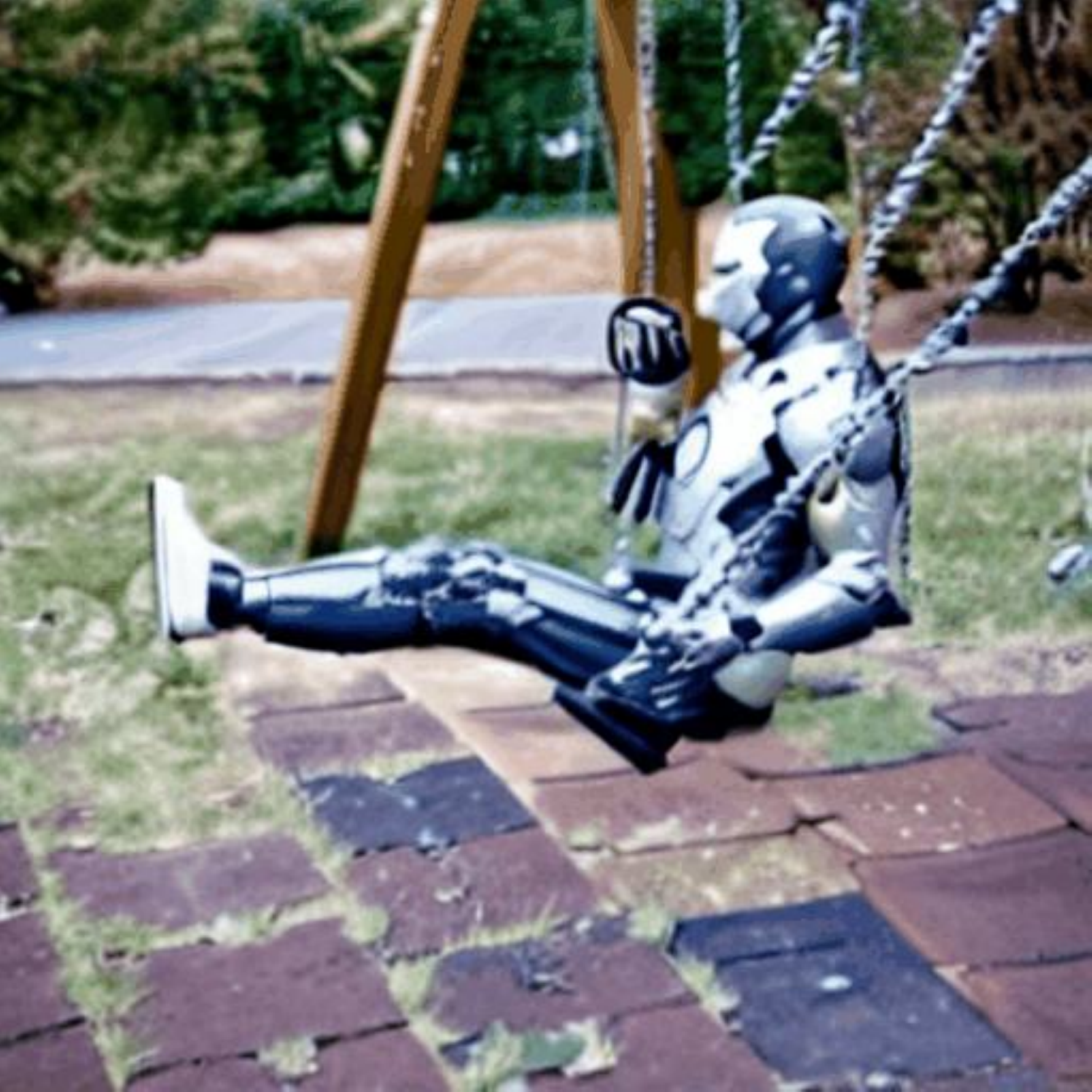}
\includegraphics[width=0.11\textwidth]{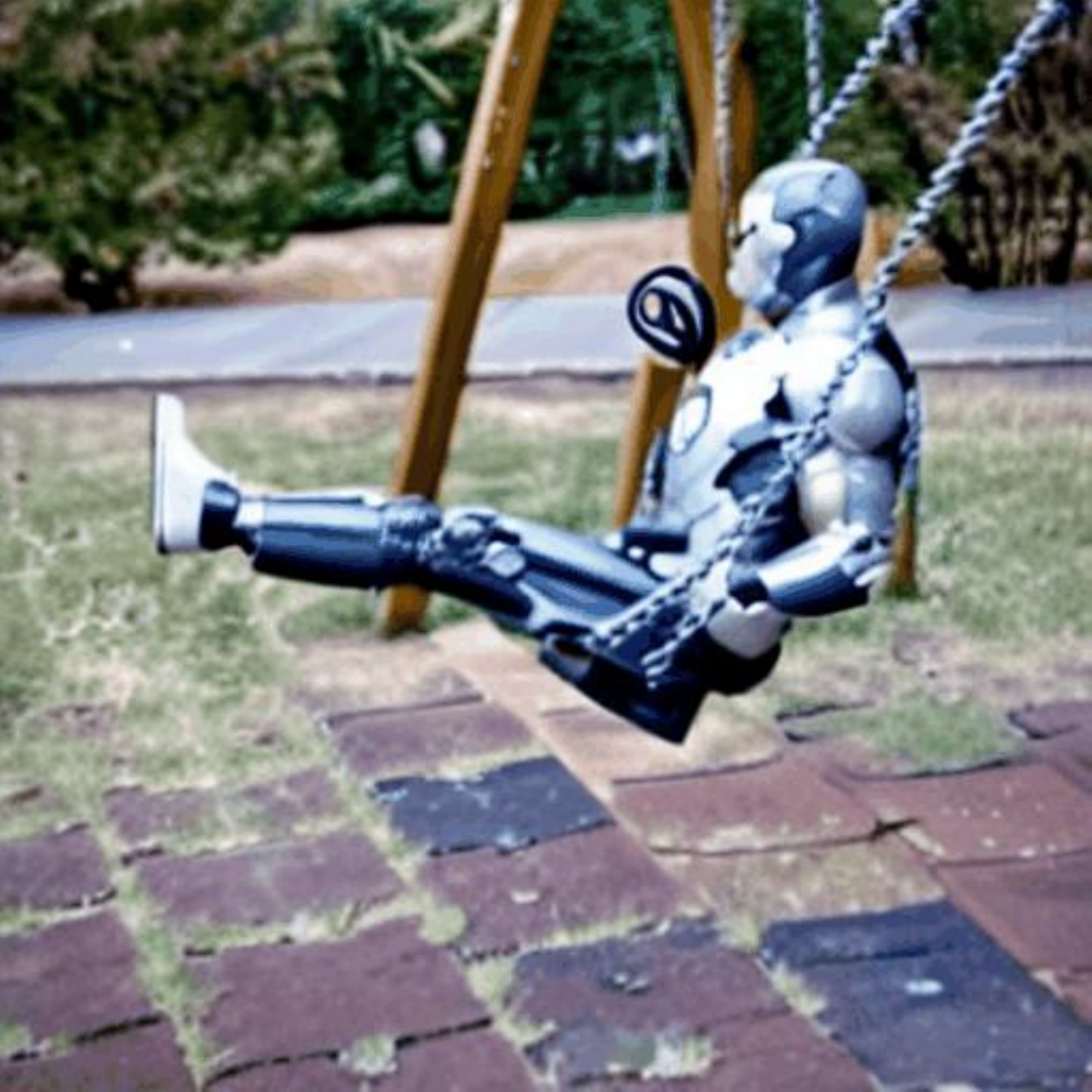}
\includegraphics[width=0.11\textwidth]{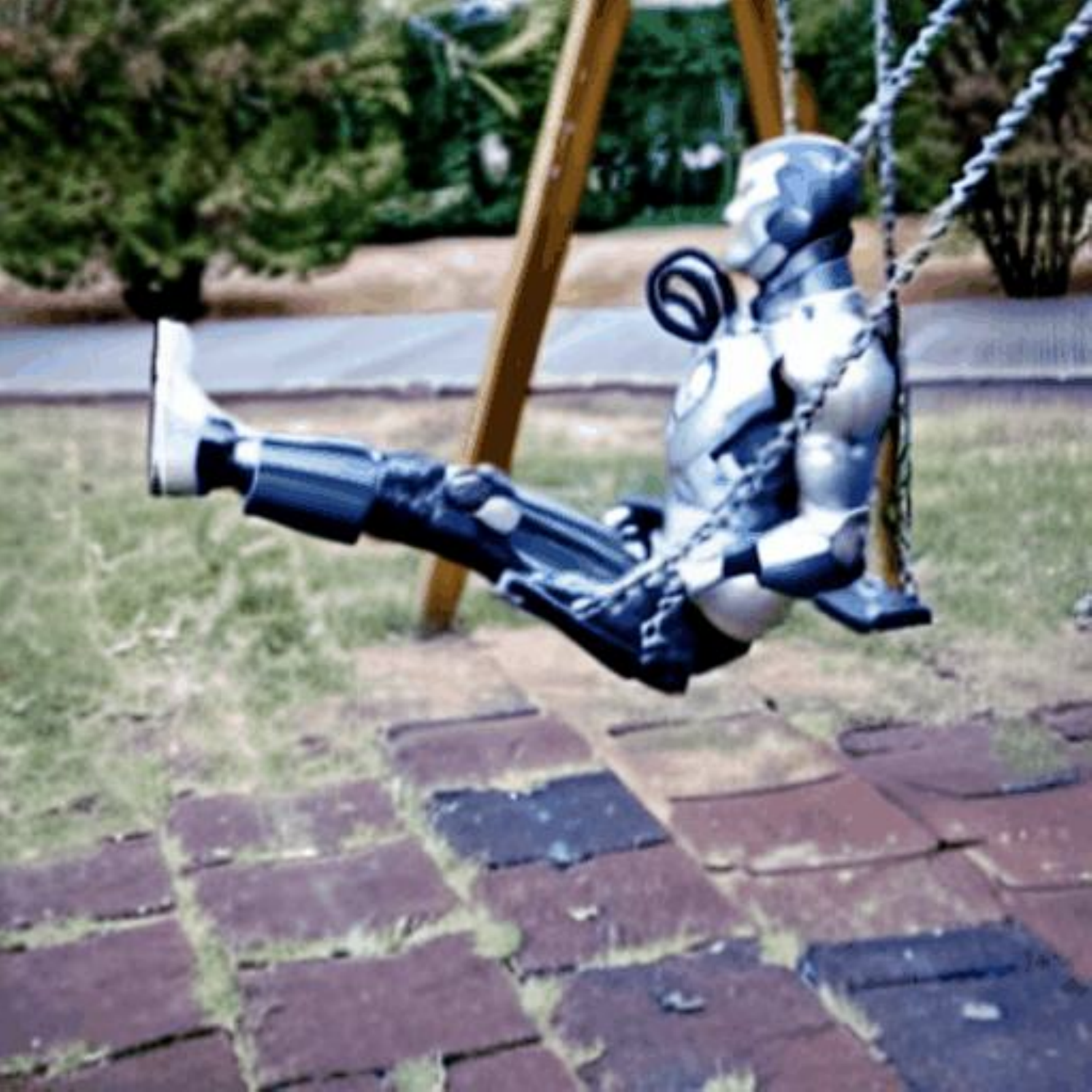}
\includegraphics[width=0.11\textwidth]{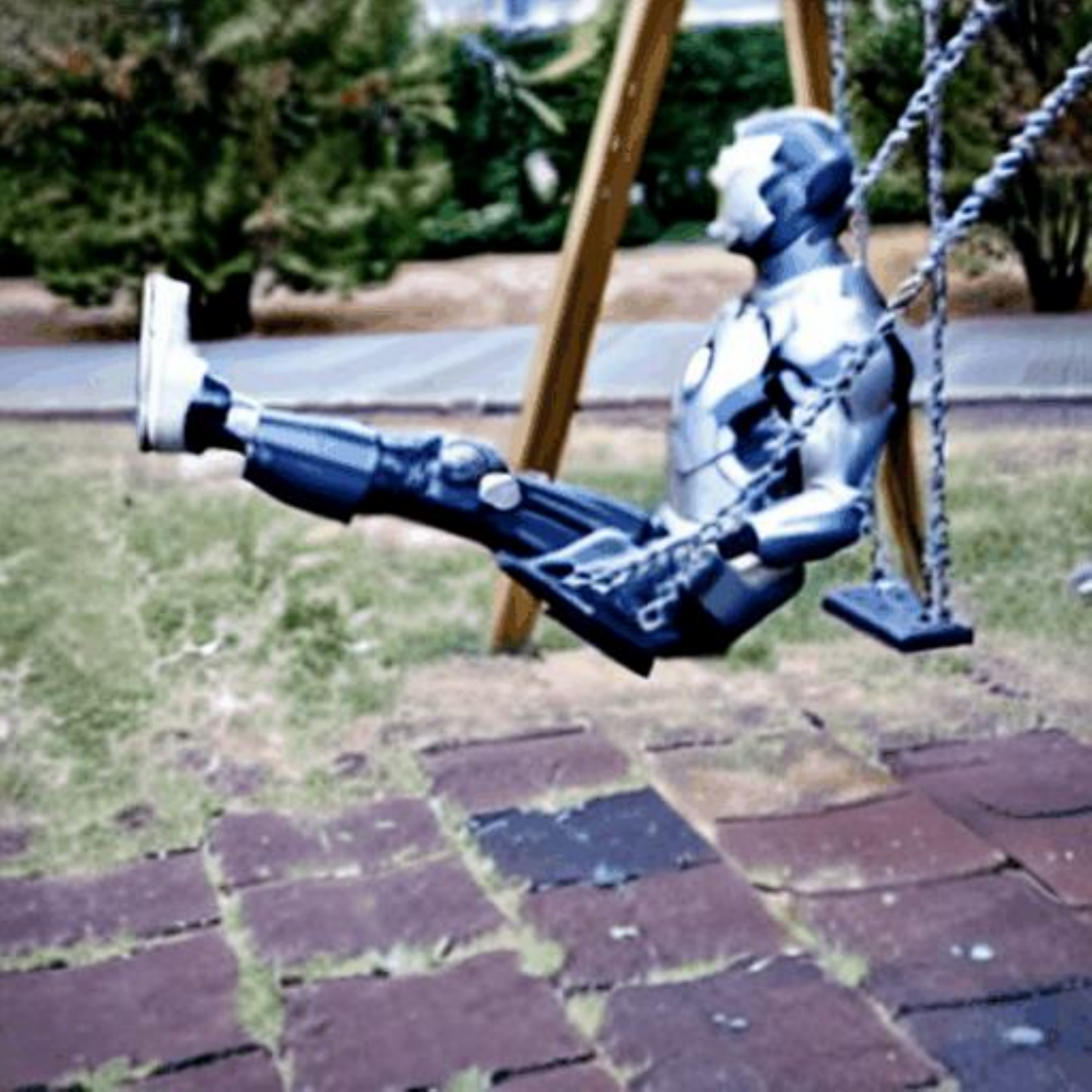}
\includegraphics[width=0.11\textwidth]{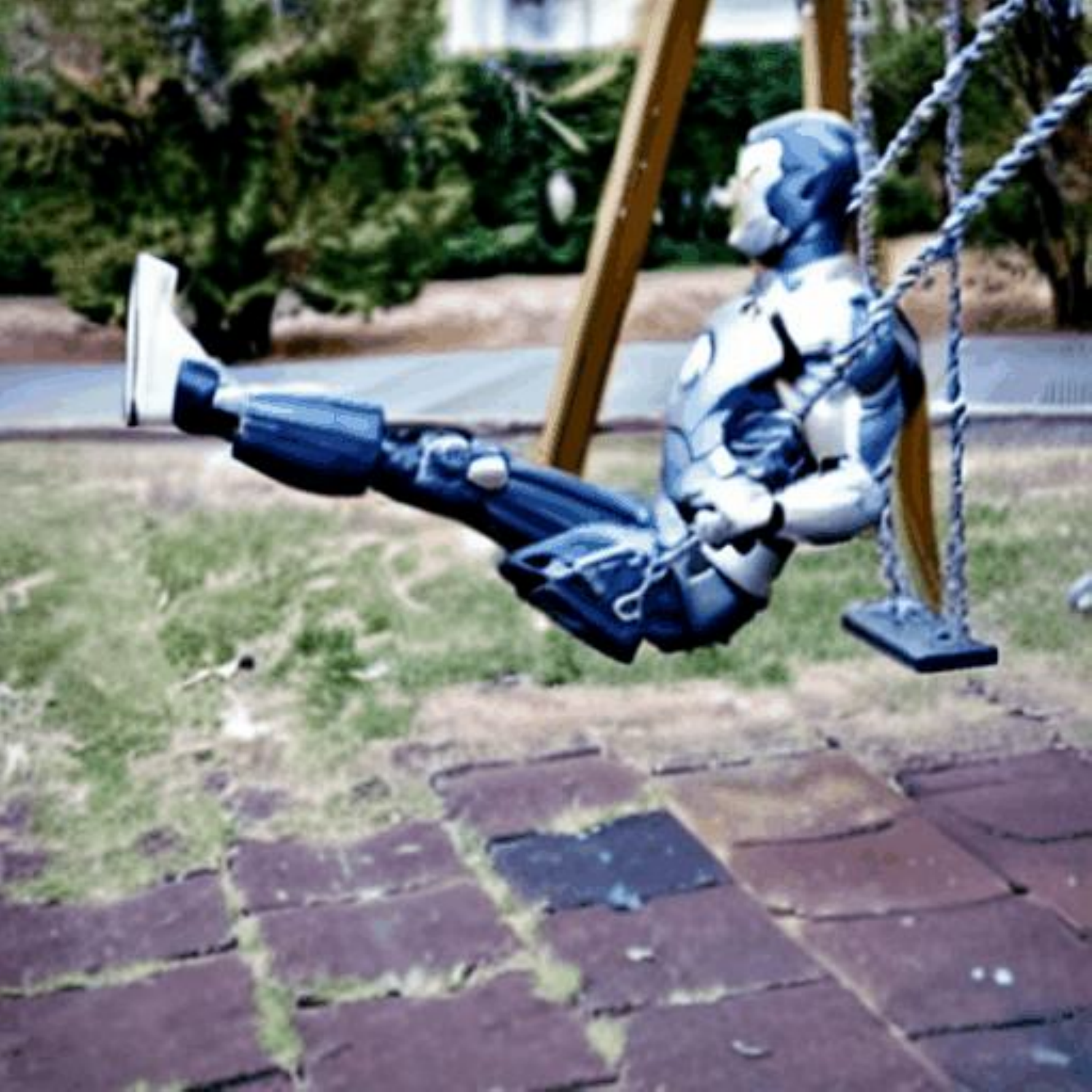}
\includegraphics[width=0.11\textwidth]{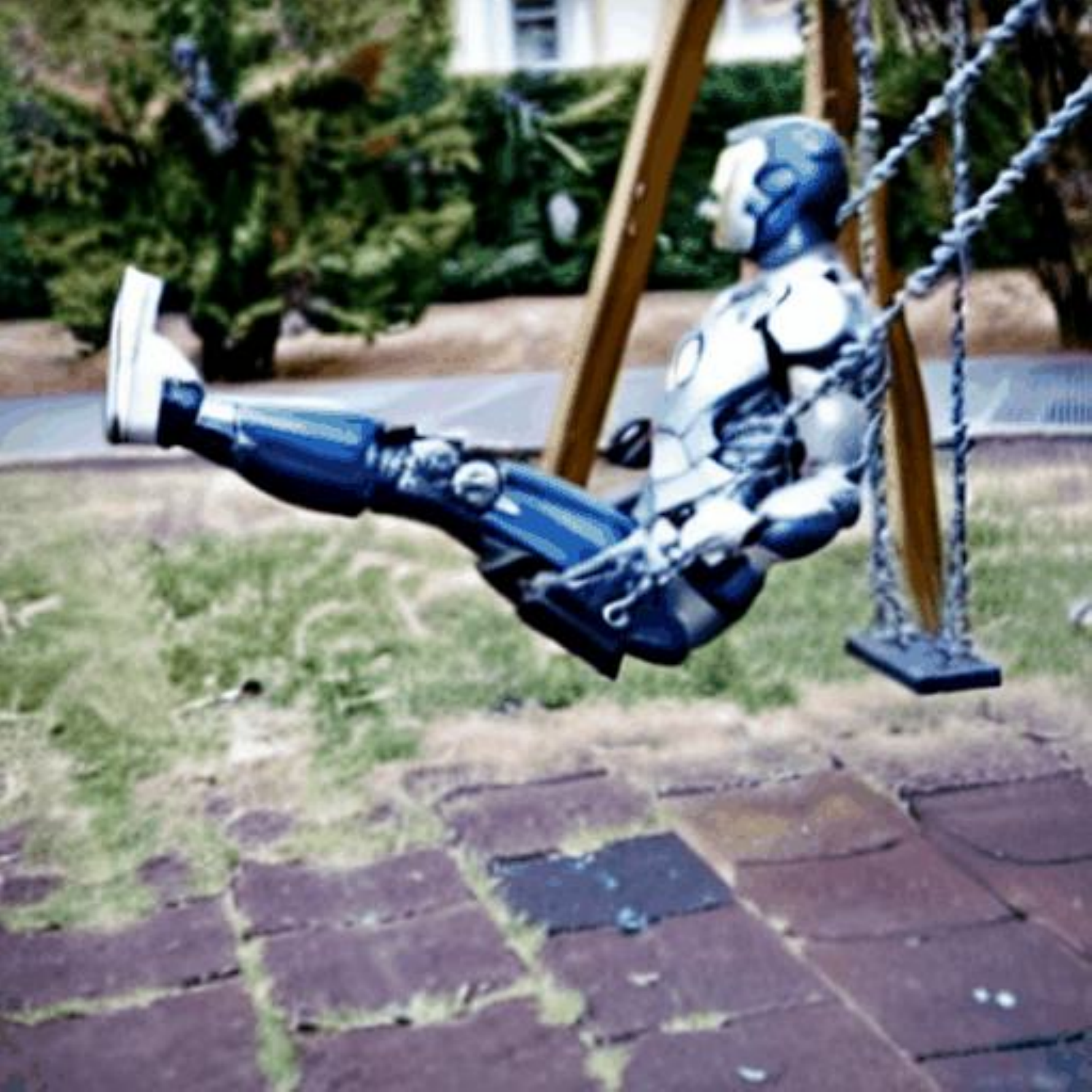}
\includegraphics[width=0.11\textwidth]{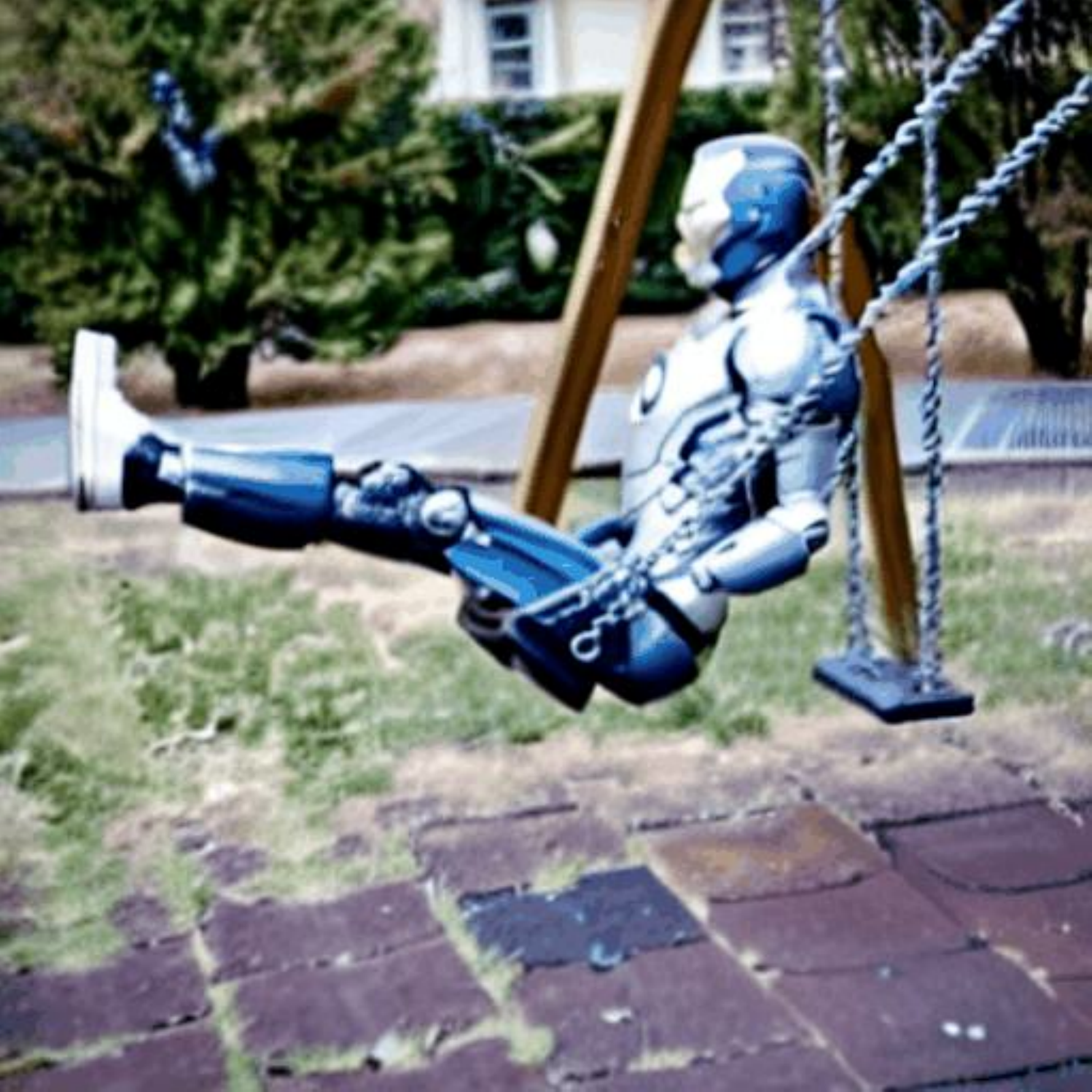}

\makebox[0.12\textwidth]{\colorbox{green}{\textbf{SDEdit}} A \textcolor{blue}{\textbf{Iron Man}} is on the swing}\\

\includegraphics[width=0.11\textwidth]{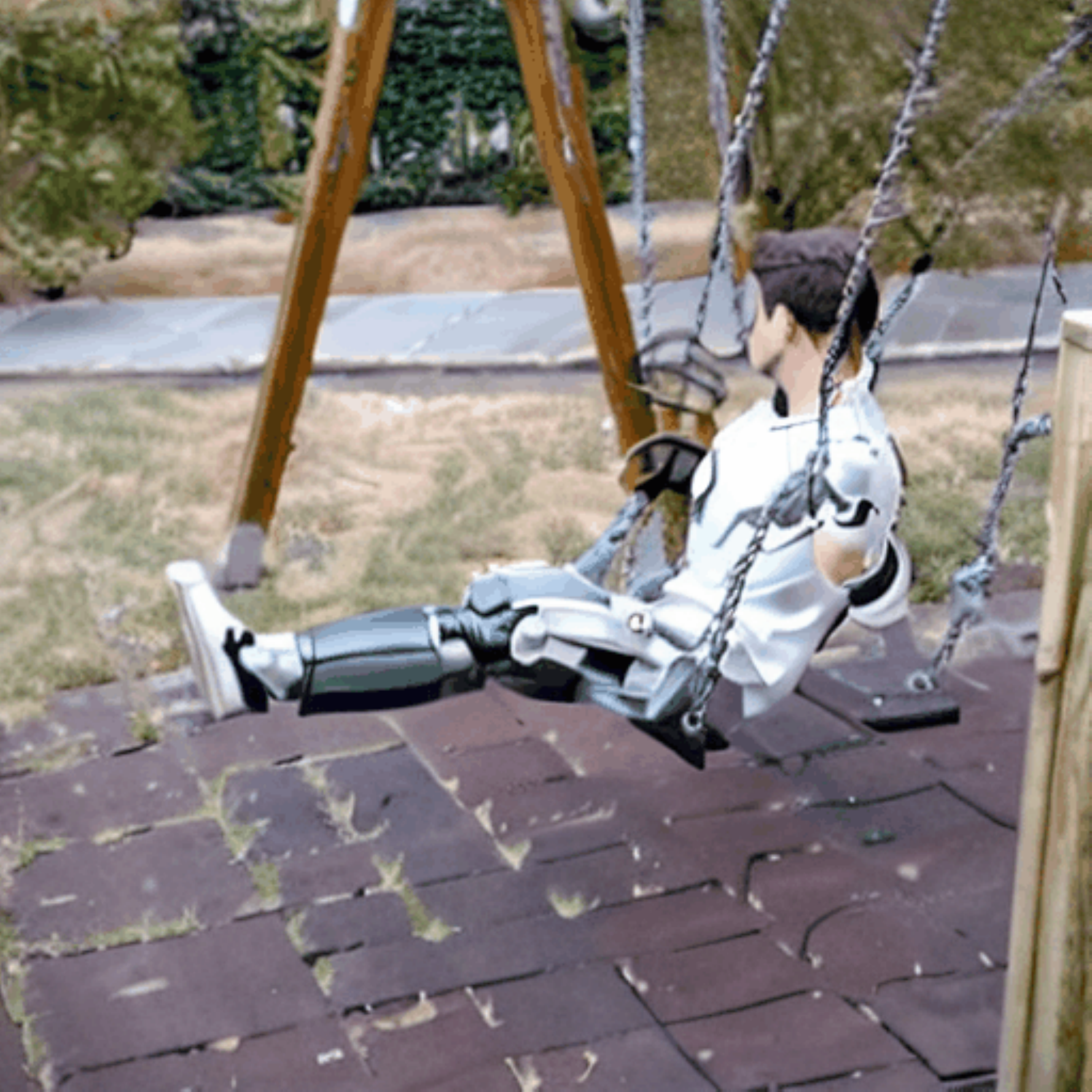}
\includegraphics[width=0.11\textwidth]{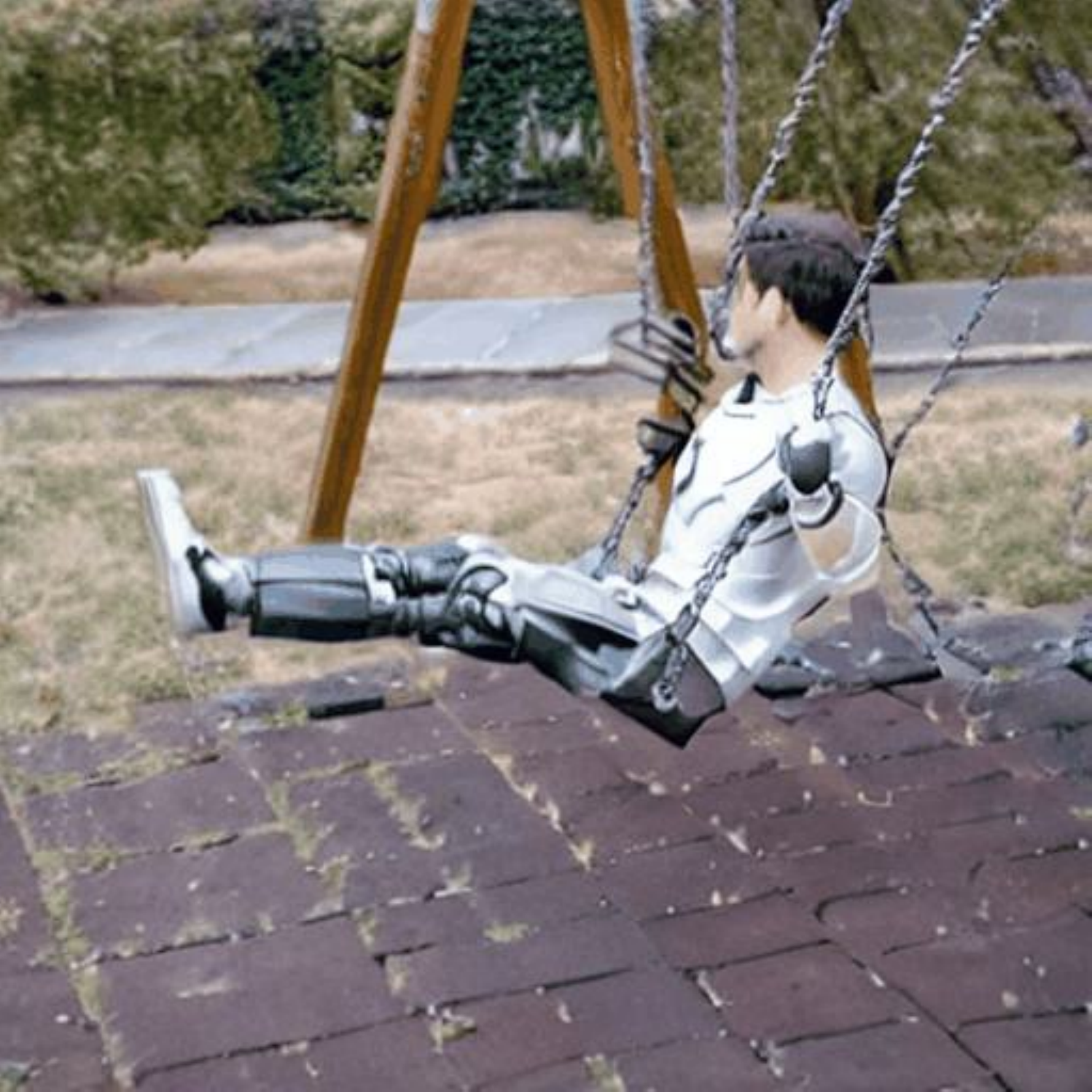}
\includegraphics[width=0.11\textwidth]{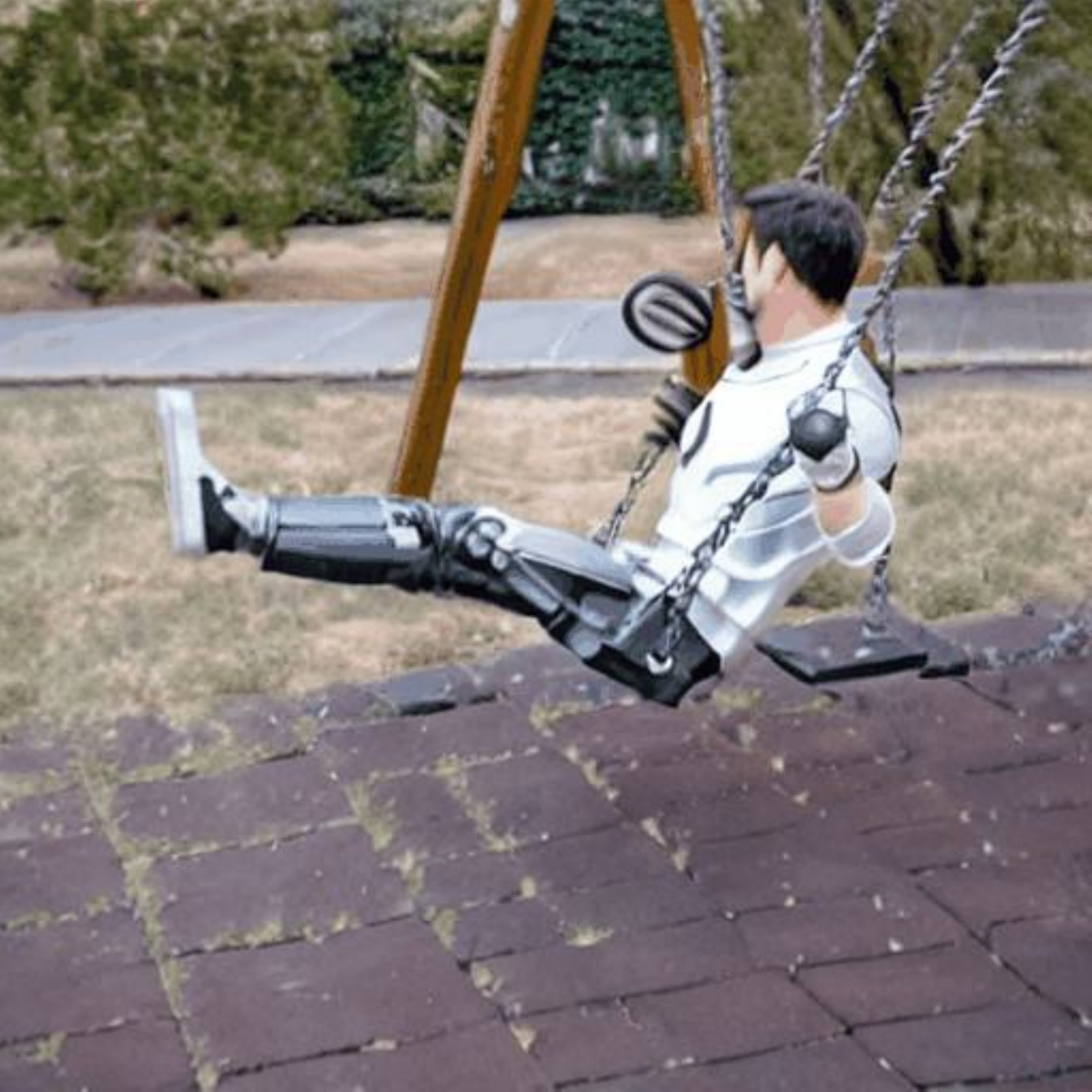}
\includegraphics[width=0.11\textwidth]{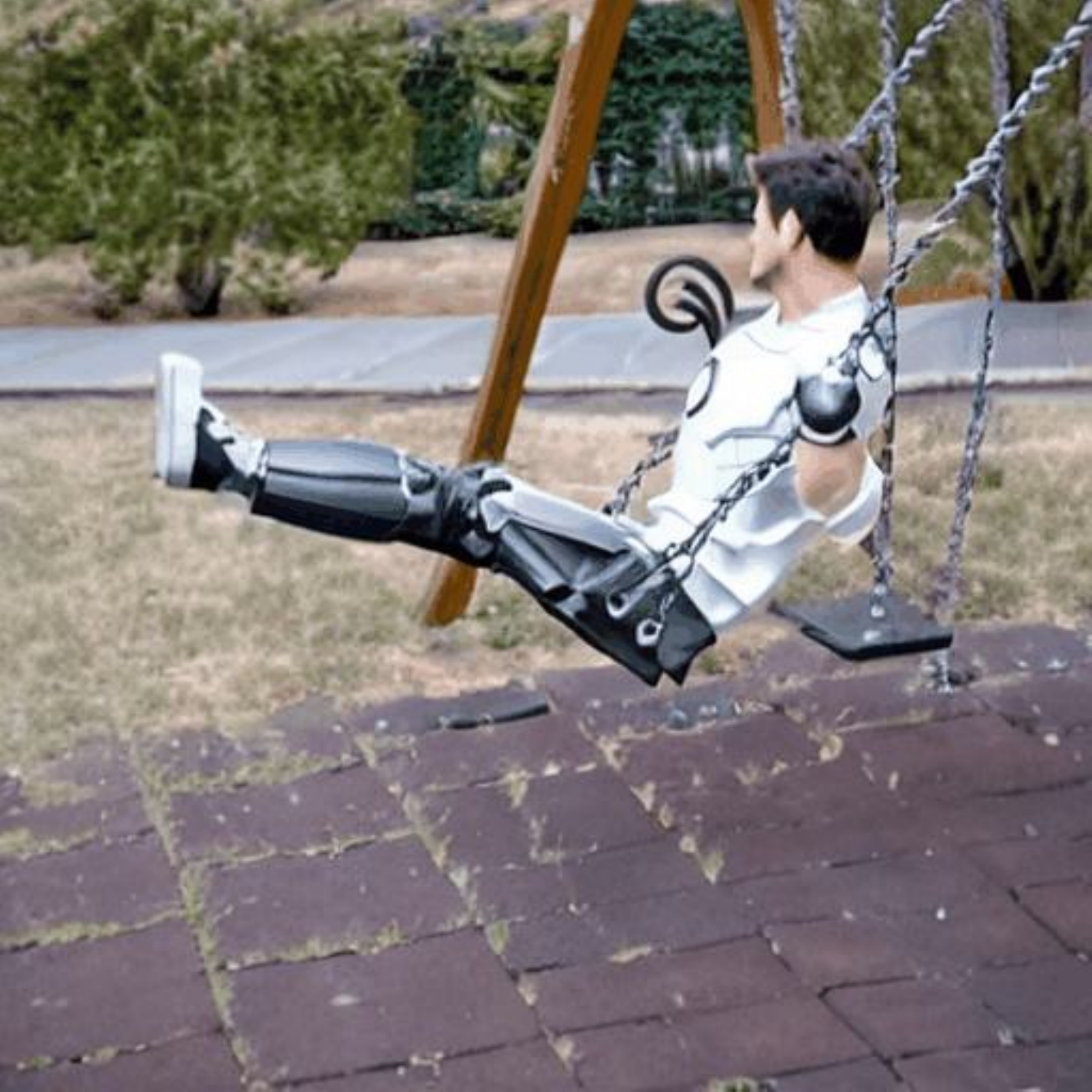}
\includegraphics[width=0.11\textwidth]{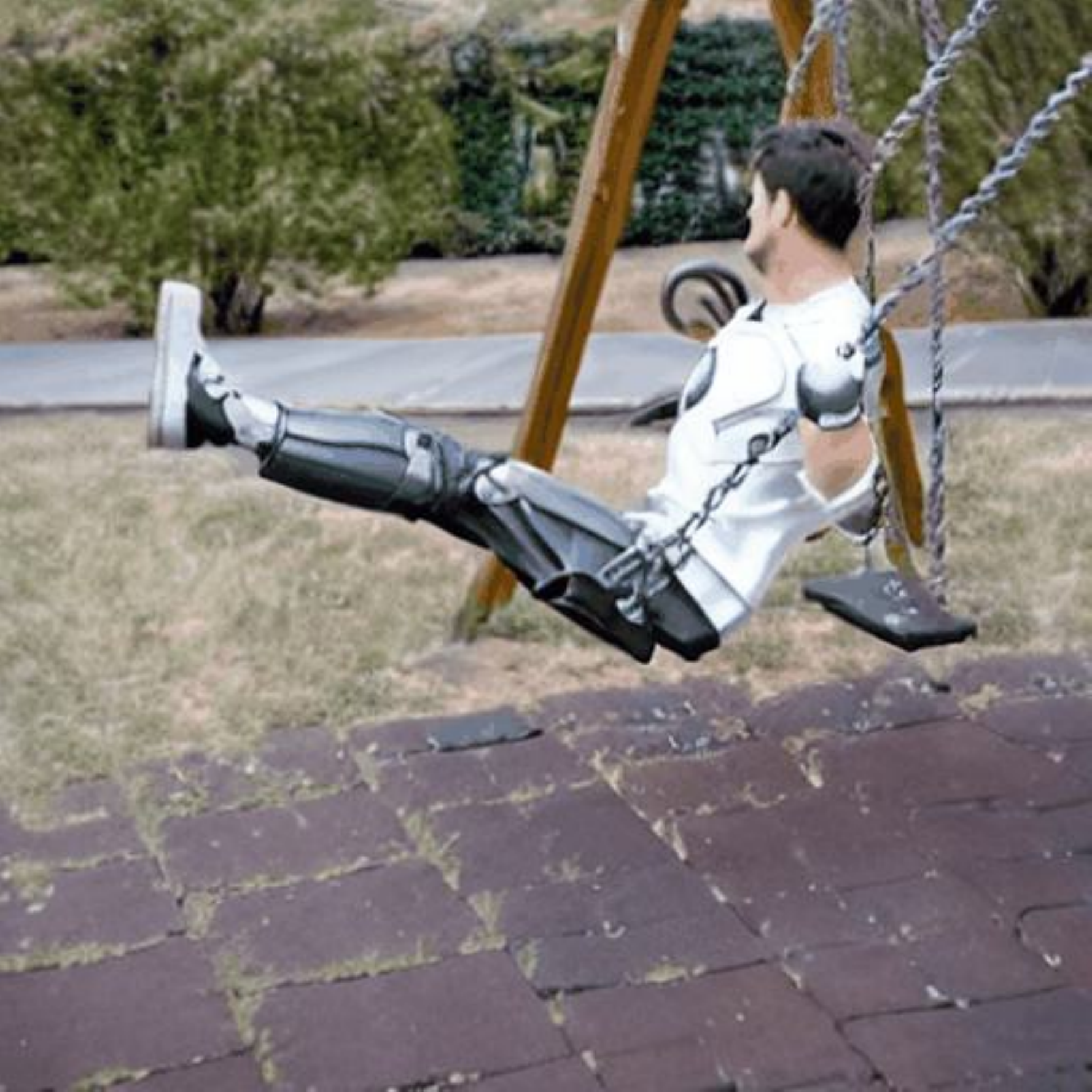}
\includegraphics[width=0.11\textwidth]{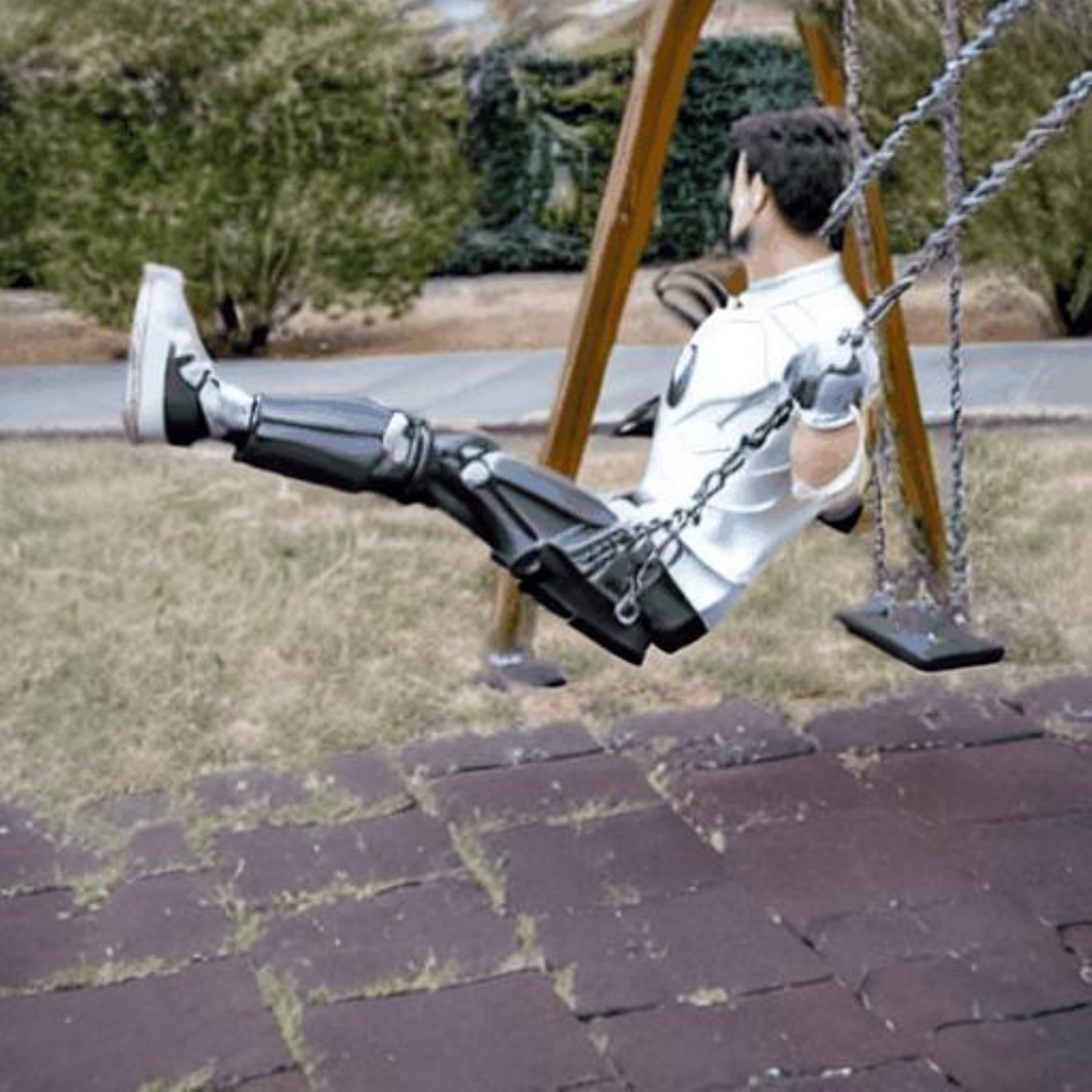}
\includegraphics[width=0.11\textwidth]{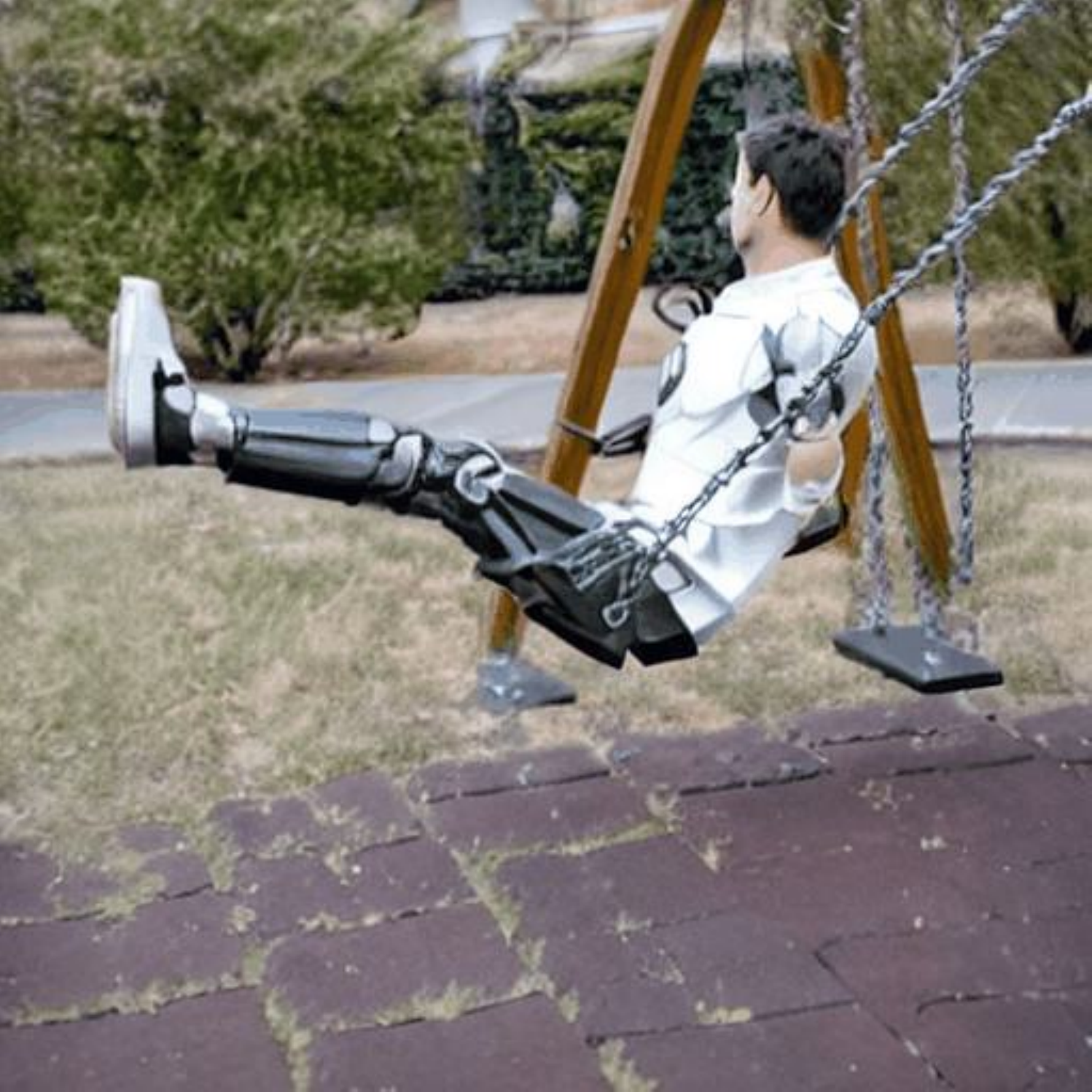}
\includegraphics[width=0.11\textwidth]{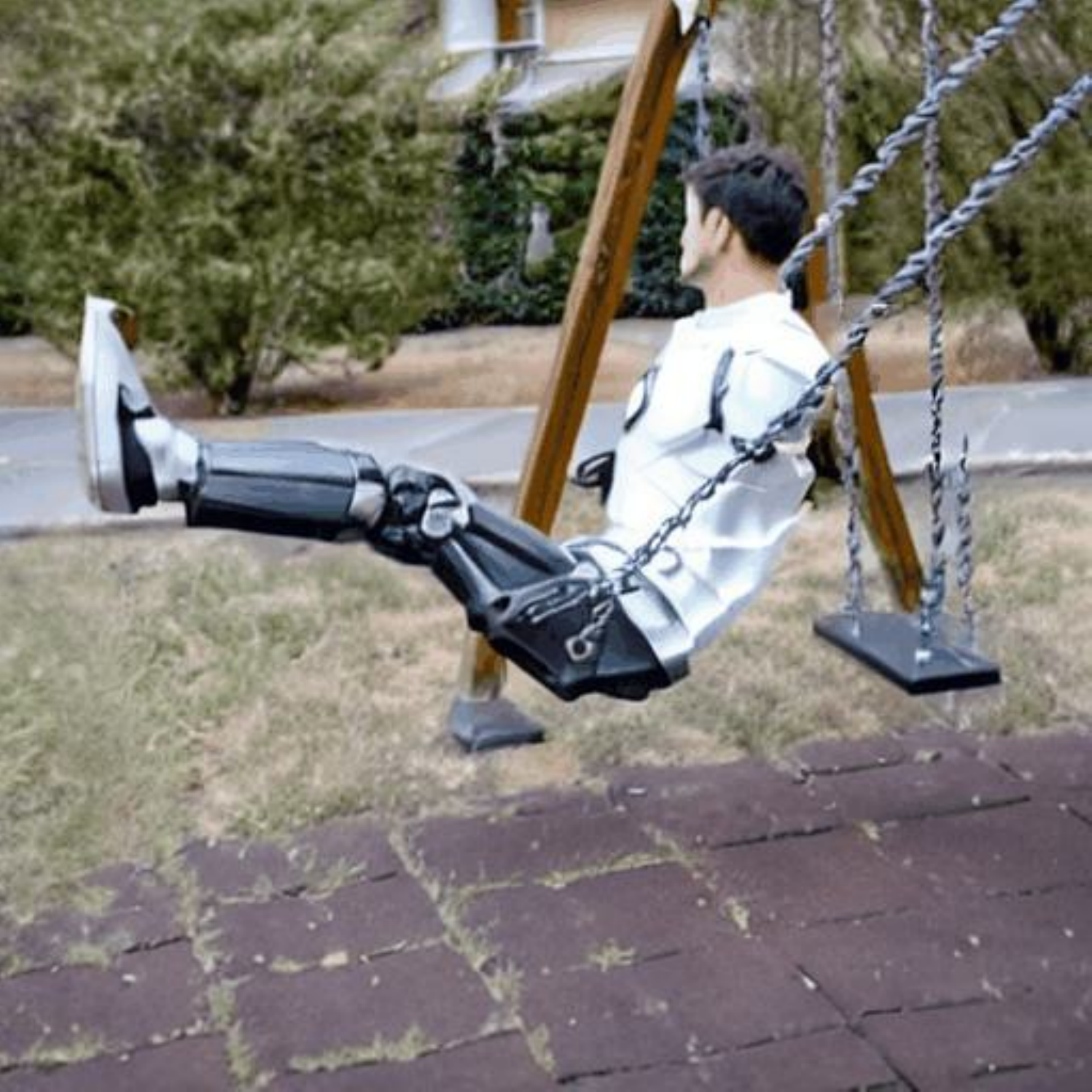}

\makebox[0.12\textwidth]{\colorbox{green}{\textbf{Video-P2P}} A \textcolor{blue}{\textbf{Iron Man}} is on the swing}\\

\includegraphics[width=0.11\textwidth]{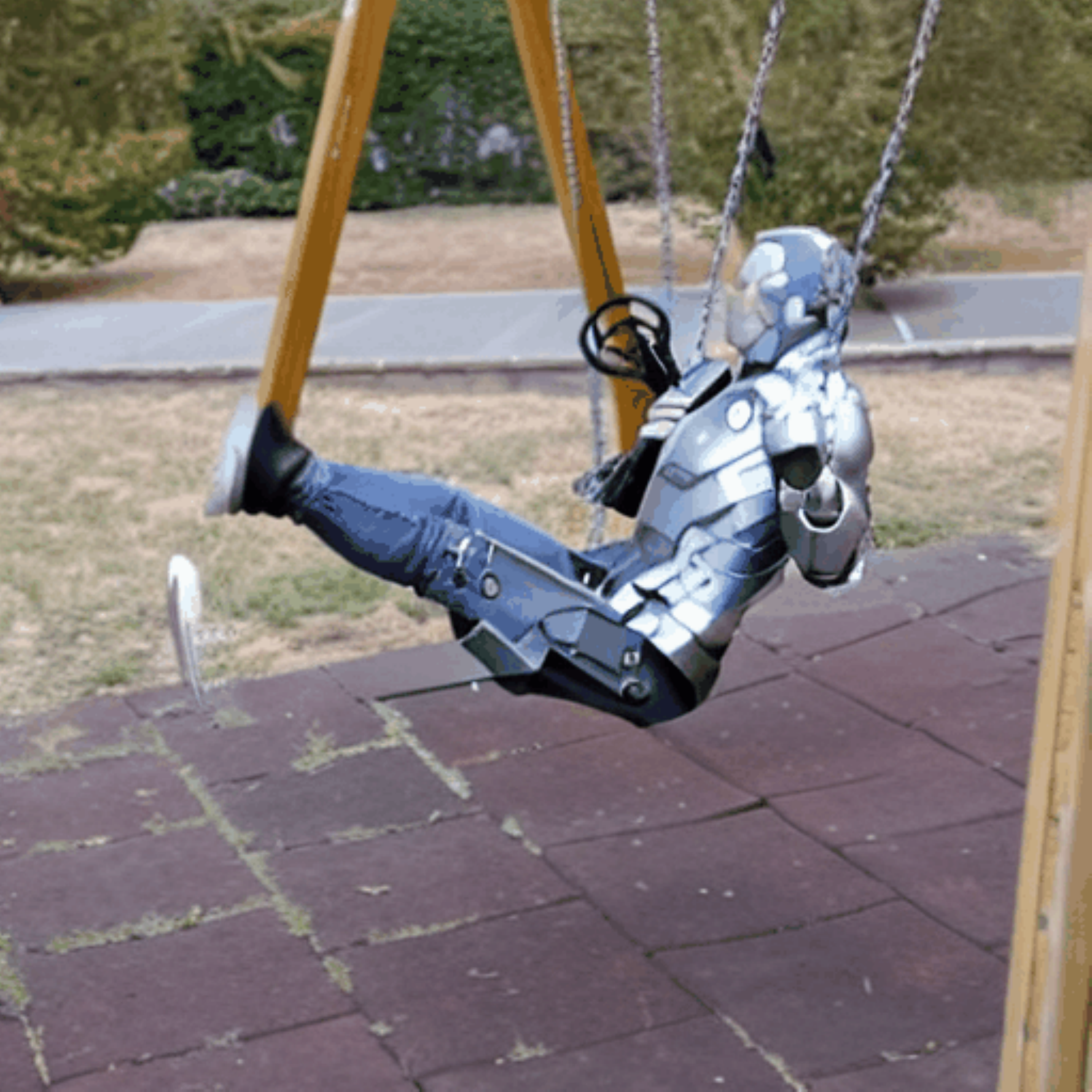}
\includegraphics[width=0.11\textwidth]{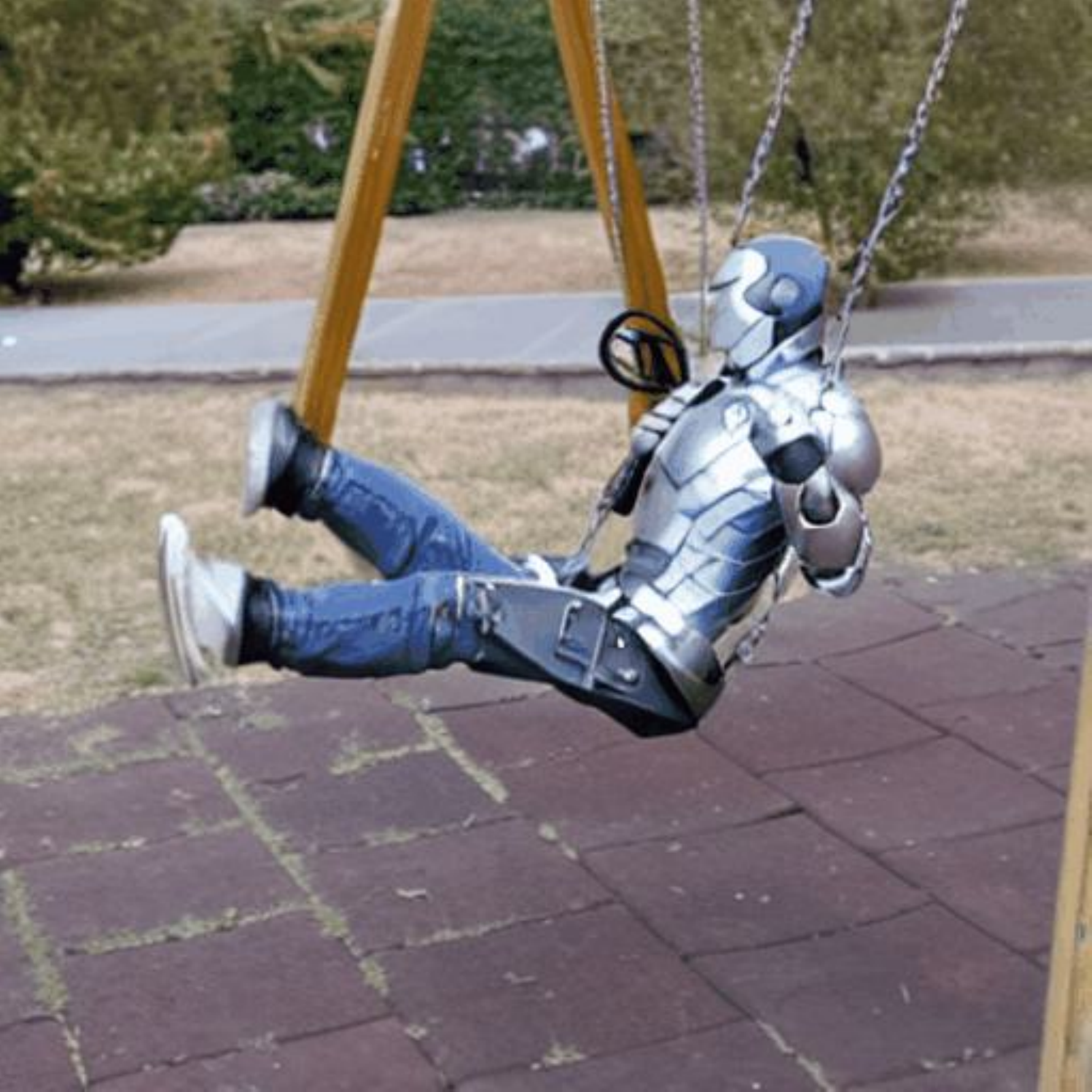}
\includegraphics[width=0.11\textwidth]{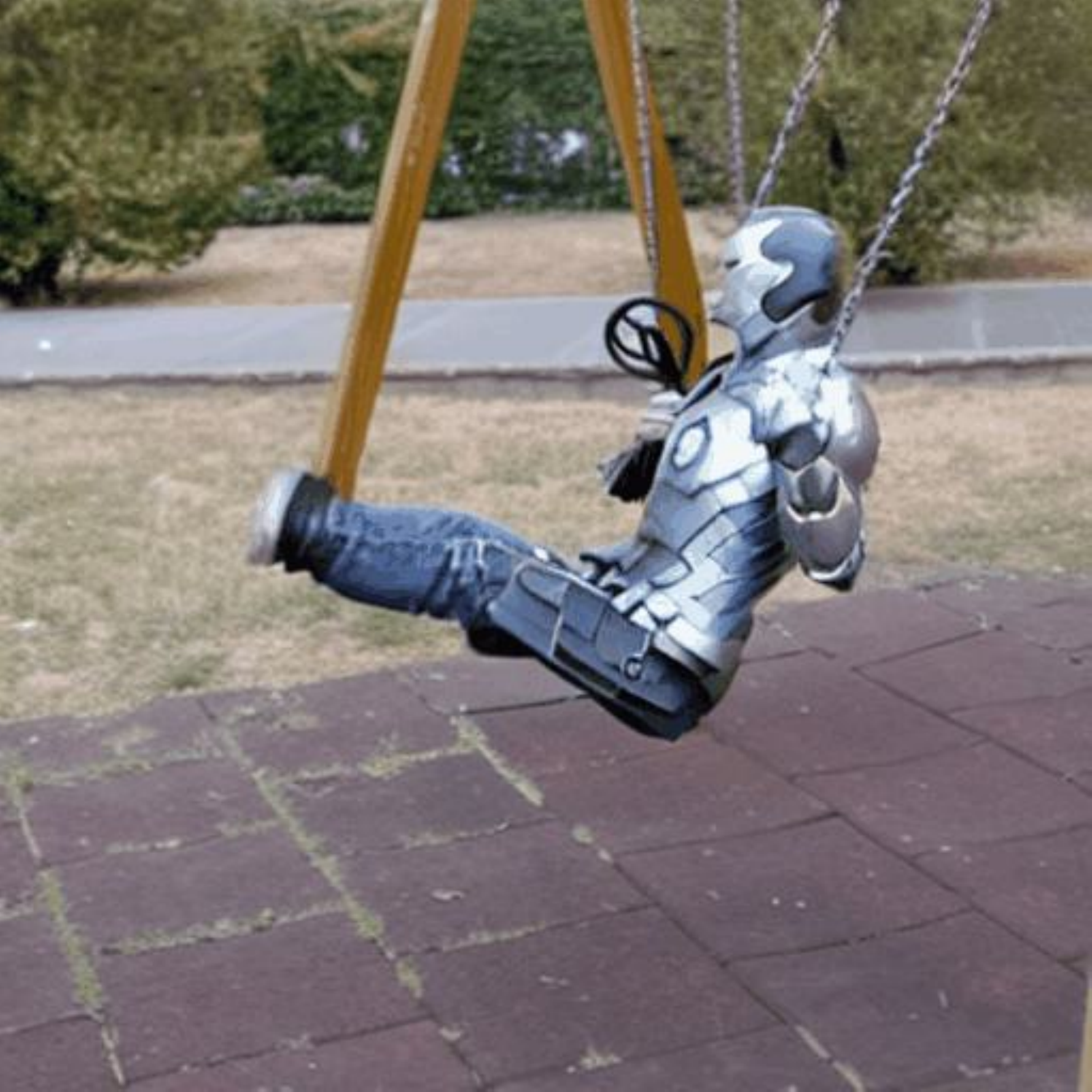}
\includegraphics[width=0.11\textwidth]{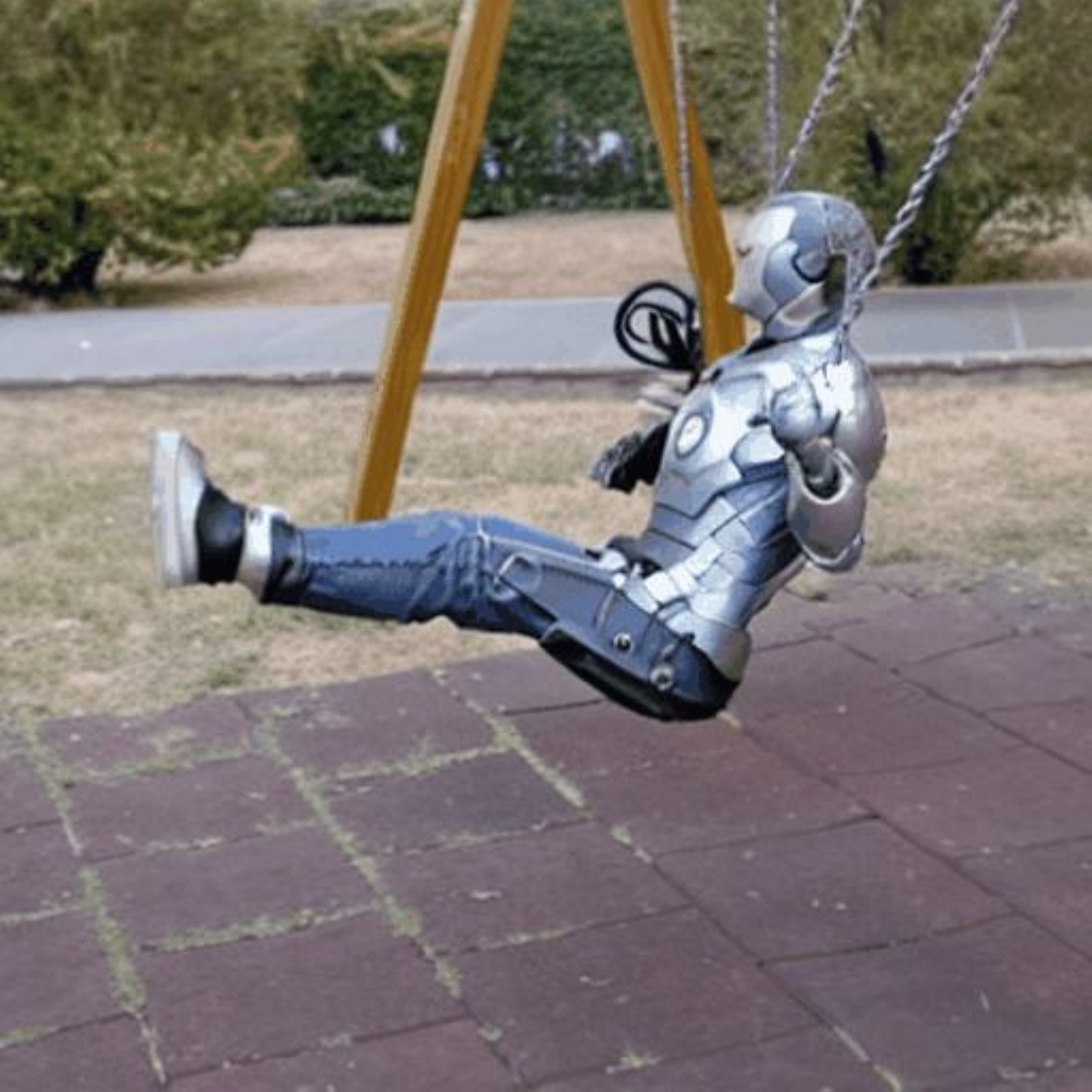}
\includegraphics[width=0.11\textwidth]{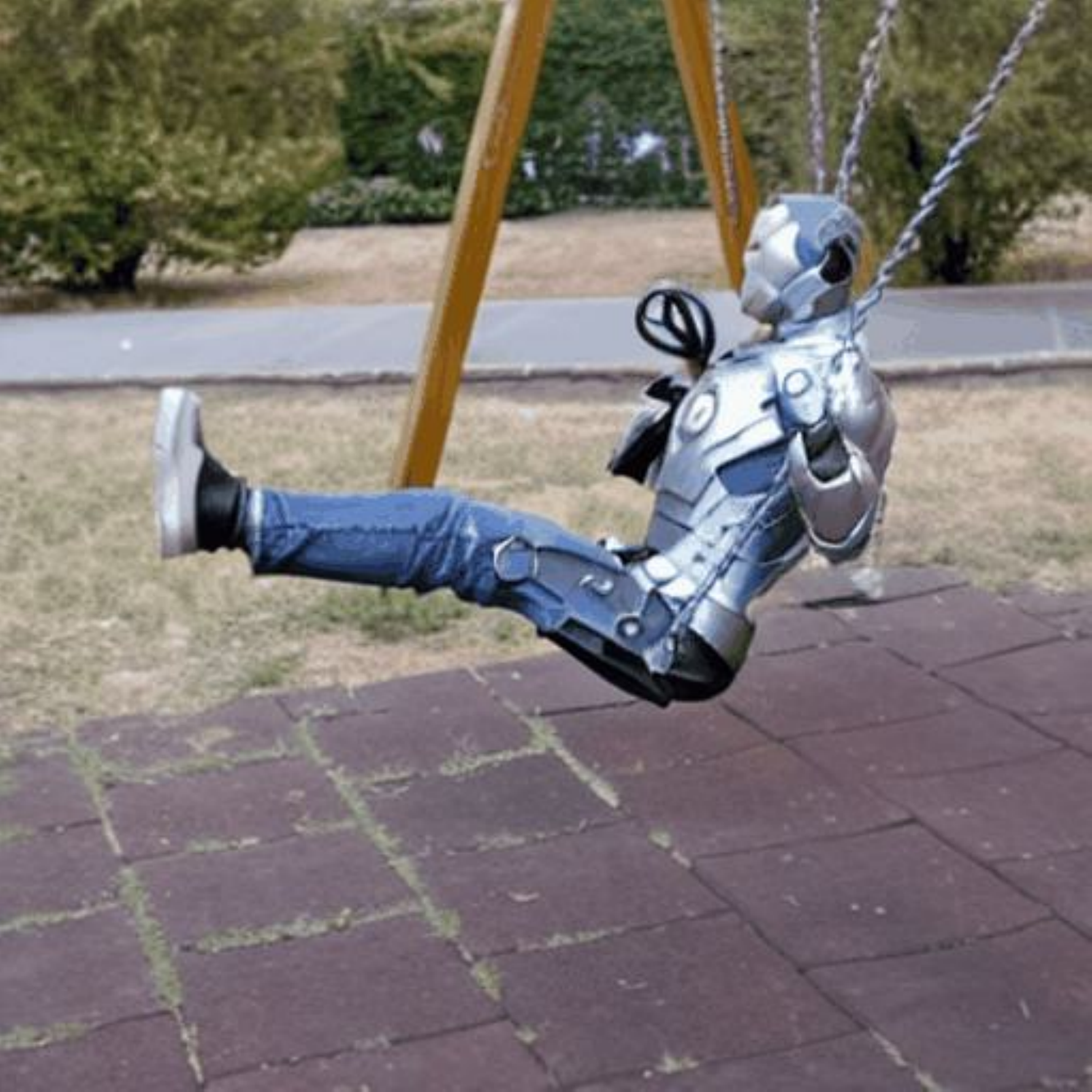}
\includegraphics[width=0.11\textwidth]{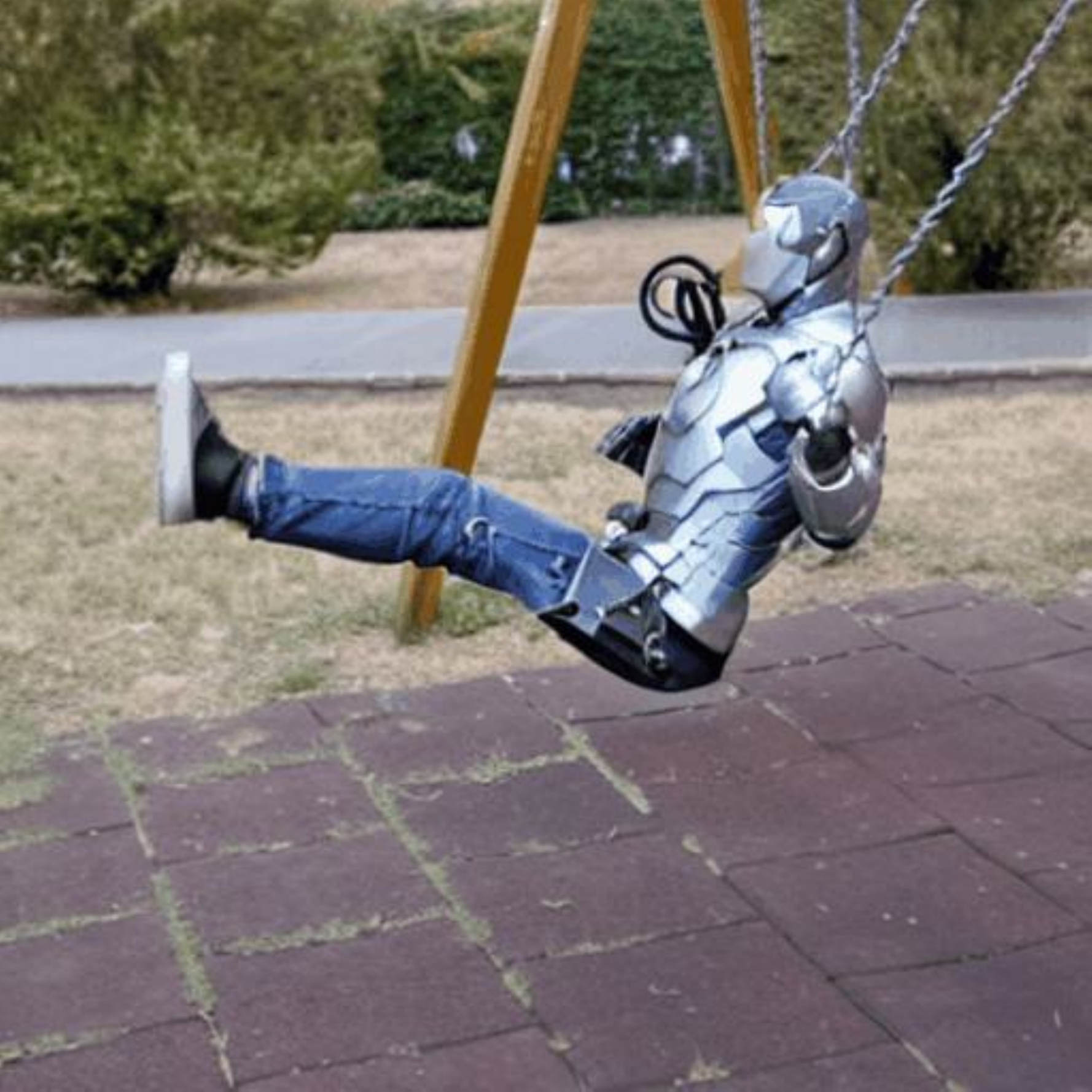}
\includegraphics[width=0.11\textwidth]{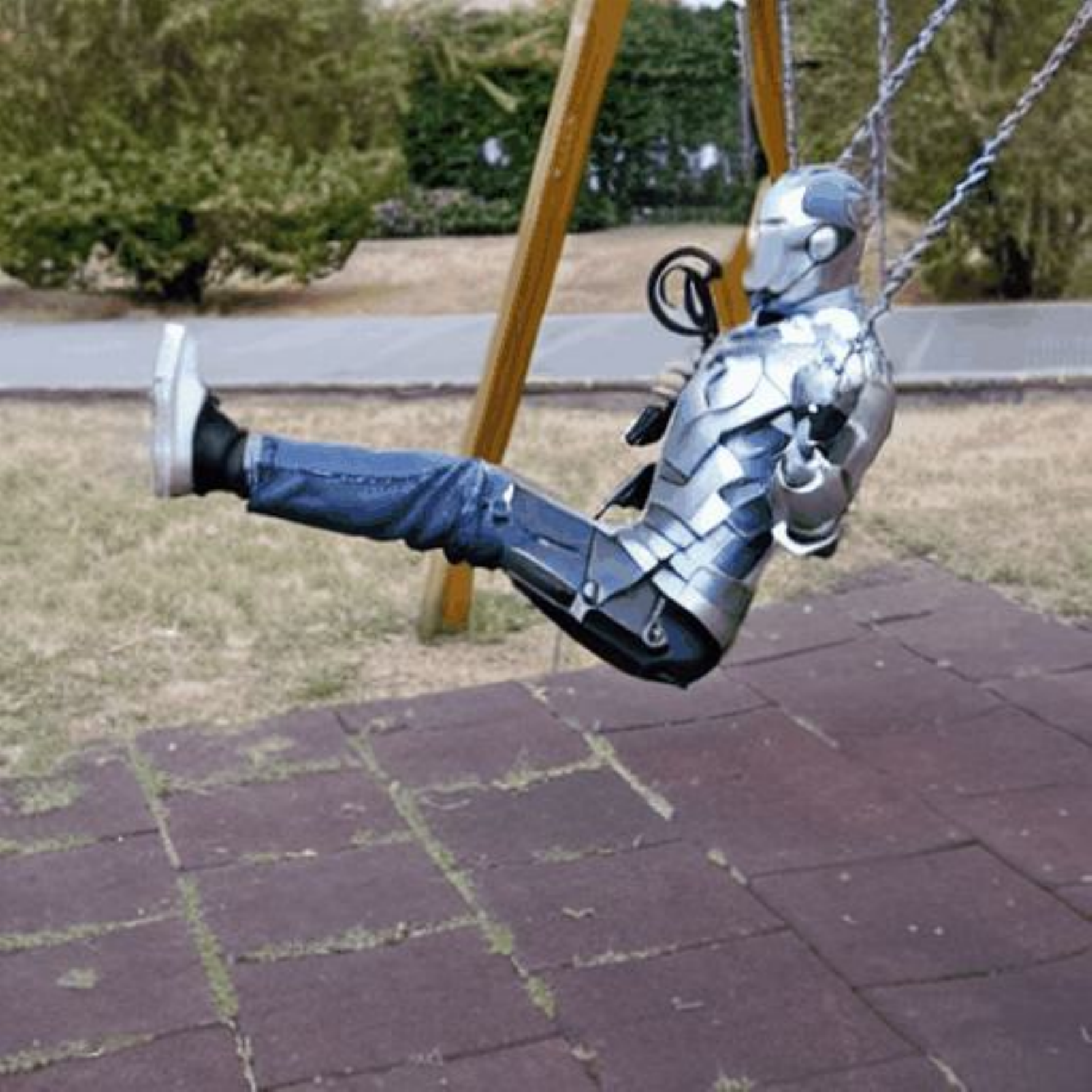}
\includegraphics[width=0.11\textwidth]{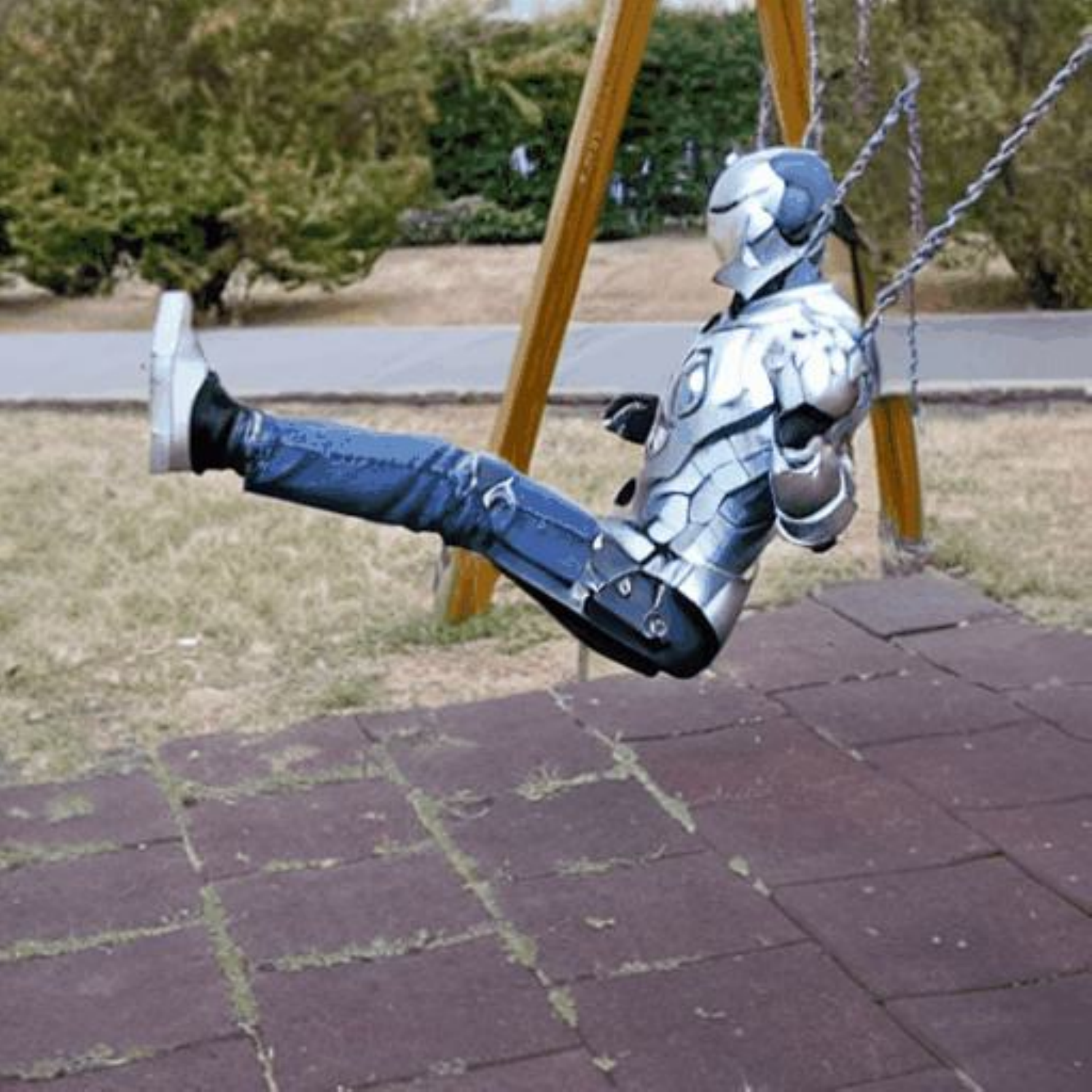}

\caption{\textbf{Qualitative Results} Edit-A-Video outperforms in editing compared to other baselines.}
\label{fig:main_result}
\end{center}
\end{figure*}

\begin{table}[t]
\small{
\caption{\textbf{Qualitative Comparisons to Baselines} We measure the overall human preference score (User Score (O)) and automatic metric scores for the comparisons to baselines.}
\begin{center}
\begin{tabular}{|c|c|c|c|c|}
\hline
Method & User Score (O) ($\uparrow$) & Text Alignment ($\uparrow$) & LPIPS ($\downarrow$) & PSNR ($\uparrow$) \\
\hline\hline
Edit-A-Video (Ours) & $3.80\pm0.10$ & $30.2688$ & $0.2625$ & $20.0992$ \\
Tune-A-Video & $3.46\pm0.10$ & $30.0514$ & $0.4482$ & $14.5753$ \\
SDEdit & $3.40\pm0.10$ & $28.4203$ & $0.2711$ & $20.4767$ \\
Video-P2P & $3.66\pm0.10$ & $30.0842$ & $0.3047$ &$17.5760$ \\
\hline
\end{tabular}
\end{center}
\vspace{-1.0em}
\label{tab:comparisons}
}
\end{table}

\subsection{Implementation Details}
We implement our method based on the stable-diffusion-v1-4\footnote{Stable Diffusion: \href{https://github.com/CompVis/stable-diffusion}{https://github.com/CompVis/stable-diffusion}}, publicly available TTI model~\citep{rombach2022high}.
We finetune only the latent diffusion model in TTI model on $8$ frame $512\times512$ video for $300$ steps in temporal modeling and $500$ steps in inversion, while fixing the autoencoder to encode each frame independently.
At inversion and sampling, we use $50$ step DDIM sampler and set the classifier-free guidance scale to $7.5$.
From the analysis introduced in Sec.~\ref{hparams}, we use cross-attention injection duration as $0.2$, spatio-temporal attention injection duration as $0.5$, and temporal attention injection duration as $0.8$. We set TC blending threshold ($\tau$) as $0.25$ and show the effects of the value in Sec.~\ref{ablation}.

For the quantitative evaluation, we edit a total $100$ $<$text, video$>$ pairs (four captions for each of $25$ videos), and the videos are collected from the web and DAVIS dataset~\citep{davis} as other works~\citep{esser2023structure, bar2022text2live}.
We set two of the four sentences for each video to change the style including the background, and the other two sentences to change the object.
We compare our model to baselines by human preference study, which we refer to as User Score (O), and automatic evaluation metrics.
In the human preference study, we ask $62$ users to grade the overall quality score of the edited video on a scale of $1-5$ considering three aspects: background preservation, text alignment, and video realism.
We include detailed explanations in the Supplementary Materials.

For the detailed analysis, we evaluate all models with three automatic metrics, one for editing performance and two for background preservation.
We measure the text alignment for the editing performance, which estimates how much the editing reflects the target text by averaging the cosine similarity between CLIP embedding of target text and CLIP image embeddings of all frames.
We further measure the distance between the source video and the target video for the background preservation by LPIPS~\citep{zhang2018unreasonable} and PSNR following the previous works~\citep{esser2023structure,mokady2022null,hertz2022prompt}.

\subsection{Baseline Comparisons}
\label{baseline_comparison}
We compare our method with three baselines quantitatively and qualitatively:
(1) \textit{Tune-A-Video}: generating the video from target prompt after tuning the inflated 3D model with source text-video pair.
(2) \textit{SDEdit}: Based on Tune-A-Video, editing the video with another method, SDEdit~\citep{meng2021sdedit} which injects the noise to video until intermediate timestep $t_{0}=25$ among $50$ steps following \cite{meng2021sdedit} and denoises from it conditioned on target prompt.
(3) \textit{Video-P2P}: concurrent single video editing method, which inflates the 2D model by replacing self-attention with first-frame attention, where the attention matrix of the current frame is calculated only based on the first frame.

\textbf{Quantitative Results}
We perform the user evaluation to let the participants grade the scores on the edited videos. 
Owing to the proposed techniques, Edit-A-Video achieves superior performance compared to baselines with statistical significance (p-value $<$ $0.05$ from the Wilcoxon signed-rank test), as shown in table~\ref{tab:comparisons}.
We further measure automatic evaluation metrics to support the user score, which are also included in table~\ref{tab:comparisons}.
Since Tune-A-Video generates entire frames of video corresponding to the target text, the synthesized sample accurately reflects the target text. However, we observe that Tune-A-Video modifies even the background to be preserved, which is in line with the results obtained from LPIPS and PSNR measurements.
SDEdit, another baseline, preserves the contents of the source video, yet shows the lowest text alignment score, which indicates that it does not reflect the target text faithfully.
Unlike the aforementioned two baselines, Video-P2P preserves the property that is independent of the target editing to some extent while reflecting the target text. 
Edit-A-Video exhibited superior performance compared to Video-P2P in all metrics. 
Through these results, we confirm that Edit-A-Video is capable of editing video correspond to target text while preserving details that should be remained.

\textbf{Qualitative Results}
Fig.~\ref{fig:main_result} presents qualitative results of ours and baselines.
Tune-A-Video fails to maintain the background content, such as the color of the brick on the floor, while SDEdit fails to generate samples that reflect the target text. 
Although Video-P2P generates higher-quality samples than previous baselines, its imprecise blending mask leads to unintended changes in regions that should not be edited, such as the positioning of the legs.
Compared to these baselines, Edit-A-Video generates samples that preserve the property that should remain unedited and align accurately with the target prompt.
More examples are included in the Supplementary Materials.

\begin{table}[t]
\small{
\caption{\textbf{TC Blending Ablation Study} We demonstrate TC Blending's impact through subjective scores (User Scores) and automatic metrics. User Score (O) represents overall editing quality, while User Score (P) assesses non-target content preservation, including the background. Mask IoU measures IoU between the foreground mask from the saliency detector and the blending mask.
}
\begin{center}
\begin{tabular}{|c|c|c|c|c|c|}
\hline
Method & User Score (O) ($\uparrow$) & User Score (P) ($\uparrow$) & LPIPS 
($\downarrow$) & PSNR ($\uparrow$) & Mask IoU ($\uparrow$) \\
\hline\hline
Edit-A-Video & $3.80\pm0.10$ & $4.13\pm0.13$ & $0.2625$ & $20.0992$ & $0.3805$ \\
w/o TC-Bld & $3.69\pm0.10$ & $3.96\pm0.13$ & $0.2723$ & $19.8628$ & $0.2371$ \\

\hline
\end{tabular}
\end{center}
\vspace{-0.5em}
\label{tab:ablation}
}
\vspace{-0.5em}
\end{table}

\subsection{Ablation}
\label{ablation}

\begin{figure}[t]
\begin{center}
\begin{minipage}{0.45\textwidth}
\centering
\makebox[0.12\textwidth]{\colorbox{yellow}{\textbf{Baseline w/o TC Blending}}}\\
\rotatebox{90}{\parbox{0.20\textwidth}{\centering threshold \\ 0.10}}
\includegraphics[width=0.20\textwidth]{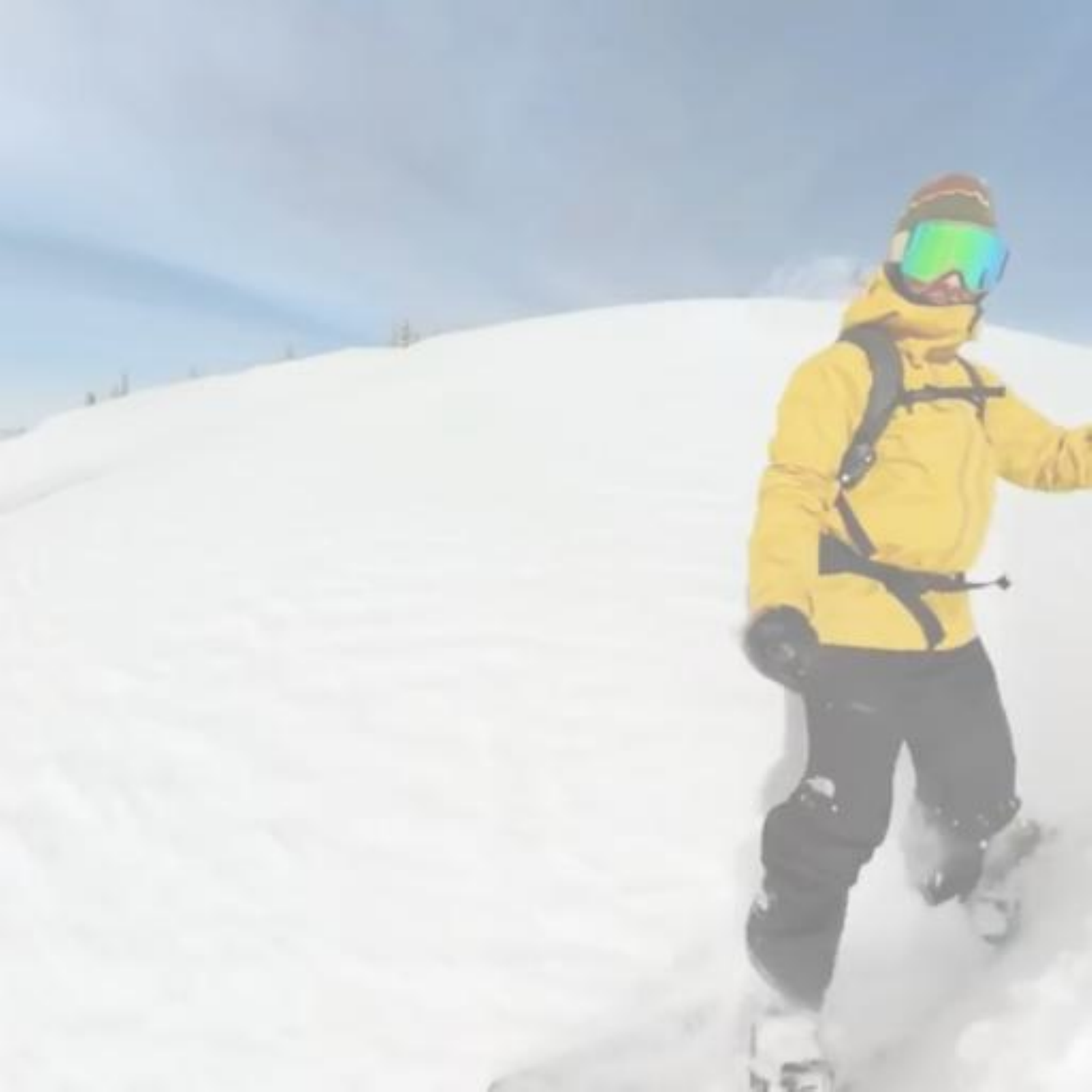}
\includegraphics[width=0.20\textwidth]{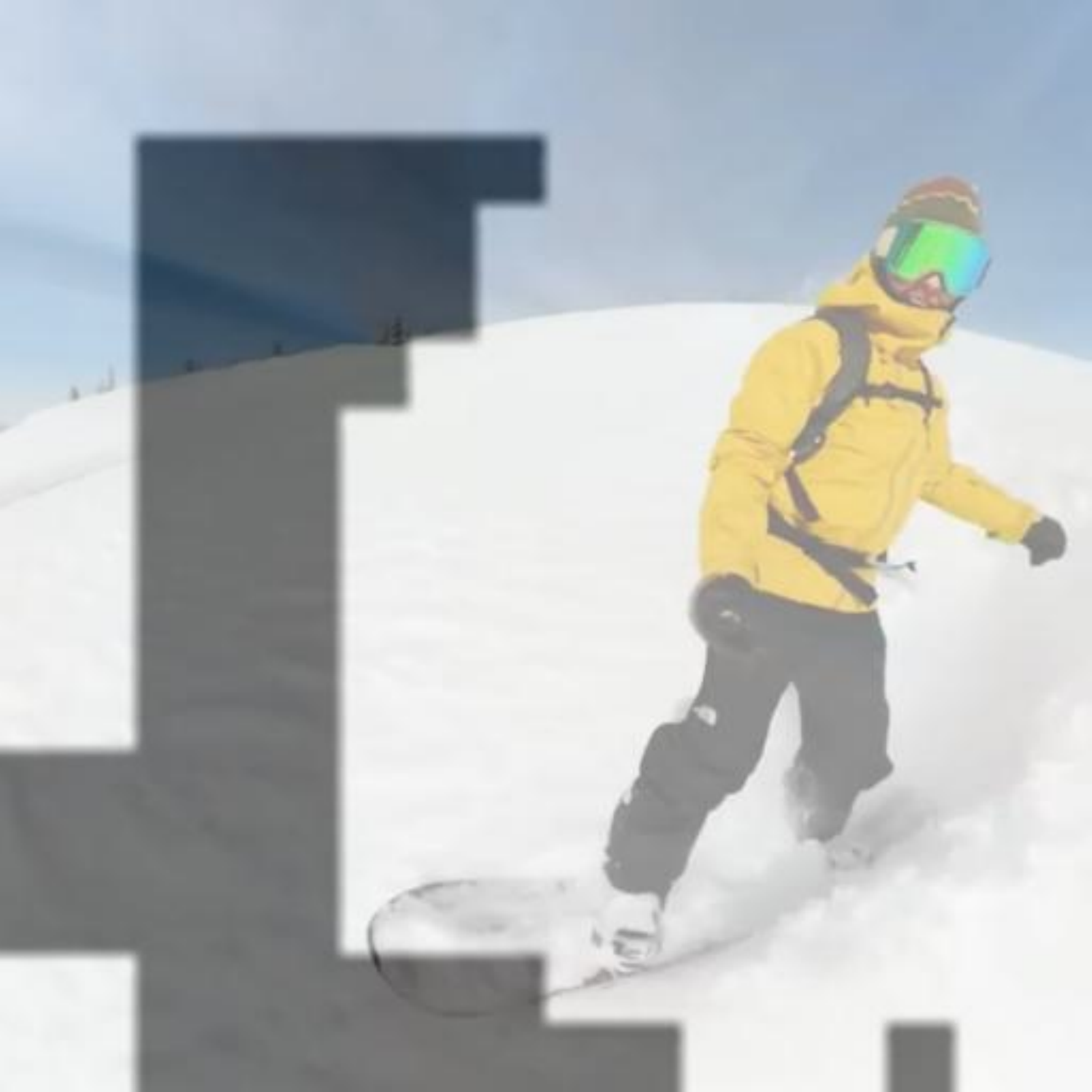}
\includegraphics[width=0.20\textwidth]{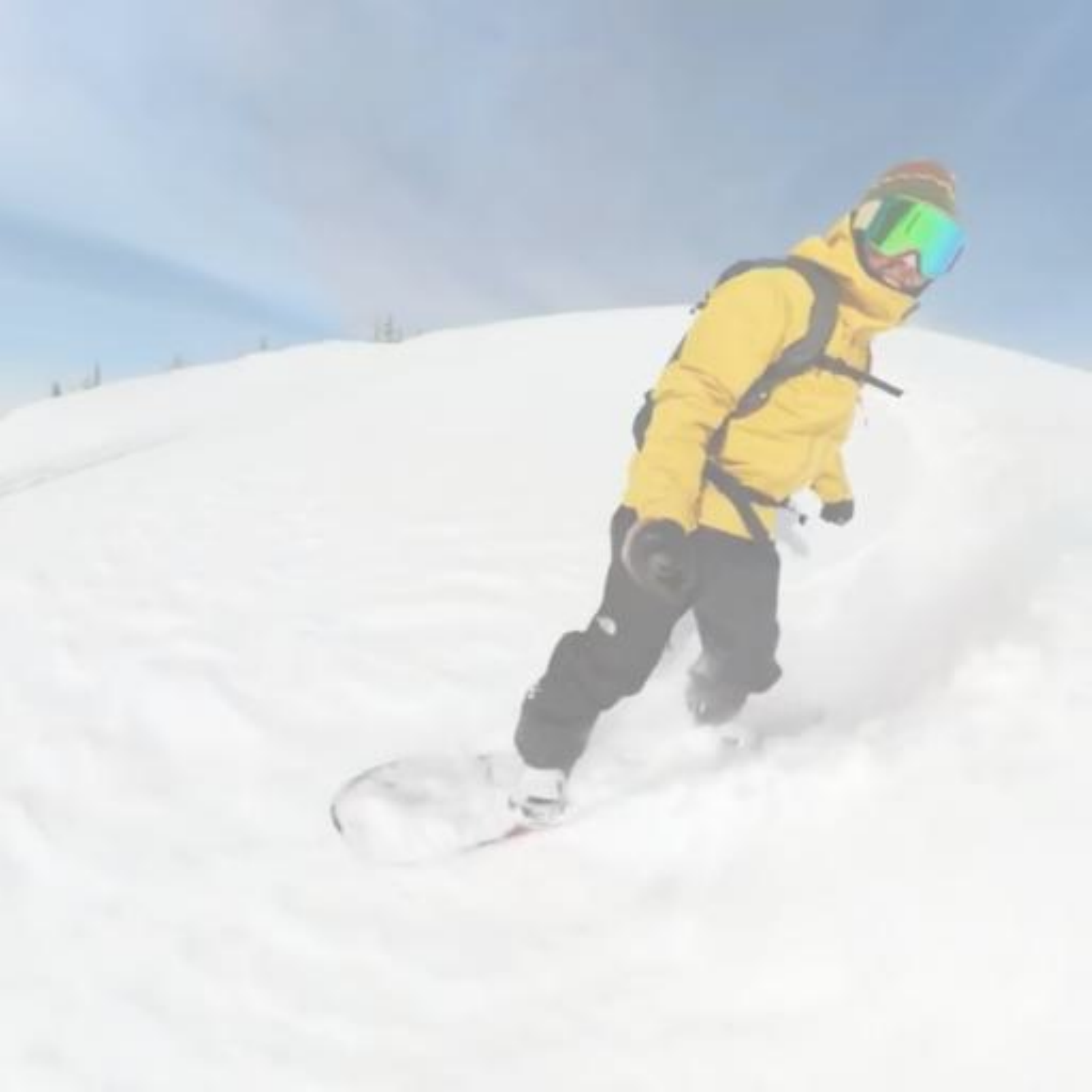}
\includegraphics[width=0.20\textwidth]{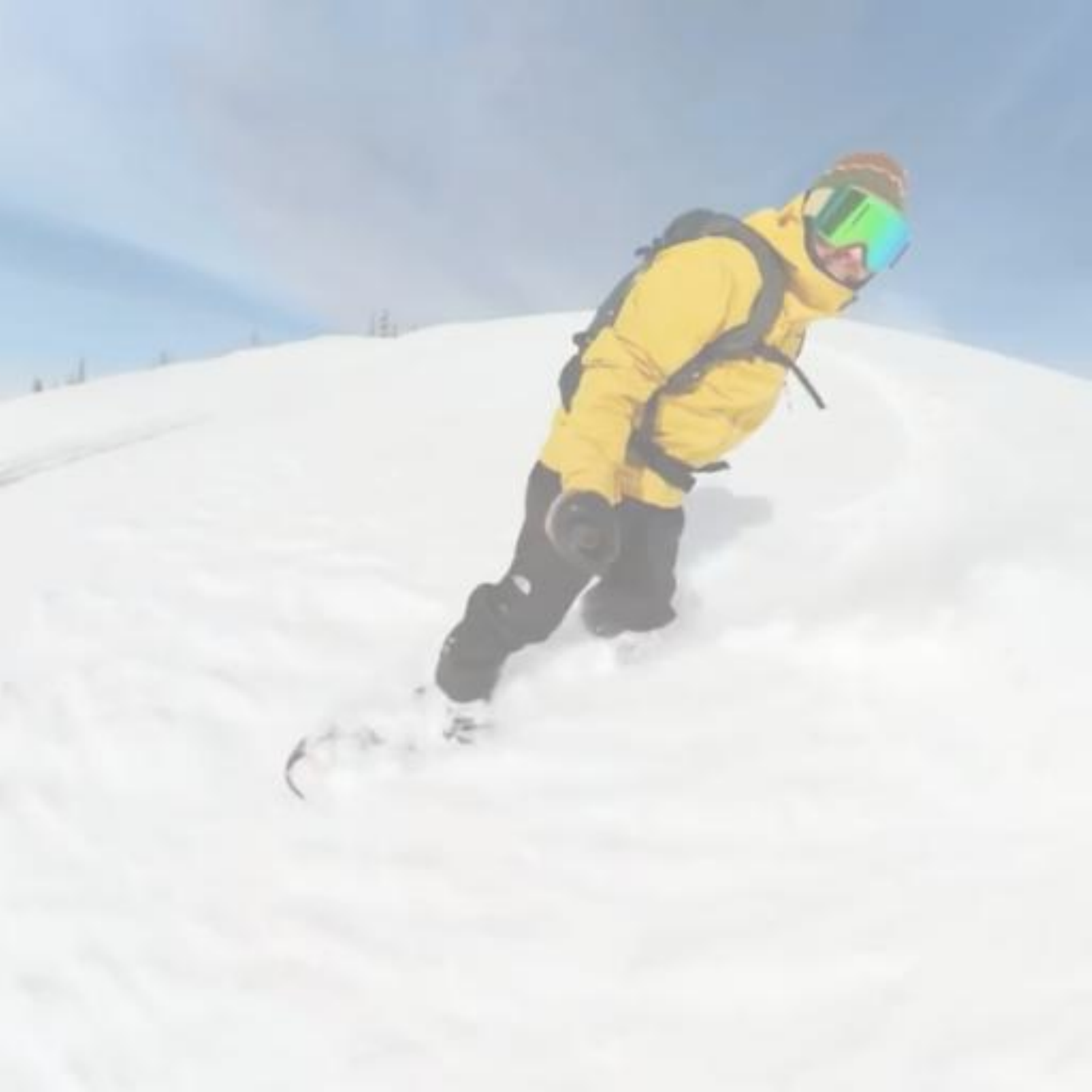}

\rotatebox{90}{\parbox{0.20\textwidth}{\centering ~ \\ ~}}
\includegraphics[width=0.20\textwidth]{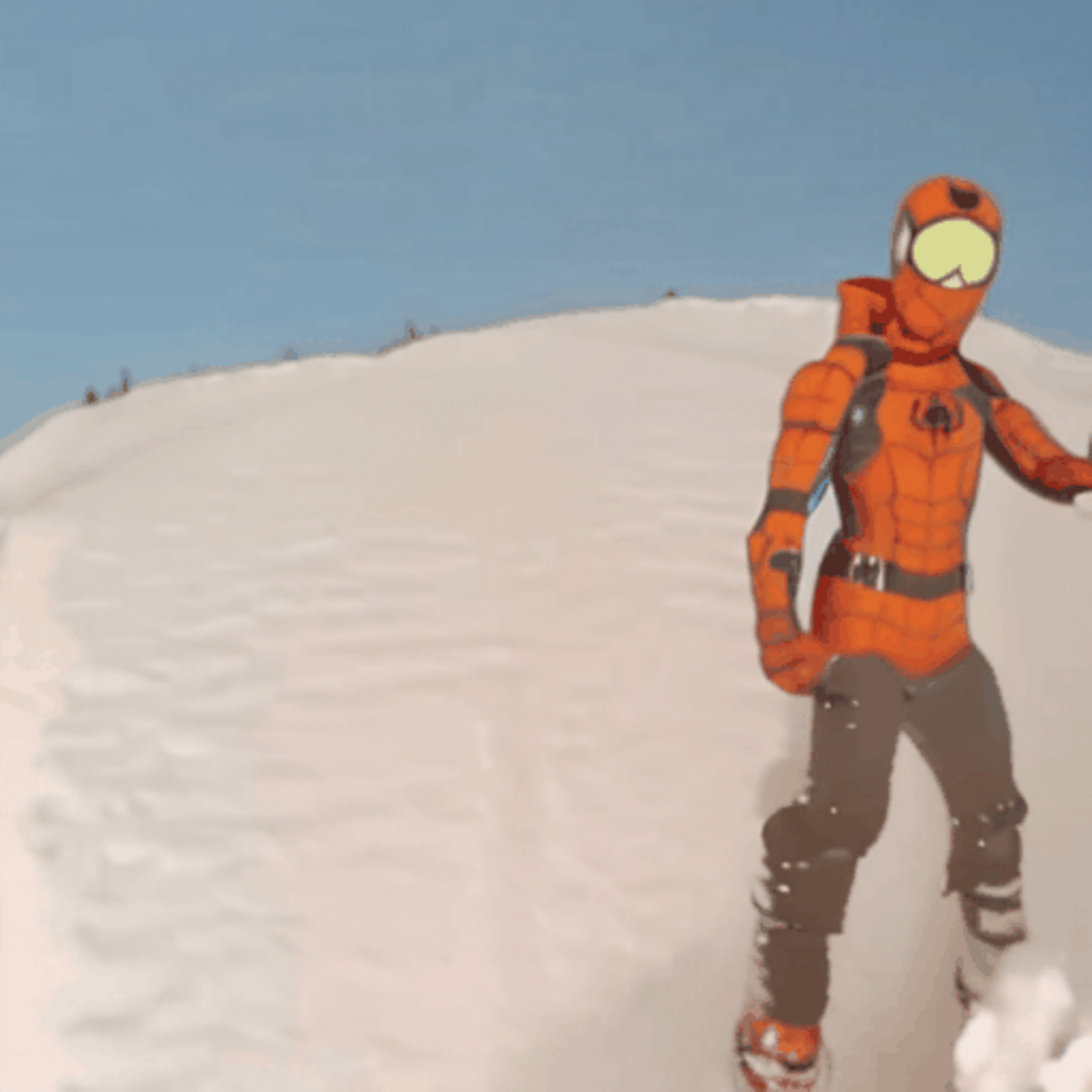}
\includegraphics[width=0.20\textwidth]{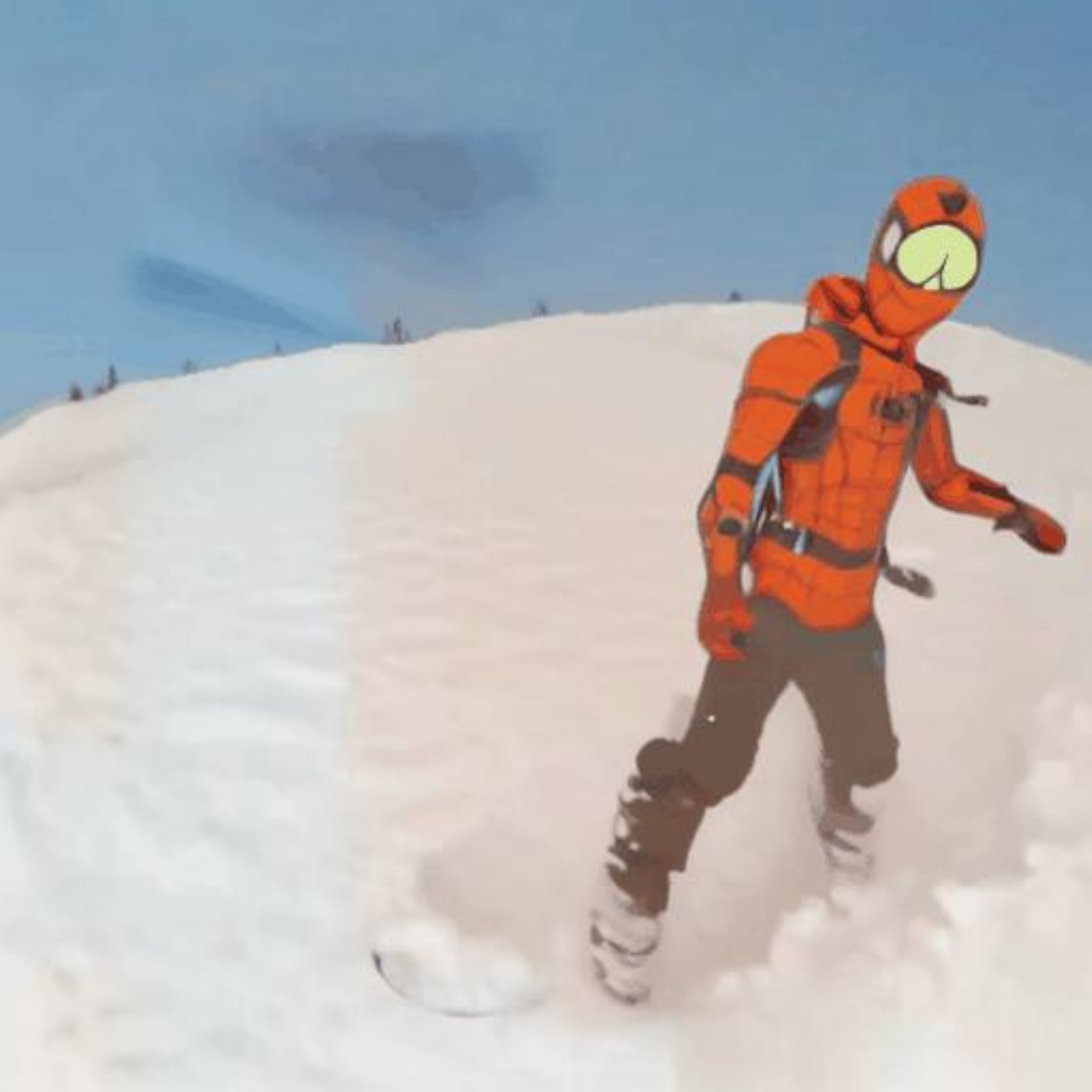}
\includegraphics[width=0.20\textwidth]{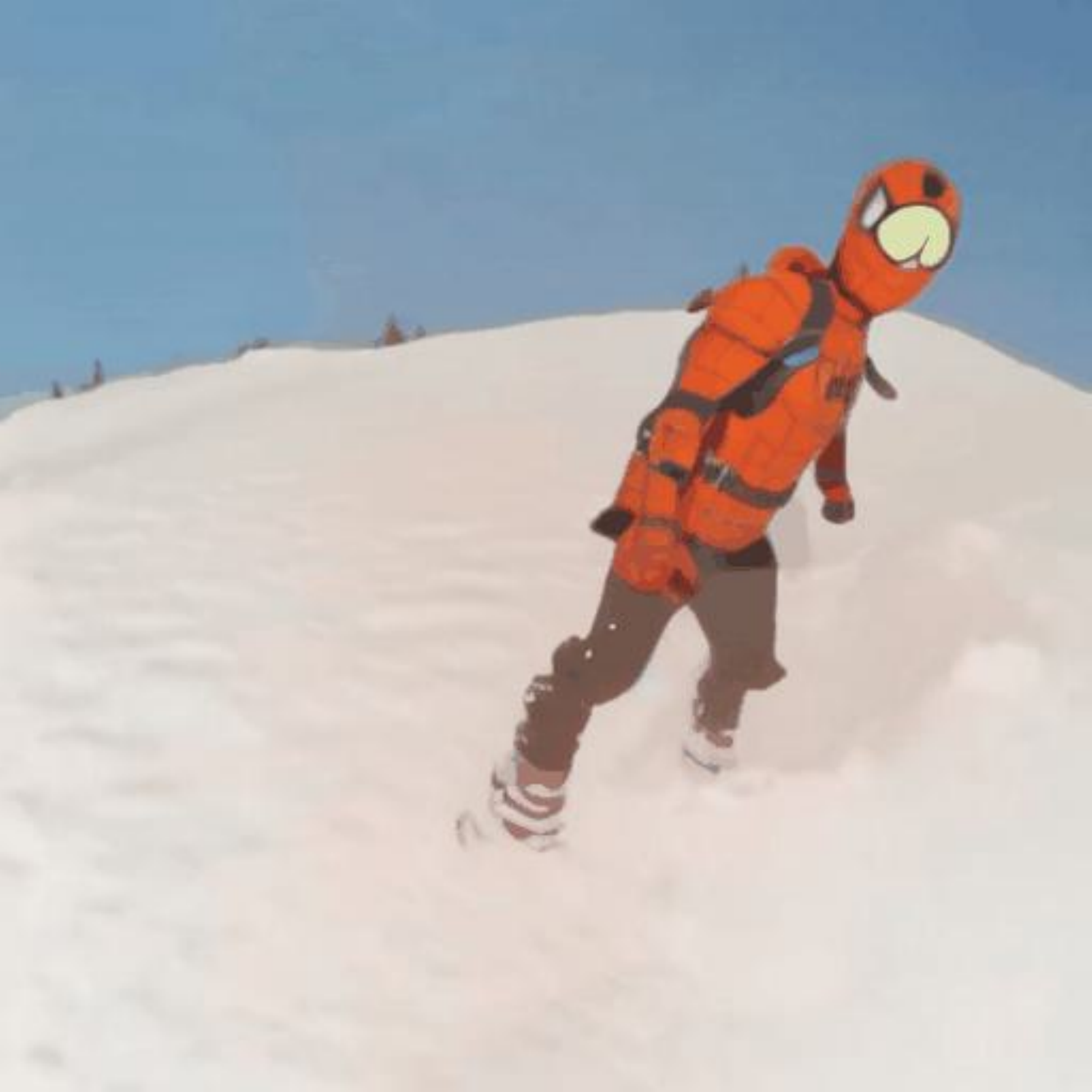}
\includegraphics[width=0.20\textwidth]{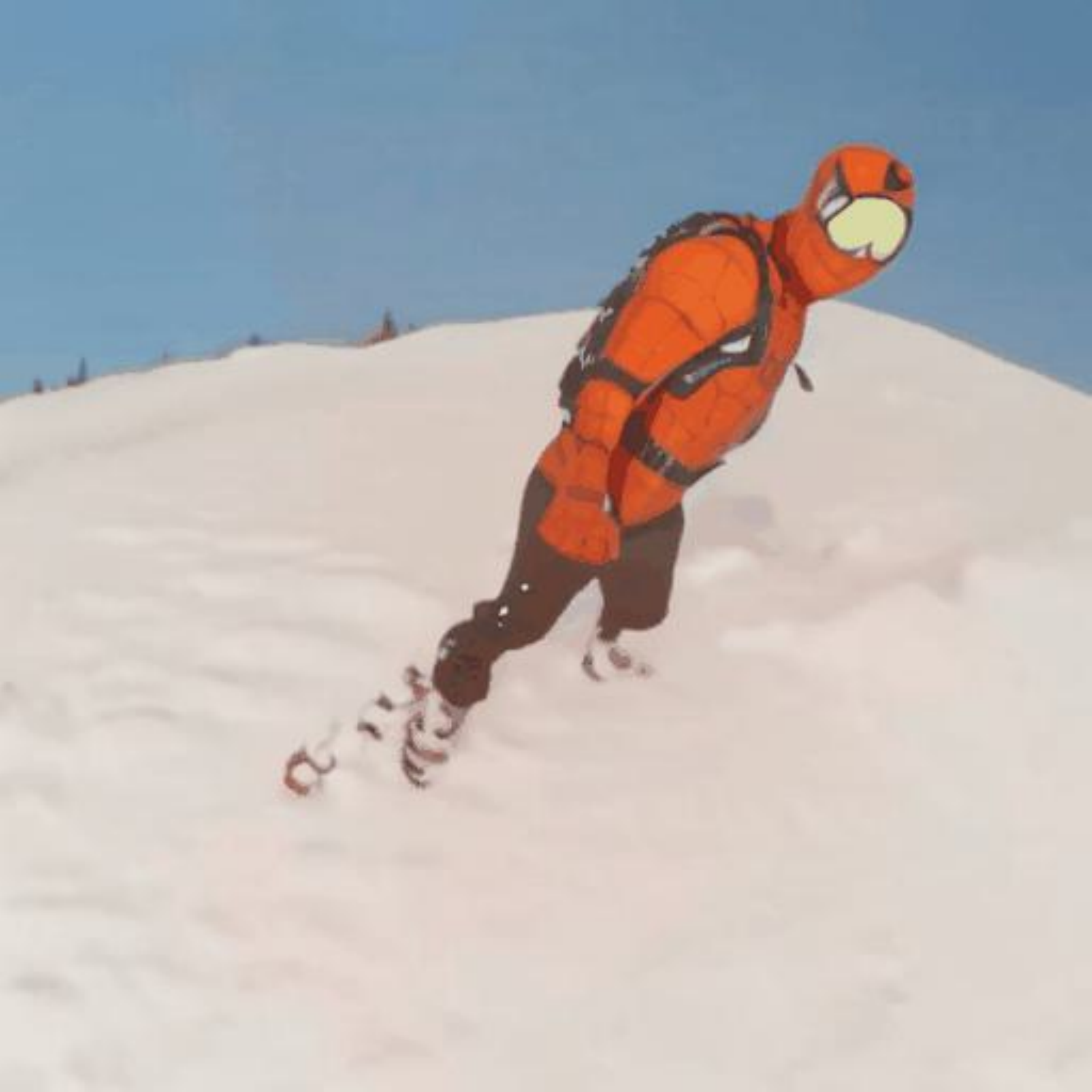}

\rotatebox{90}{\parbox{0.20\textwidth}{\centering \textbf{threshold \\ 0.25}}}
\includegraphics[width=0.20\textwidth]{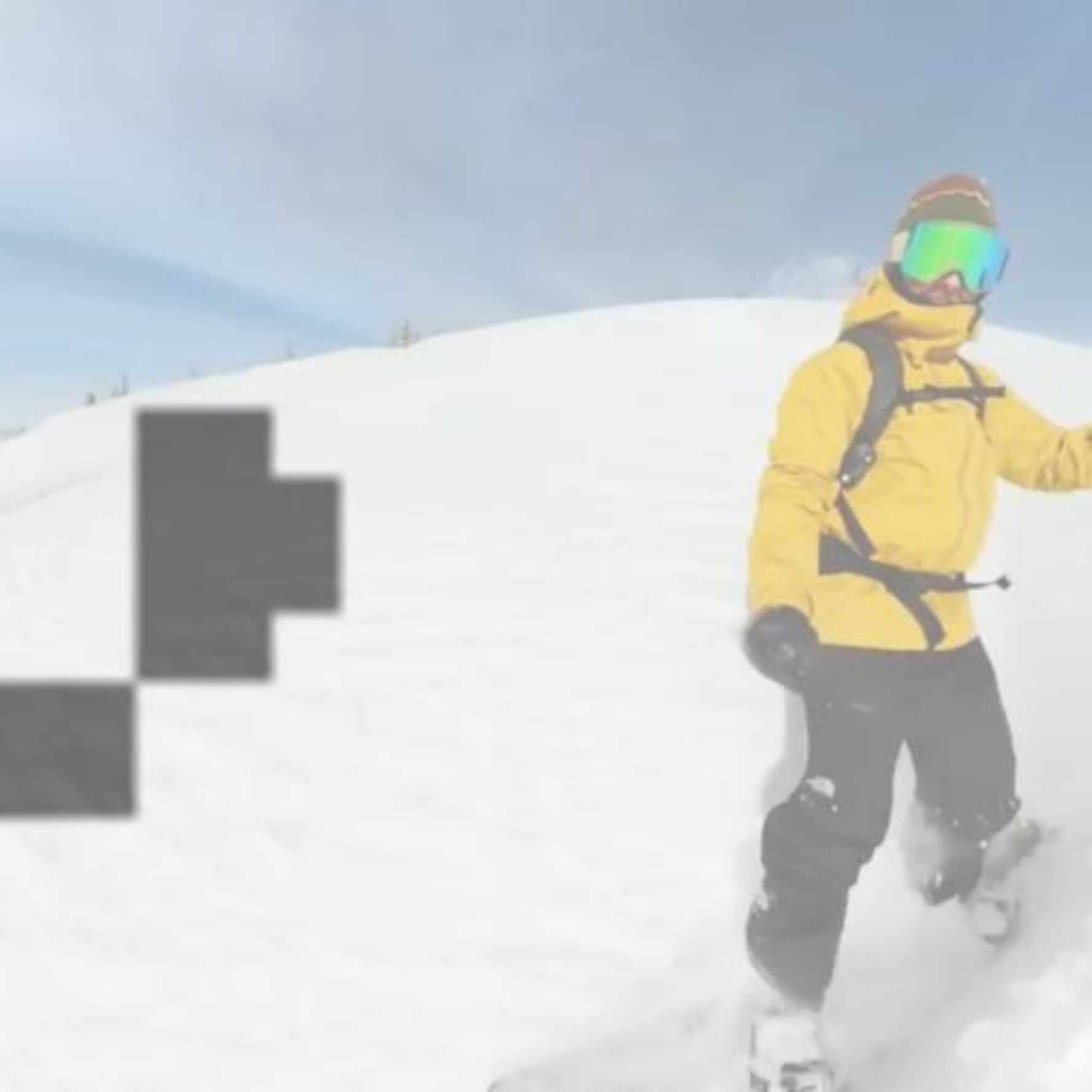}
\includegraphics[width=0.20\textwidth]{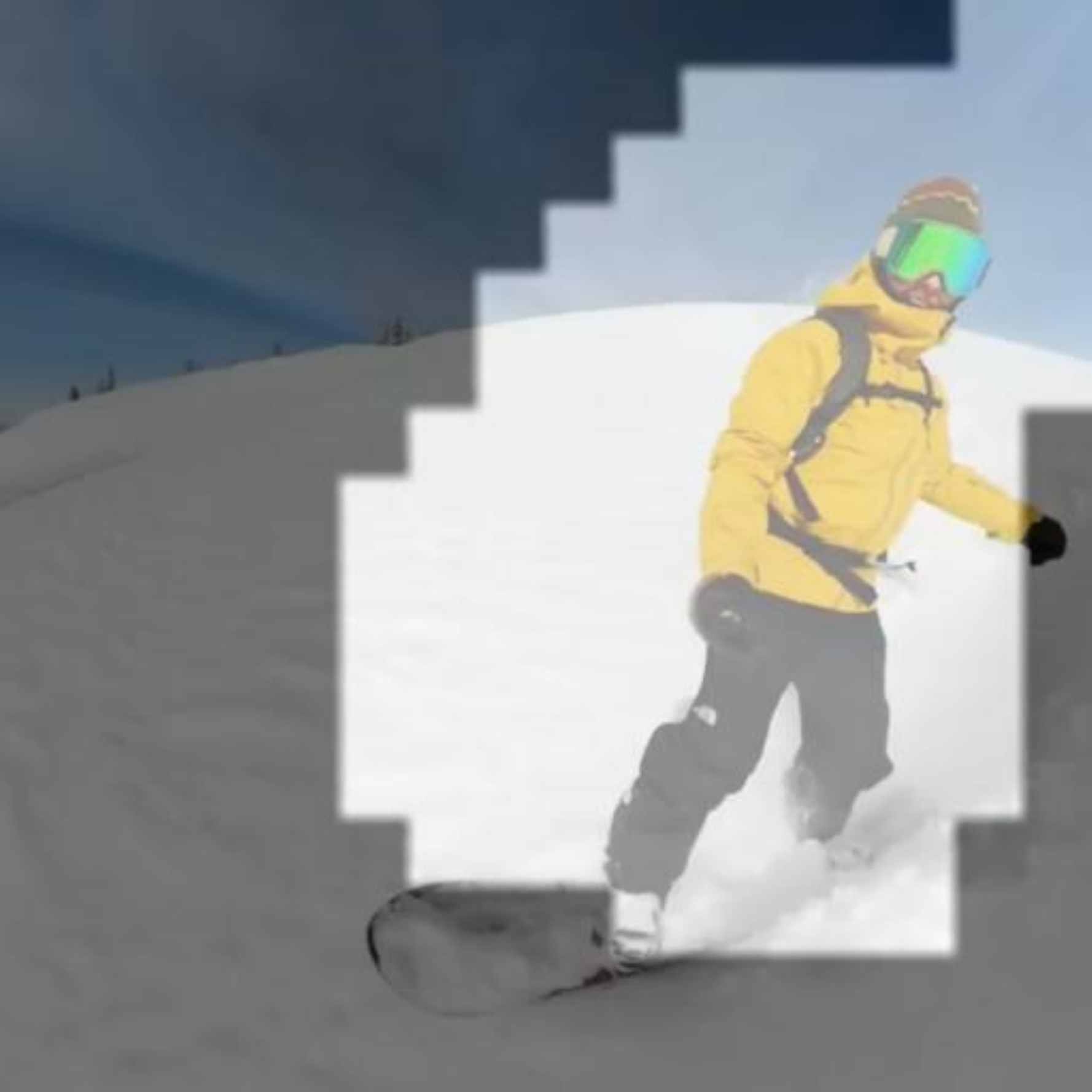}
\includegraphics[width=0.20\textwidth]{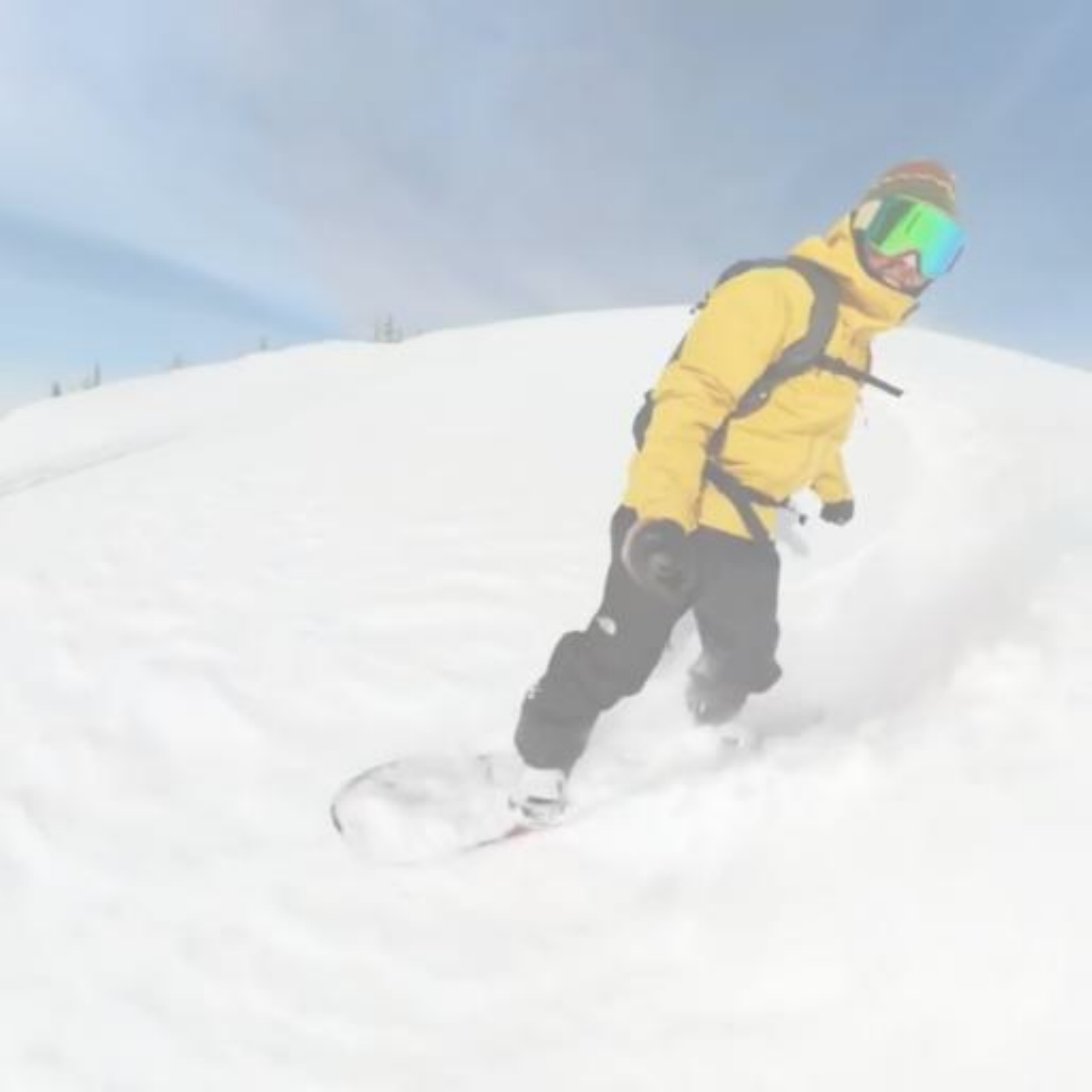}
\includegraphics[width=0.20\textwidth]{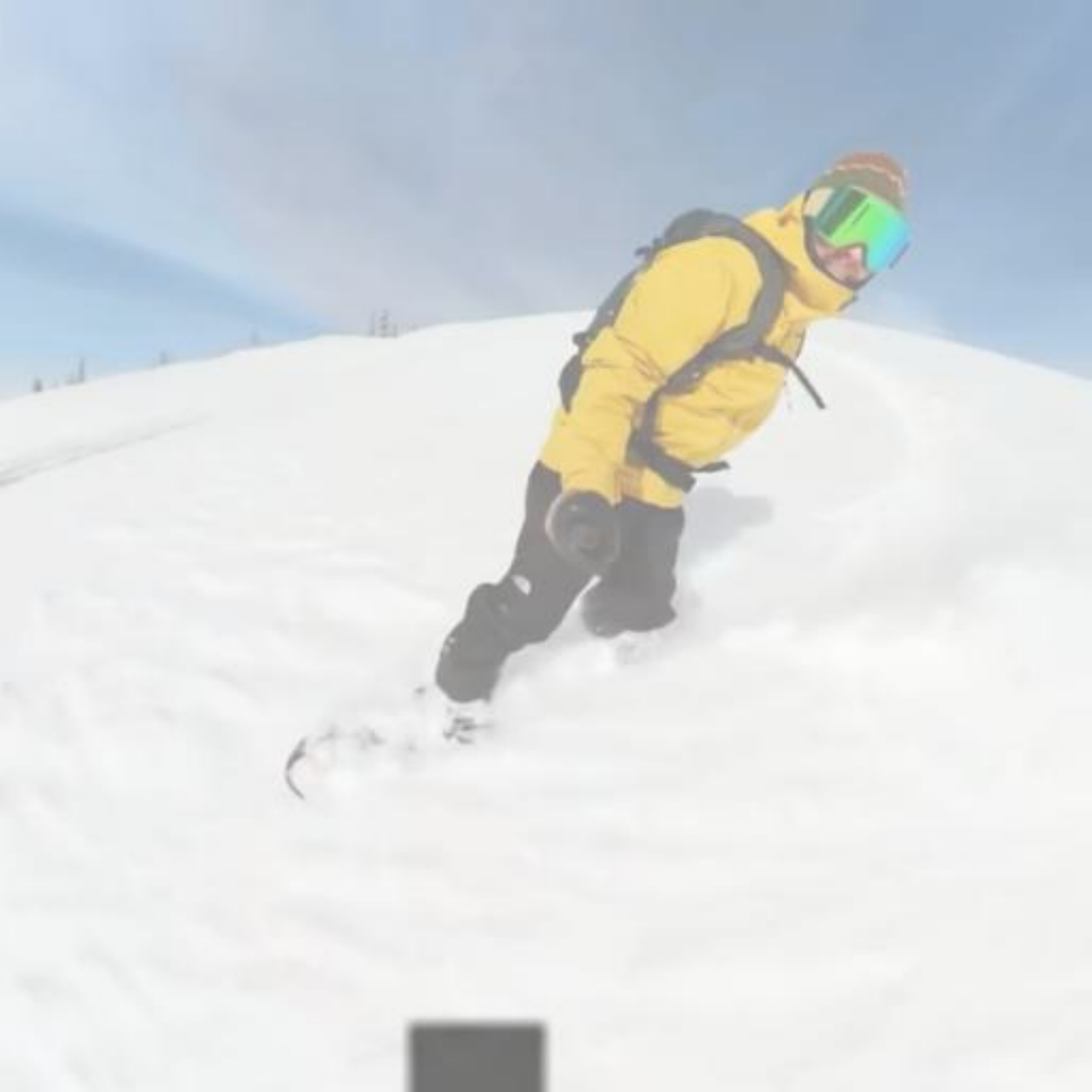}

\rotatebox{90}{\parbox{0.20\textwidth}{\centering ~ \\ ~}}
\includegraphics[width=0.20\textwidth]{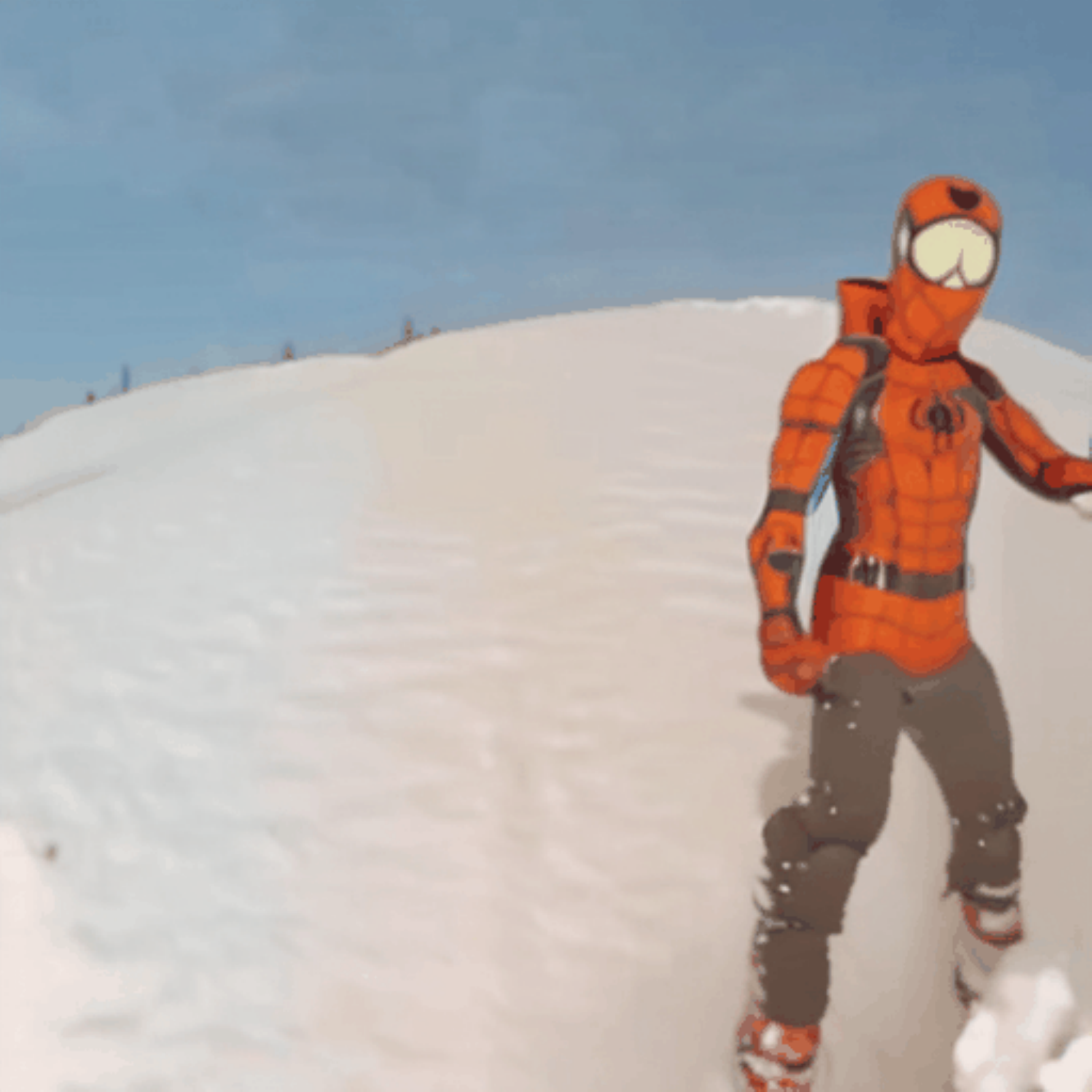}
\includegraphics[width=0.20\textwidth]{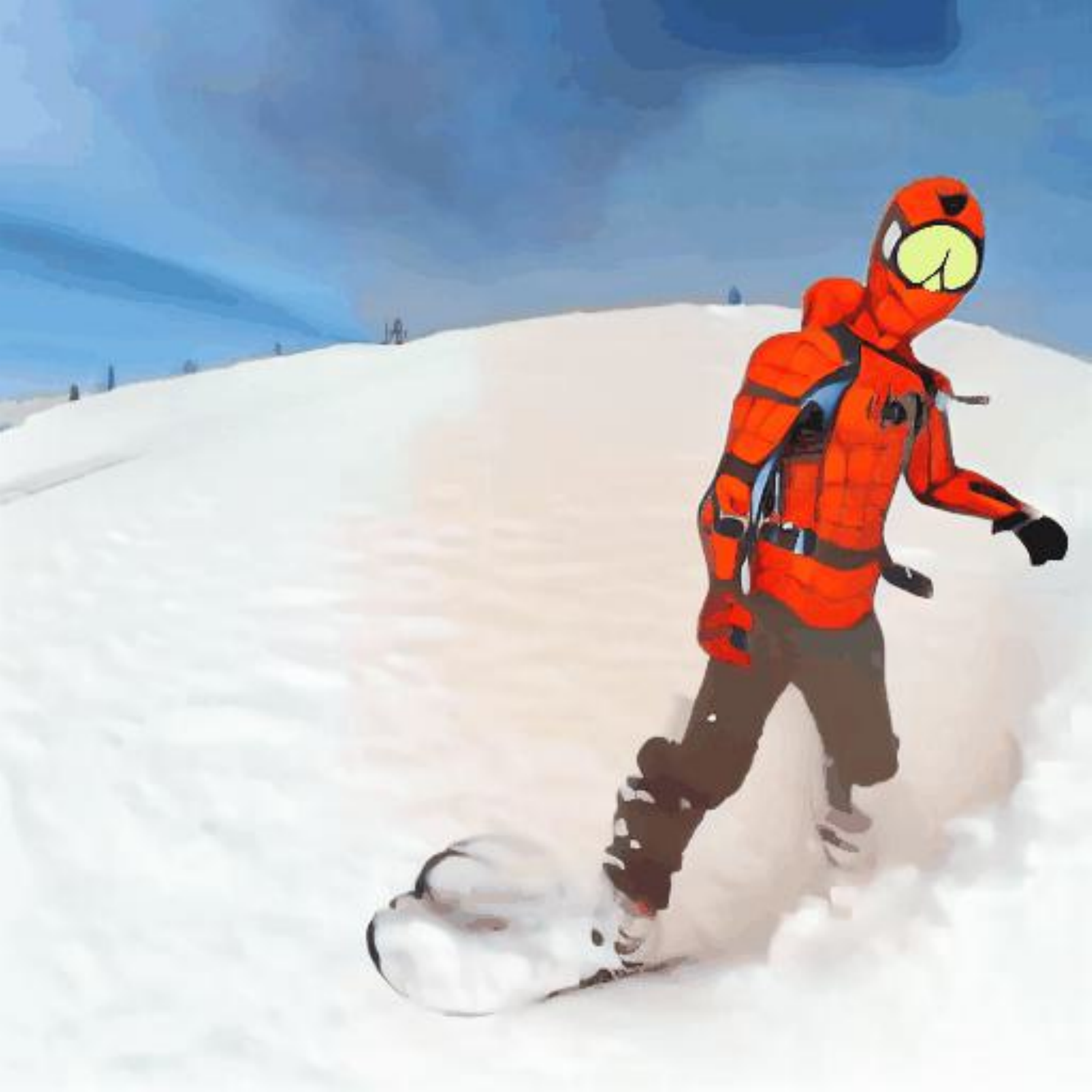}
\includegraphics[width=0.20\textwidth]{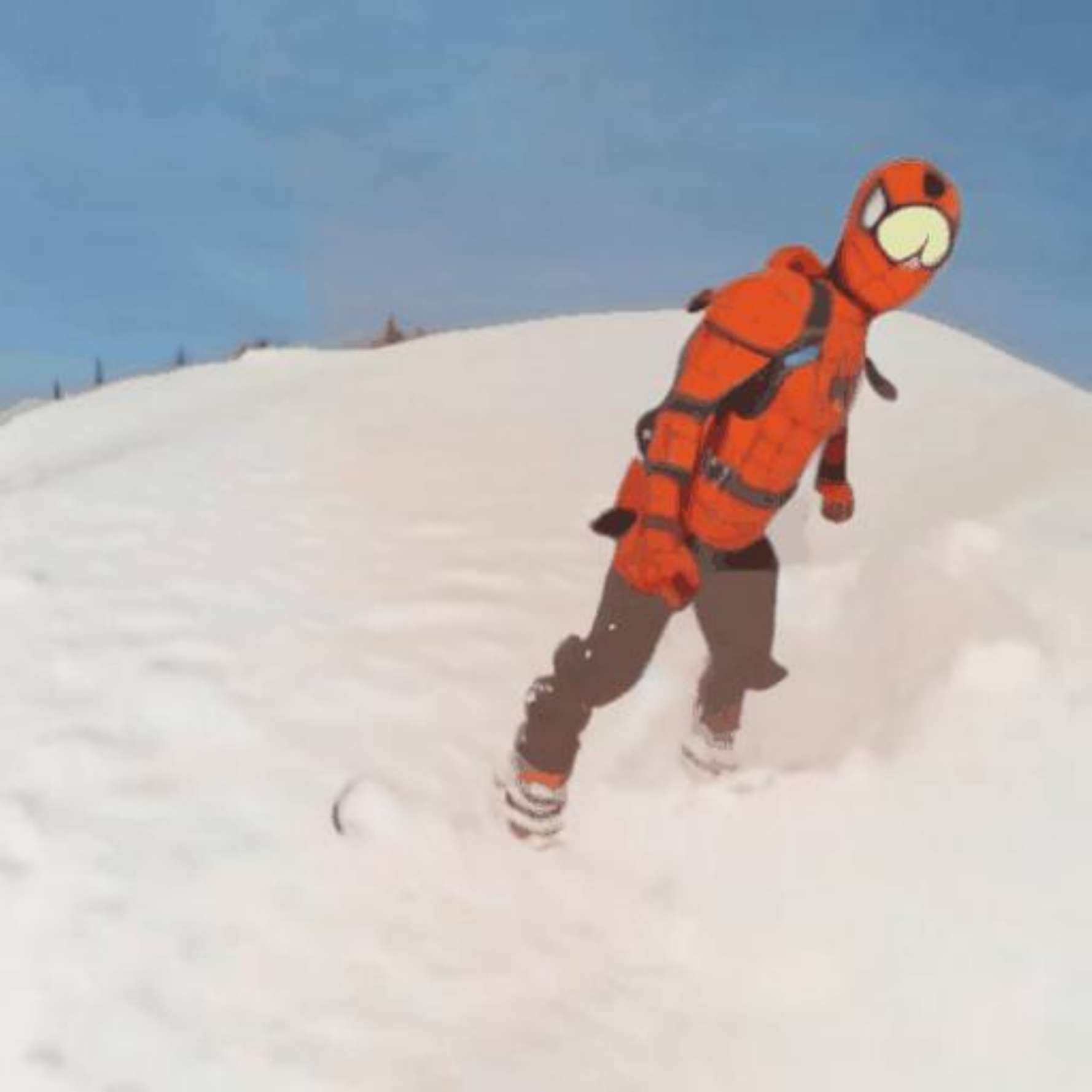}
\includegraphics[width=0.20\textwidth]{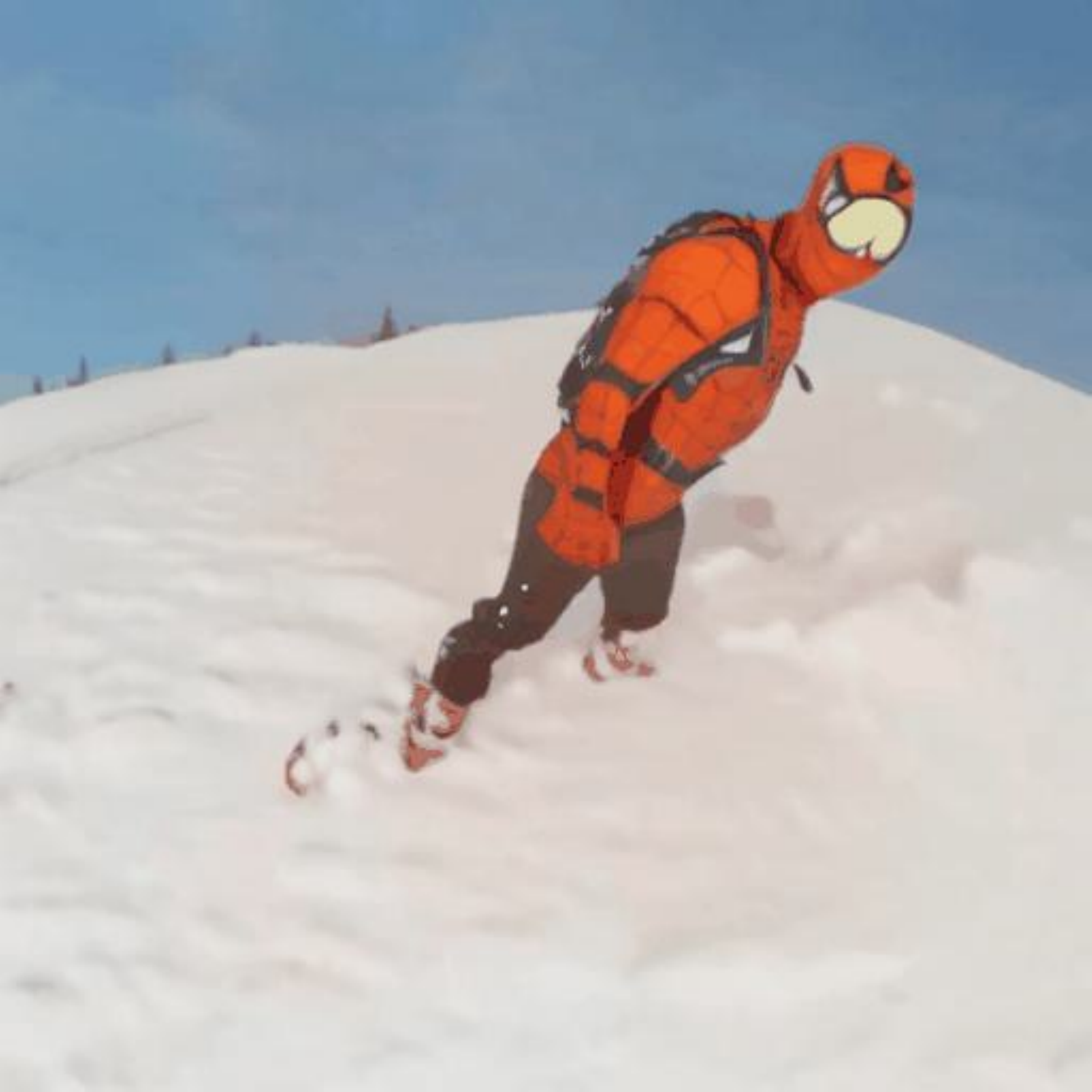}

\rotatebox{90}{\parbox{0.20\textwidth}{\centering threshold \\ 0.40}}
{
\includegraphics[width=0.20\textwidth]{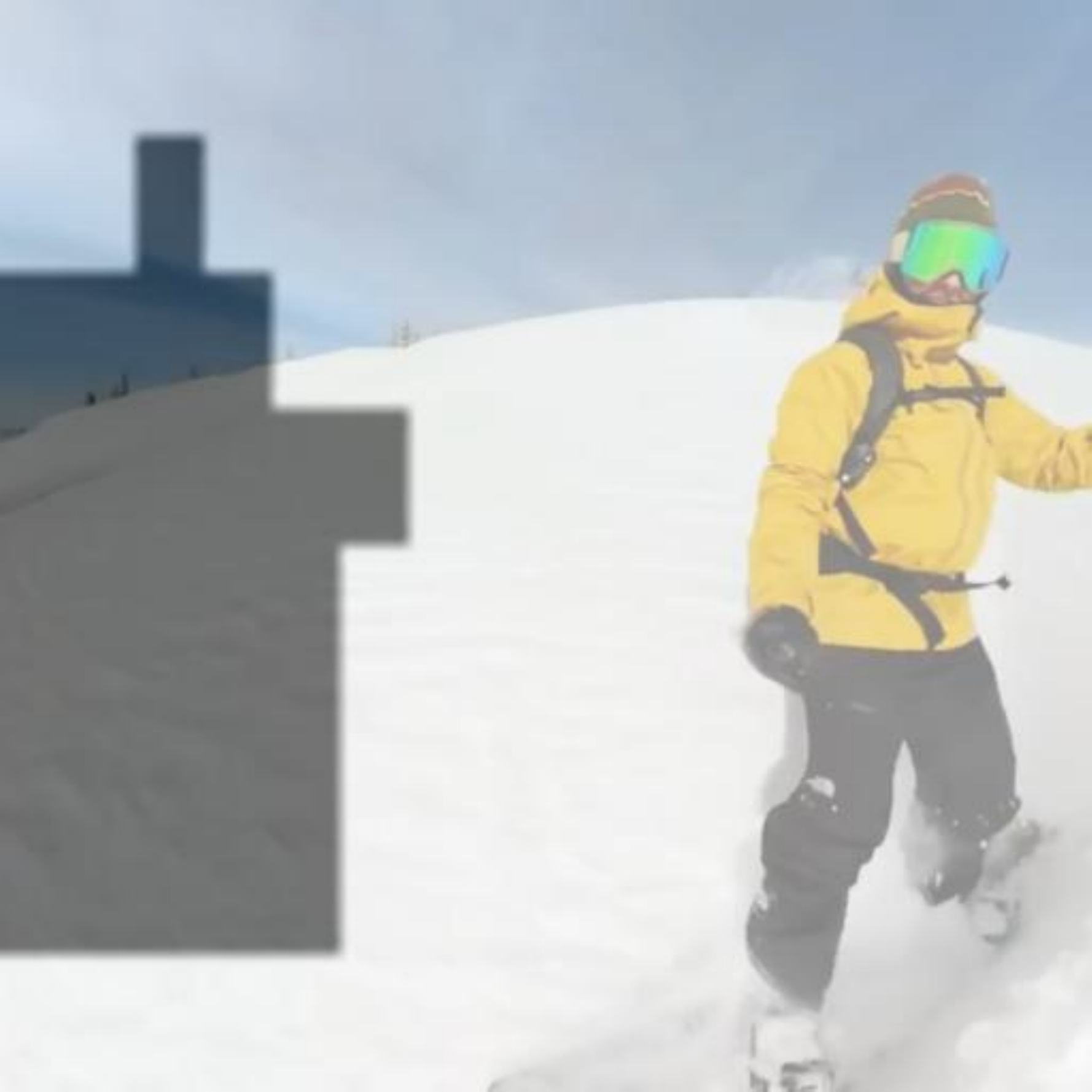}
\includegraphics[width=0.20\textwidth]{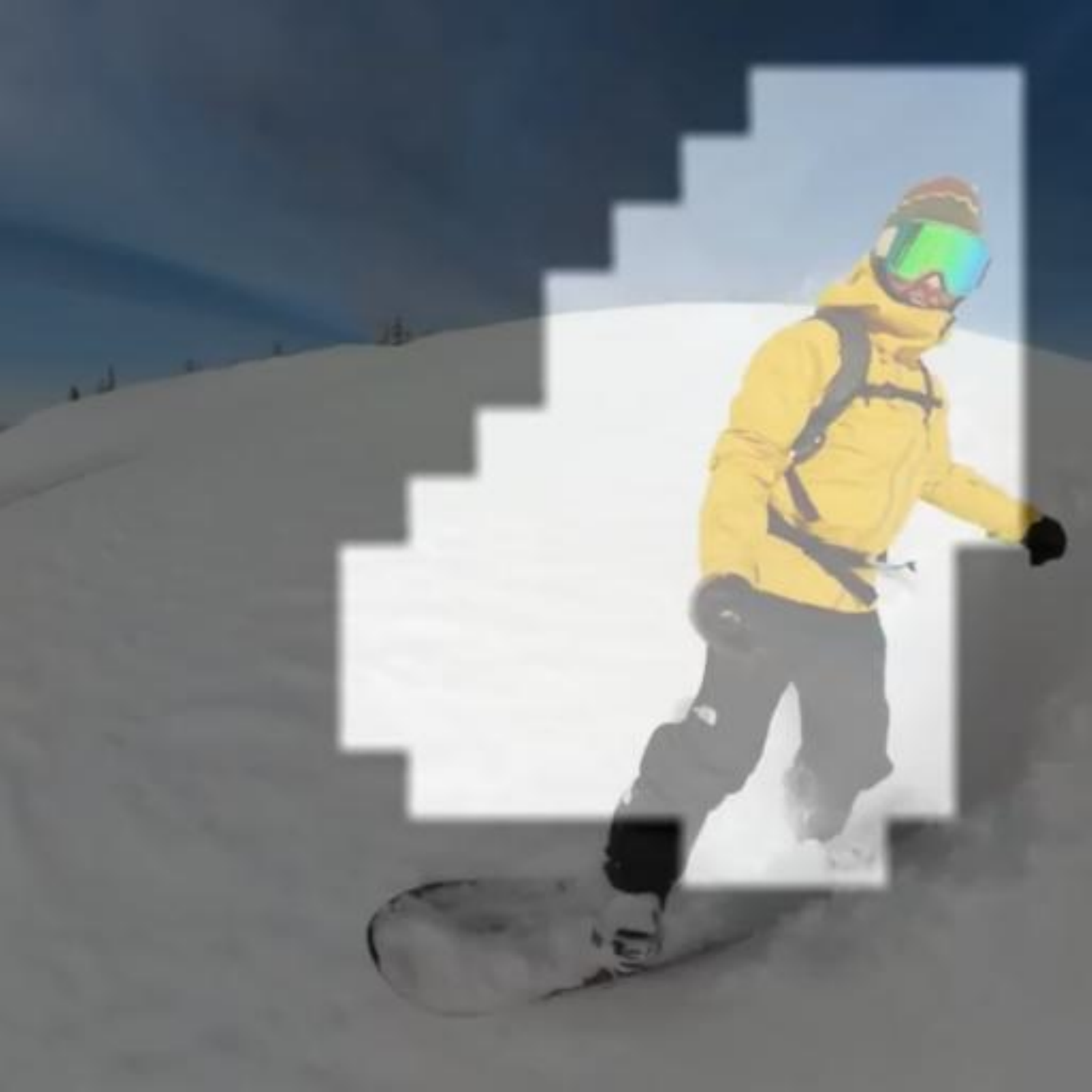}
\includegraphics[width=0.20\textwidth]{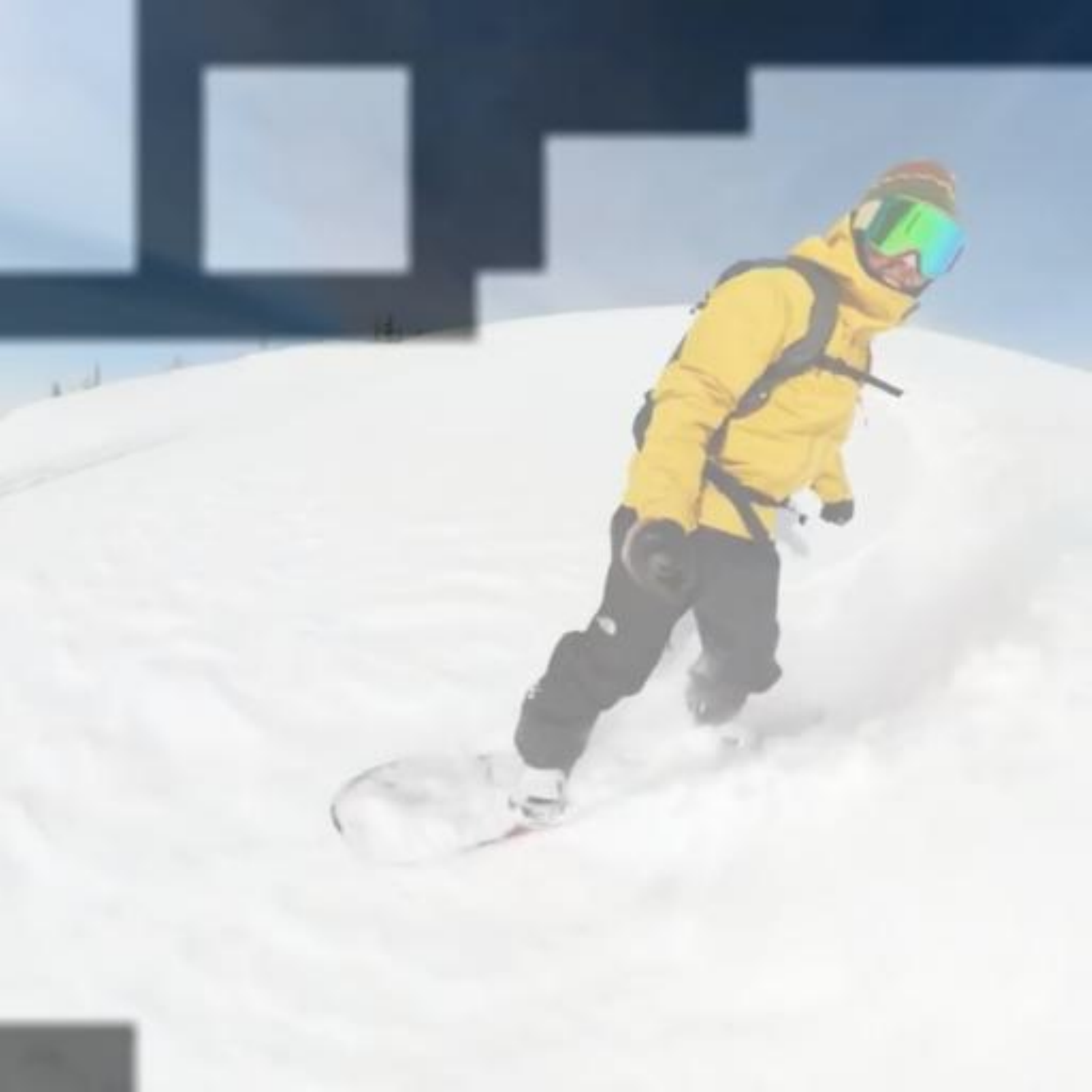}
\includegraphics[width=0.20\textwidth]{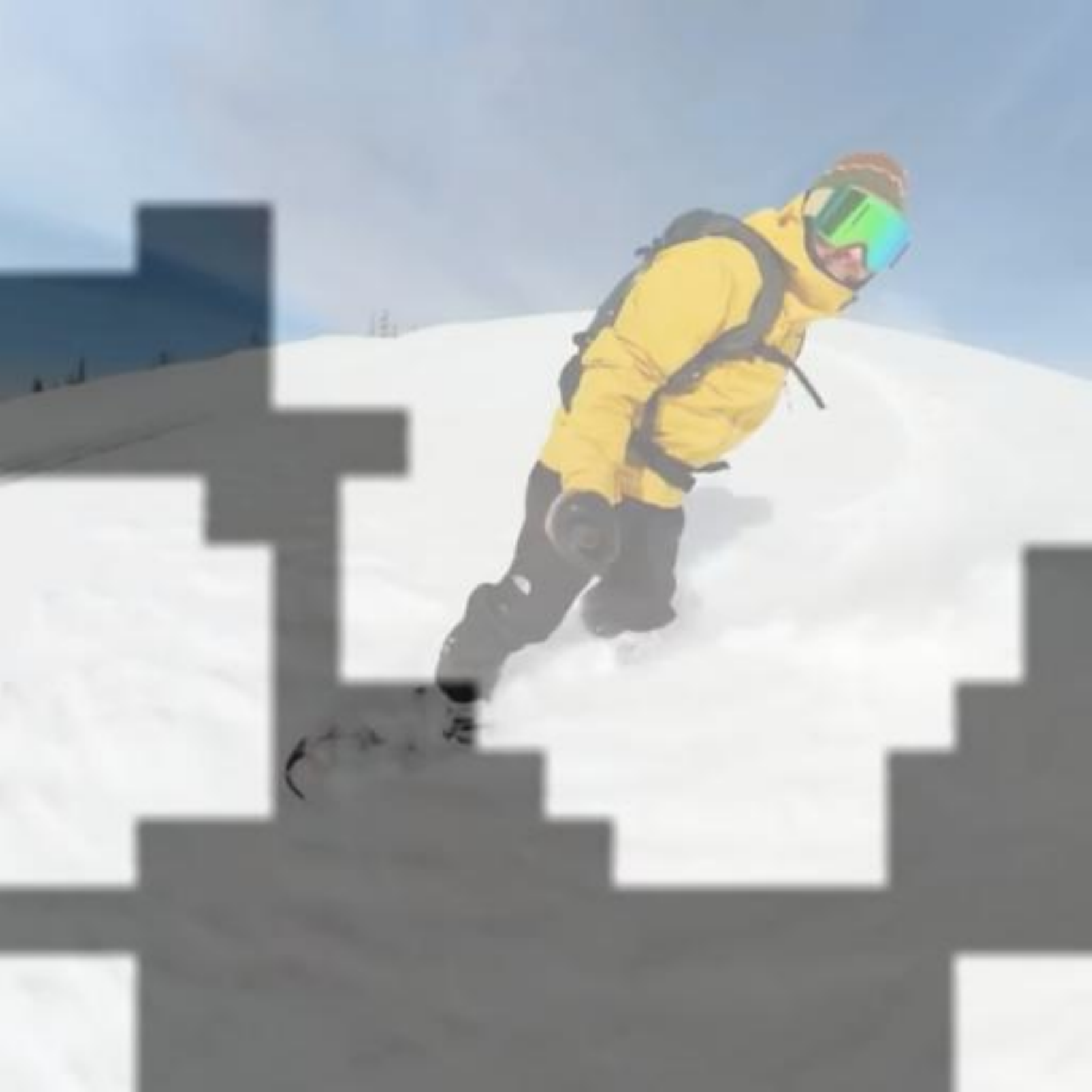}

\rotatebox{90}{\parbox{0.20\textwidth}{\centering ~ \\ ~}}
\includegraphics[width=0.20\textwidth]{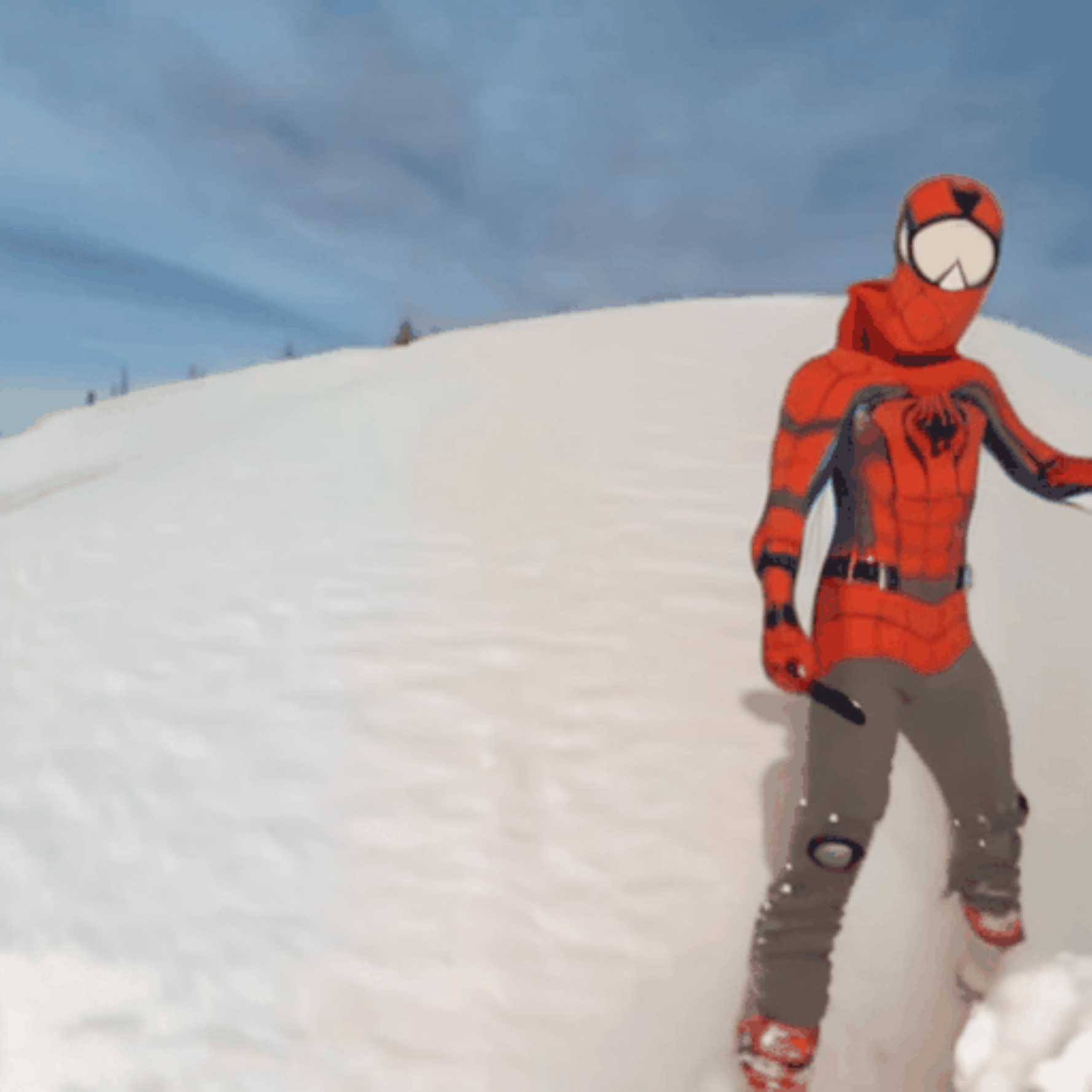}
\includegraphics[width=0.20\textwidth]{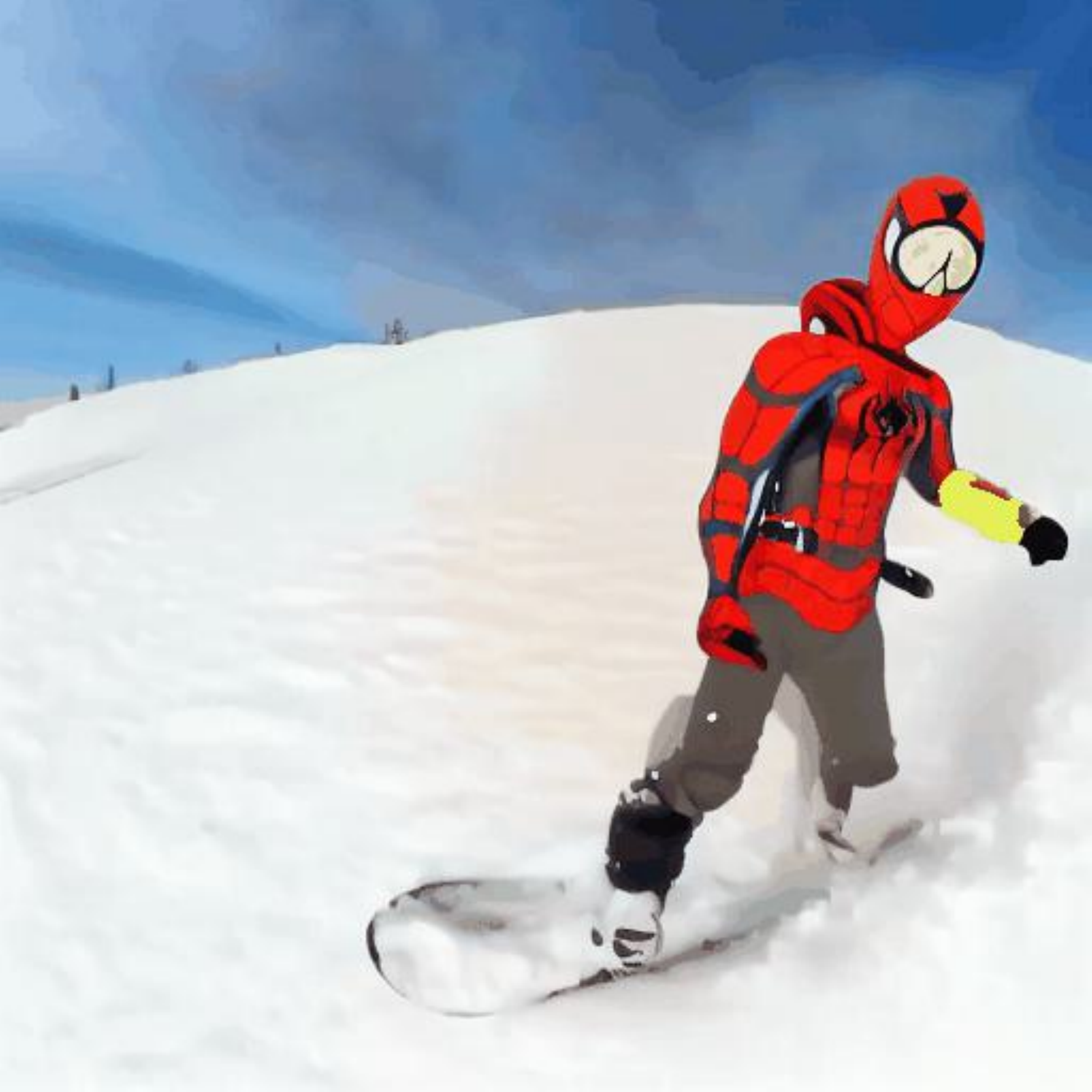}
\includegraphics[width=0.20\textwidth]{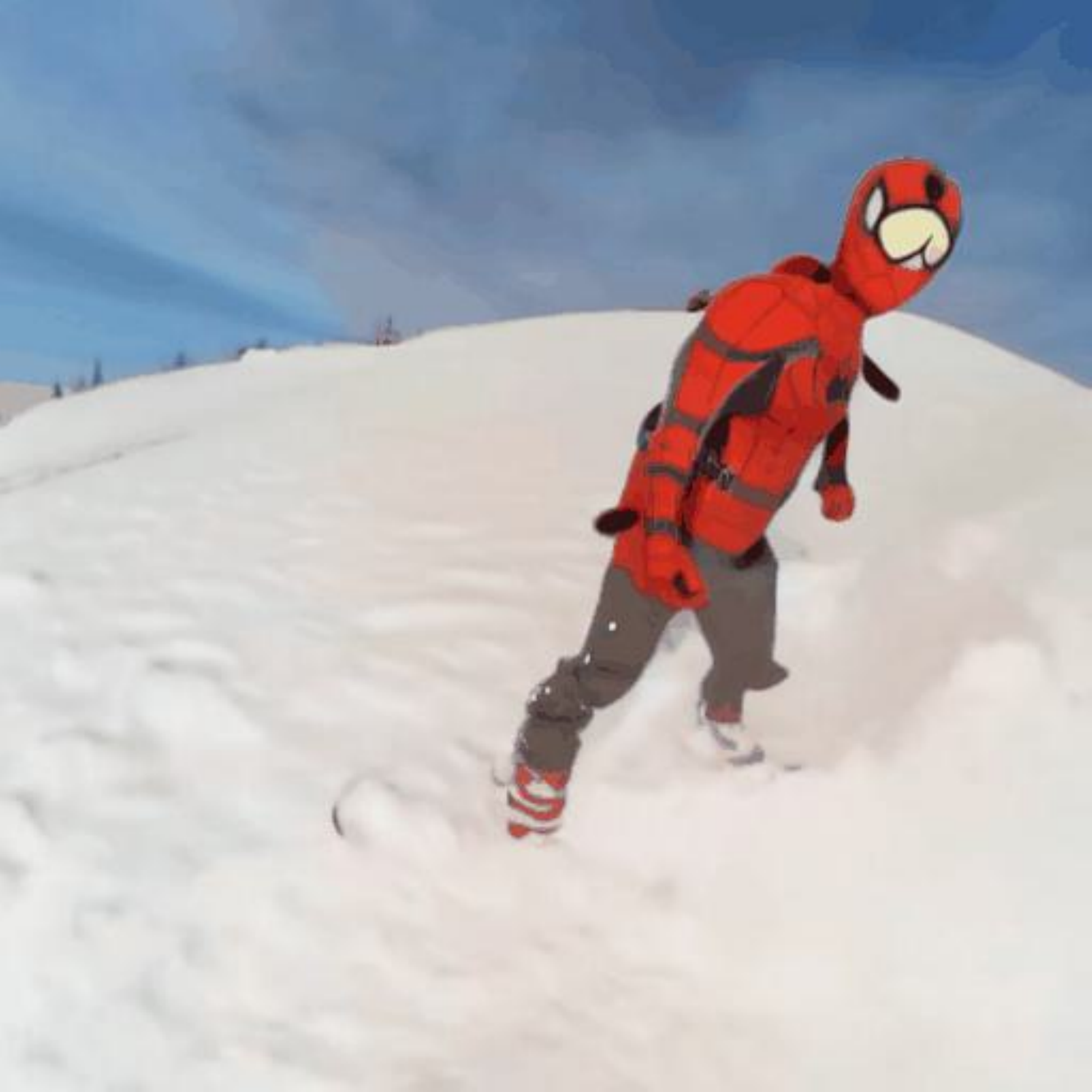}
\includegraphics[width=0.20\textwidth]{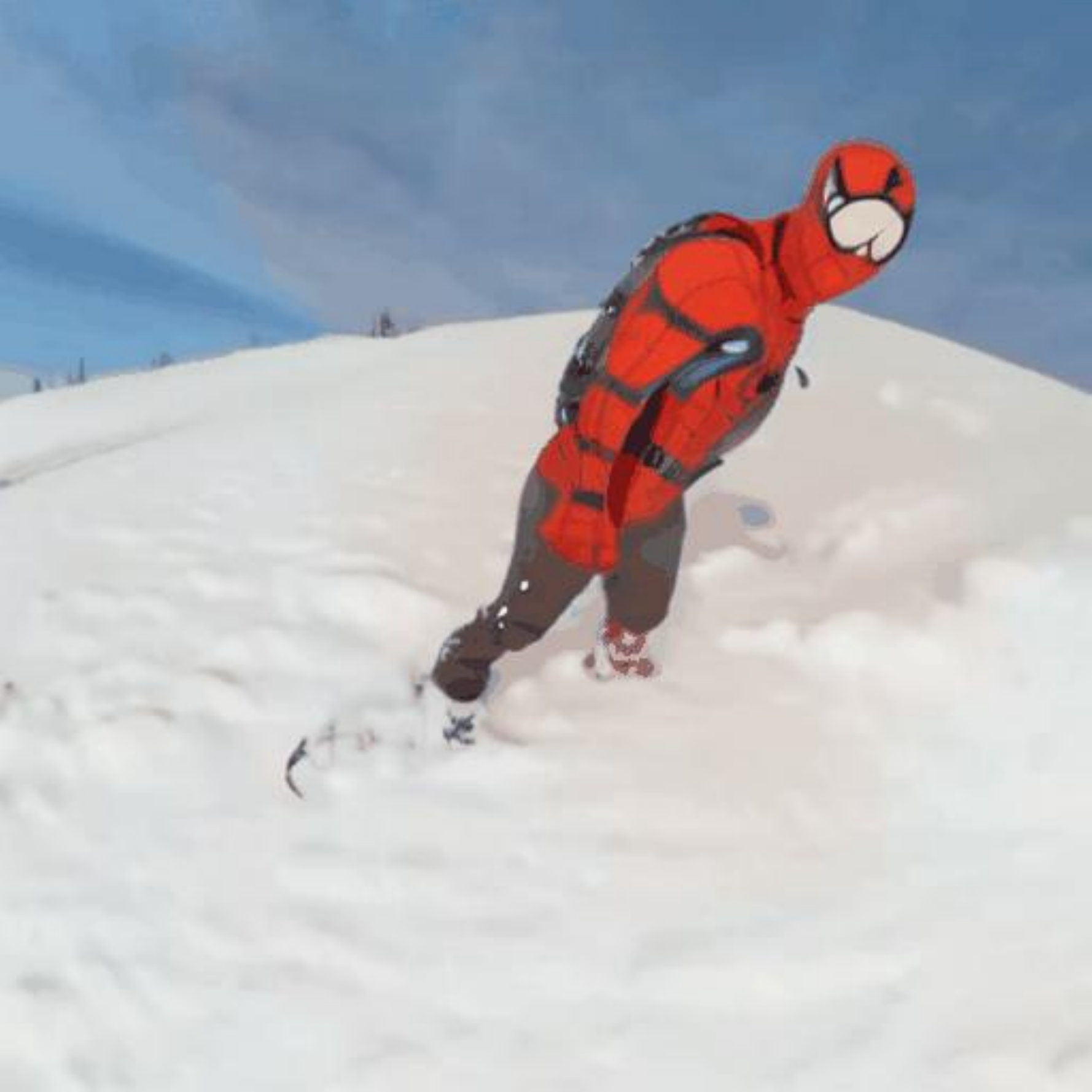}
}
\end{minipage}\hfill
\begin{minipage}{0.45\textwidth}
\centering
\makebox[0.12\textwidth]{\colorbox{green}{\textbf{Edit-A-Video}}}\\
\rotatebox{90}{\parbox{0.20\textwidth}{\centering threshold \\ 0.10}}
\includegraphics[width=0.20\textwidth]{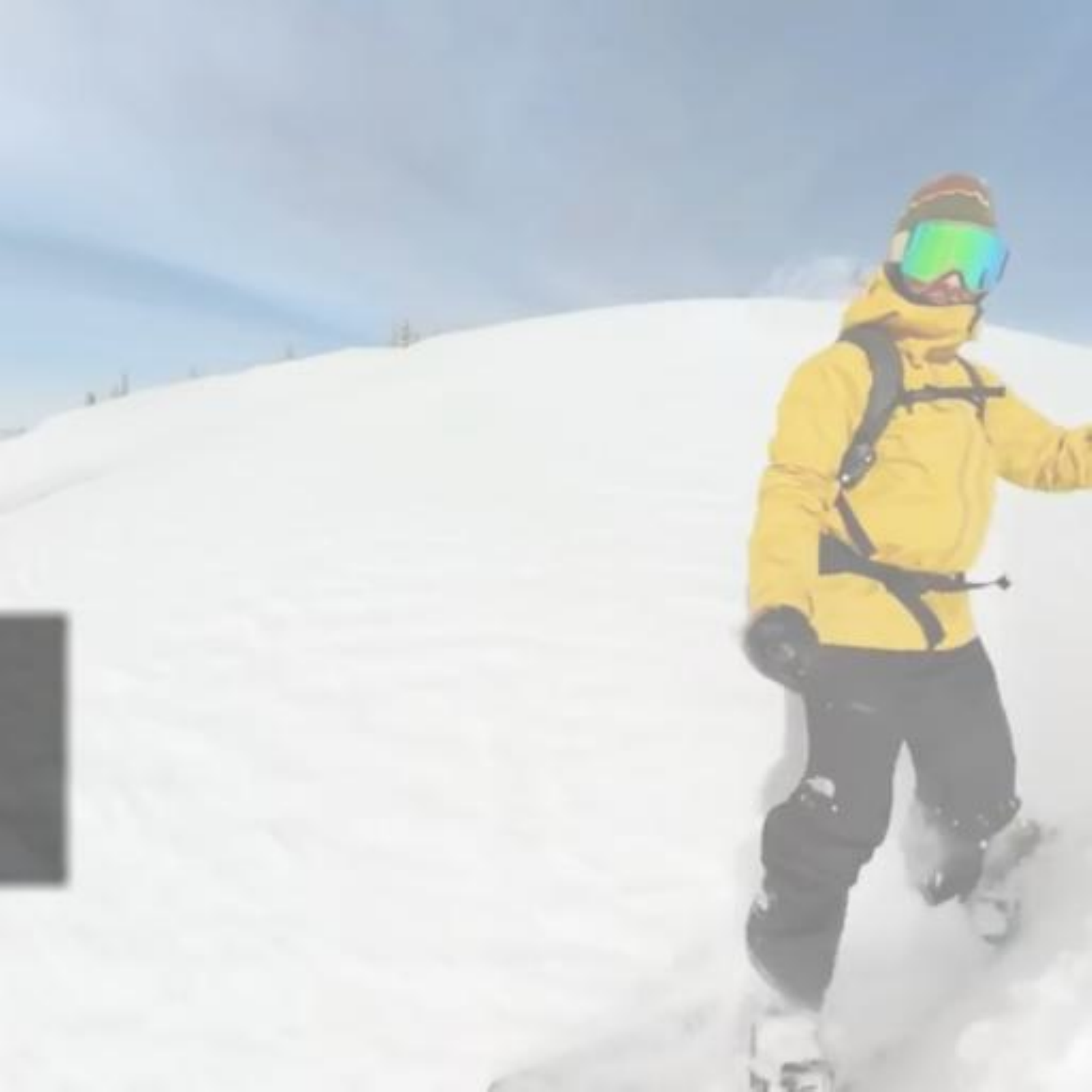}
\includegraphics[width=0.20\textwidth]{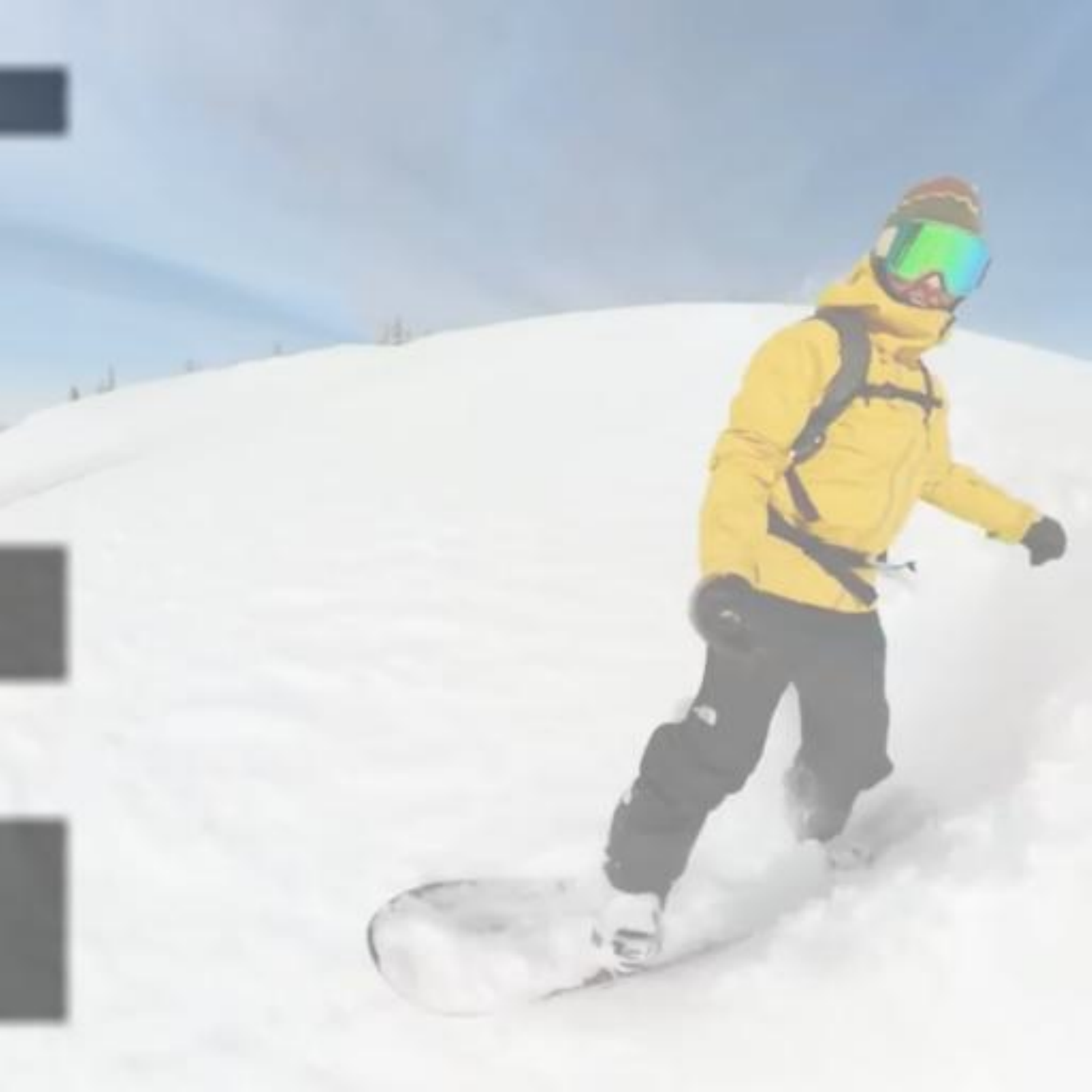}
\includegraphics[width=0.20\textwidth]{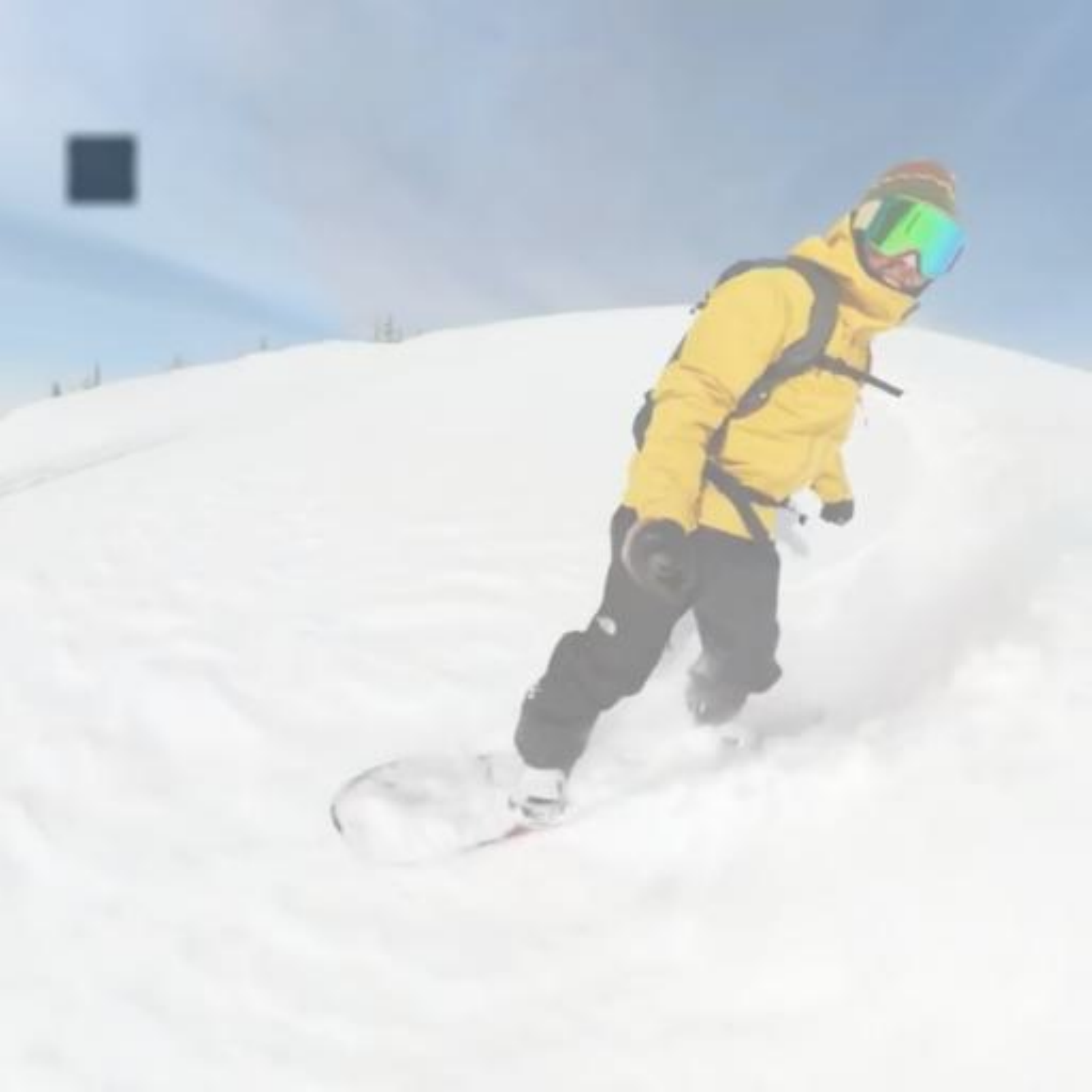}
\includegraphics[width=0.20\textwidth]{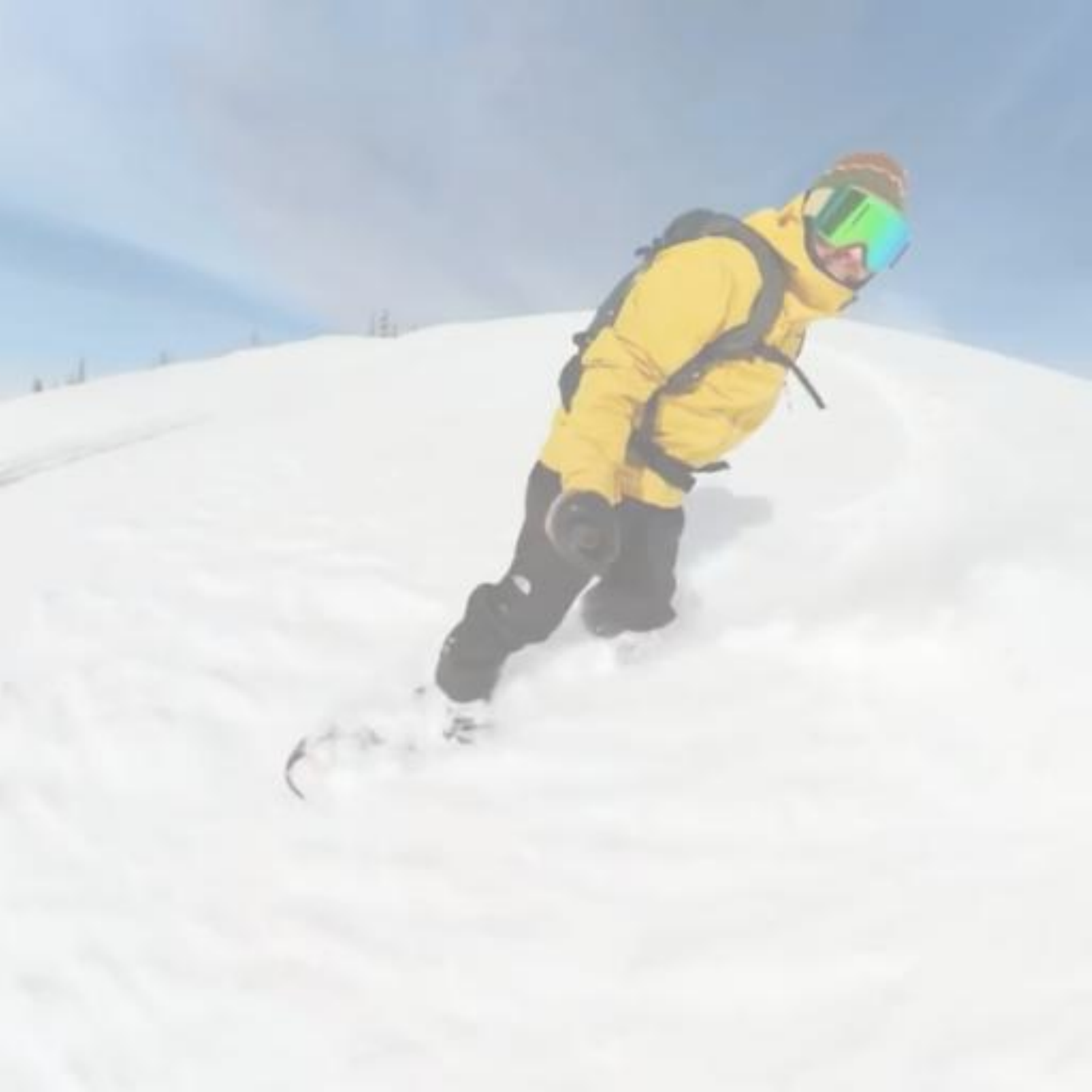}

\rotatebox{90}{\parbox{0.20\textwidth}{\centering ~ \\ ~}}
\includegraphics[width=0.20\textwidth]{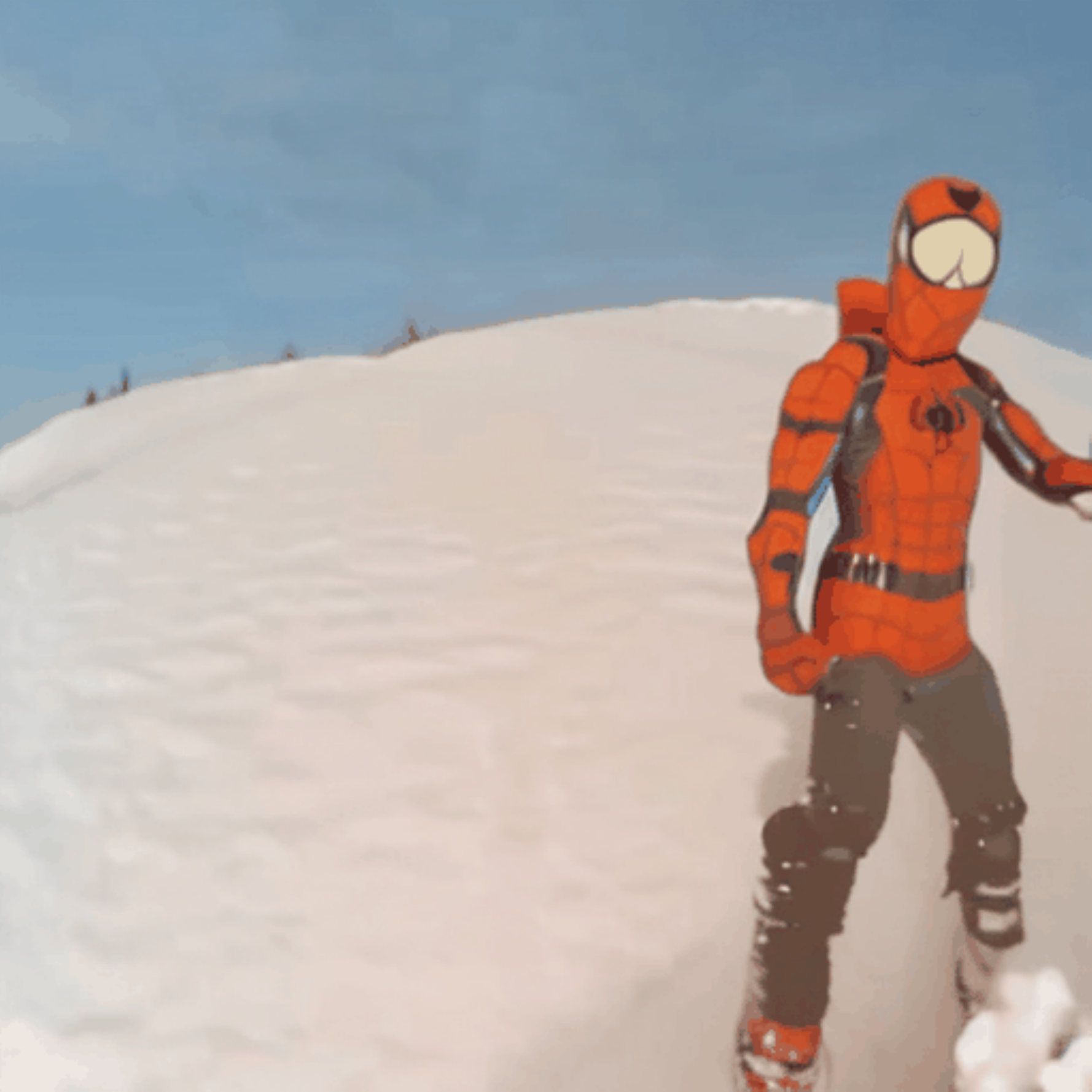}
\includegraphics[width=0.20\textwidth]{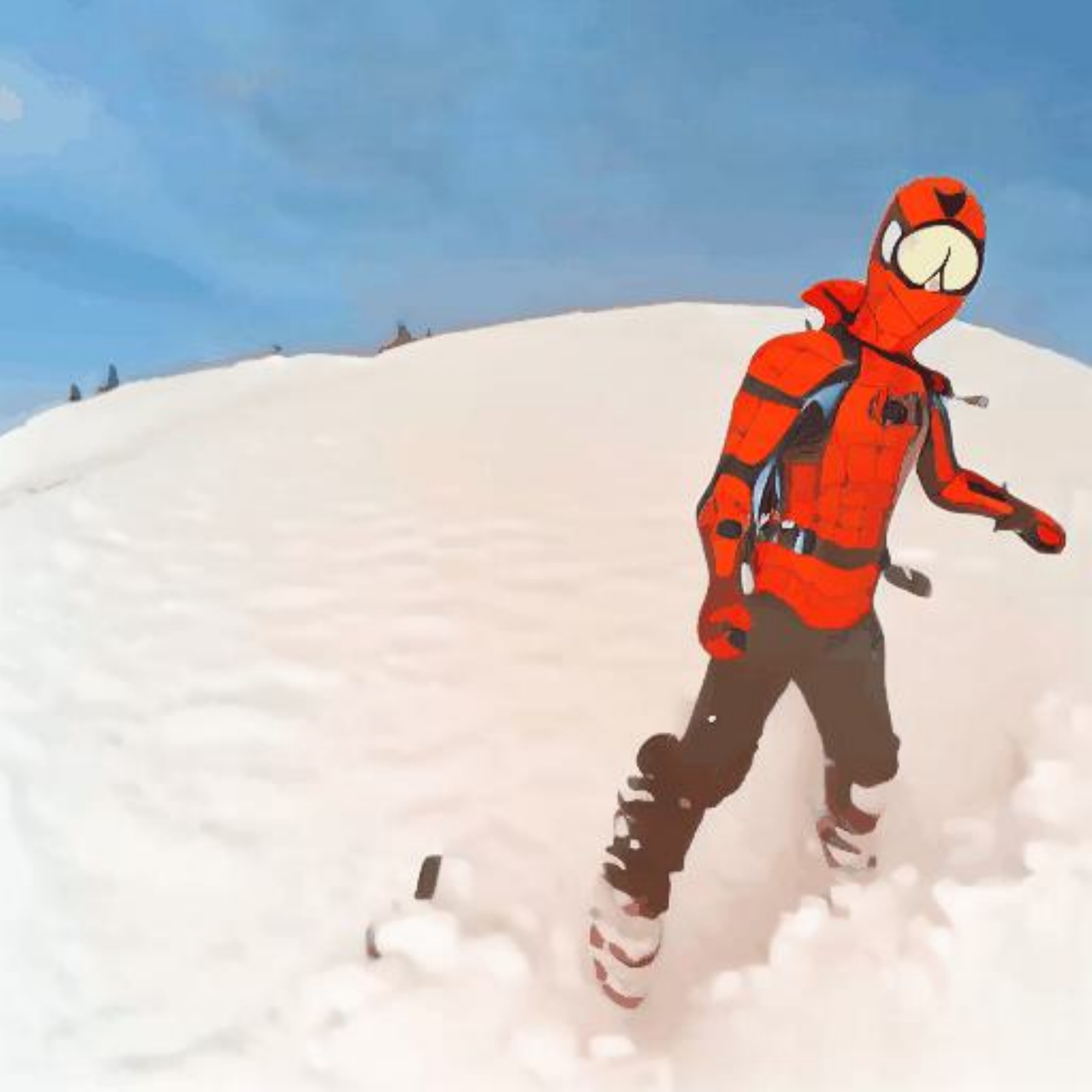}
\includegraphics[width=0.20\textwidth]{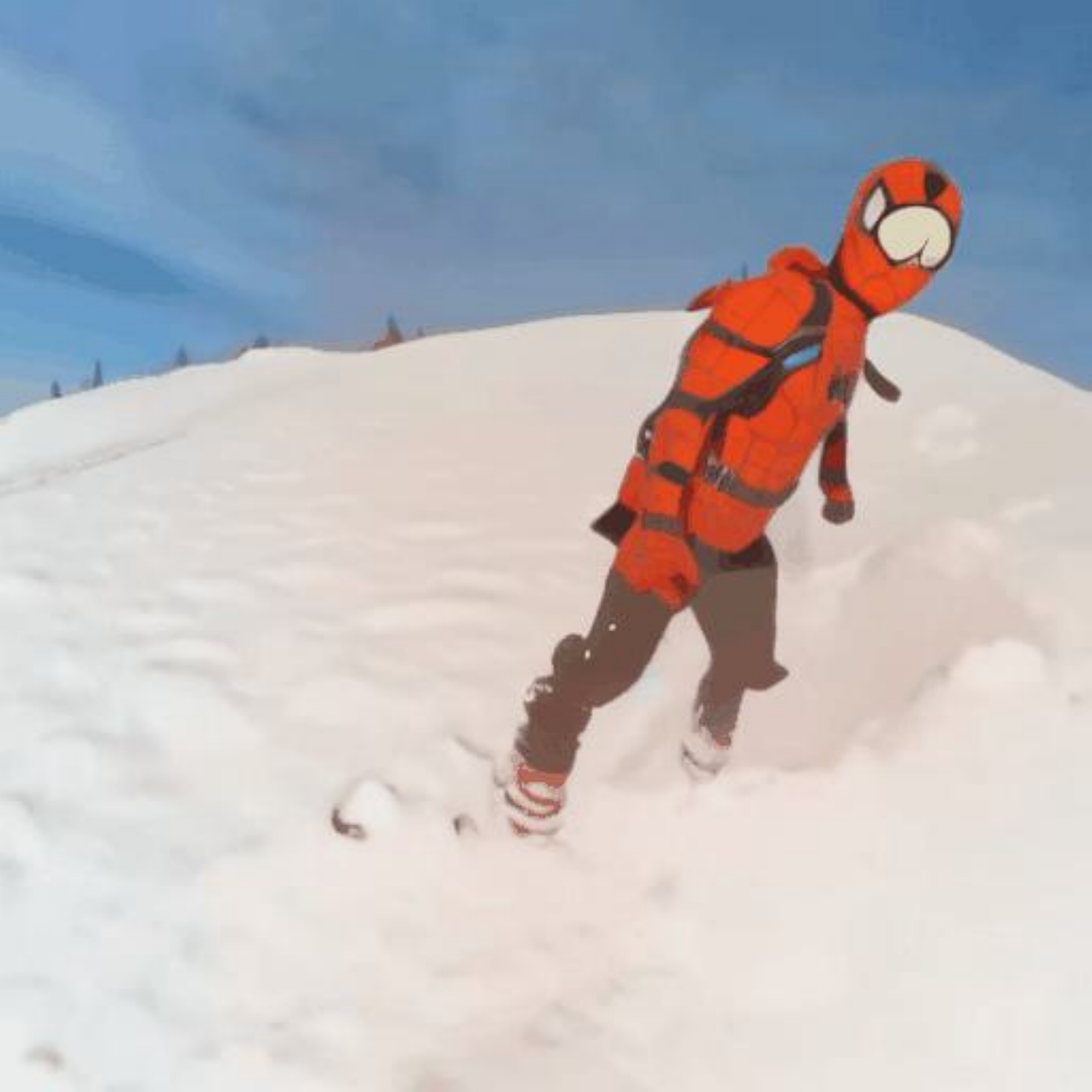}
\includegraphics[width=0.20\textwidth]{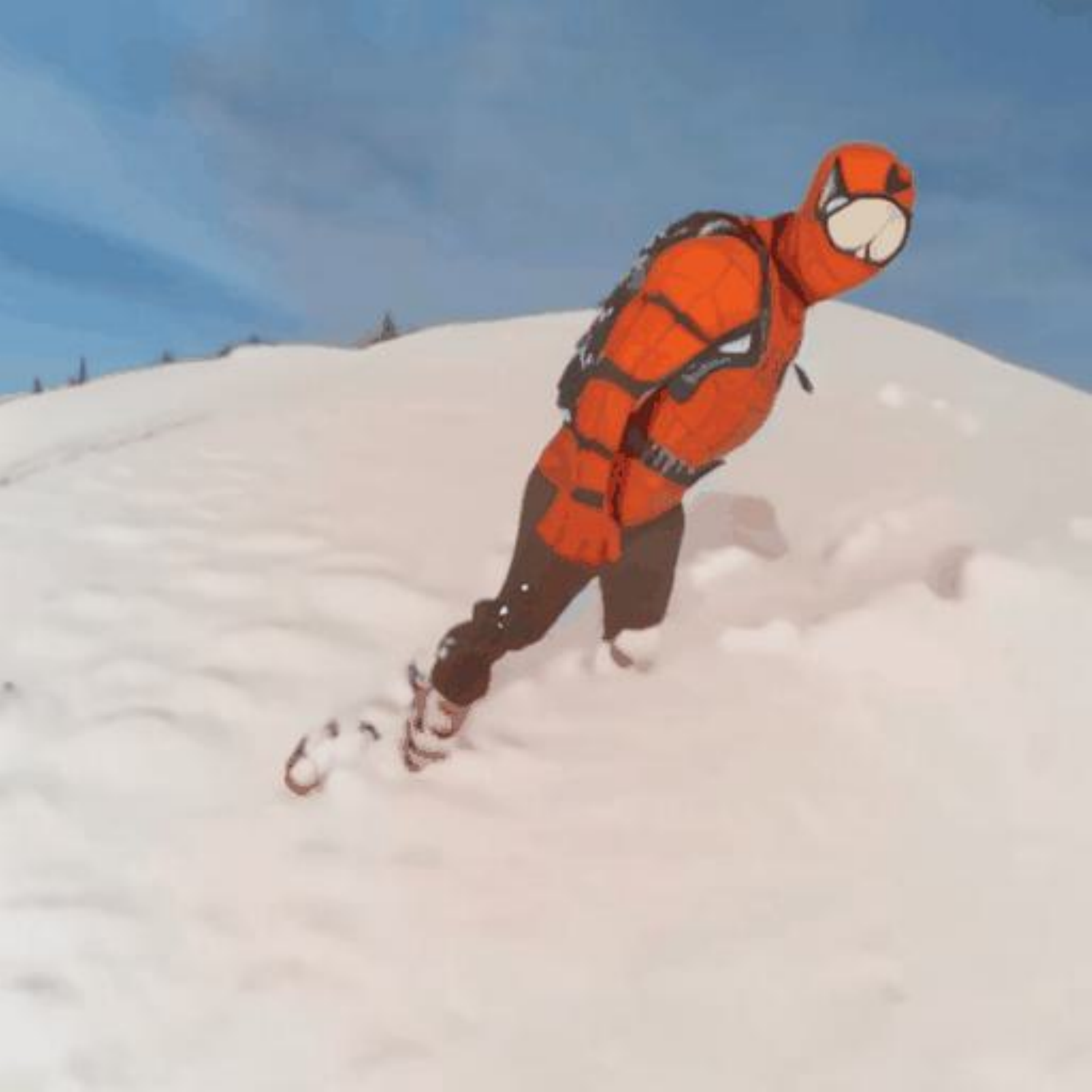}

\rotatebox{90}{\parbox{0.20\textwidth}{\centering \textbf{threshold \\ 0.25}}}
\includegraphics[width=0.20\textwidth]{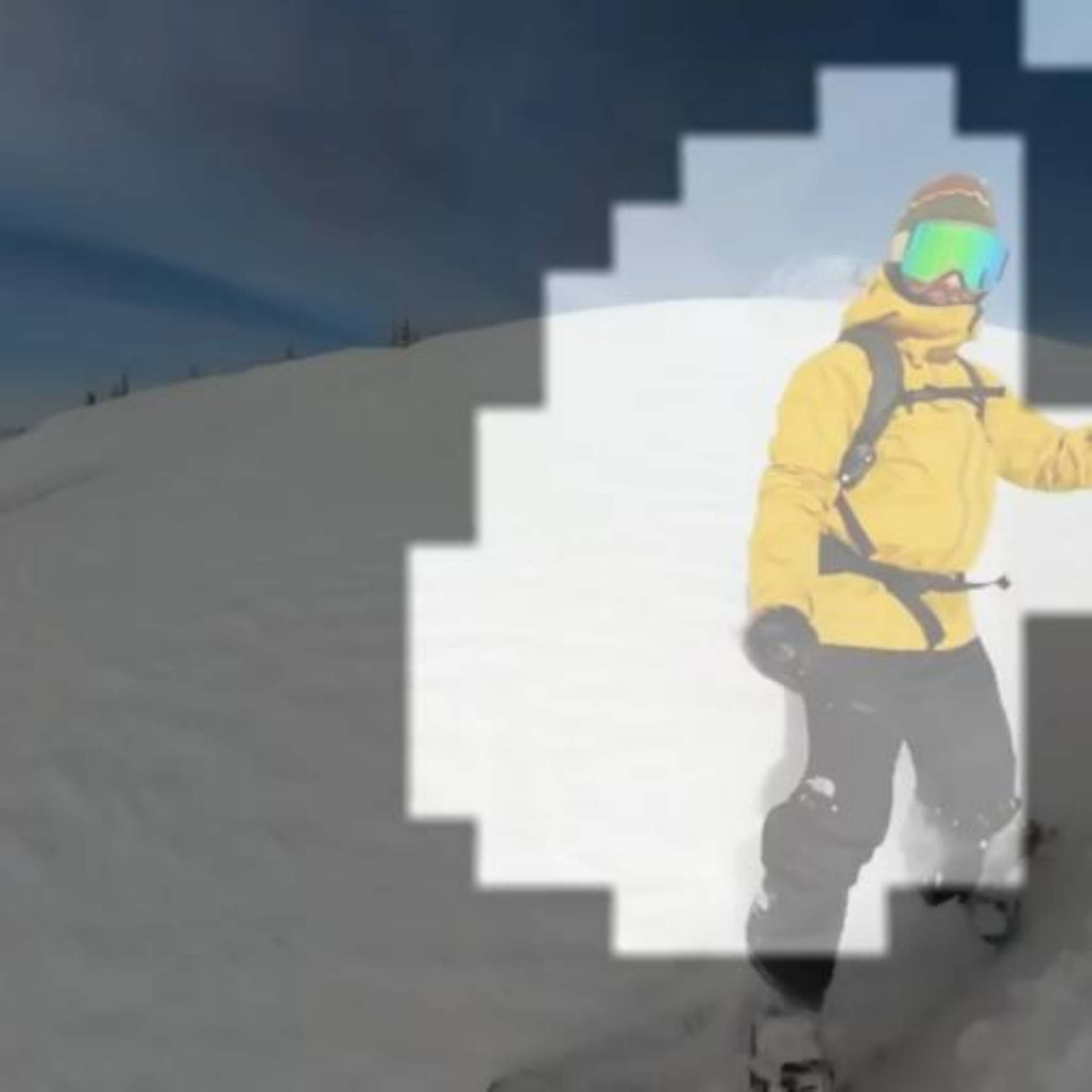}
\includegraphics[width=0.20\textwidth]{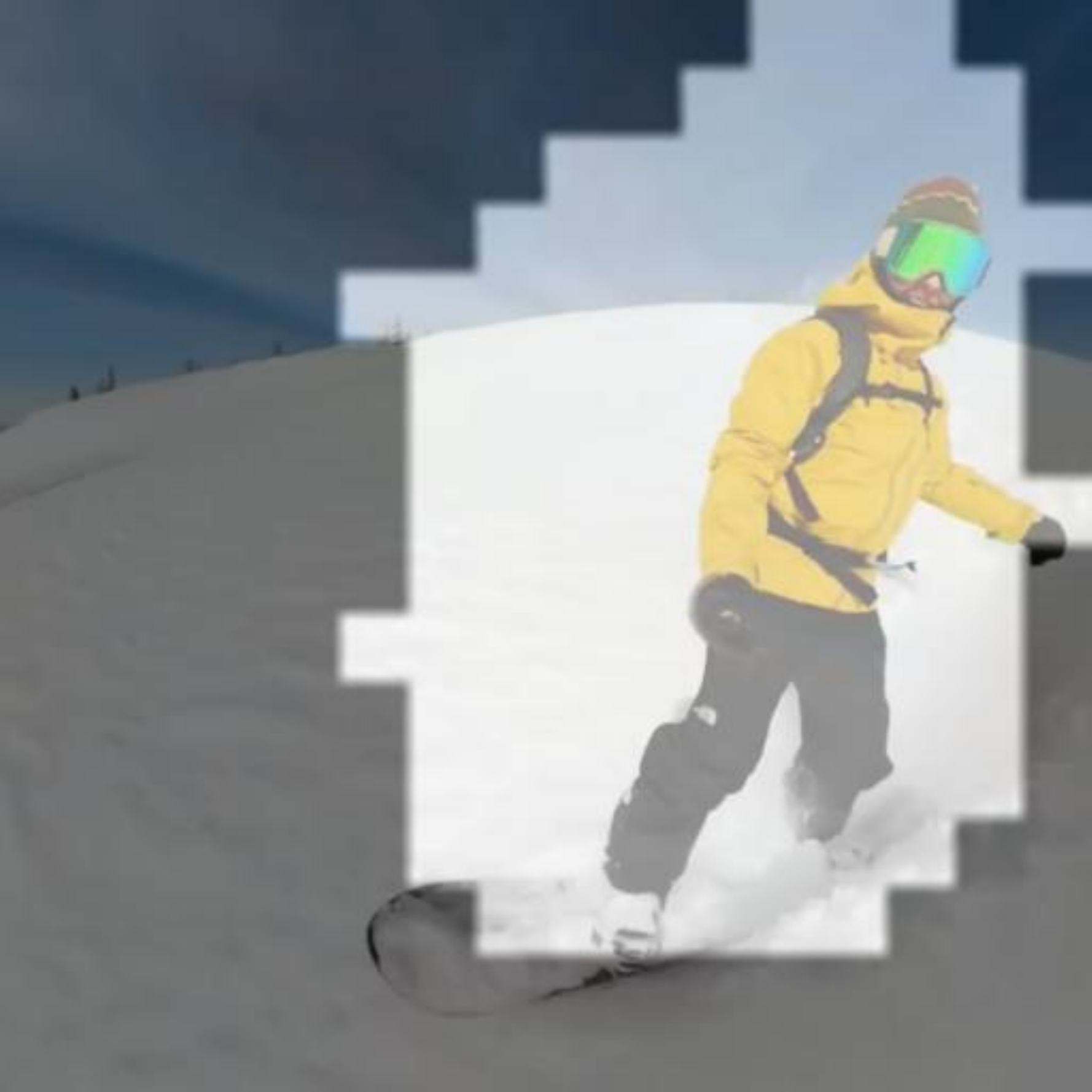}
\includegraphics[width=0.20\textwidth]{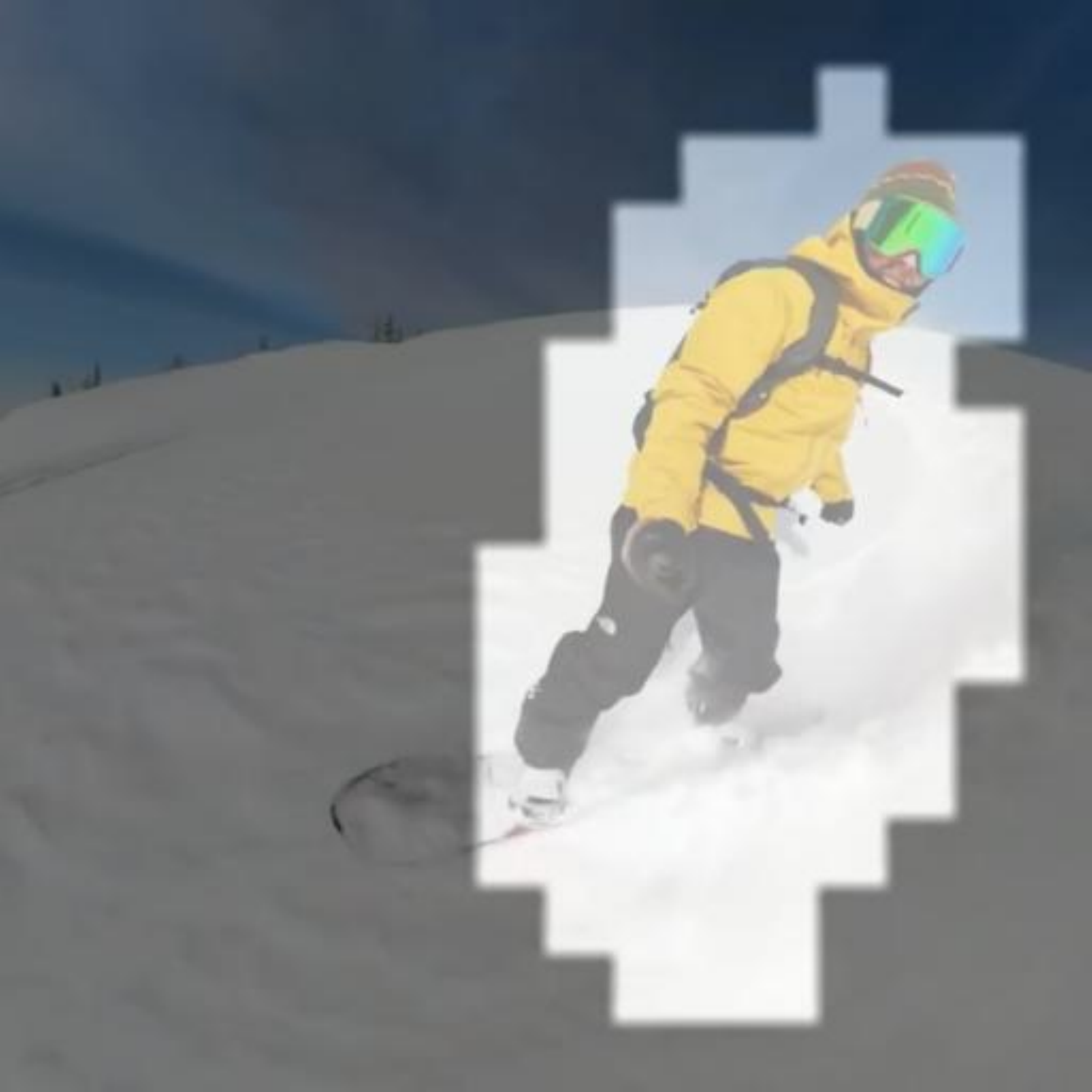}
\includegraphics[width=0.20\textwidth]{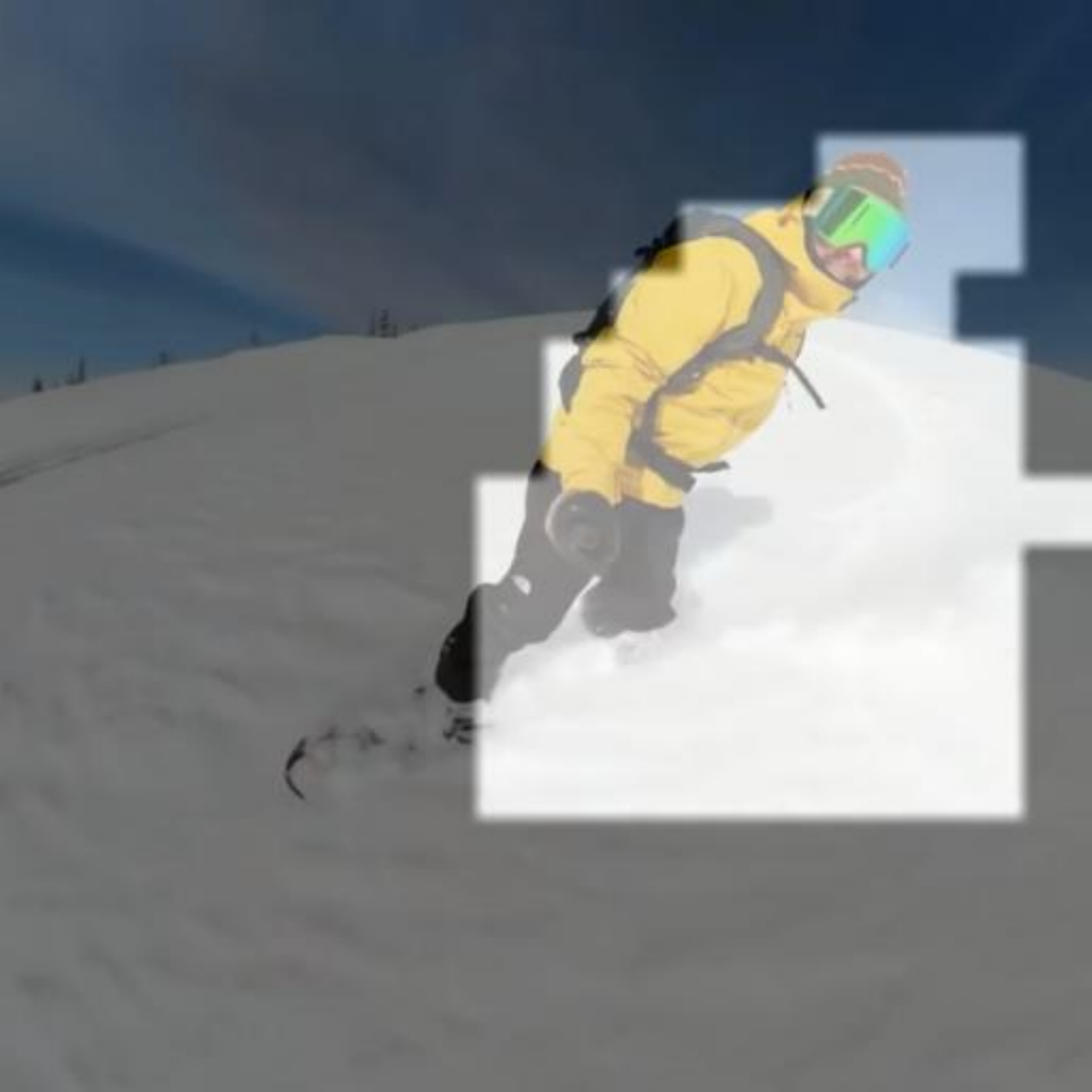}

\rotatebox{90}{\parbox{0.20\textwidth}{\centering ~ \\ ~}}
\includegraphics[width=0.20\textwidth]{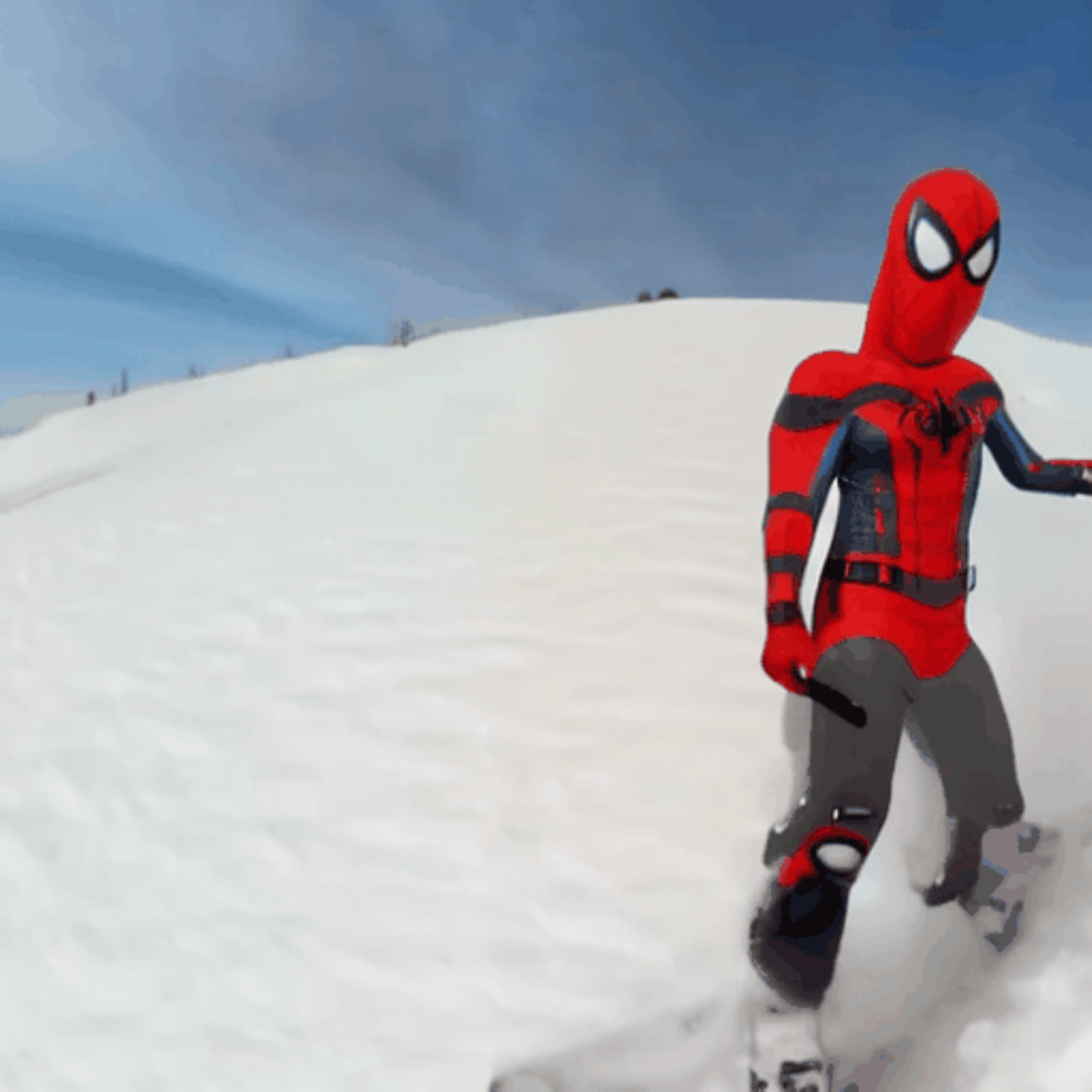}
\includegraphics[width=0.20\textwidth]{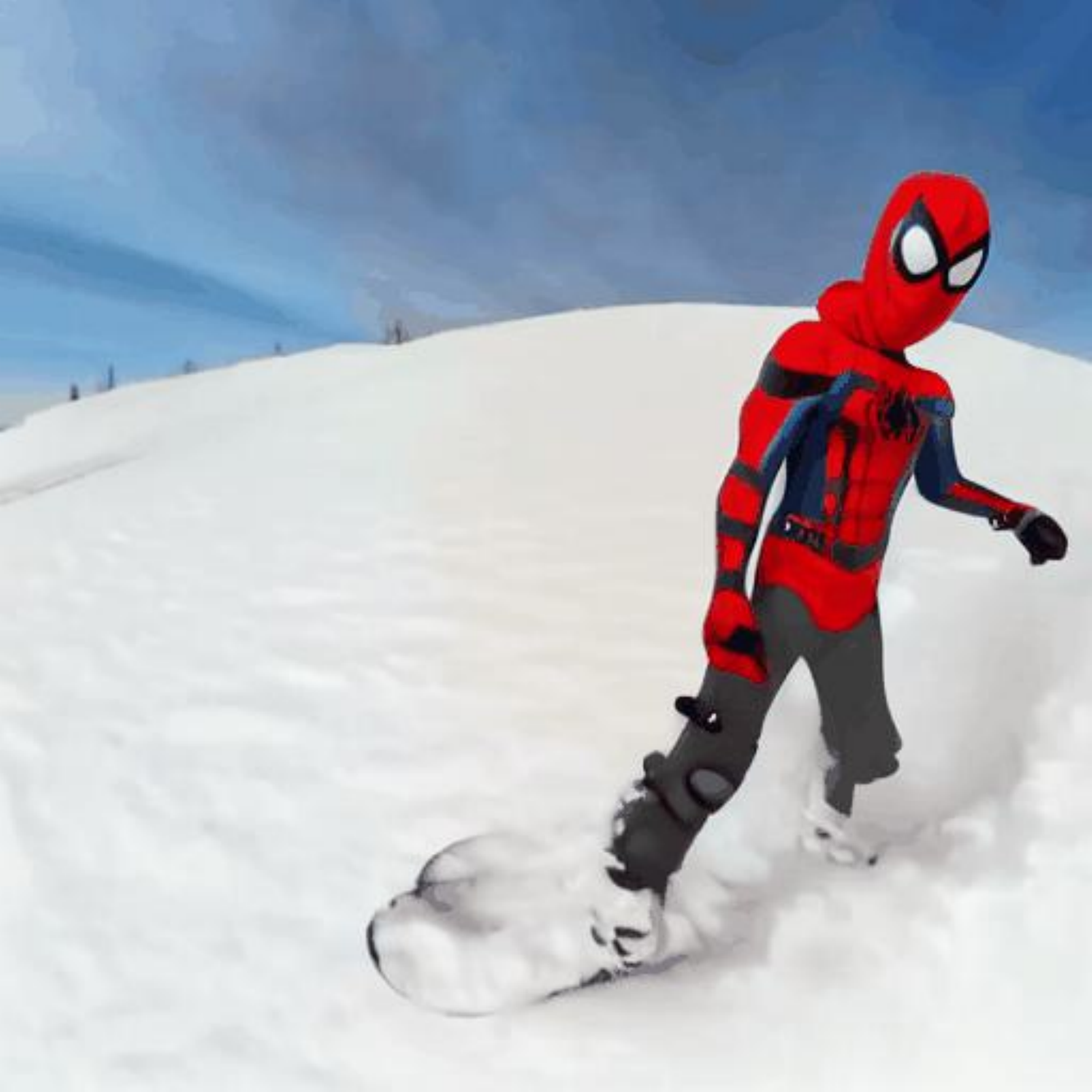}
\includegraphics[width=0.20\textwidth]{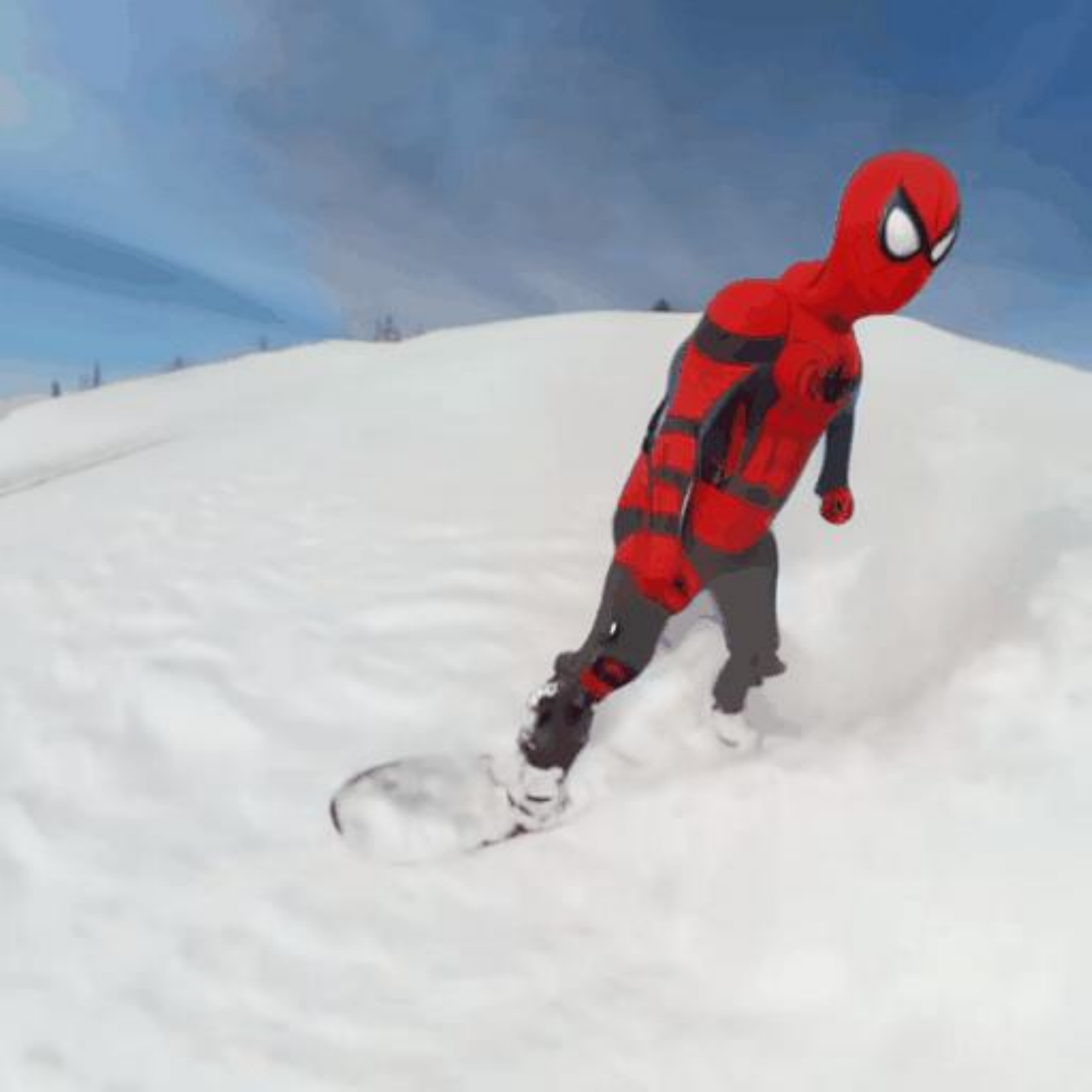}
\includegraphics[width=0.20\textwidth]{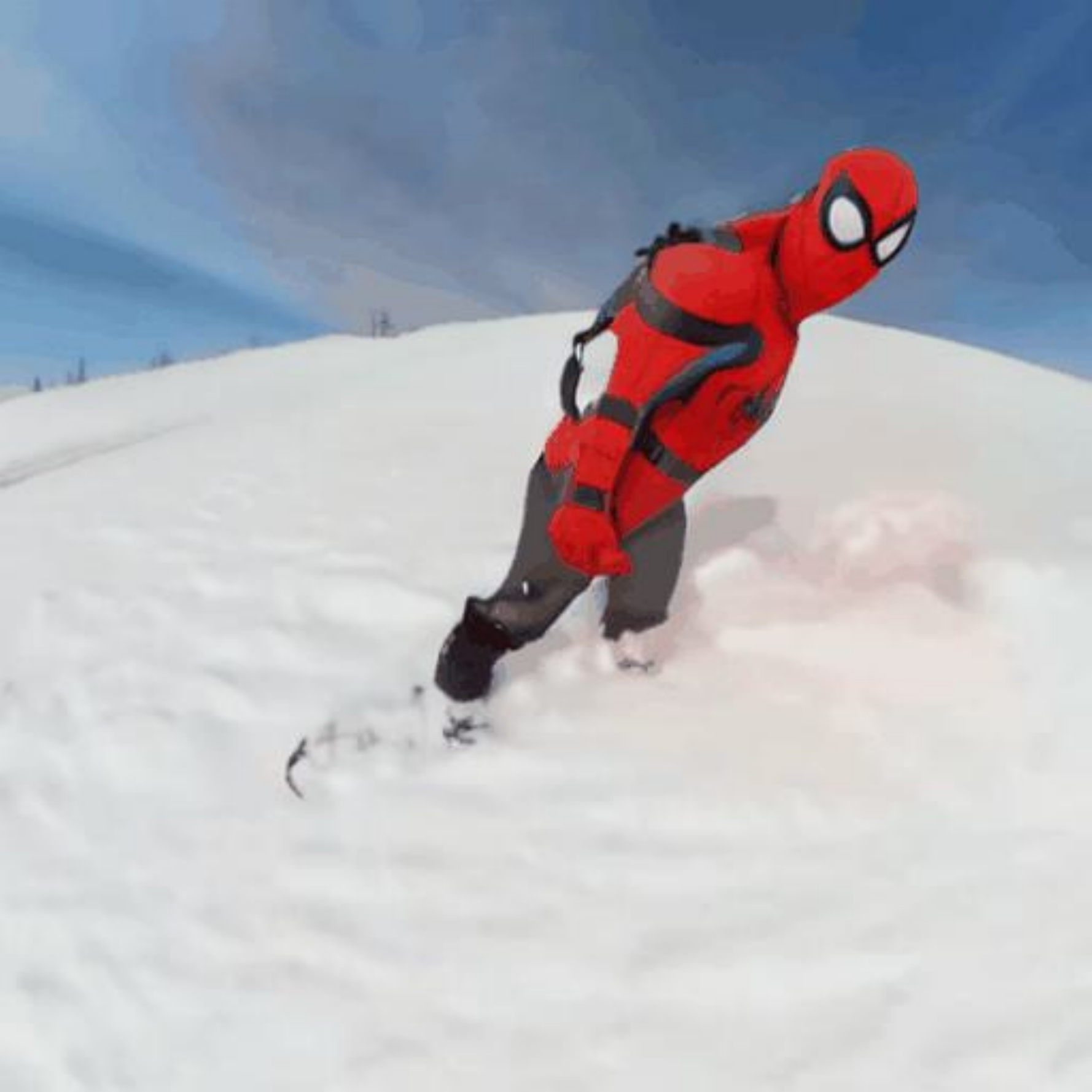}

\rotatebox{90}{\parbox{0.20\textwidth}{\centering threshold \\ 0.40}}
{
\includegraphics[width=0.20\textwidth]{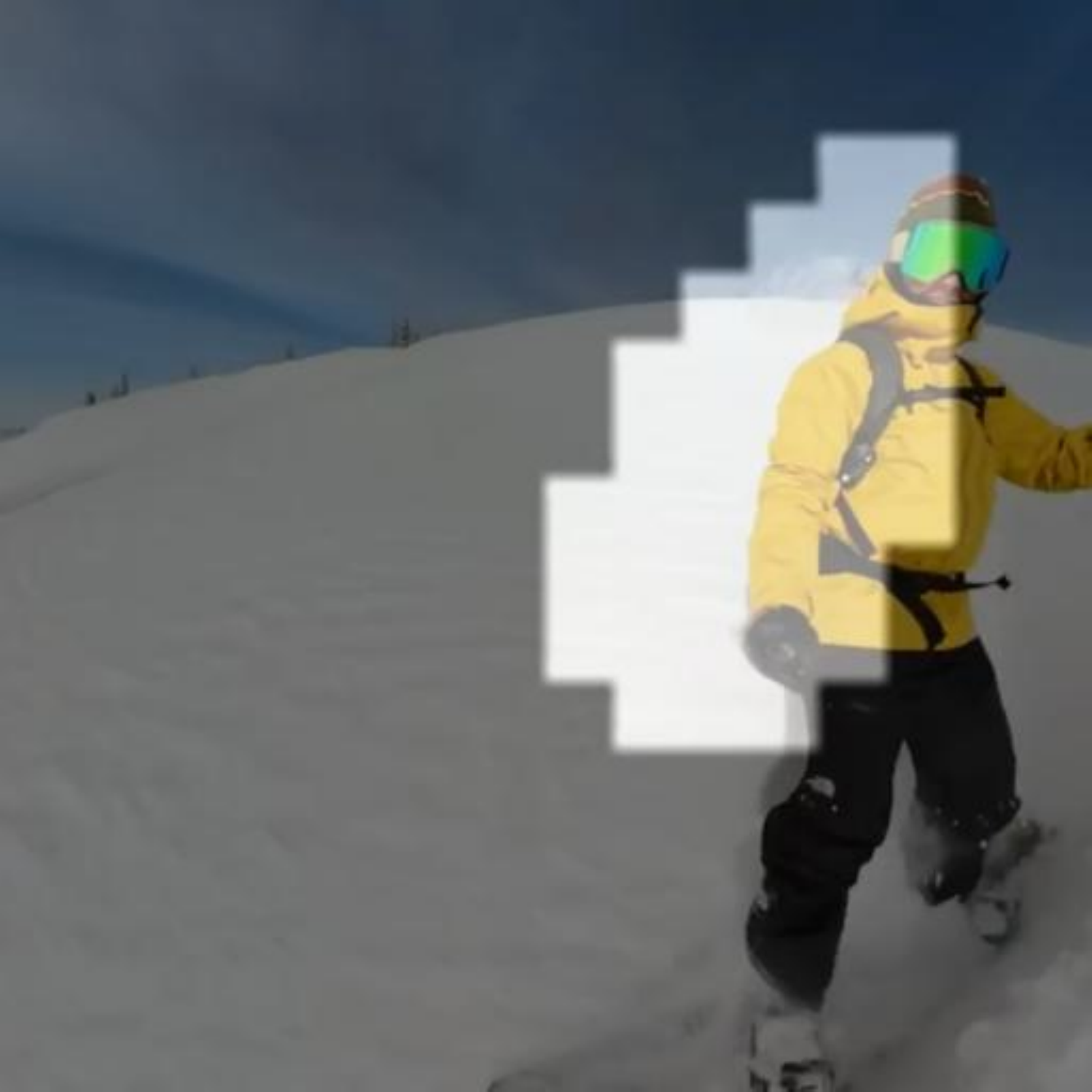}
\includegraphics[width=0.20\textwidth]{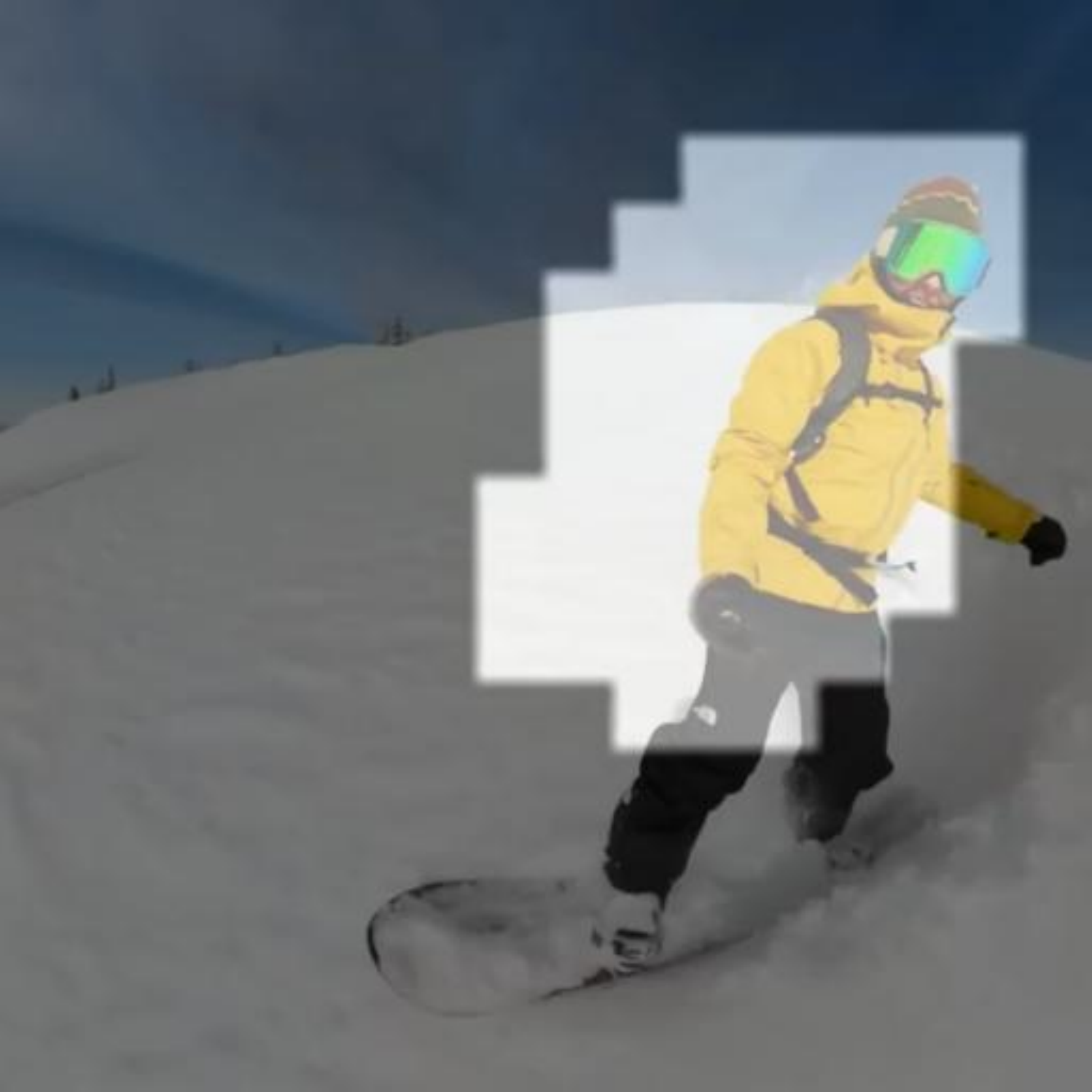}
\includegraphics[width=0.20\textwidth]{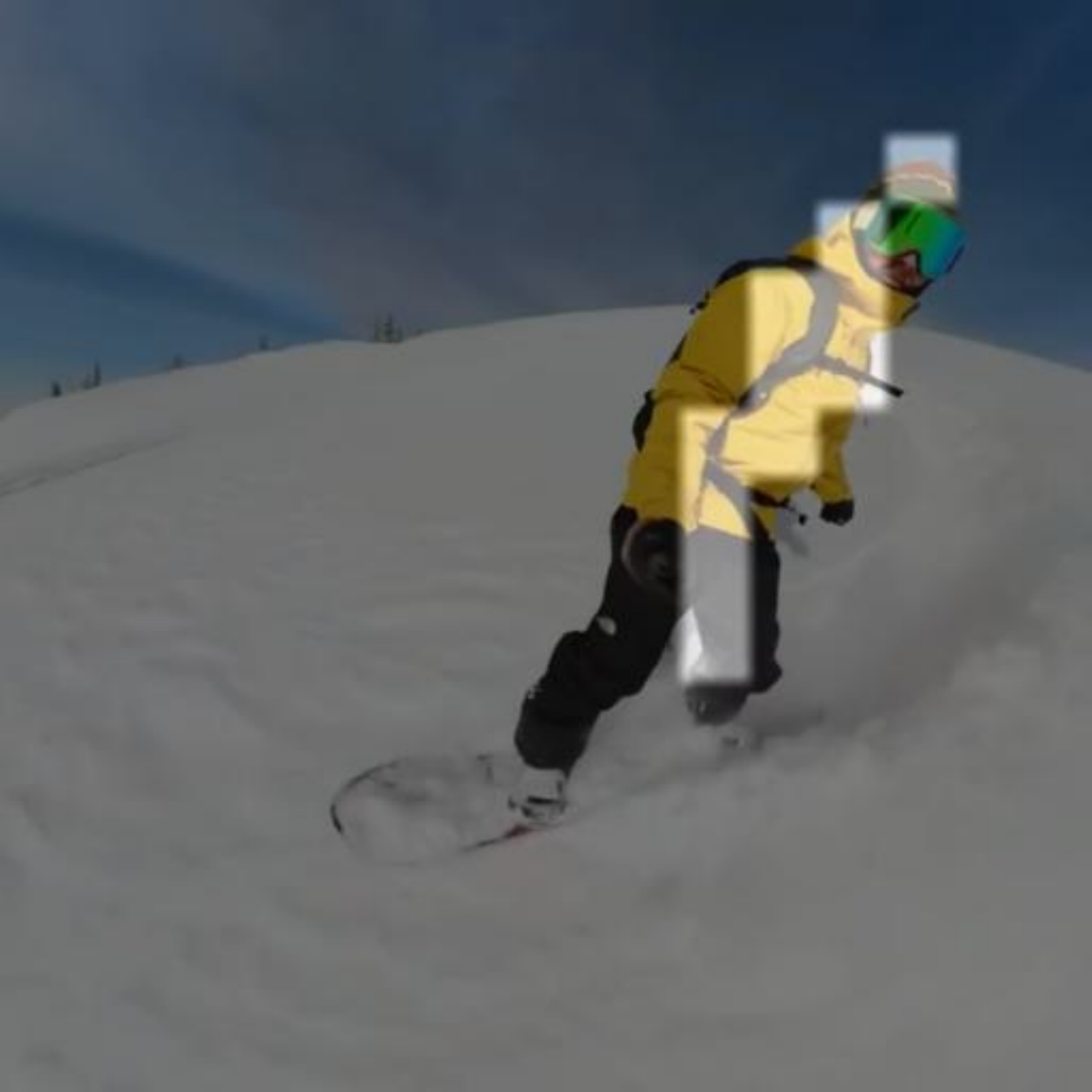}
\includegraphics[width=0.20\textwidth]{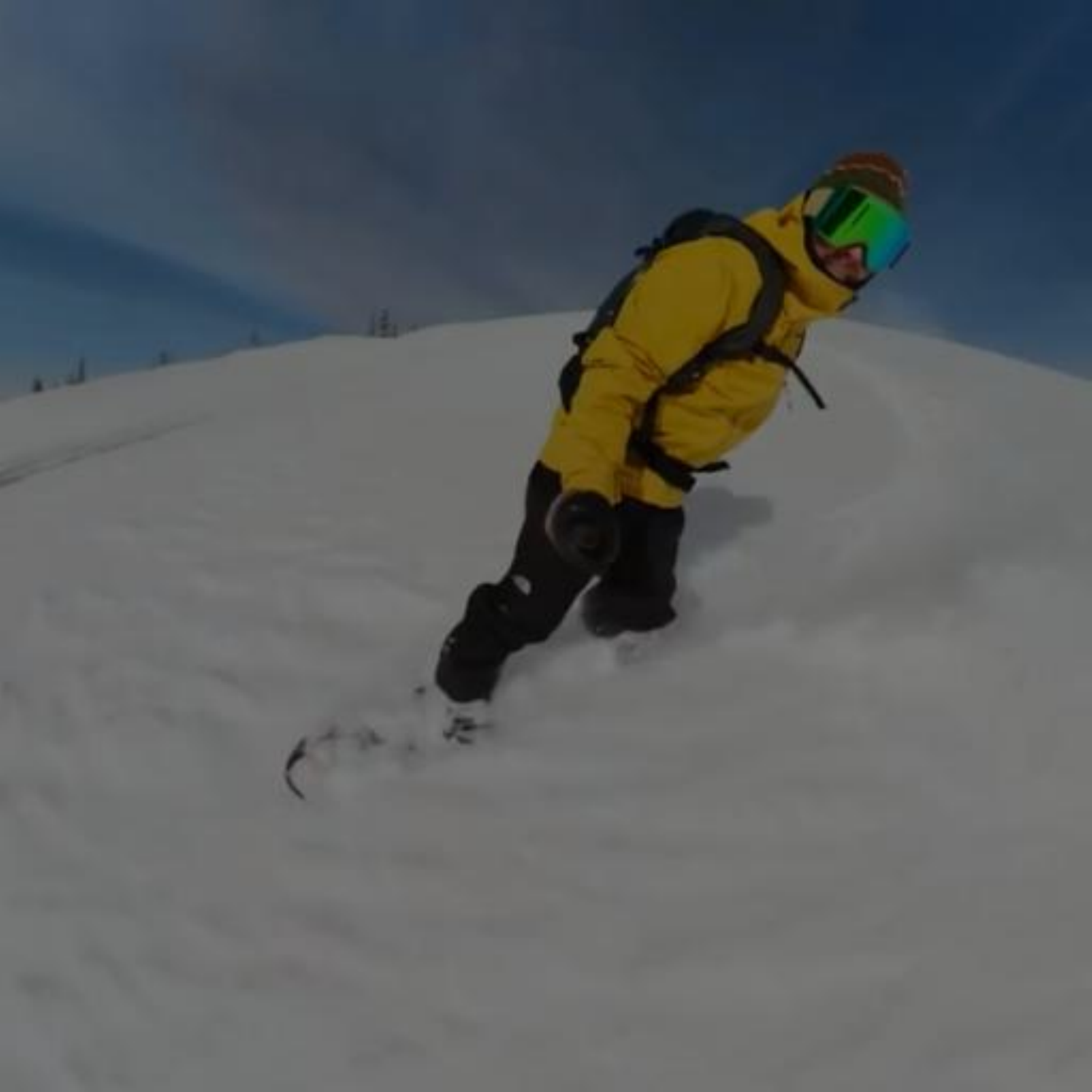}

\rotatebox{90}{\parbox{0.20\textwidth}{\centering ~ \\ ~}}
\includegraphics[width=0.20\textwidth]{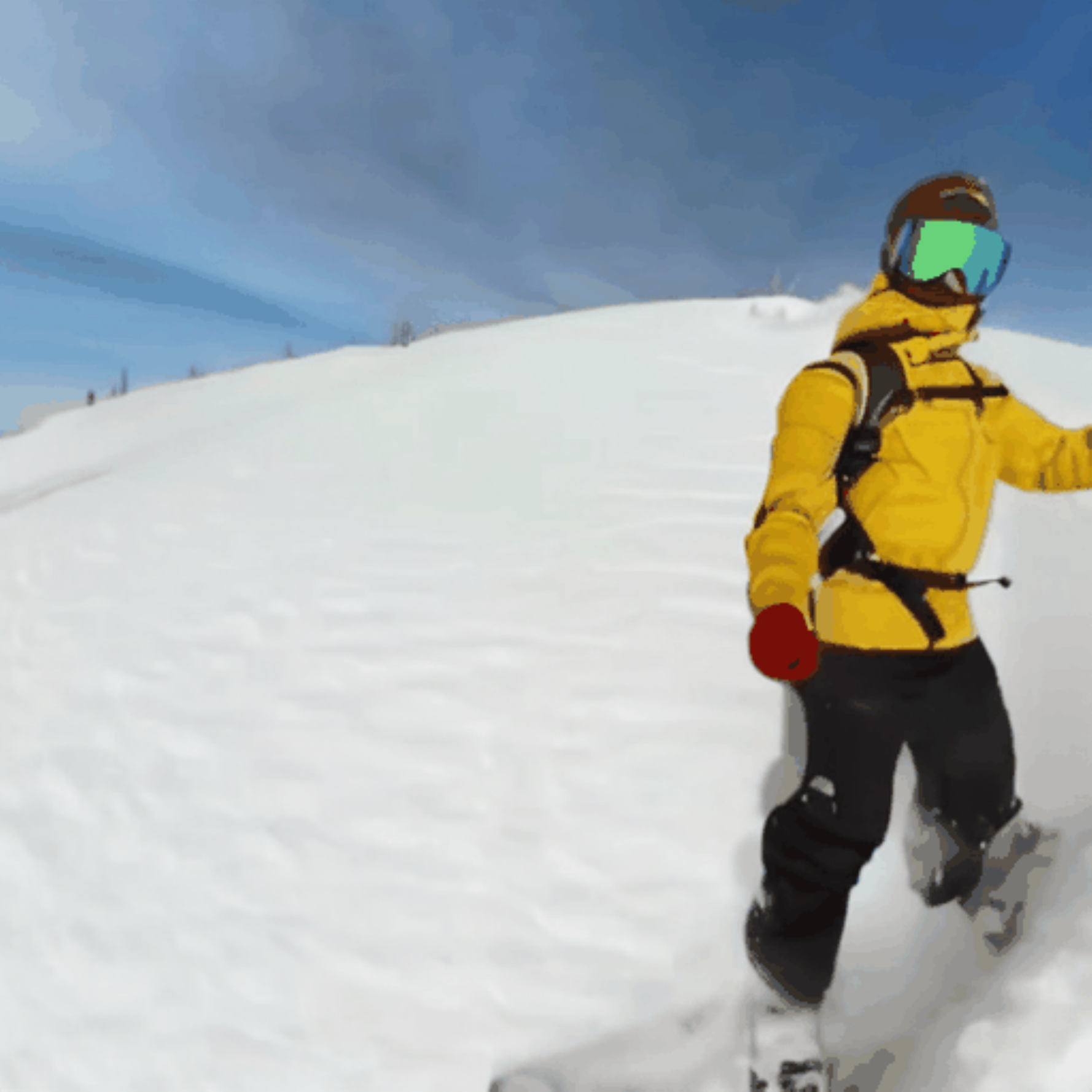}
\includegraphics[width=0.20\textwidth]{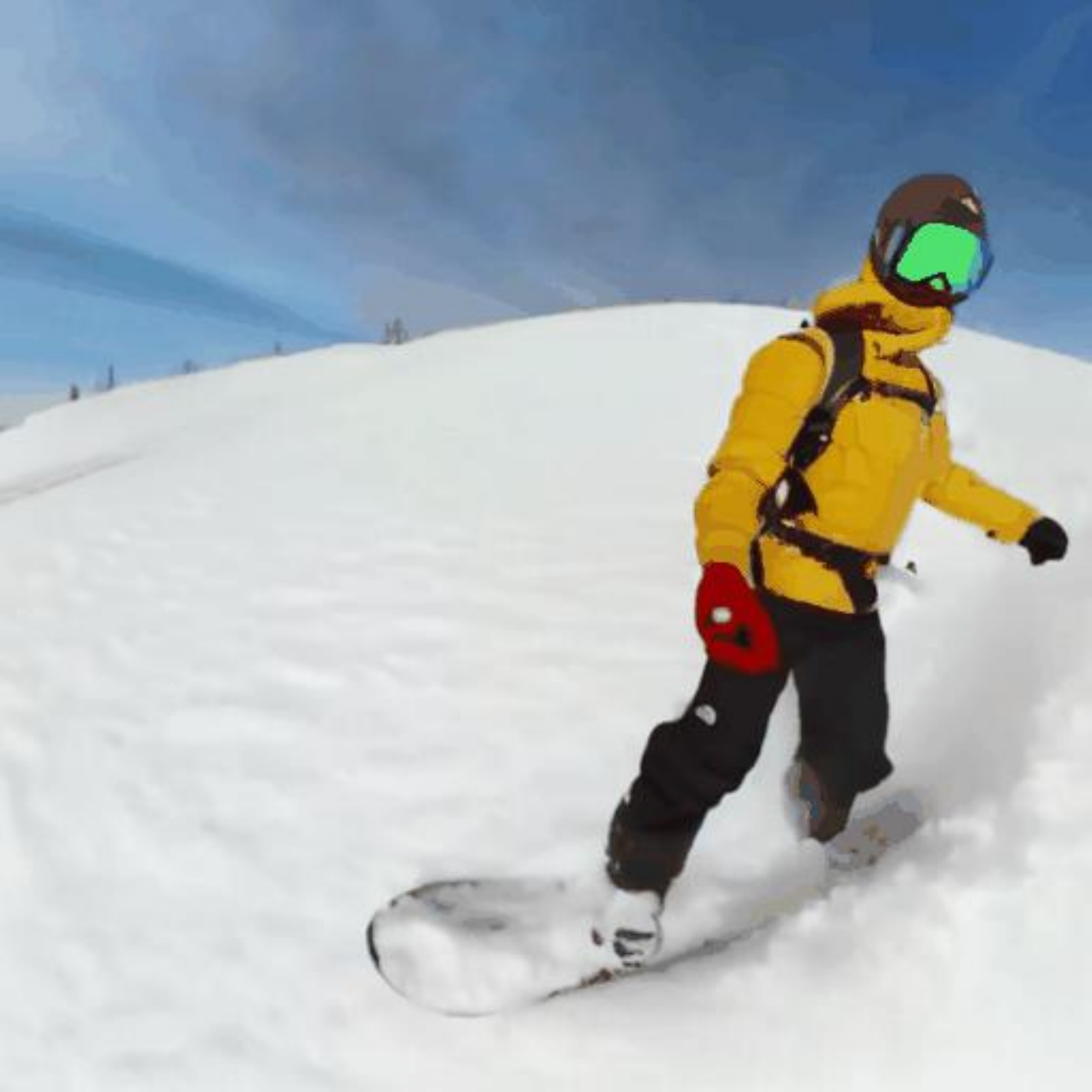}
\includegraphics[width=0.20\textwidth]{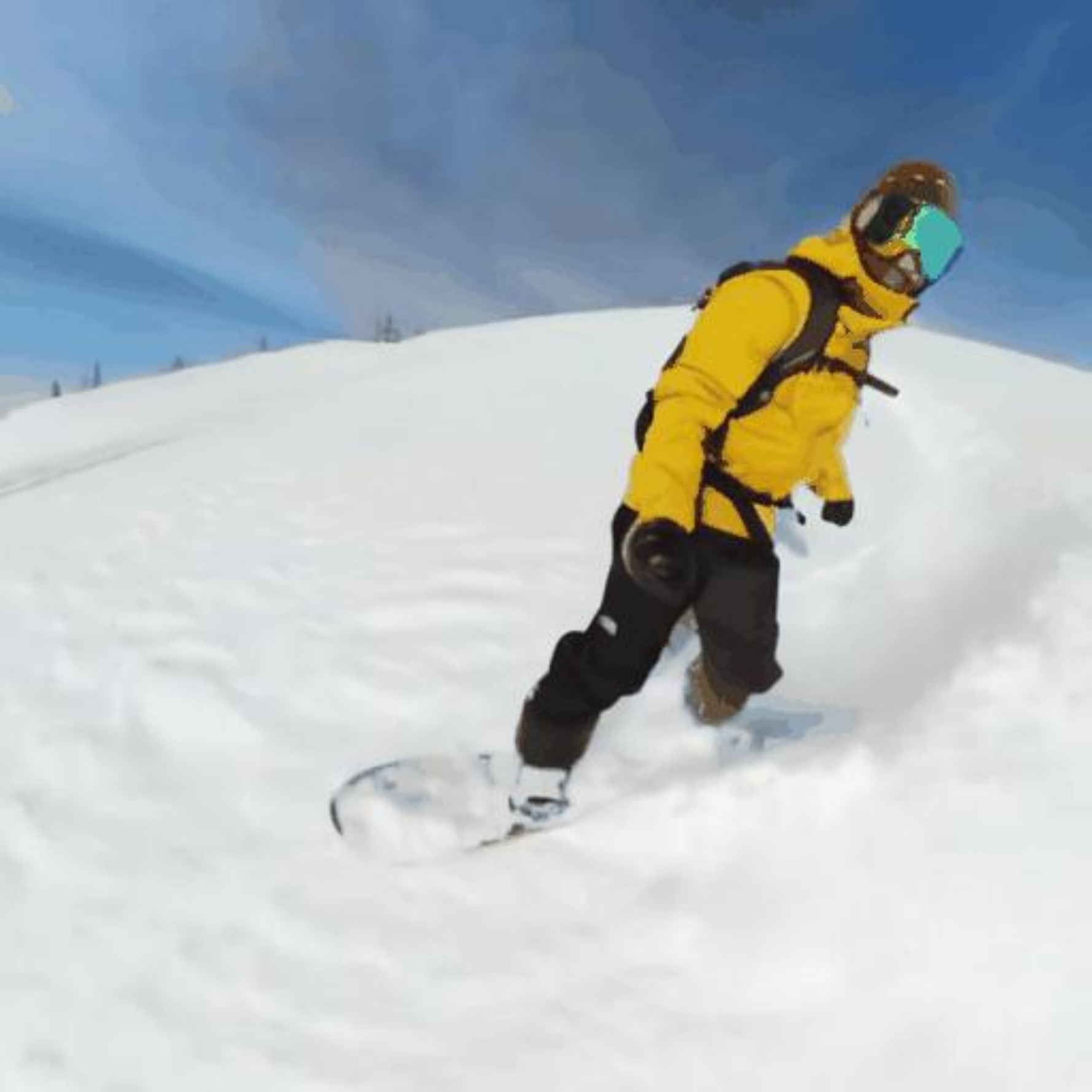}
\includegraphics[width=0.20\textwidth]{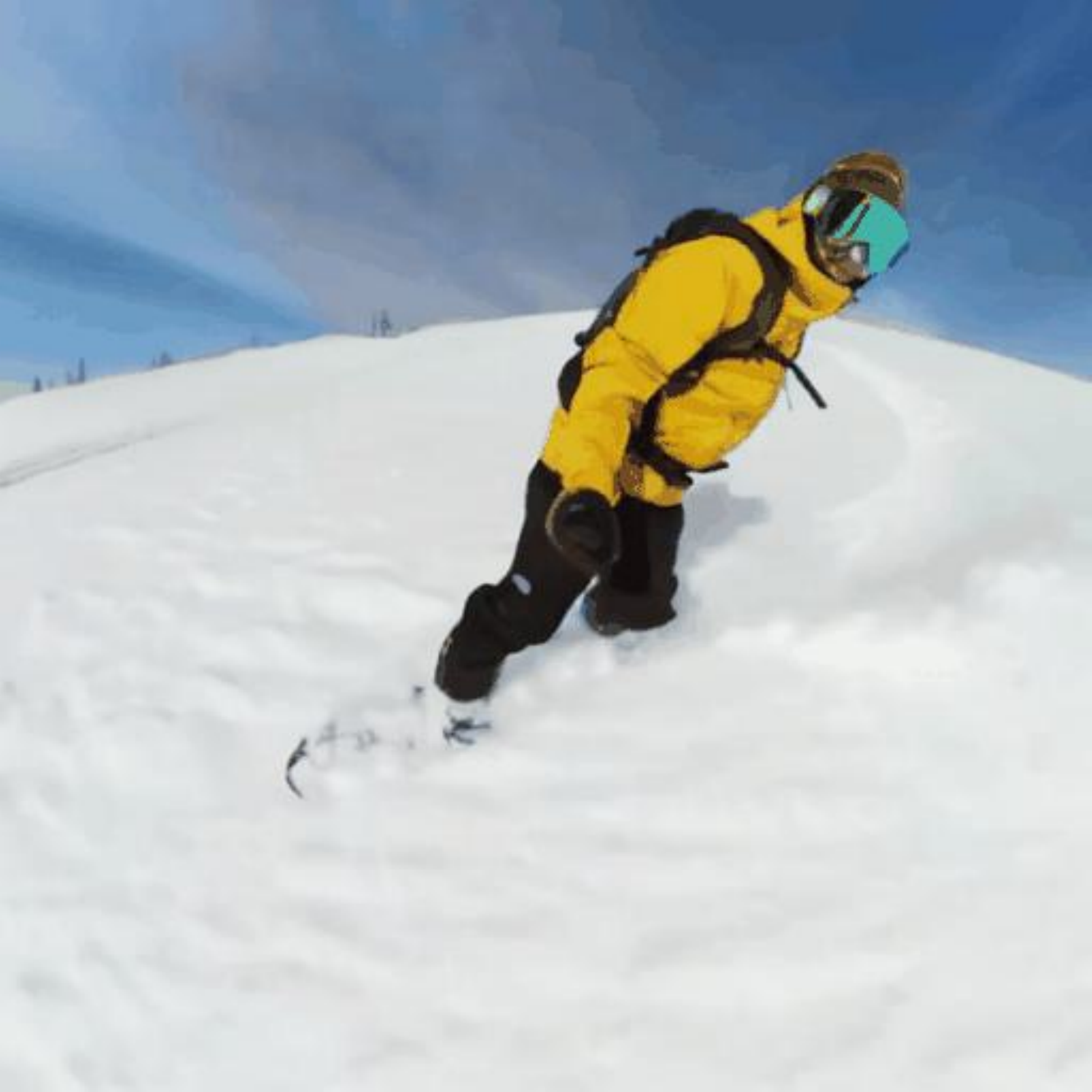}
}
\end{minipage}
\caption{\textbf{Qualitative Ablation Studies} We visualize the editing results for the target text ``A Spider Man is skiing". The samples on the left are based on the masking threshold without the proposed blending method applied, and the samples on the right are the results with the proposed blending method applied.}
\label{fig:mask_thres_ablation}
\end{center}
\vspace{-2.0em}
\end{figure}

In this section, we analyze the TC Blending mask's impact through an ablation study.
We evaluate the editing targets of 50 out of the 100 pairs of $<$text, video$>$ used in Sec~\ref{baseline_comparison}, focusing specifically on object editing.
We evaluate the proposed blending method's effectiveness using two human preference scores and three automatic metrics.
The human preference scores, User Score (O) and User Score (P), are rated on a 5-point scale, measuring overall editing quality and non-target region preservation.
For automatic metrics, we use LPIPS, PSNR, and a newly proposed metric, Mask Intersection over Union (Mask IoU).
Mask IoU quantifies the overlap between the blending mask and the target object region, calculated as the Intersection over Union (IoU) between these two regions.
To capture the target object region, we use a publicly available saliency detector~\citep{zheng2022mccl}.

As shown in table~\ref{tab:ablation}, Edit-A-Video achieves higher user scores than the model without TC Blending (p-value $<$ $0.01$ from the Wilcoxon signed-rank test).
Furthermore, through various automatic metrics and qualitative results in Fig~\ref{fig:mask_thres_ablation}, we confirm that TC Blending is effective in accurate target object masking and background preservation.

We also demonstrate the effect of mask threshold value $\tau$. 
In Fig~\ref{fig:mask_thres_ablation}, $\tau$ controls the editing region size. 
Without TC Blending, adjusting the threshold fails to capture sharp object regions effectively, causing abrupt frame-wise mask changes and undesirable artifacts. 
In contrast, Edit-A-Video achieves sharp and smooth masks by properly setting the threshold. These results confirm that our proposed TC Blending mitigates background inconsistency.

\section{Conclusion}

We propose Edit-A-Video, the video editing framework only given the single $<$text, video$>$ pair and pretrained text-to-image (TTI) model.
Edit-A-Video inflates the TTI model and tunes the model on a given source video for the temporal modeling, and enables editing the video with target prompt by the inversion and attention map injection.
We also suggest the temporal-consistent blending method which ensures content preservation and temporal coherence by making use of the temporal modeling capability in the model.
Our framework achieves superior performance to the baselines in various aspects.
We anticipate that this method will provide a simple and intuitive video editing method.

\acks{This work was supported by Institute of Information \& communications Technology Planning \& Evaluation grant funded by the Korea government (MSIT) [2021-0-01343, AI Graduate School Program (SNU)], National Research Foundation of Korea grant funded by MSIT (2022R1A3B1077720), and the BK21 FOUR program of the Education and Research Program for Future ICT Pioneers, SNU in 2023.}

\bibliography{main}

\clearpage
\newpage

{
\centering
\Large
\textbf{Edit-A-Video: Single Video Editing with Object-Aware Consistency} \\
\vspace{0.5em}-- Supplementary Materials --\\
\vspace{1.0em}
}

\appendix
\setcounter{figure}{0}
\renewcommand{\thefigure}{\Alph{figure}}

\appendix

\section{Human Preference Study}
\label{sec:human_preference}
We conduct two distinct human preference studies. The first one involves evaluating the overall video editing quality of our model and baselines. We collect feedback from $62$ participants who are asked to rate scores on a 5-point scale from 1-5 by taking into account three key factors as follows:

\begin{itemize}
    \item \textbf{Background Preservation} Edited video preserves unedited details of the original video.
    \item \textbf{Text Alignment} Edited video matches the target edit description provided.
    \item \textbf{Video Realism} The overall visual quality and smoothness of the edited video.
\end{itemize}

In the second study, we measure the effectiveness of our proposed TC Blending technique in terms of the \textit{background inconsistency problem}. We check this by assessing the degree of preservation and consistency of the background in the presence or absence of the TC Blending technique. Once again, we ask $49$ participants to provide feedback, with scores related to background preservation.

\section{Additional Samples}
\label{sec:add_samples}
\begin{figure*}
\vspace{0.8em}
\begin{center}
\makebox[0.12\textwidth]{\colorbox{pink}{\textbf{Training video}} A man is dribbling a basketball}\\
\includegraphics[width=0.10\textwidth]{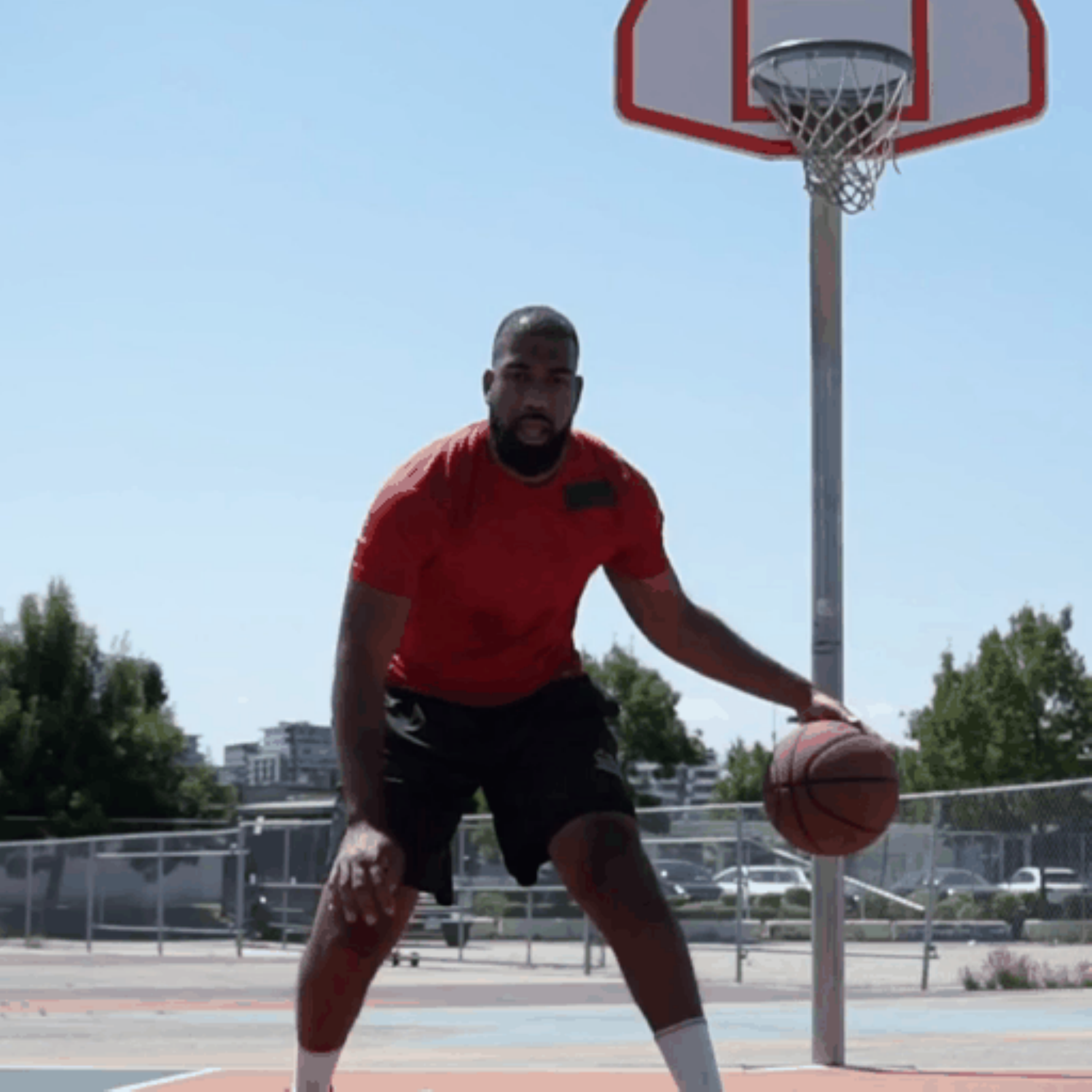}
\includegraphics[width=0.10\textwidth]{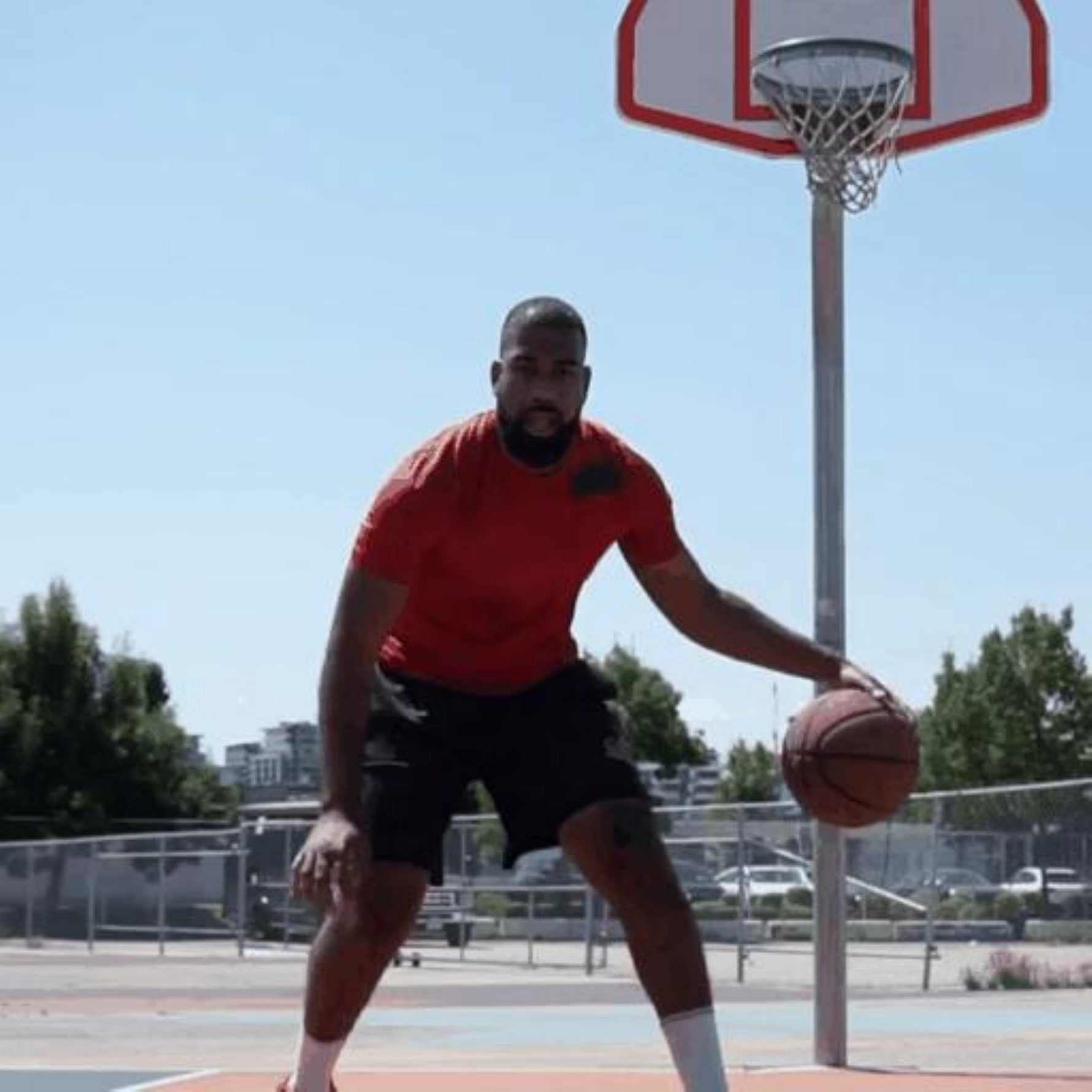}
\includegraphics[width=0.10\textwidth]{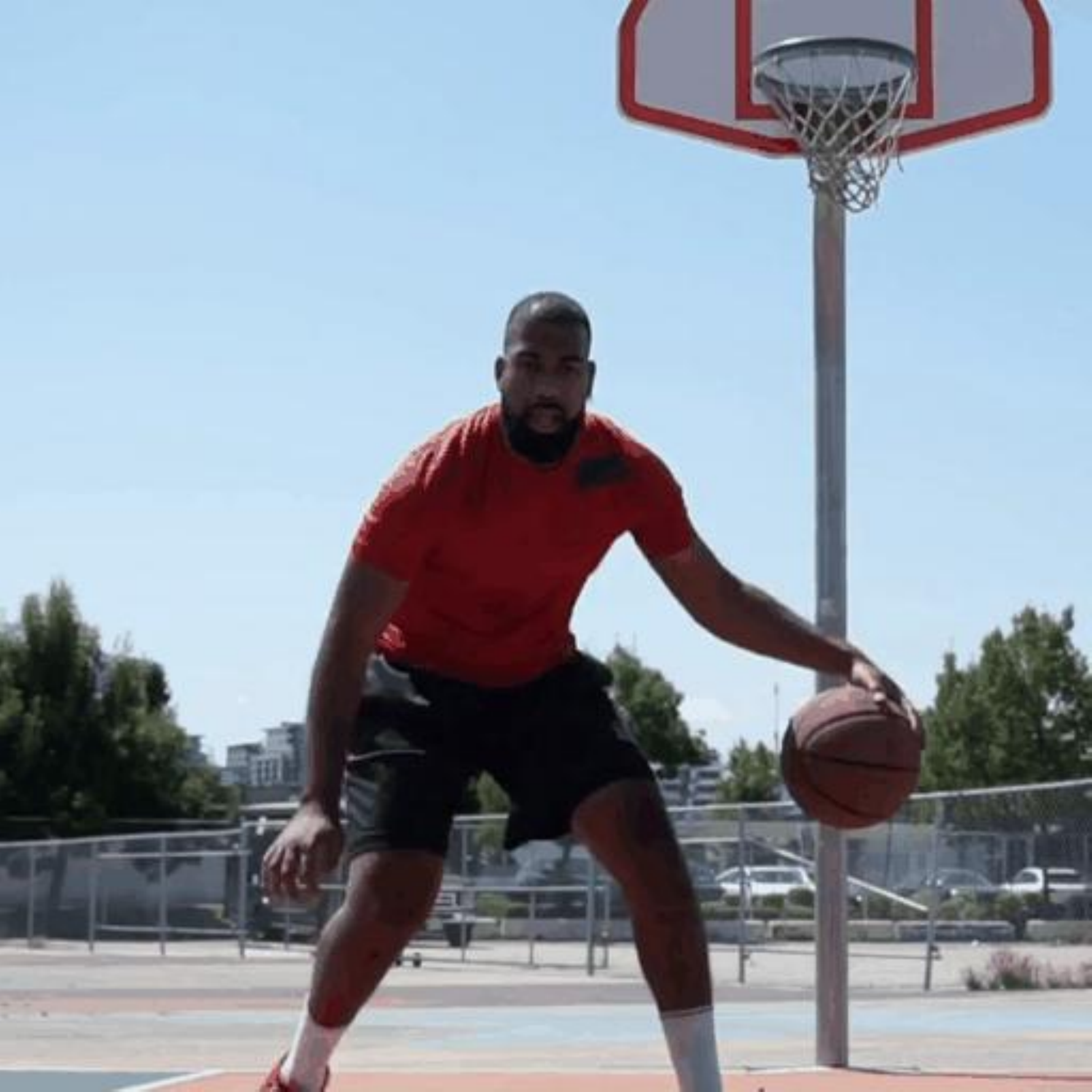}
\includegraphics[width=0.10\textwidth]{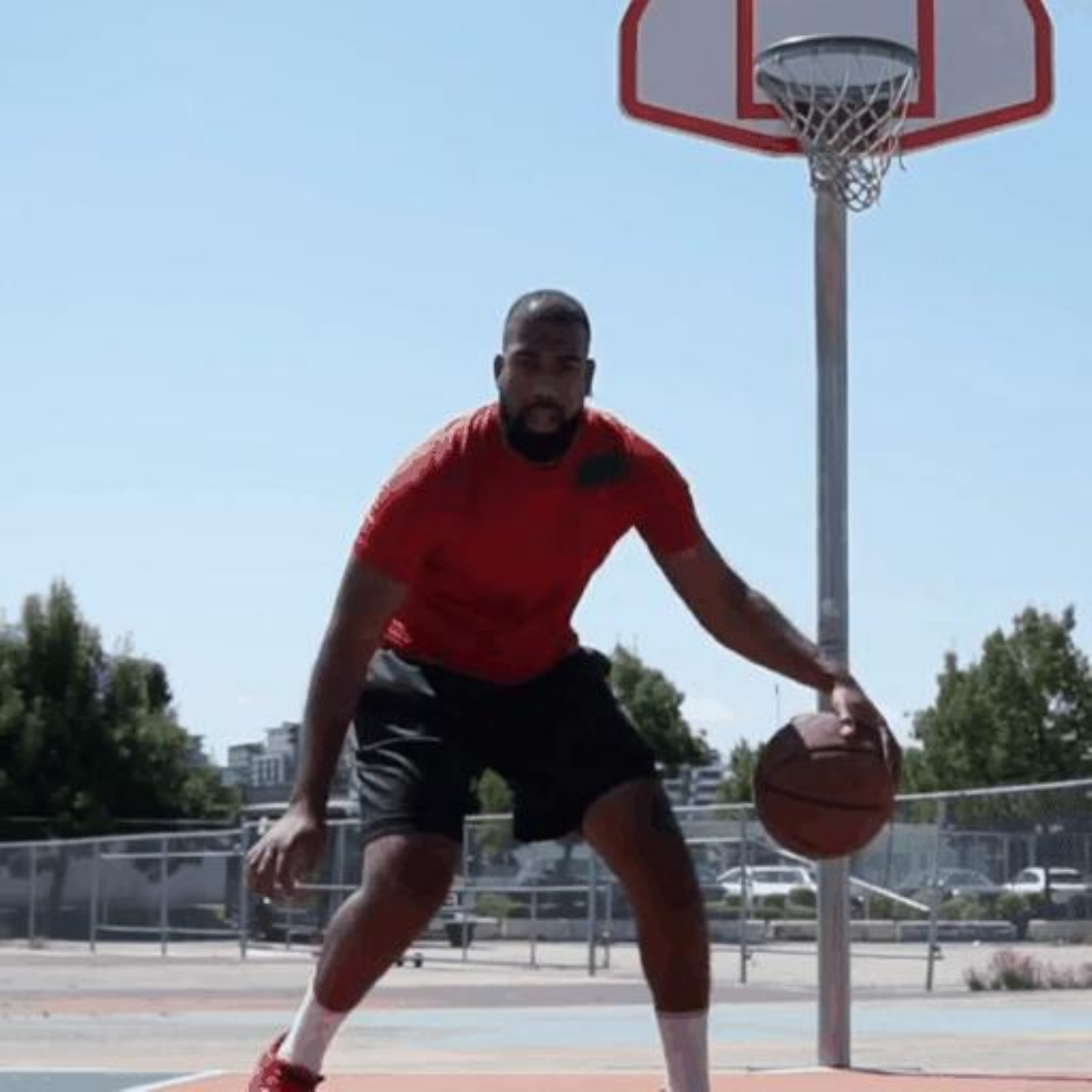}
\includegraphics[width=0.10\textwidth]{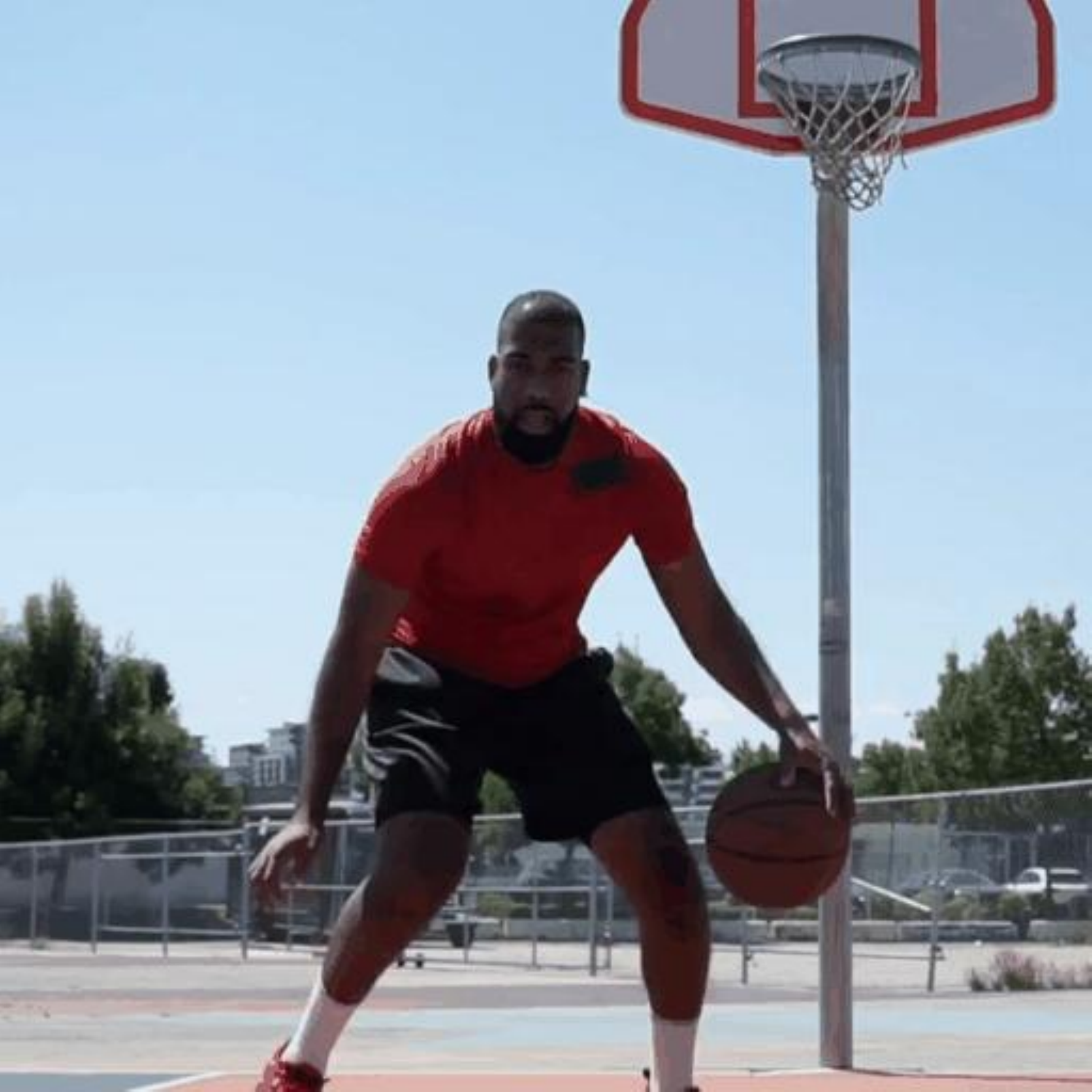}
\includegraphics[width=0.10\textwidth]{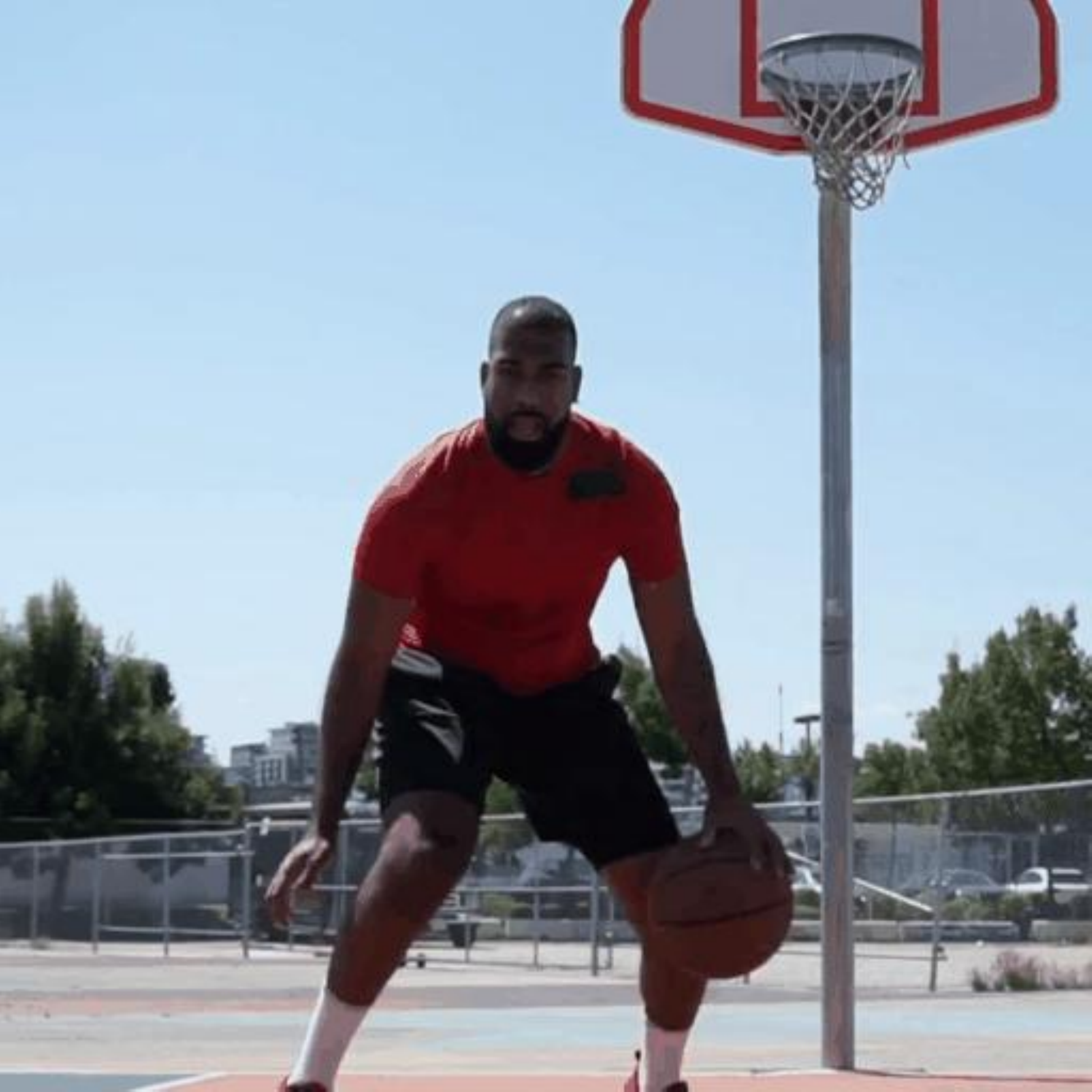}
\includegraphics[width=0.10\textwidth]{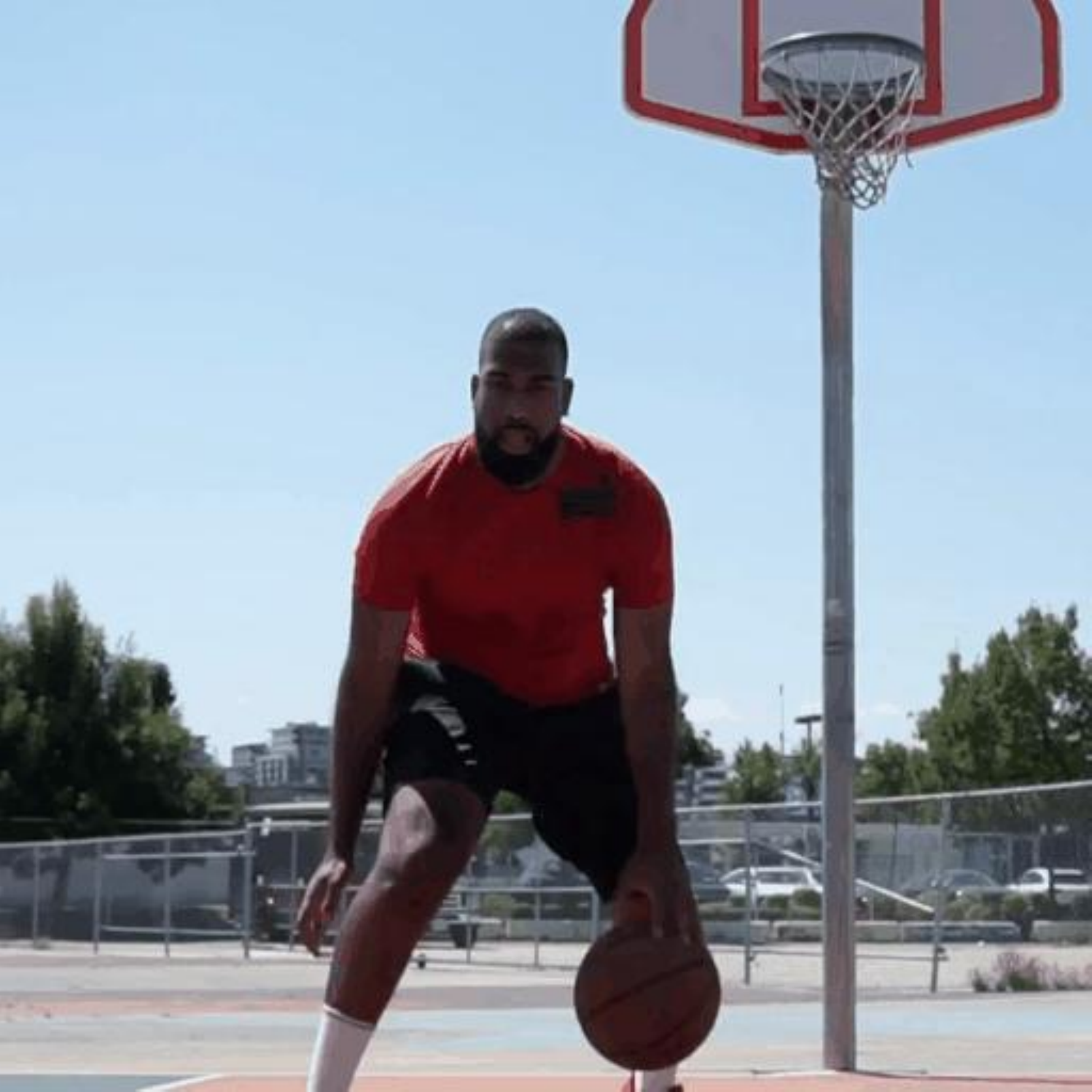}
\includegraphics[width=0.10\textwidth]{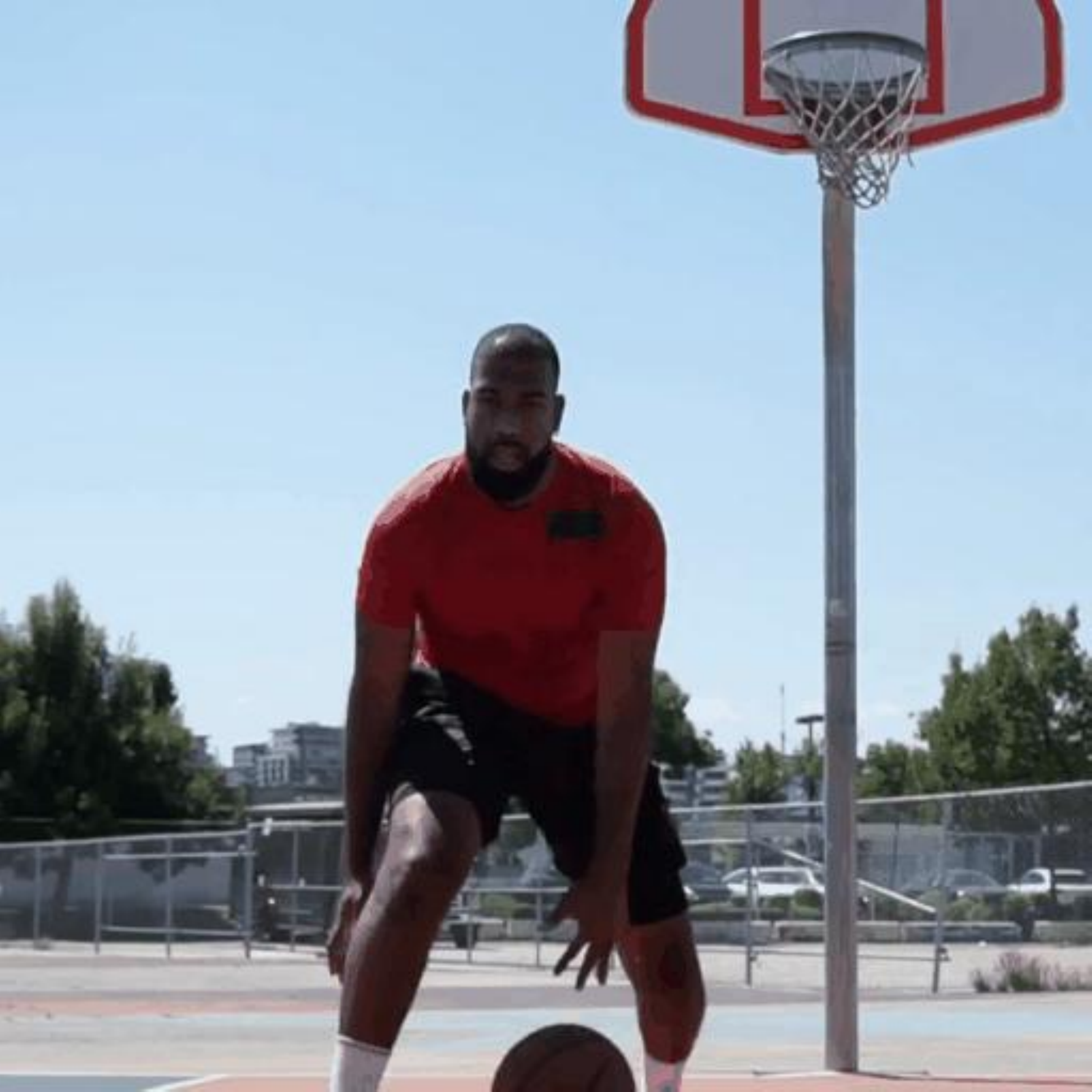}

\makebox[0.12\textwidth]{A \textcolor{blue}{\textbf{Lionel Messi}} is dribbling a basketball.}\\
\includegraphics[width=0.10\textwidth]{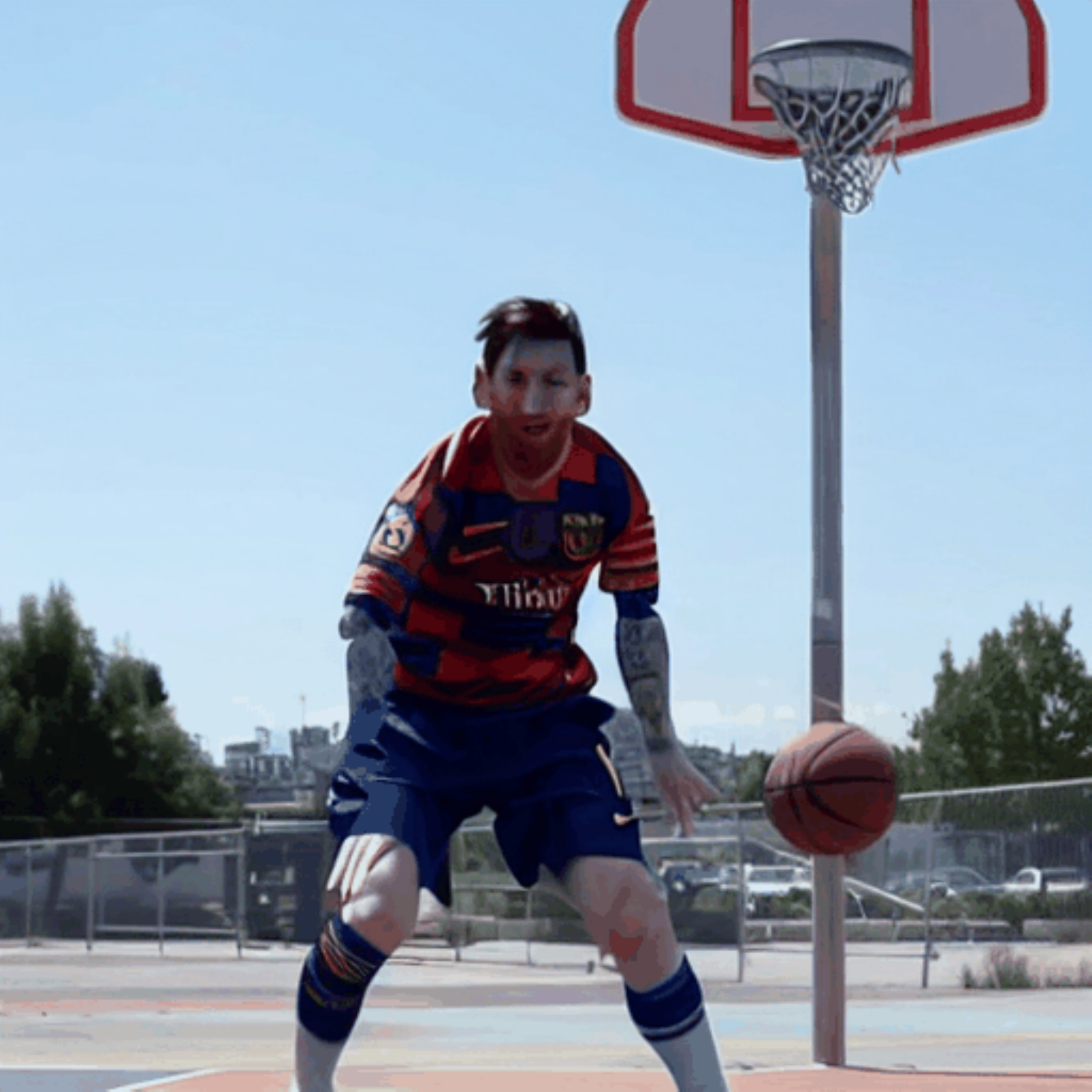}
\includegraphics[width=0.10\textwidth]{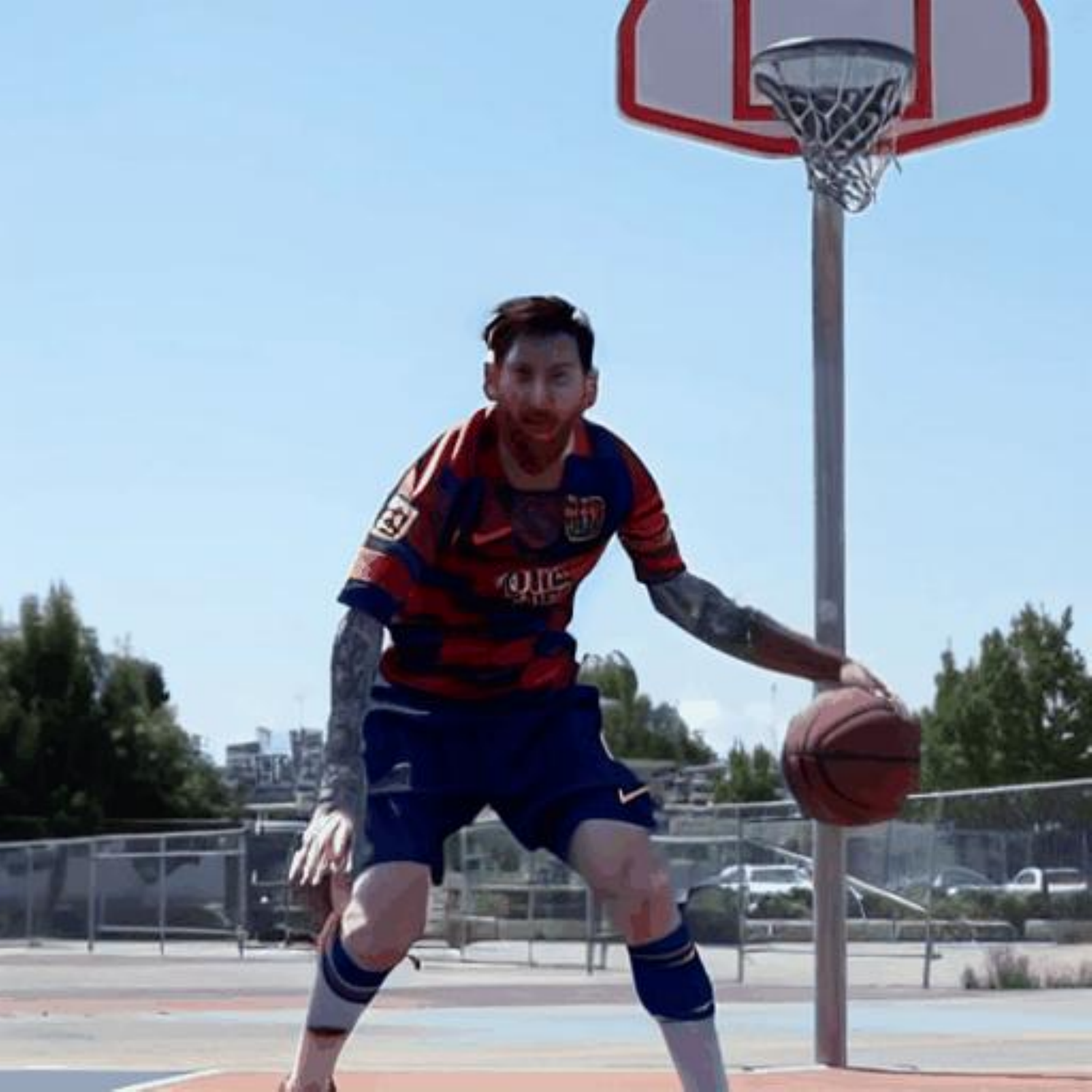}
\includegraphics[width=0.10\textwidth]{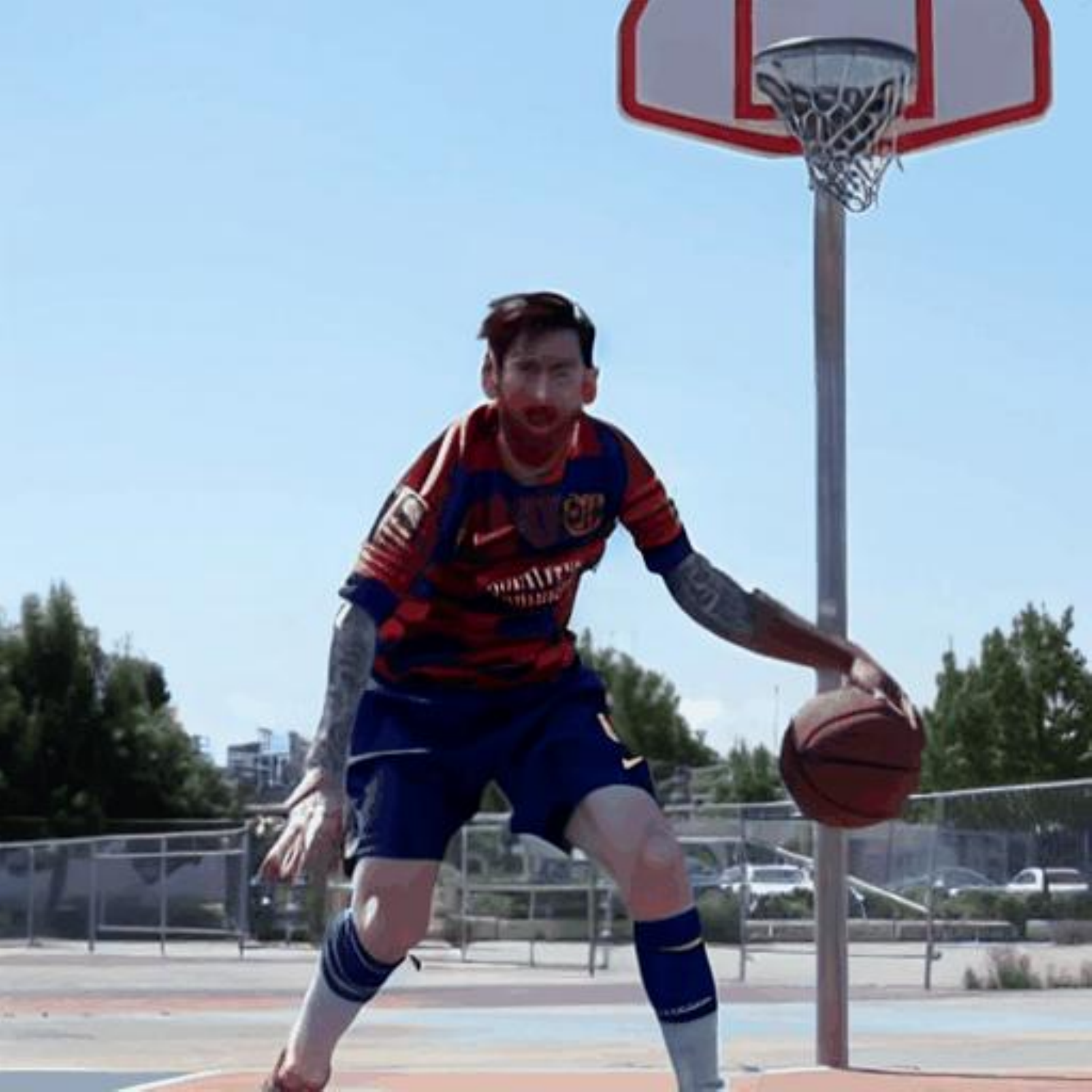}
\includegraphics[width=0.10\textwidth]{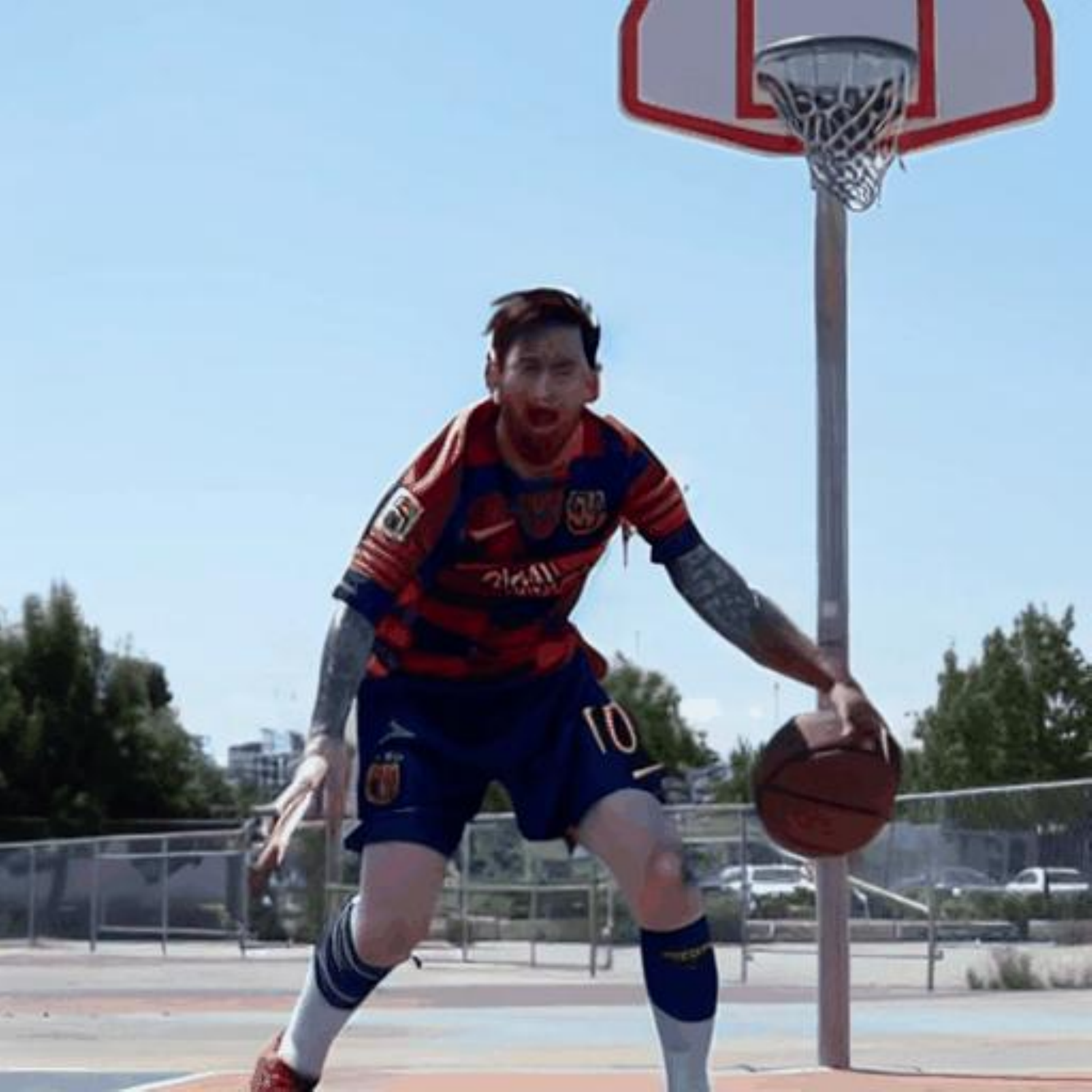}
\includegraphics[width=0.10\textwidth]{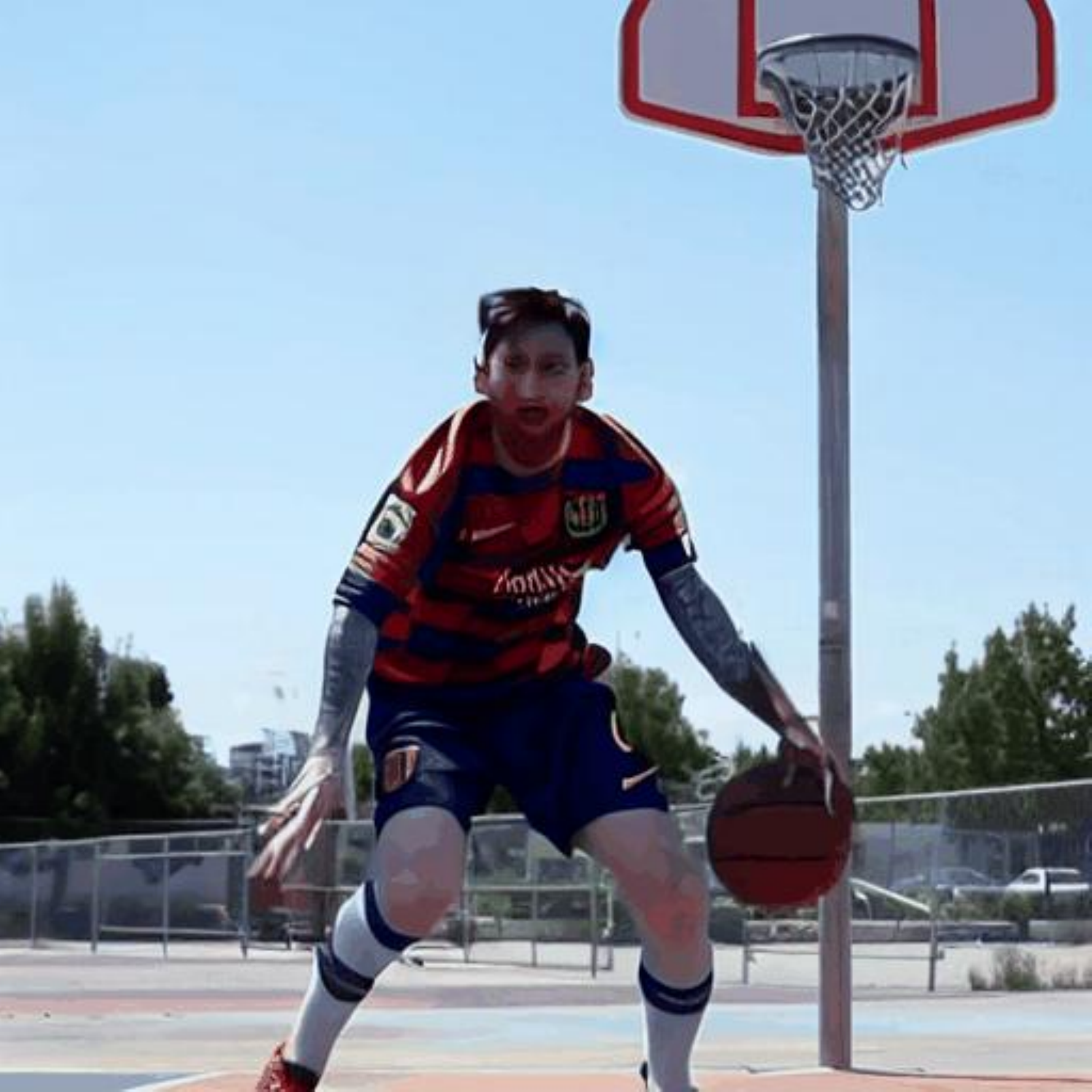}
\includegraphics[width=0.10\textwidth]{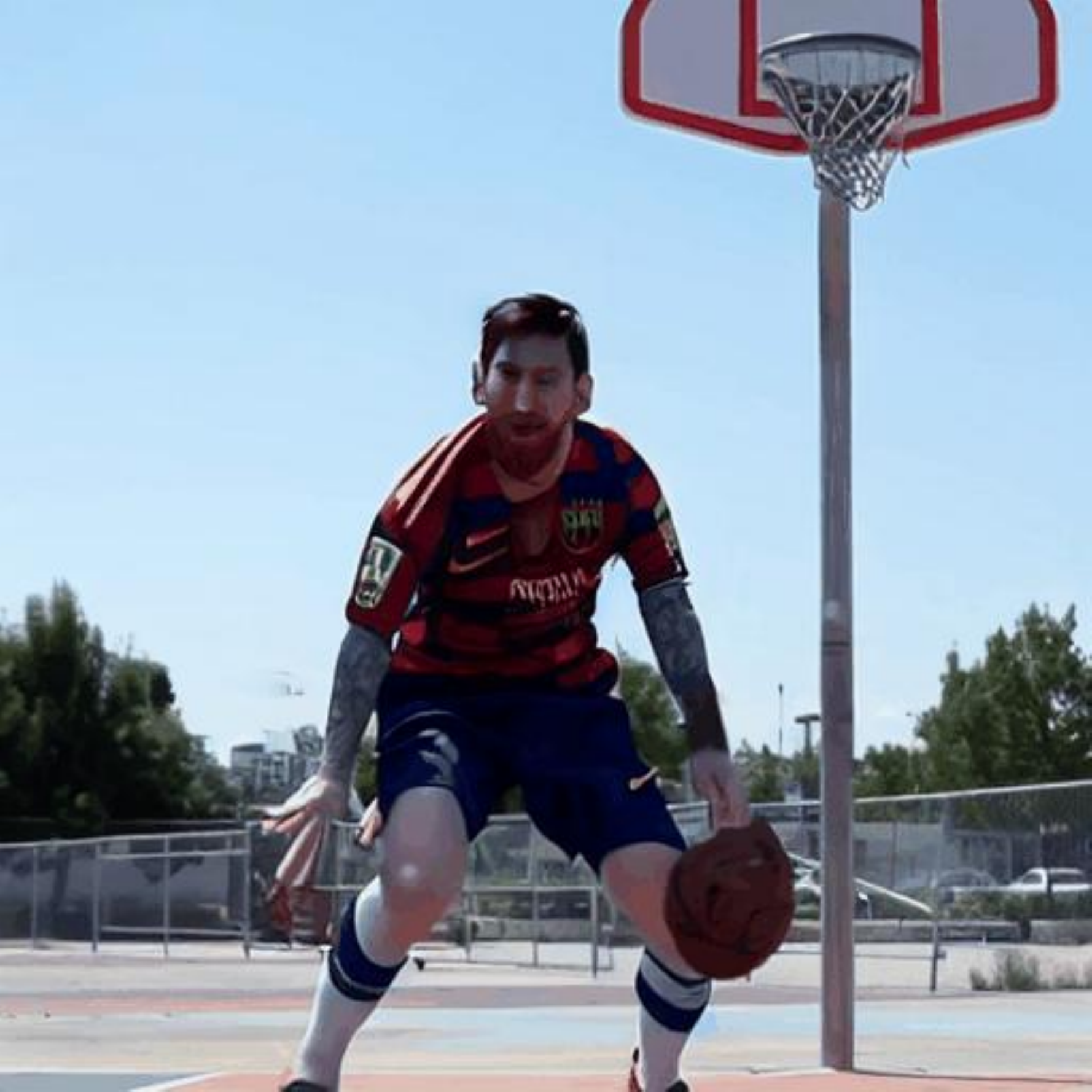}
\includegraphics[width=0.10\textwidth]{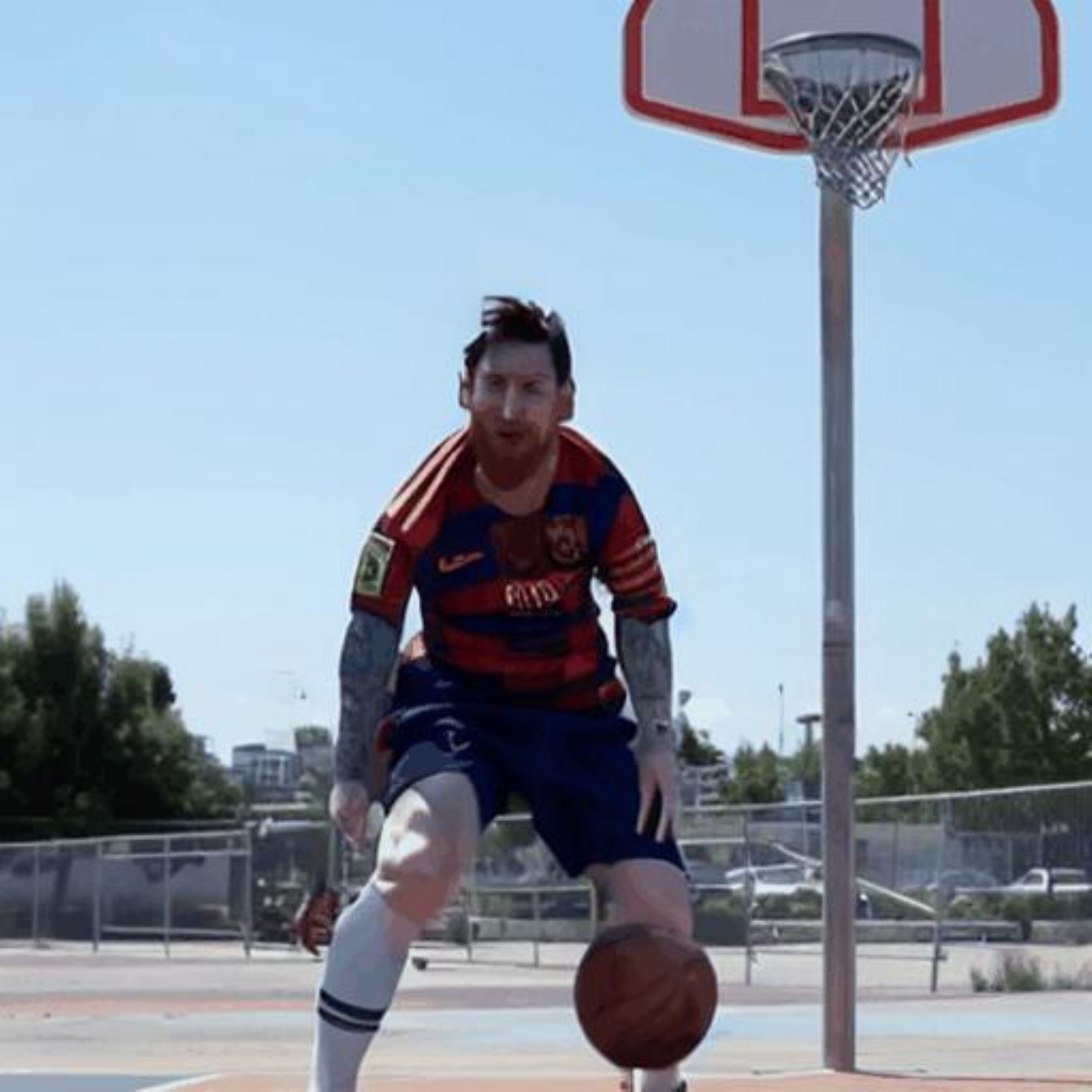}
\includegraphics[width=0.10\textwidth]{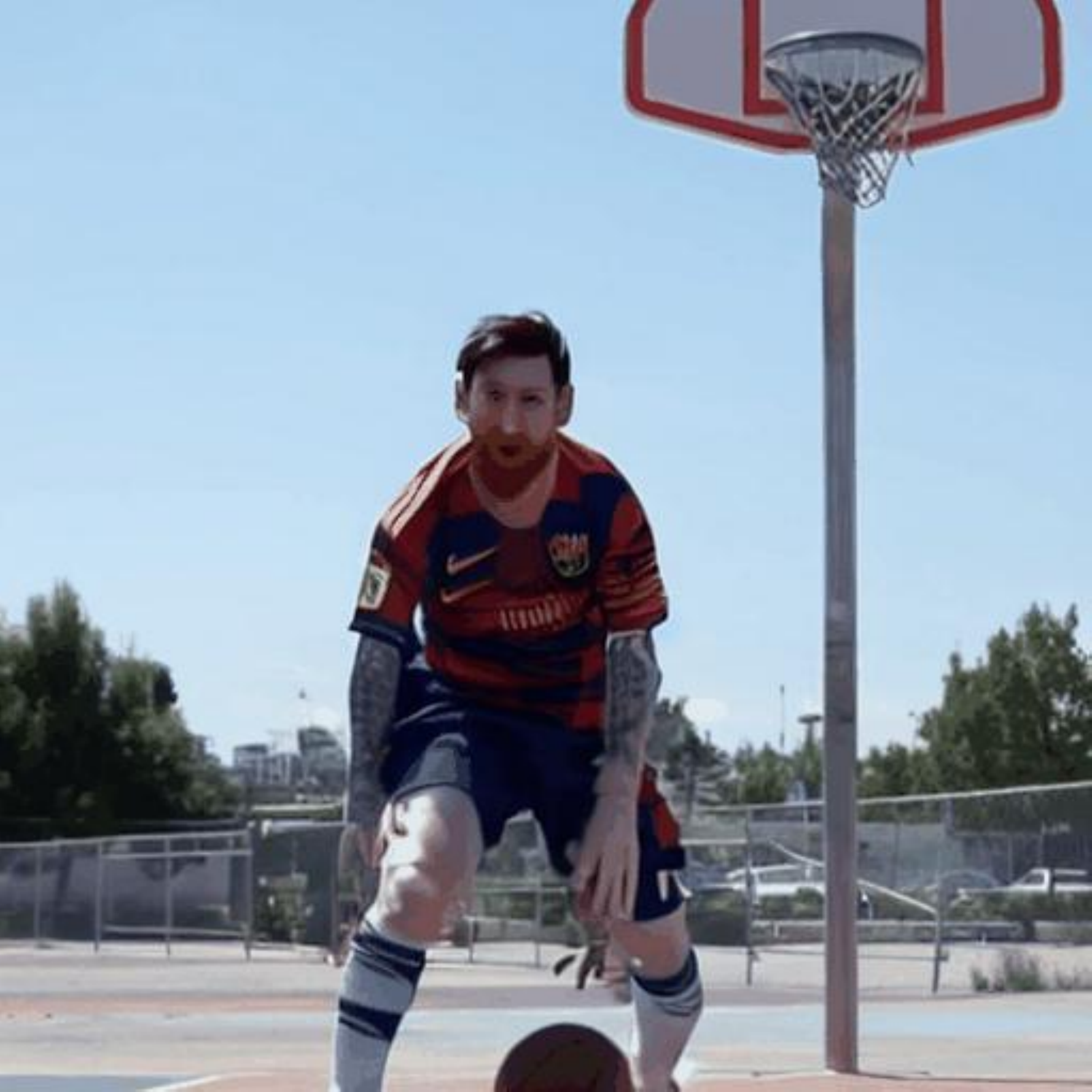}

\makebox[0.12\textwidth]{A \textcolor{blue}{\textbf{racoon}} is dribbling a basketball.}\\
\includegraphics[width=0.10\textwidth]{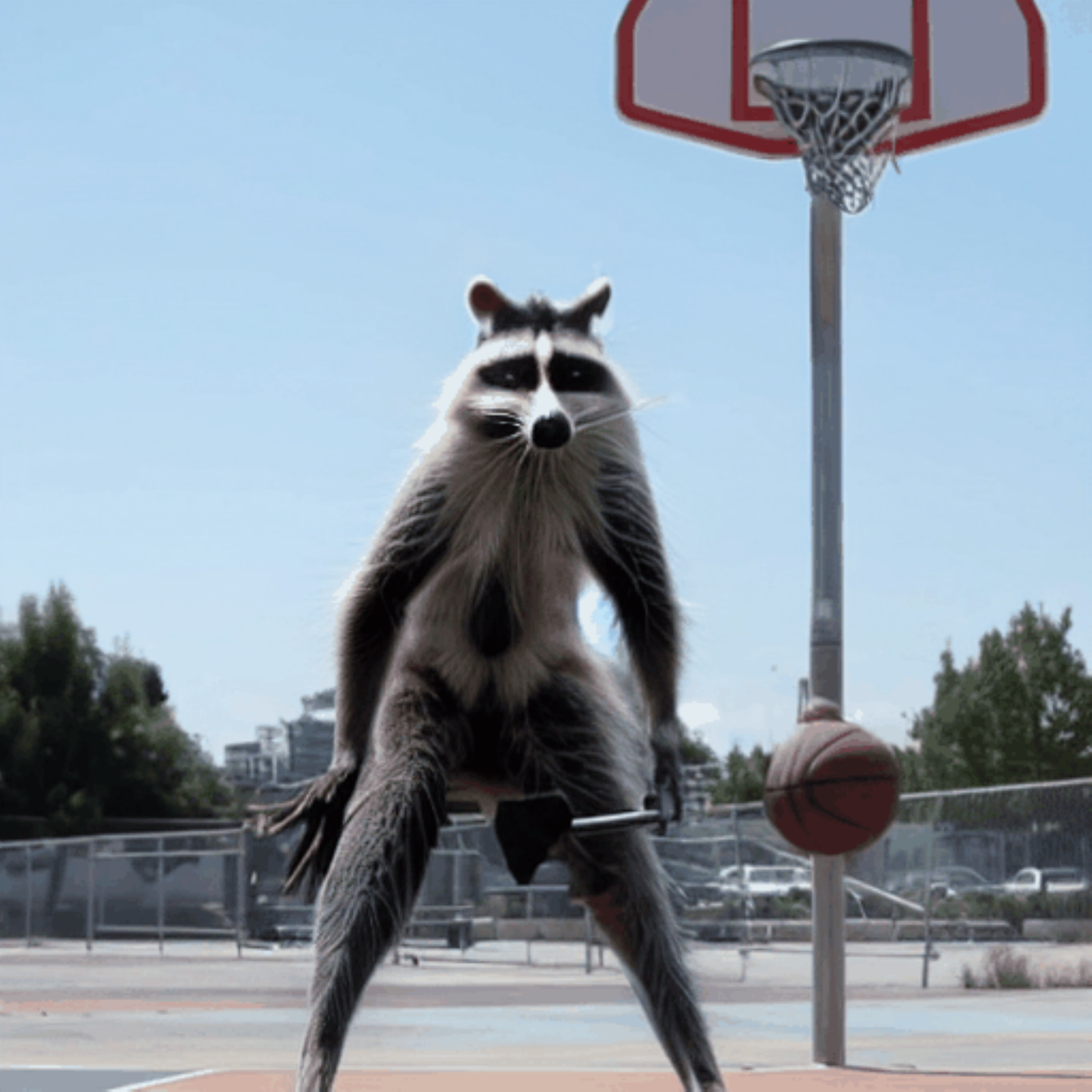}
\includegraphics[width=0.10\textwidth]{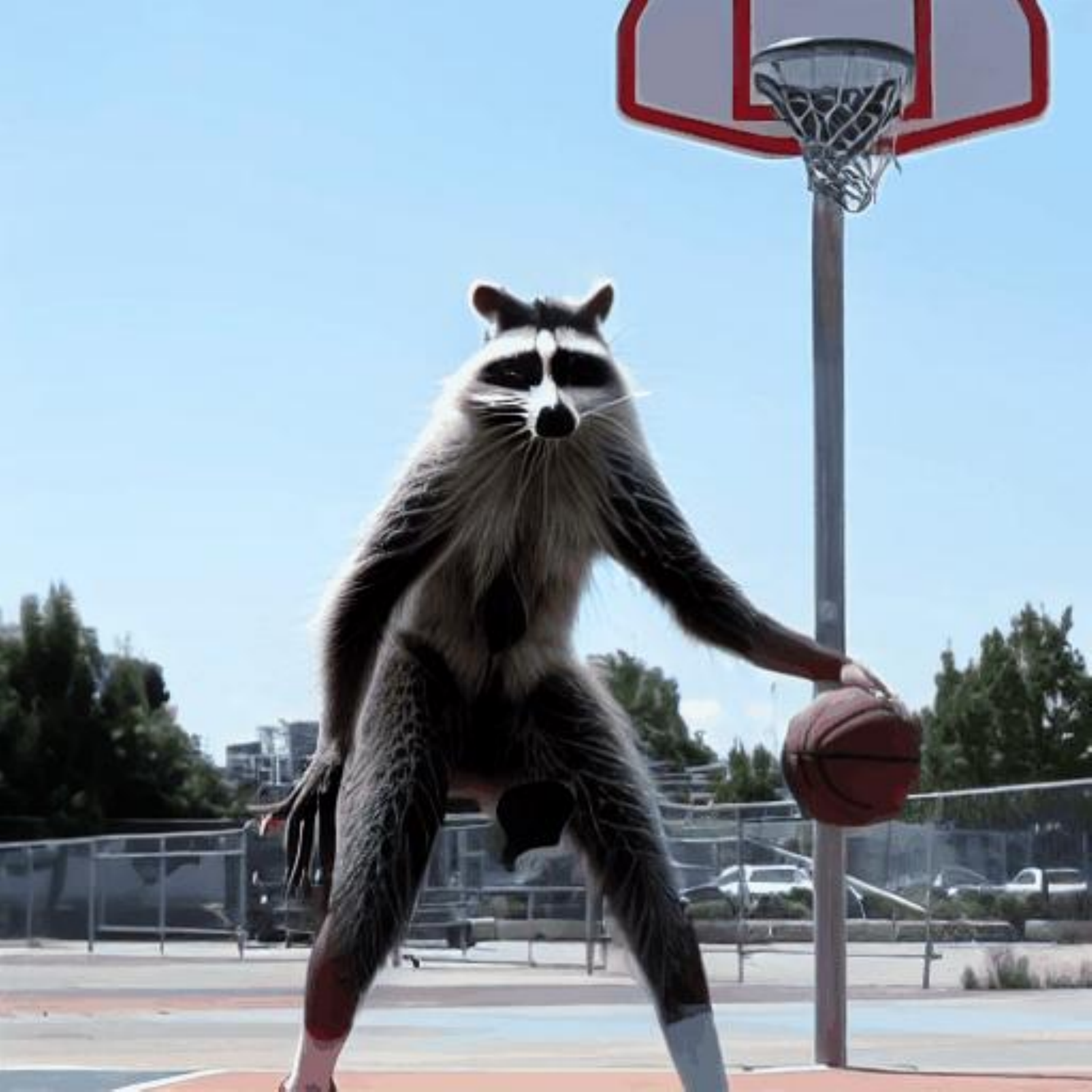}
\includegraphics[width=0.10\textwidth]{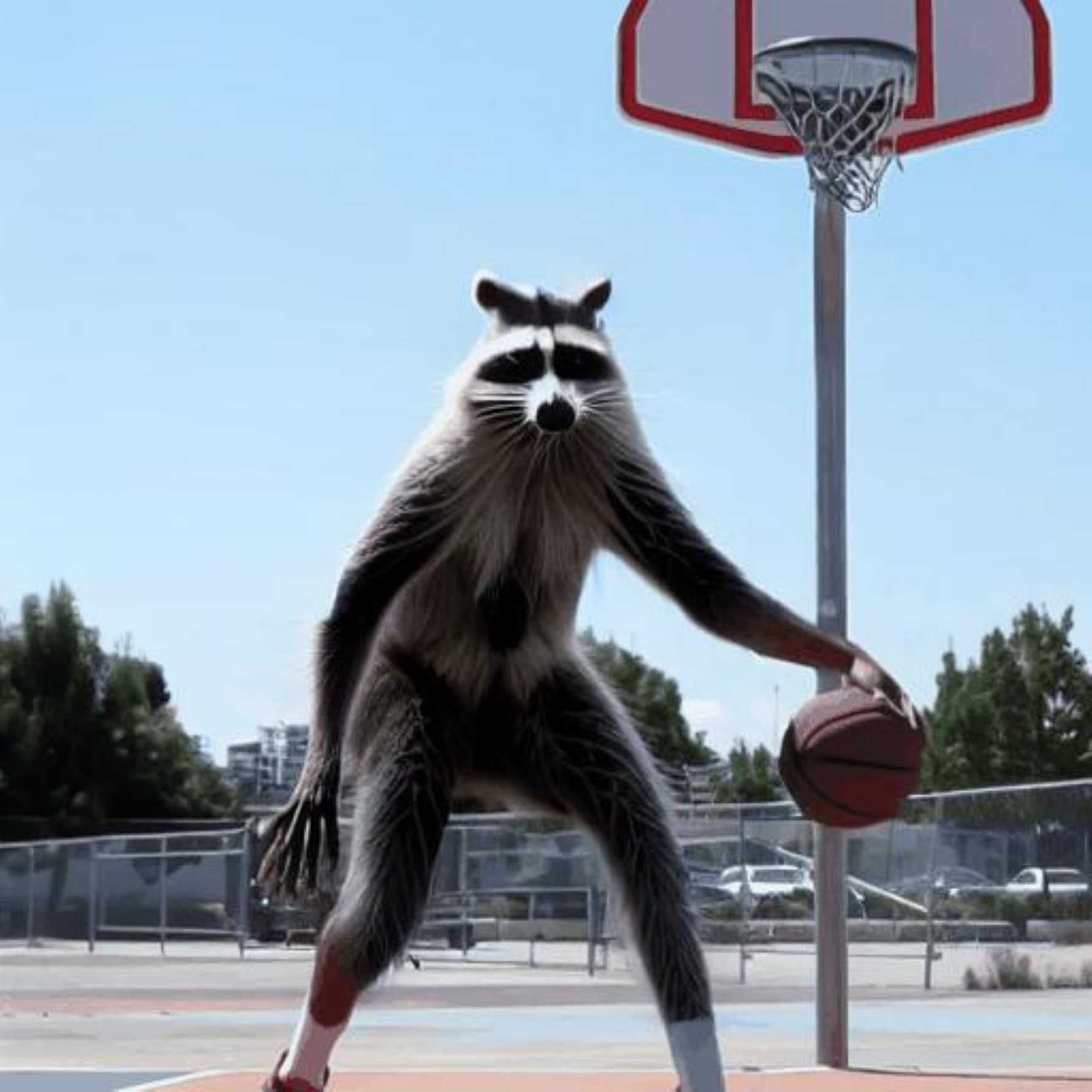}
\includegraphics[width=0.10\textwidth]{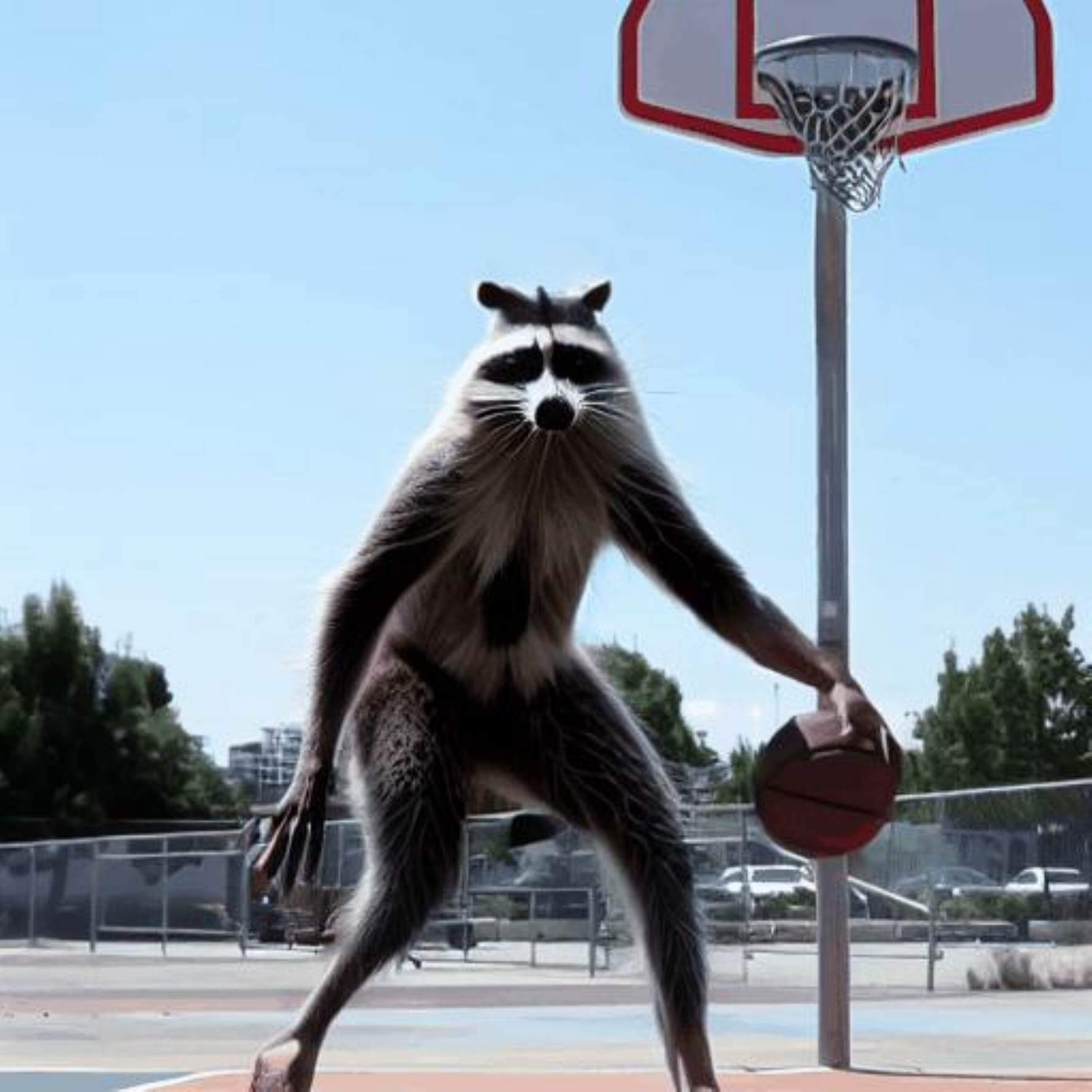}
\includegraphics[width=0.10\textwidth]{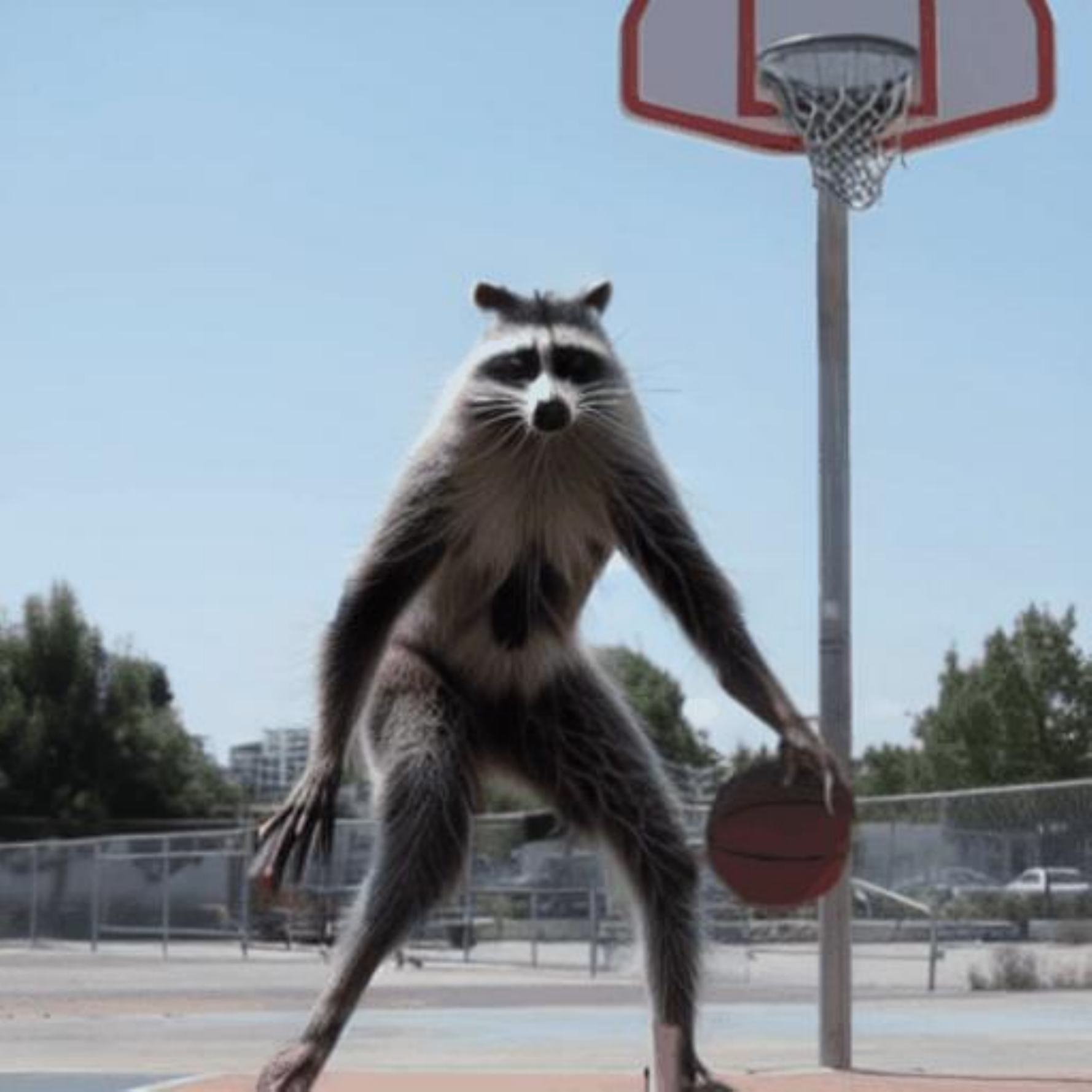}
\includegraphics[width=0.10\textwidth]{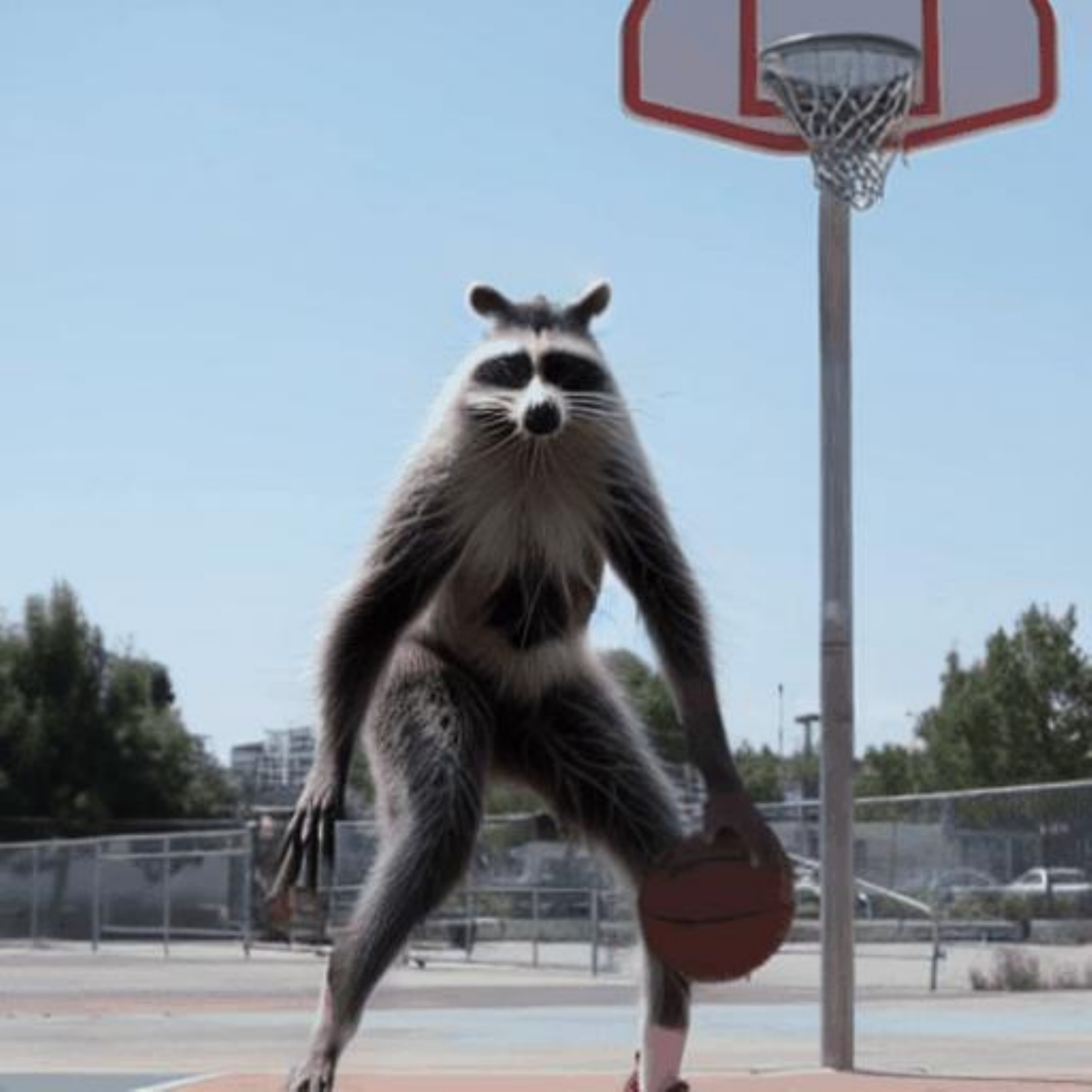}
\includegraphics[width=0.10\textwidth]{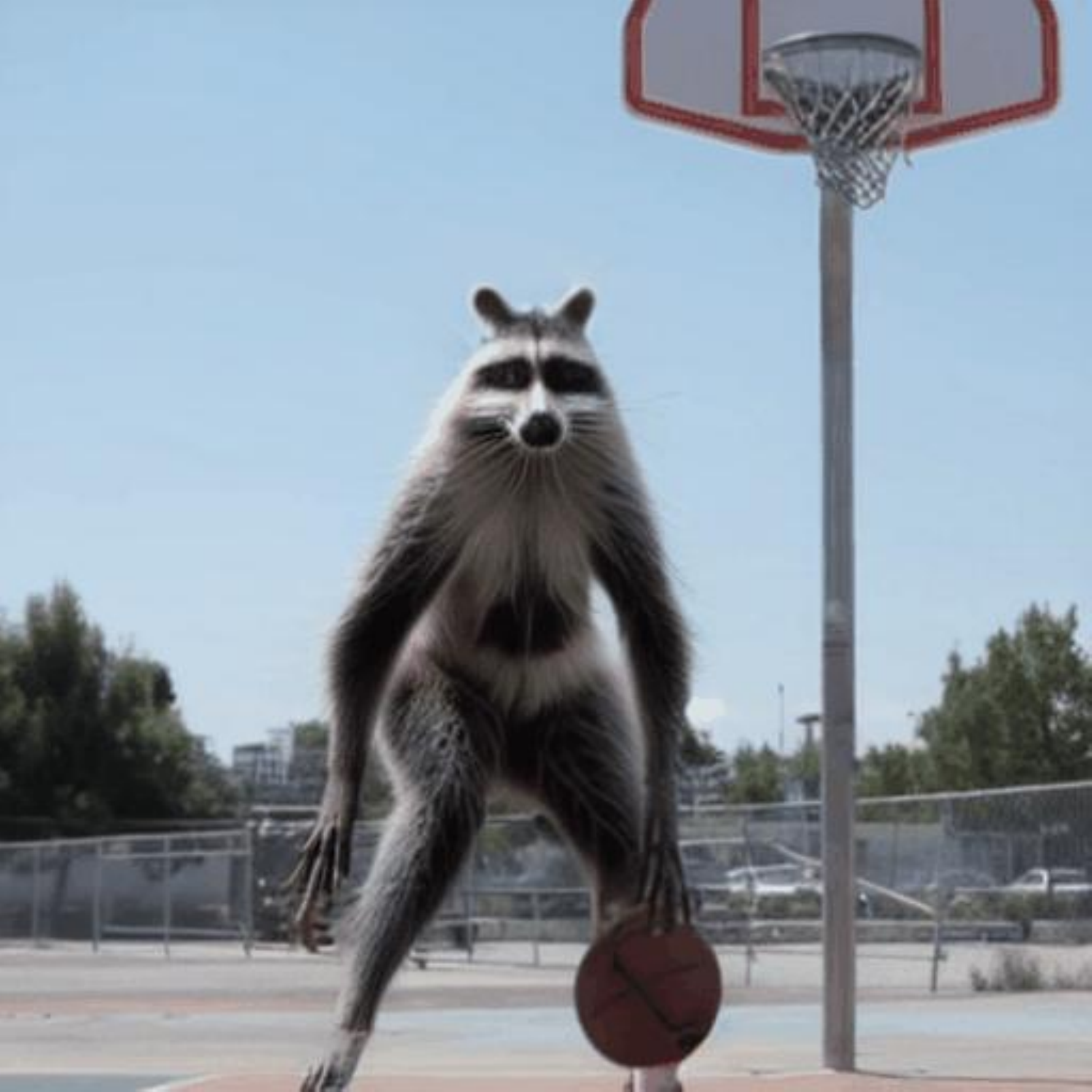}
\includegraphics[width=0.10\textwidth]{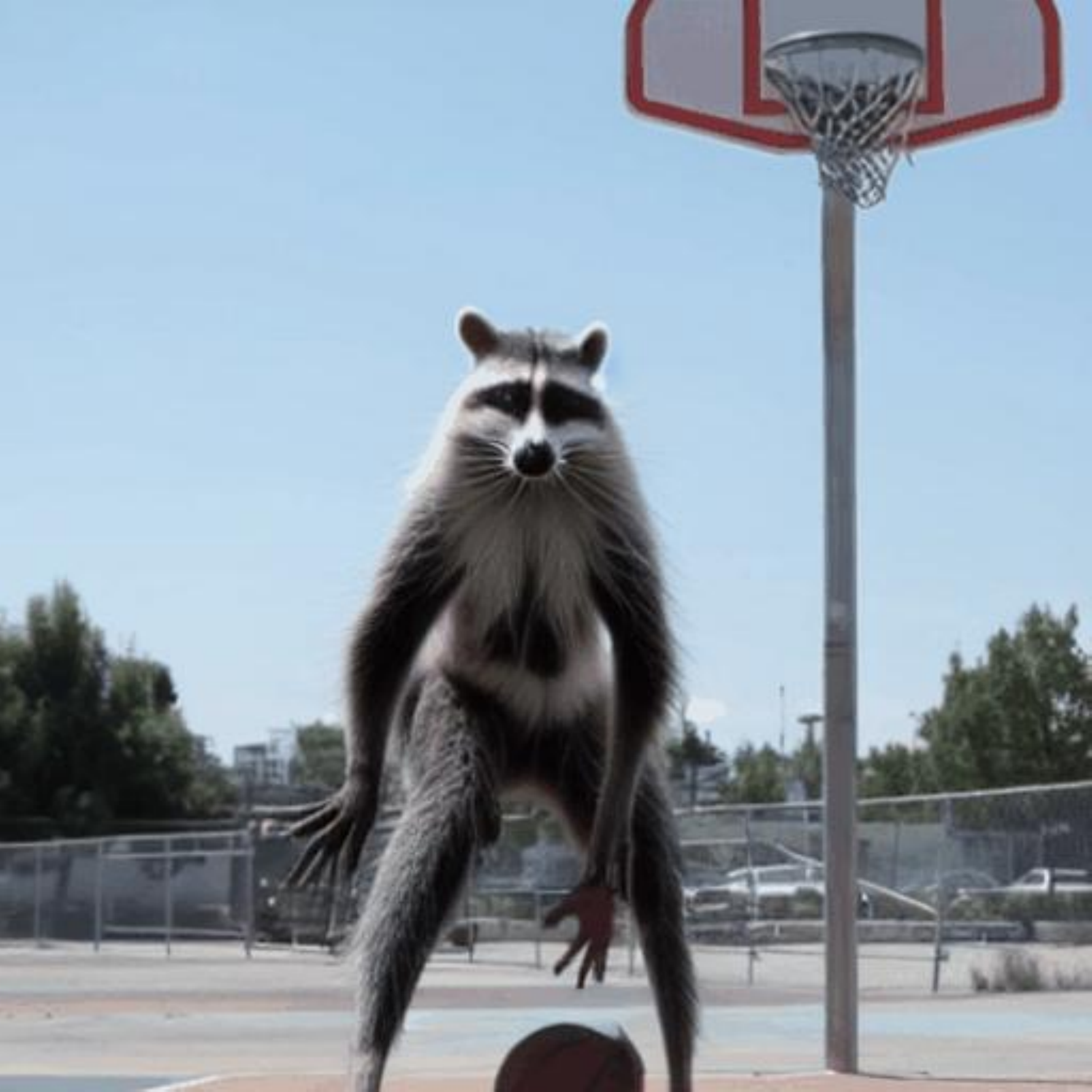}

\makebox[0.12\textwidth]{A \textcolor{blue}{\textbf{oil painting}} that a man is dribbling a basketball.}\\
\includegraphics[width=0.10\textwidth]{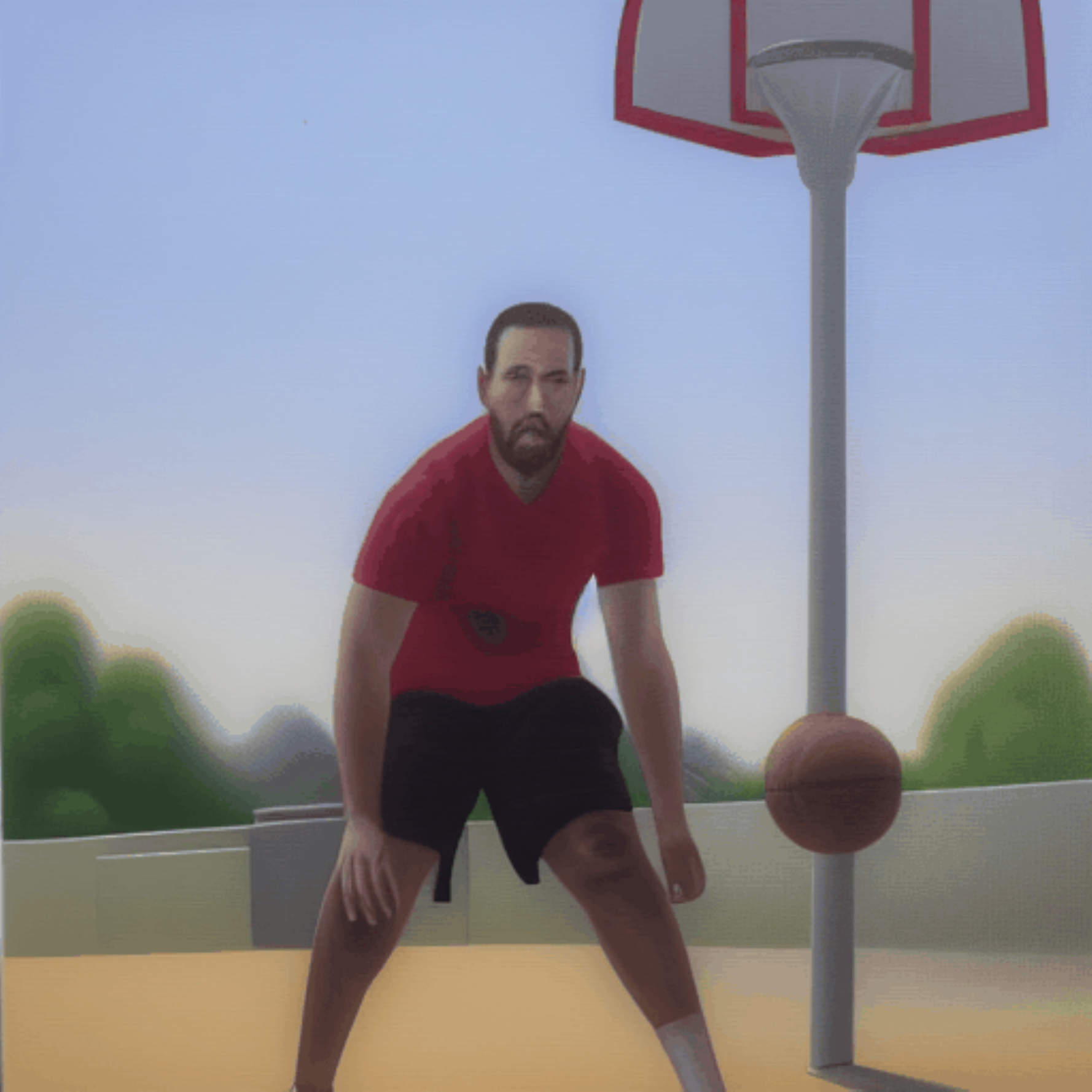}
\includegraphics[width=0.10\textwidth]{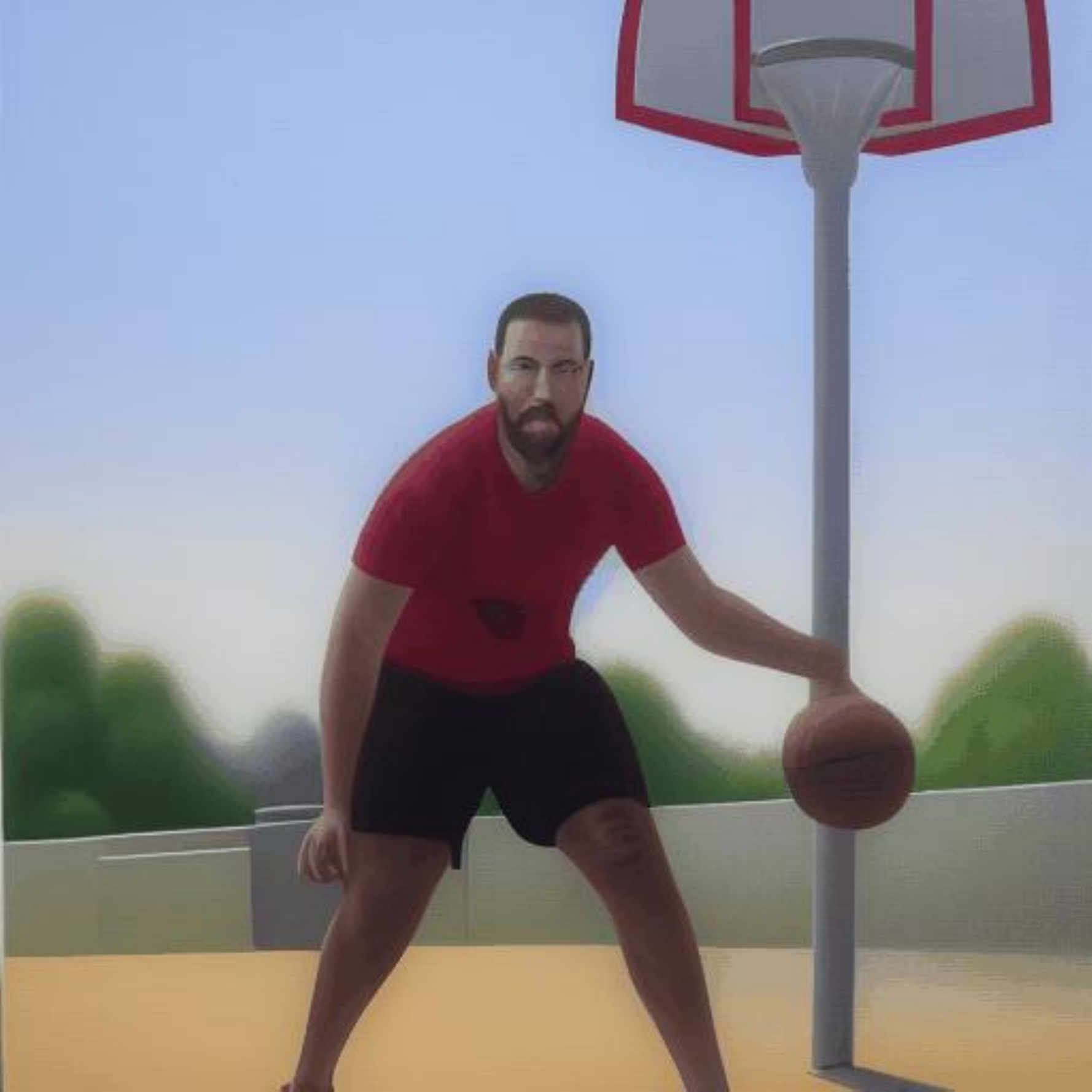}
\includegraphics[width=0.10\textwidth]{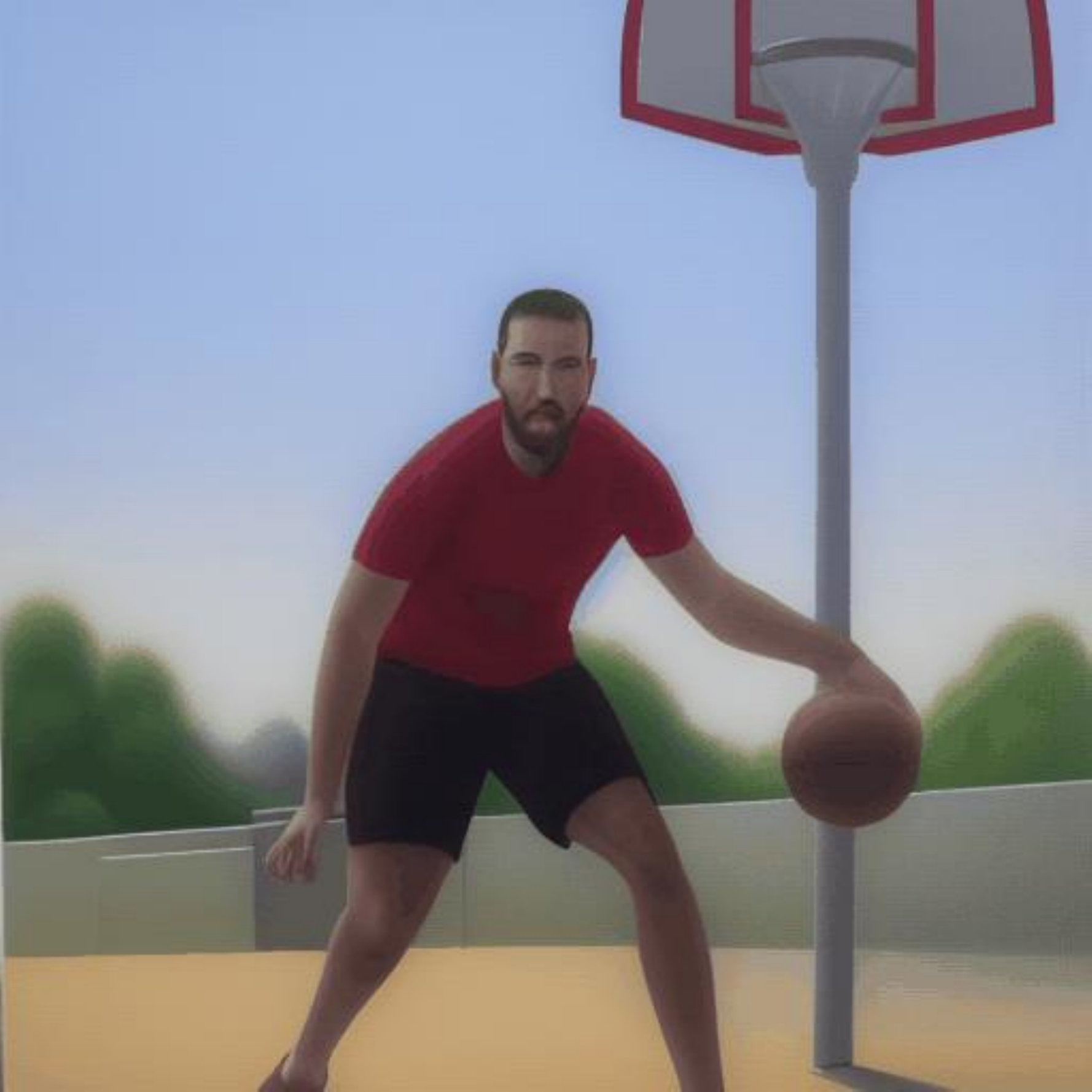}
\includegraphics[width=0.10\textwidth]{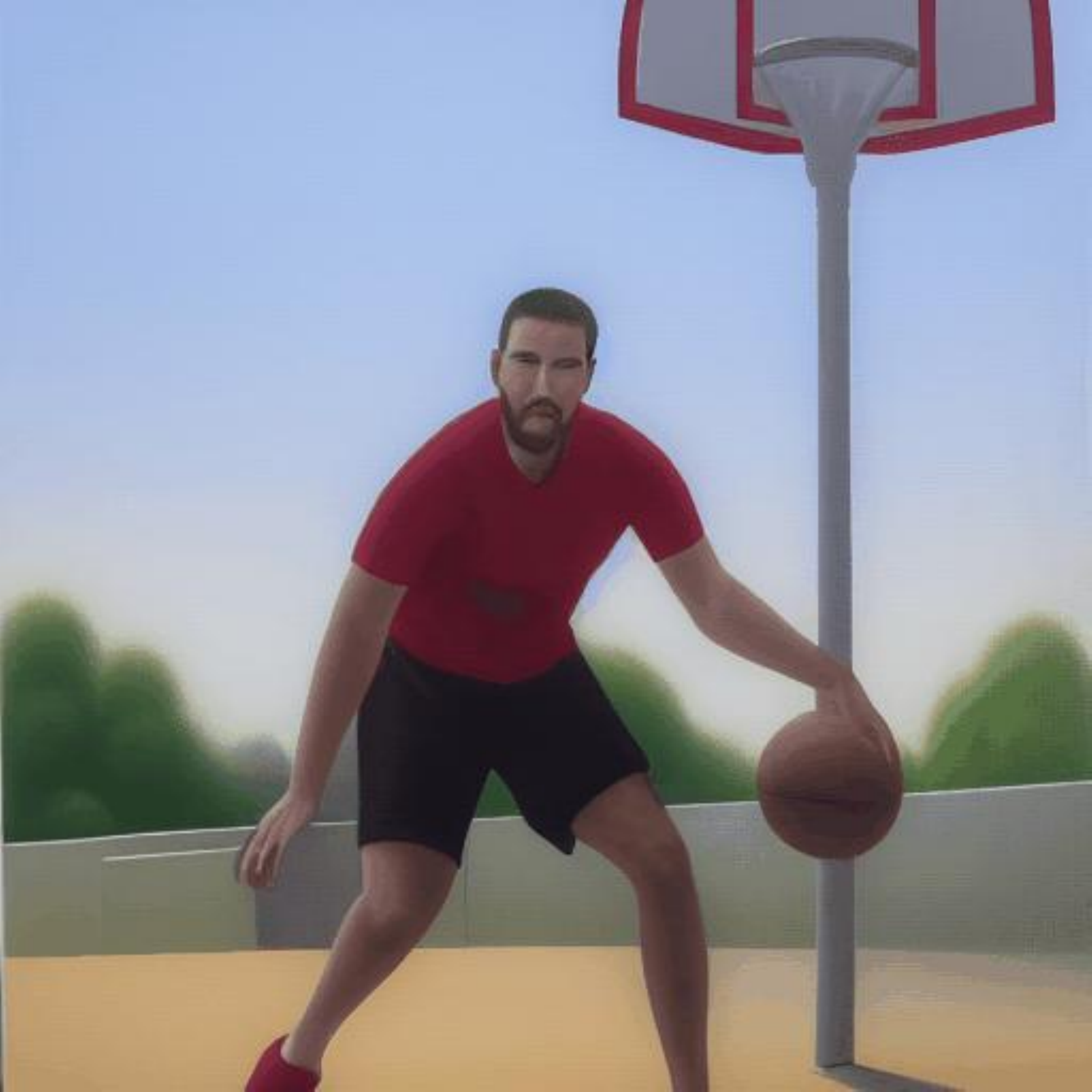}
\includegraphics[width=0.10\textwidth]{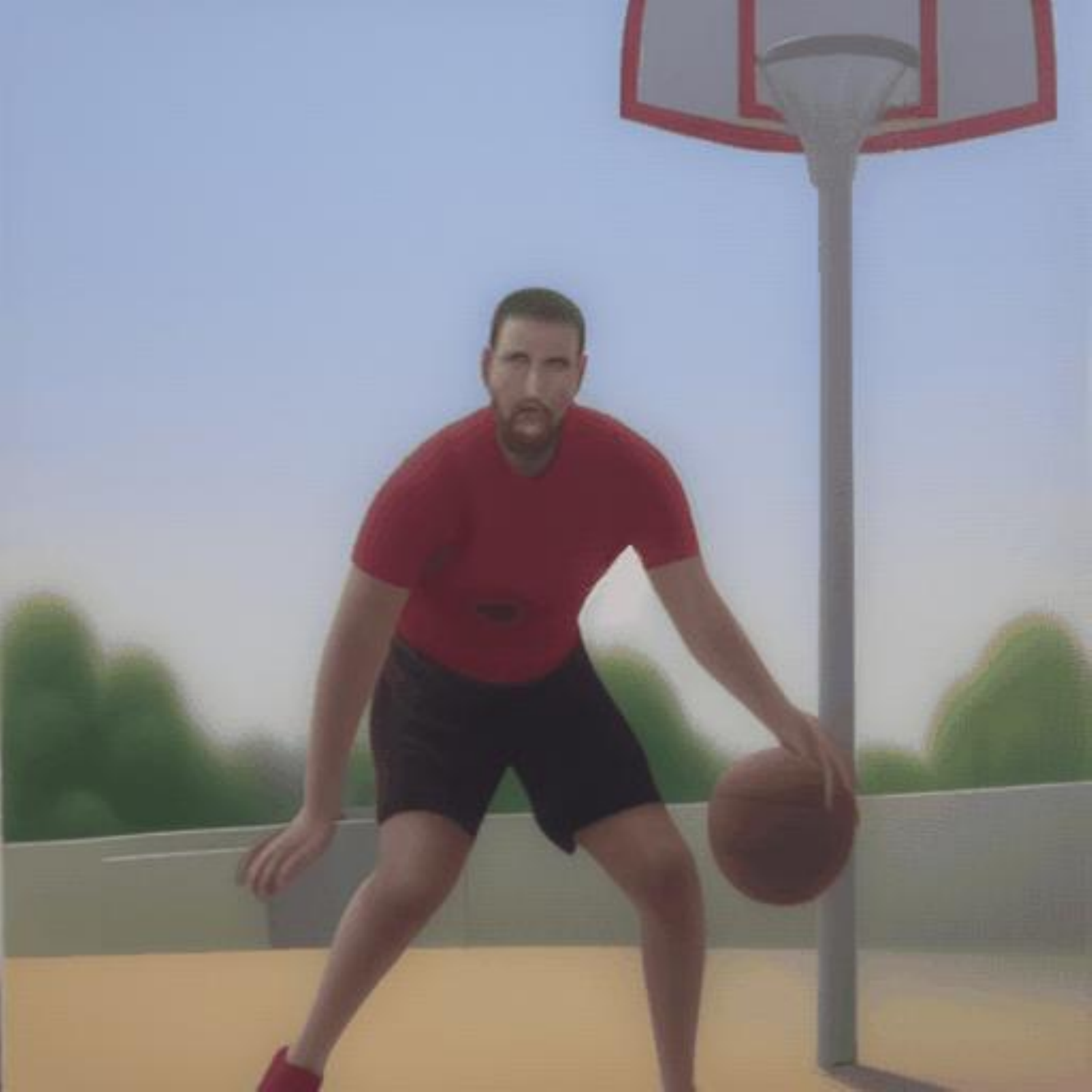}
\includegraphics[width=0.10\textwidth]{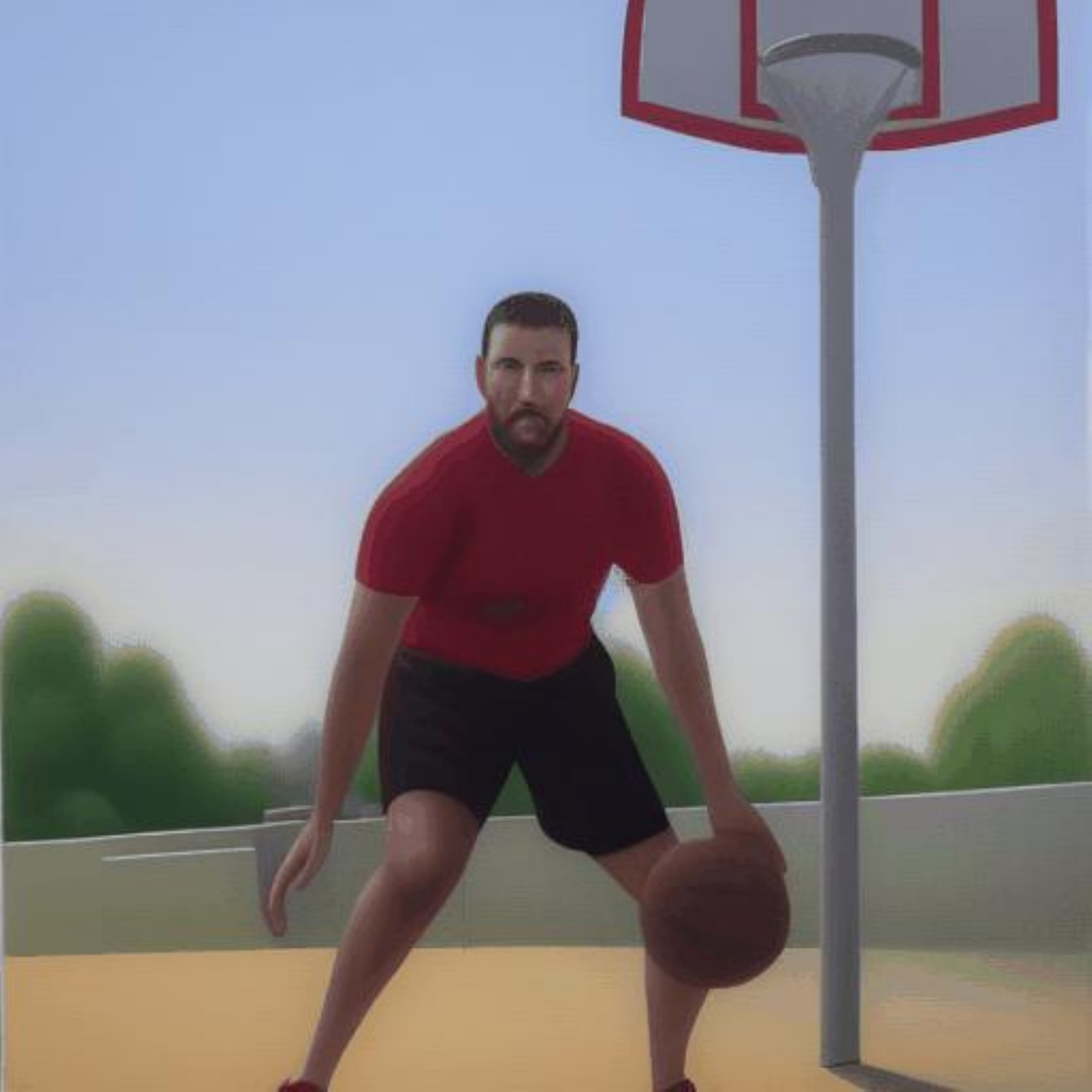}
\includegraphics[width=0.10\textwidth]{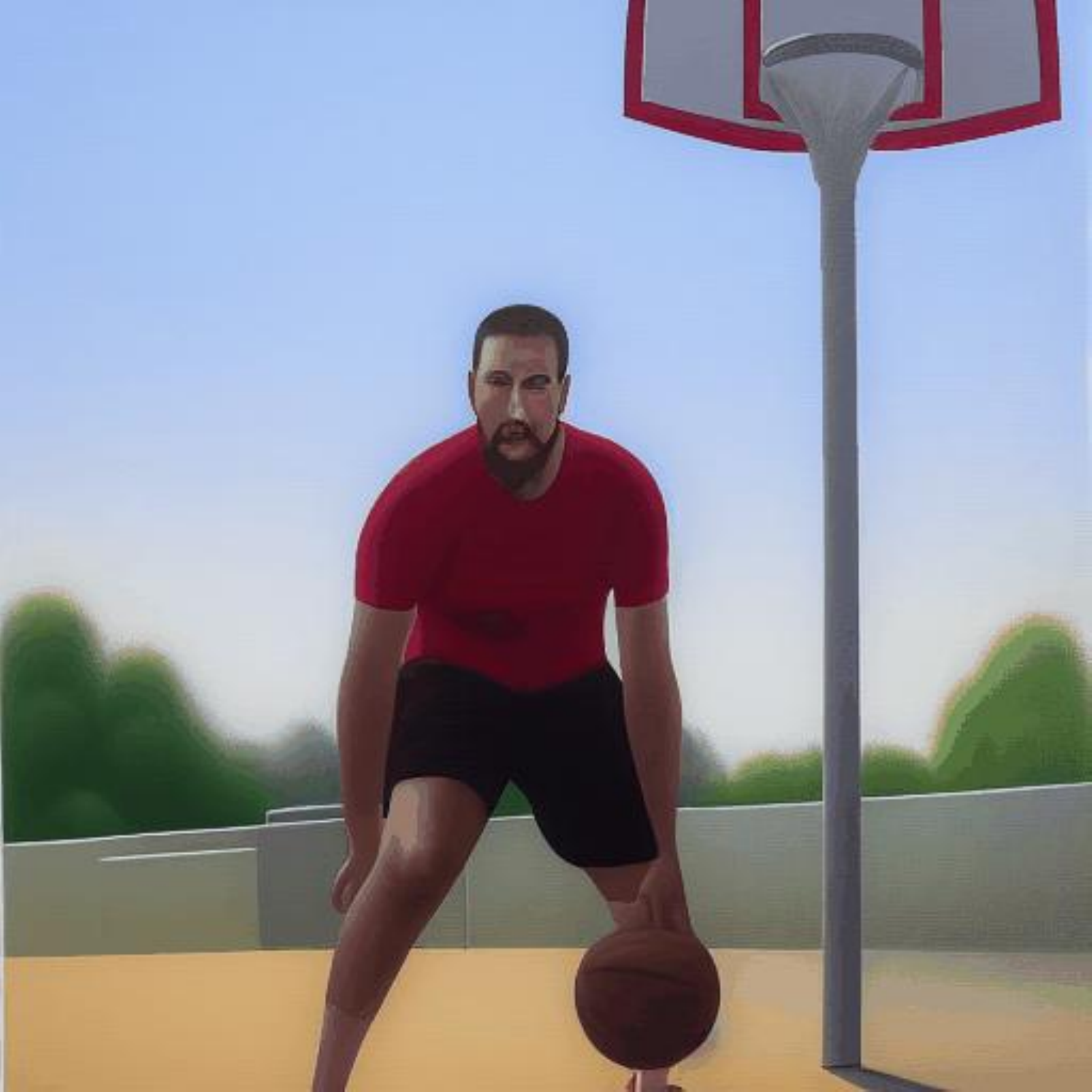}
\includegraphics[width=0.10\textwidth]{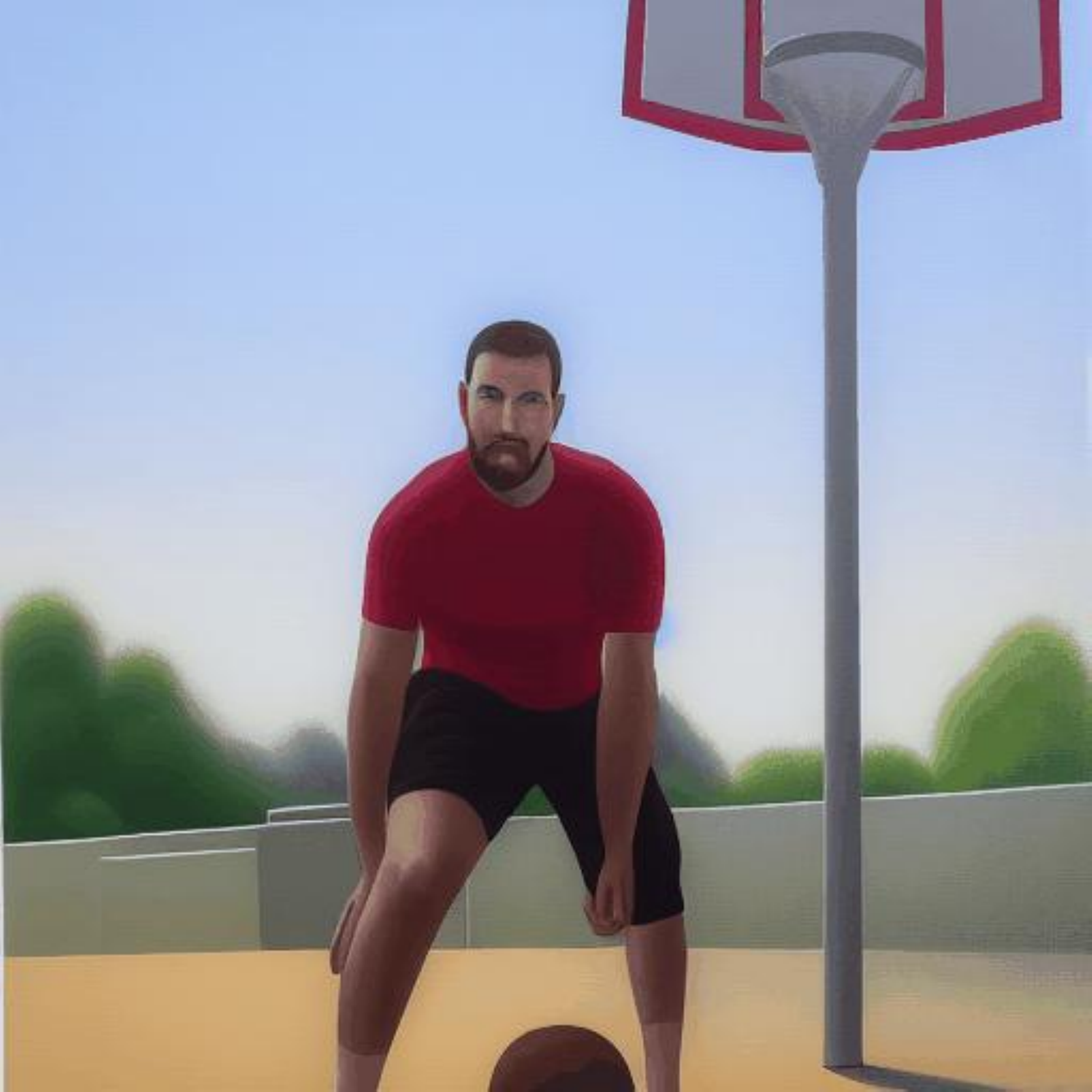}

\makebox[0.12\textwidth]{A man is dribbling a basketball, \textcolor{blue}{\textbf{Renoir style}}.}\\
\includegraphics[width=0.10\textwidth]{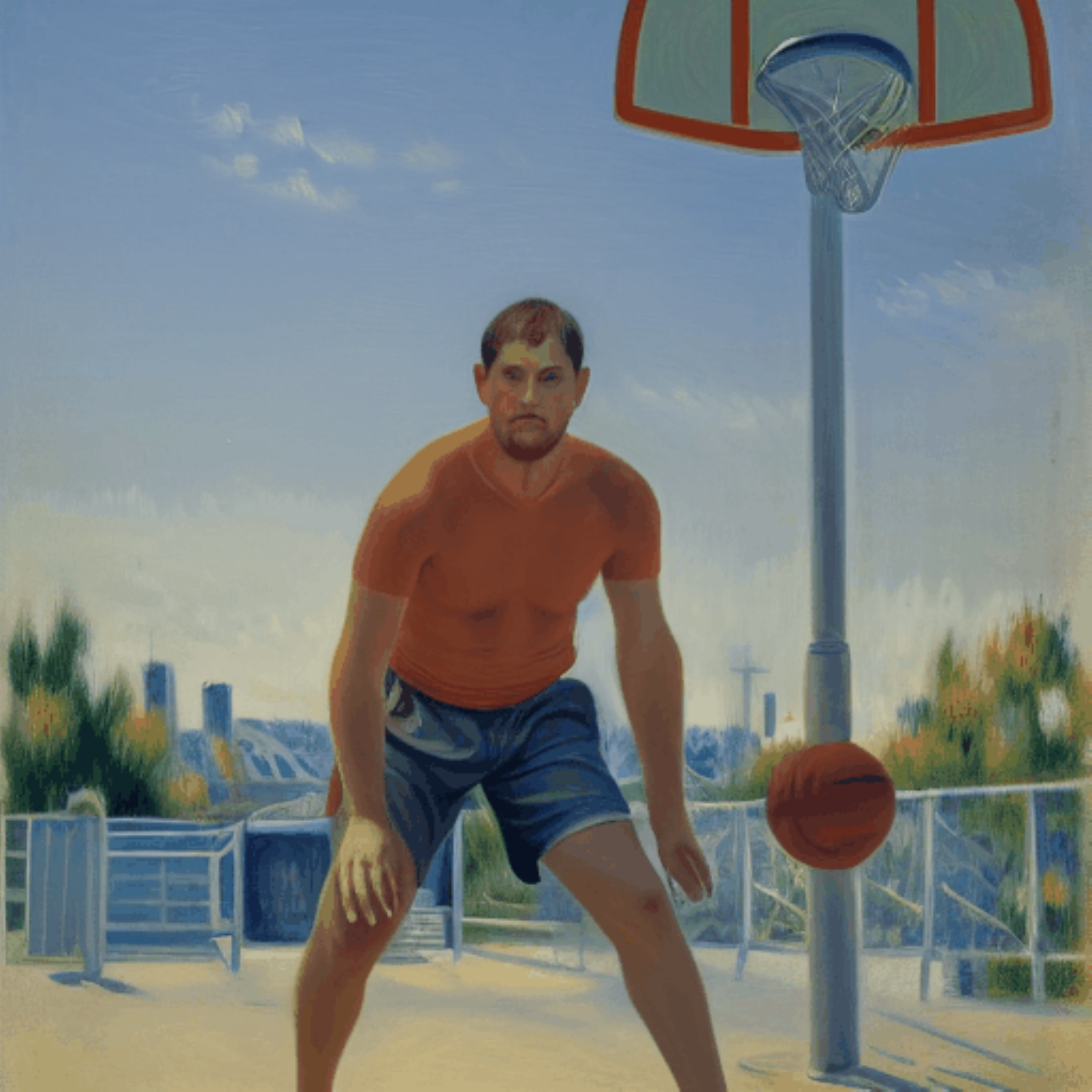}
\includegraphics[width=0.10\textwidth]{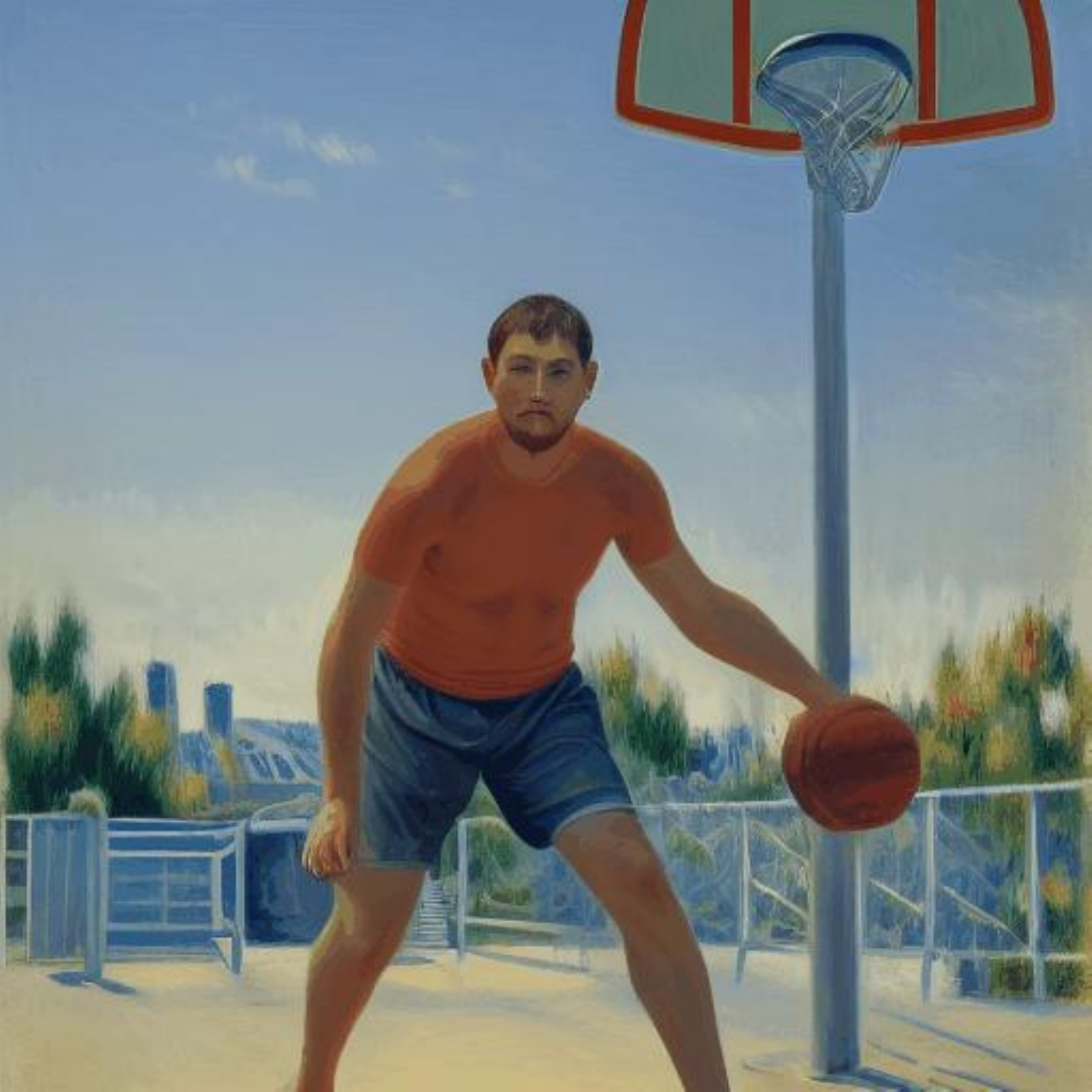}
\includegraphics[width=0.10\textwidth]{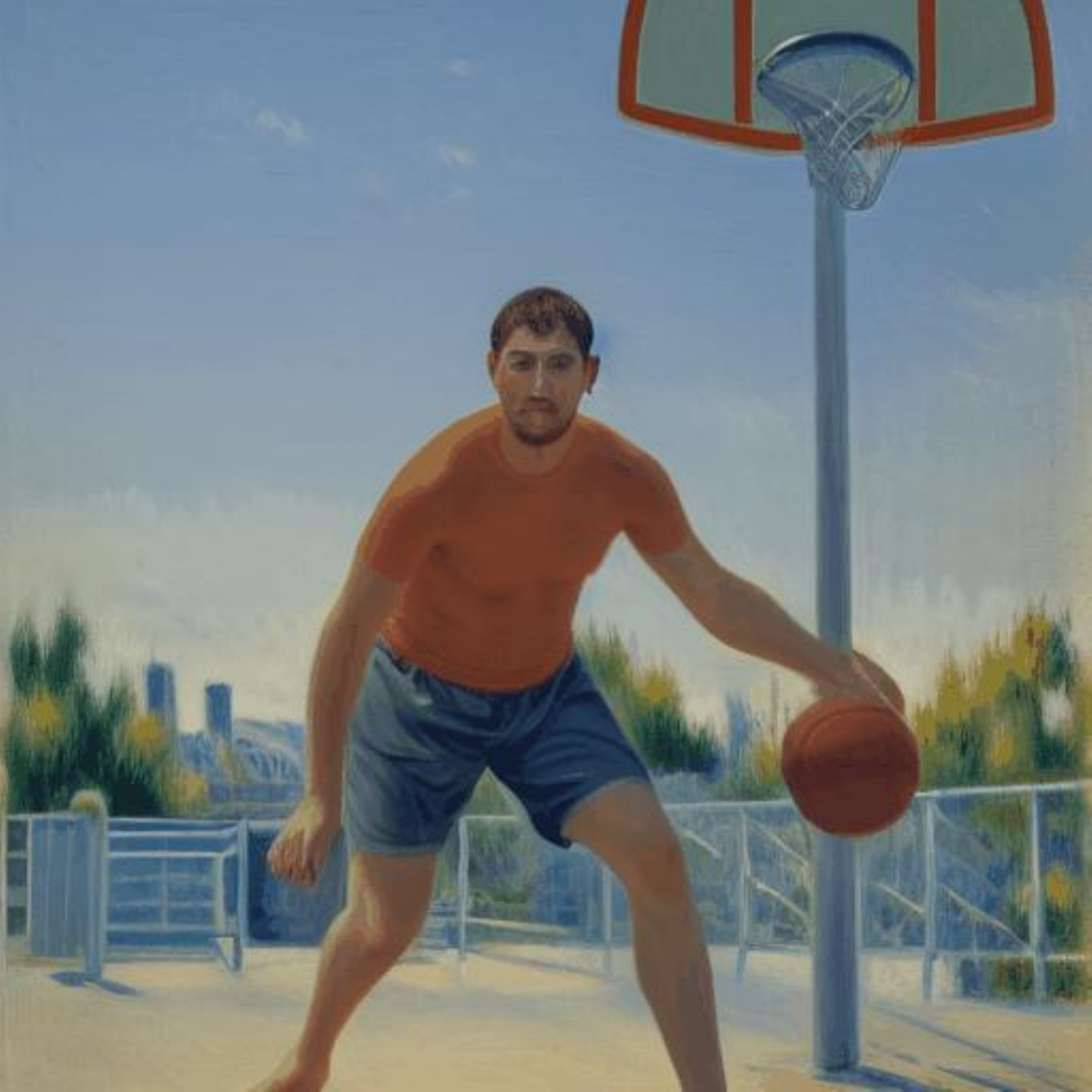}
\includegraphics[width=0.10\textwidth]{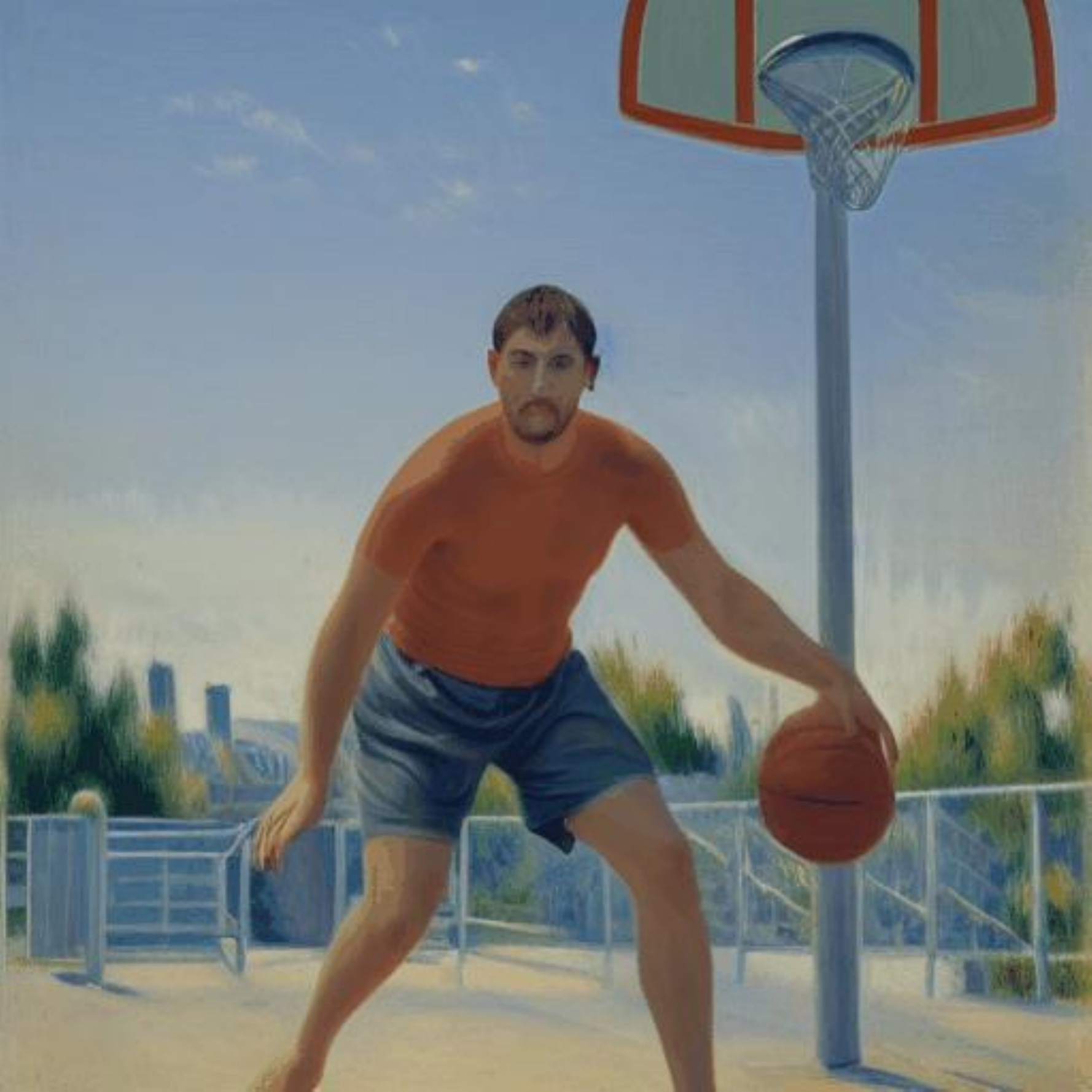}
\includegraphics[width=0.10\textwidth]{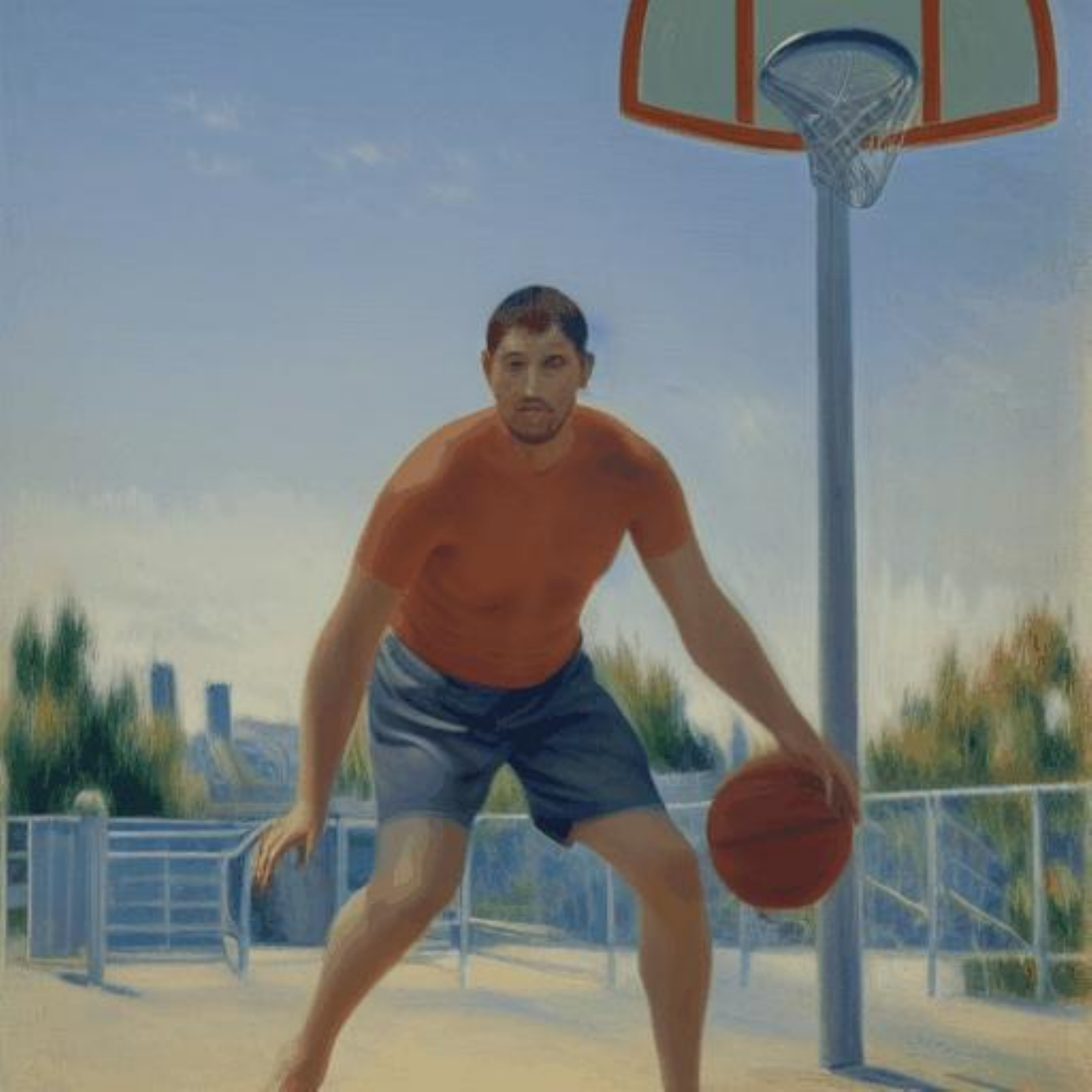}
\includegraphics[width=0.10\textwidth]{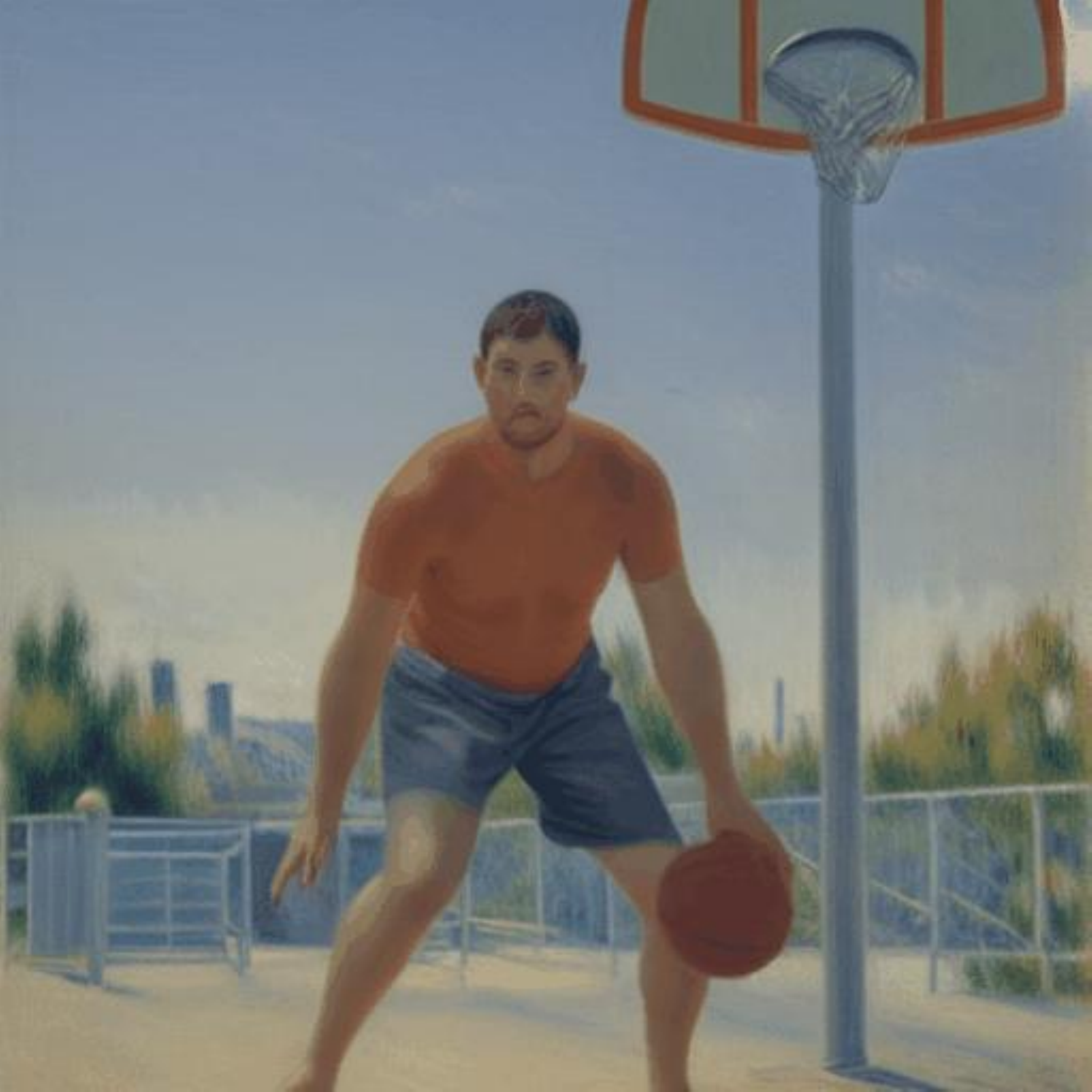}
\includegraphics[width=0.10\textwidth]{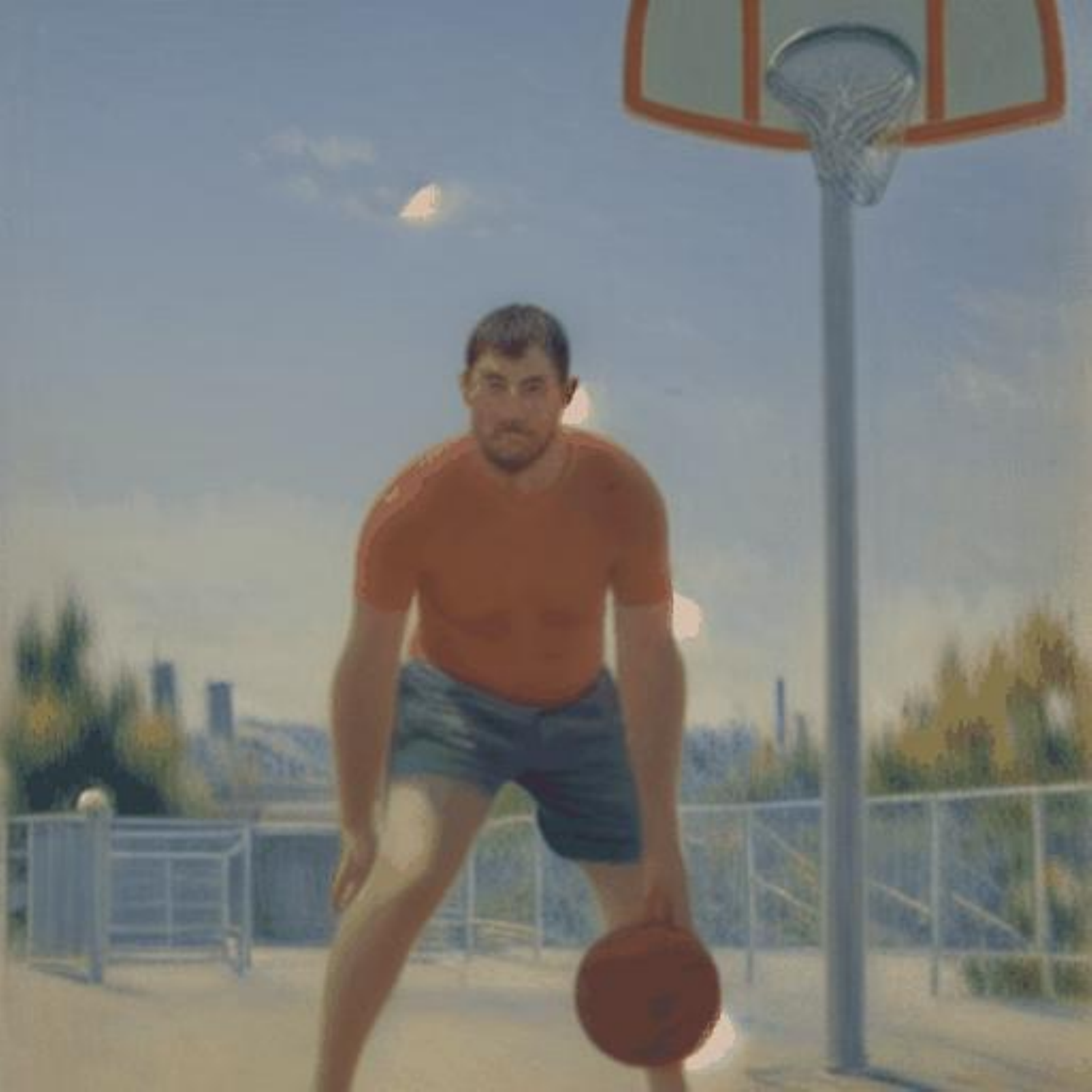}
\includegraphics[width=0.10\textwidth]{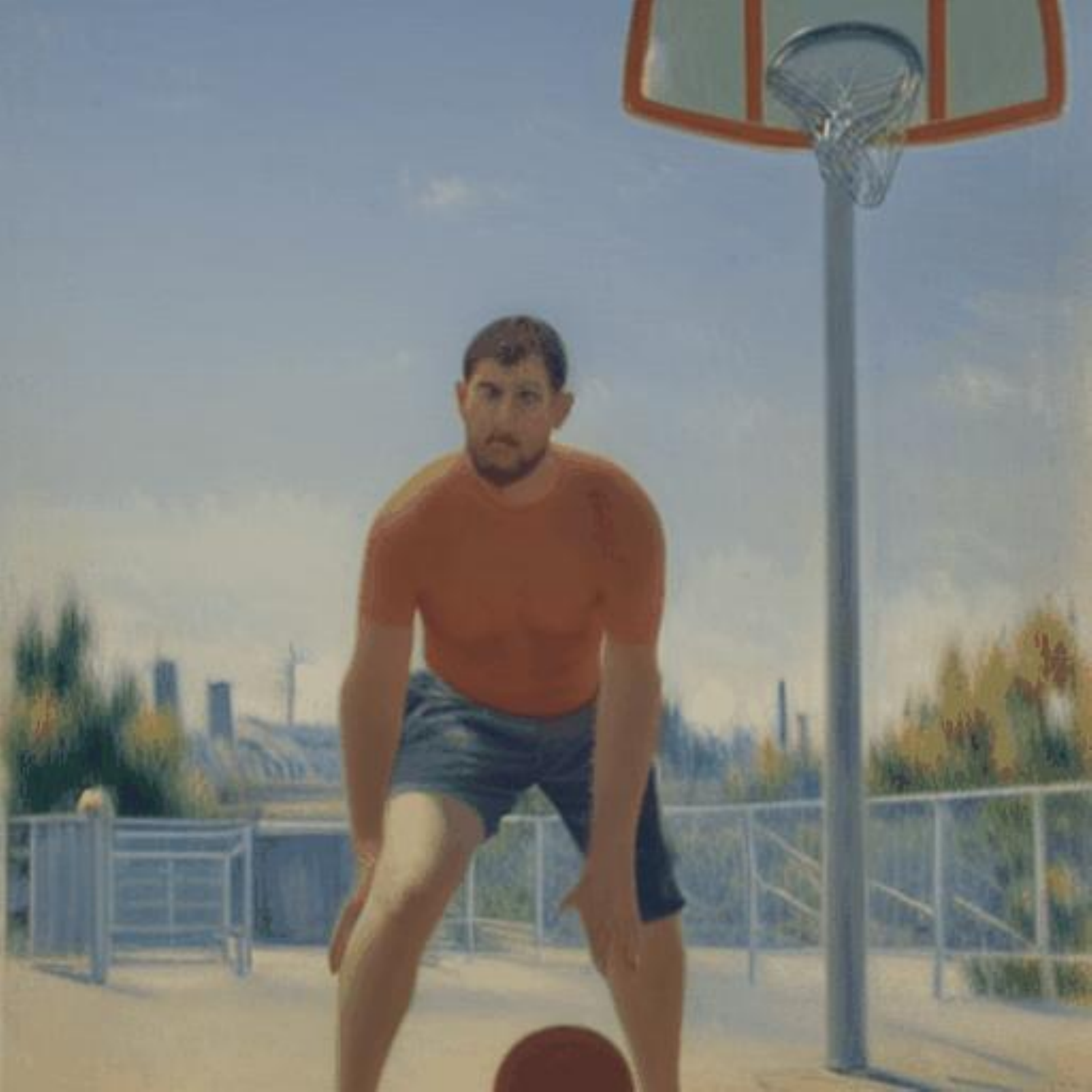}

\makebox[0.12\textwidth]{\colorbox{pink}{\textbf{Training video}} A man is running}\\
\includegraphics[width=0.10\textwidth]{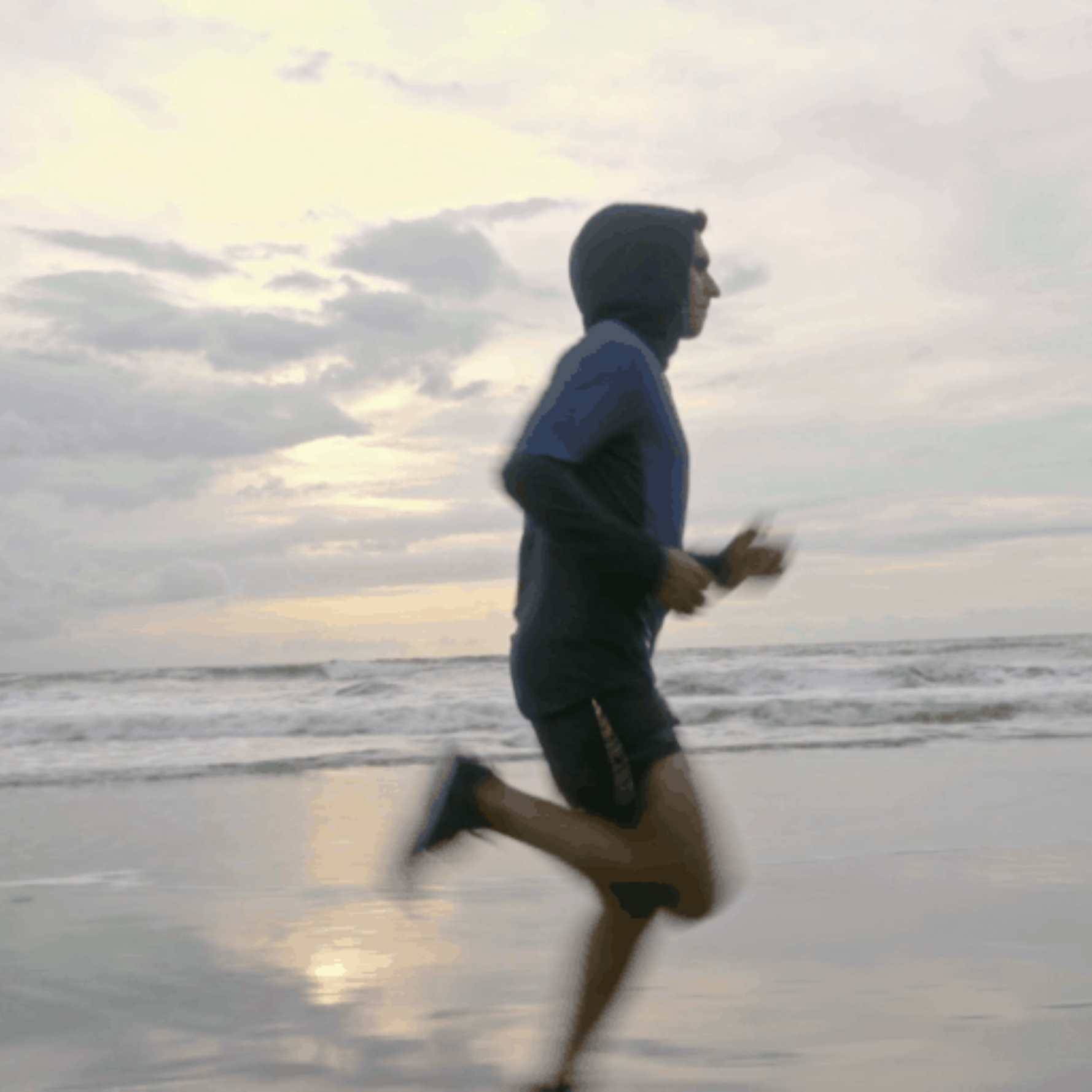}
\includegraphics[width=0.10\textwidth]{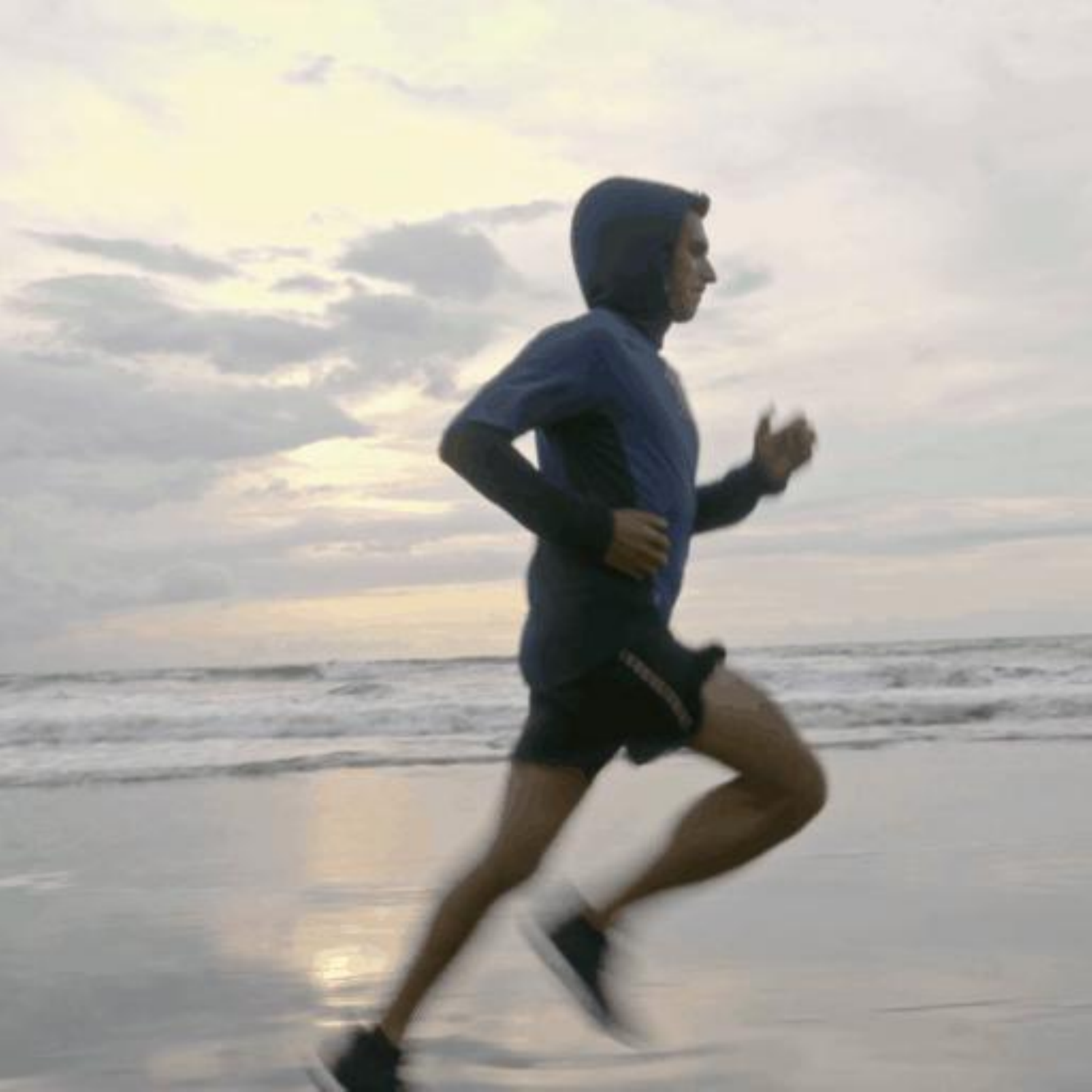}
\includegraphics[width=0.10\textwidth]{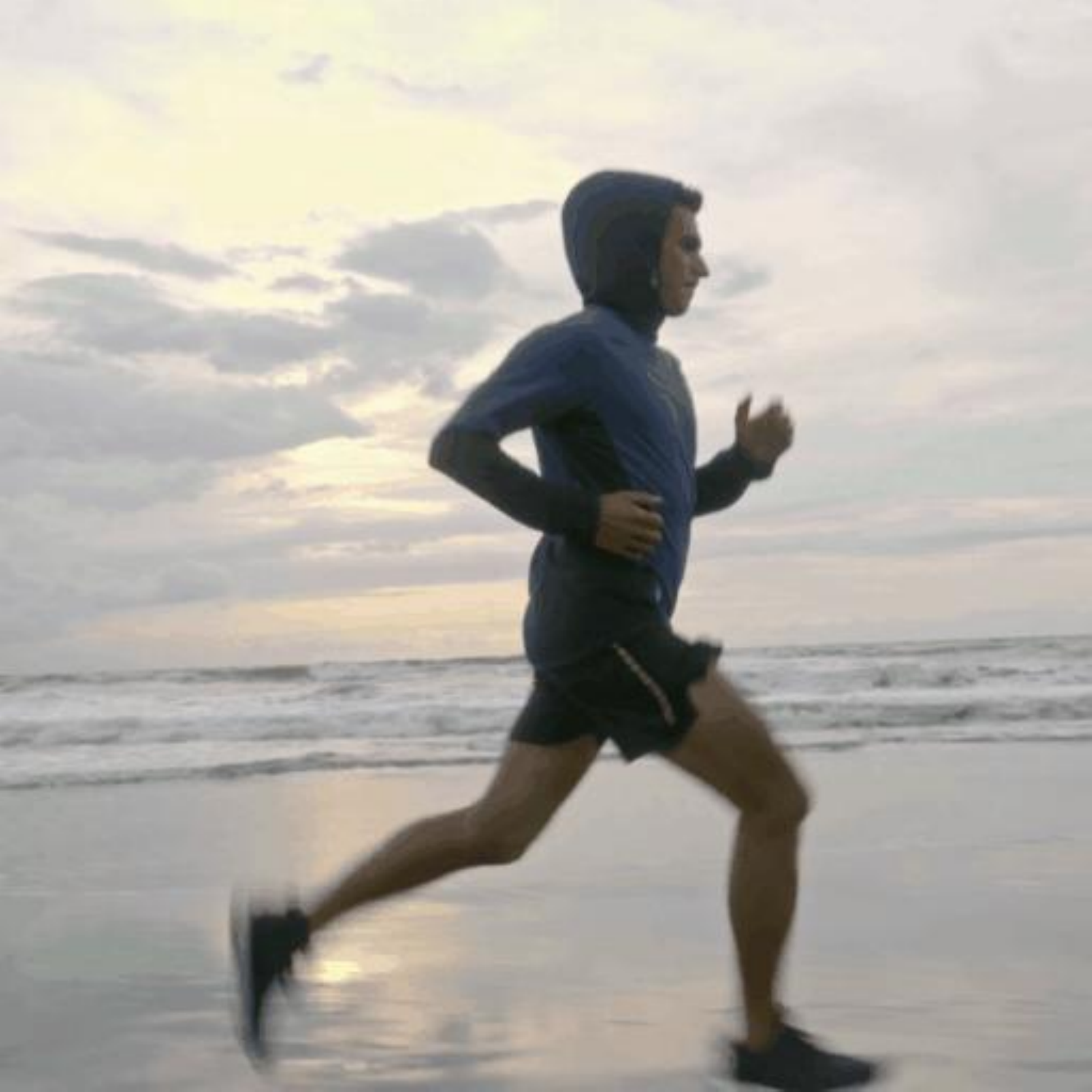}
\includegraphics[width=0.10\textwidth]{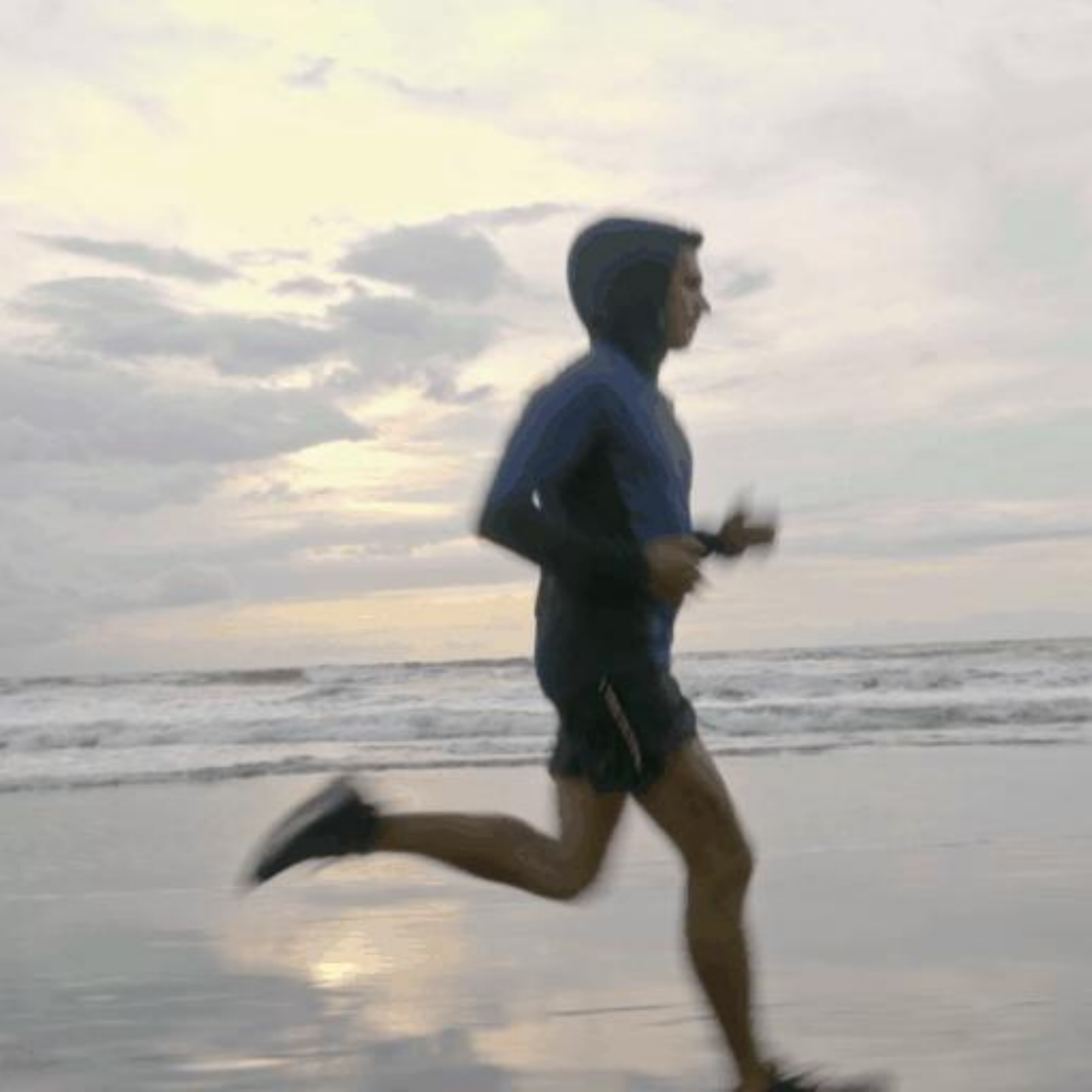}
\includegraphics[width=0.10\textwidth]{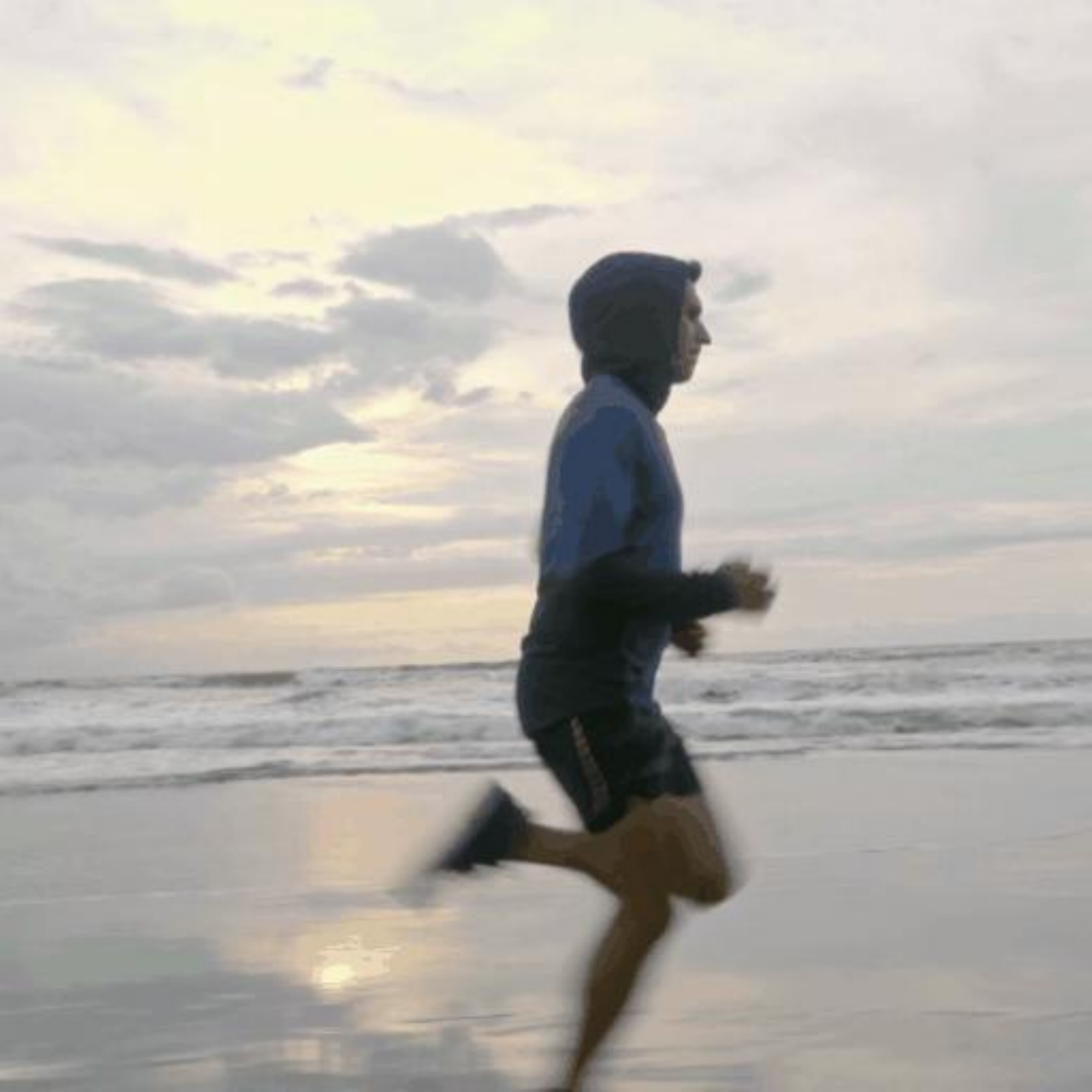}
\includegraphics[width=0.10\textwidth]{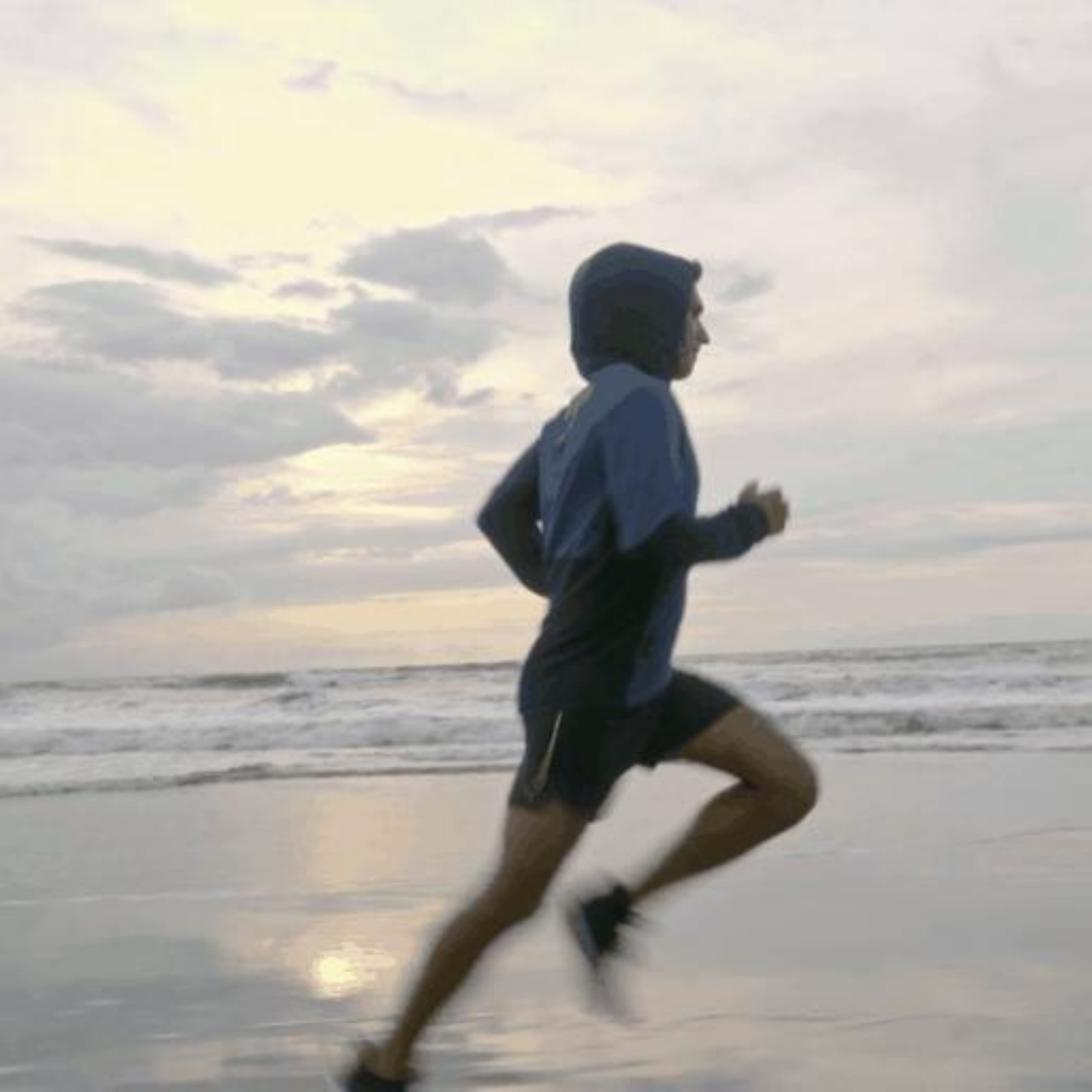}
\includegraphics[width=0.10\textwidth]{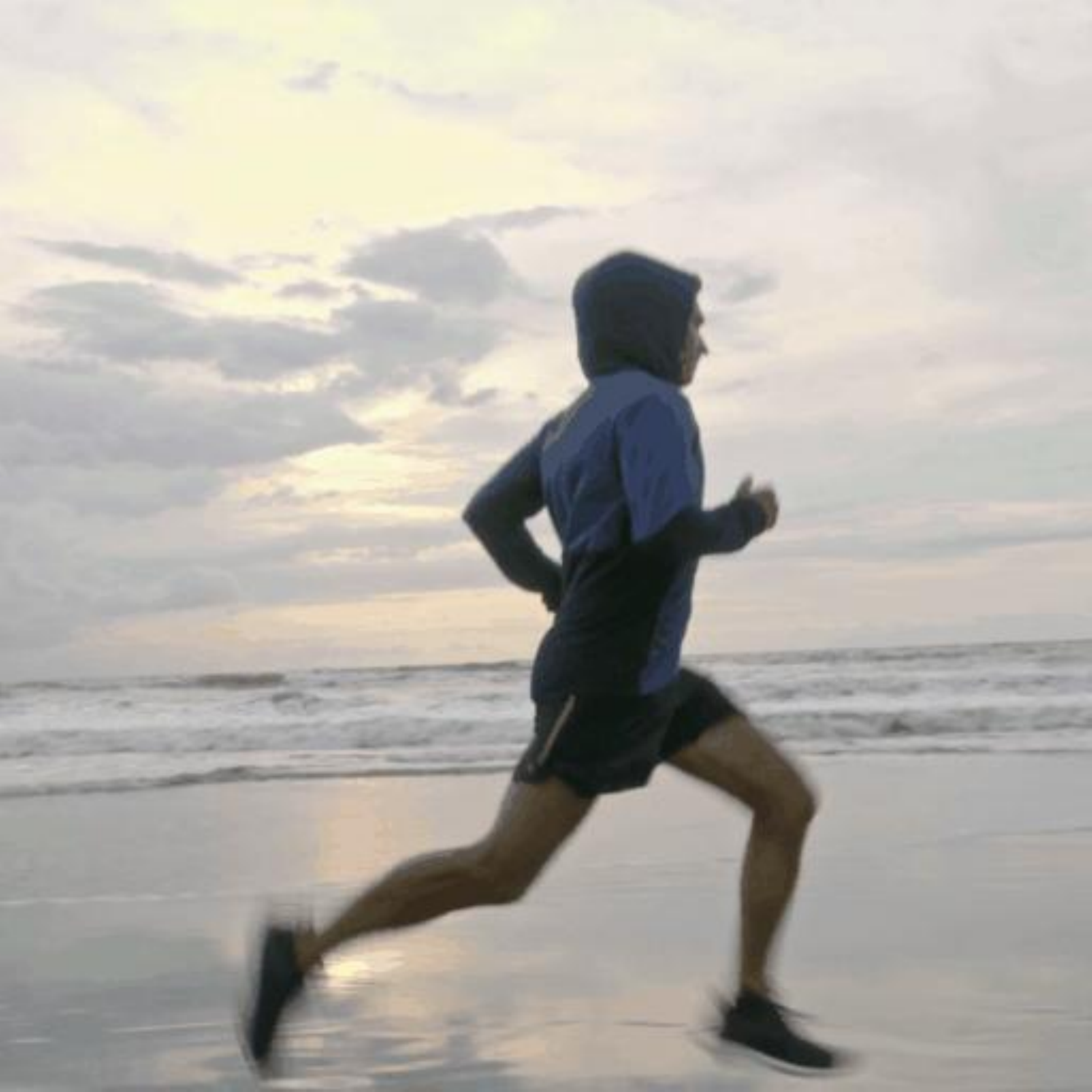}
\includegraphics[width=0.10\textwidth]{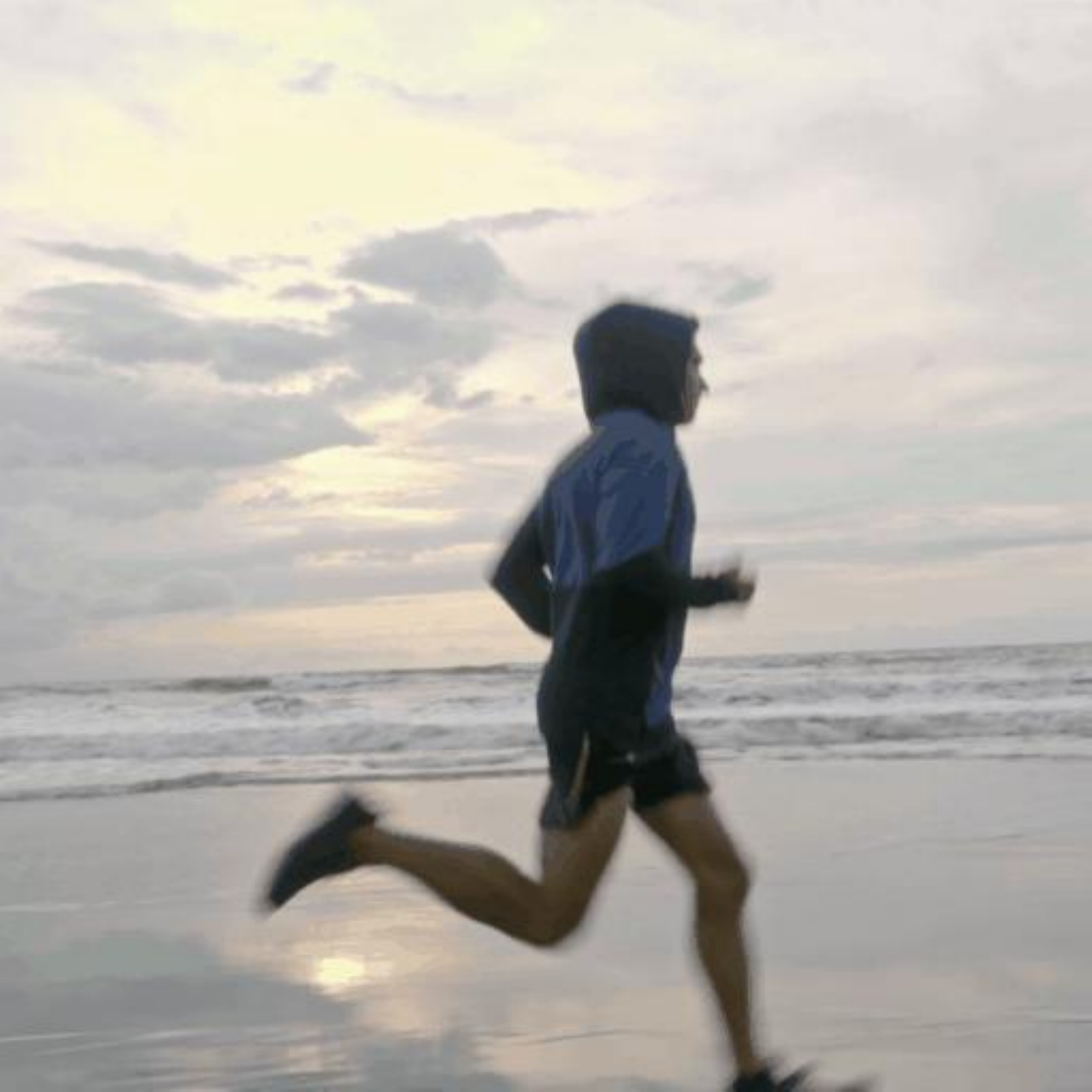}

\makebox[0.12\textwidth]{A \textcolor{blue}{\textbf{zombie}} is running.}\\
\includegraphics[width=0.10\textwidth]{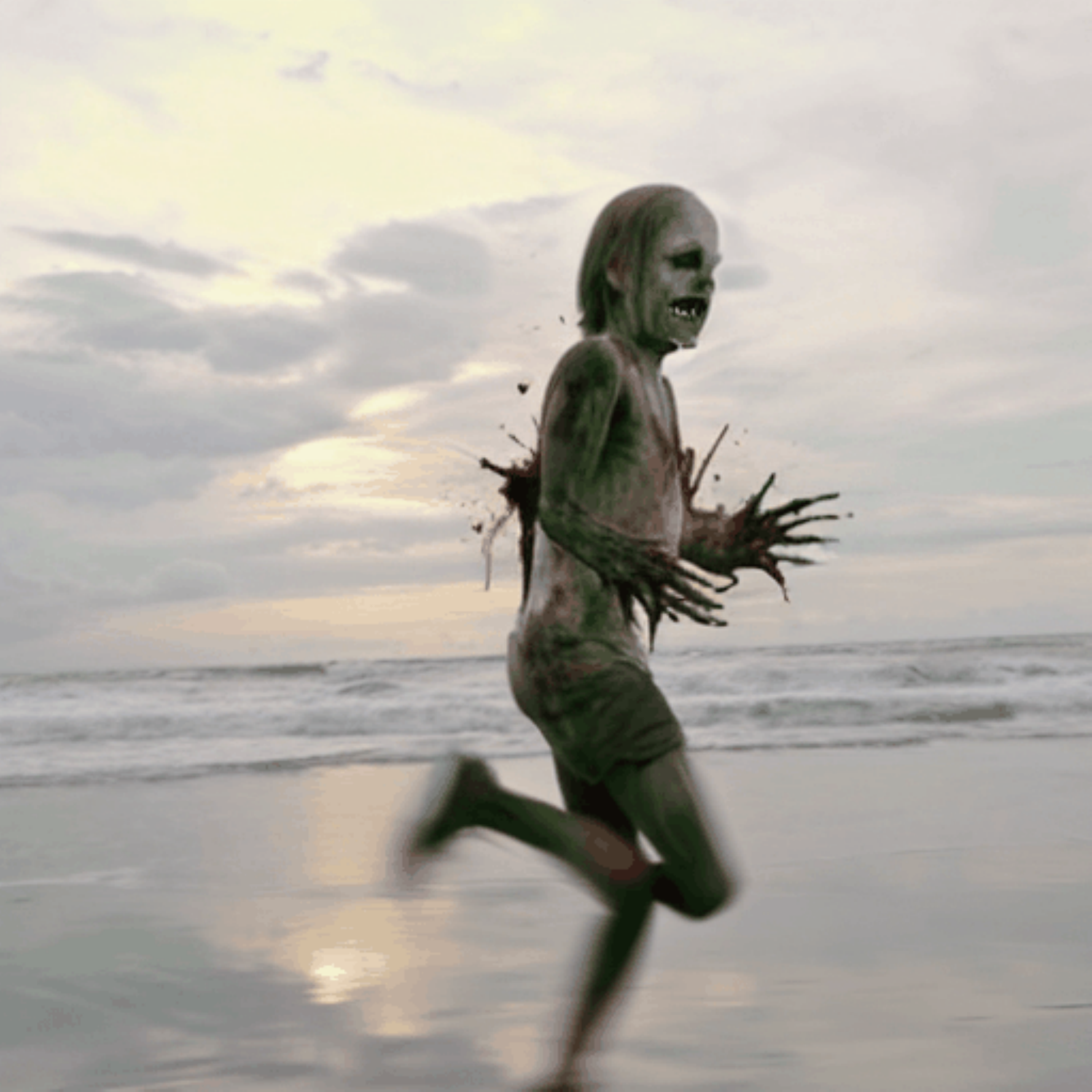}
\includegraphics[width=0.10\textwidth]{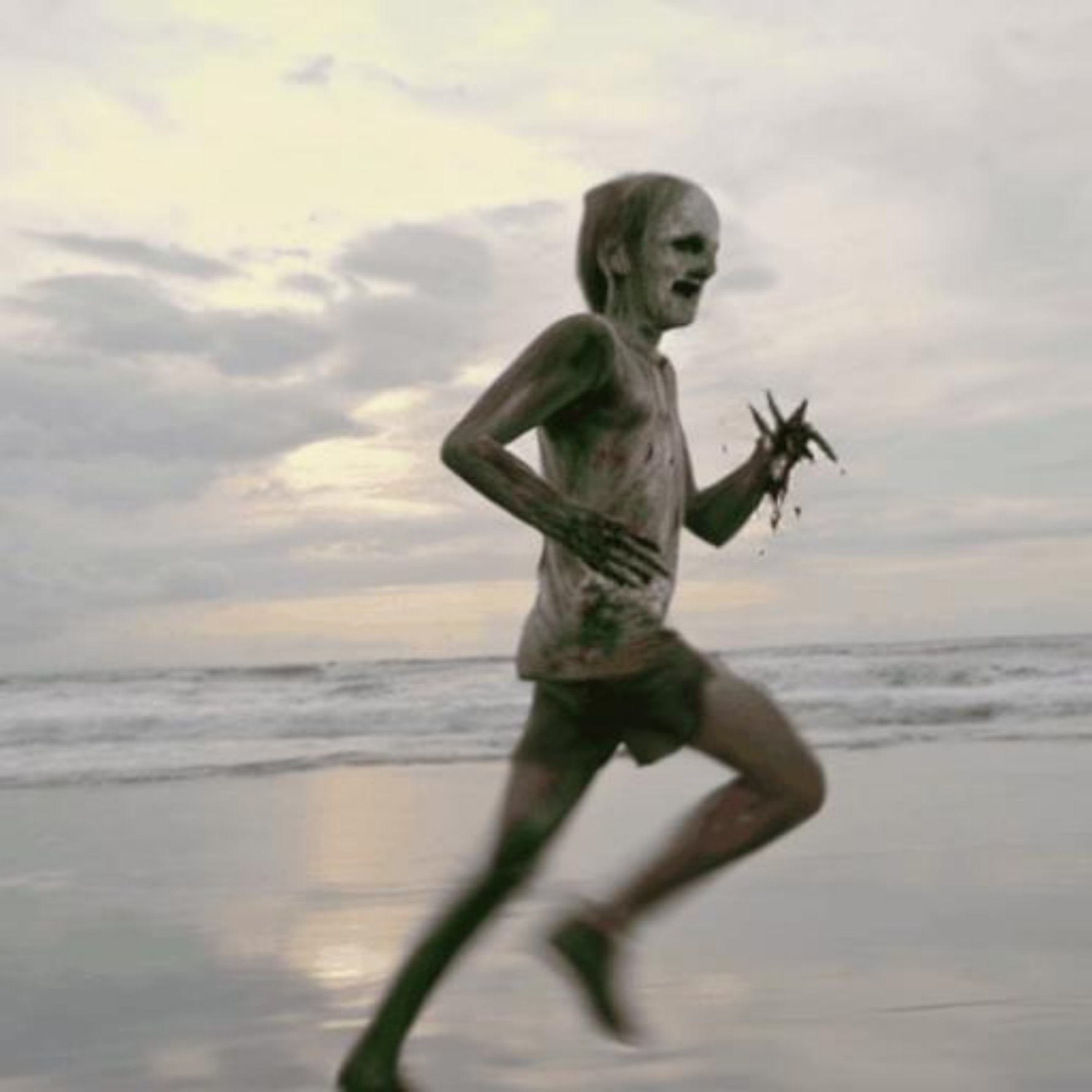}
\includegraphics[width=0.10\textwidth]{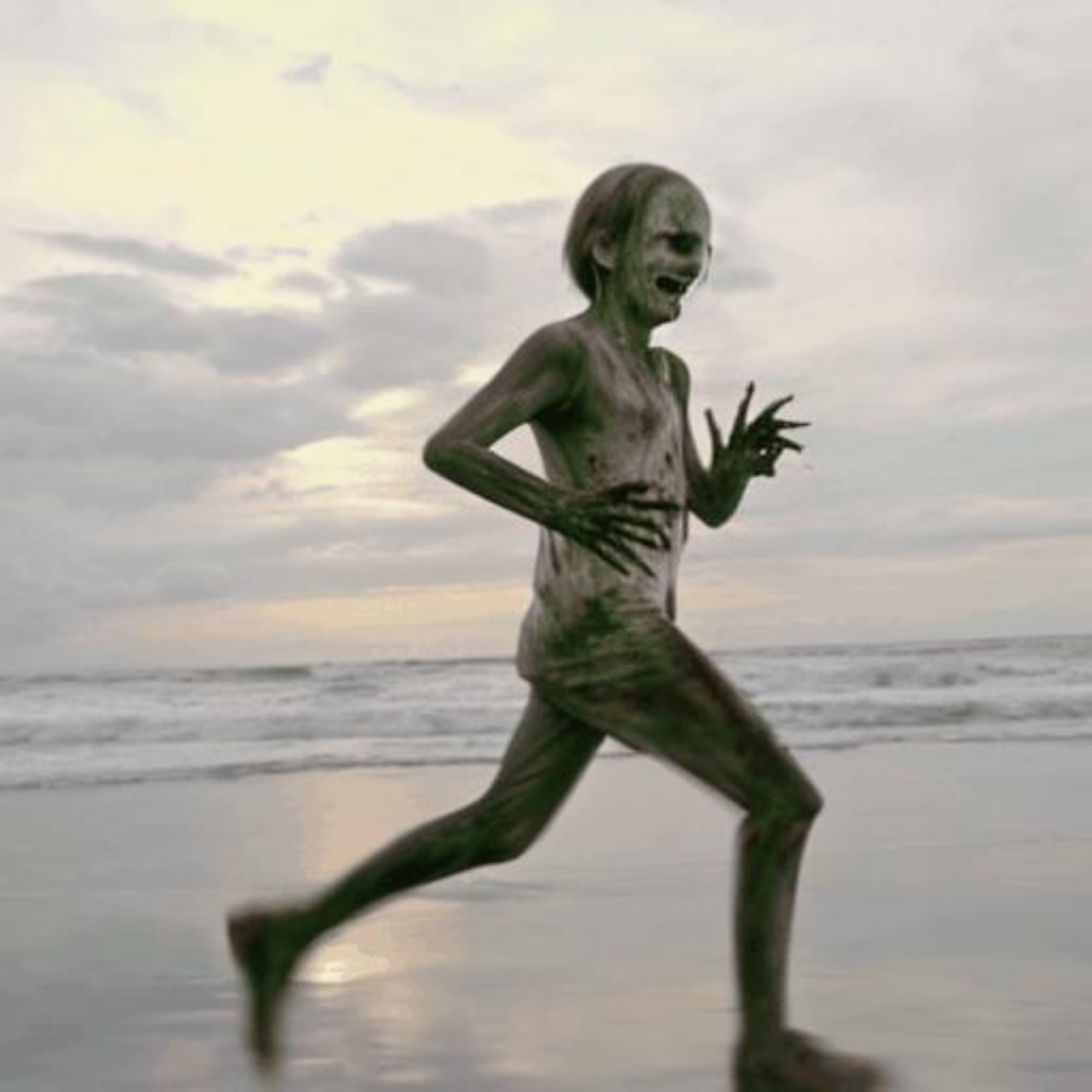}
\includegraphics[width=0.10\textwidth]{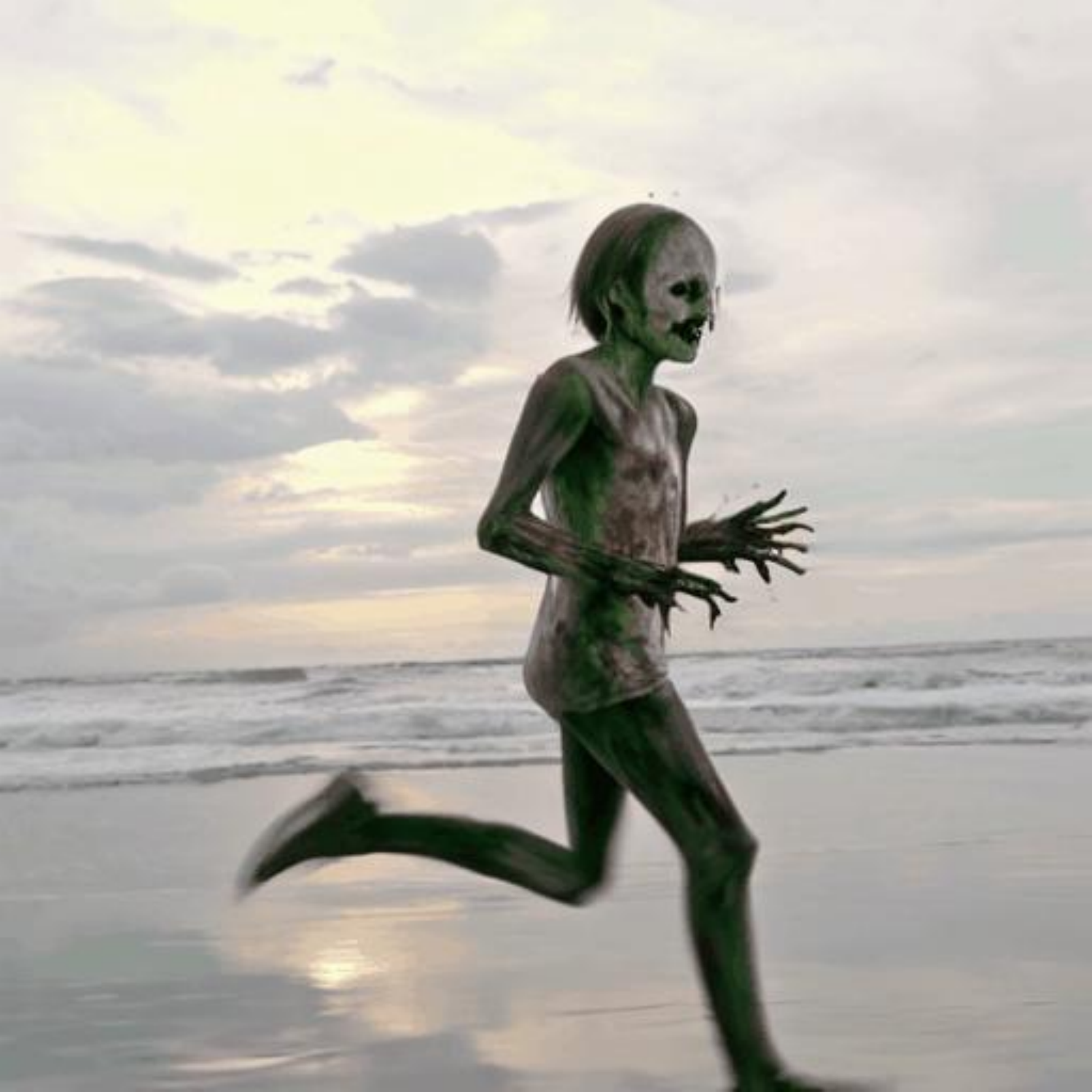}
\includegraphics[width=0.10\textwidth]{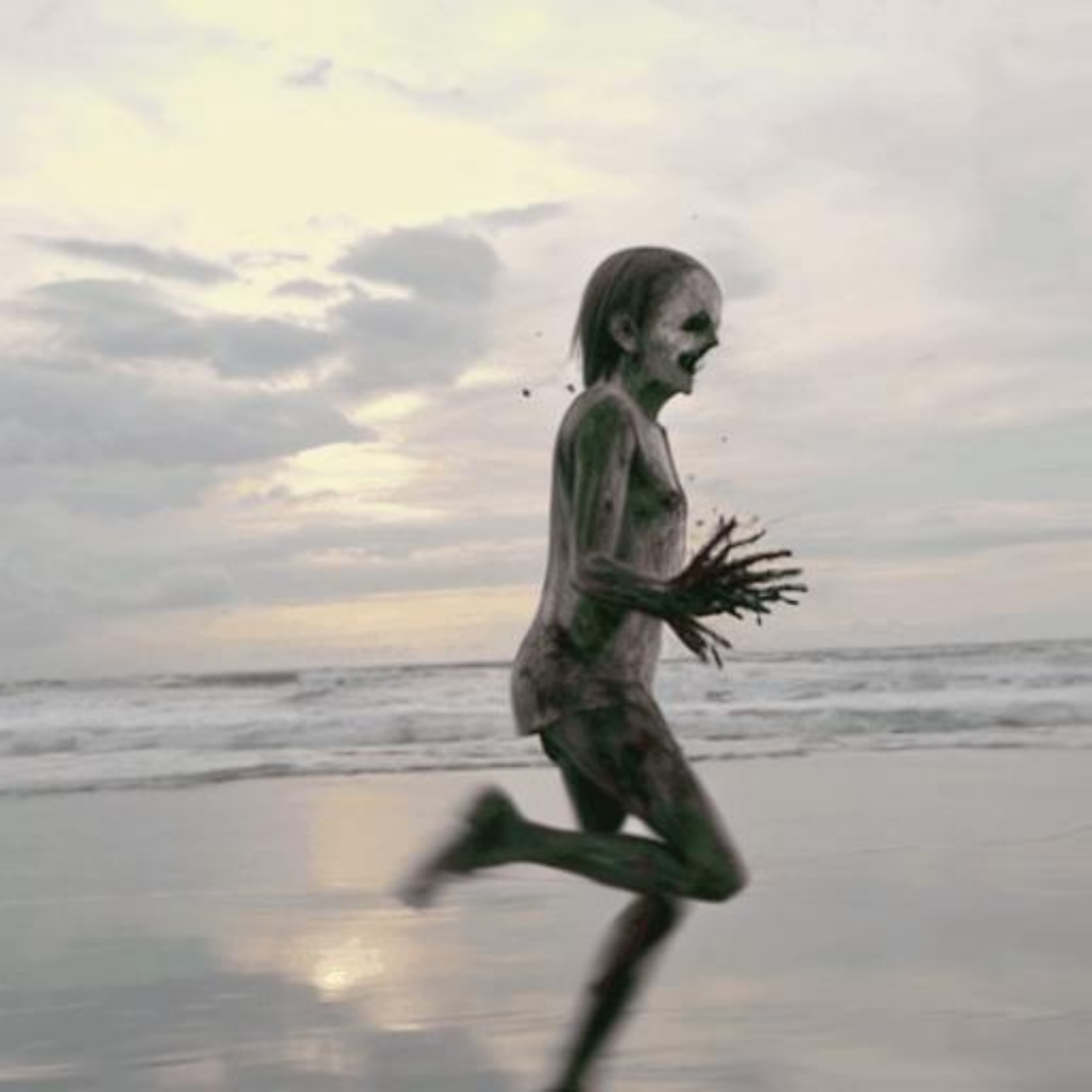}
\includegraphics[width=0.10\textwidth]{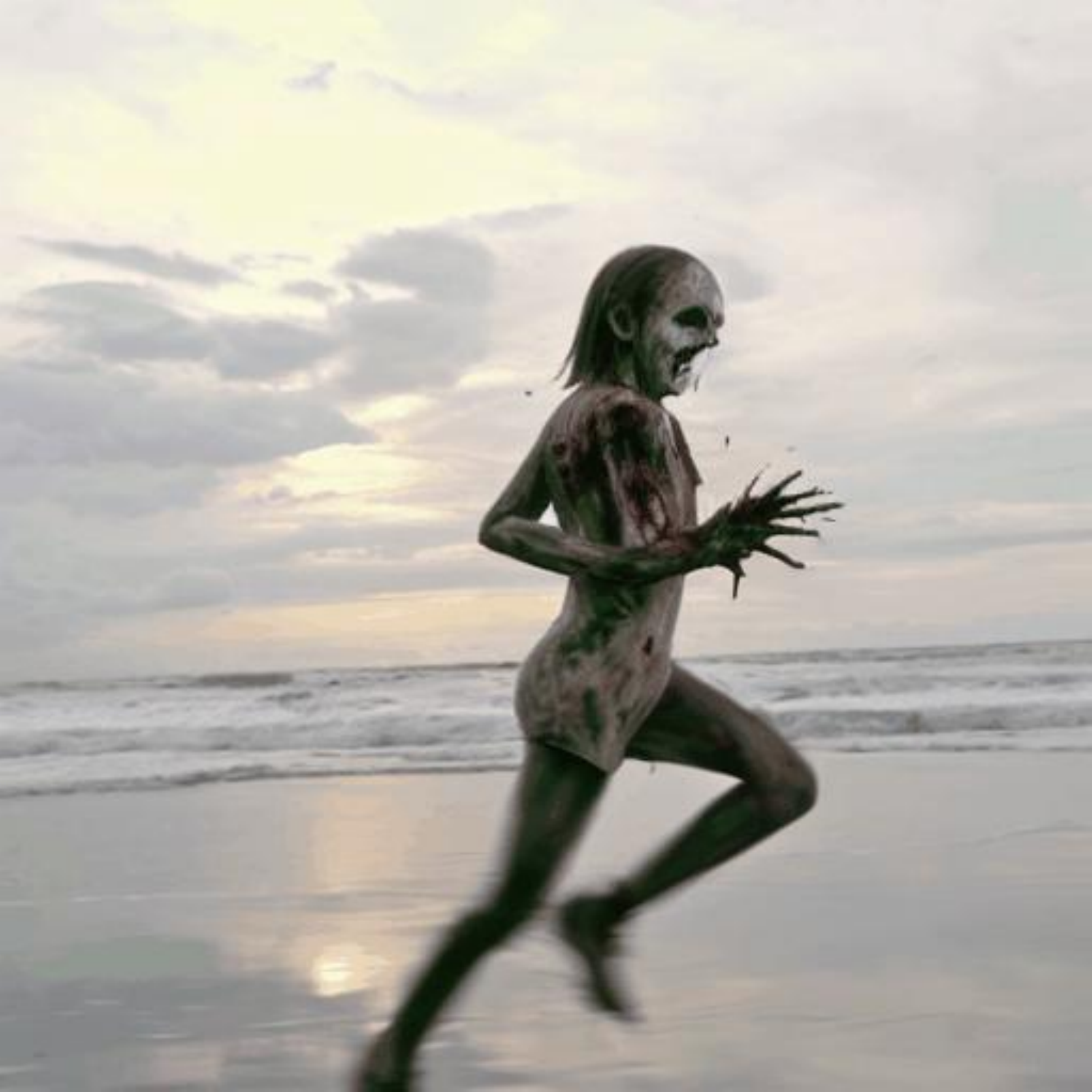}
\includegraphics[width=0.10\textwidth]{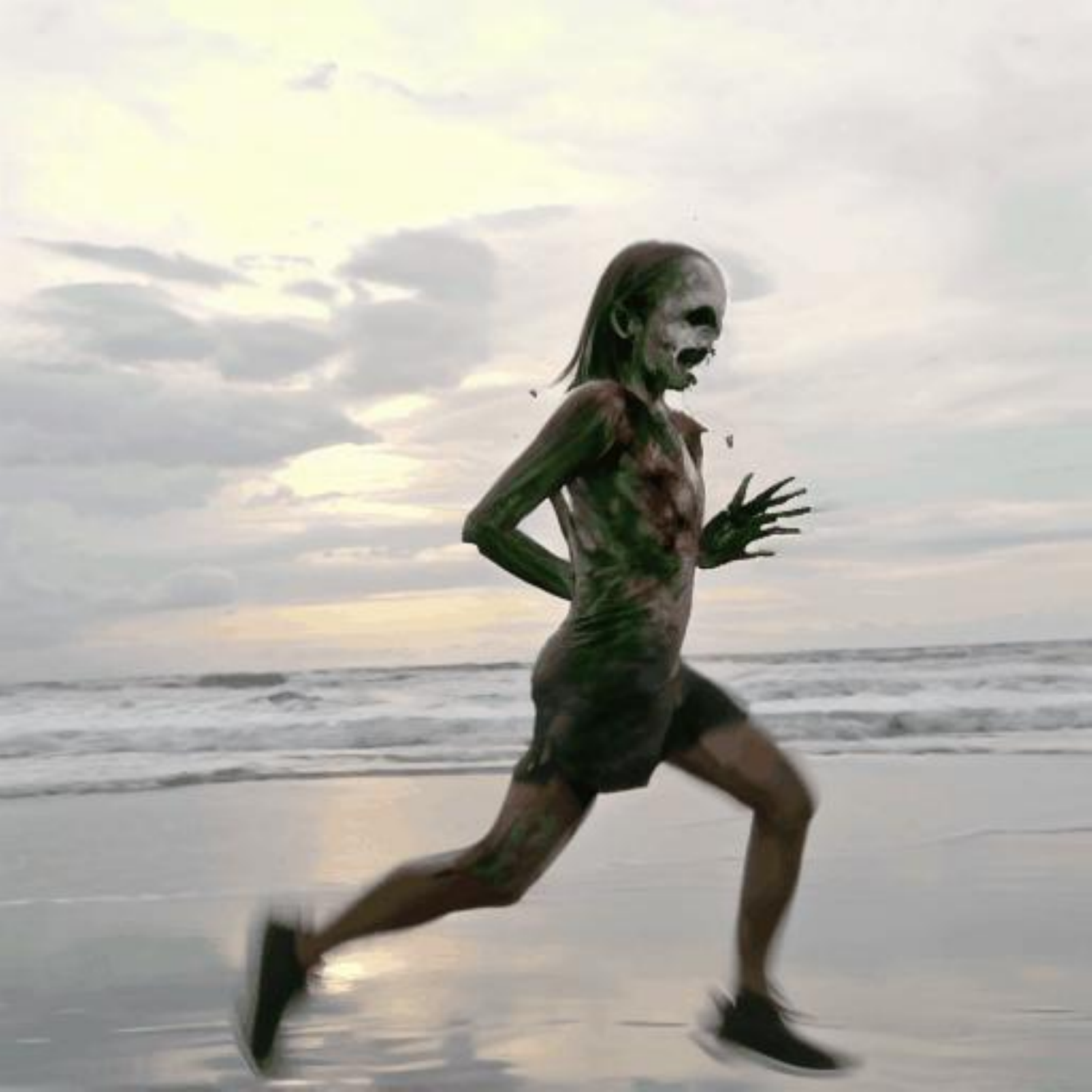}
\includegraphics[width=0.10\textwidth]{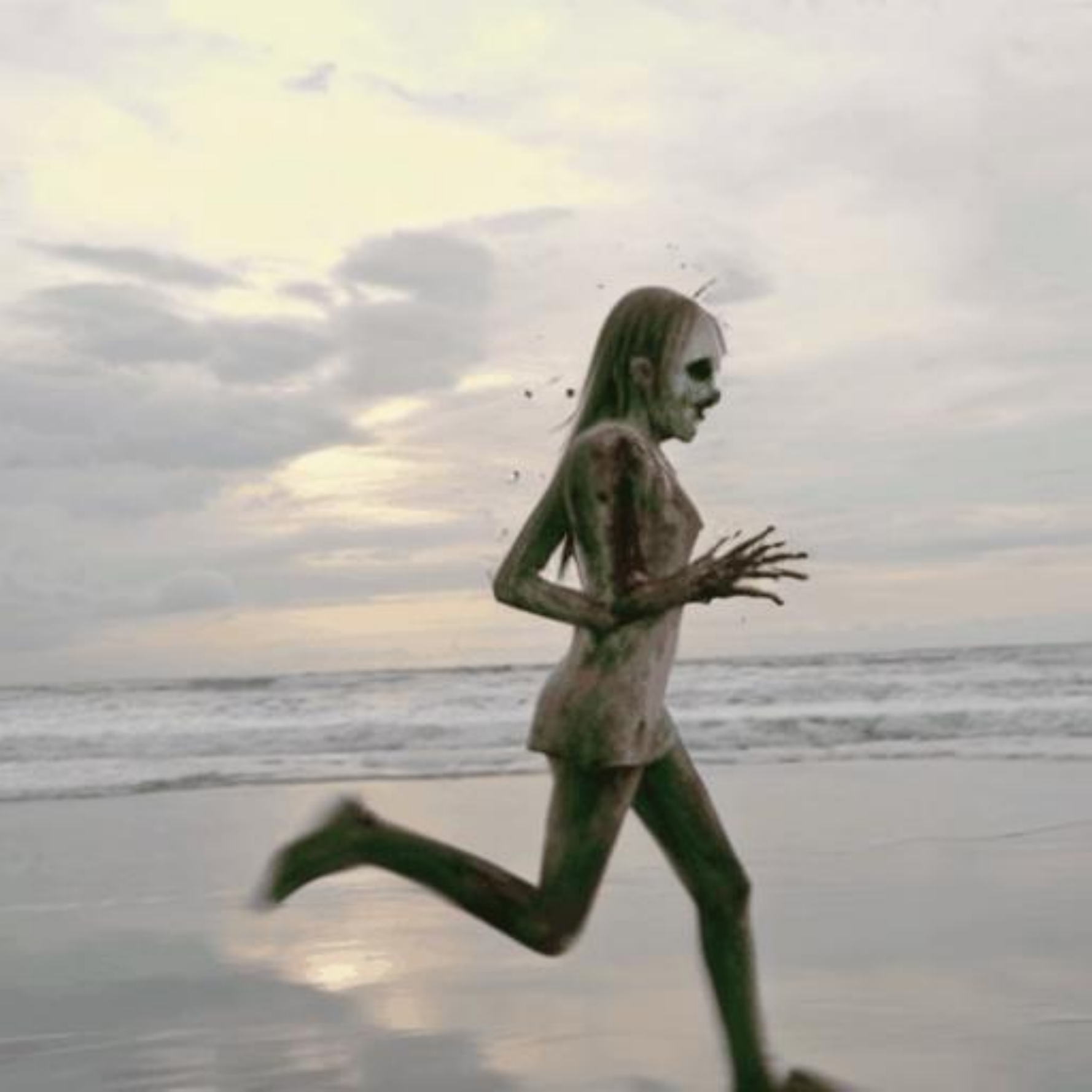}

\makebox[0.12\textwidth]{A \textcolor{blue}{\textbf{werewolf}} is running.}\\
\includegraphics[width=0.10\textwidth]{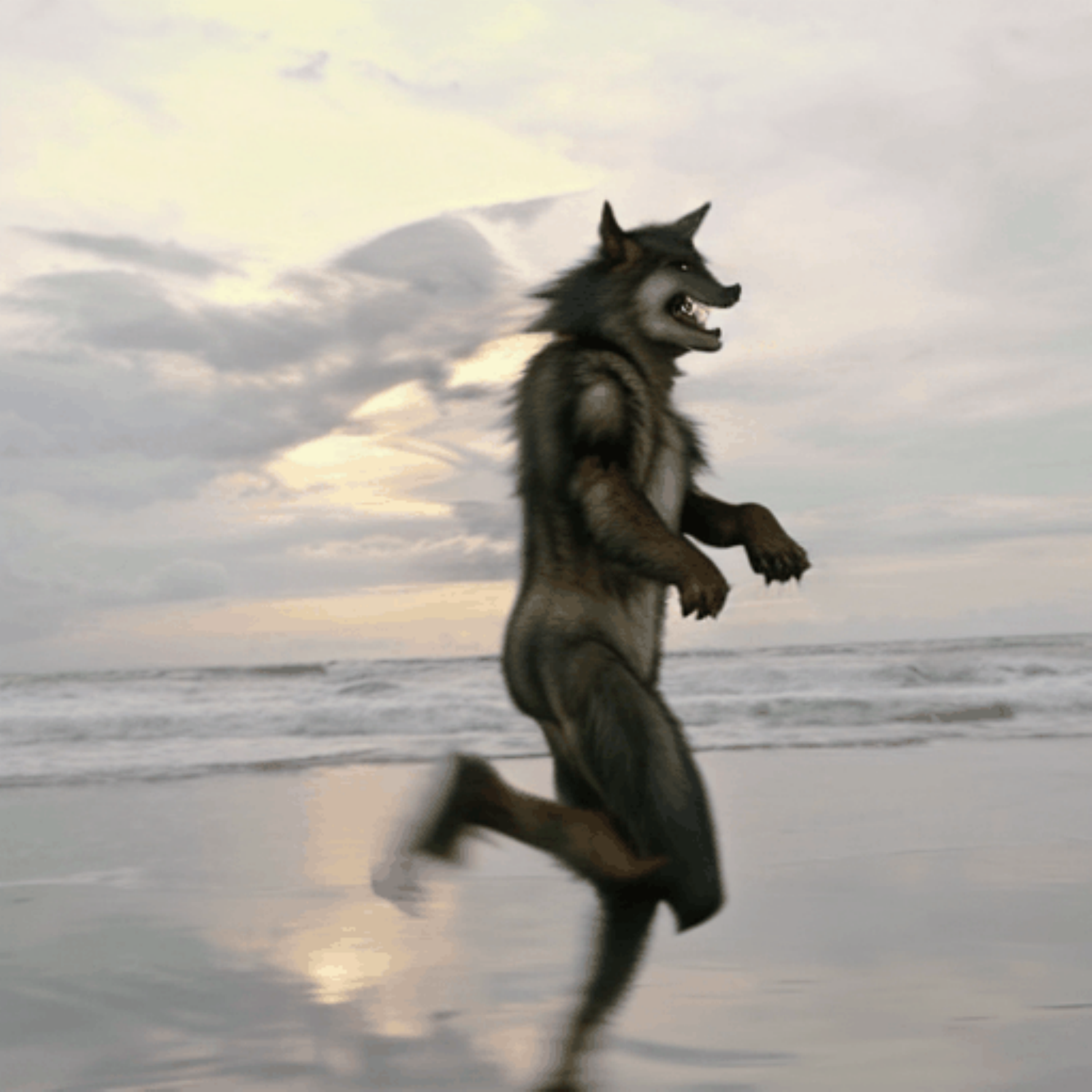}
\includegraphics[width=0.10\textwidth]{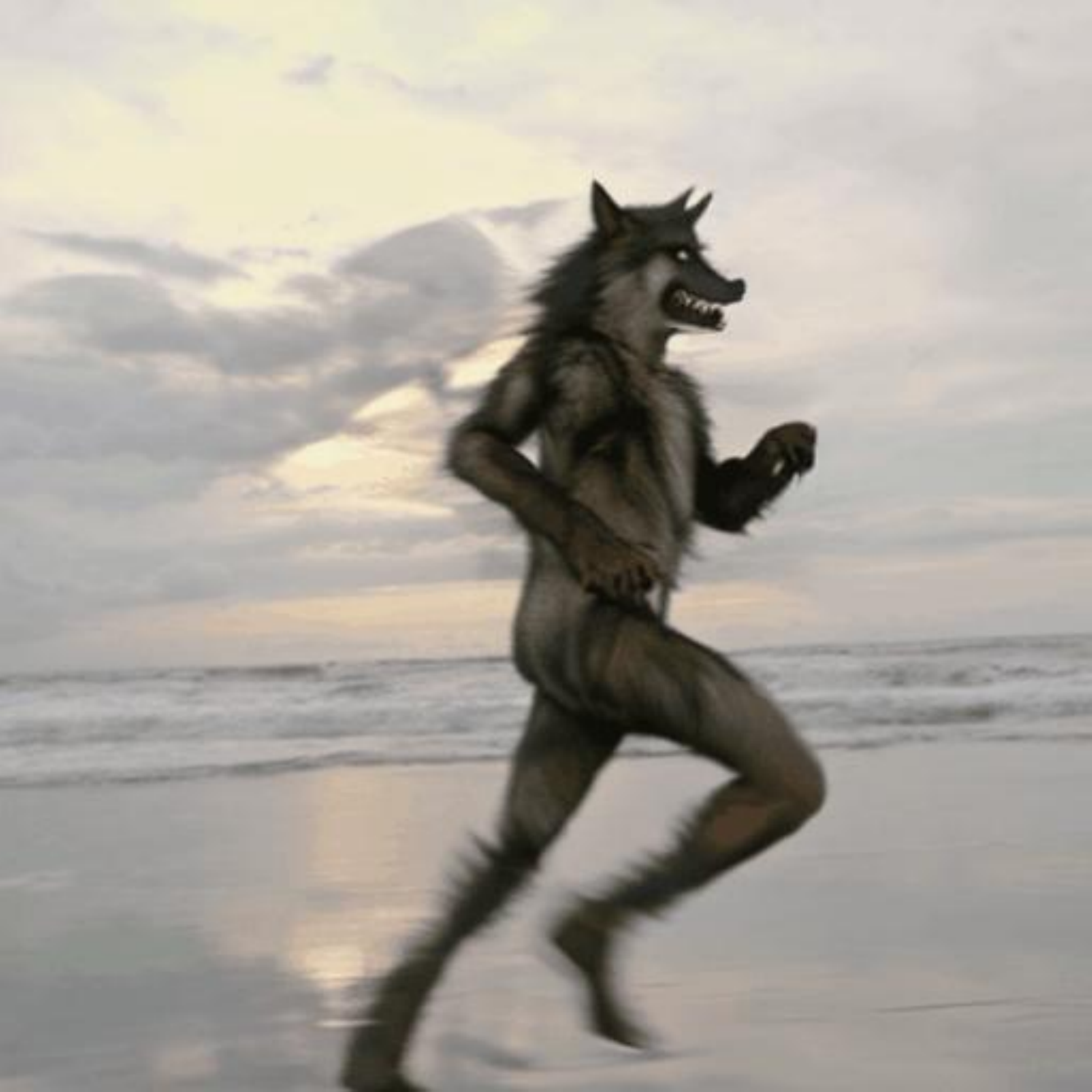}
\includegraphics[width=0.10\textwidth]{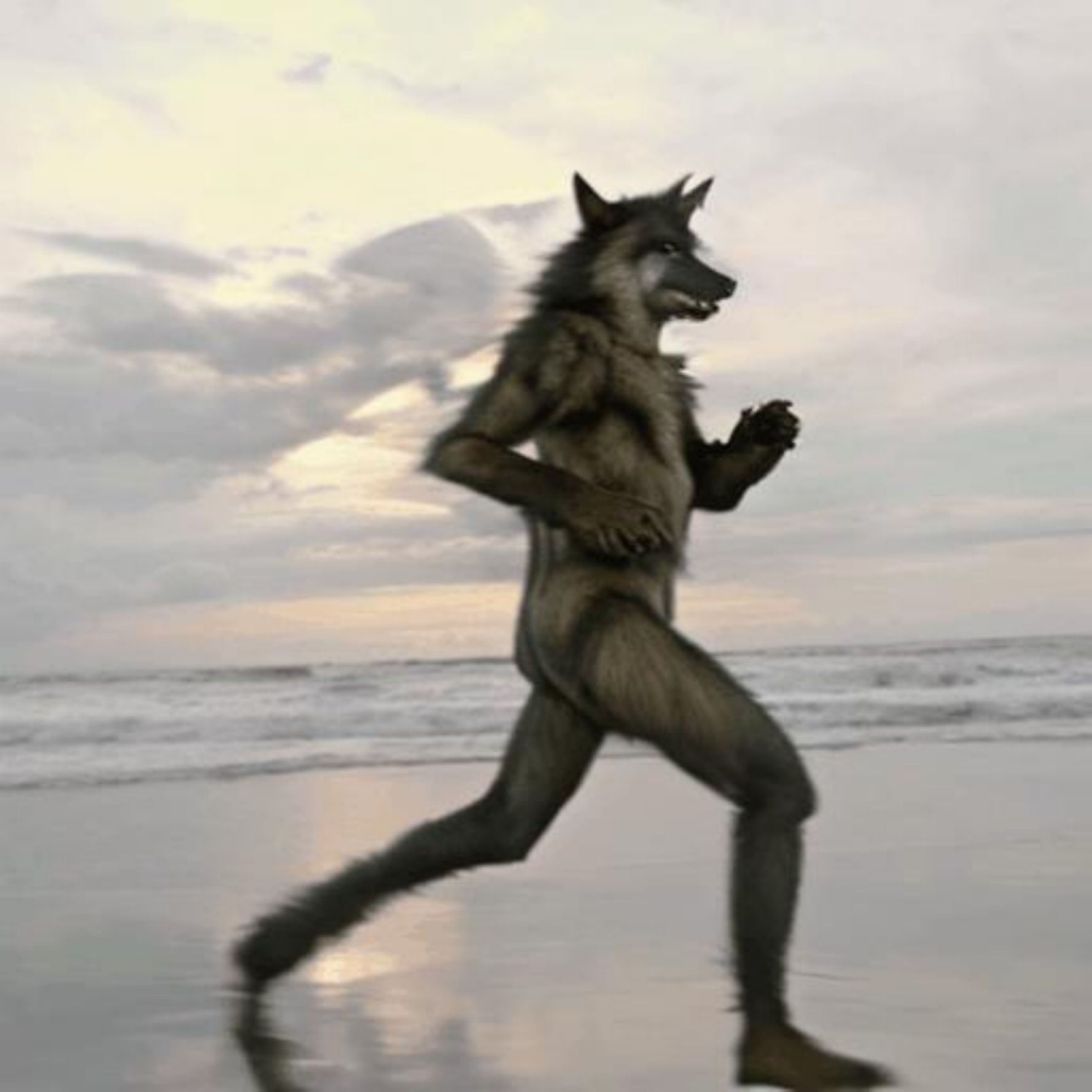}
\includegraphics[width=0.10\textwidth]{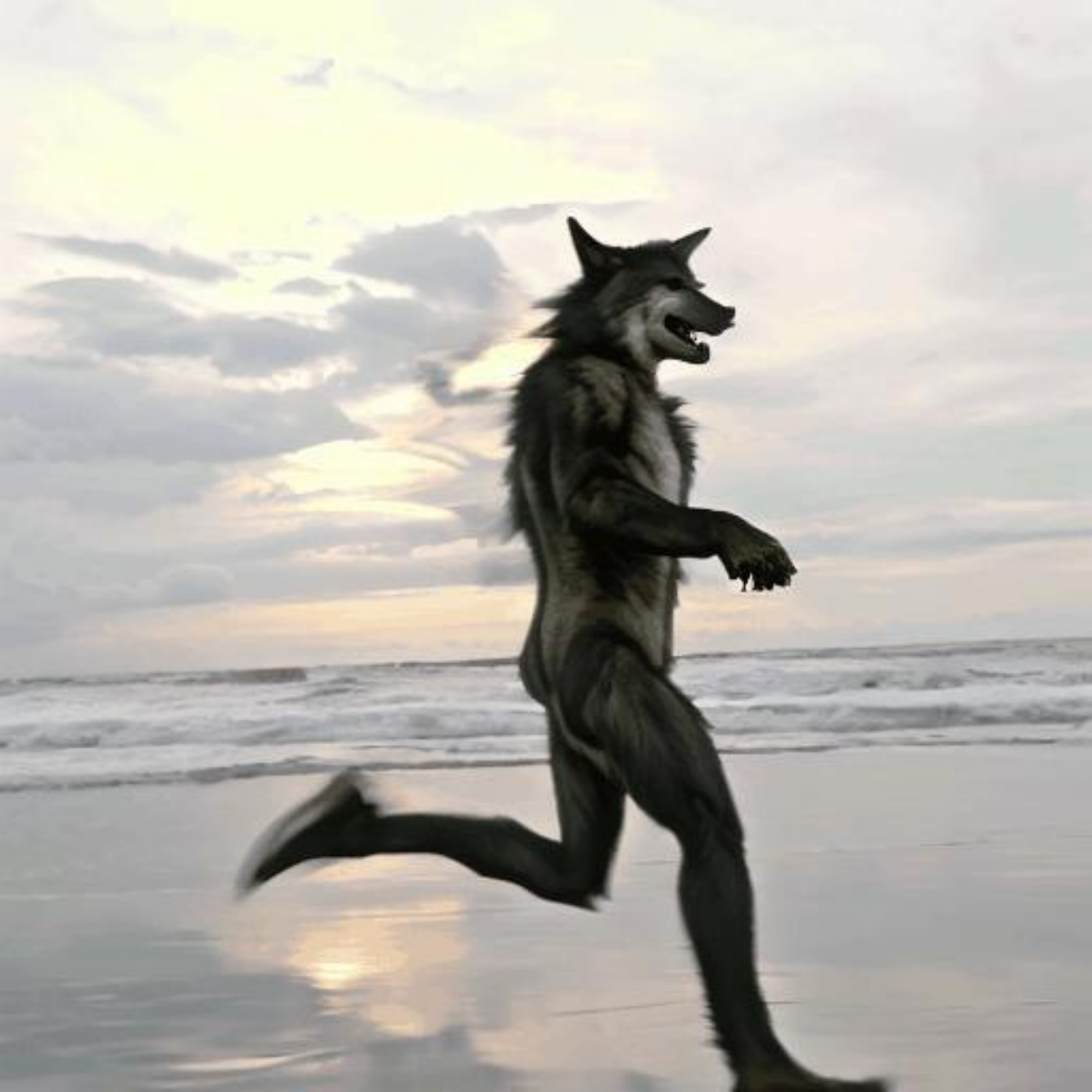}
\includegraphics[width=0.10\textwidth]{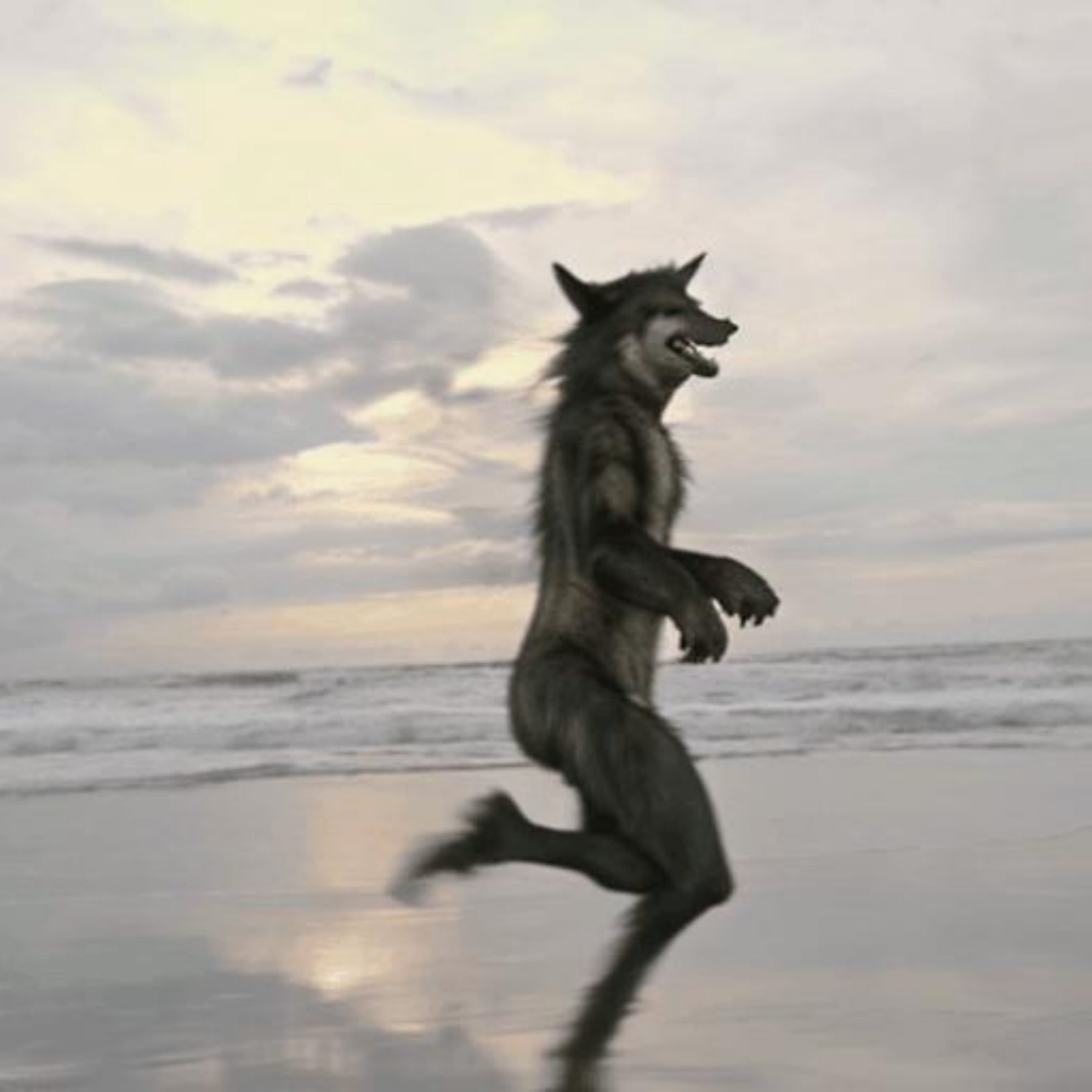}
\includegraphics[width=0.10\textwidth]{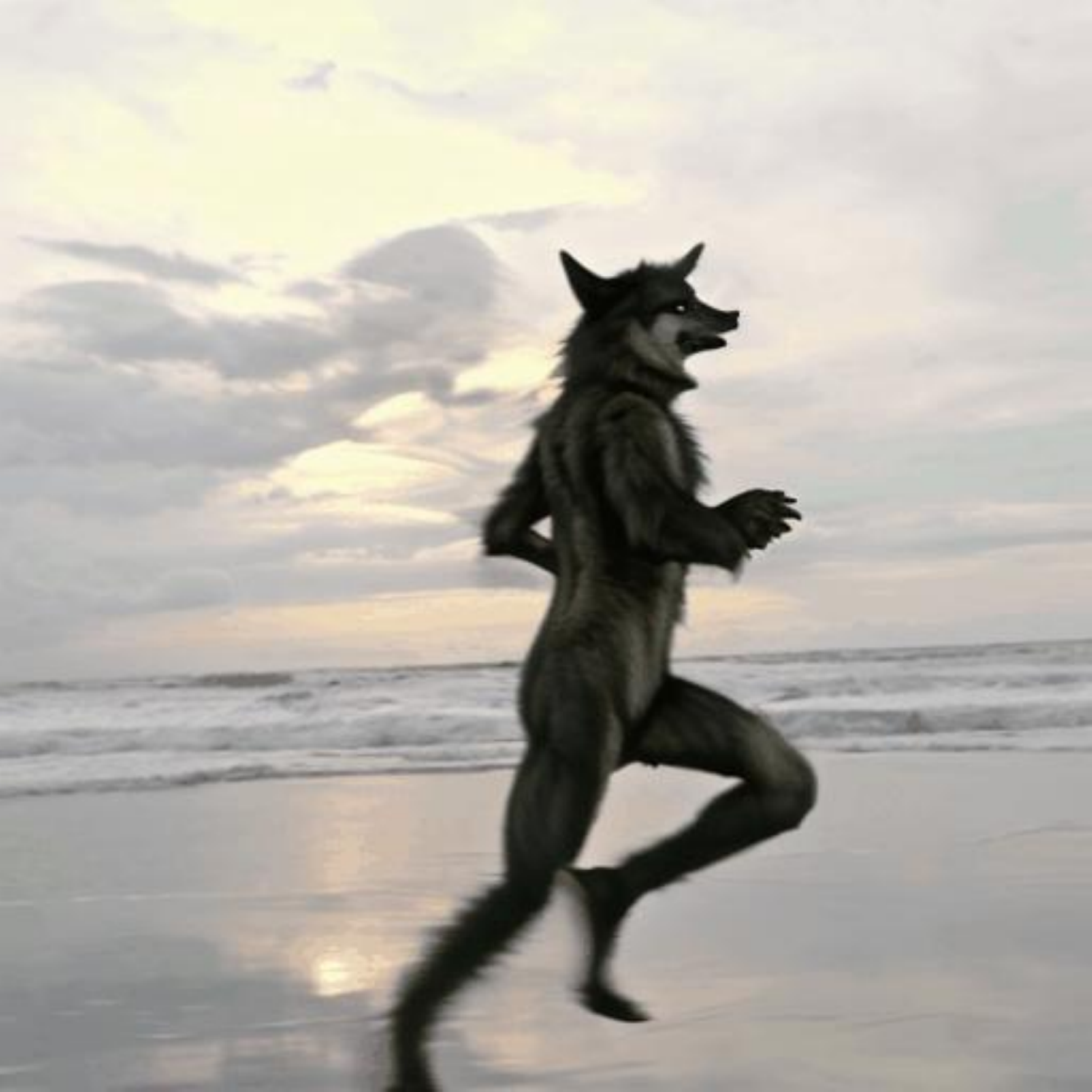}
\includegraphics[width=0.10\textwidth]{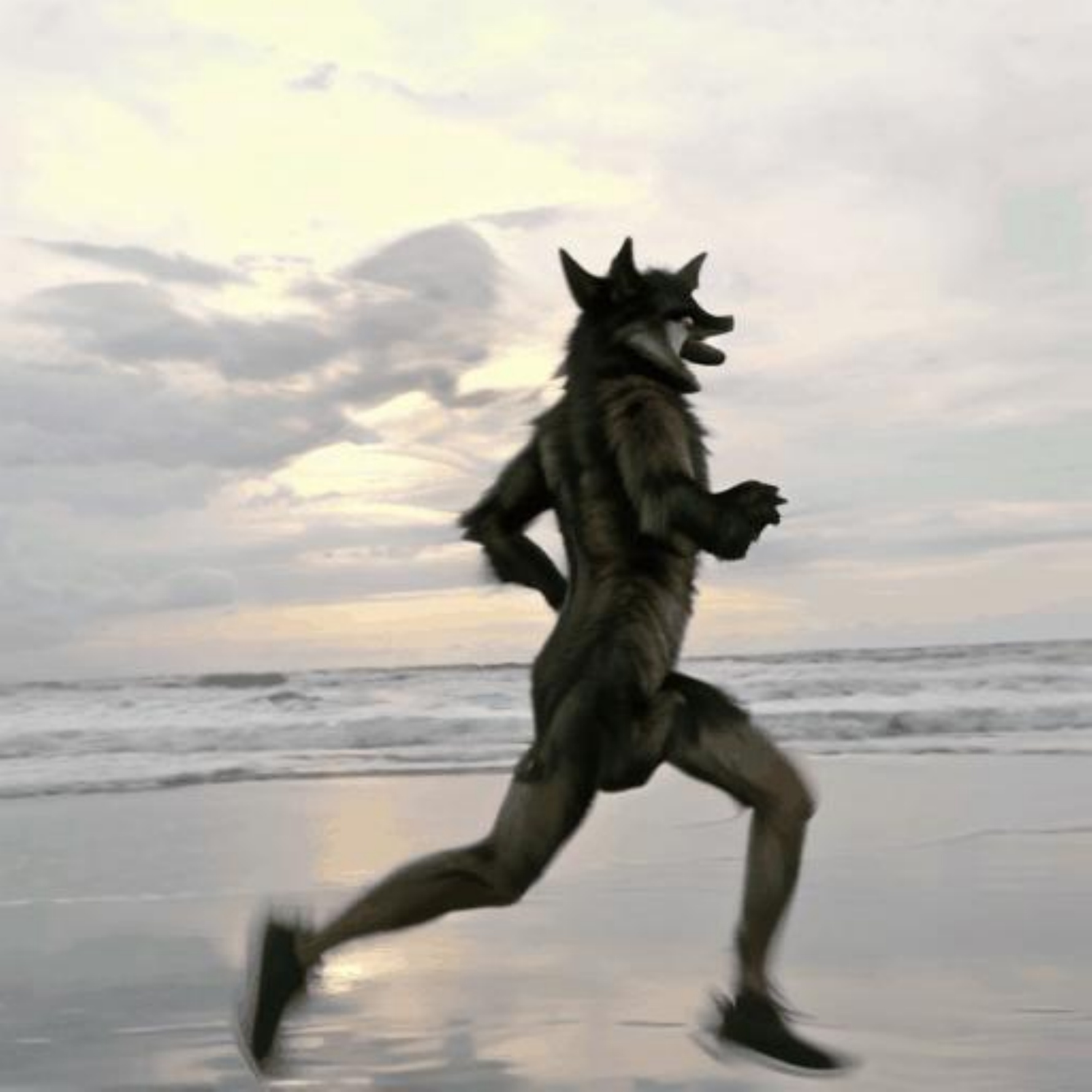}
\includegraphics[width=0.10\textwidth]{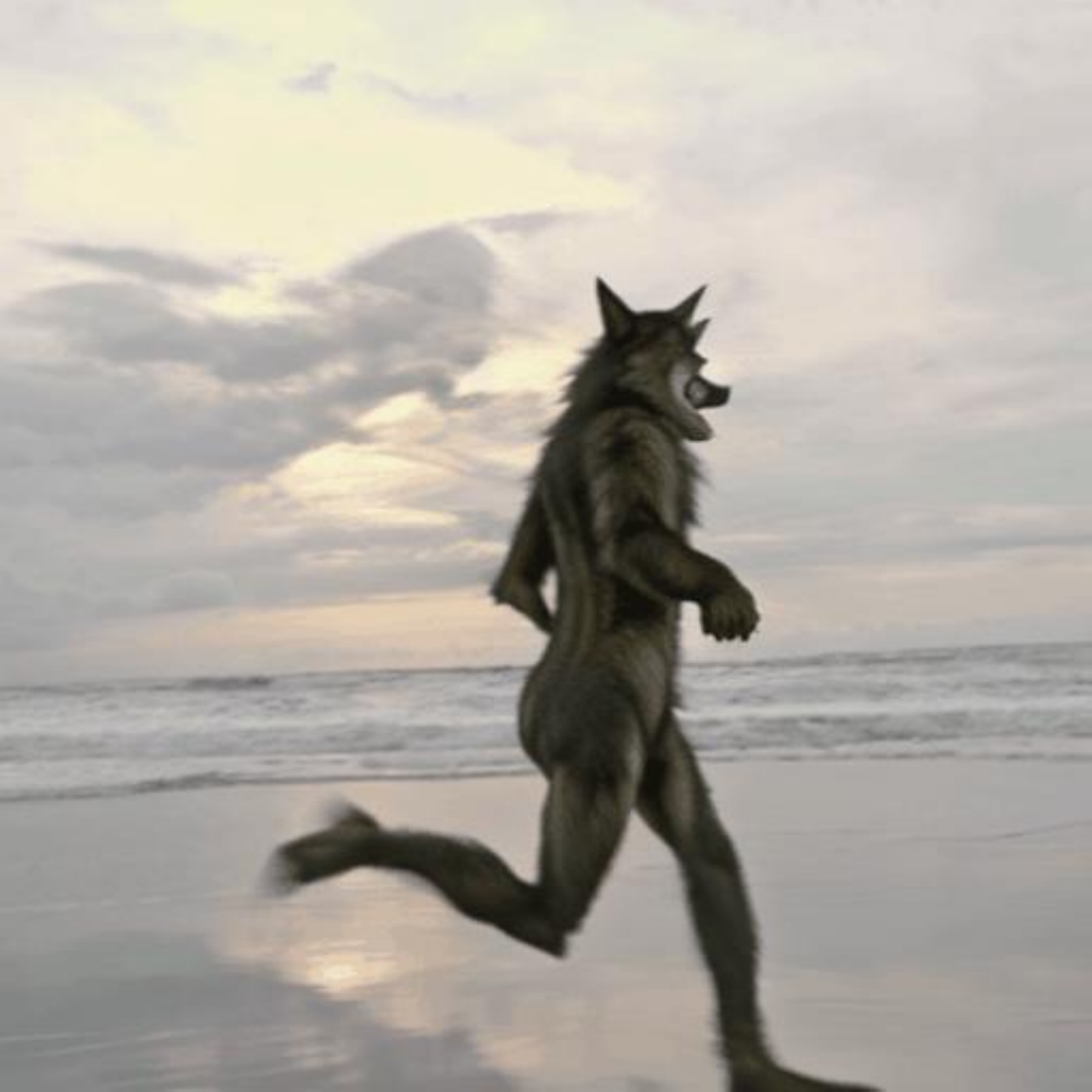}

\makebox[0.12\textwidth]{A \textcolor{blue}{\textbf{pencil sketch}} that a man is running.}\\
\includegraphics[width=0.10\textwidth]{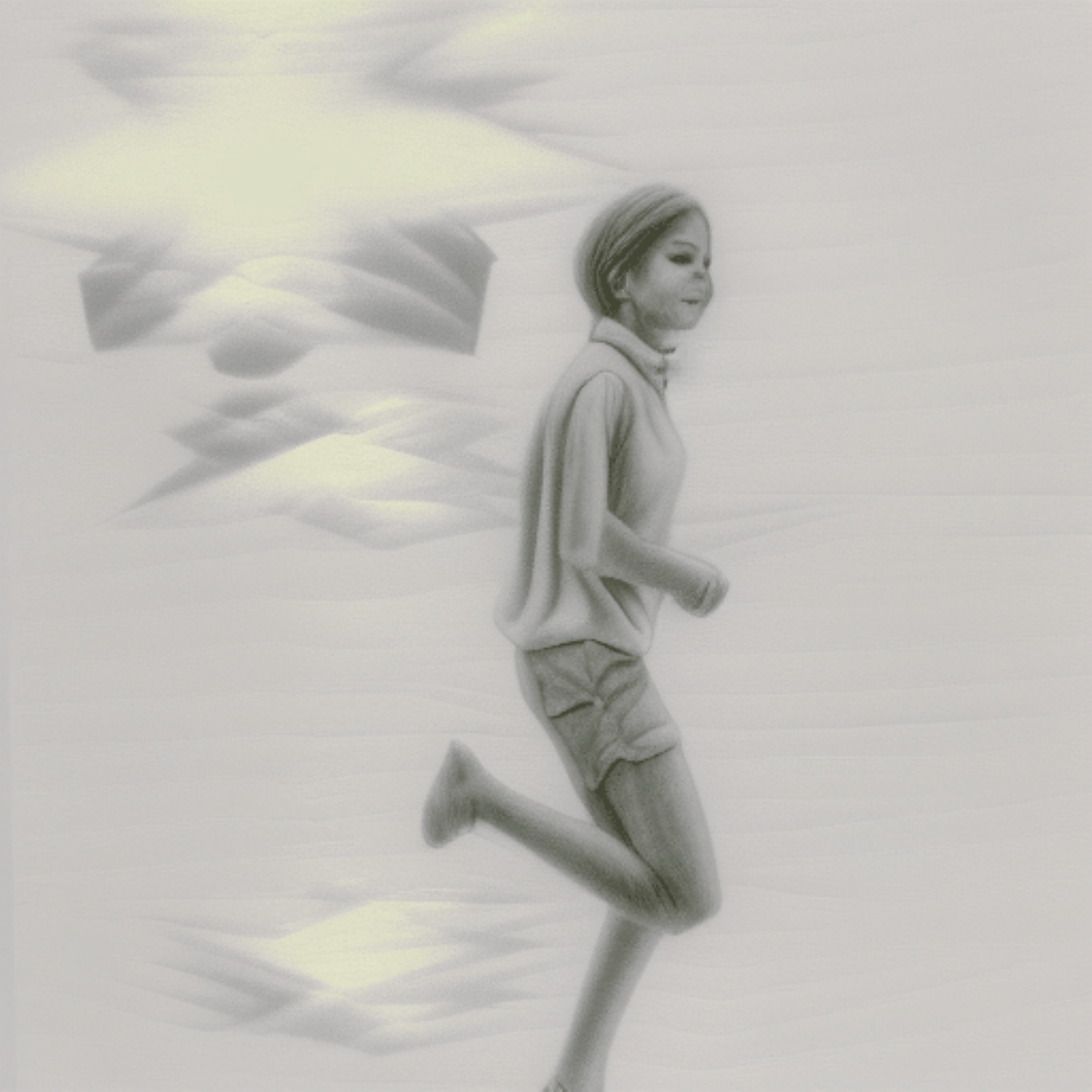}
\includegraphics[width=0.10\textwidth]{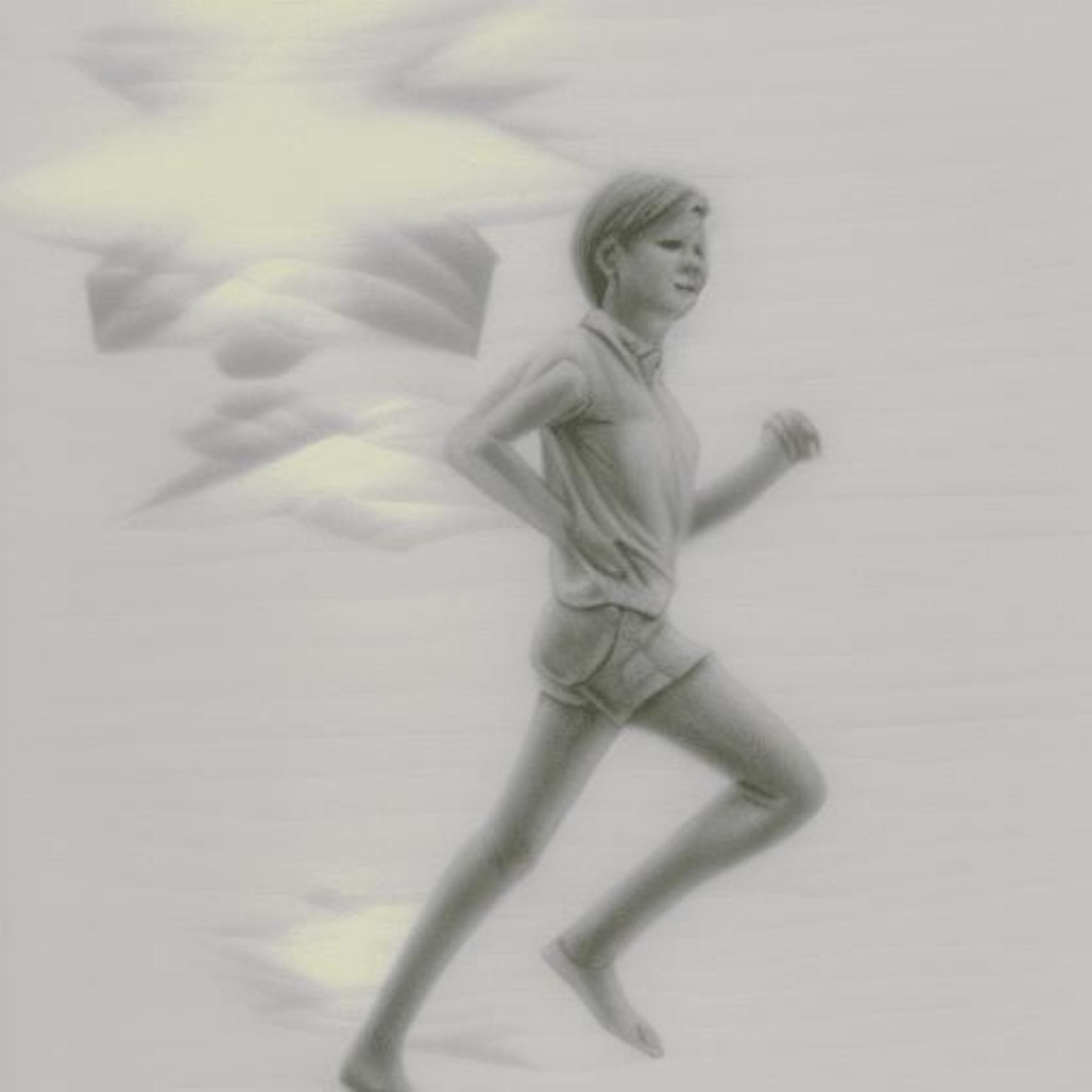}
\includegraphics[width=0.10\textwidth]{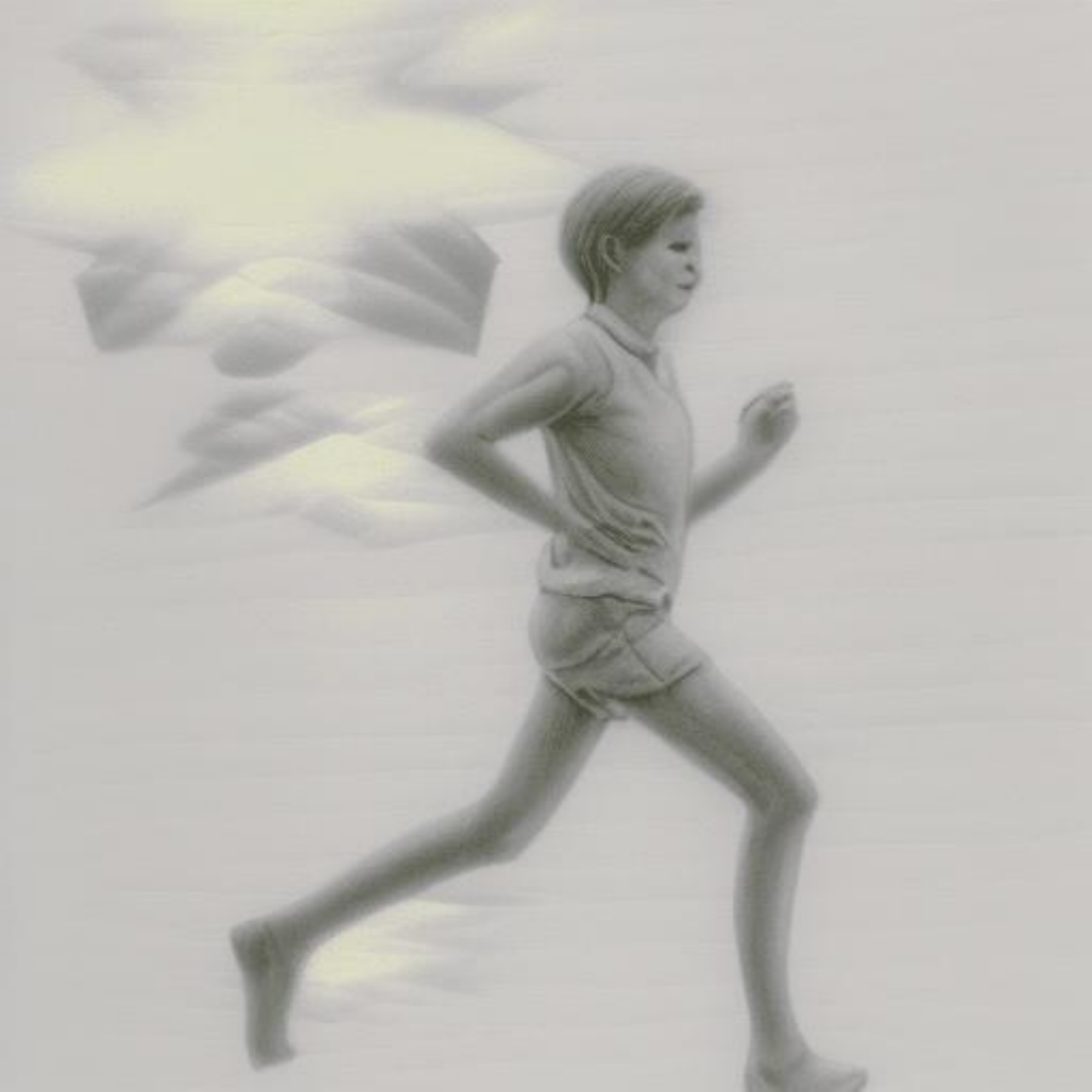}
\includegraphics[width=0.10\textwidth]{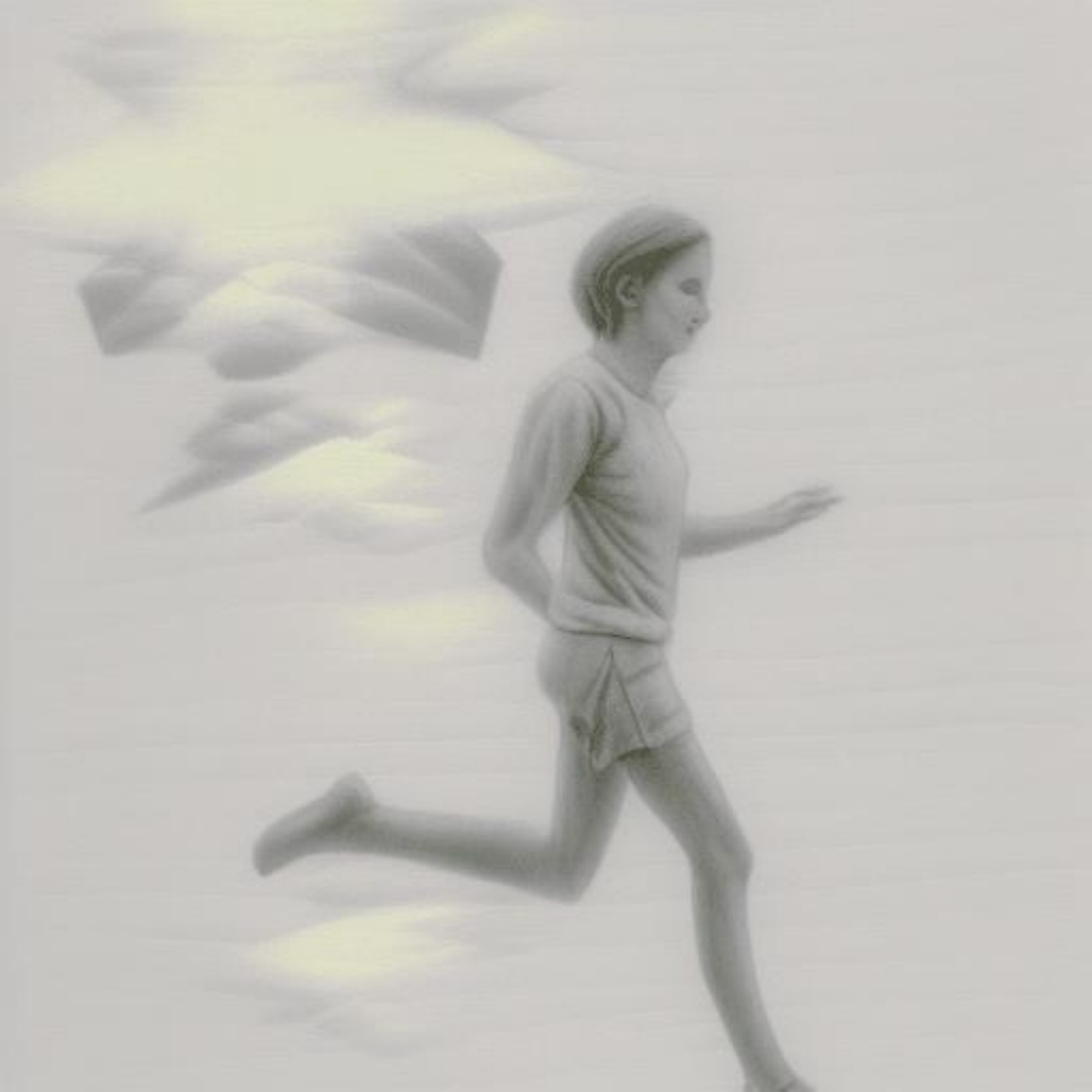}
\includegraphics[width=0.10\textwidth]{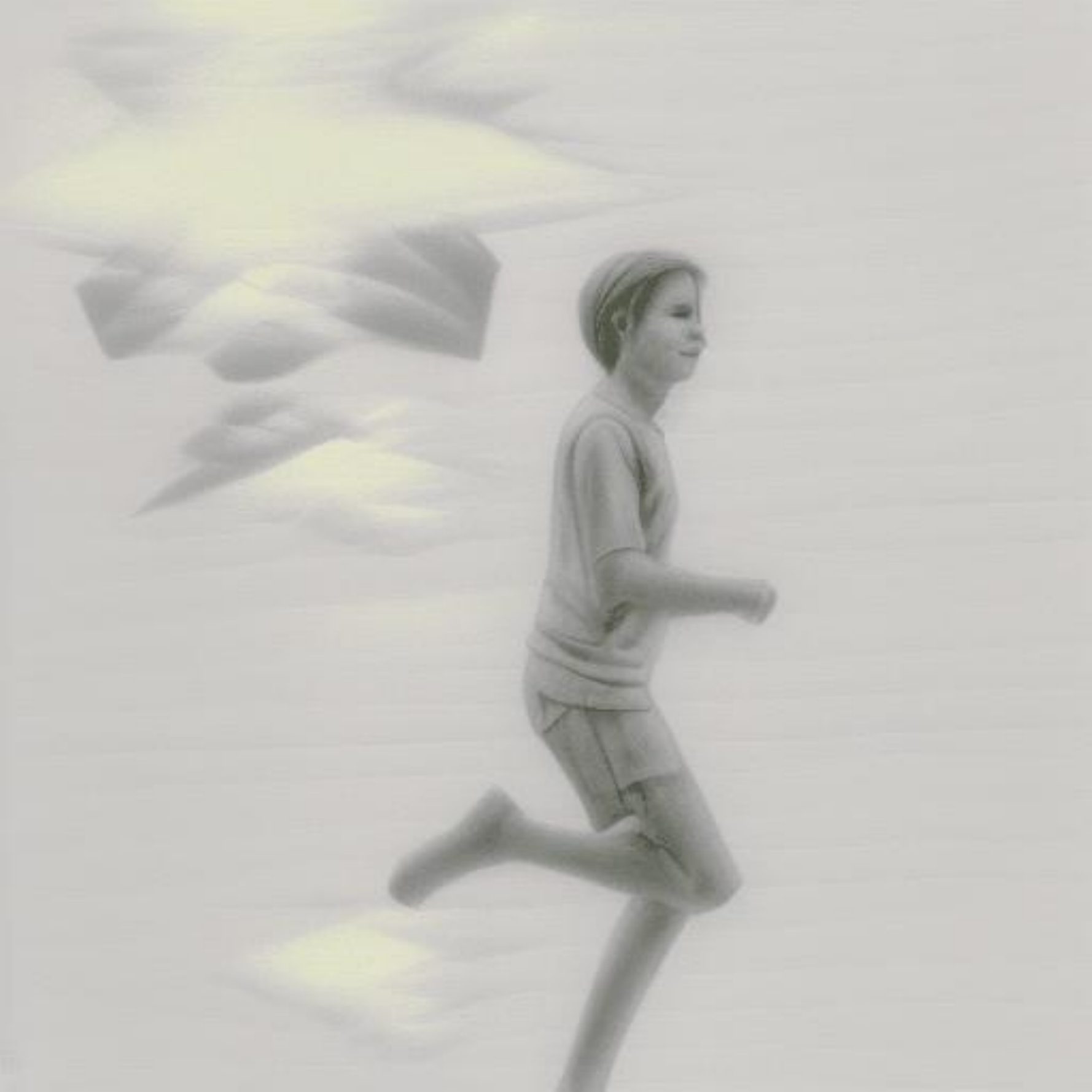}
\includegraphics[width=0.10\textwidth]{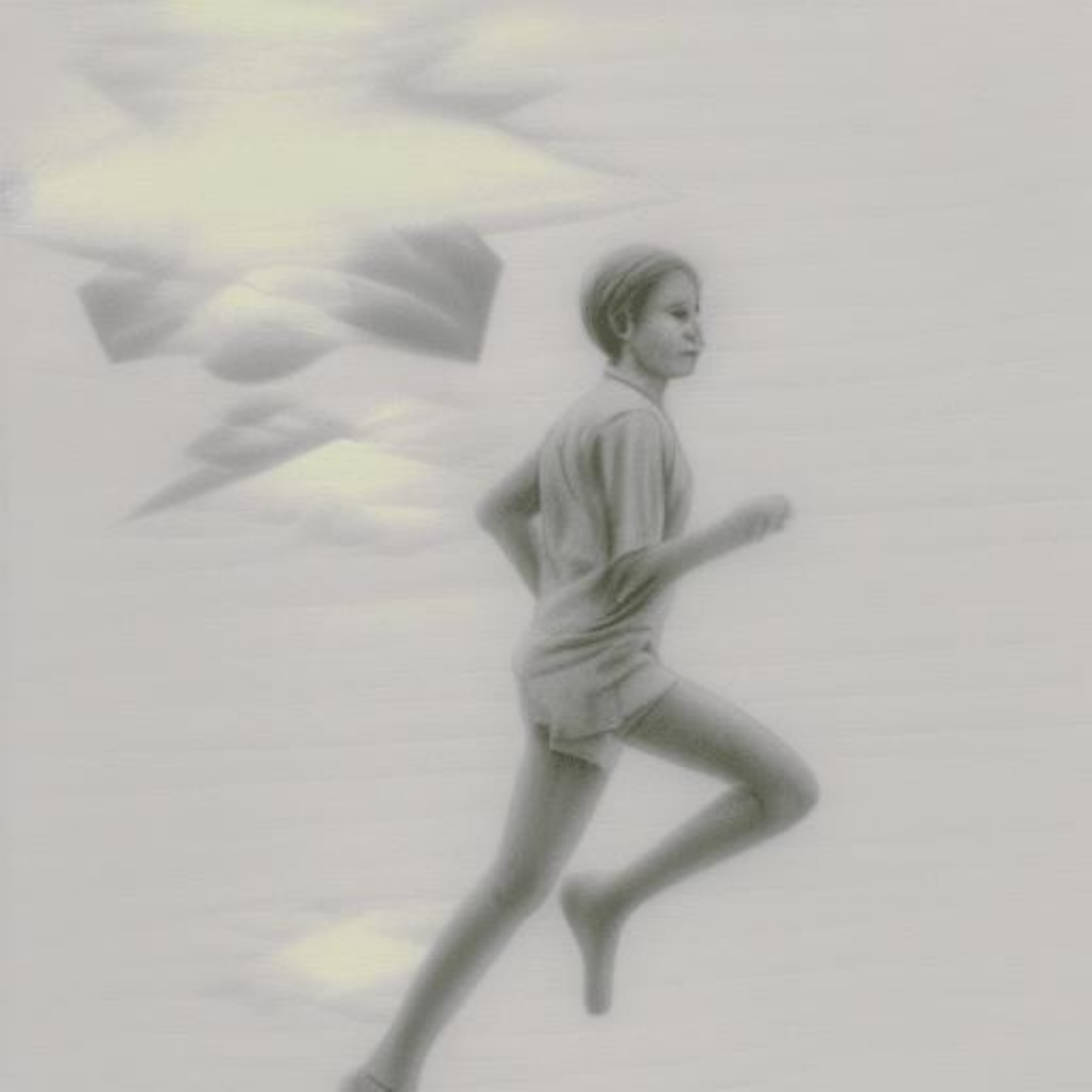}
\includegraphics[width=0.10\textwidth]{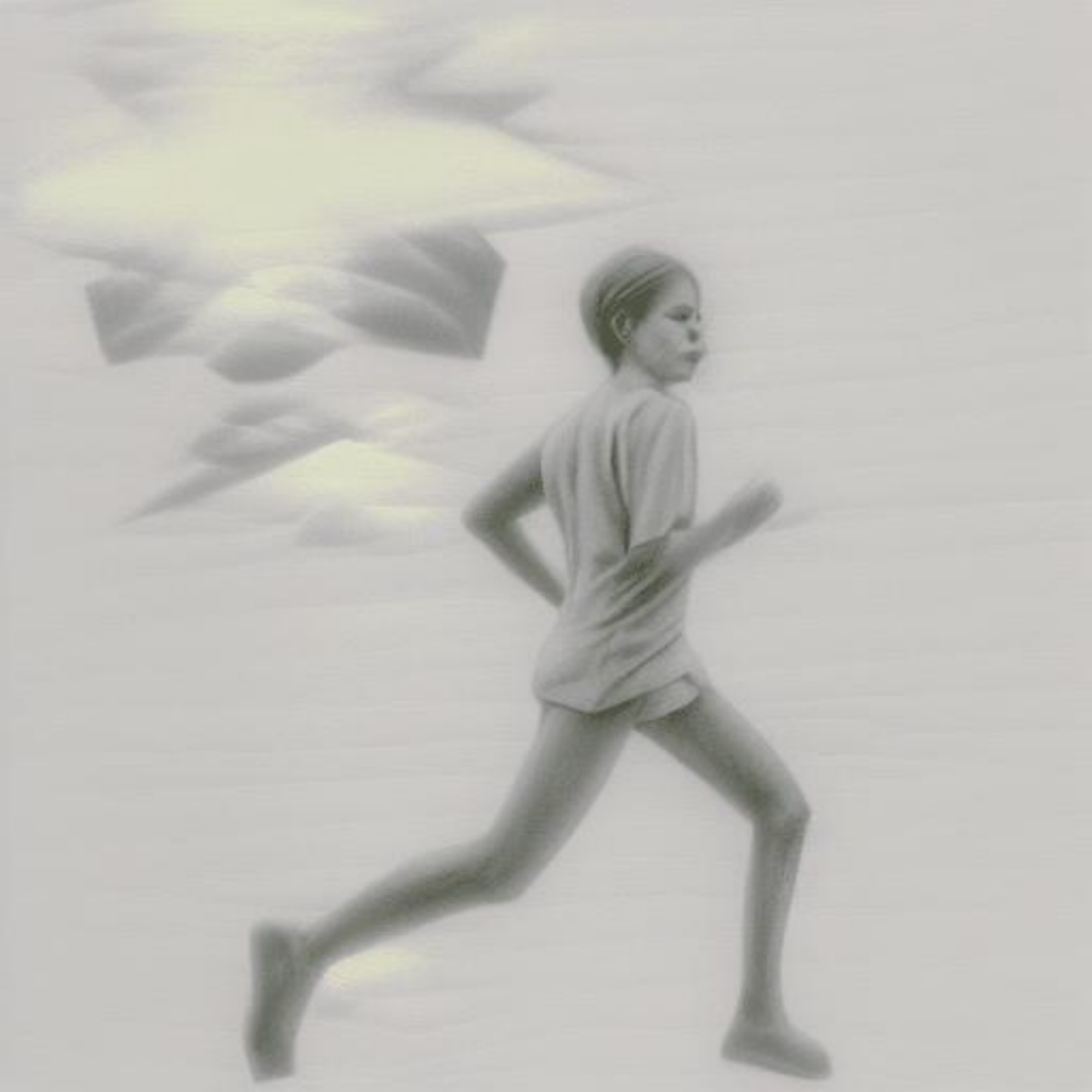}
\includegraphics[width=0.10\textwidth]{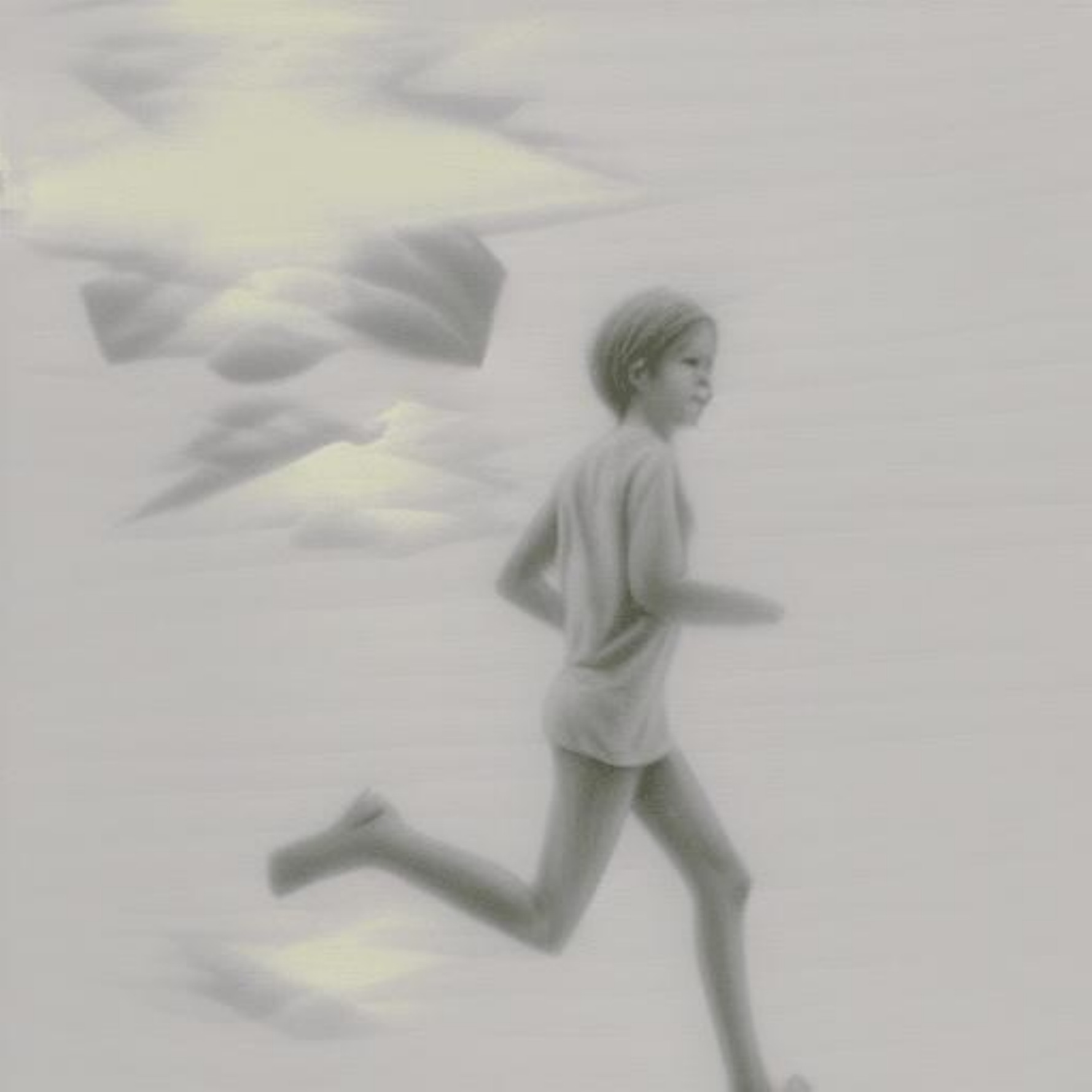}

\makebox[0.12\textwidth]{A man is running, \textcolor{blue}{\textbf{pixar style}}.}\\
\includegraphics[width=0.10\textwidth]{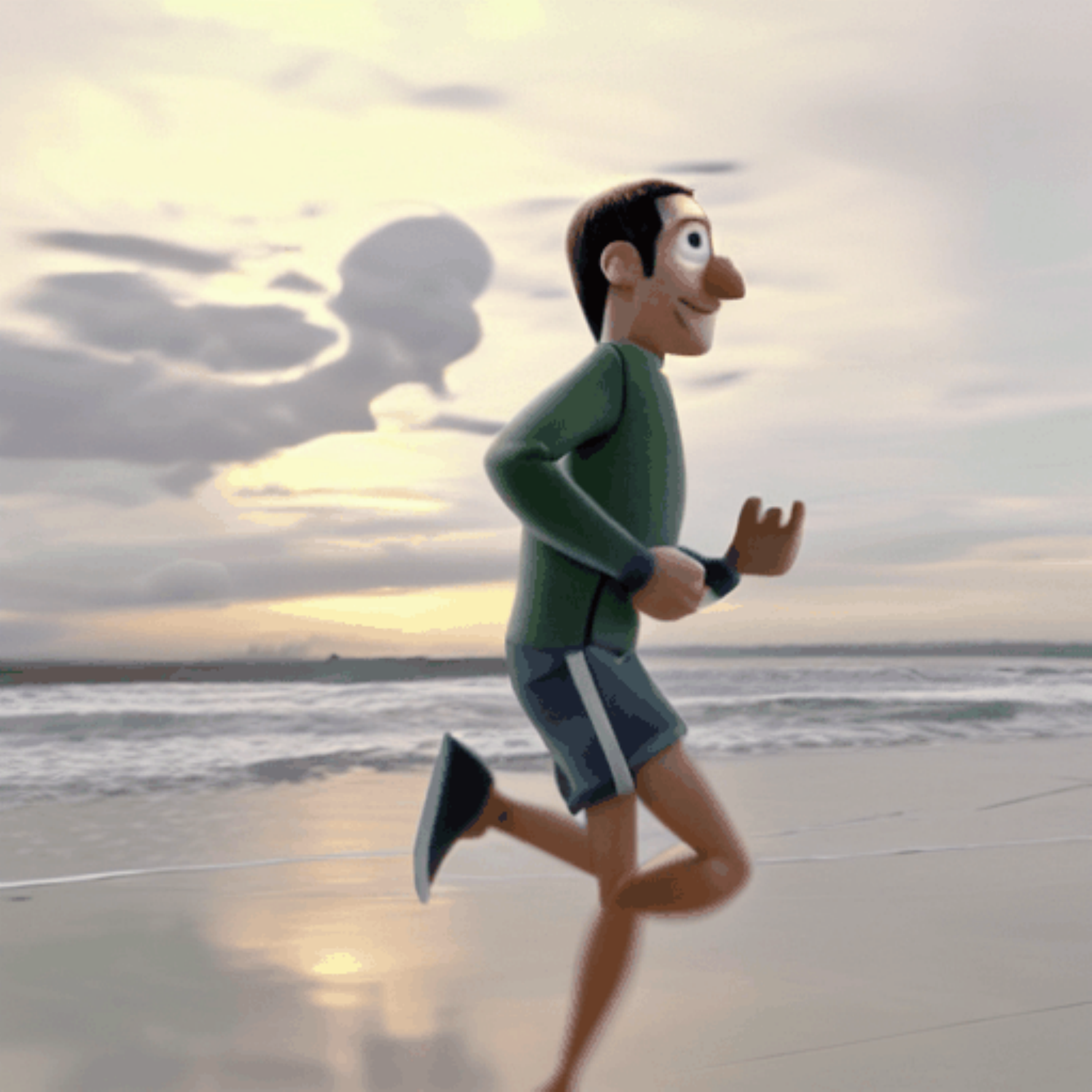}
\includegraphics[width=0.10\textwidth]{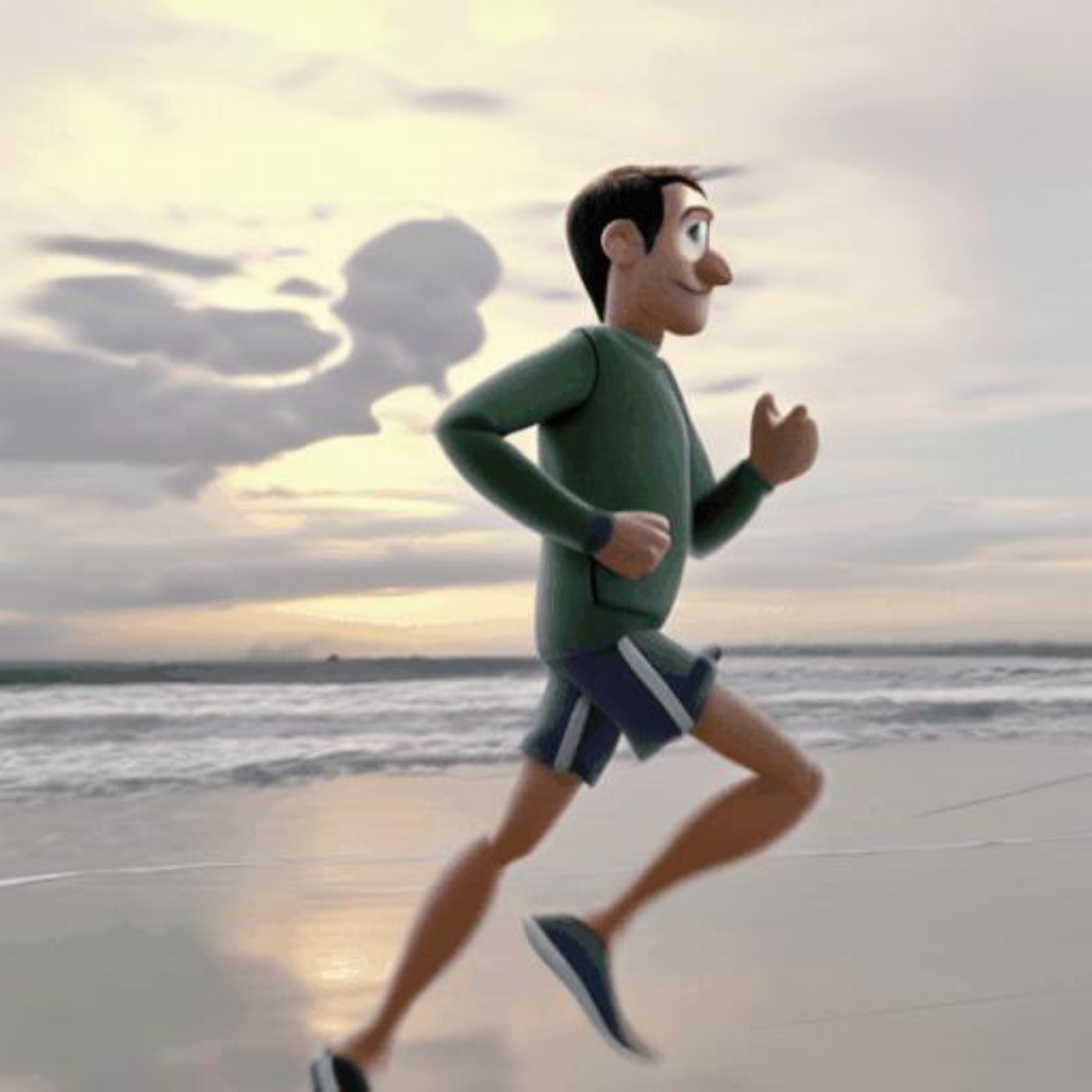}
\includegraphics[width=0.10\textwidth]{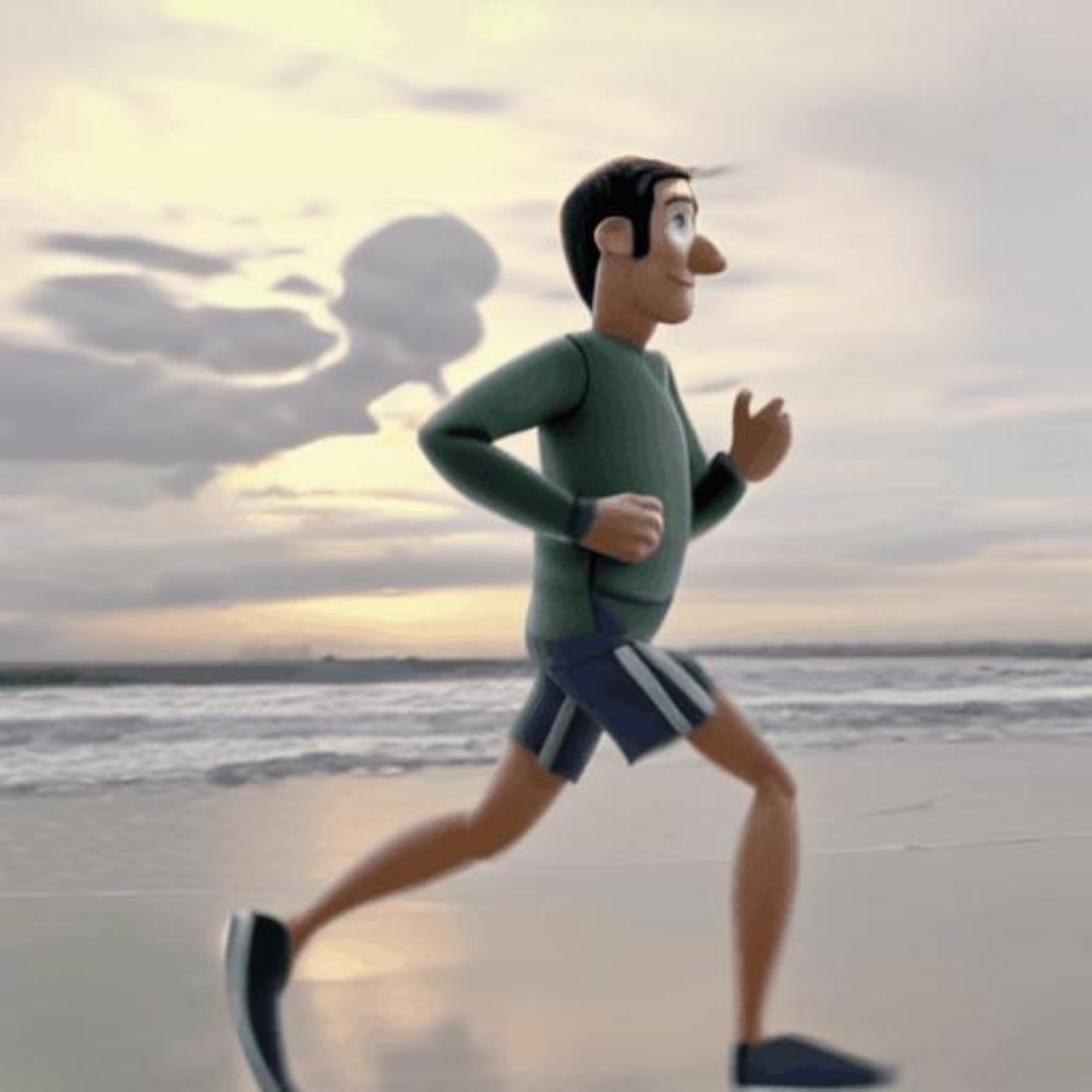}
\includegraphics[width=0.10\textwidth]{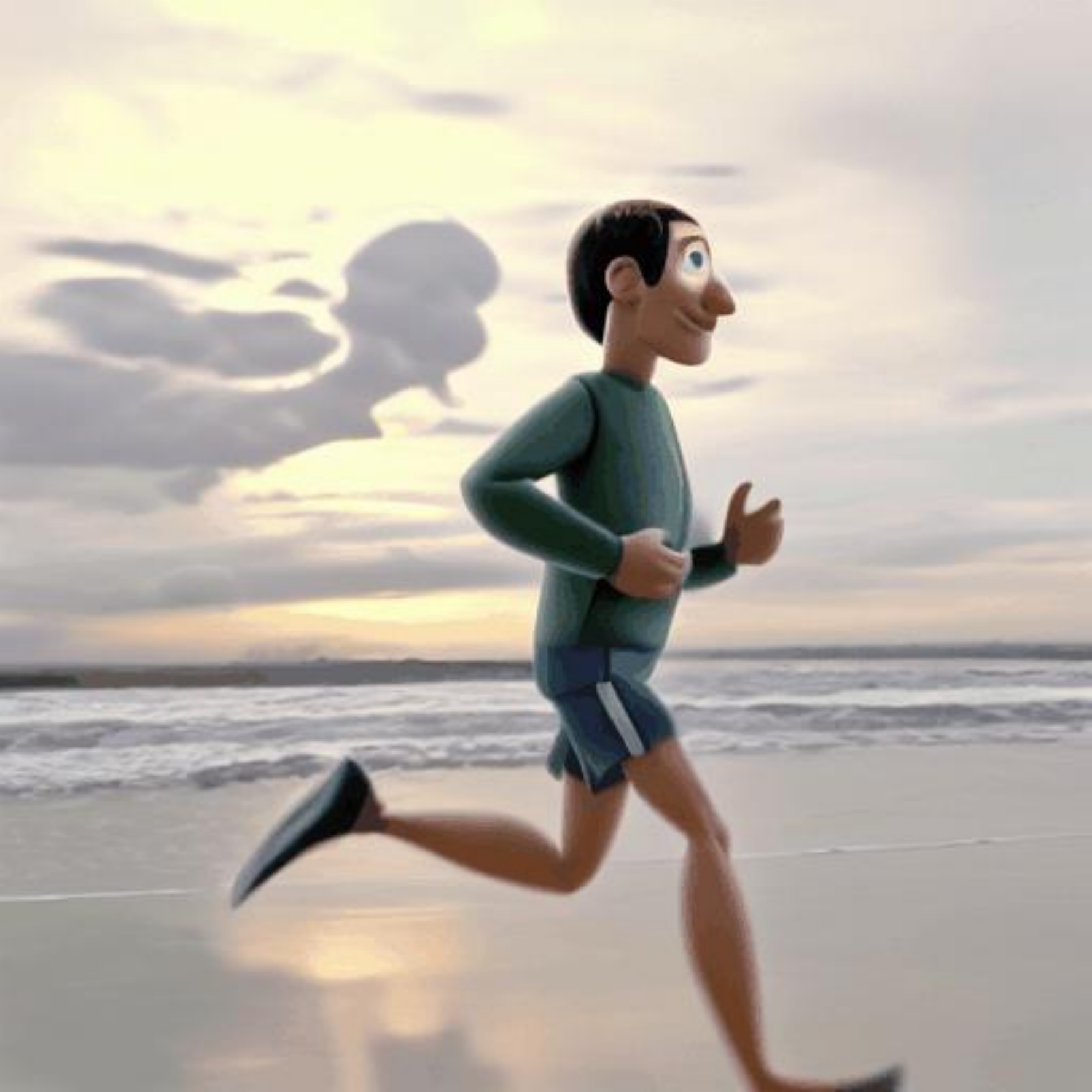}
\includegraphics[width=0.10\textwidth]{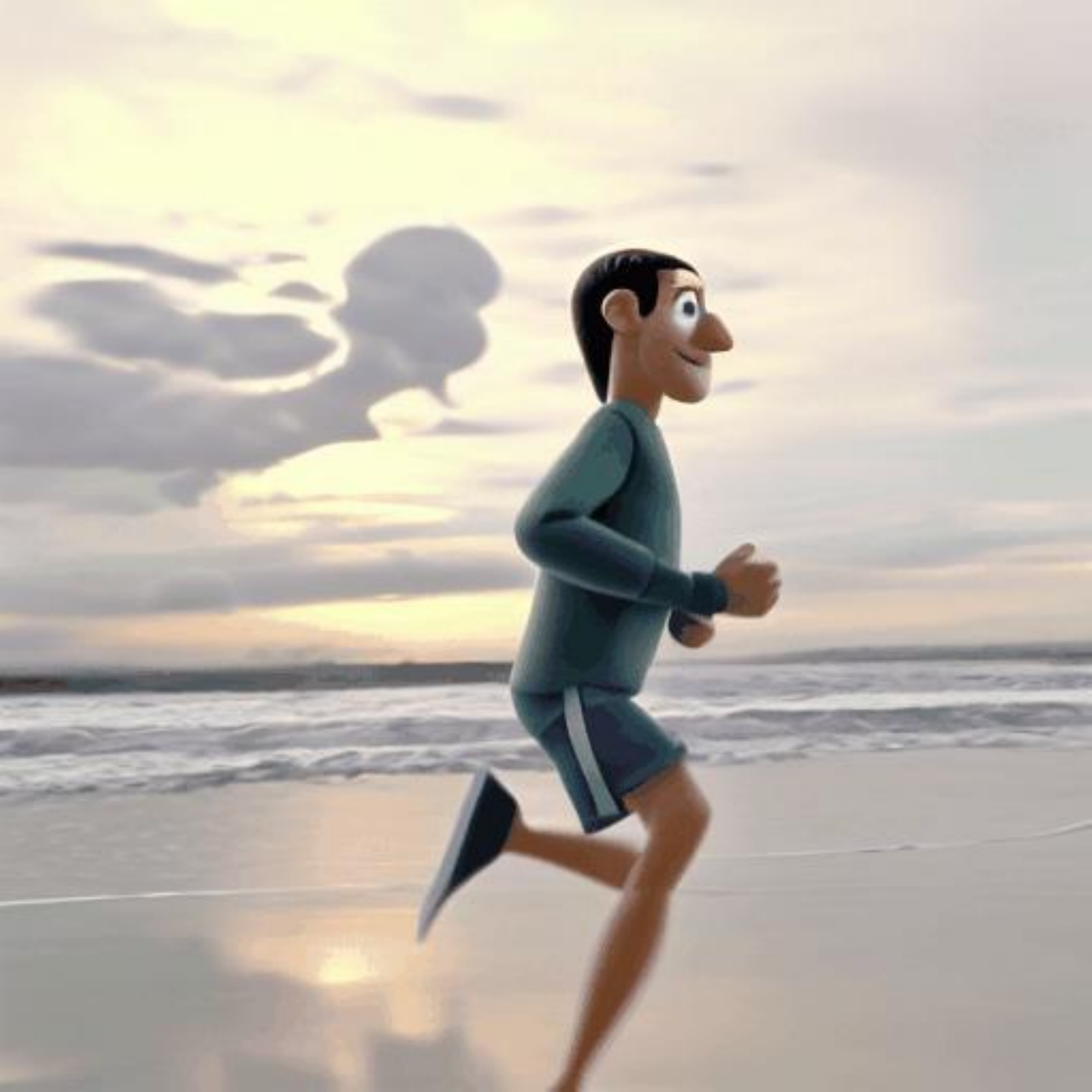}
\includegraphics[width=0.10\textwidth]{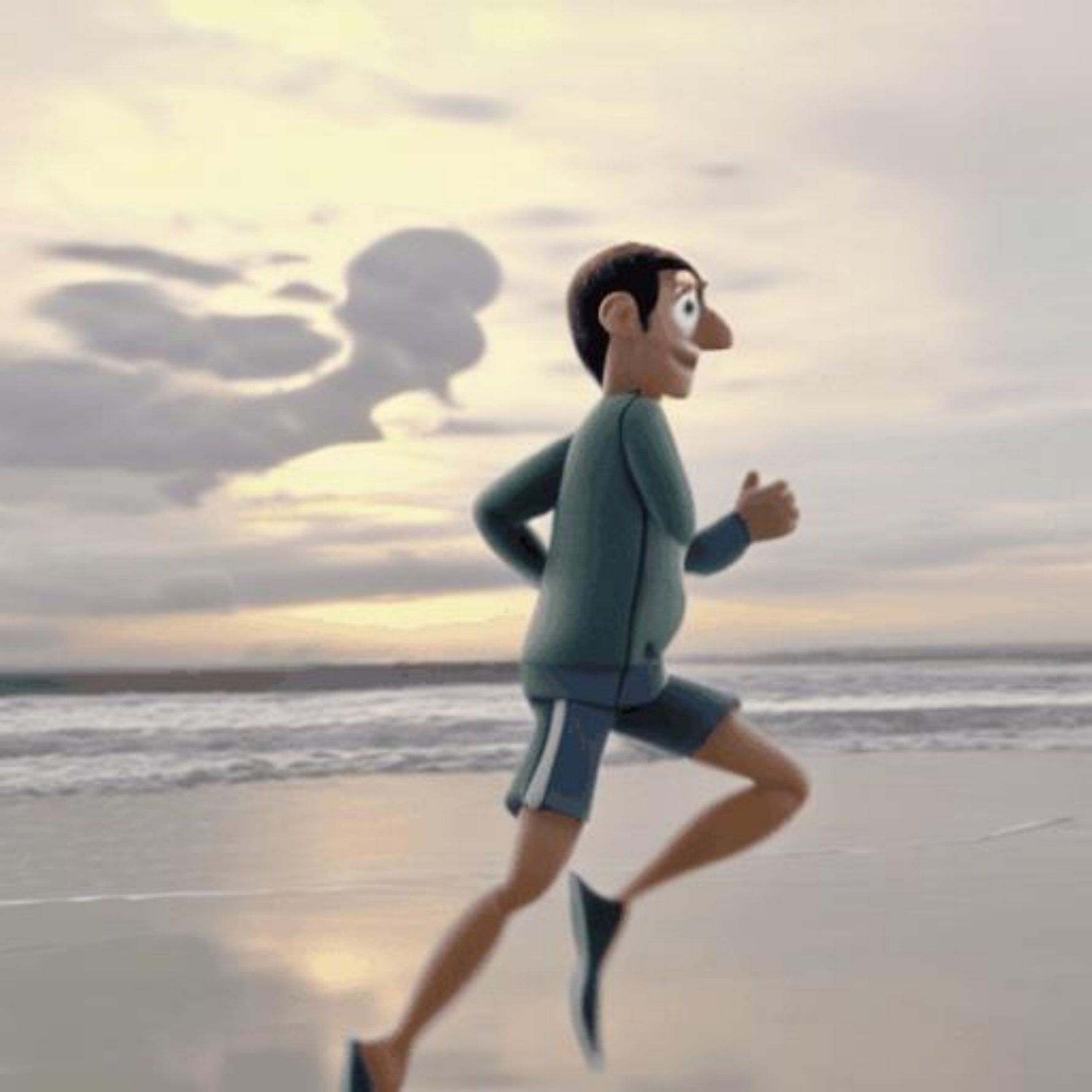}
\includegraphics[width=0.10\textwidth]{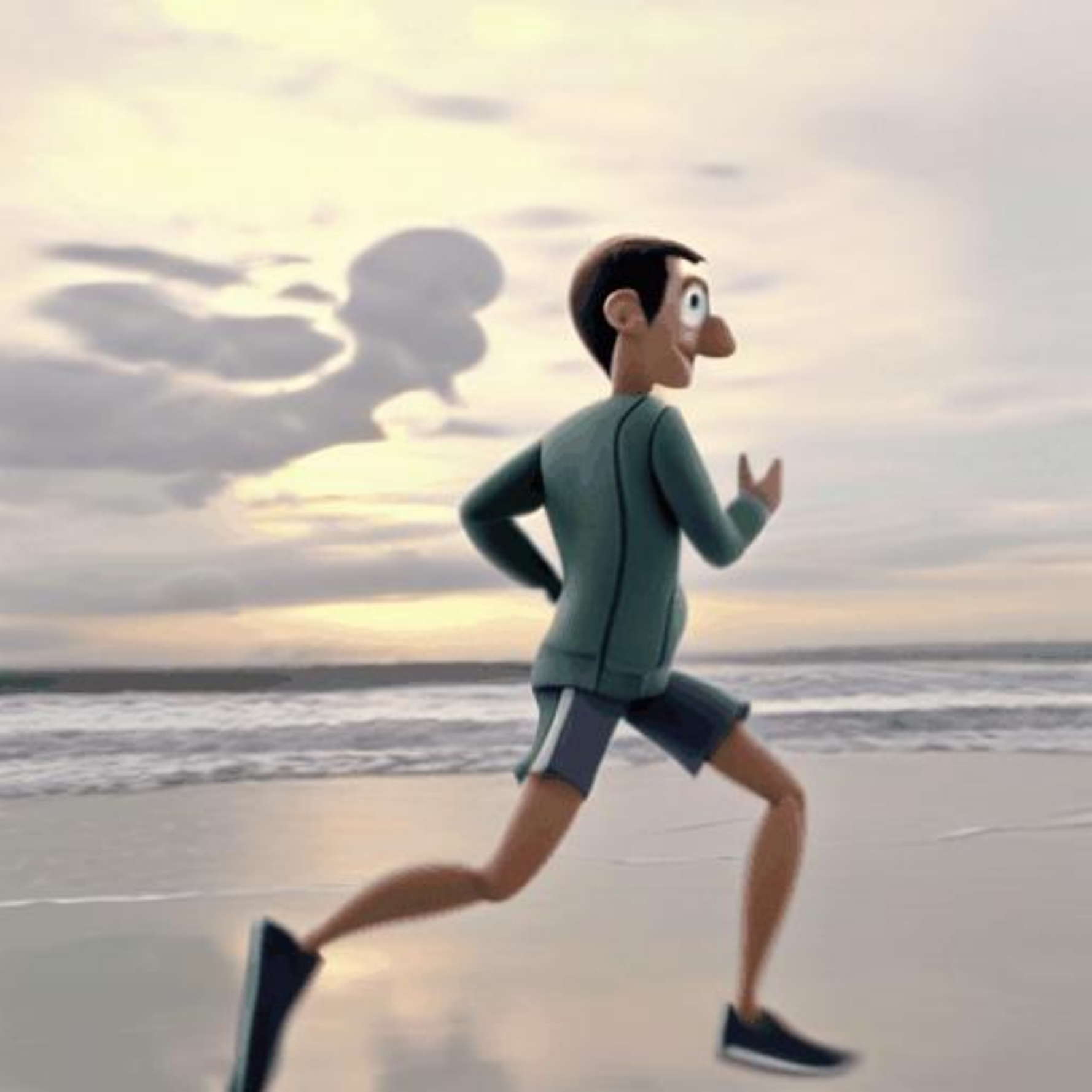}
\includegraphics[width=0.10\textwidth]{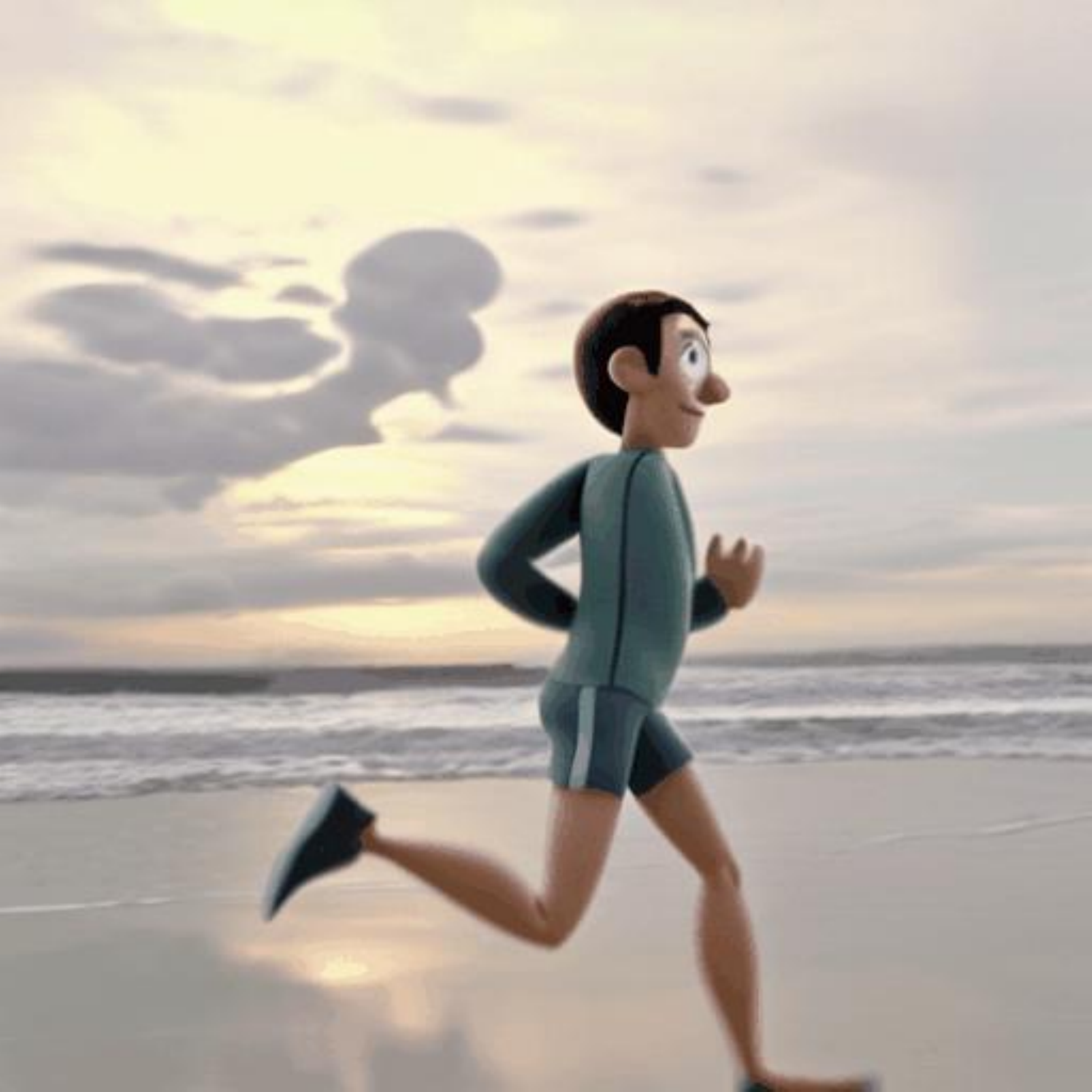}

\caption{\textbf{Qualitative Results} Additional selected samples for our model.}
\label{fig:supp_qual1}
\end{center}
\end{figure*}

\begin{figure*}
\vspace{0.8em}
\begin{center}
\makebox[0.12\textwidth]{\colorbox{pink}{\textbf{Training video}} A man is playing a guitar}\\
\includegraphics[width=0.10\textwidth]{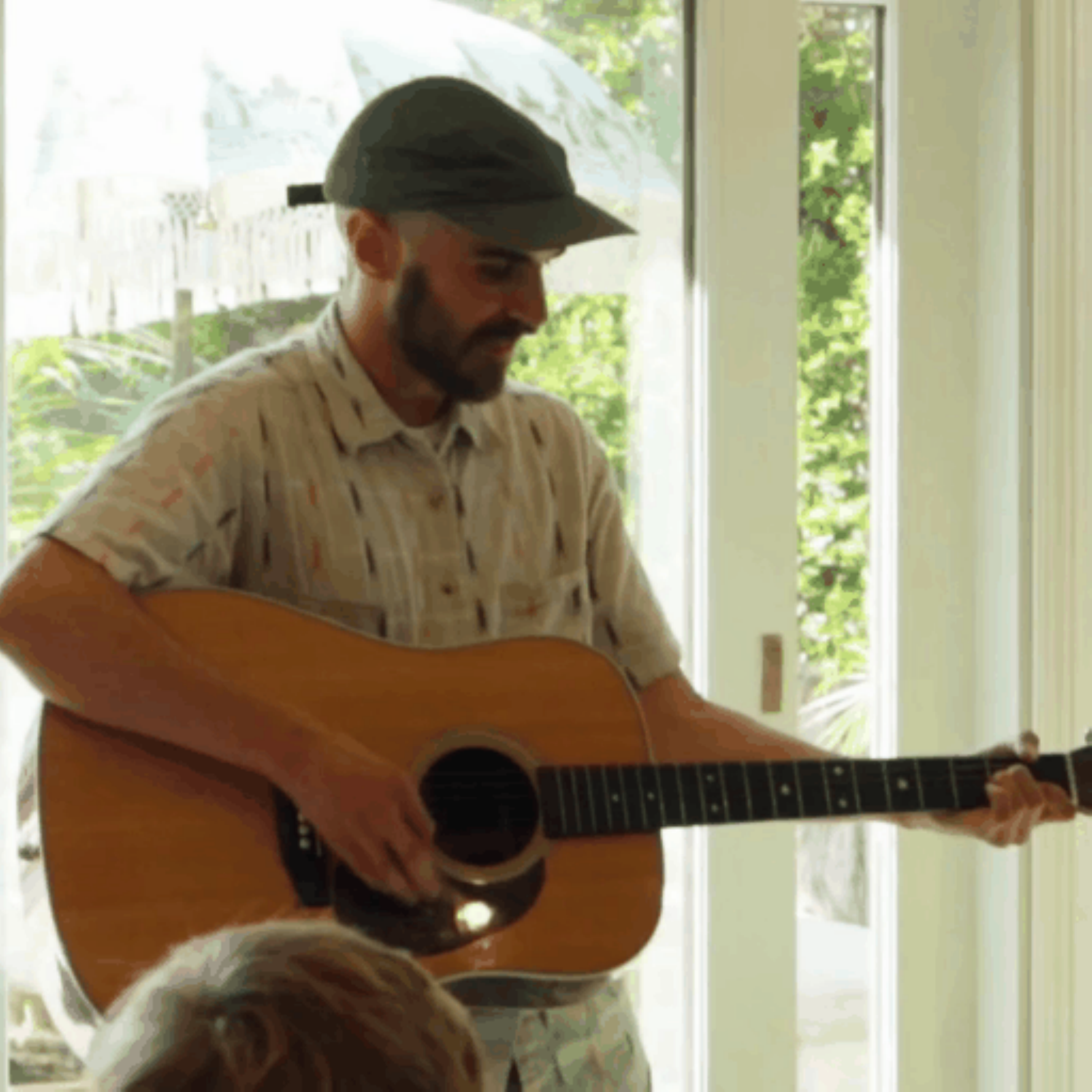}
\includegraphics[width=0.10\textwidth]{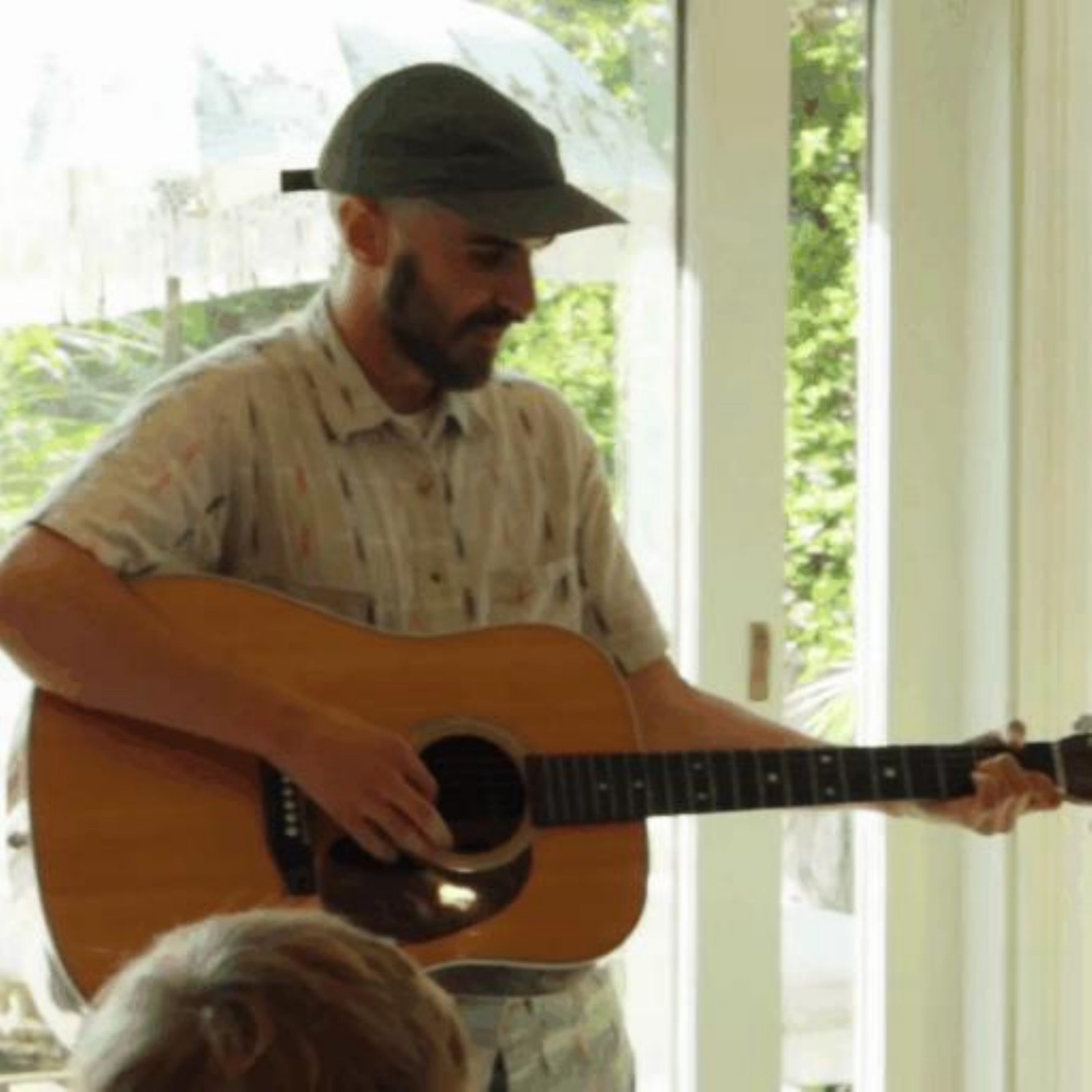}
\includegraphics[width=0.10\textwidth]{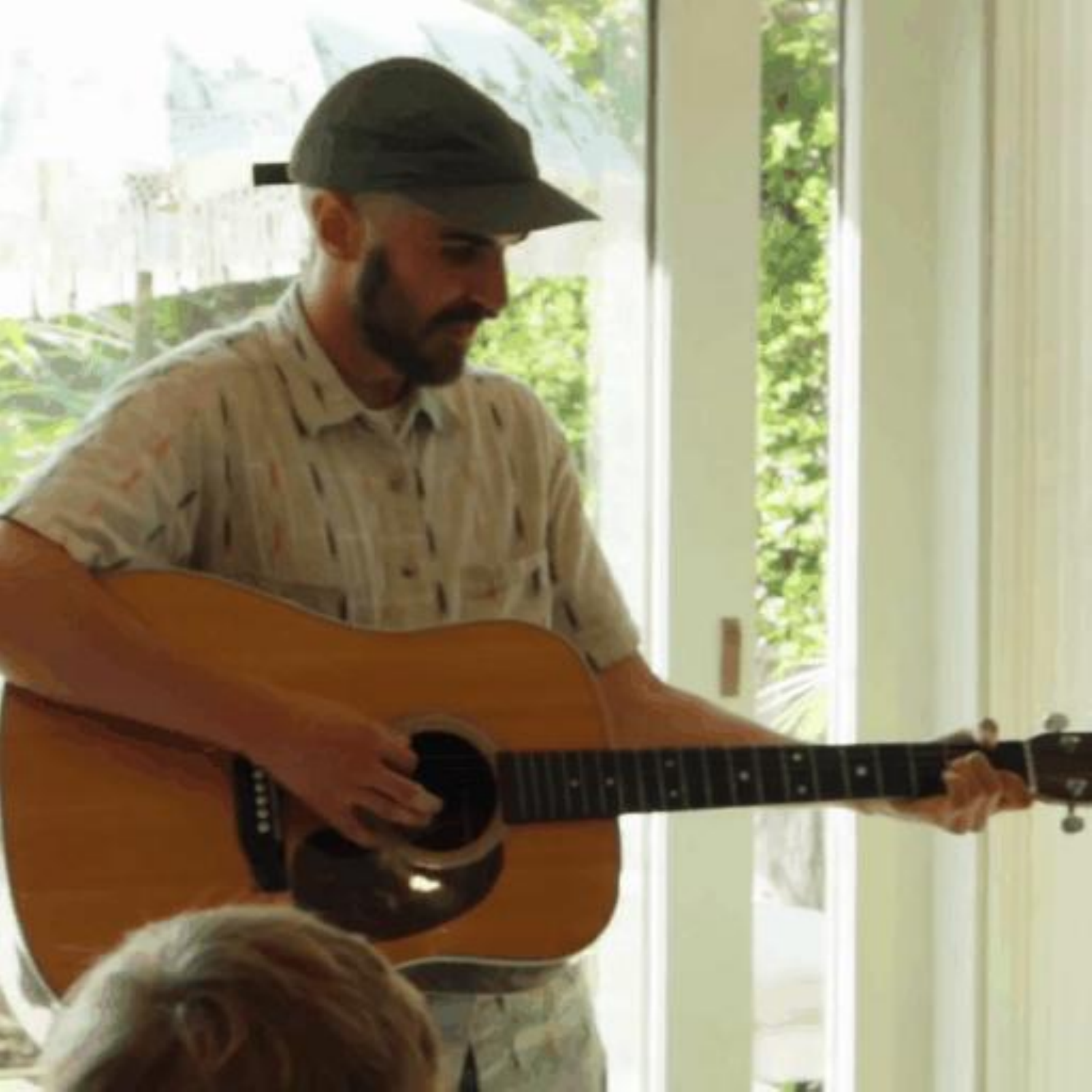}
\includegraphics[width=0.10\textwidth]{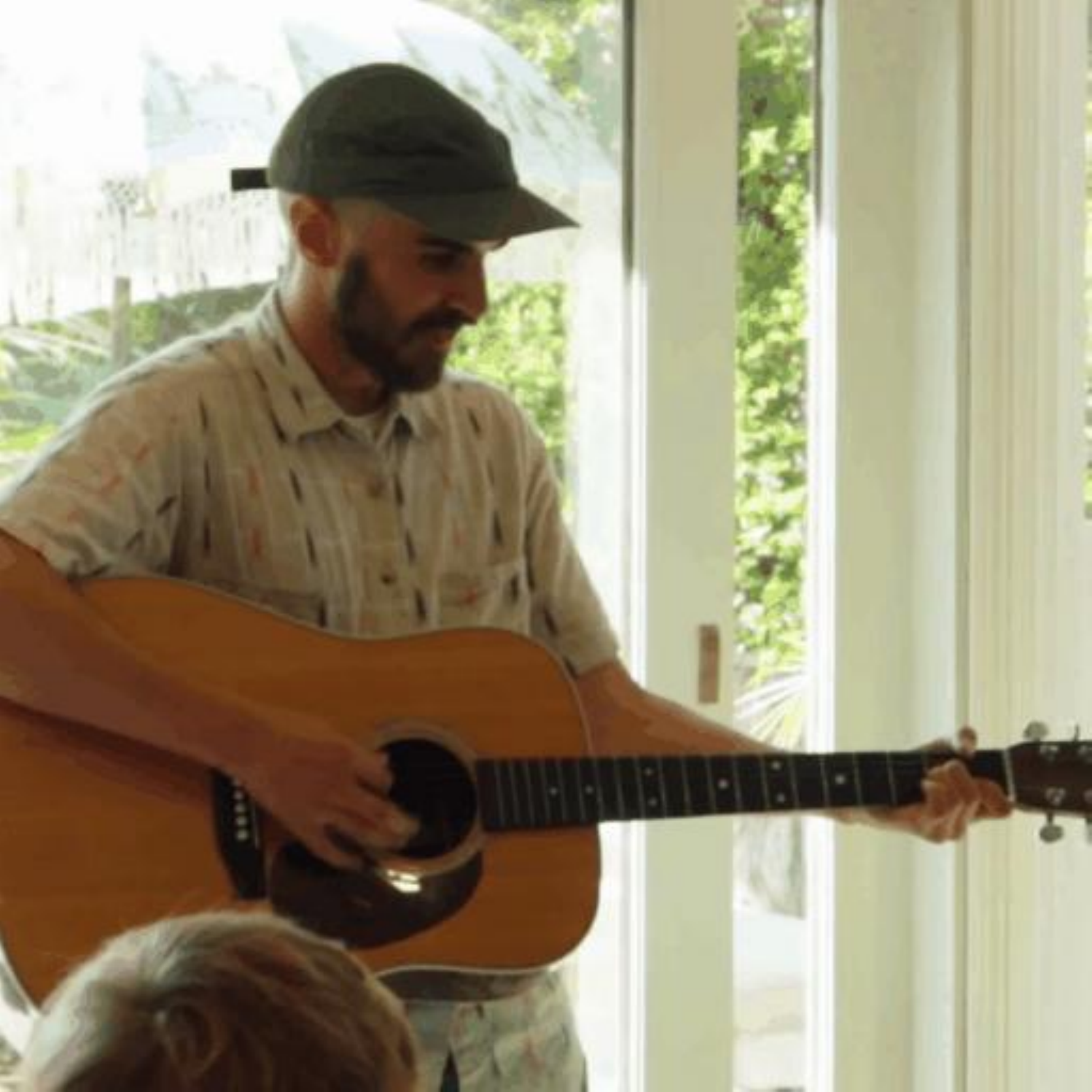}
\includegraphics[width=0.10\textwidth]{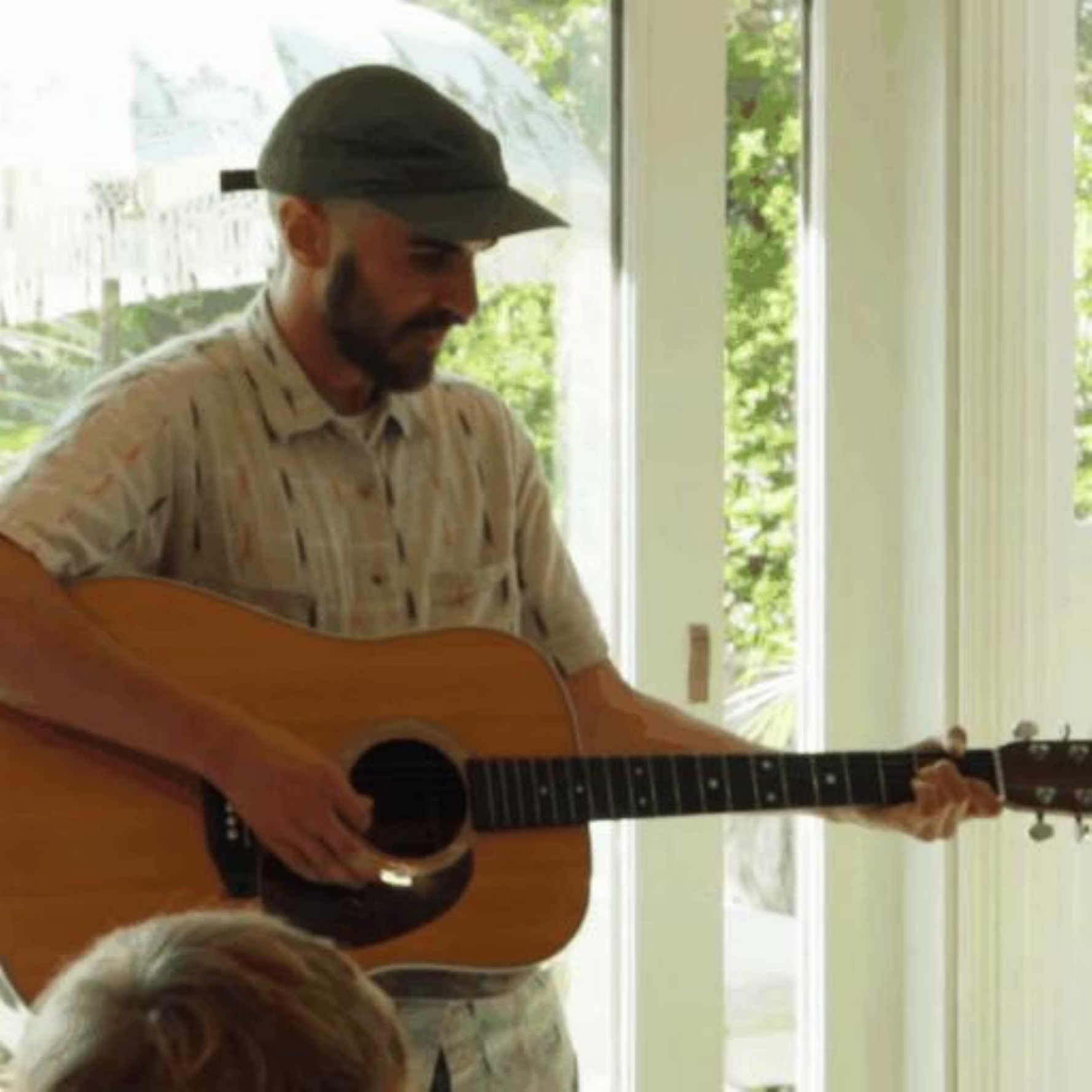}
\includegraphics[width=0.10\textwidth]{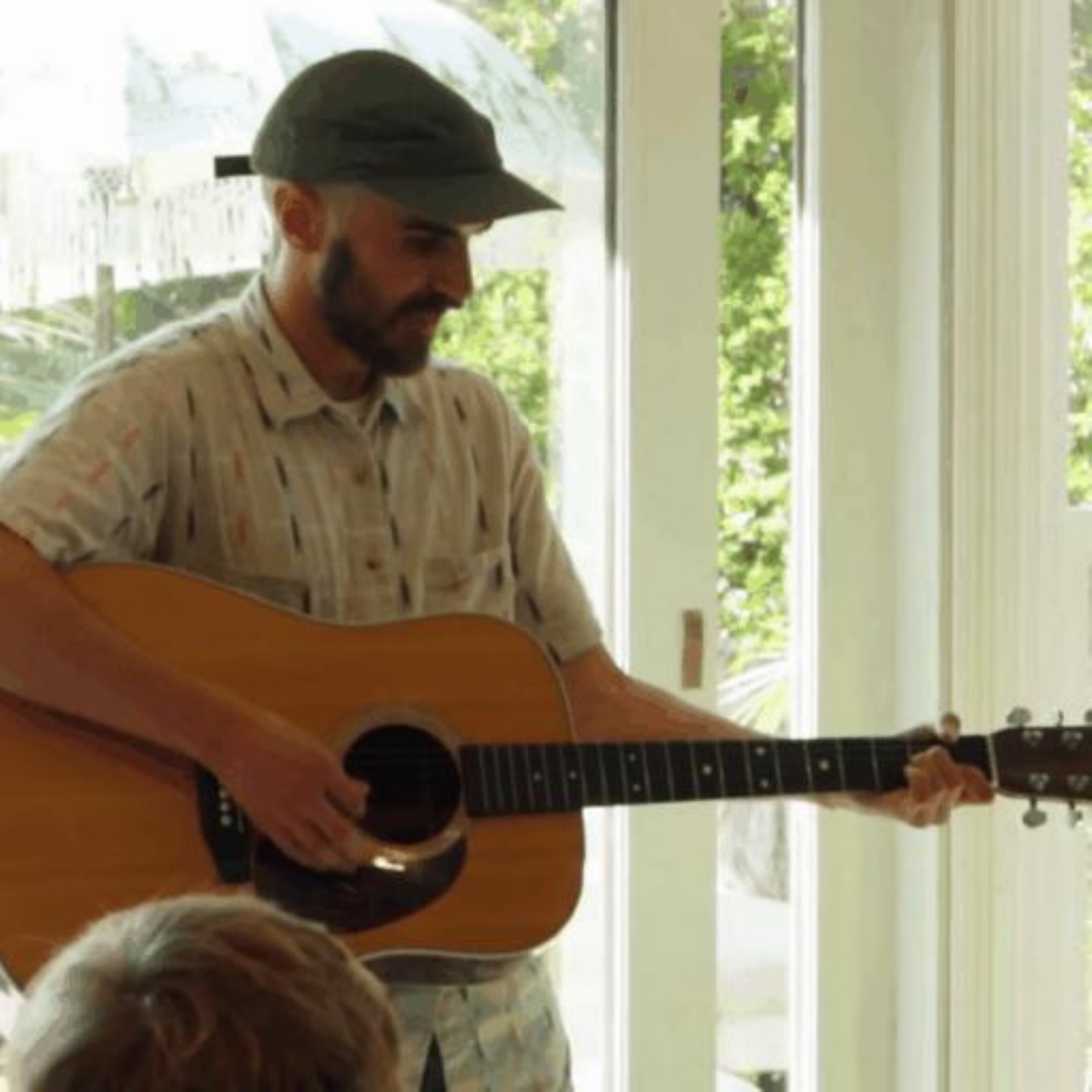}
\includegraphics[width=0.10\textwidth]{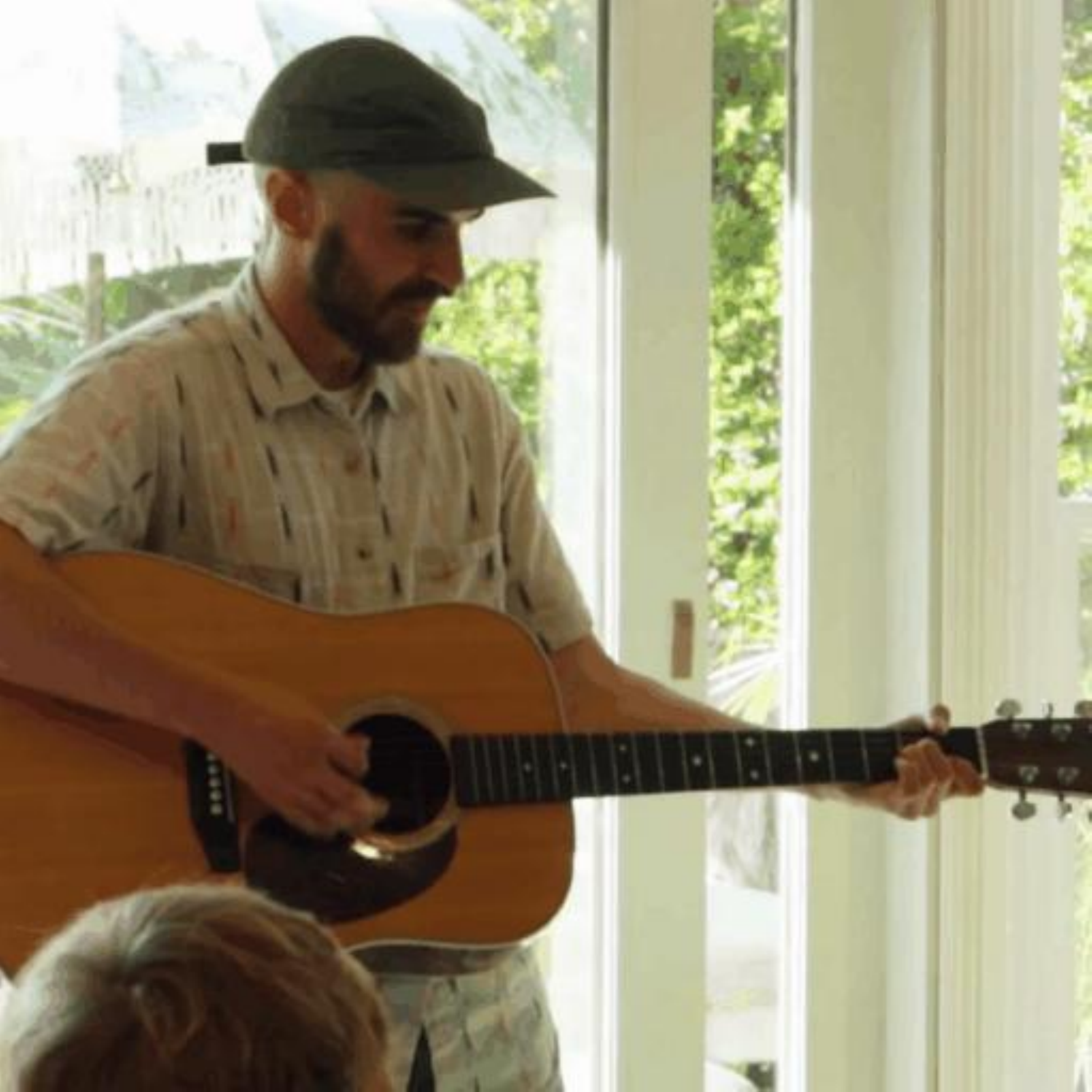}
\includegraphics[width=0.10\textwidth]{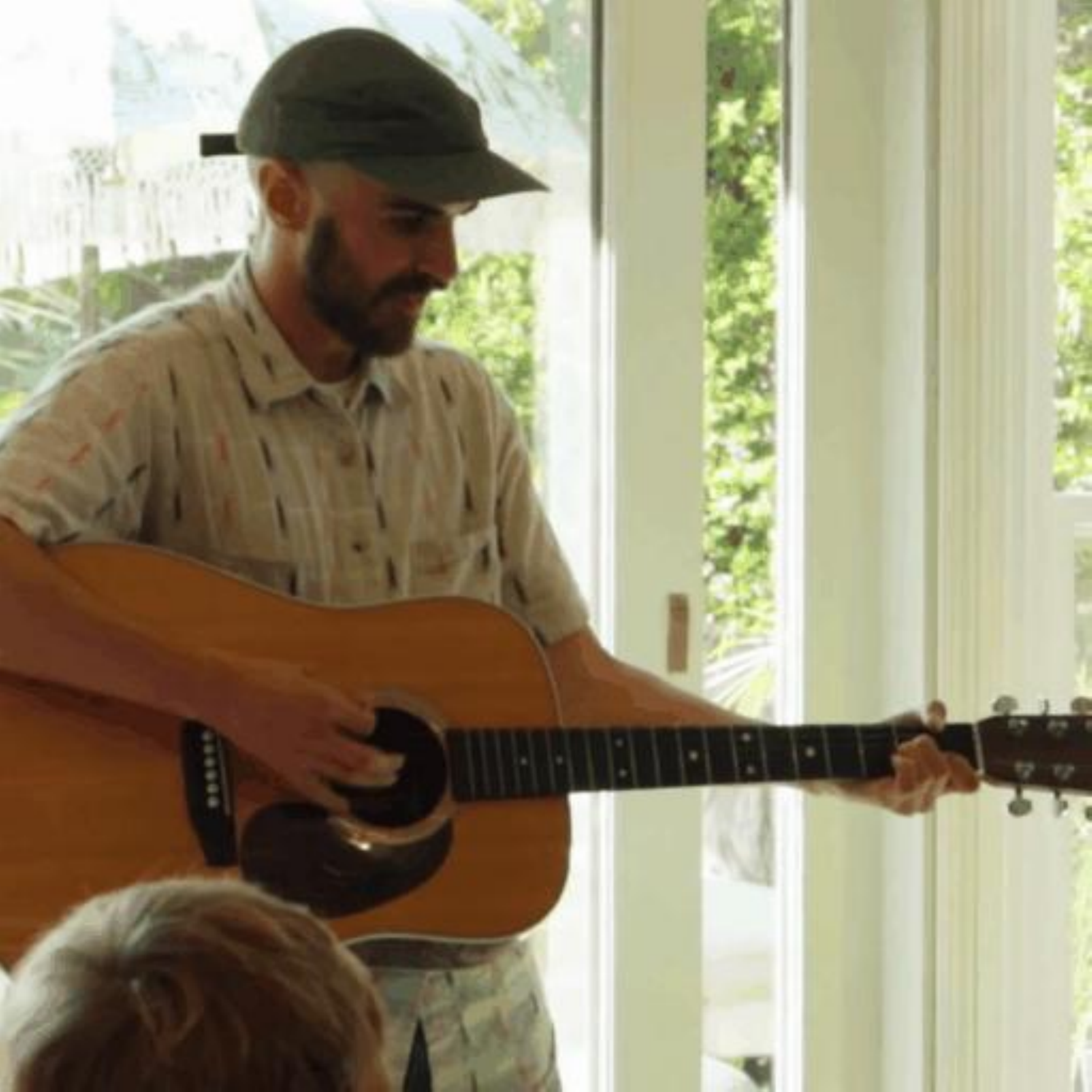}

\makebox[0.12\textwidth]{A \textcolor{blue}{\textbf{bear}} is playing a guitar.}\\
\includegraphics[width=0.10\textwidth]{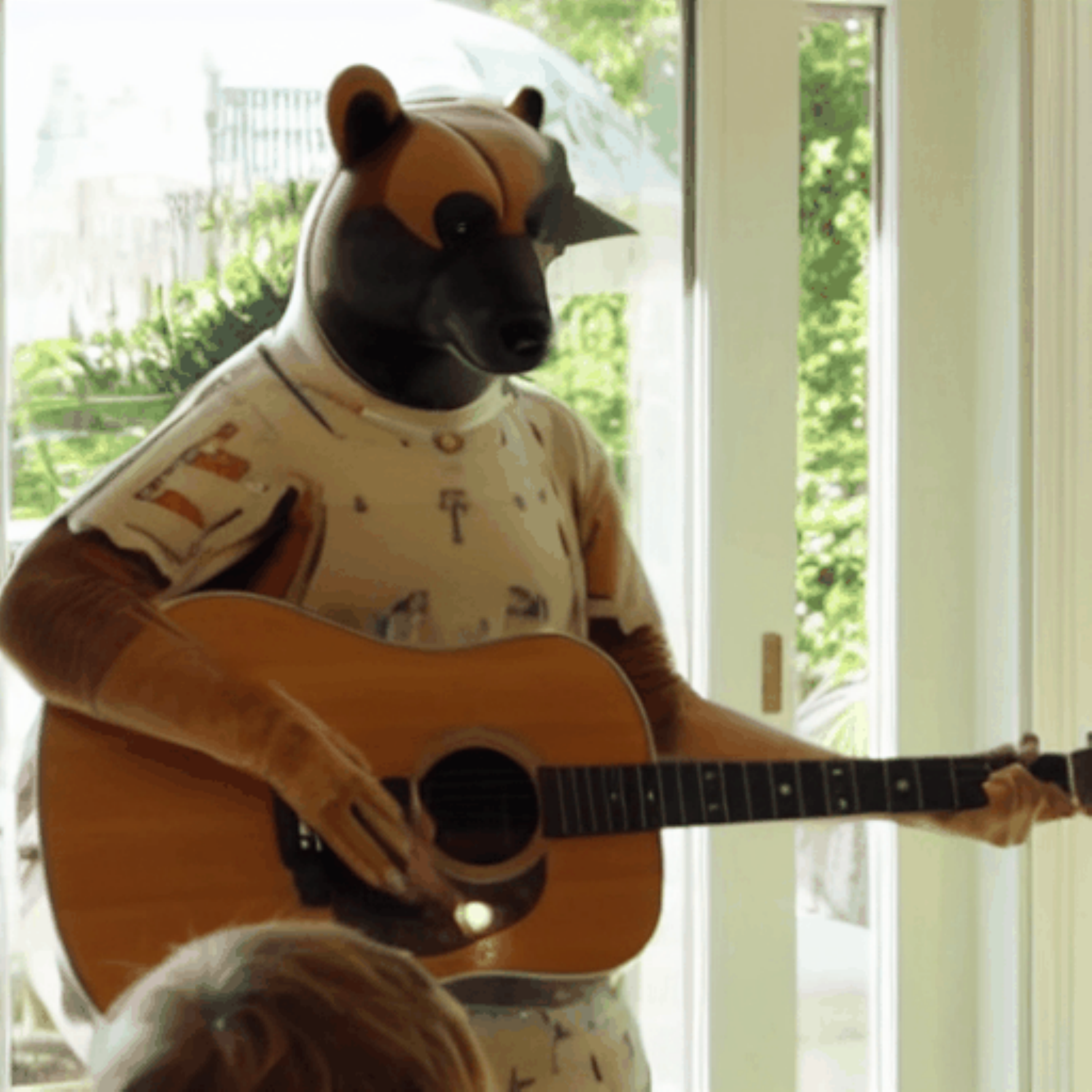}
\includegraphics[width=0.10\textwidth]{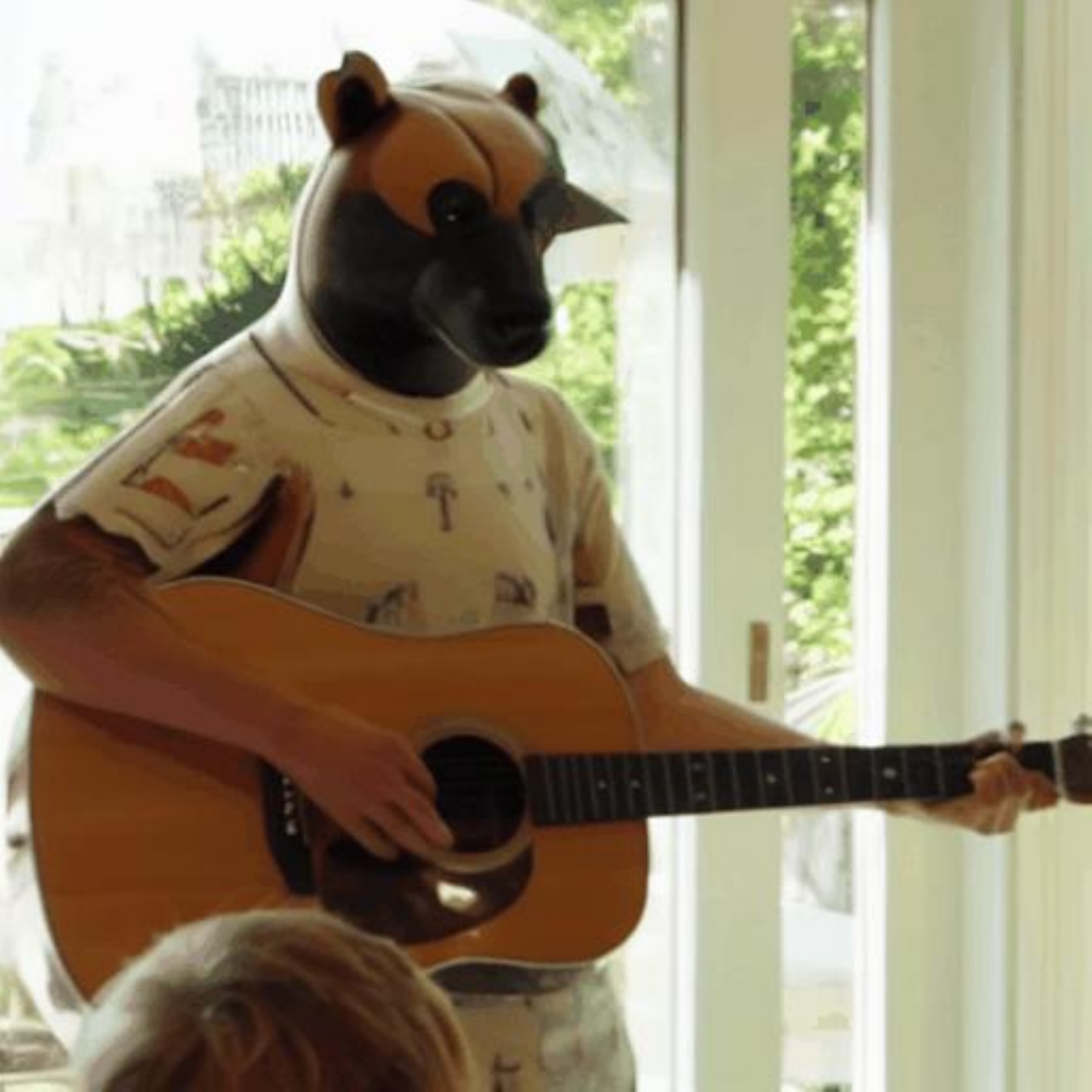}
\includegraphics[width=0.10\textwidth]{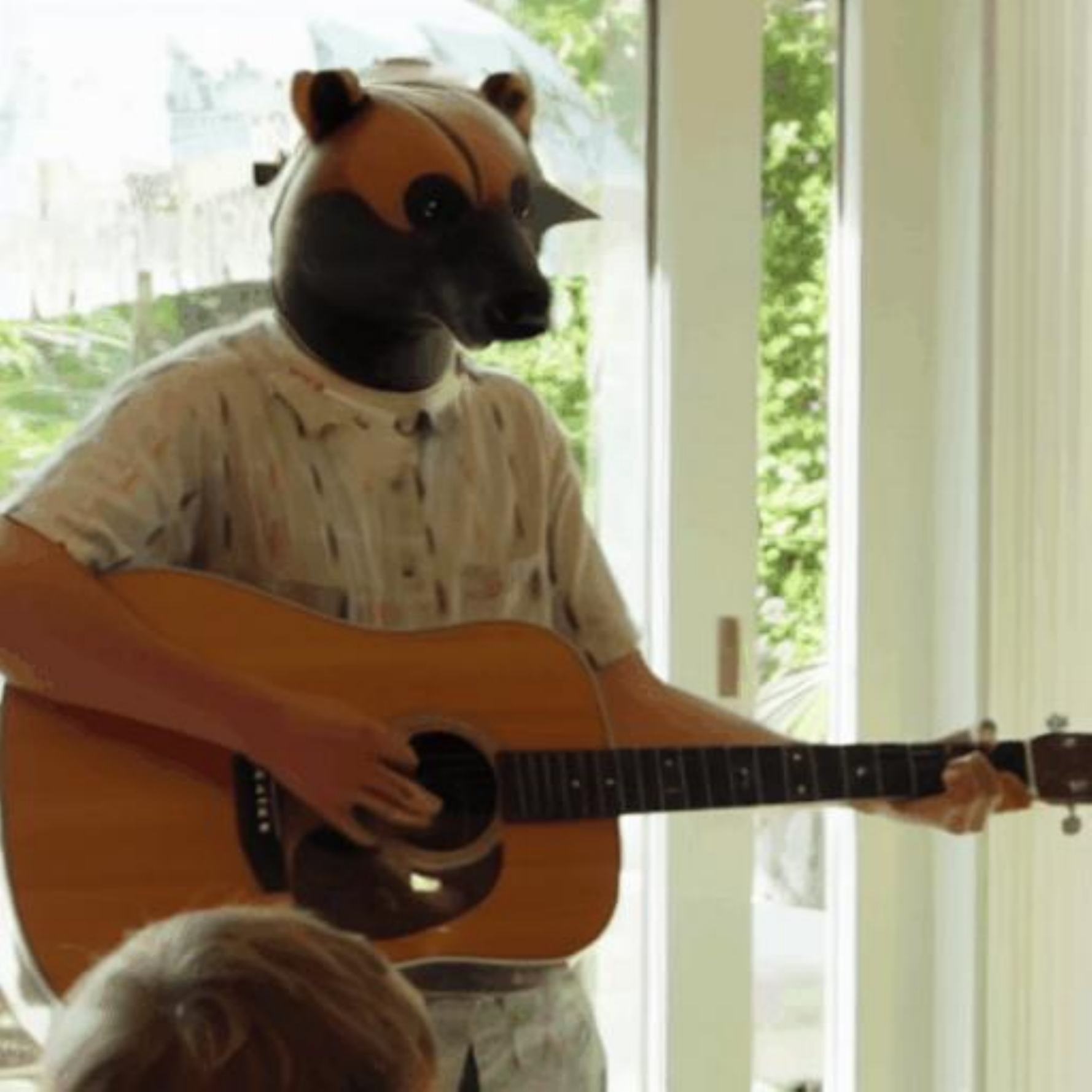}
\includegraphics[width=0.10\textwidth]{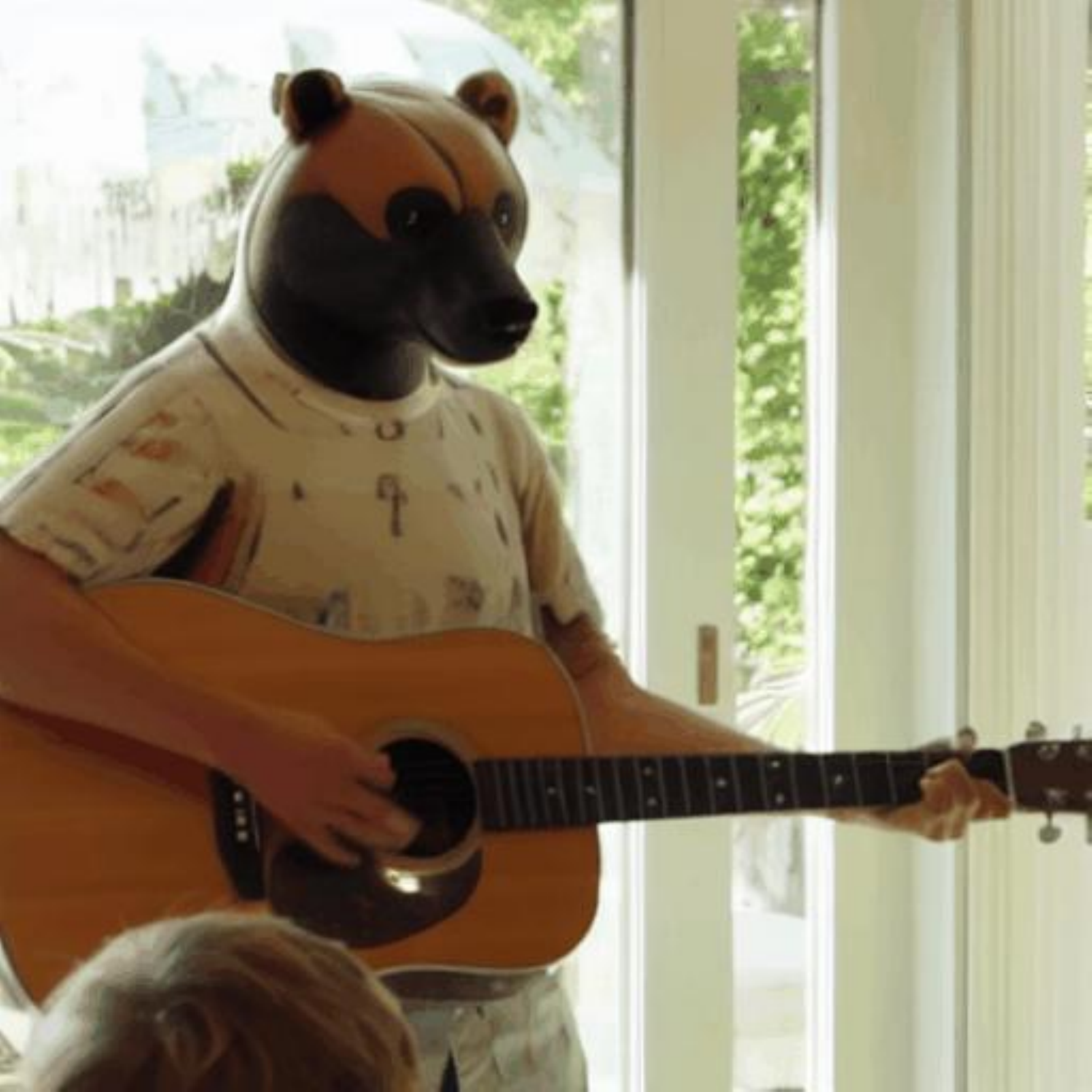}
\includegraphics[width=0.10\textwidth]{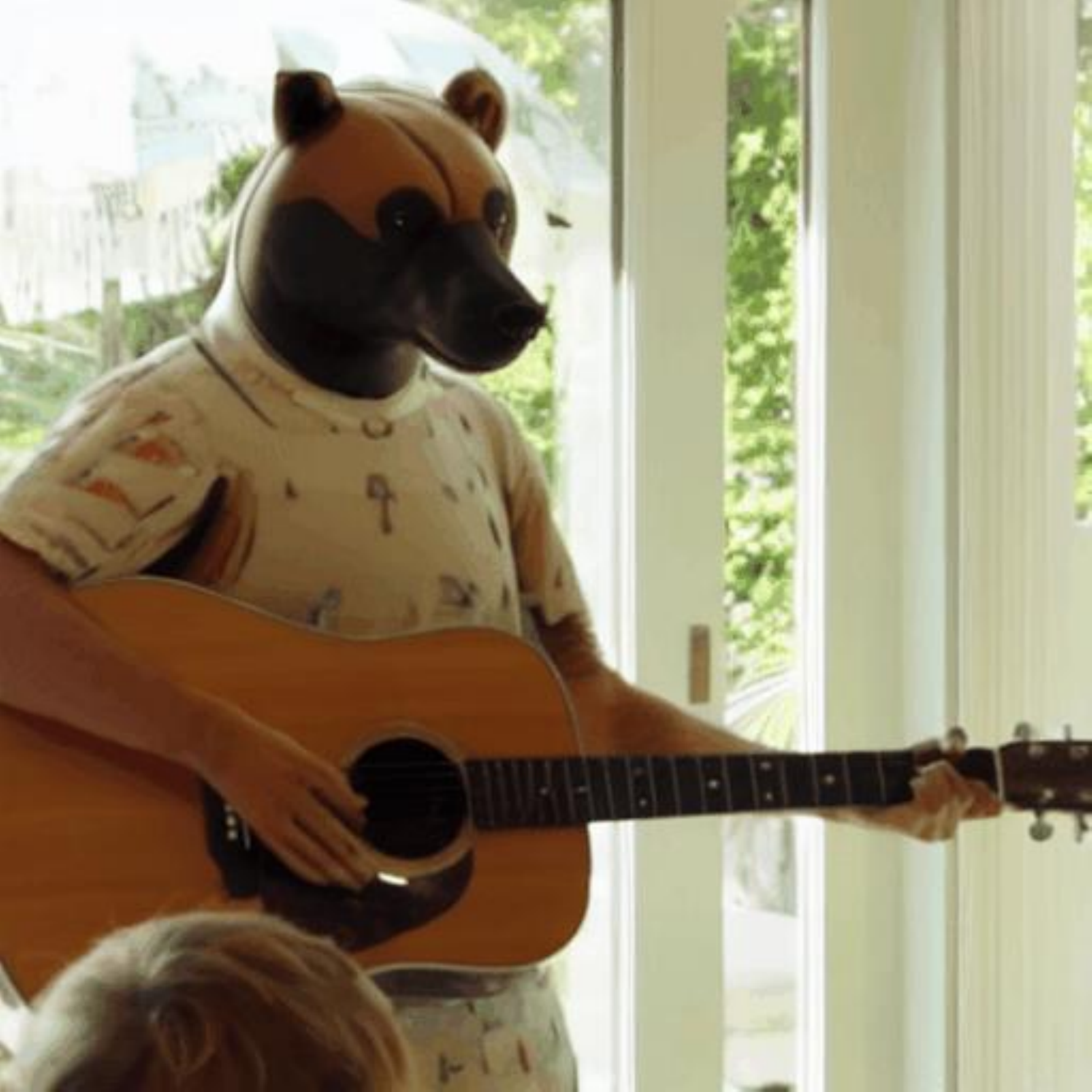}
\includegraphics[width=0.10\textwidth]{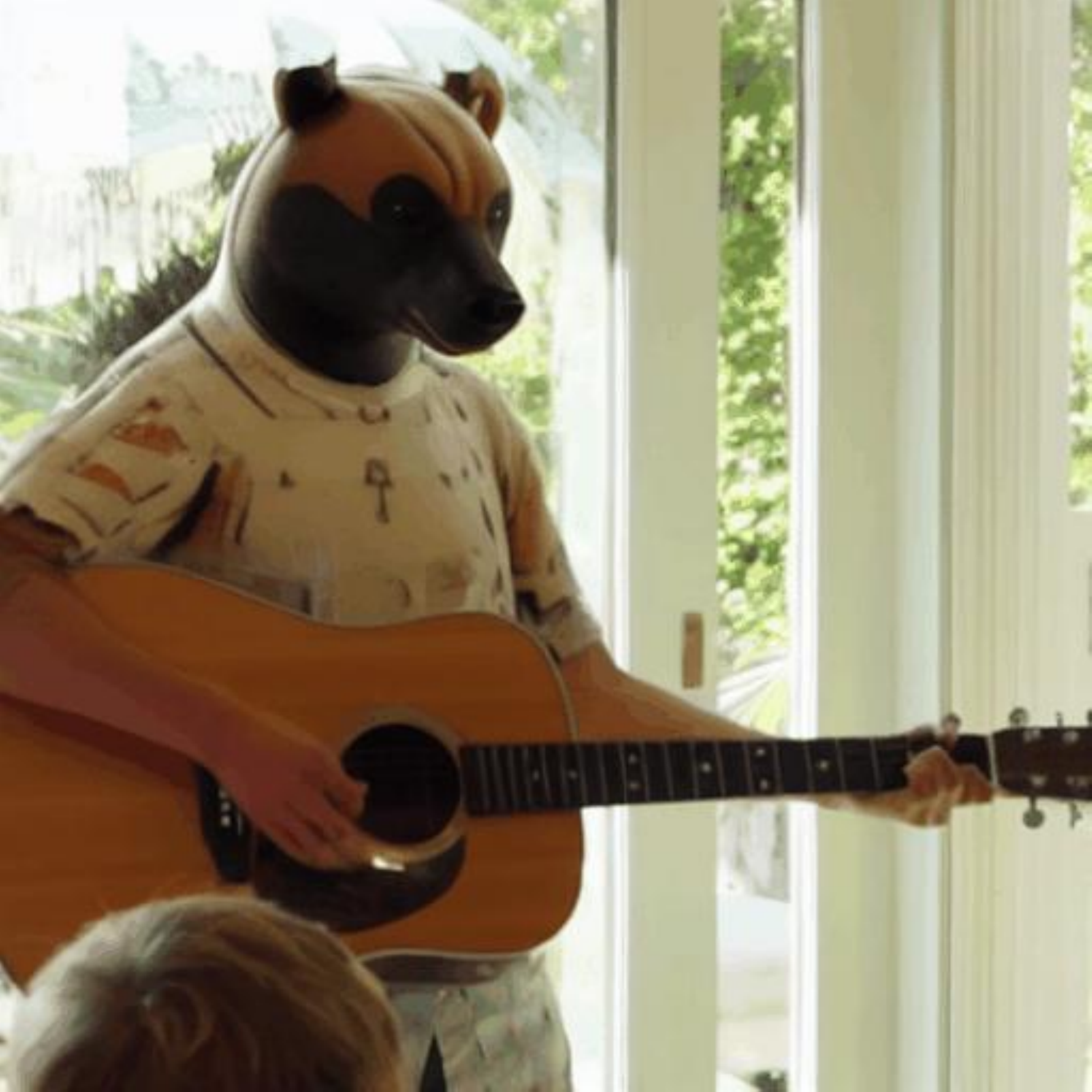}
\includegraphics[width=0.10\textwidth]{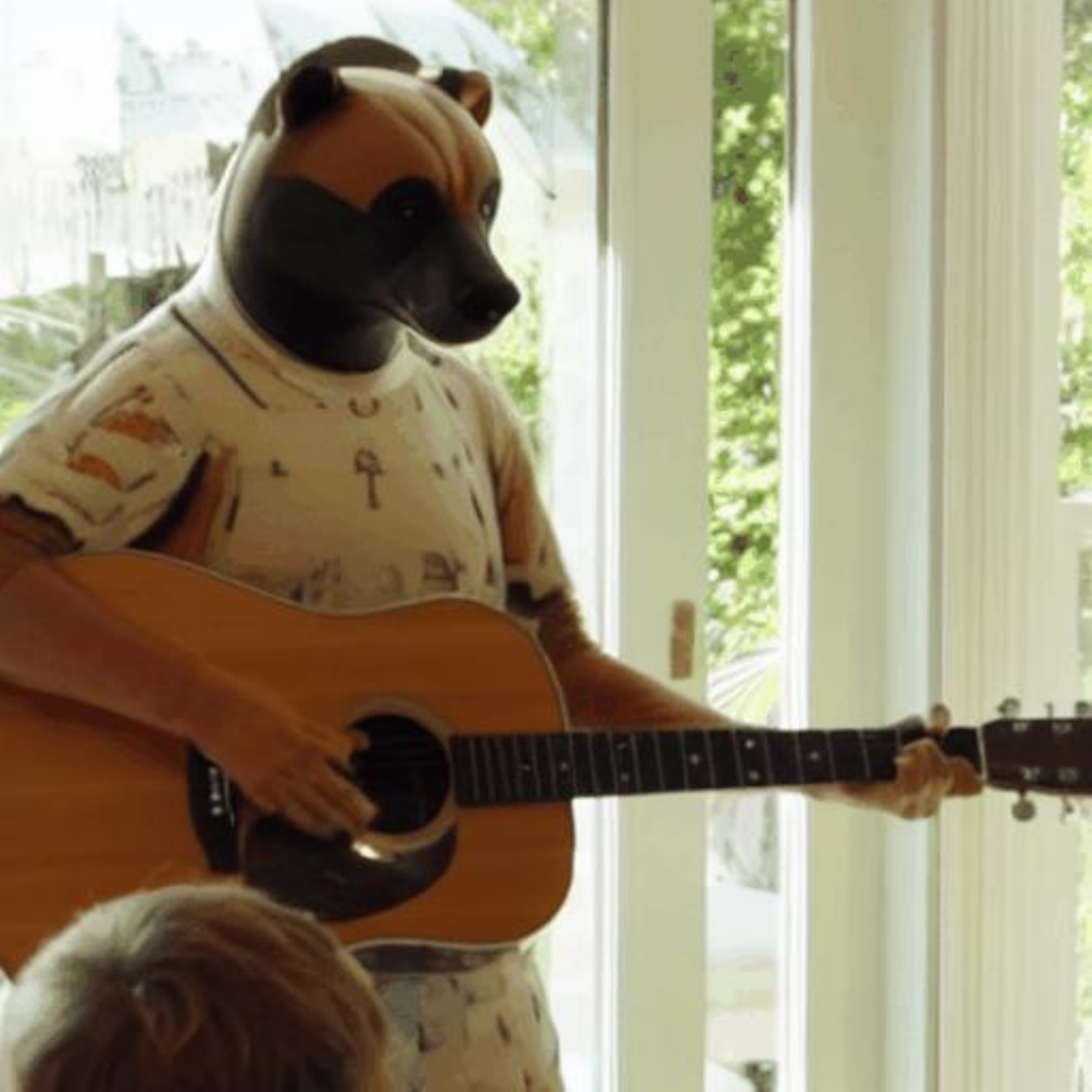}
\includegraphics[width=0.10\textwidth]{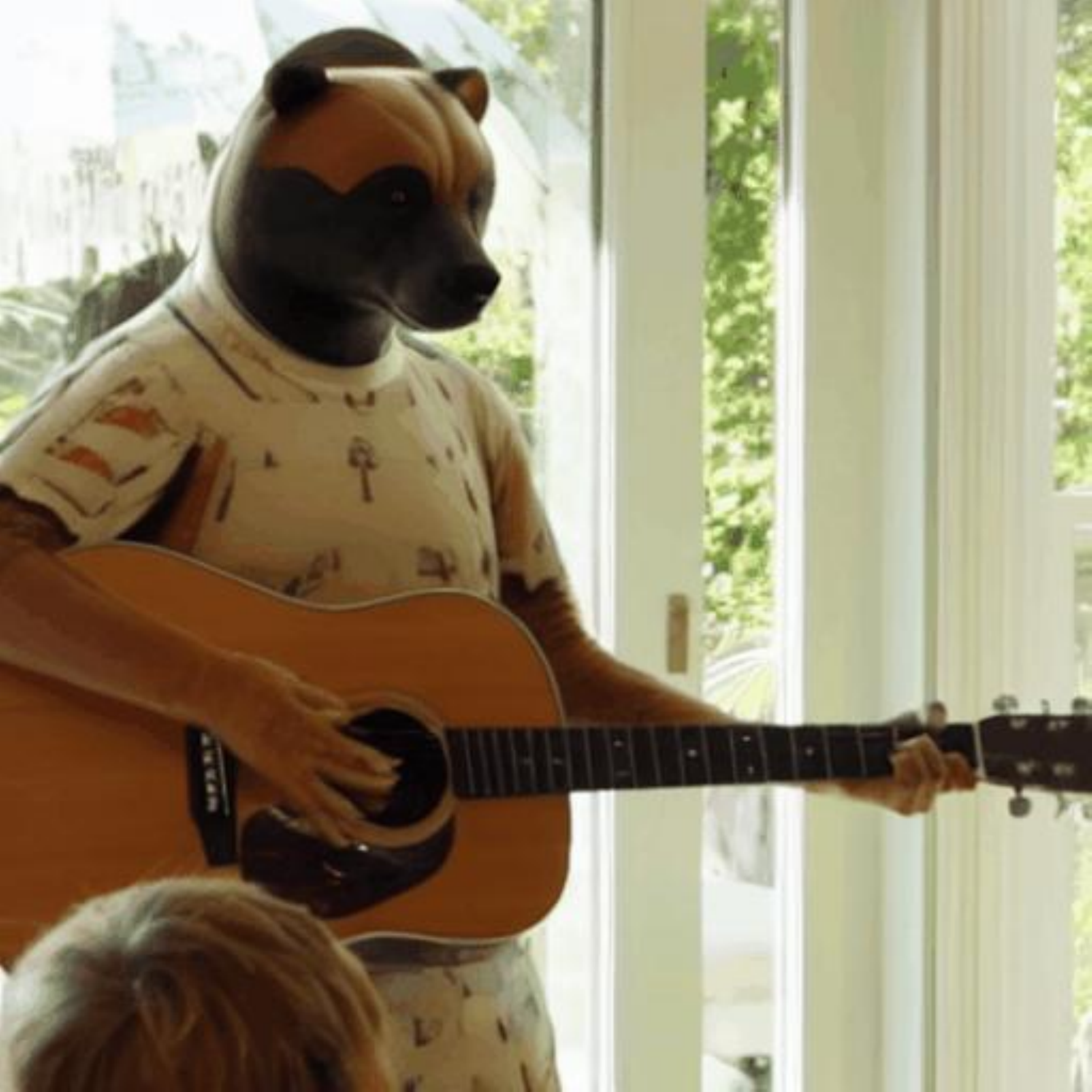}

\makebox[0.12\textwidth]{A \textcolor{blue}{\textbf{monkey}} is playing a guitar.}\\
\includegraphics[width=0.10\textwidth]{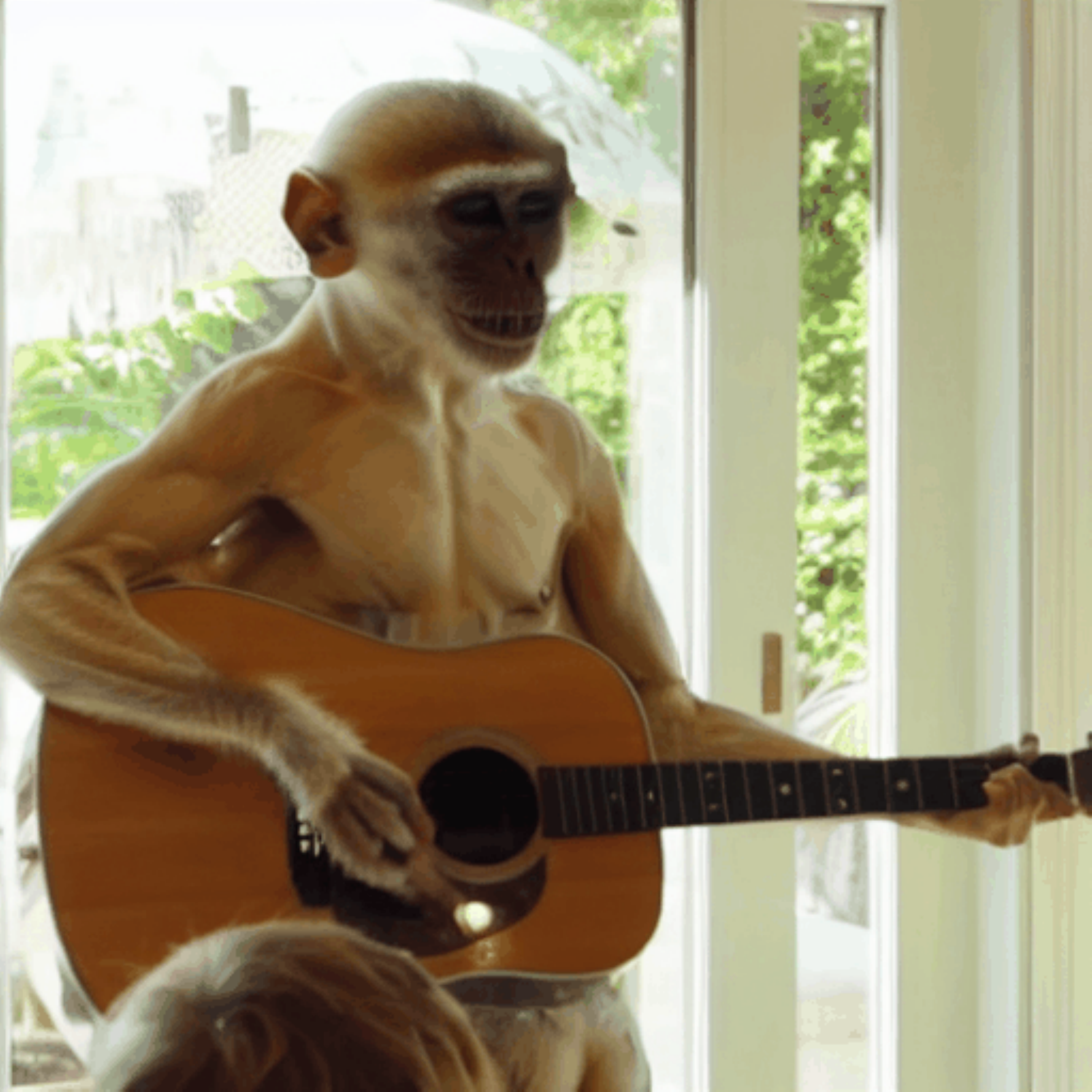}
\includegraphics[width=0.10\textwidth]{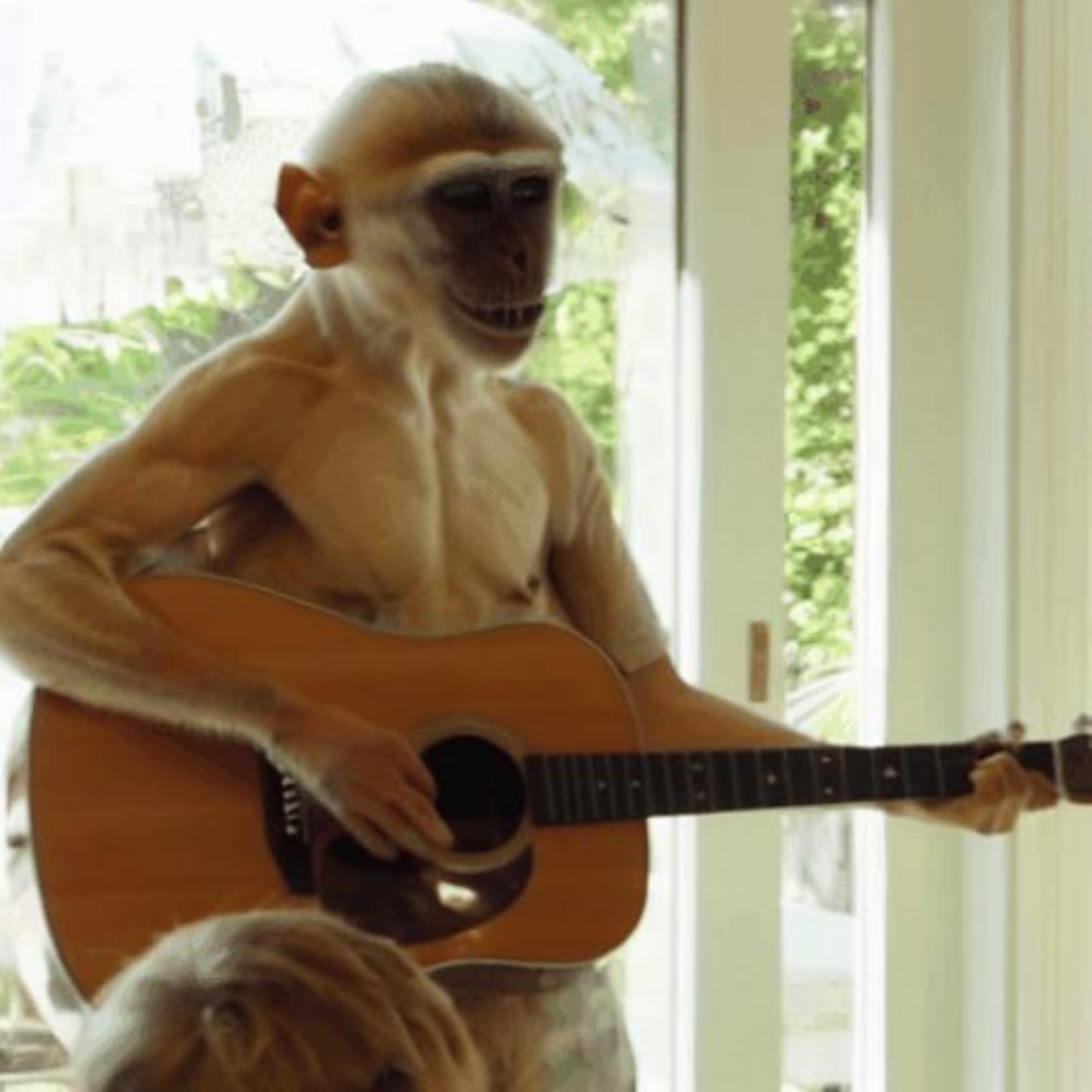}
\includegraphics[width=0.10\textwidth]{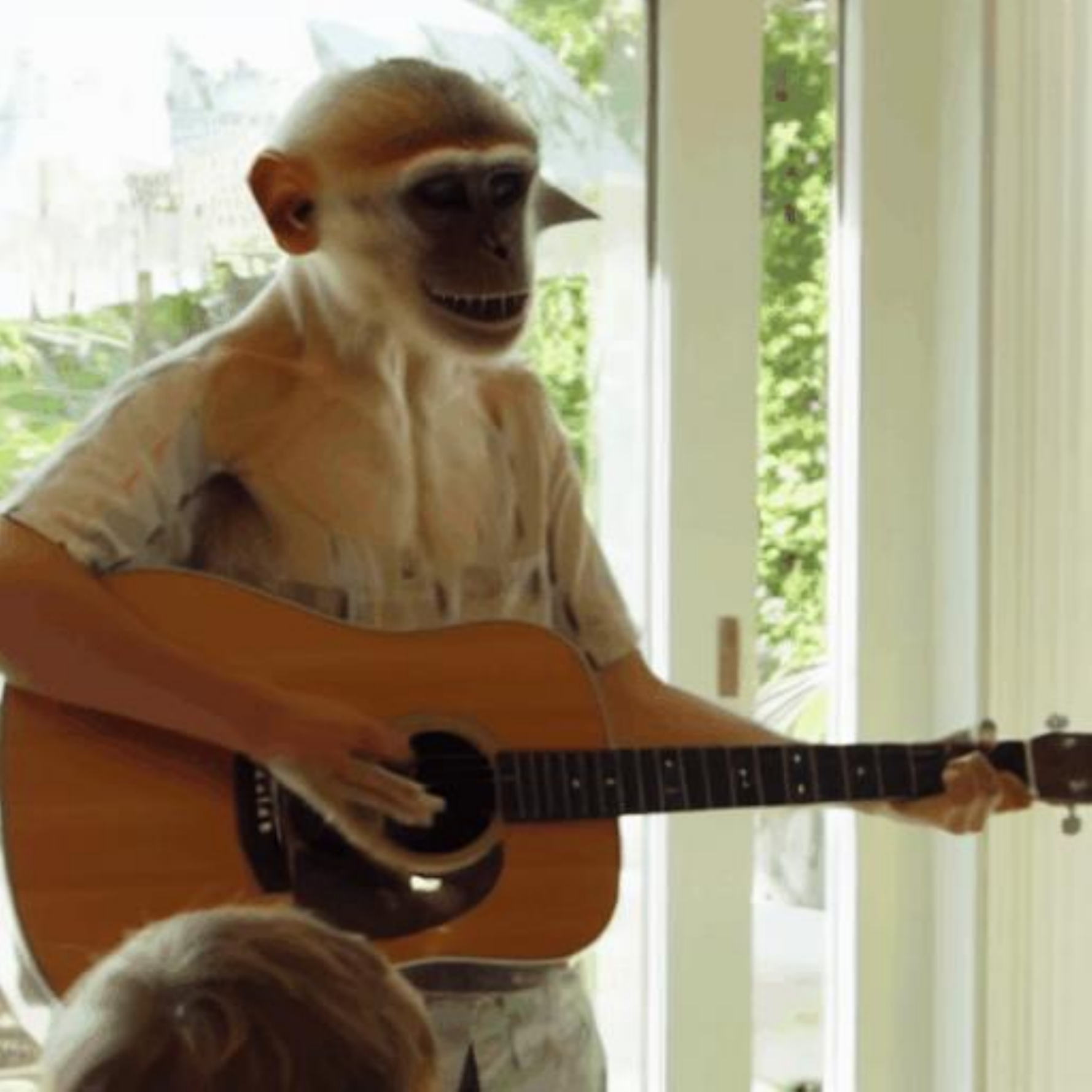}
\includegraphics[width=0.10\textwidth]{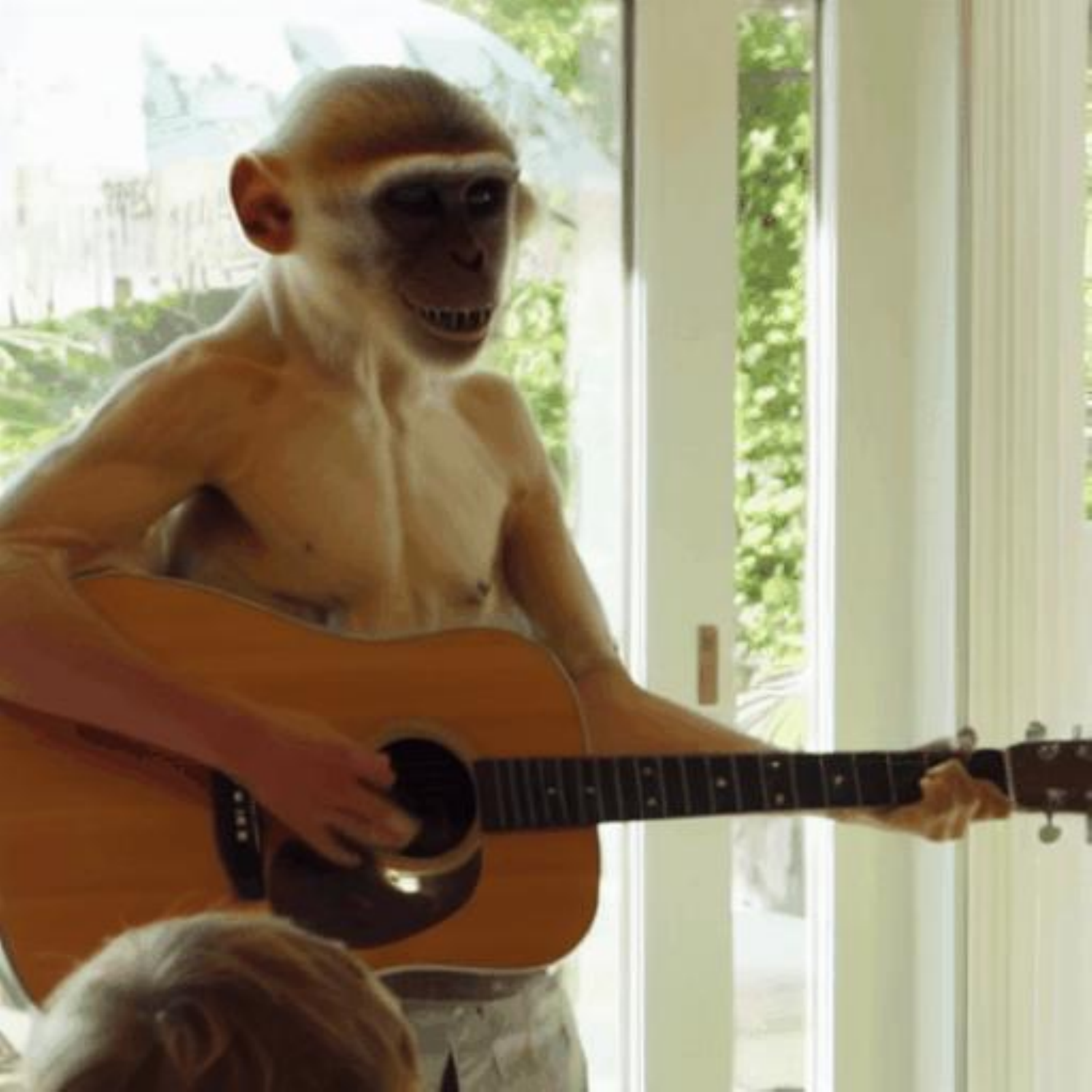}
\includegraphics[width=0.10\textwidth]{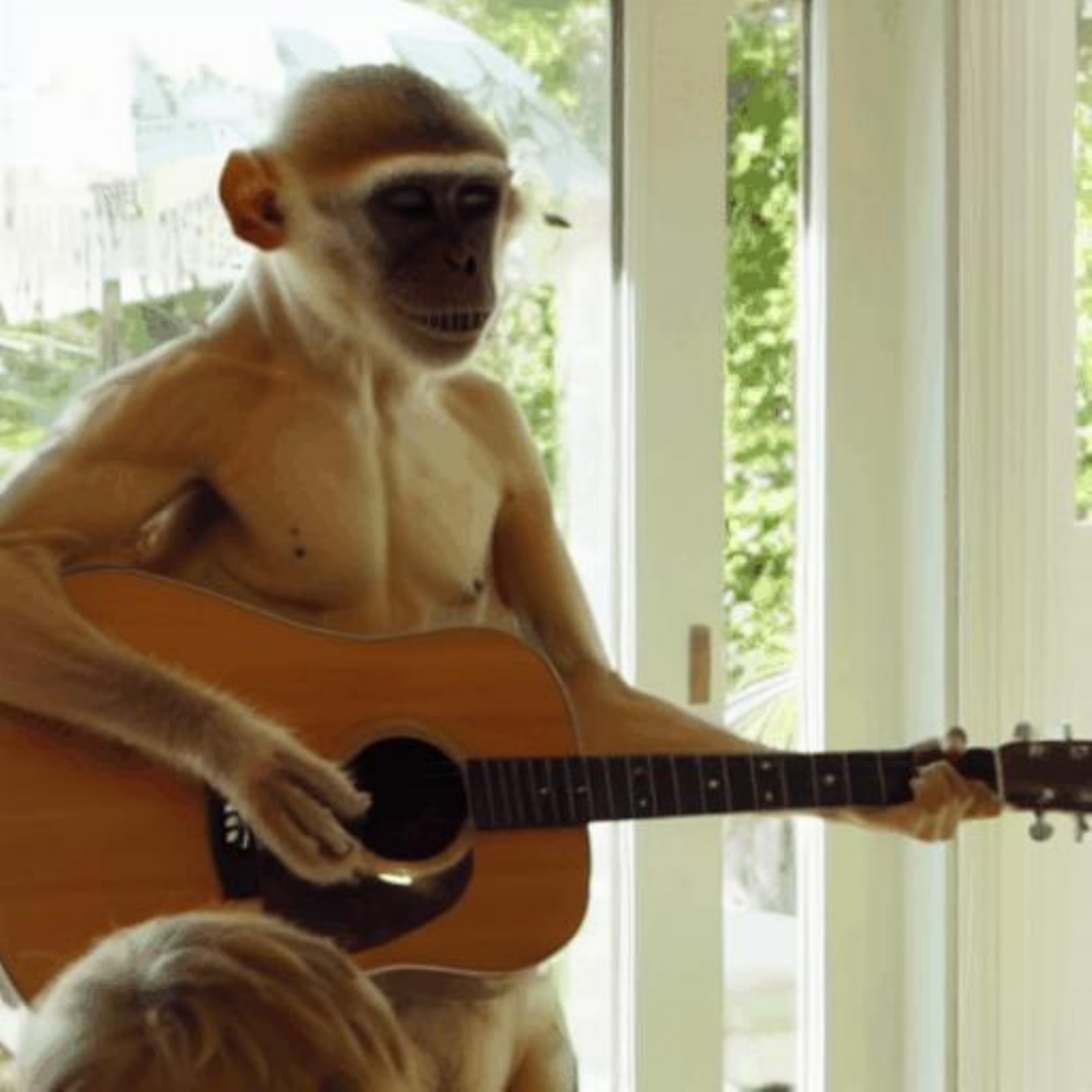}
\includegraphics[width=0.10\textwidth]{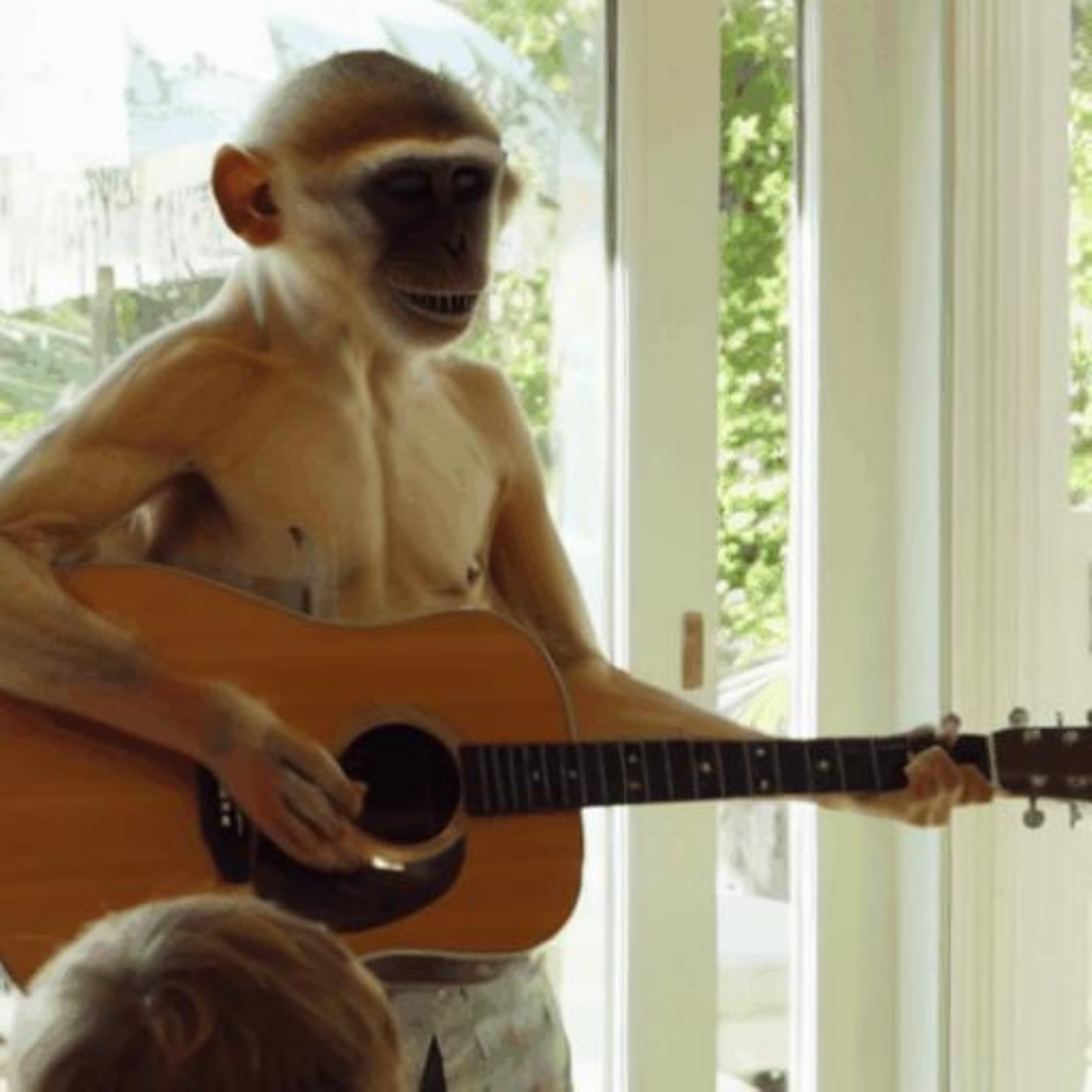}
\includegraphics[width=0.10\textwidth]{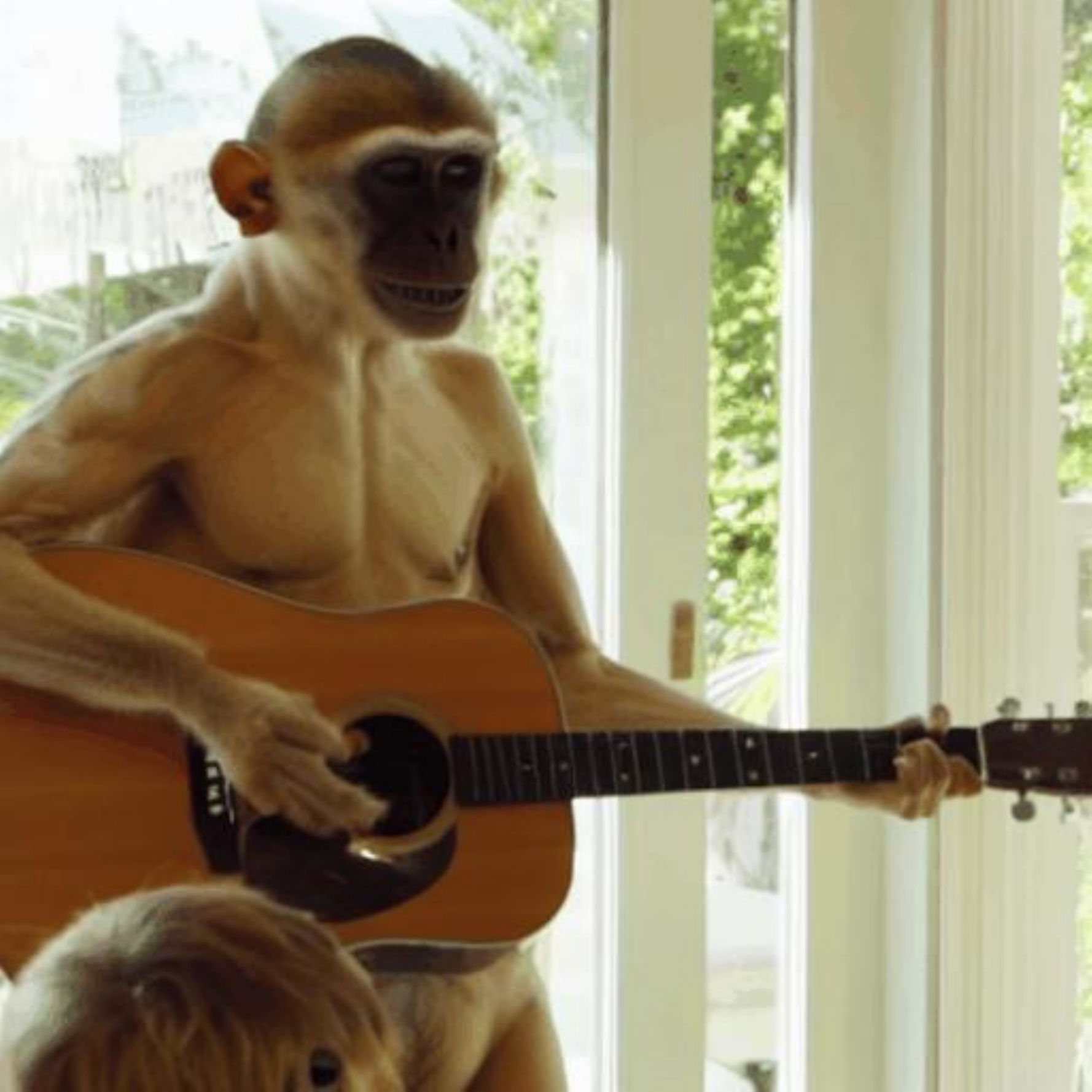}
\includegraphics[width=0.10\textwidth]{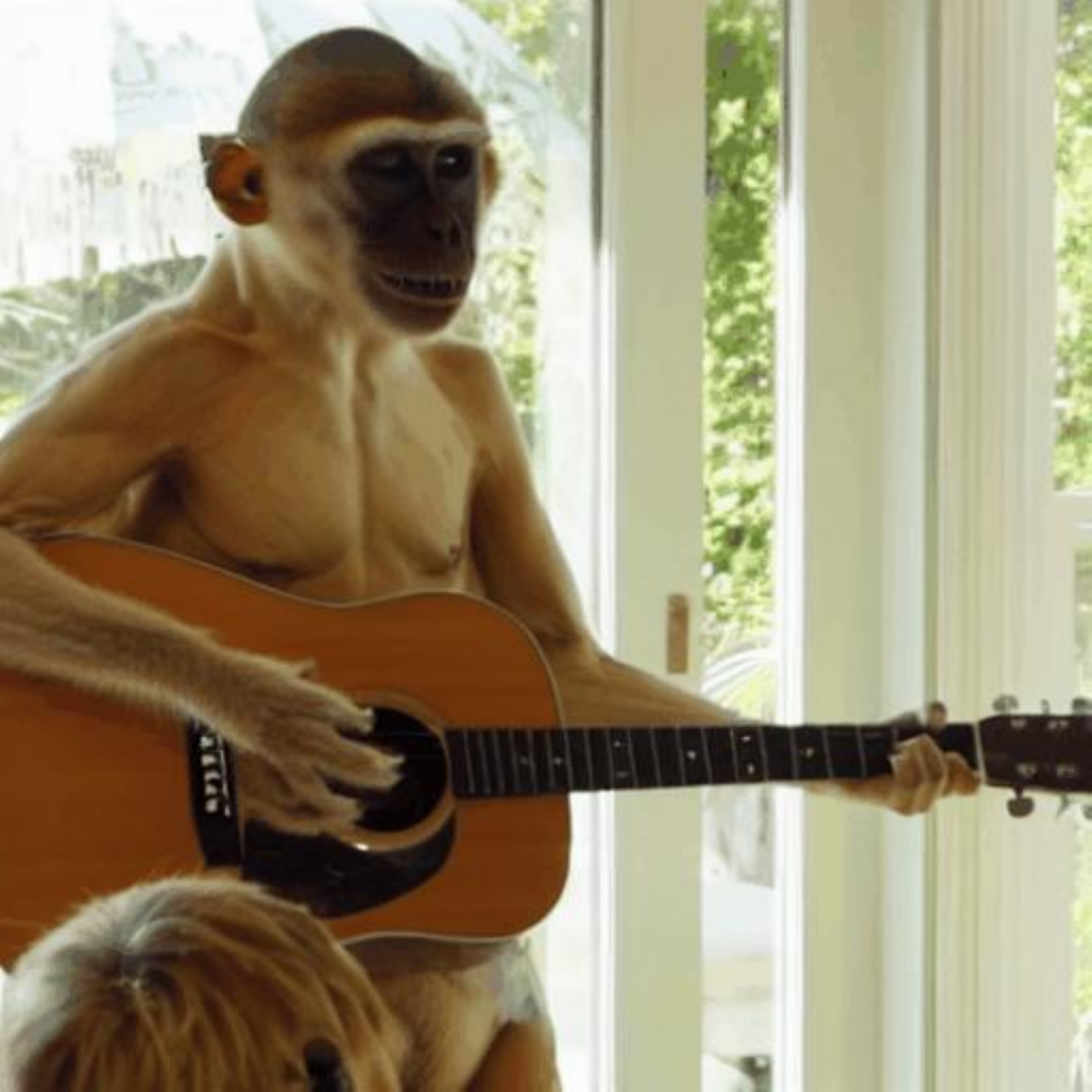}

\makebox[0.12\textwidth]{A man is playing a guitar, \textcolor{blue}{\textbf{cartoon style}}.}\\
\includegraphics[width=0.10\textwidth]{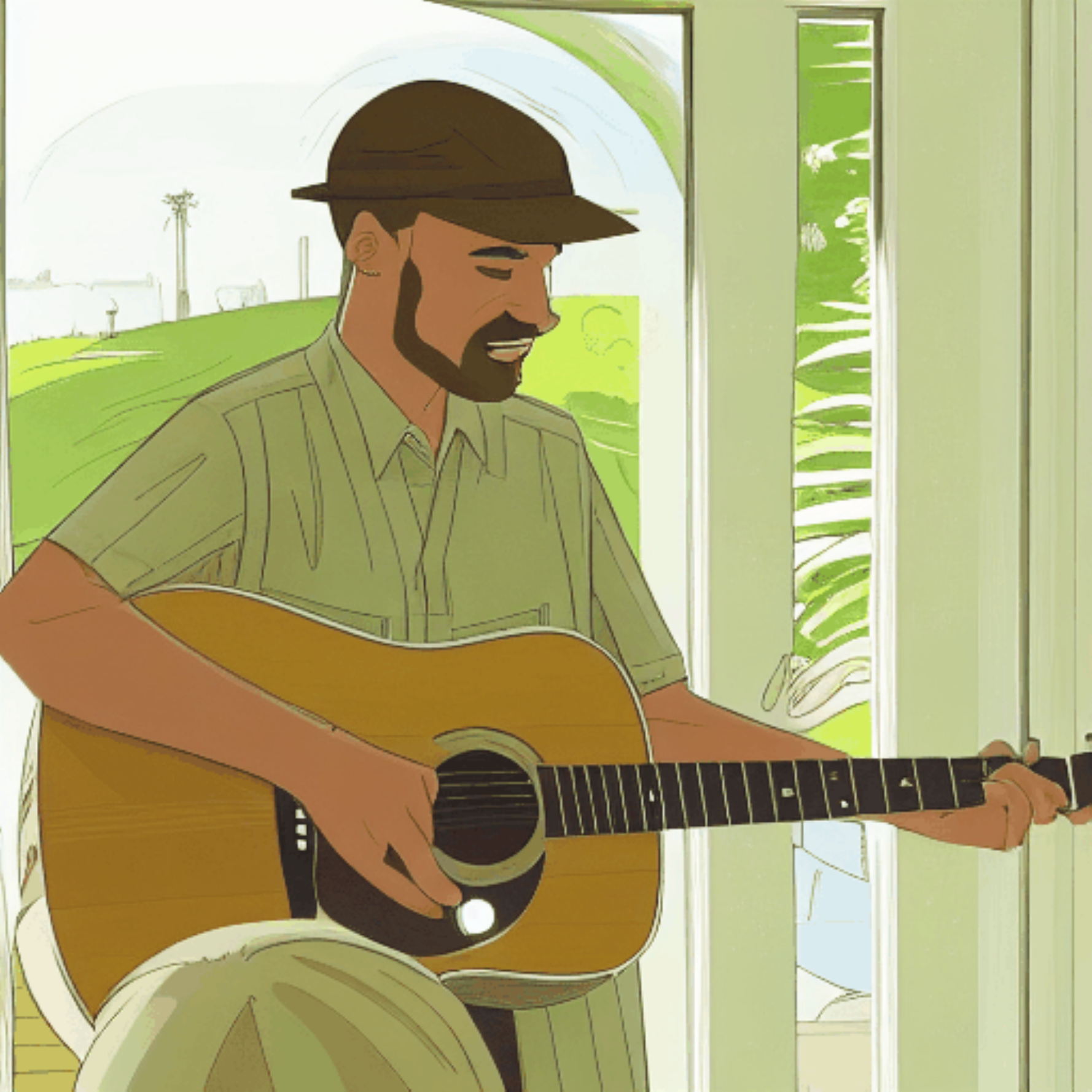}
\includegraphics[width=0.10\textwidth]{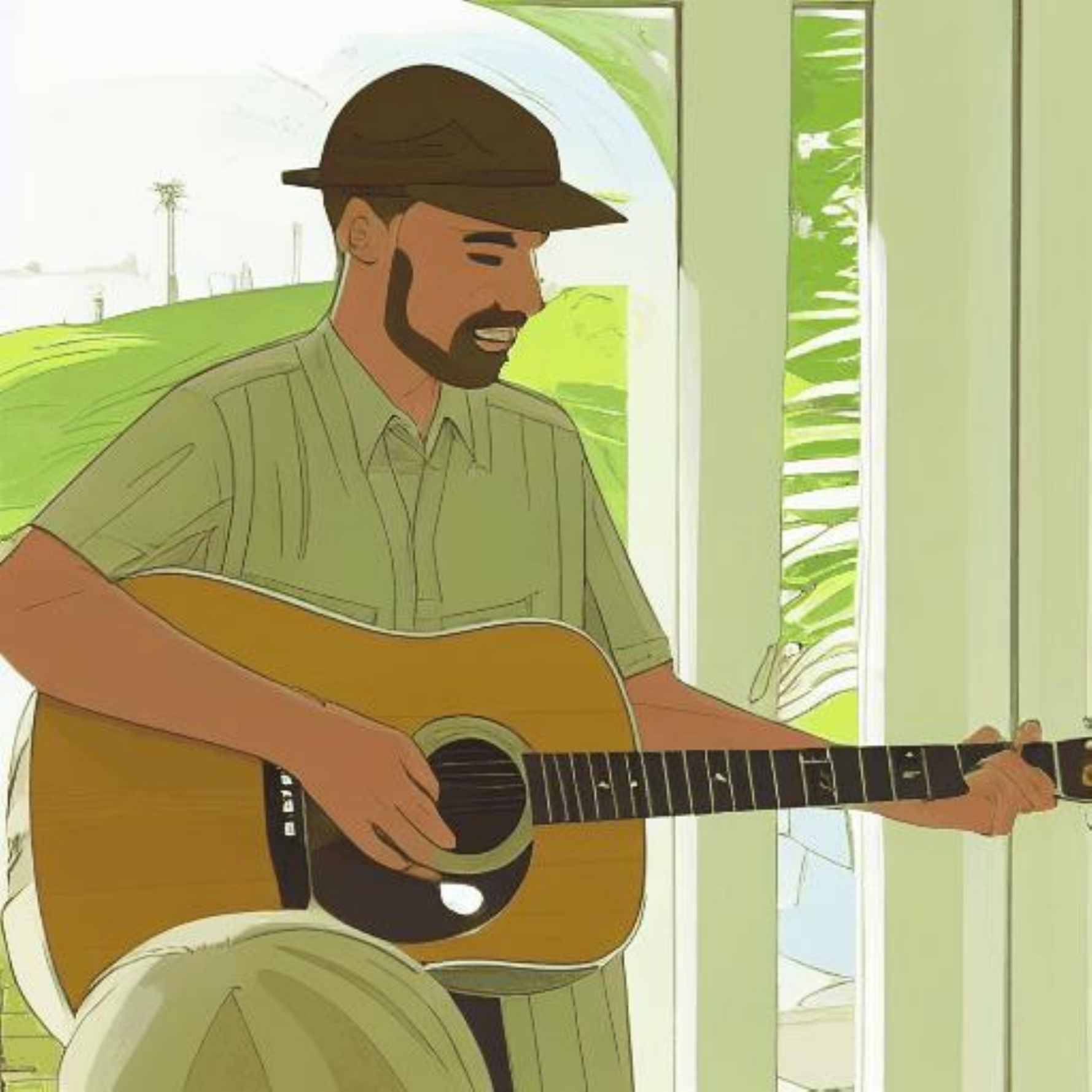}
\includegraphics[width=0.10\textwidth]{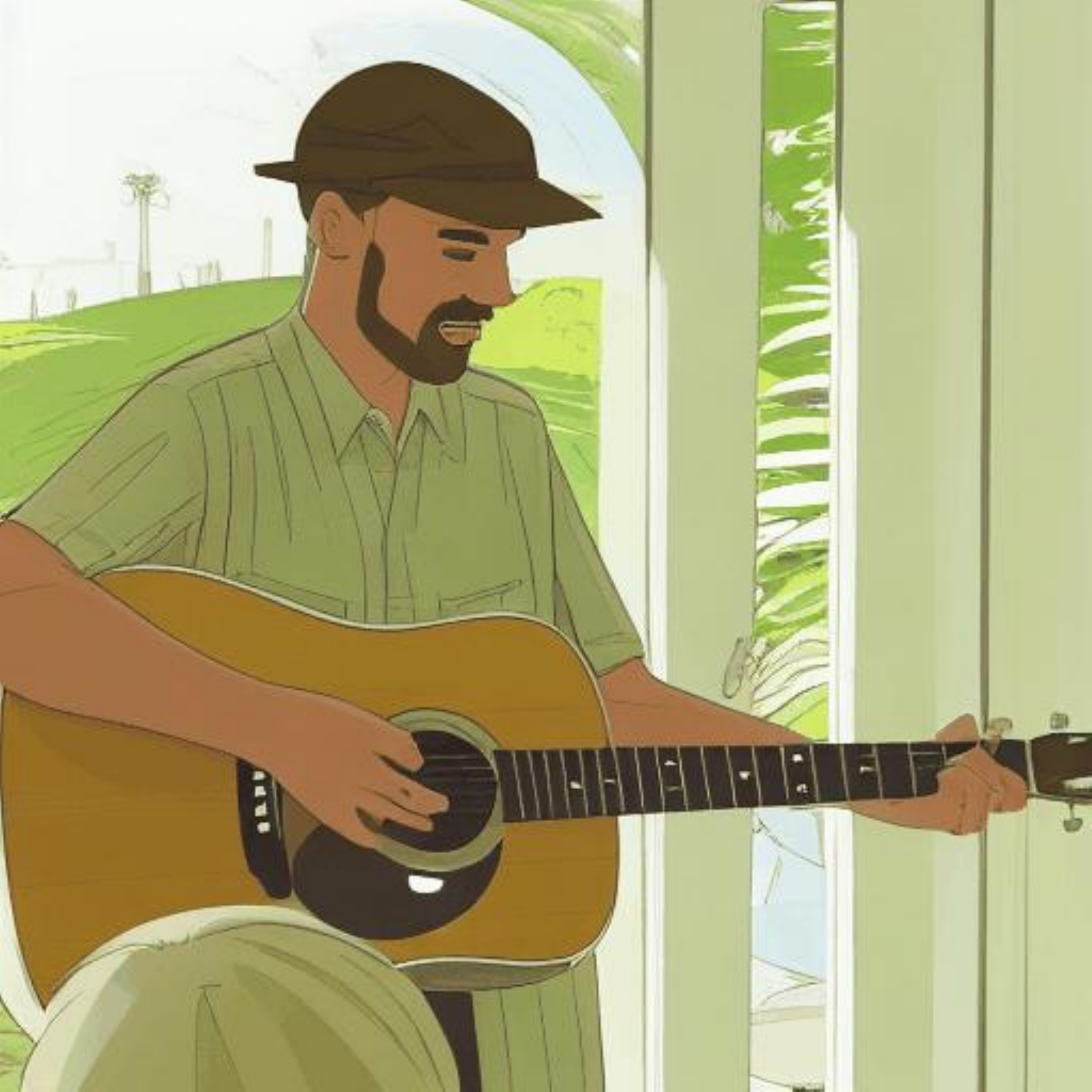}
\includegraphics[width=0.10\textwidth]{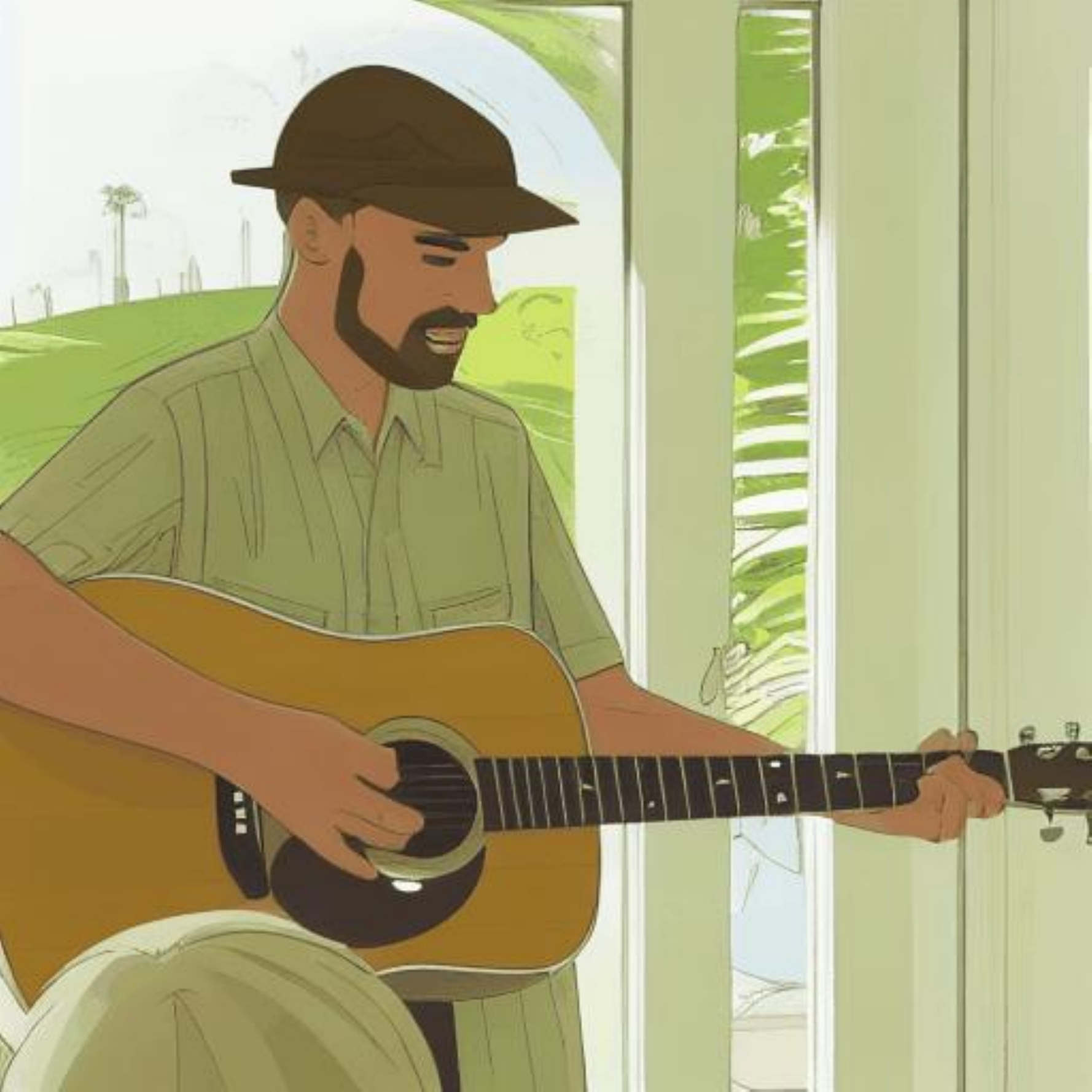}
\includegraphics[width=0.10\textwidth]{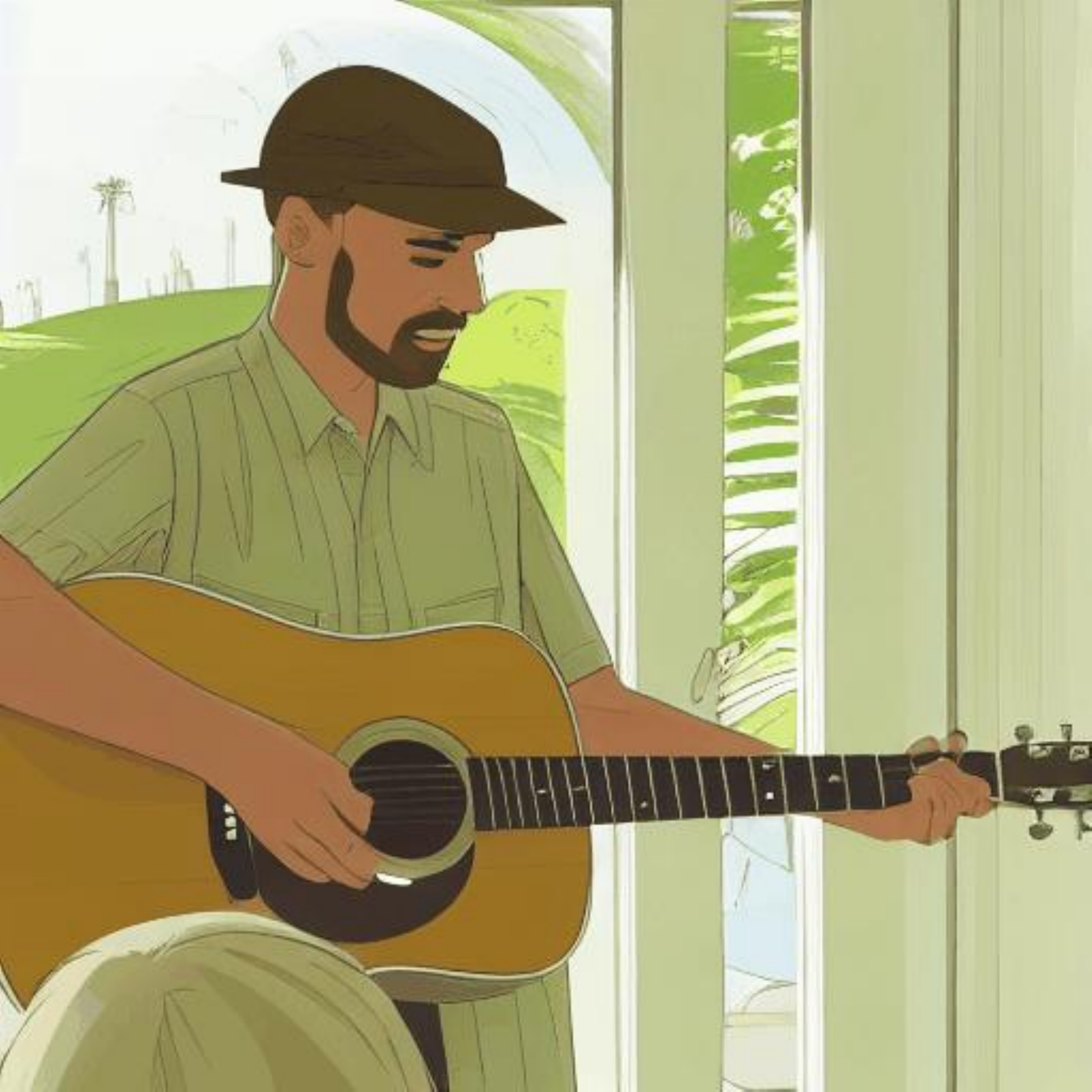}
\includegraphics[width=0.10\textwidth]{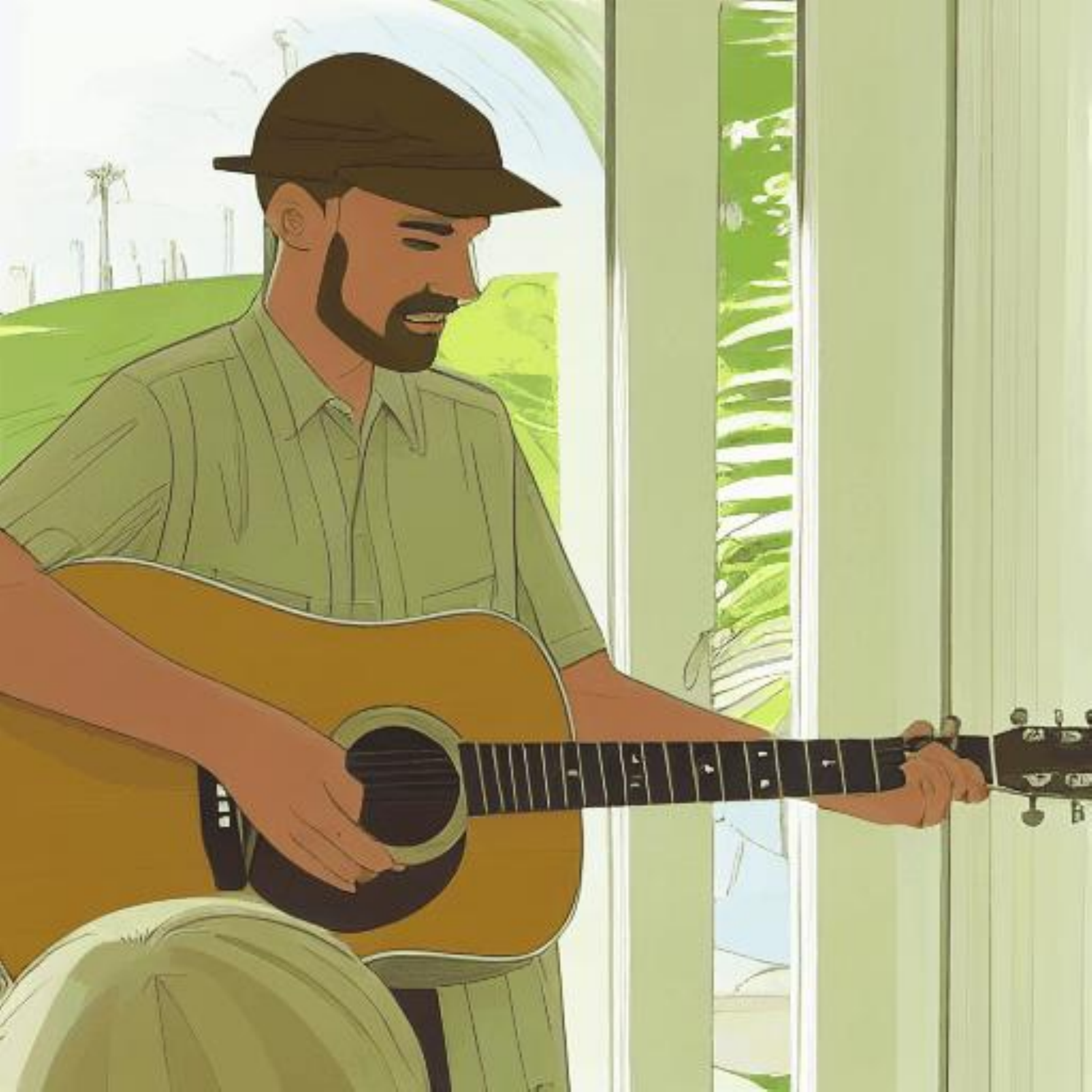}
\includegraphics[width=0.10\textwidth]{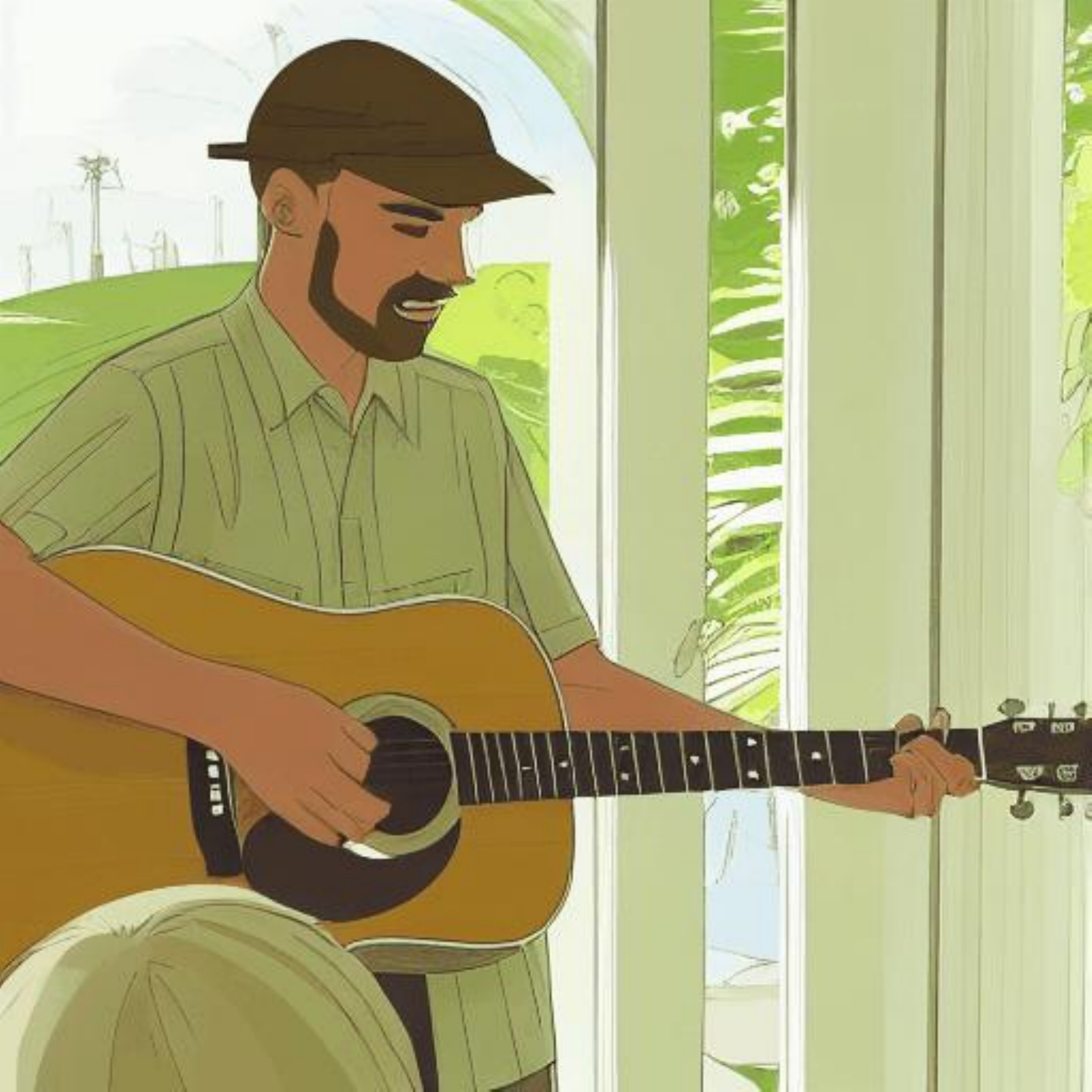}
\includegraphics[width=0.10\textwidth]{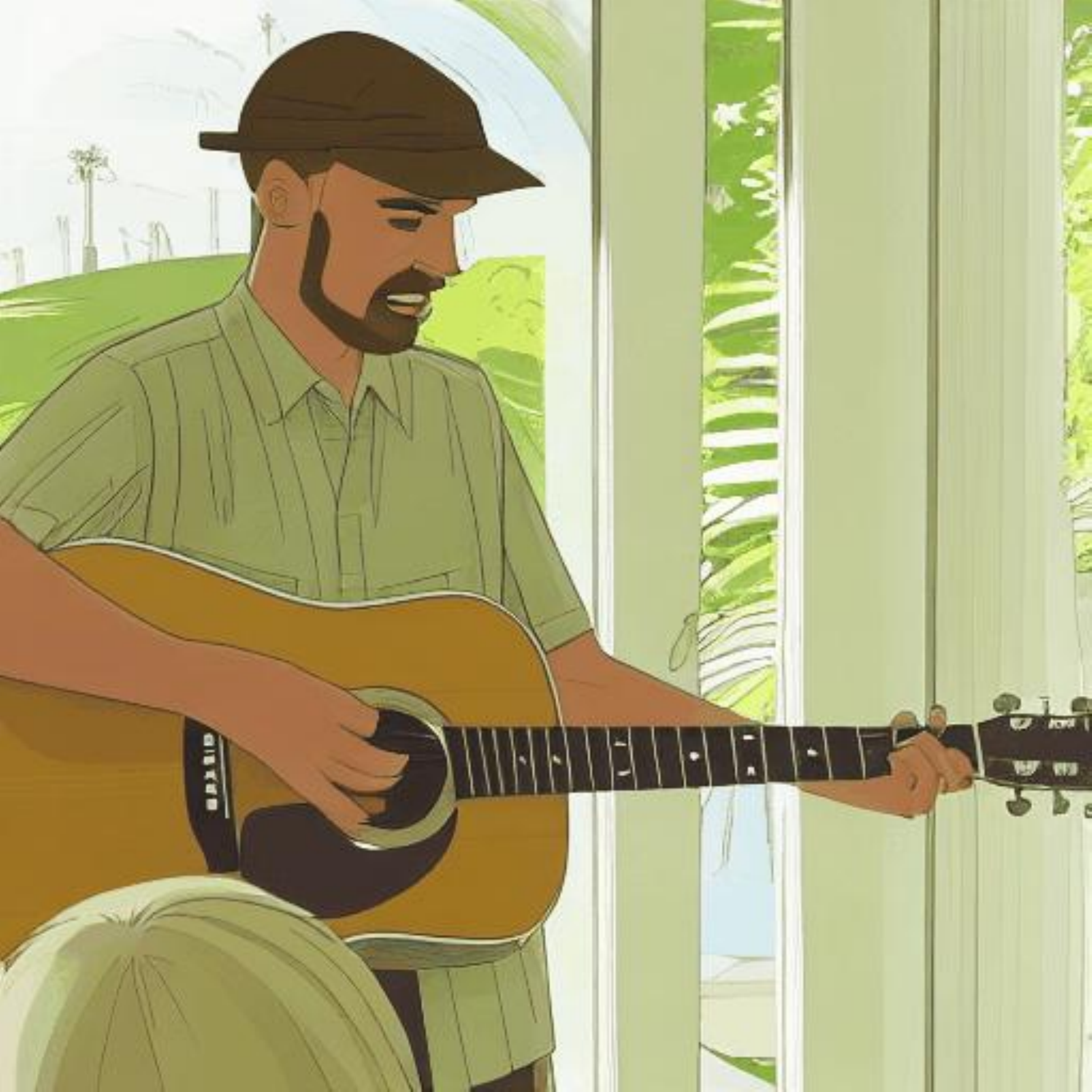}

\makebox[0.12\textwidth]{A man is playing a guitar, \textcolor{blue}{\textbf{Matisse style}}.}\\
\includegraphics[width=0.10\textwidth]{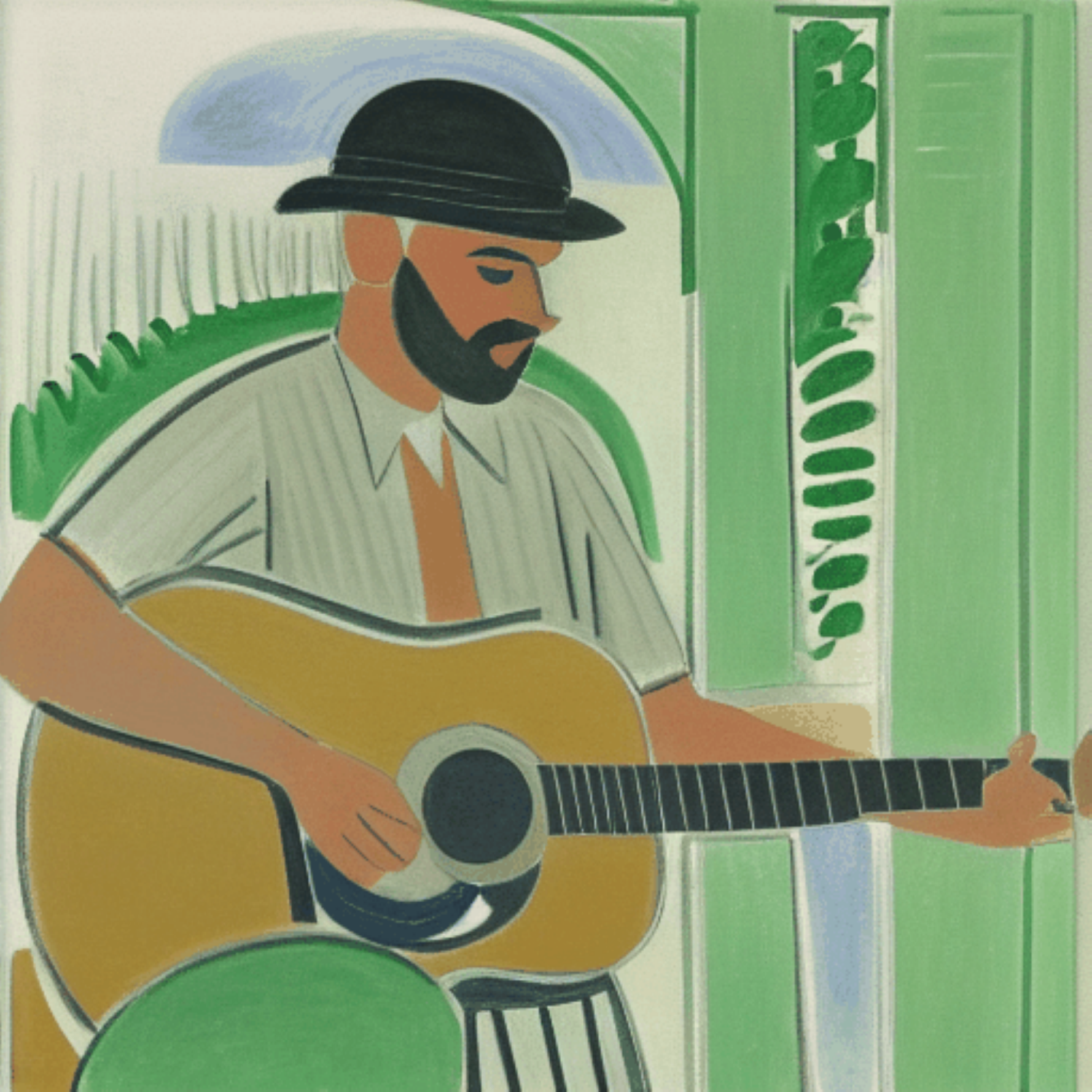}
\includegraphics[width=0.10\textwidth]{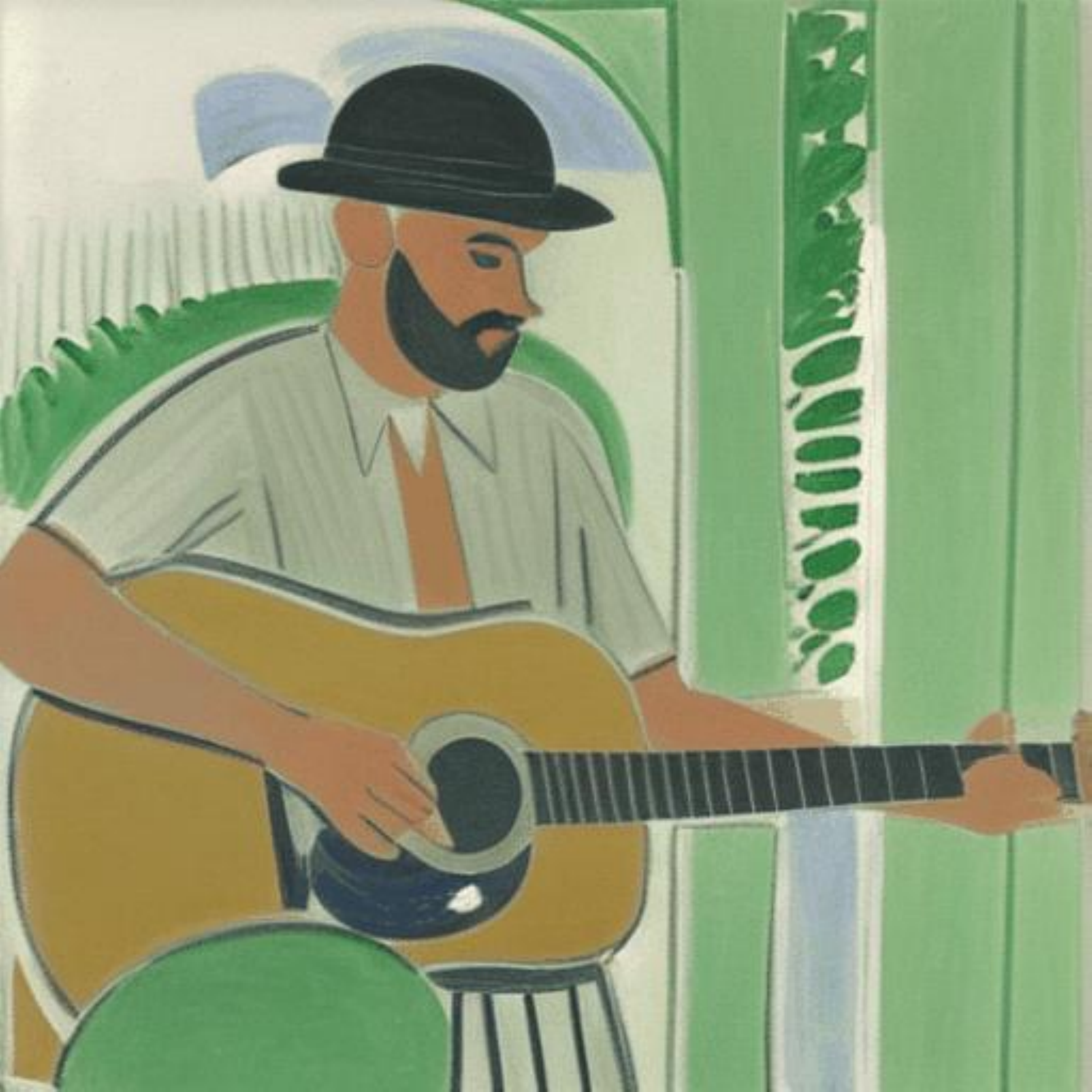}
\includegraphics[width=0.10\textwidth]{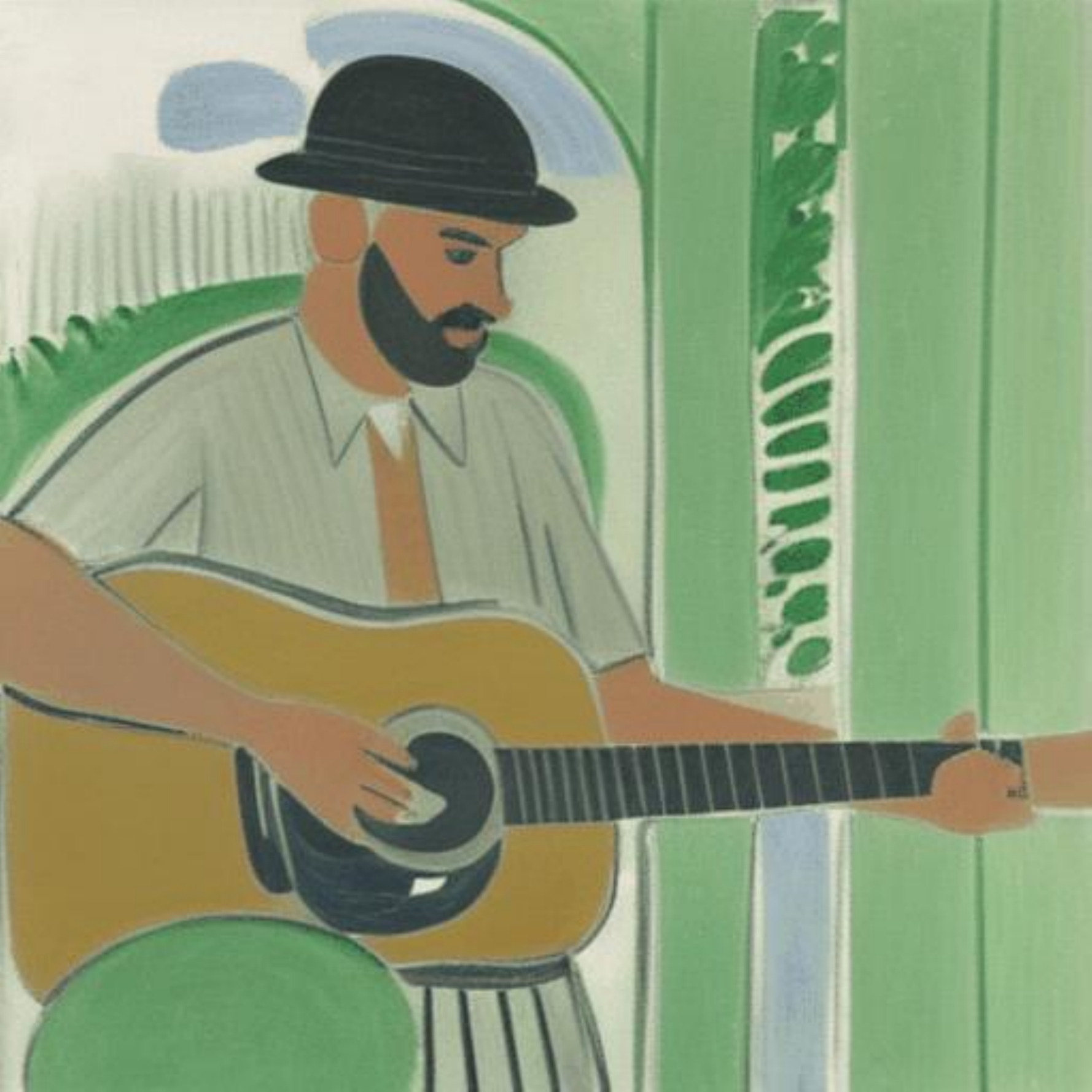}
\includegraphics[width=0.10\textwidth]{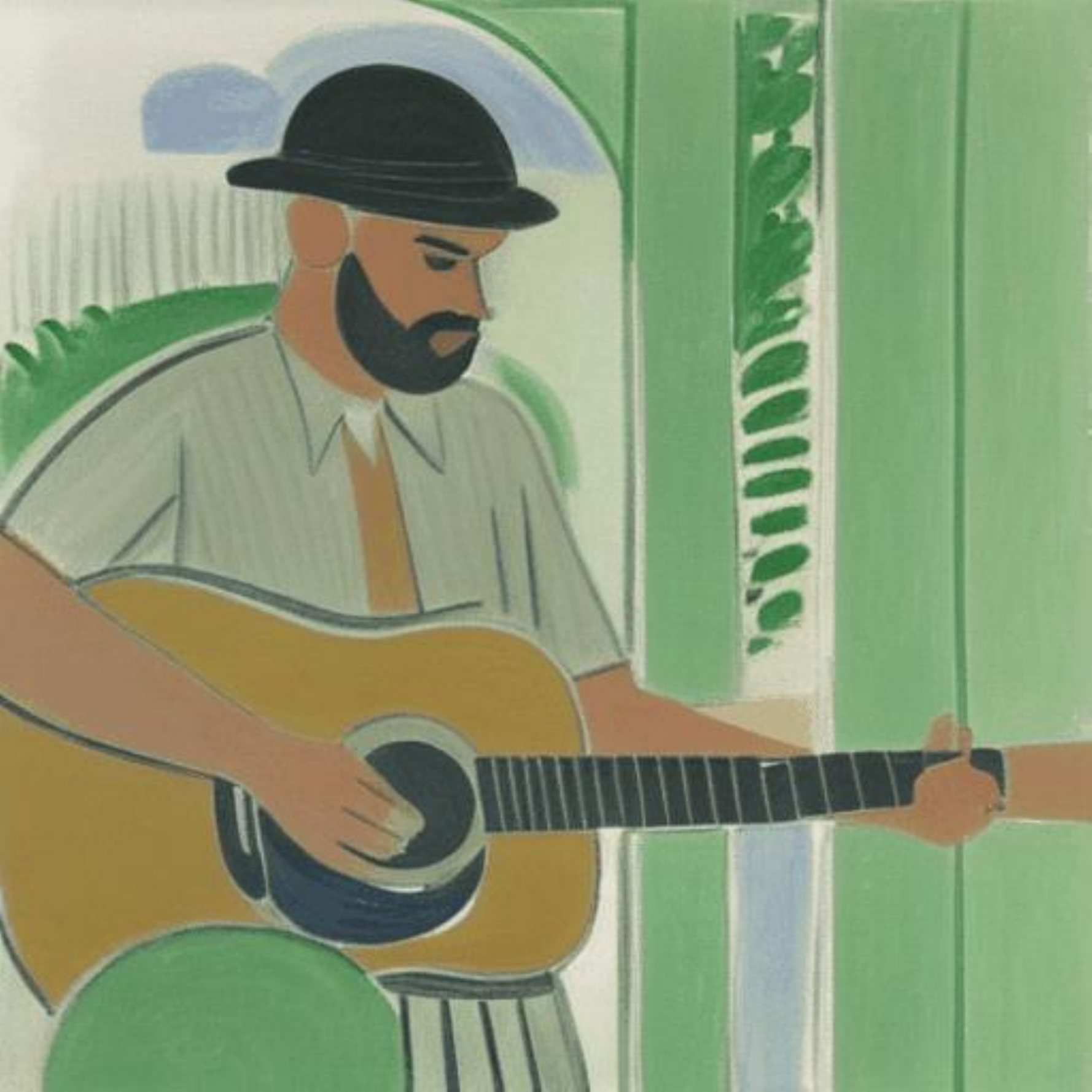}
\includegraphics[width=0.10\textwidth]{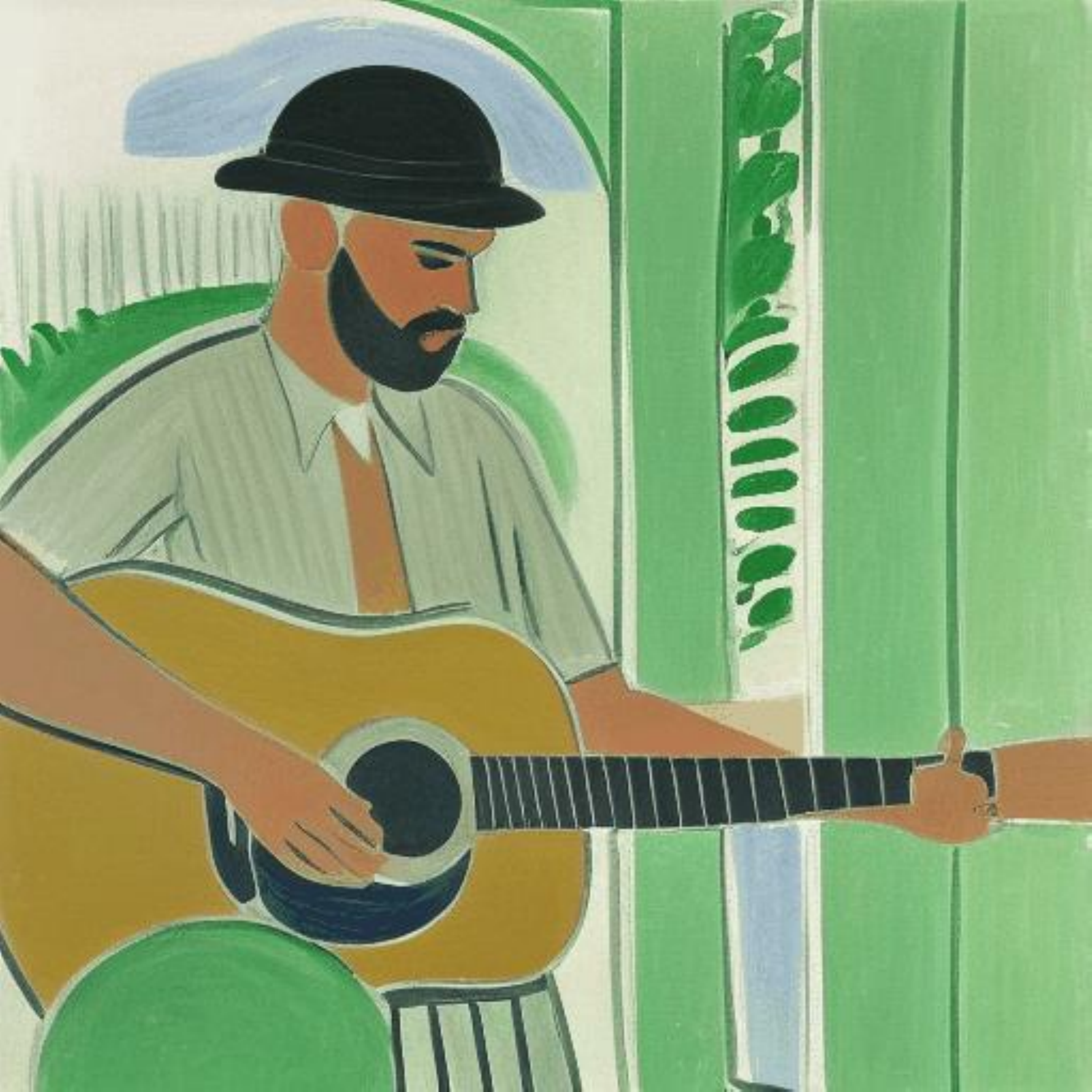}
\includegraphics[width=0.10\textwidth]{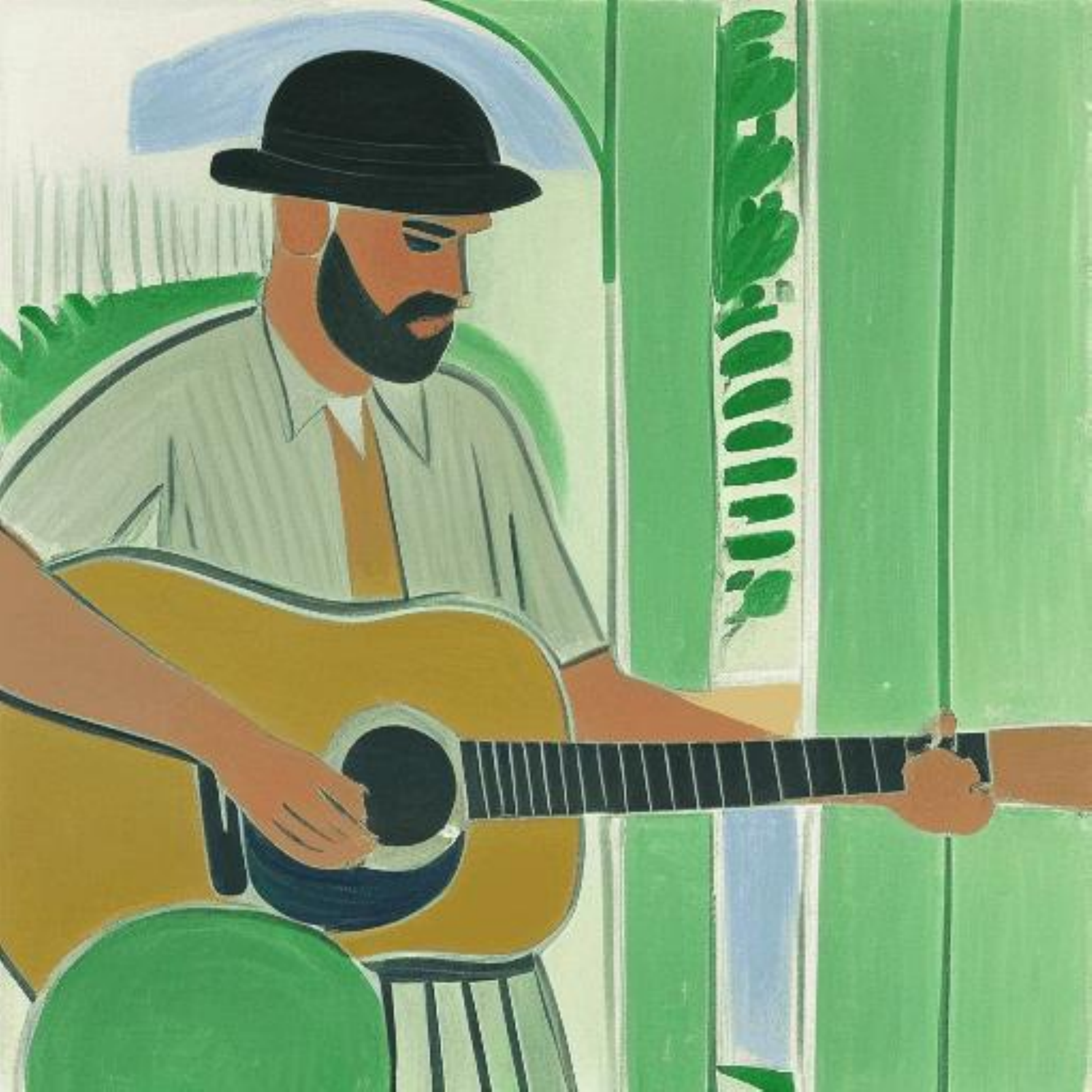}
\includegraphics[width=0.10\textwidth]{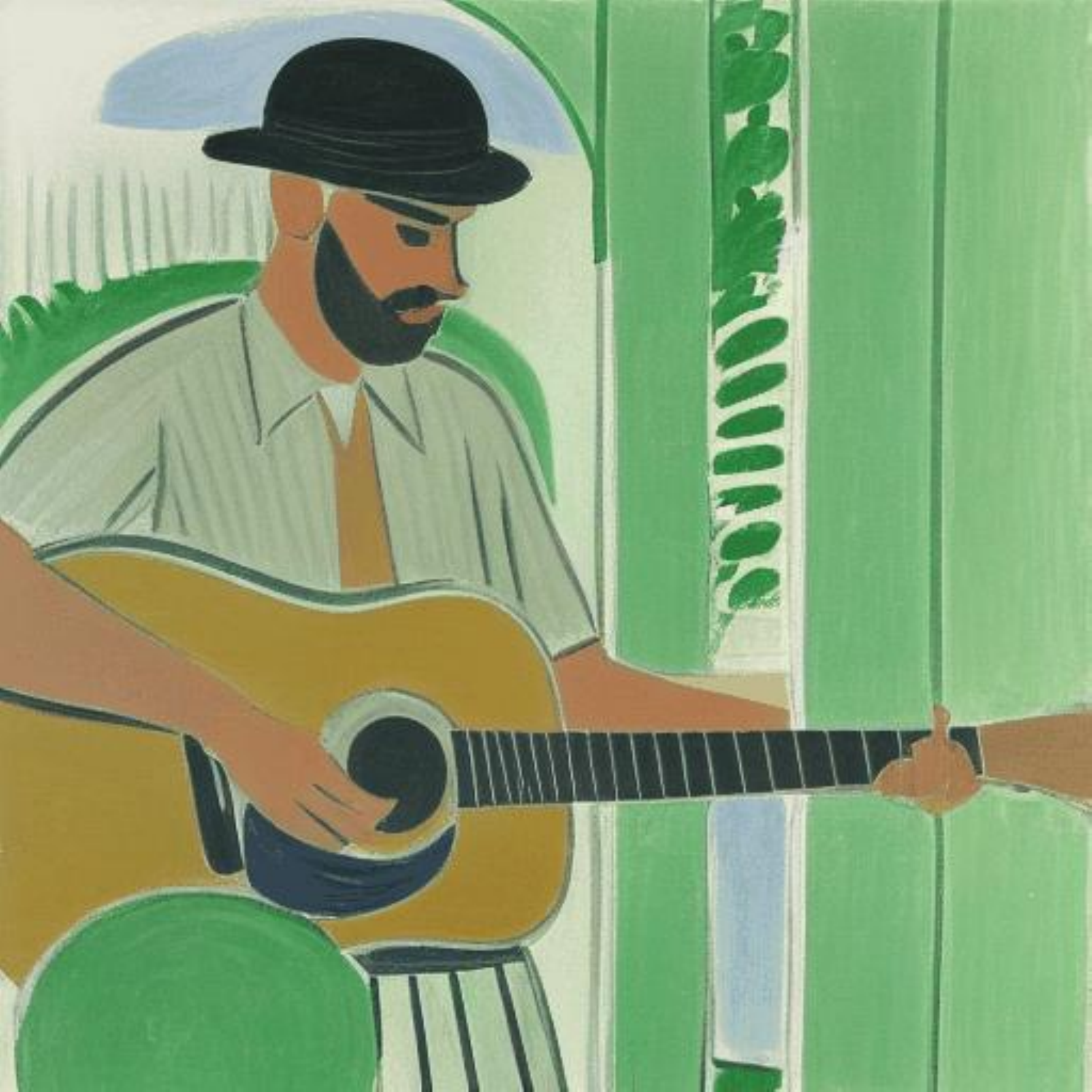}
\includegraphics[width=0.10\textwidth]{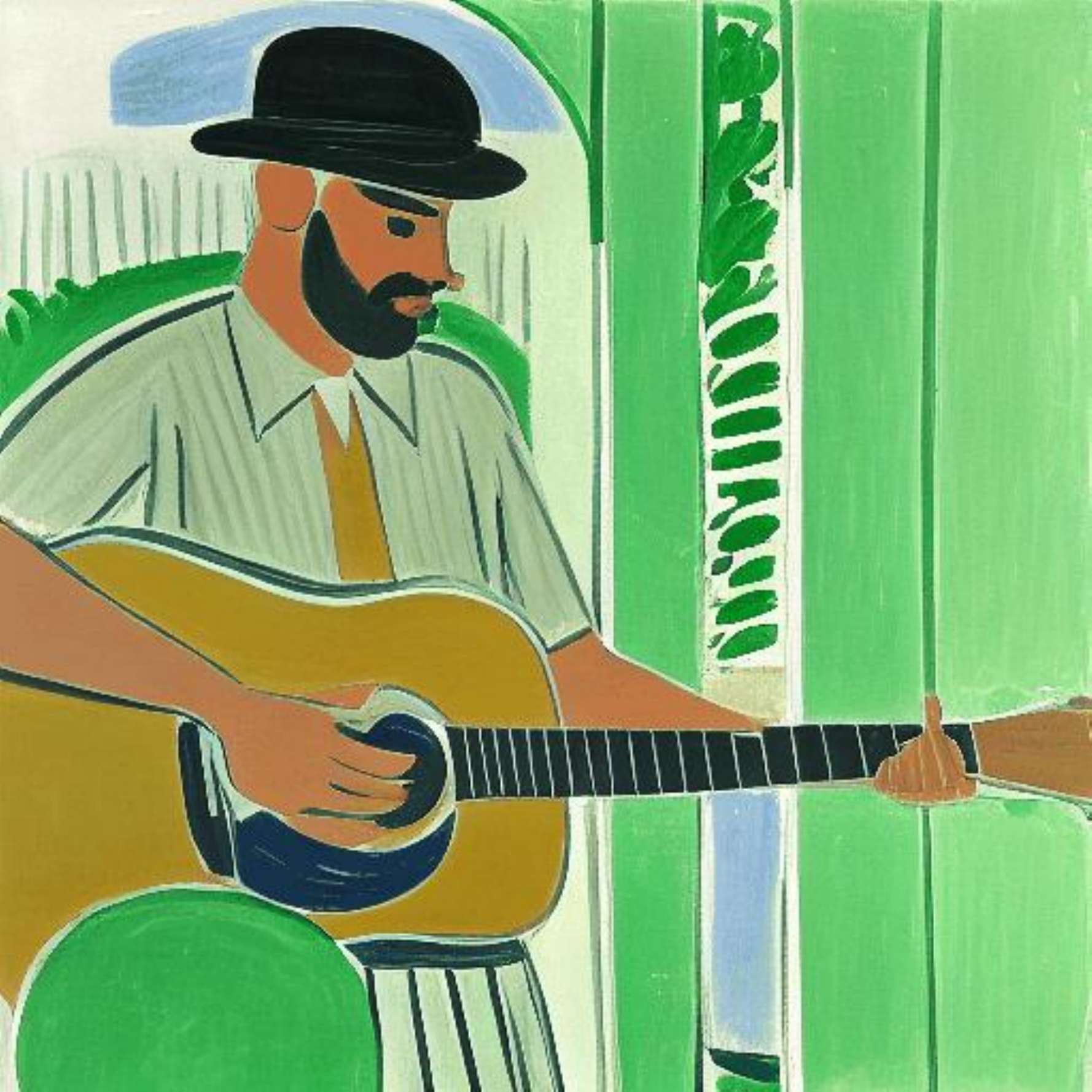}

\makebox[0.12\textwidth]{\colorbox{pink}{\textbf{Training video}} A cow is walking}\\
\includegraphics[width=0.10\textwidth]{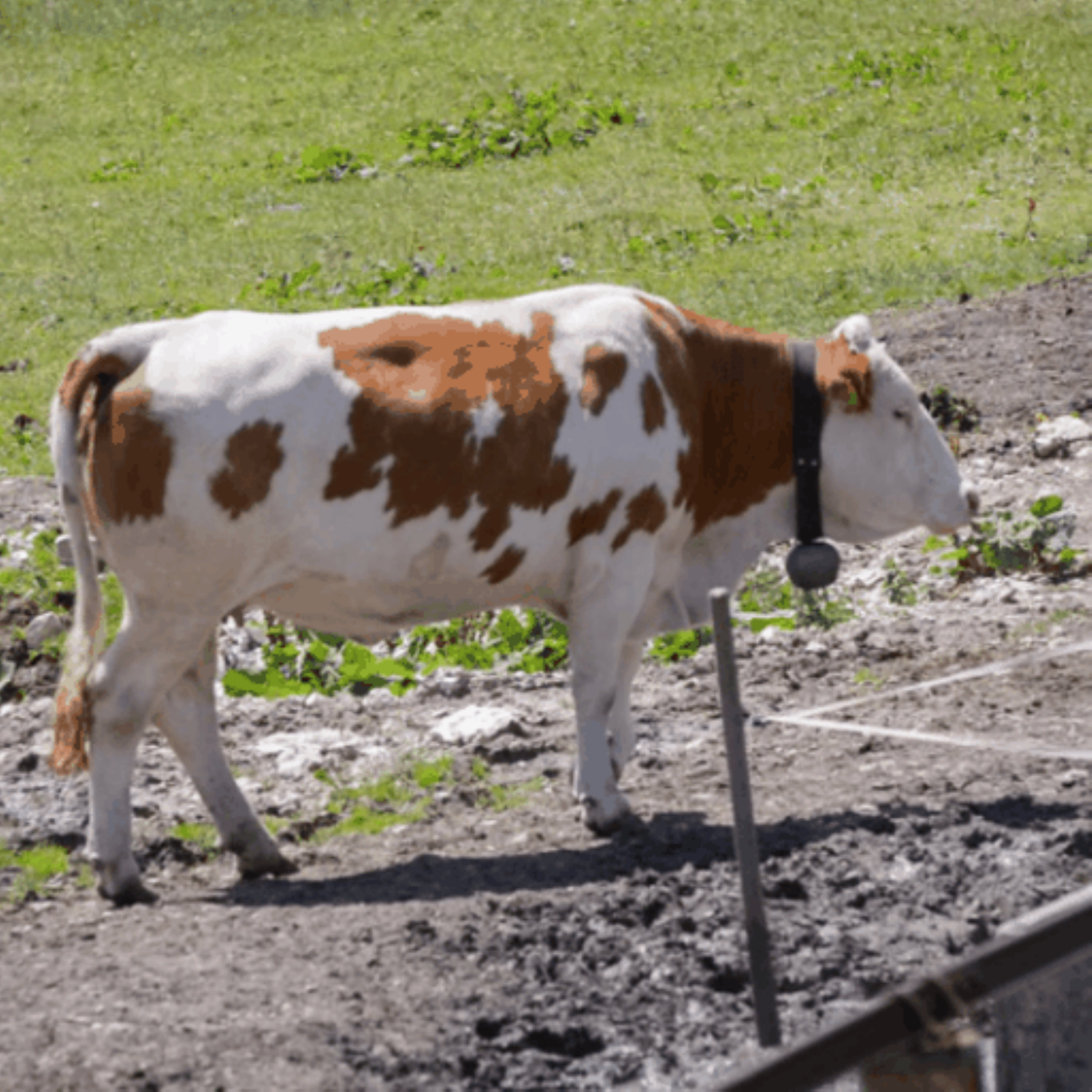}
\includegraphics[width=0.10\textwidth]{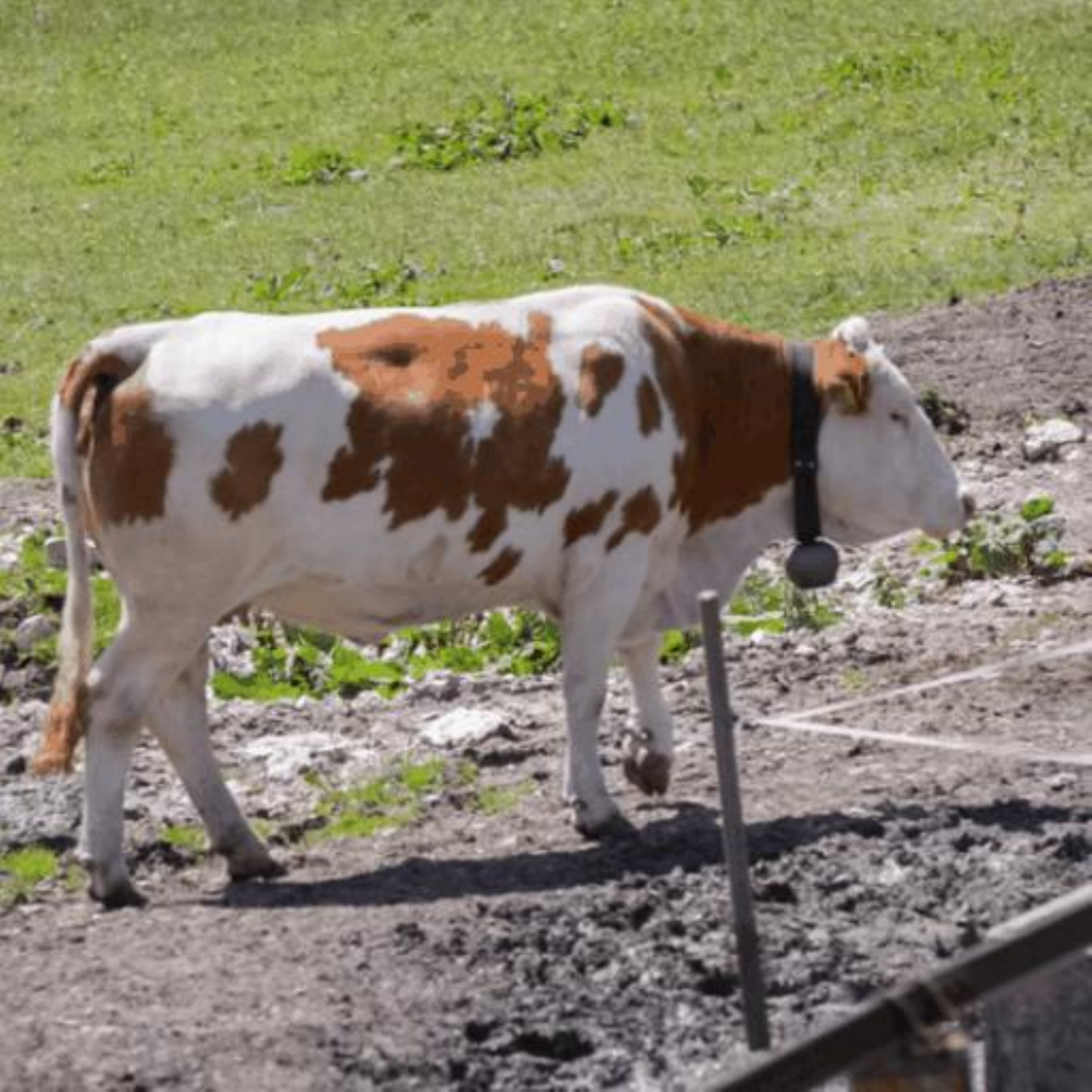}
\includegraphics[width=0.10\textwidth]{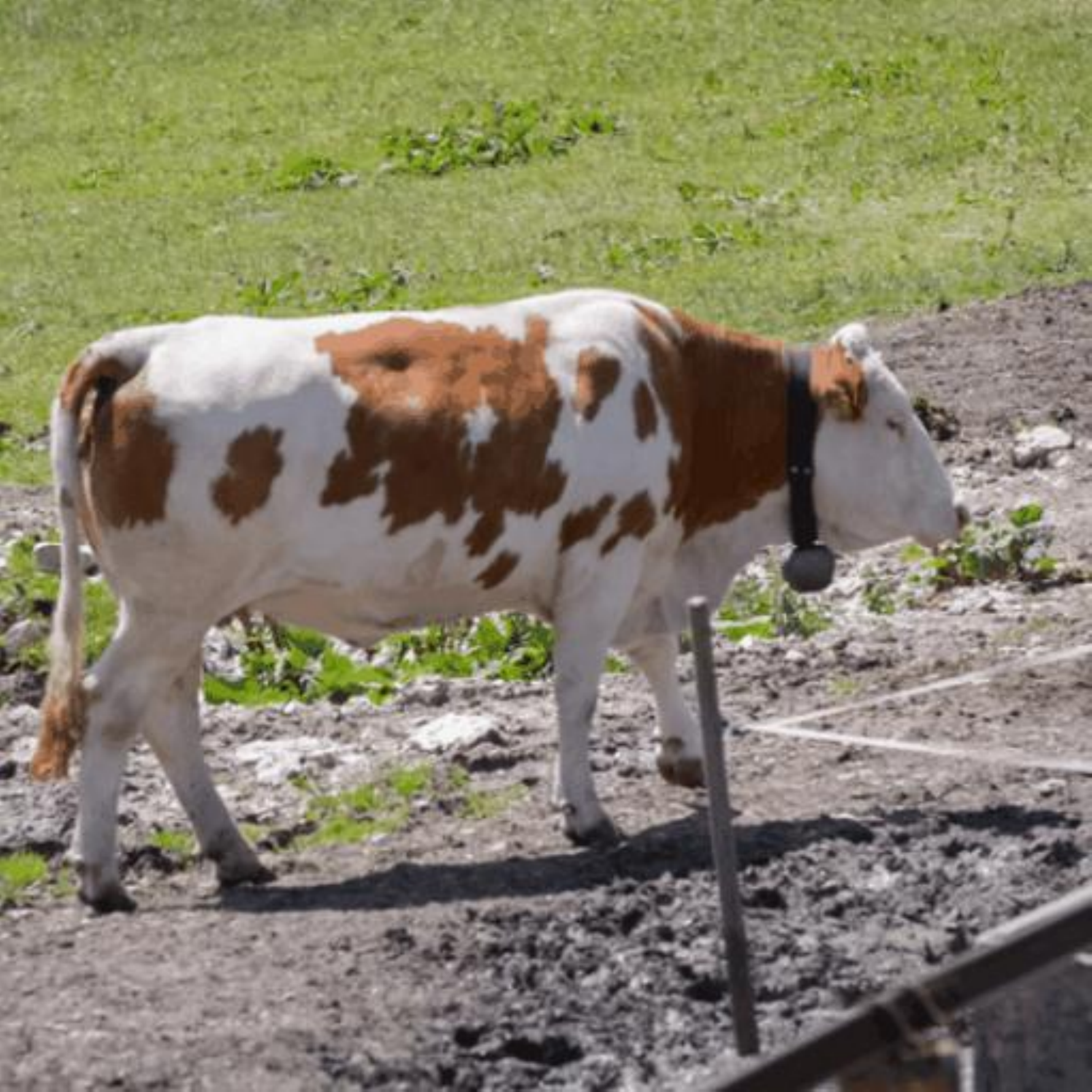}
\includegraphics[width=0.10\textwidth]{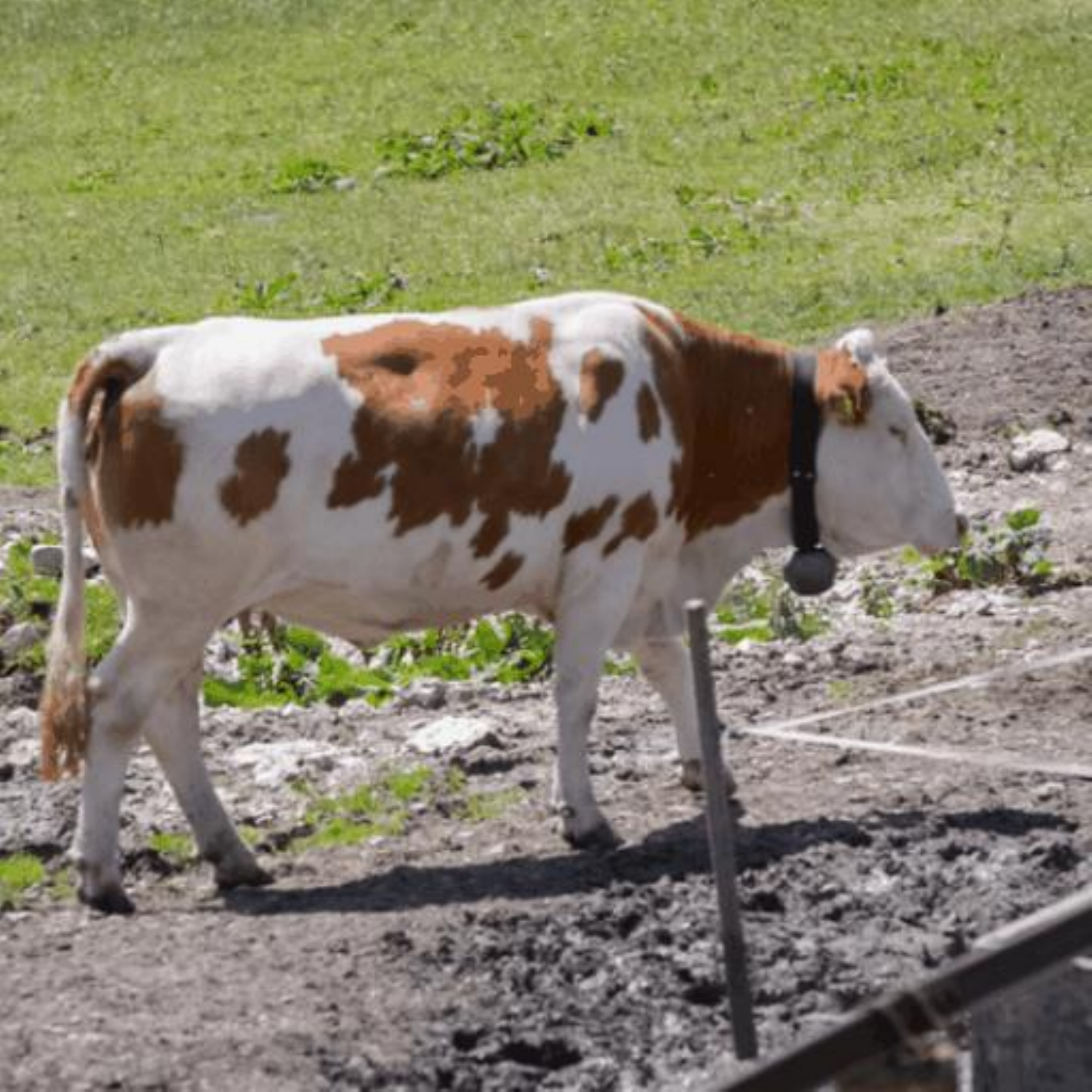}
\includegraphics[width=0.10\textwidth]{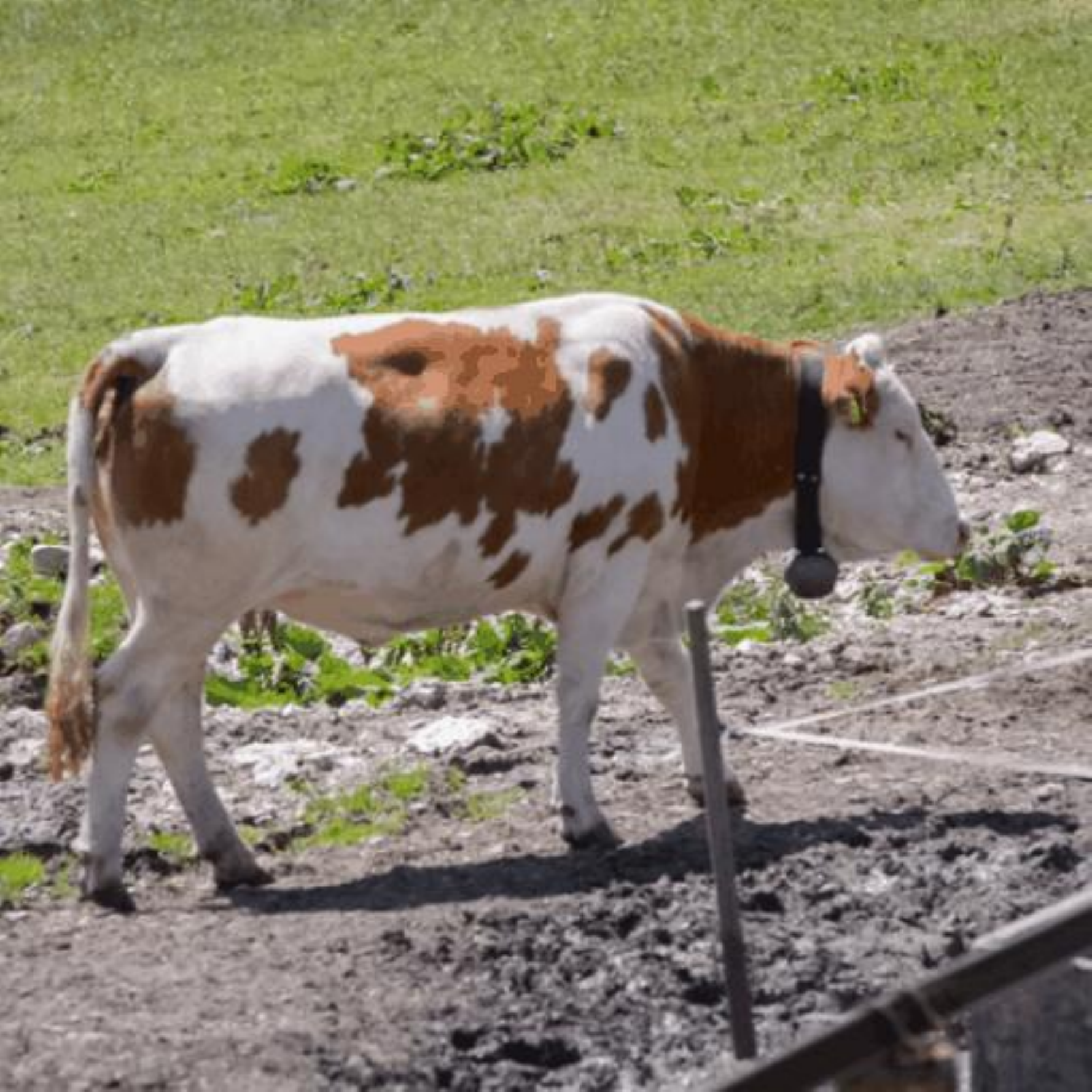}
\includegraphics[width=0.10\textwidth]{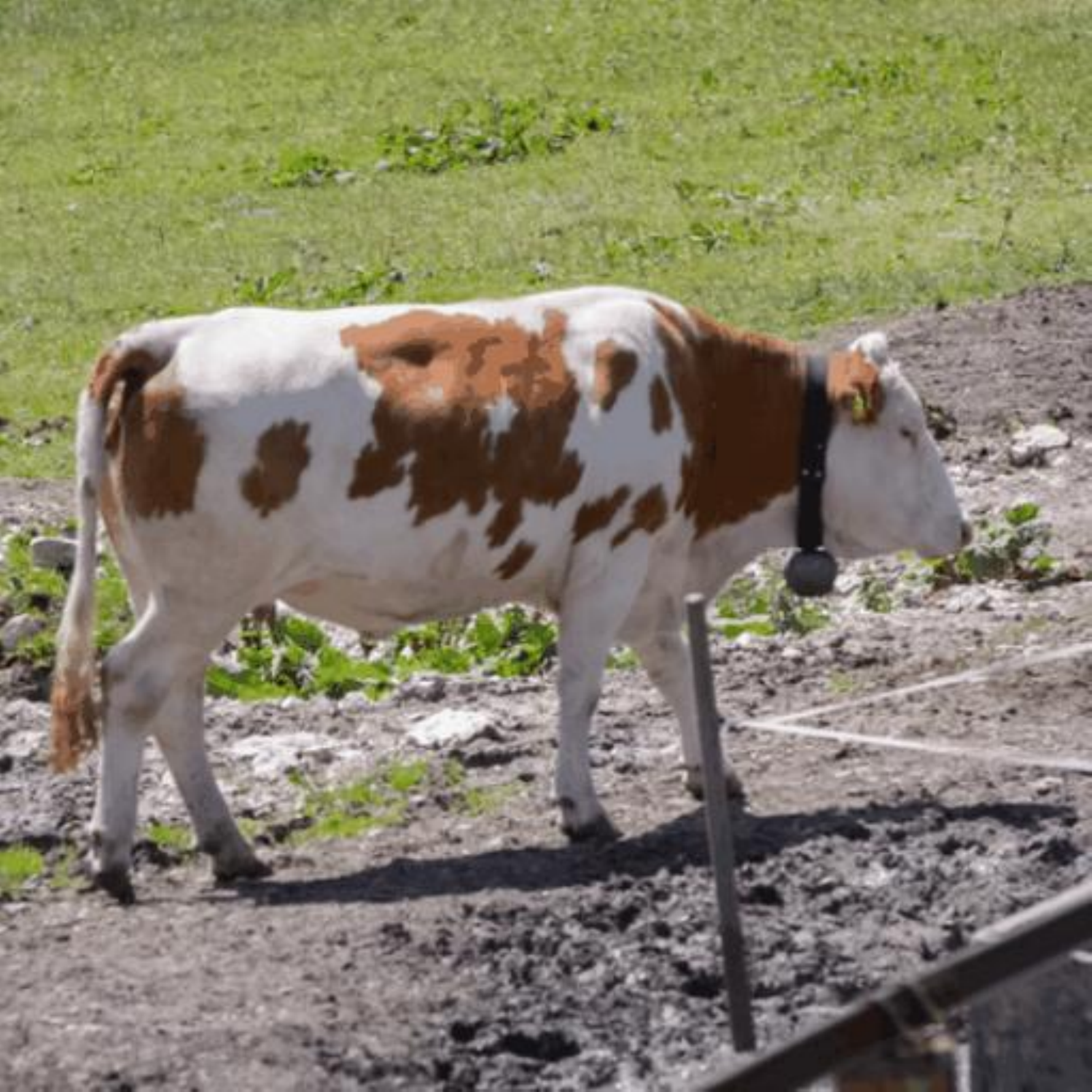}
\includegraphics[width=0.10\textwidth]{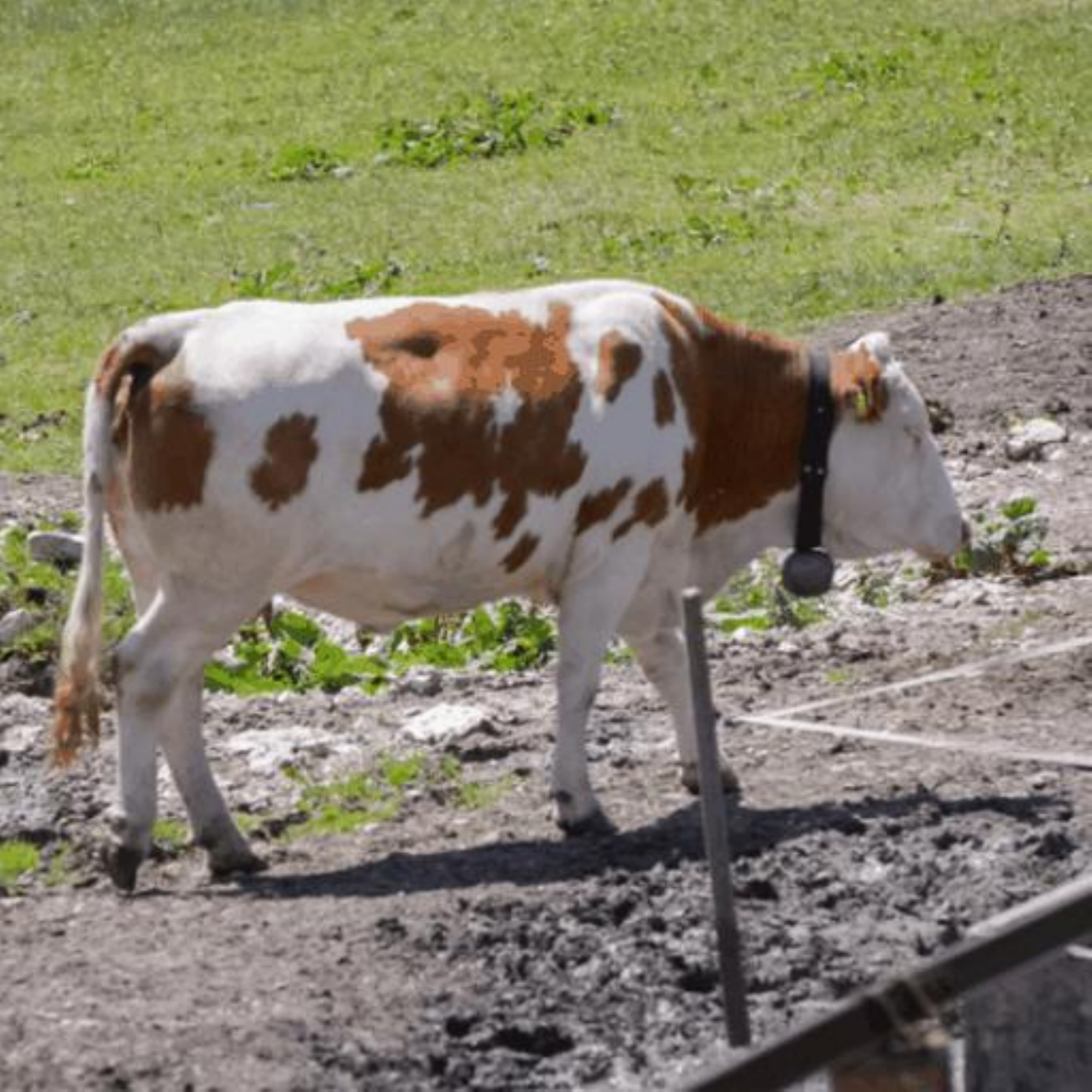}
\includegraphics[width=0.10\textwidth]{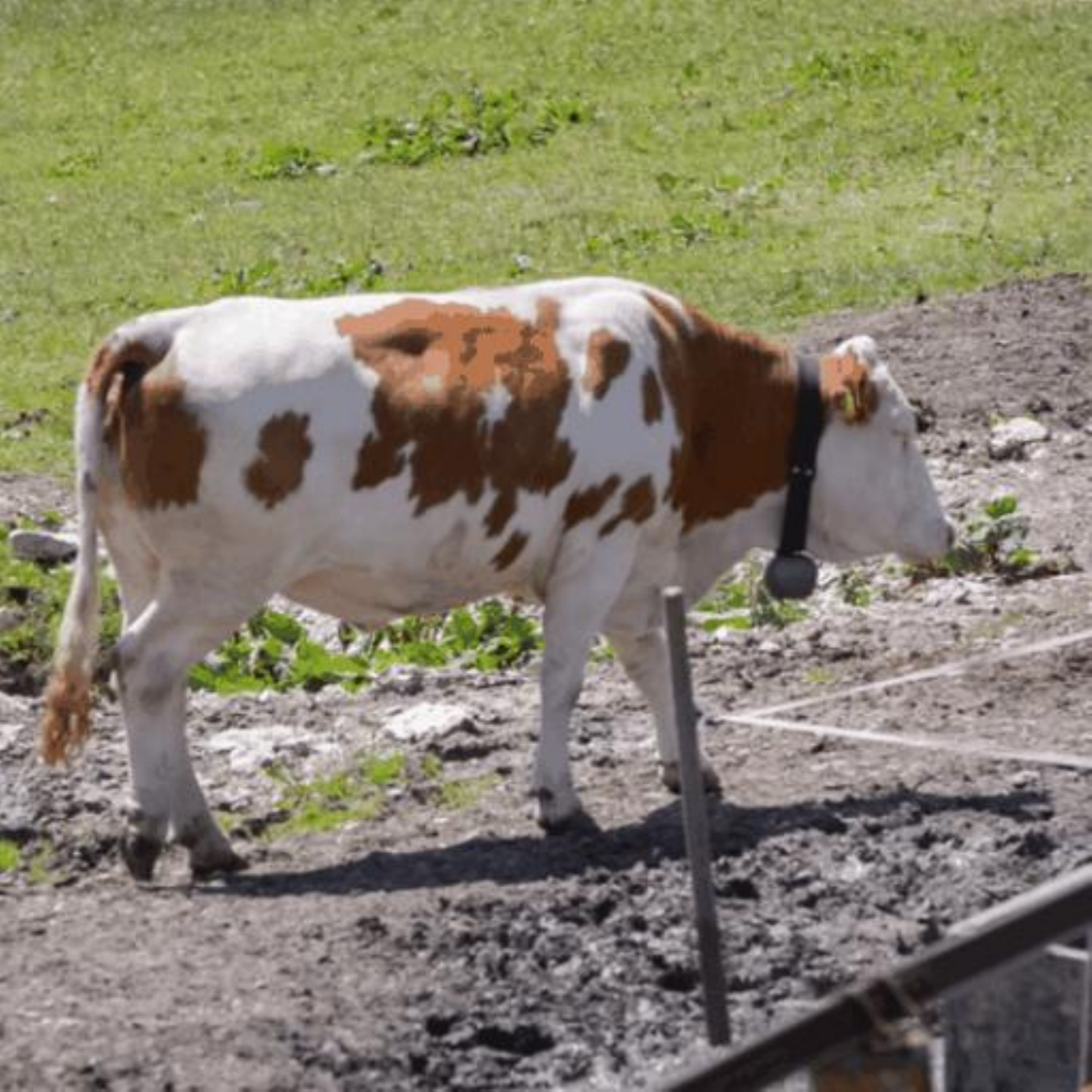}

\makebox[0.12\textwidth]{A \textcolor{blue}{\textbf{zebra}} is walking.}\\
\includegraphics[width=0.10\textwidth]{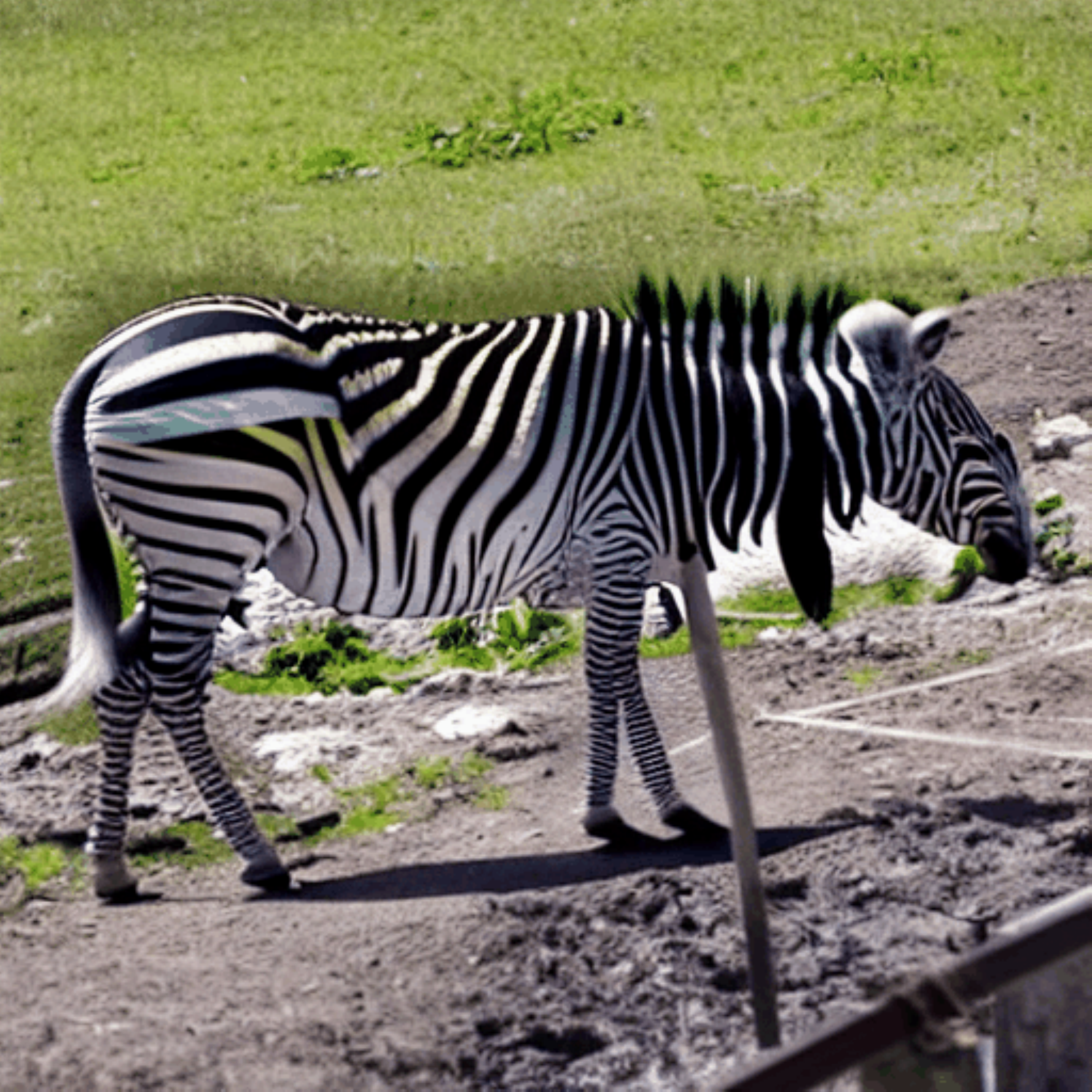}
\includegraphics[width=0.10\textwidth]{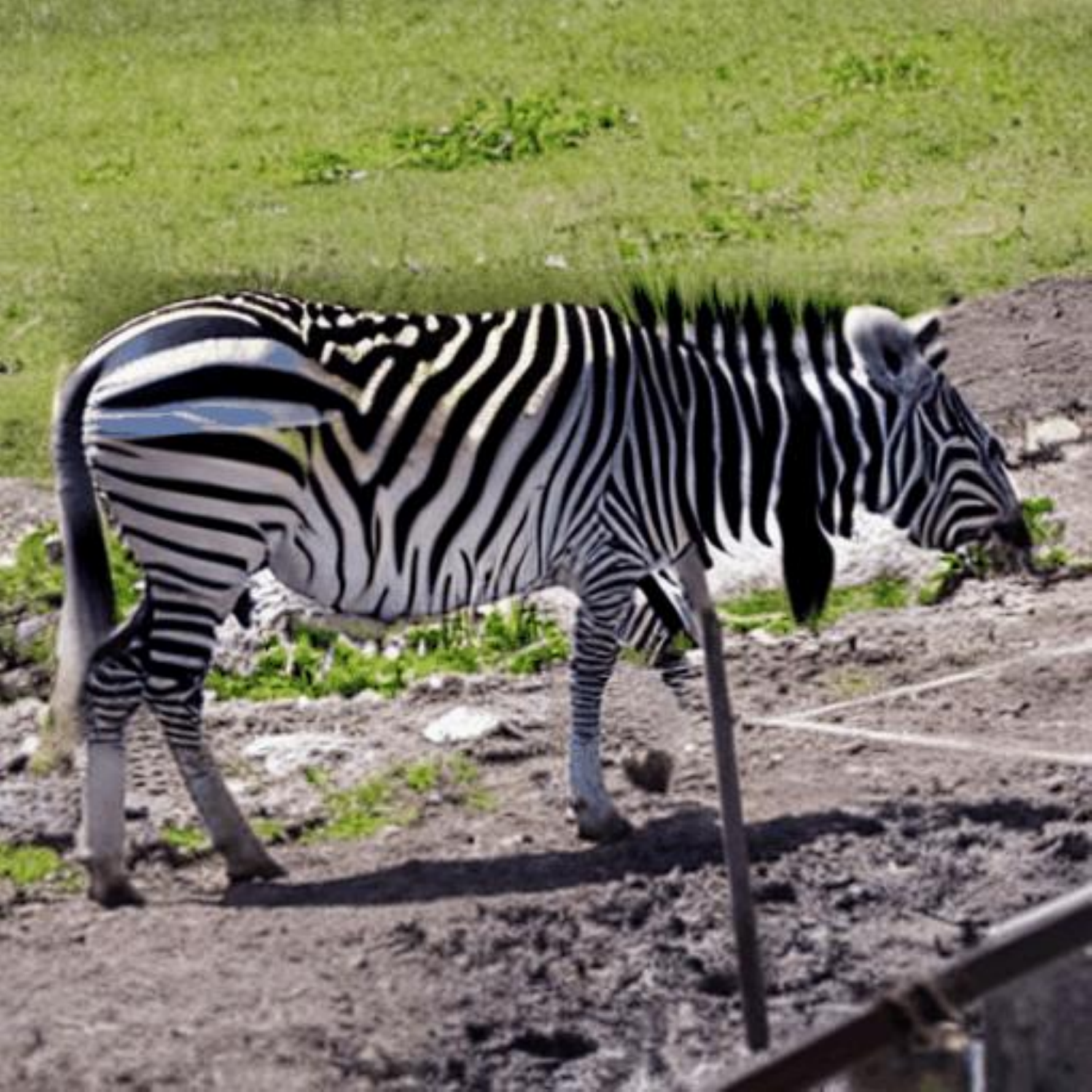}
\includegraphics[width=0.10\textwidth]{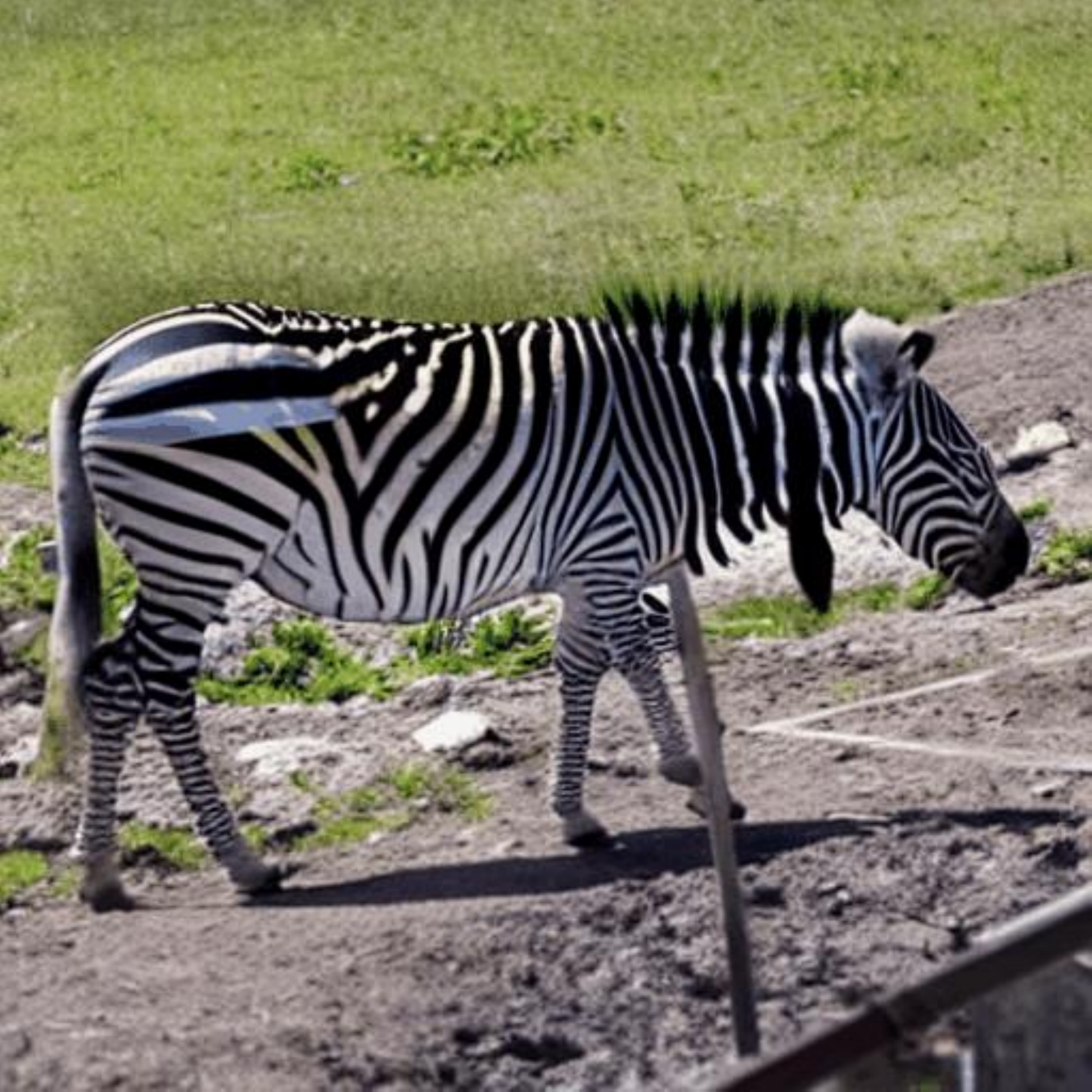}
\includegraphics[width=0.10\textwidth]{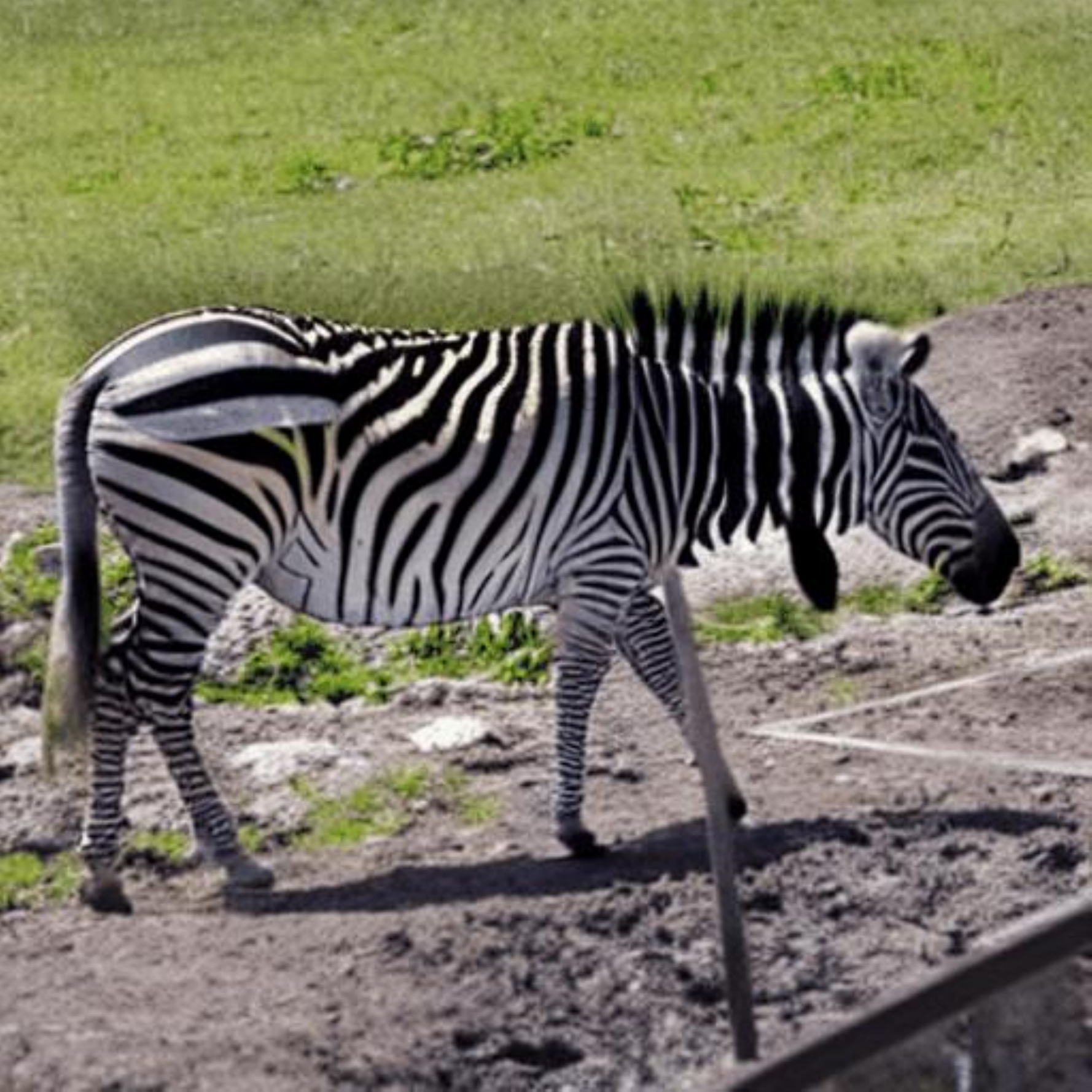}
\includegraphics[width=0.10\textwidth]{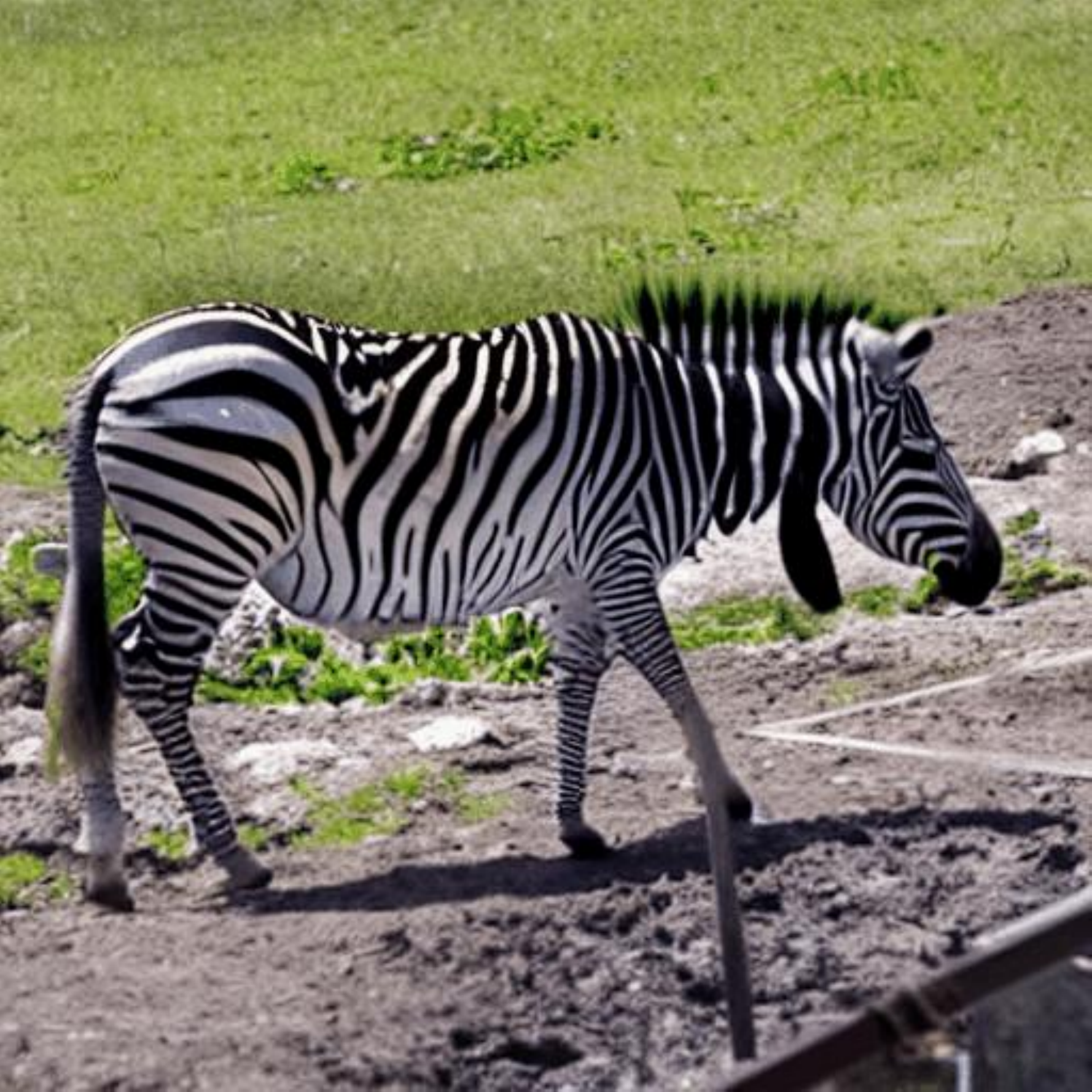}
\includegraphics[width=0.10\textwidth]{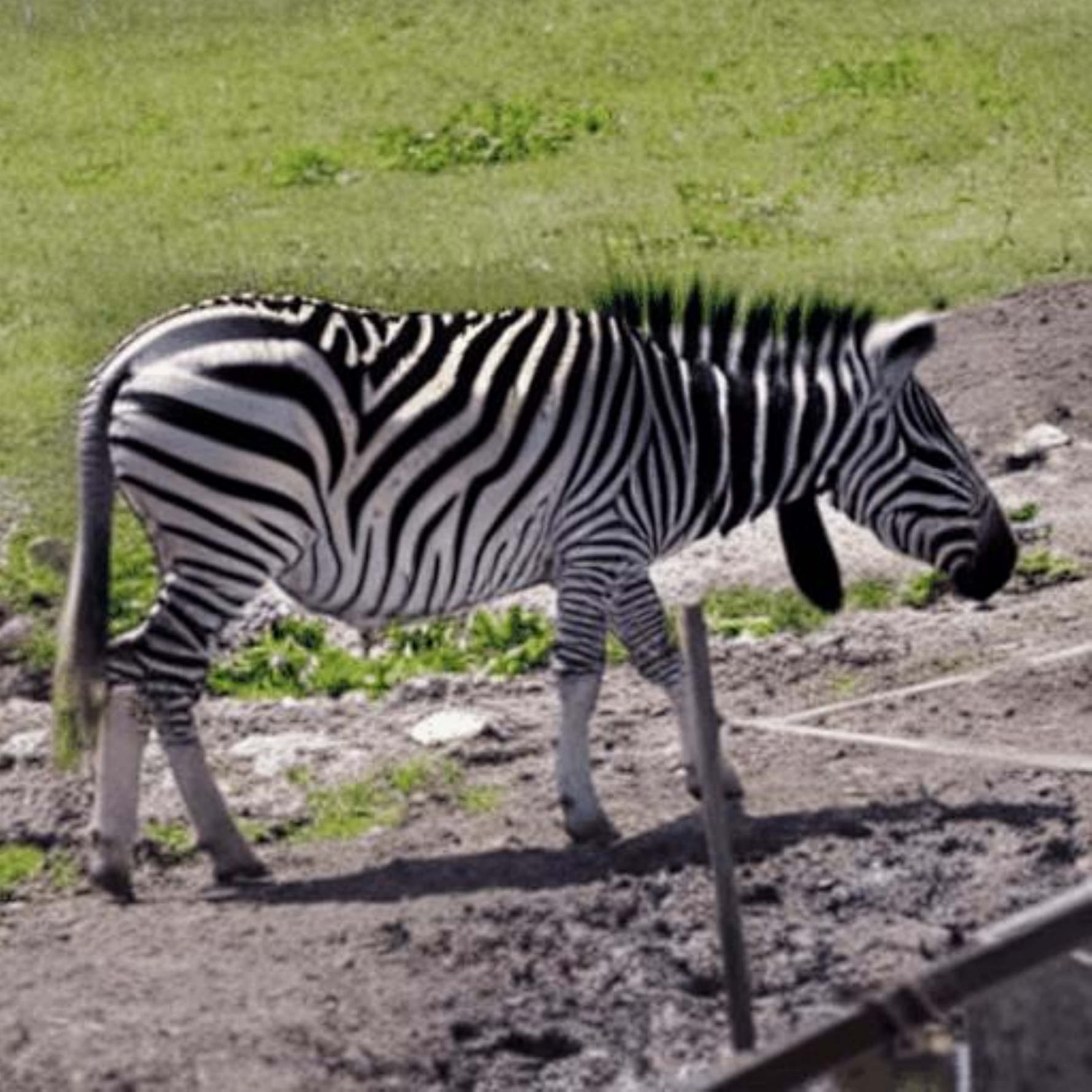}
\includegraphics[width=0.10\textwidth]{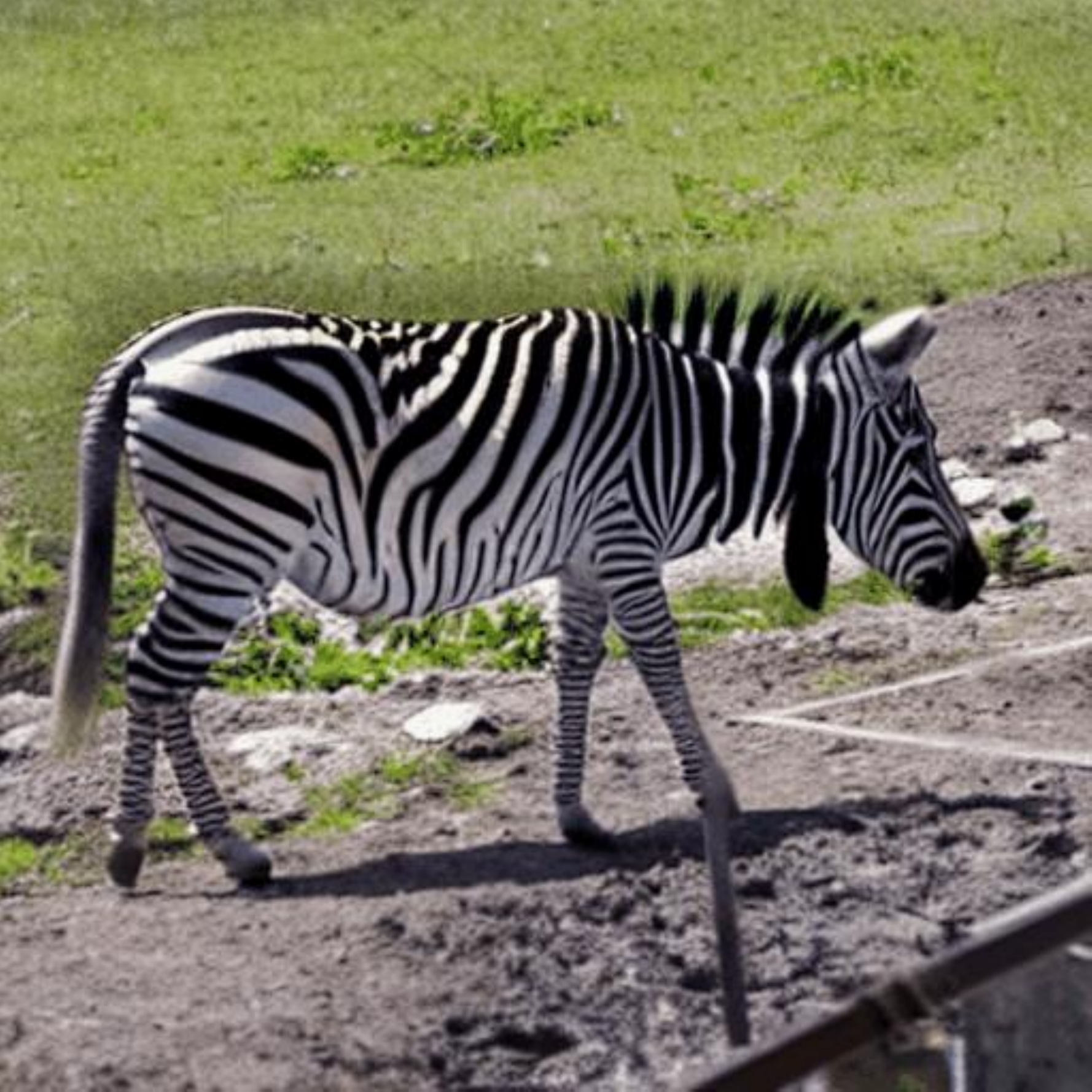}
\includegraphics[width=0.10\textwidth]{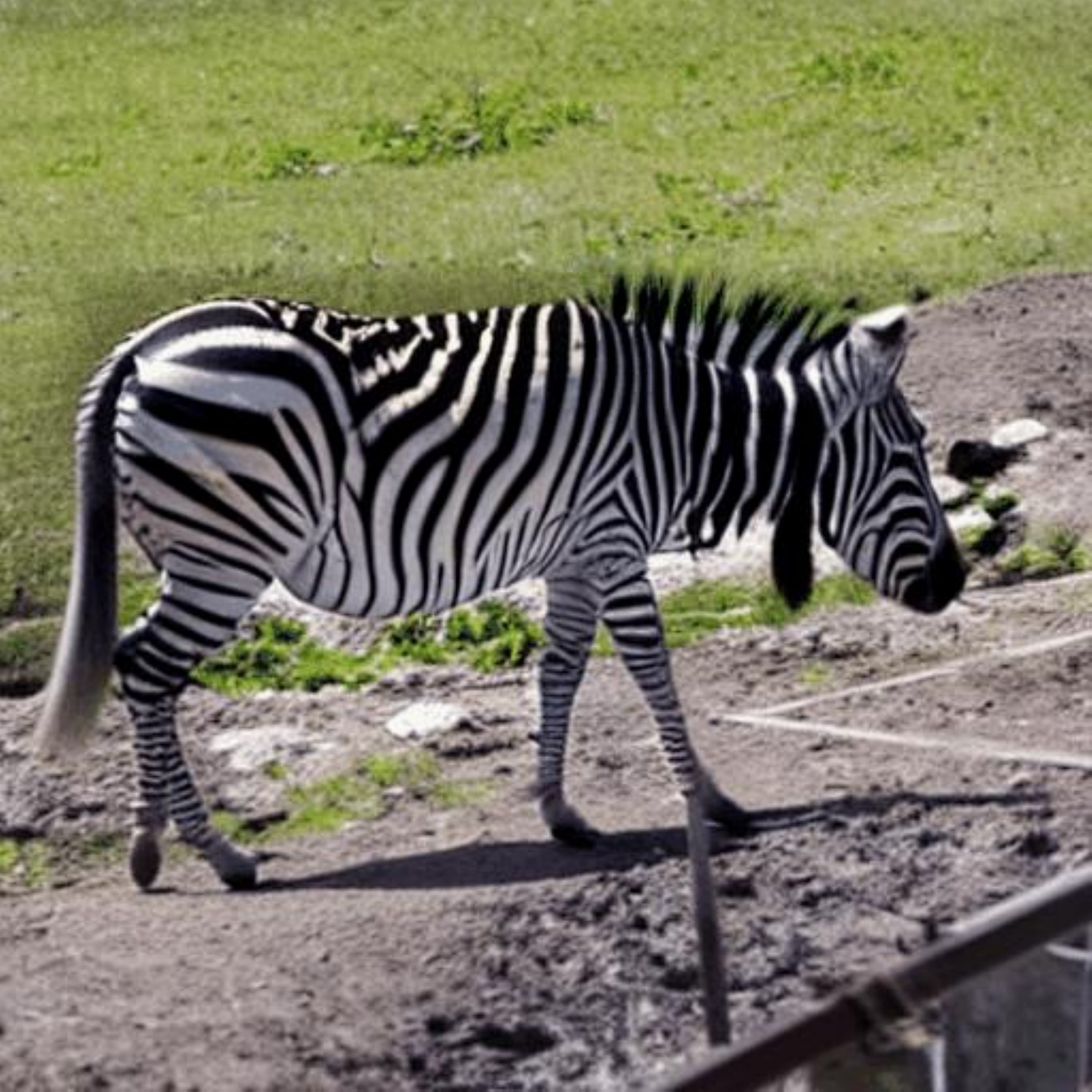}

\makebox[0.12\textwidth]{A \textcolor{blue}{\textbf{bull}} is walking.}\\
\includegraphics[width=0.10\textwidth]{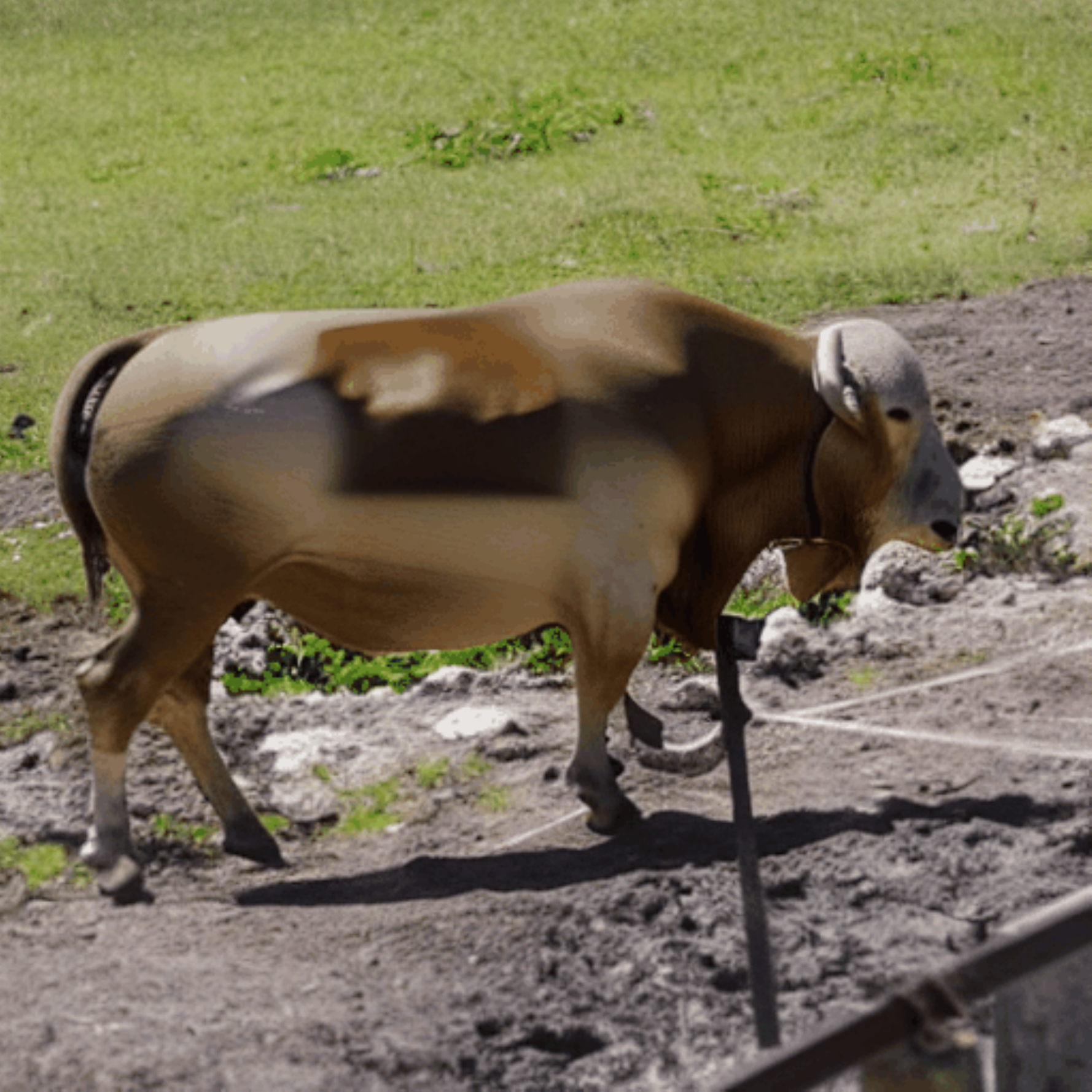}
\includegraphics[width=0.10\textwidth]{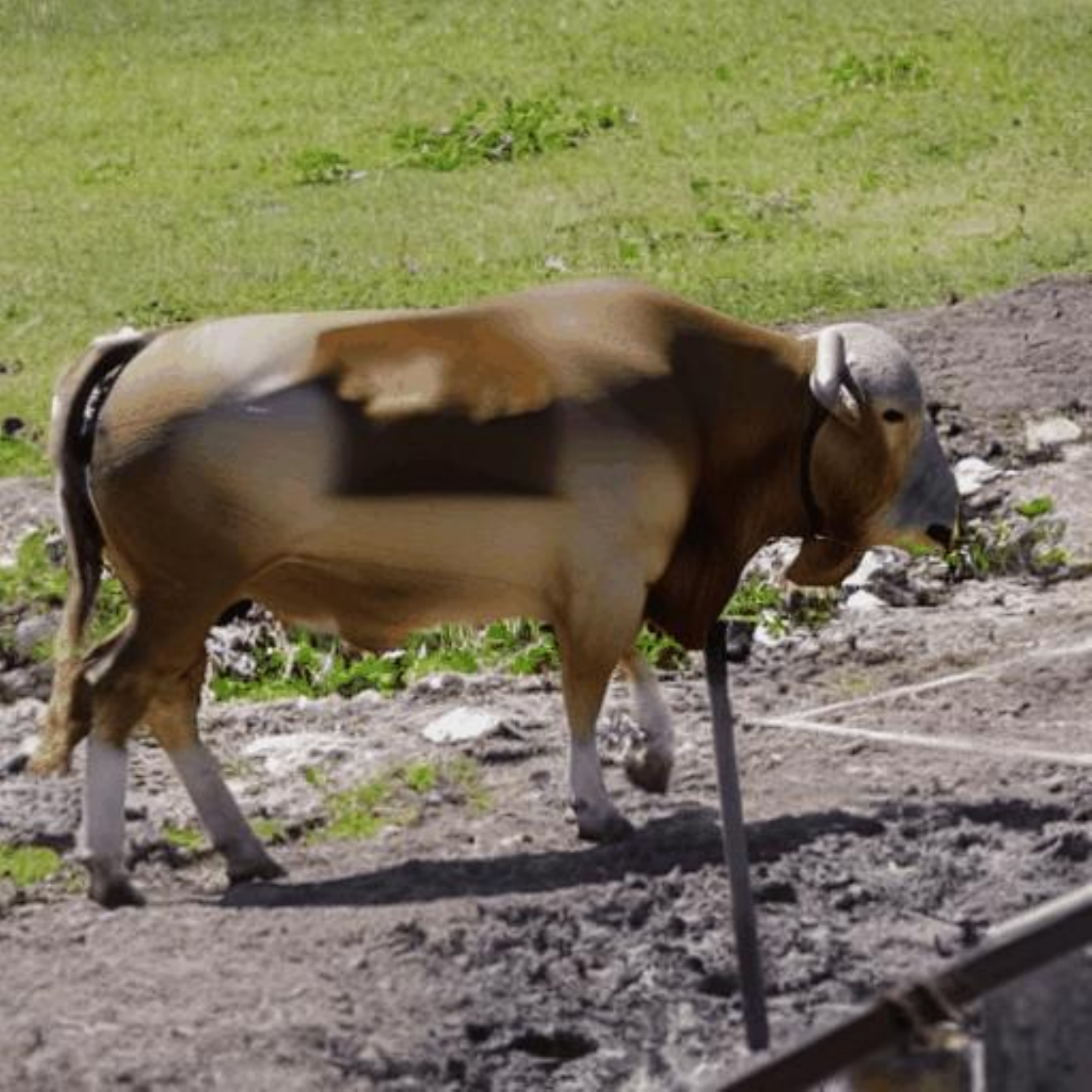}
\includegraphics[width=0.10\textwidth]{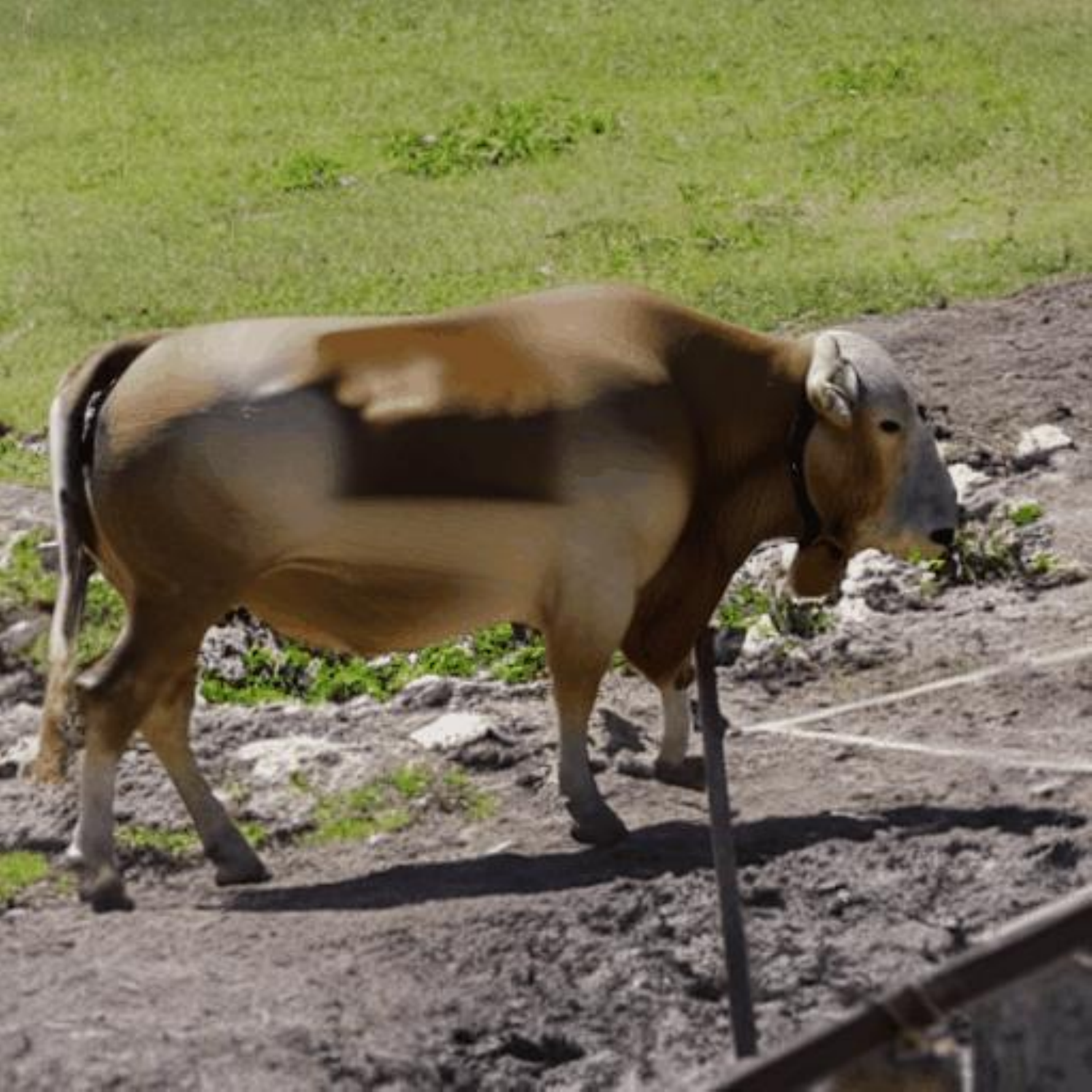}
\includegraphics[width=0.10\textwidth]{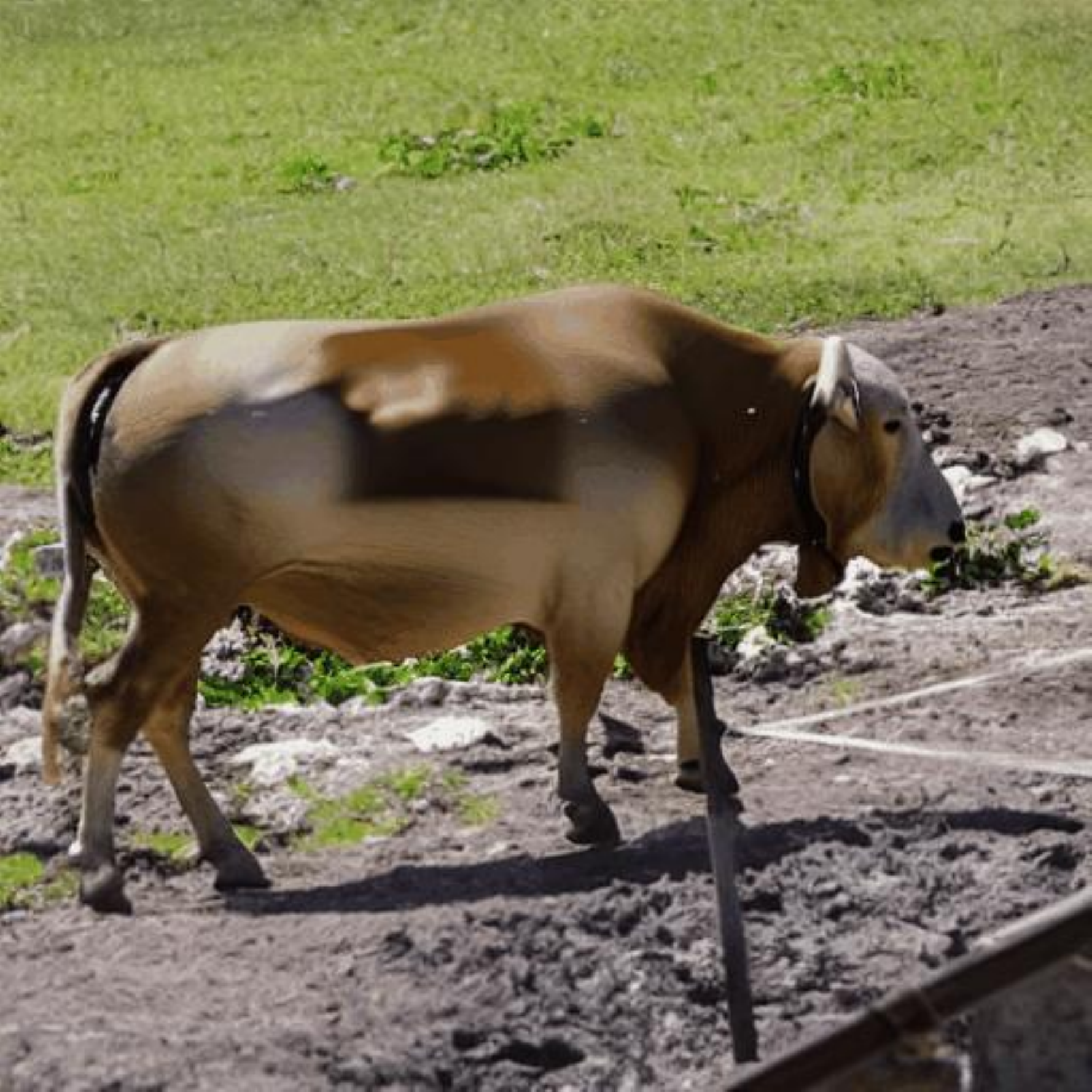}
\includegraphics[width=0.10\textwidth]{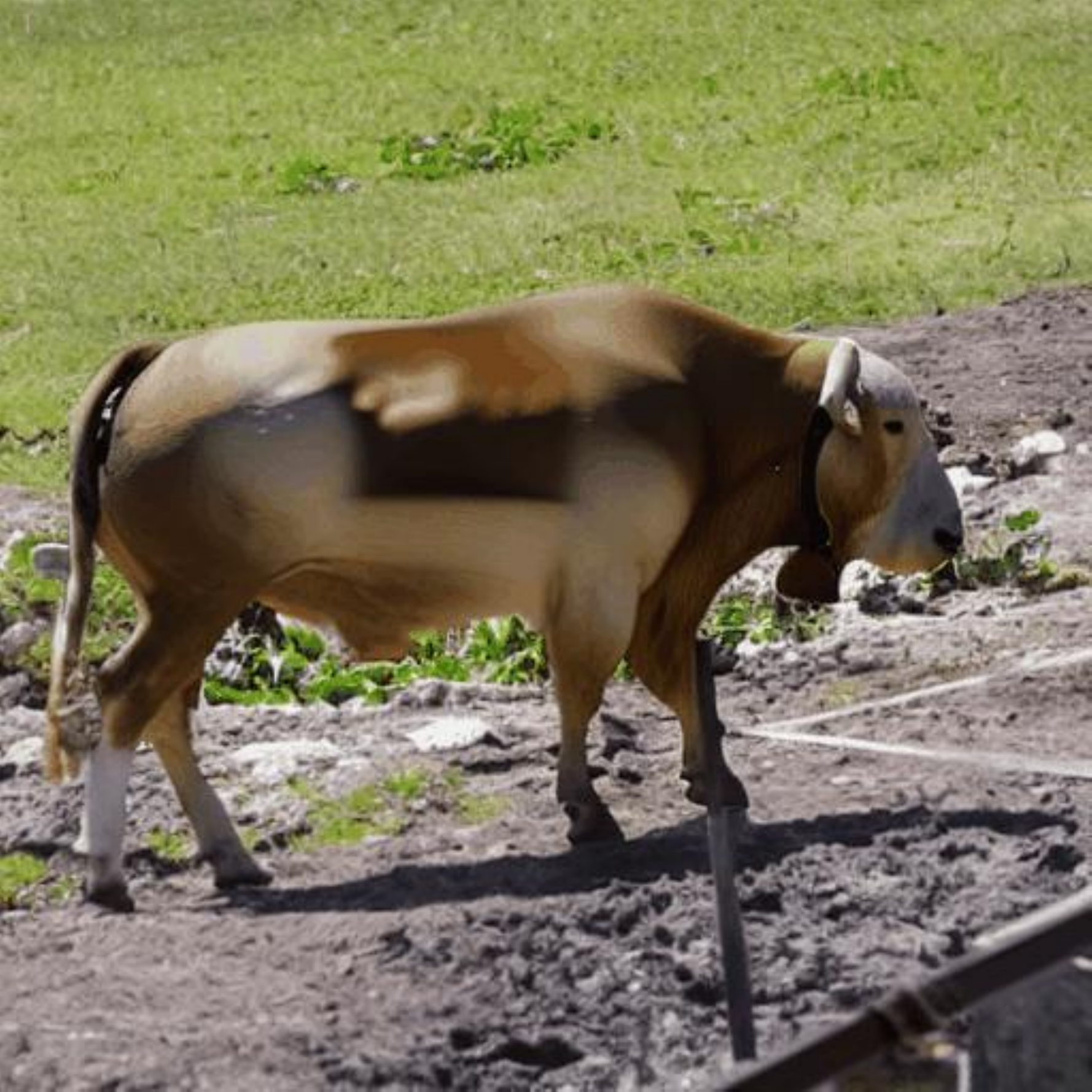}
\includegraphics[width=0.10\textwidth]{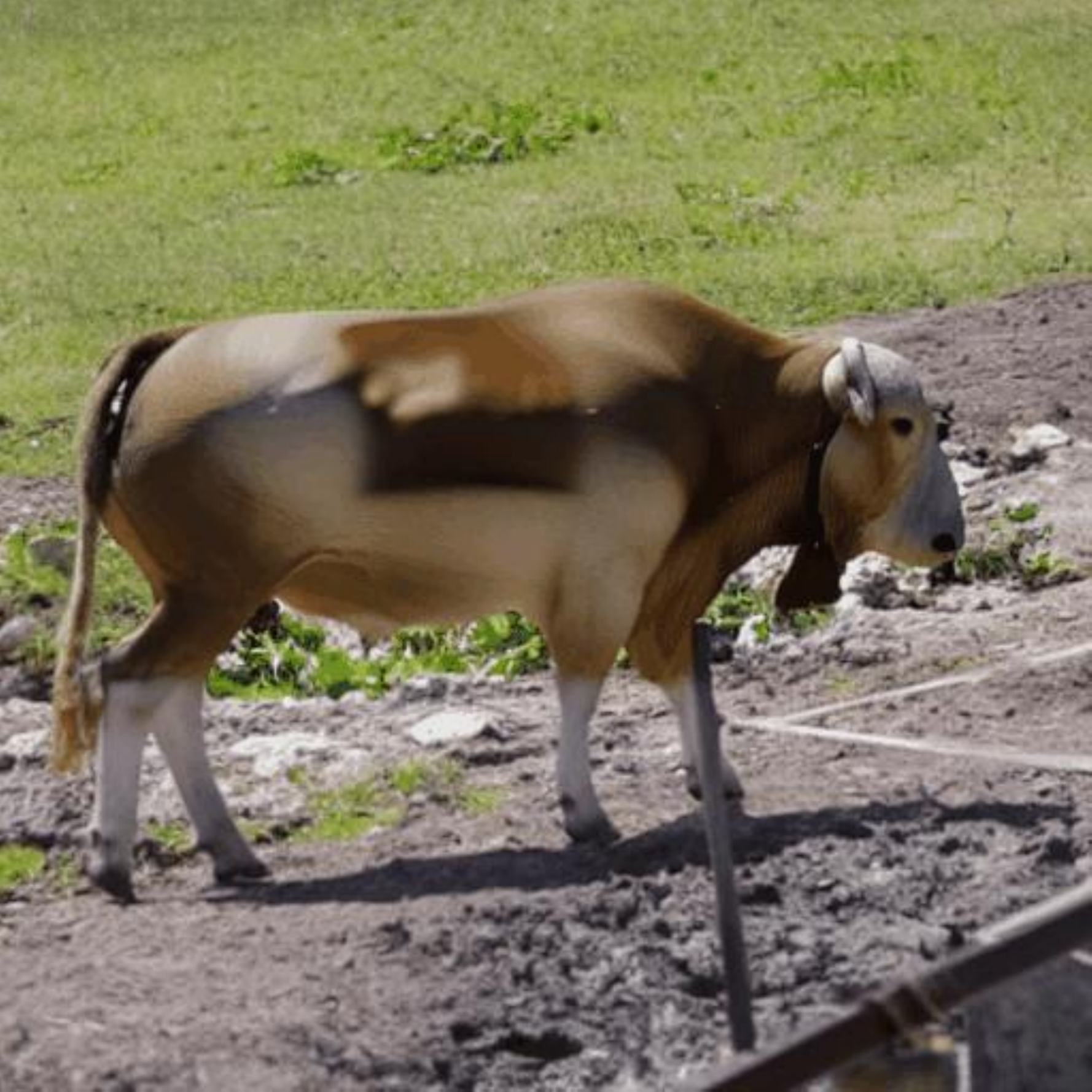}
\includegraphics[width=0.10\textwidth]{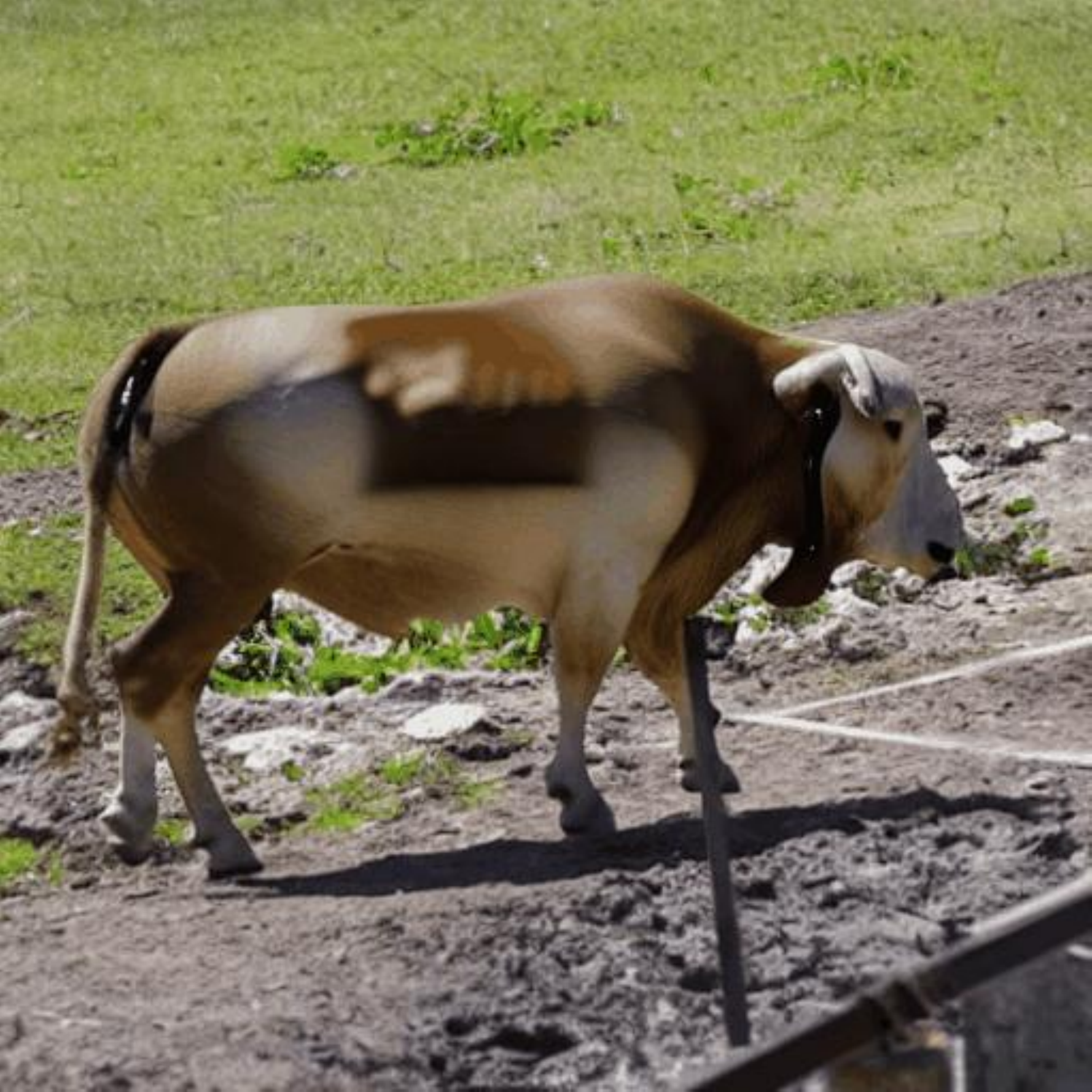}
\includegraphics[width=0.10\textwidth]{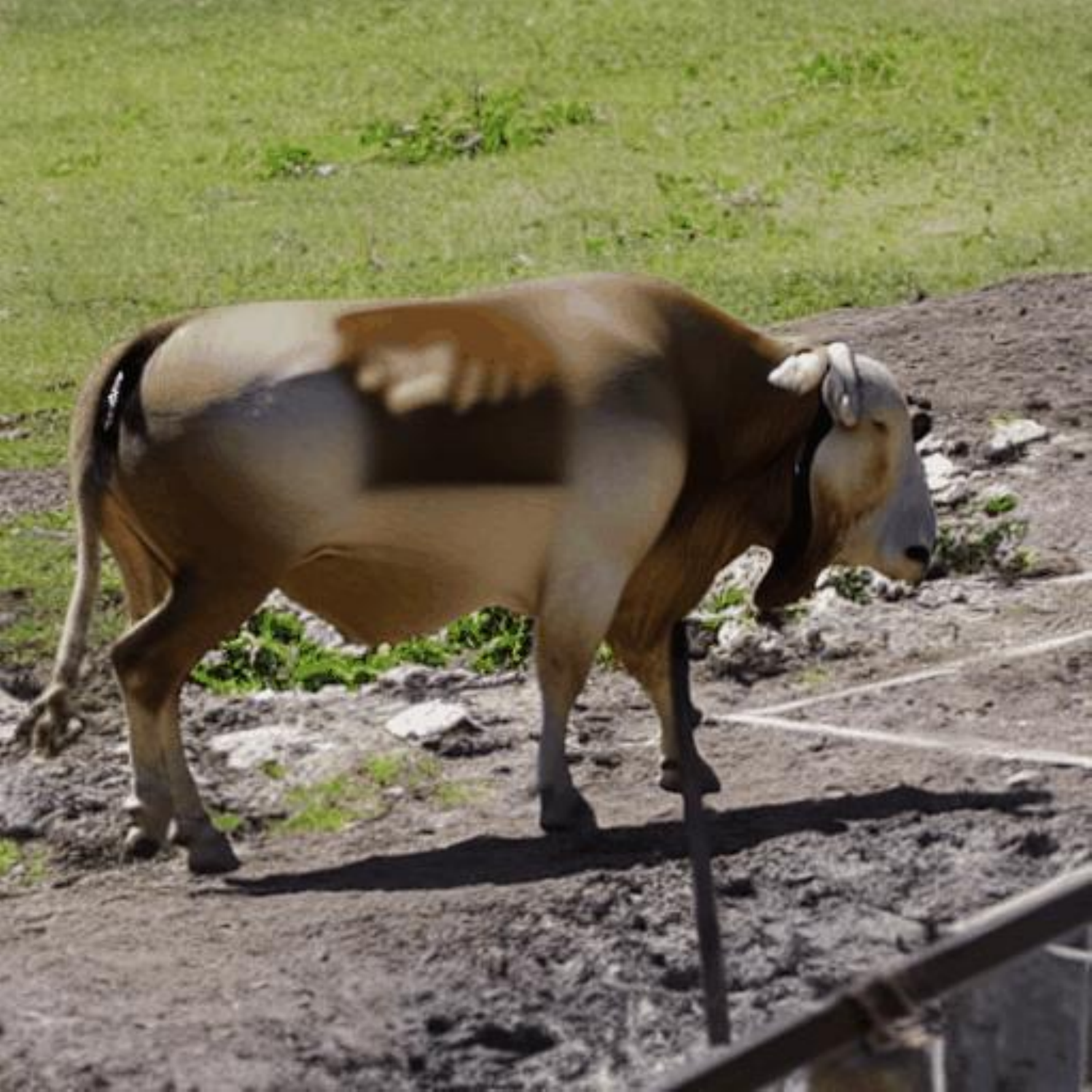}

\makebox[0.12\textwidth]{A cow is walking, \textcolor{blue}{\textbf{on the snow}}.}\\
\includegraphics[width=0.10\textwidth]{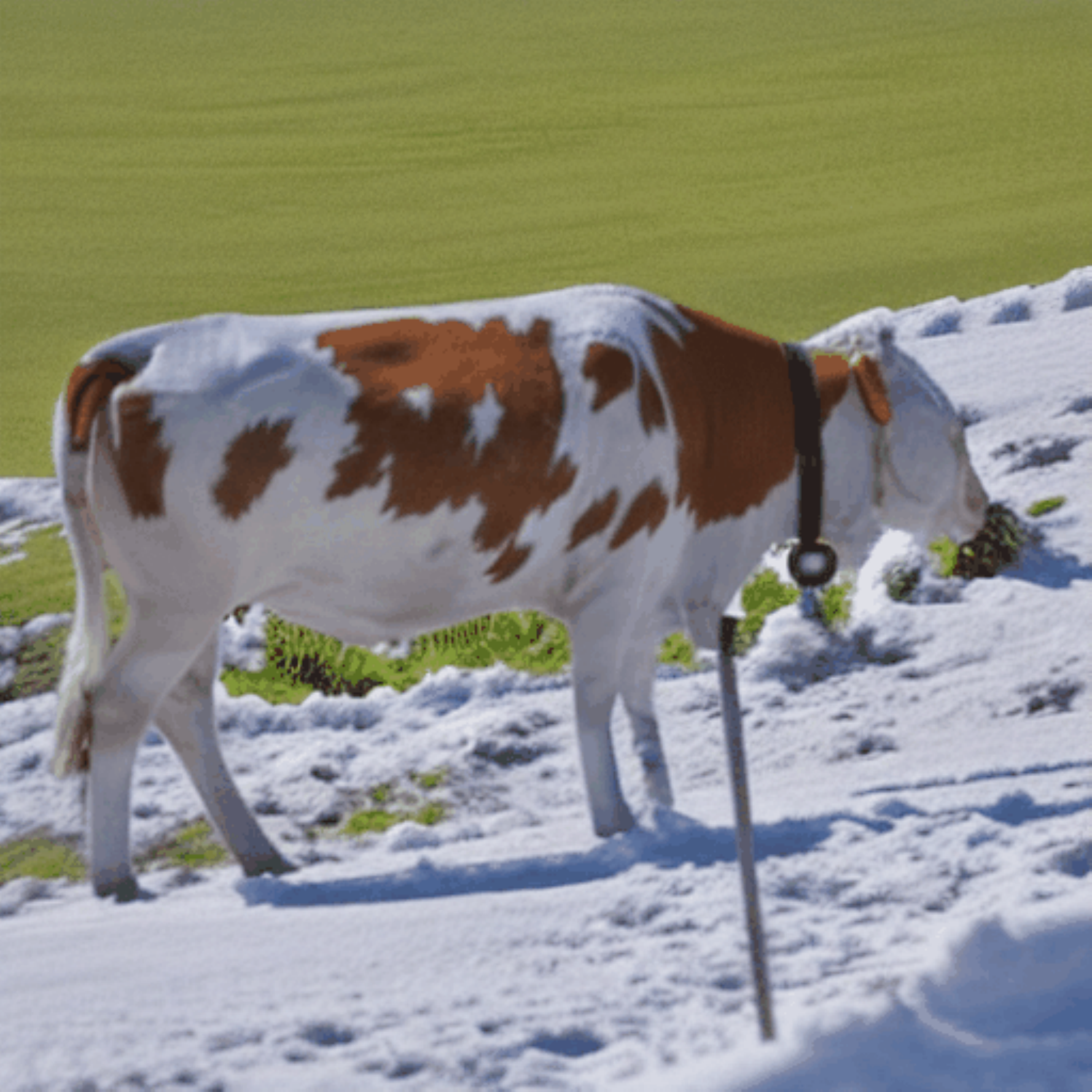}
\includegraphics[width=0.10\textwidth]{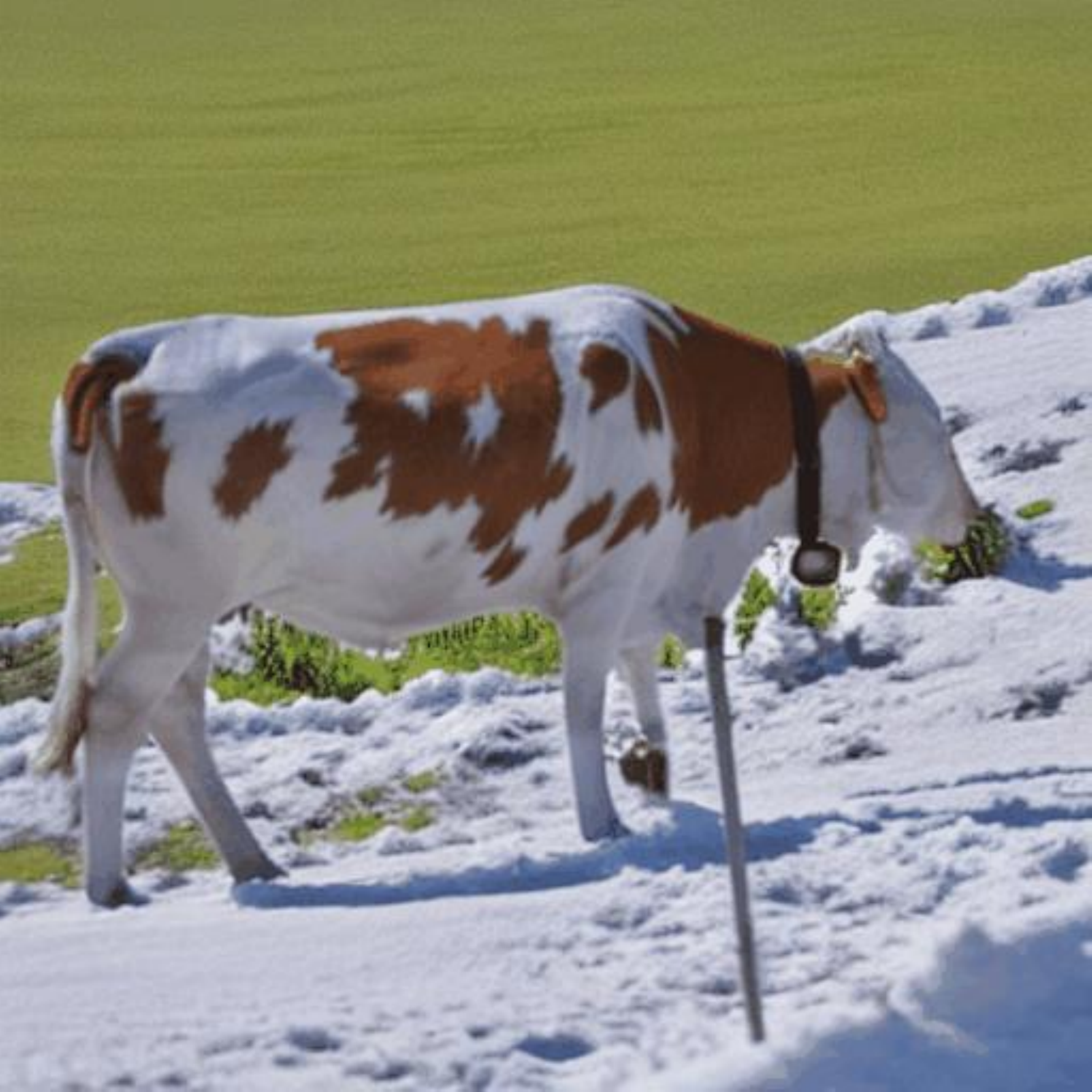}
\includegraphics[width=0.10\textwidth]{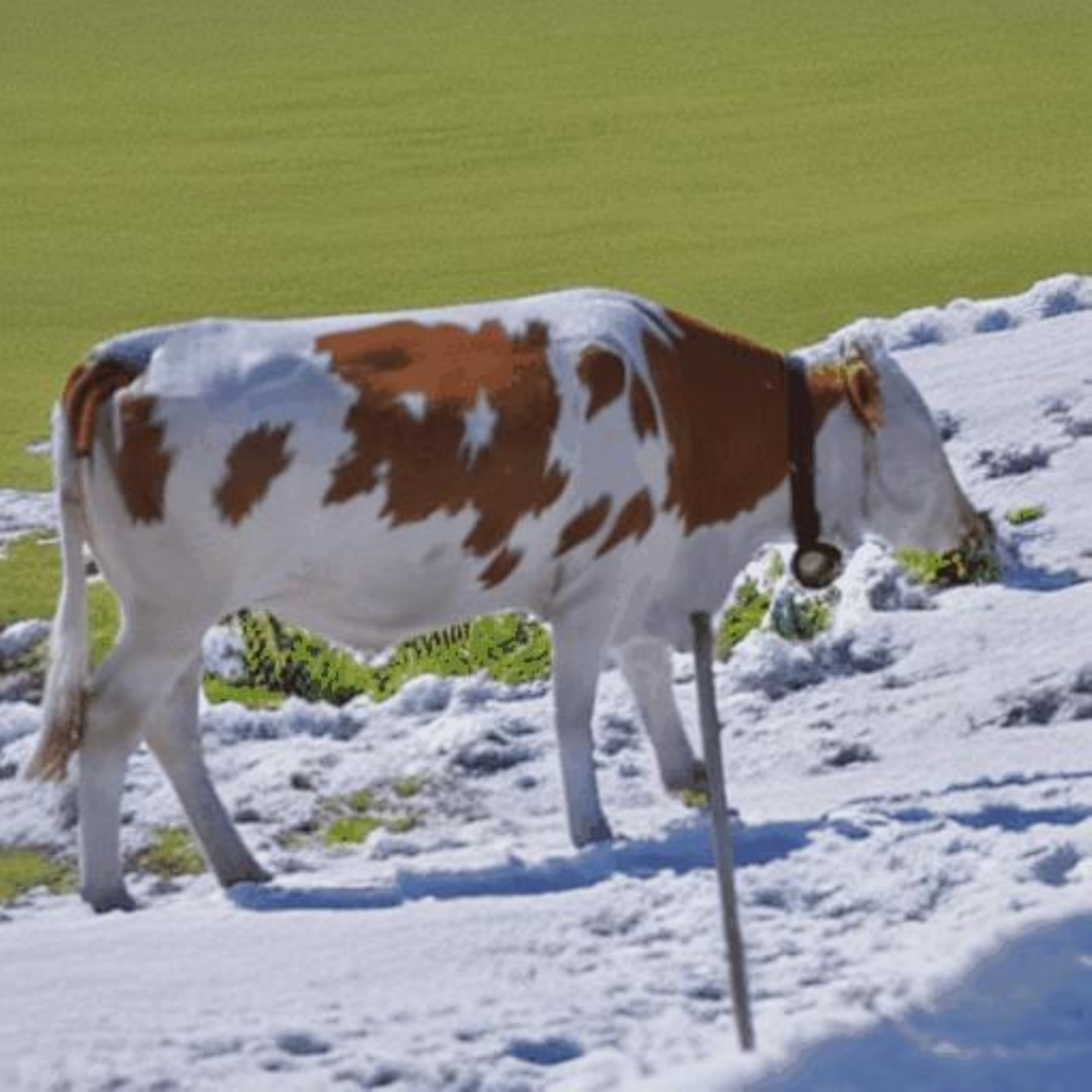}
\includegraphics[width=0.10\textwidth]{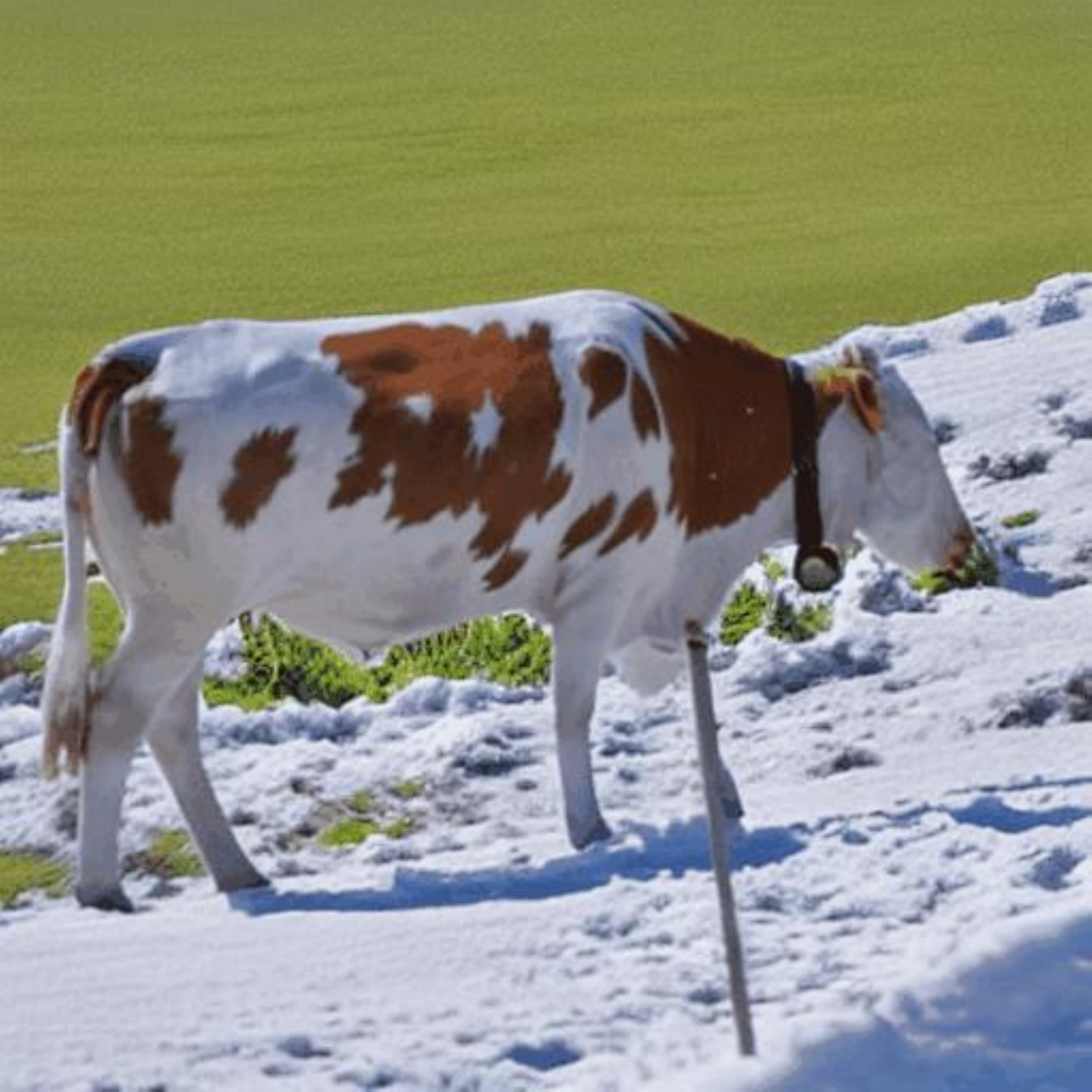}
\includegraphics[width=0.10\textwidth]{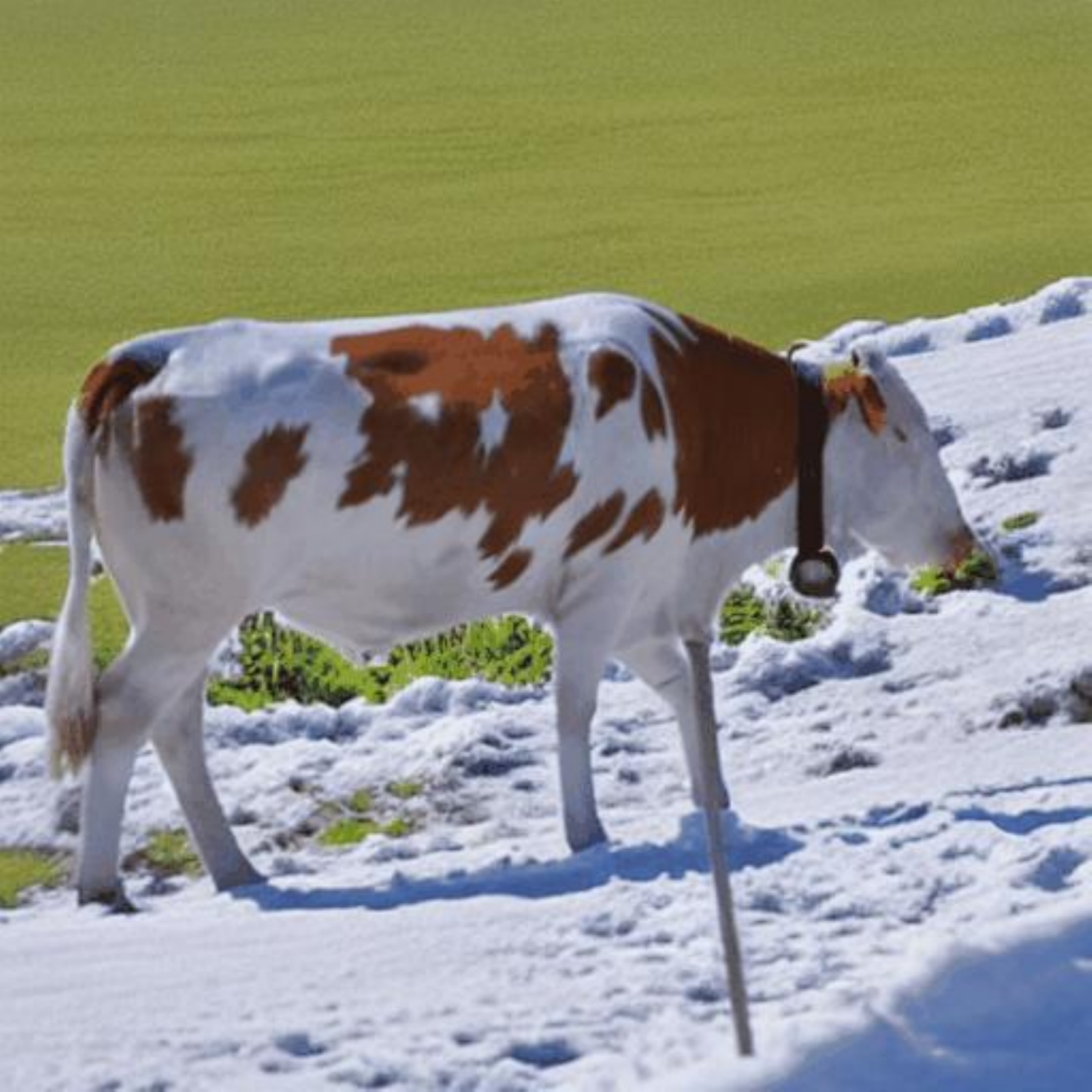}
\includegraphics[width=0.10\textwidth]{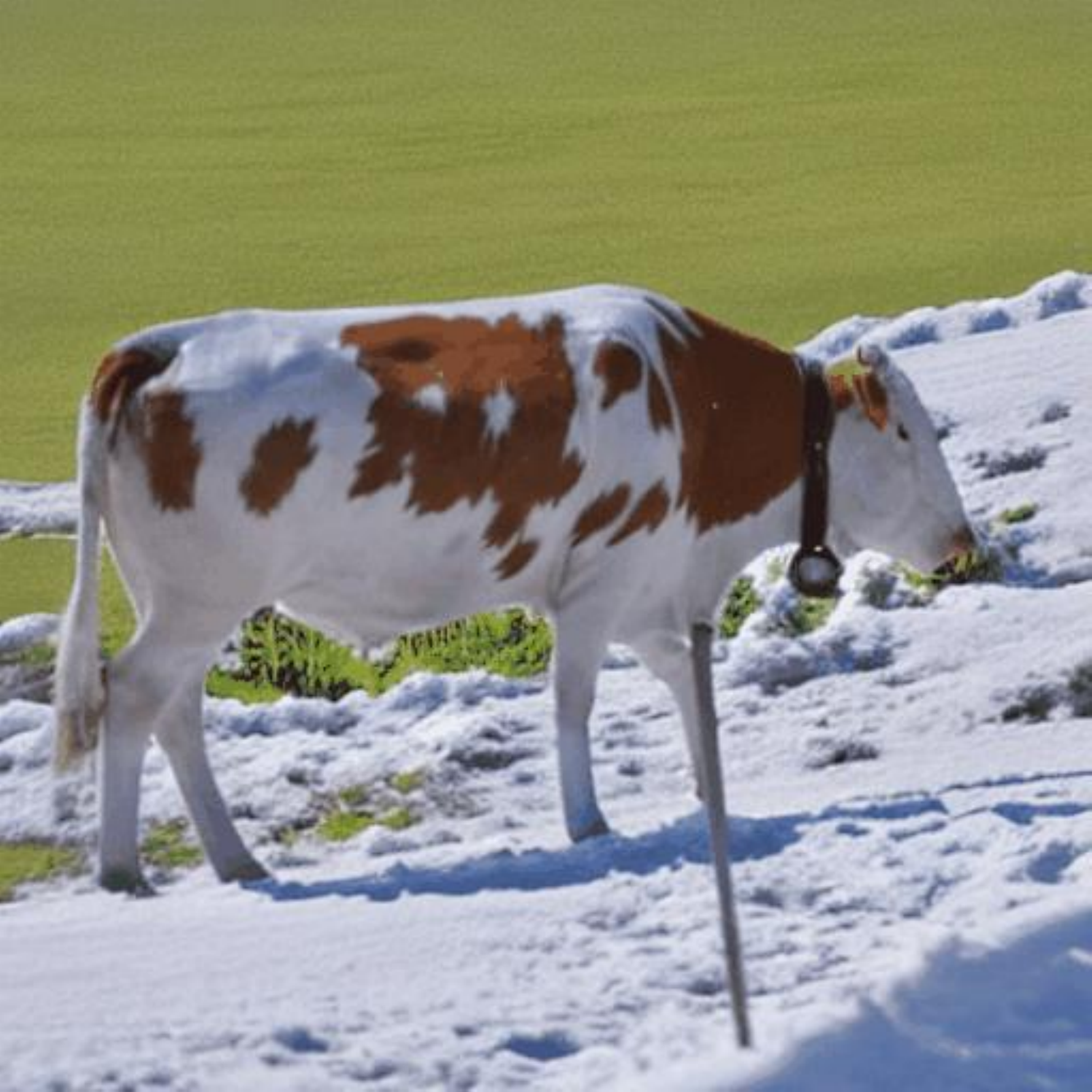}
\includegraphics[width=0.10\textwidth]{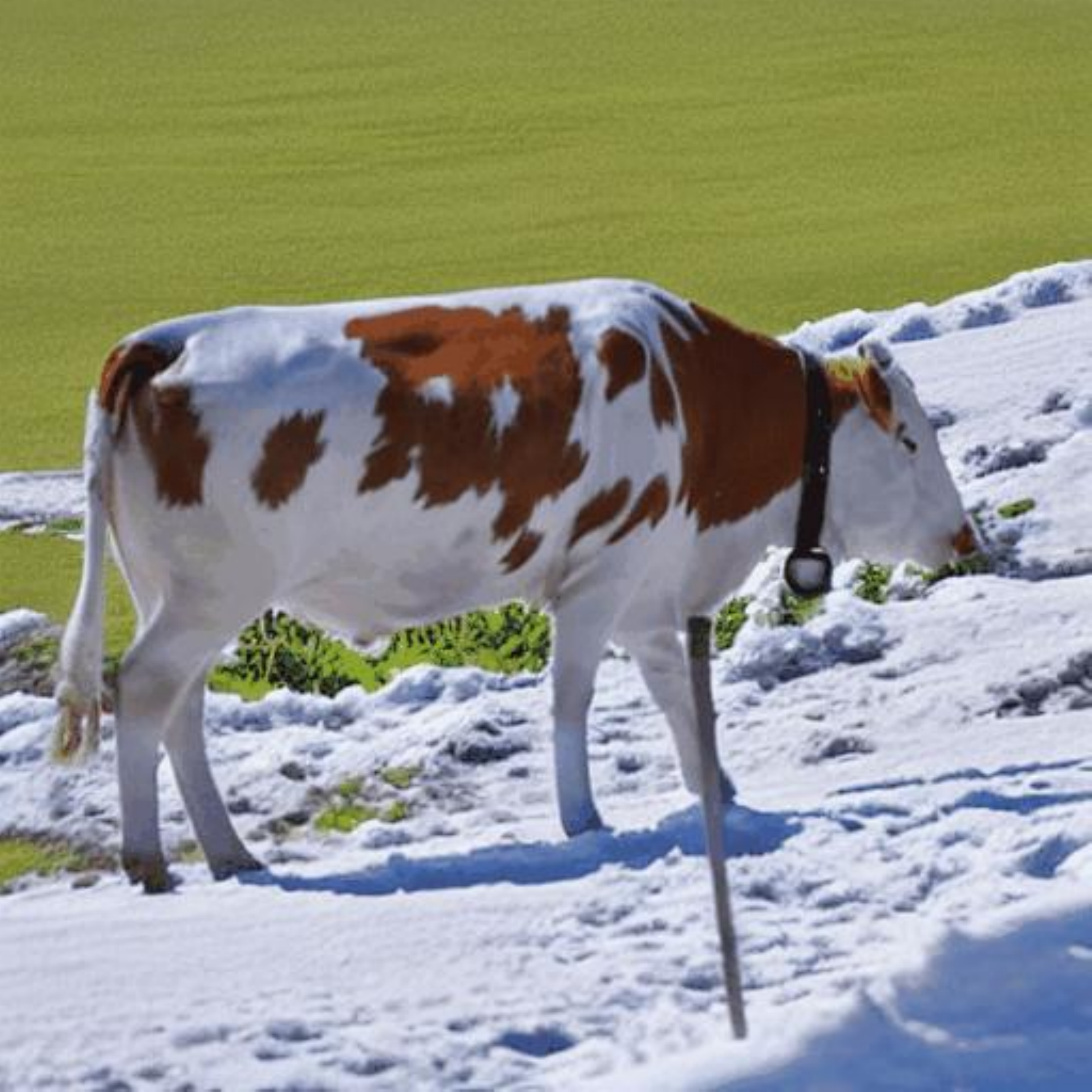}
\includegraphics[width=0.10\textwidth]{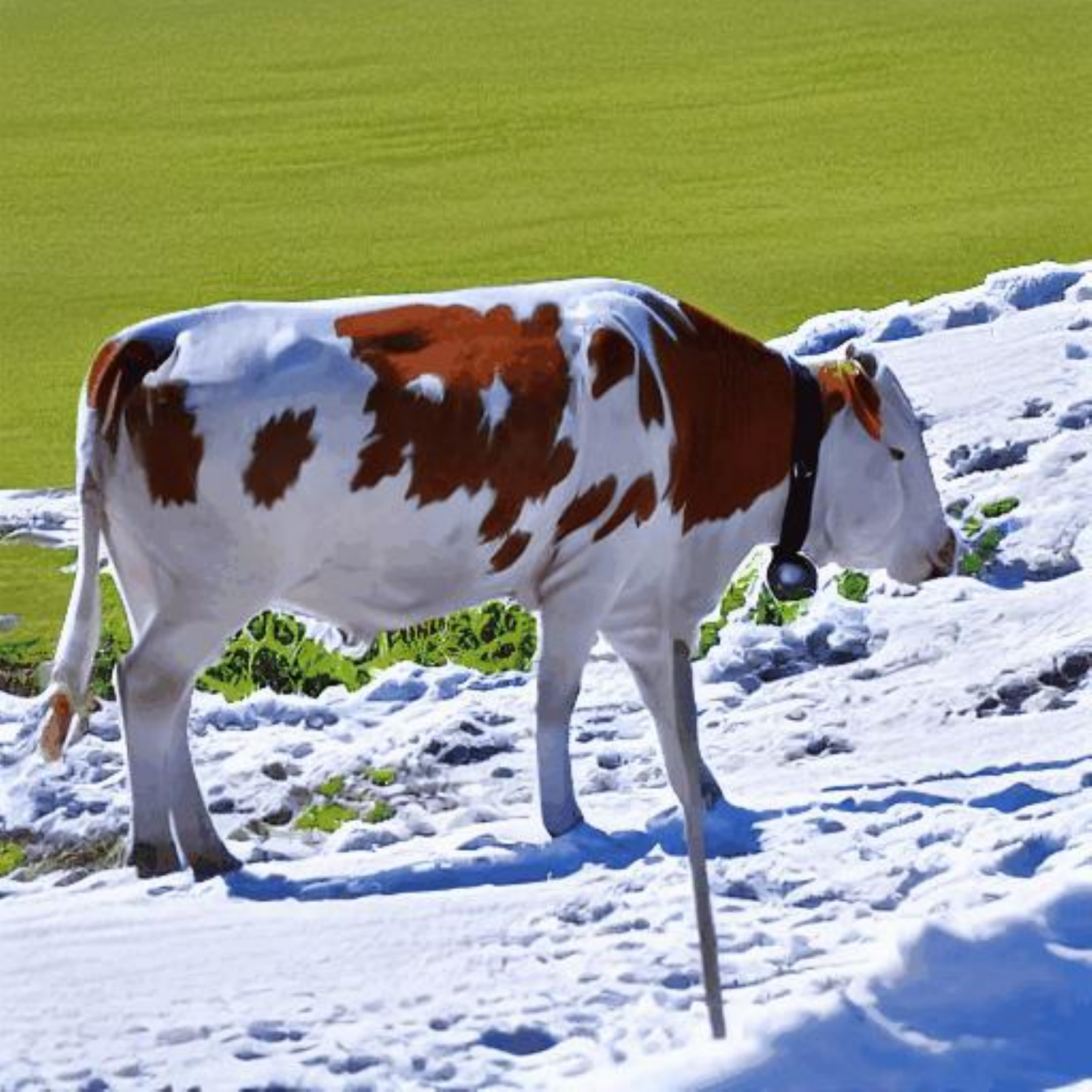}

\makebox[0.12\textwidth]{A cow is walking, \textcolor{blue}{\textbf{on the desert}}.}\\
\includegraphics[width=0.10\textwidth]{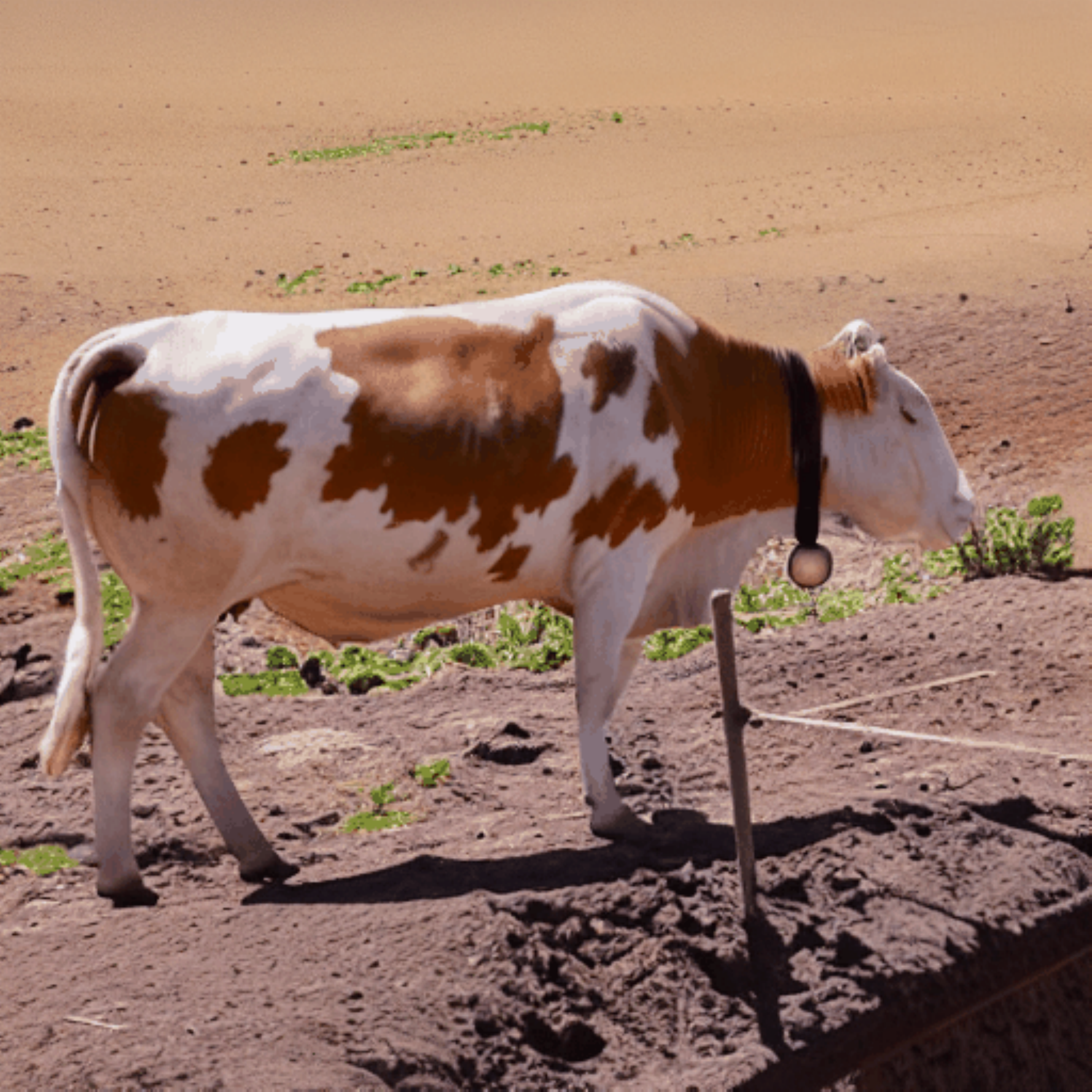}
\includegraphics[width=0.10\textwidth]{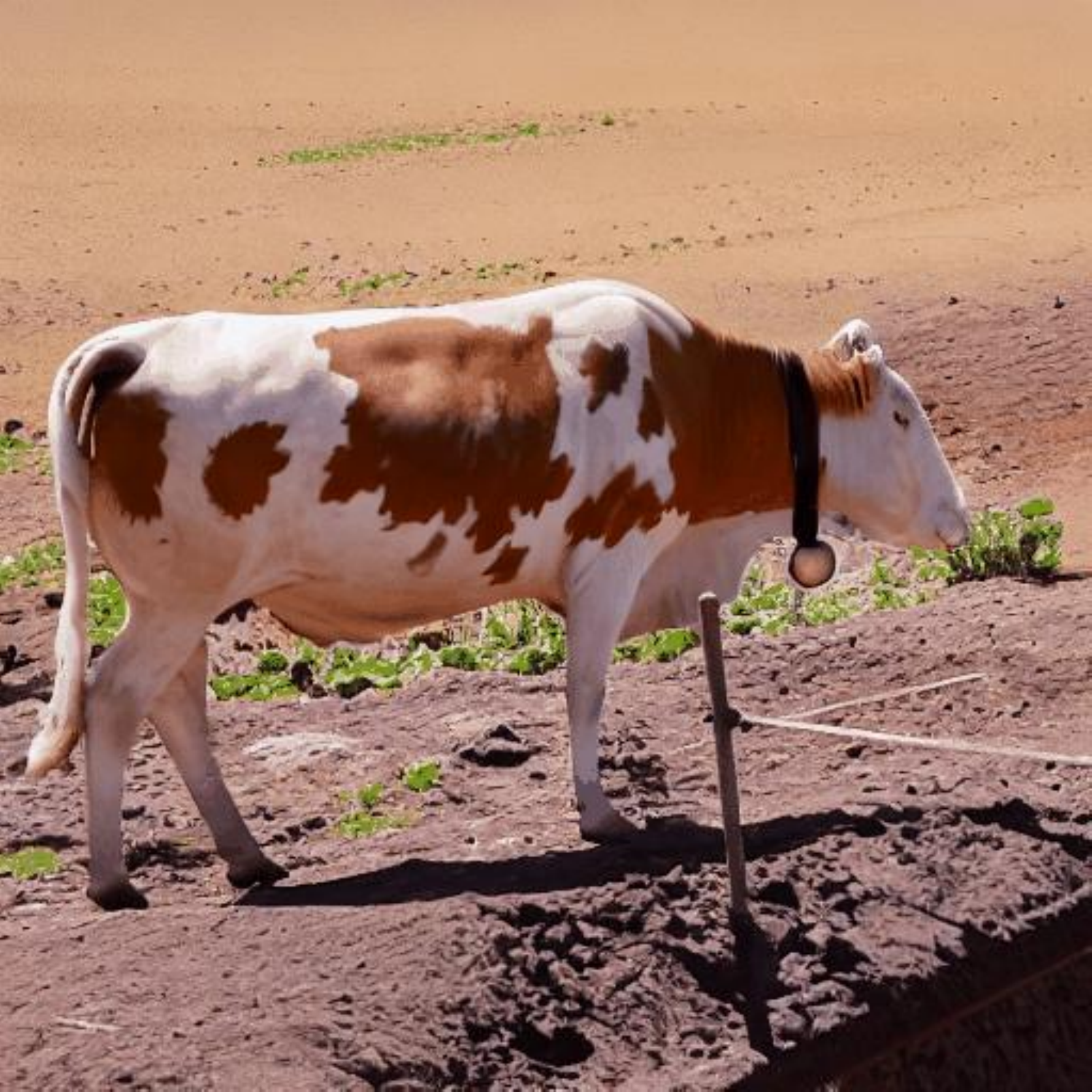}
\includegraphics[width=0.10\textwidth]{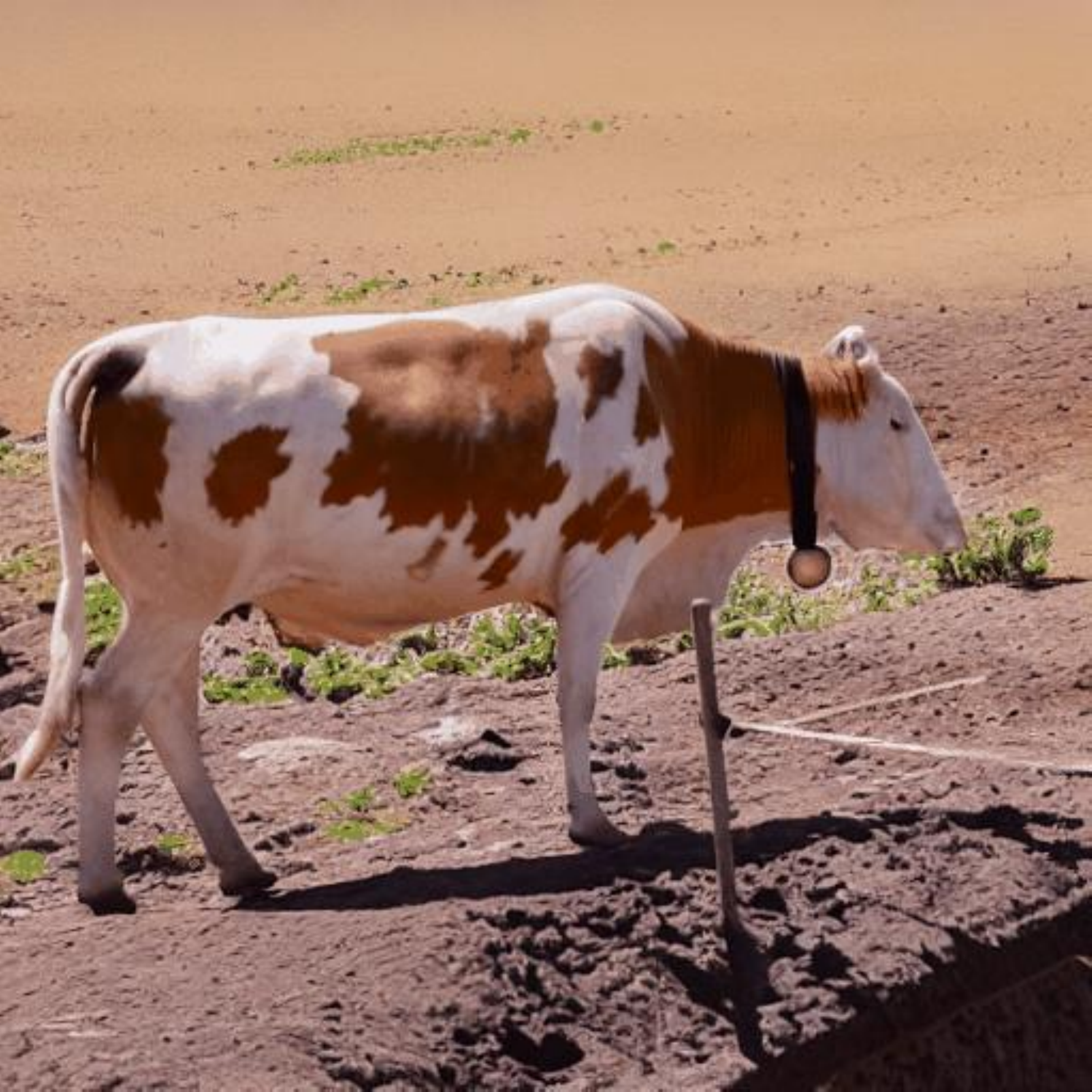}
\includegraphics[width=0.10\textwidth]{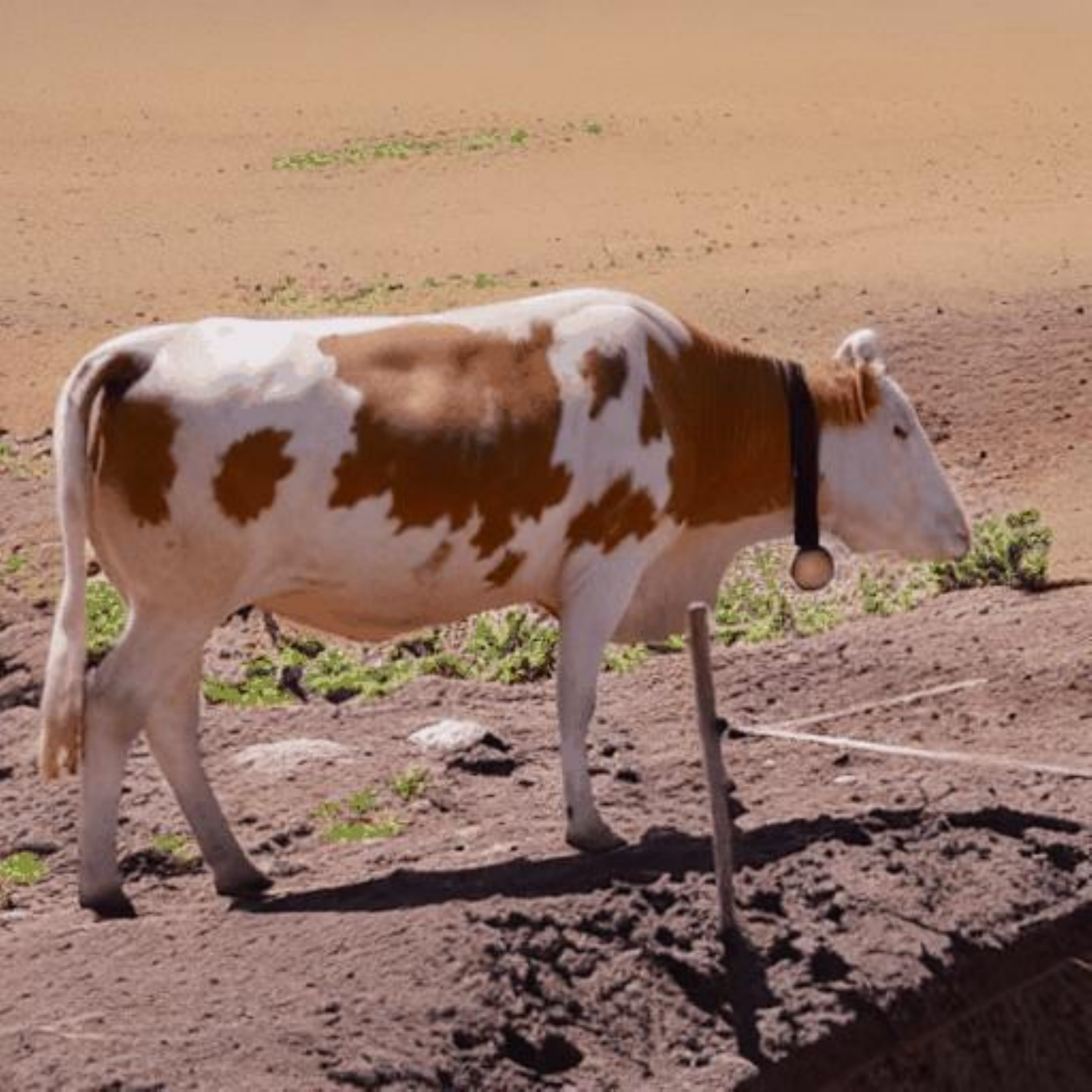}
\includegraphics[width=0.10\textwidth]{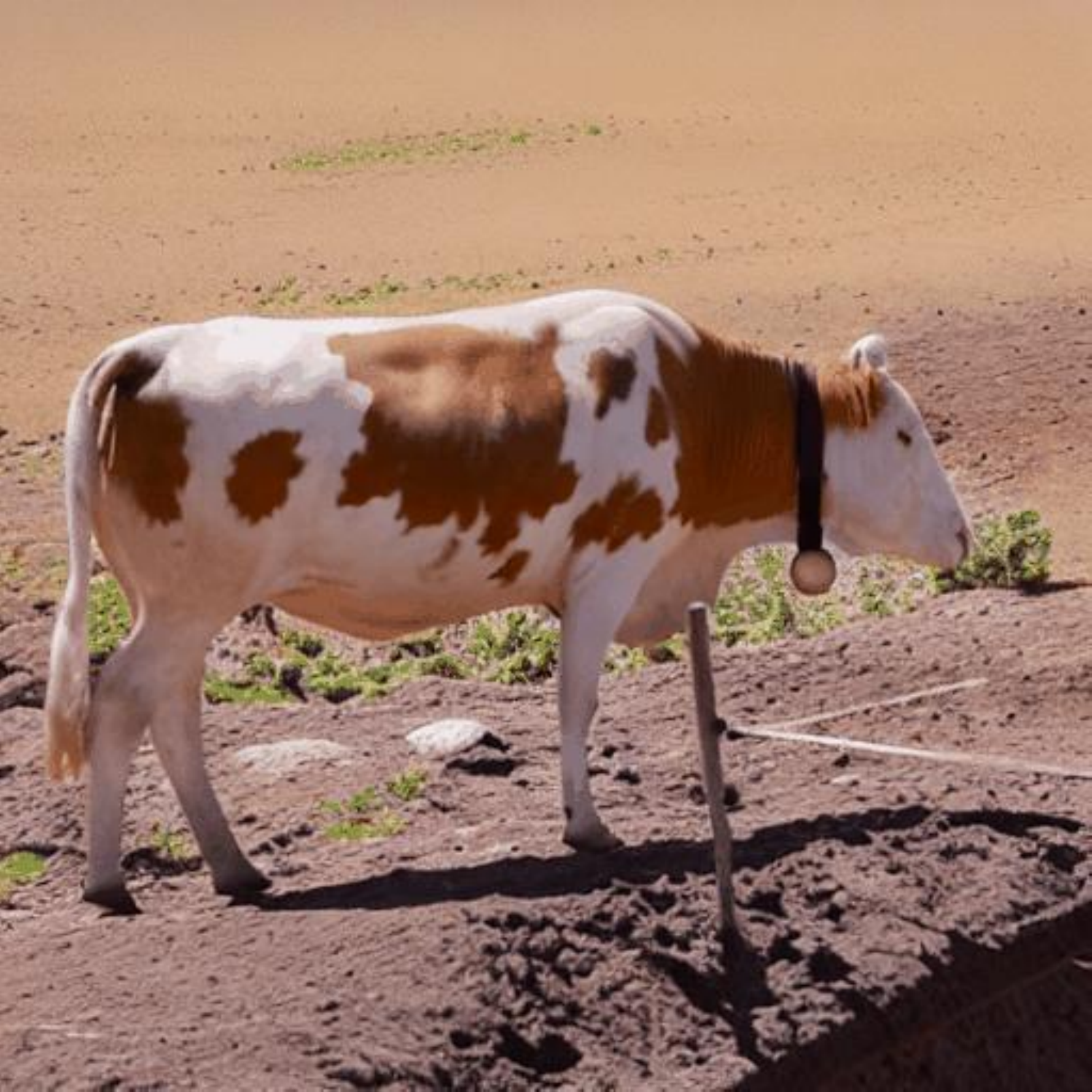}
\includegraphics[width=0.10\textwidth]{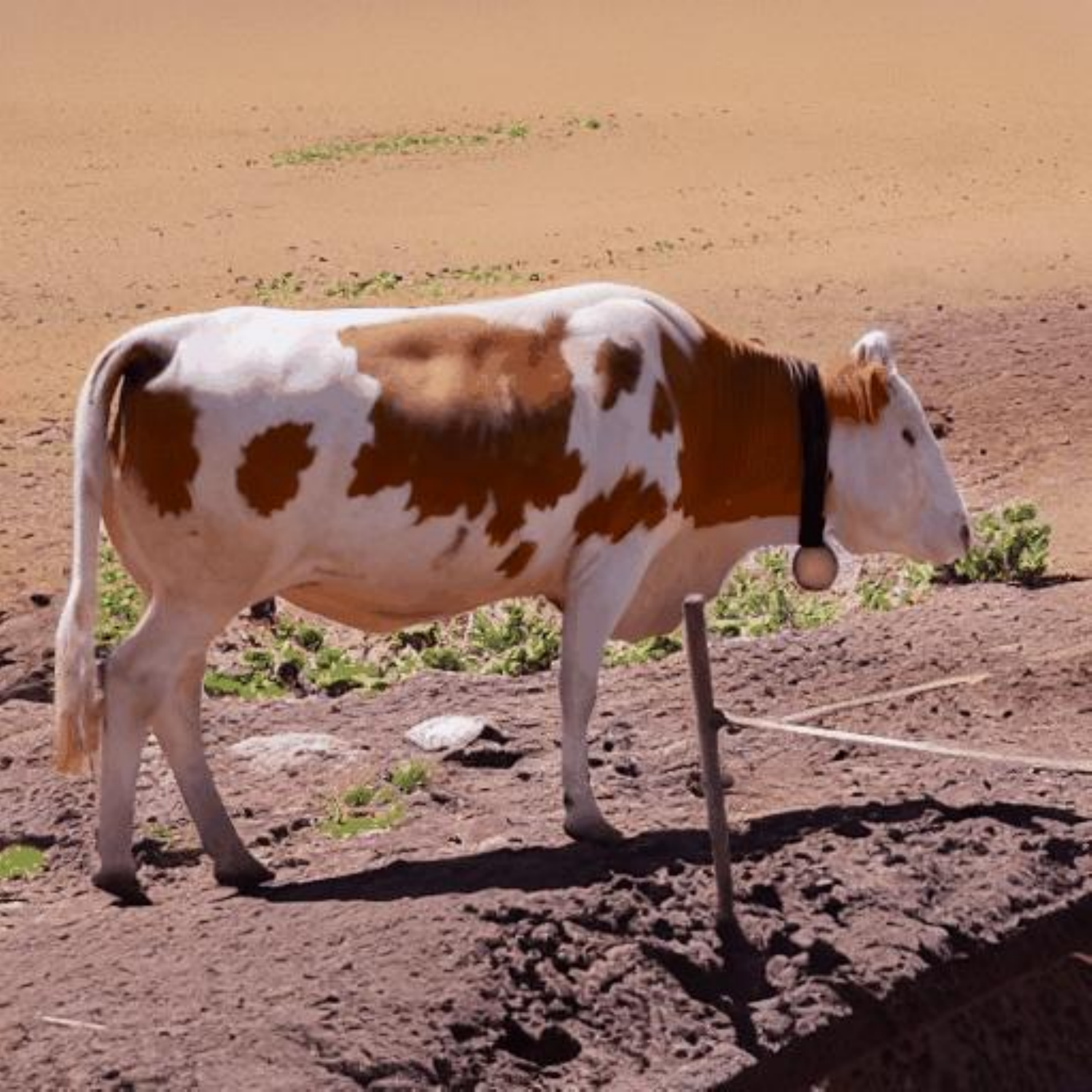}
\includegraphics[width=0.10\textwidth]{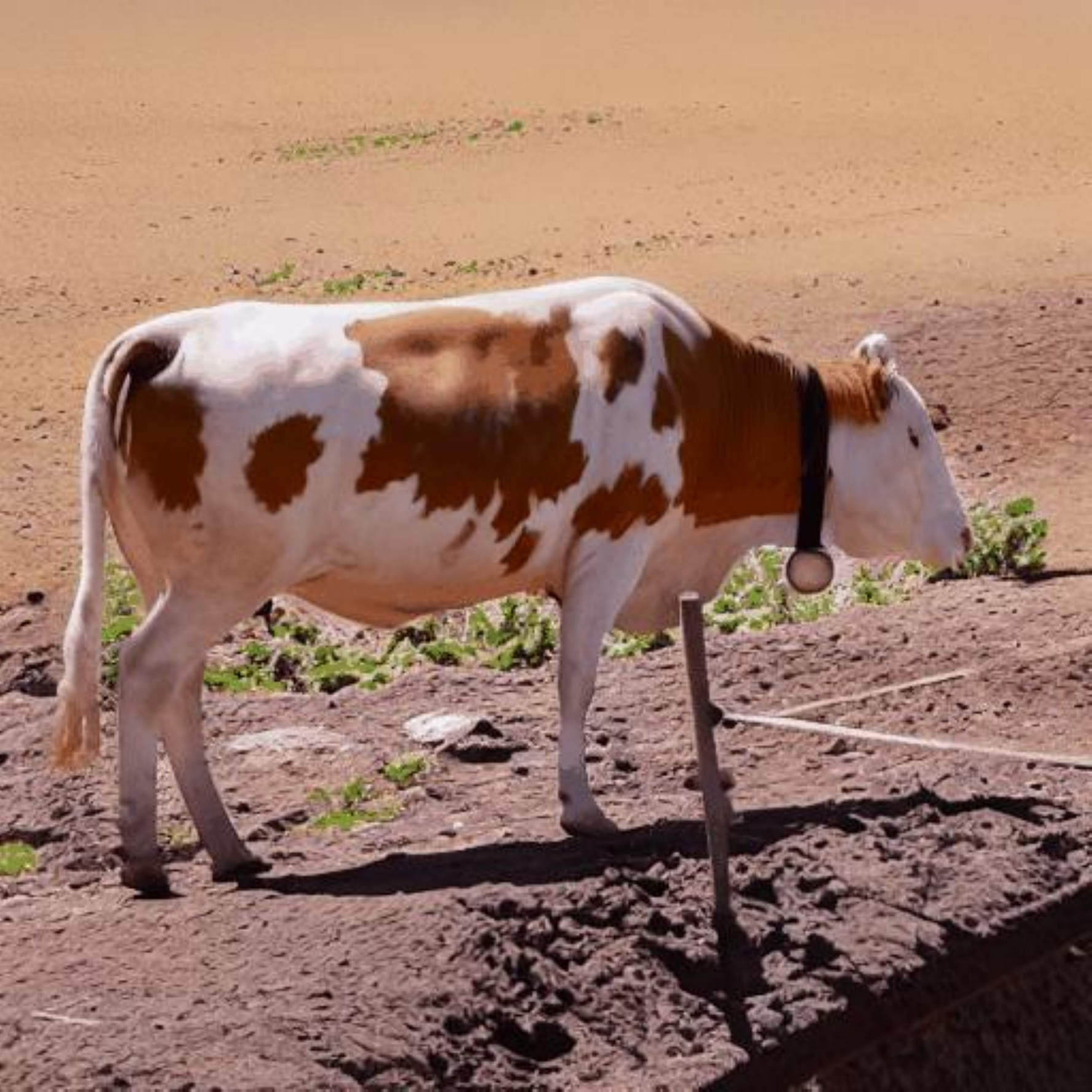}
\includegraphics[width=0.10\textwidth]{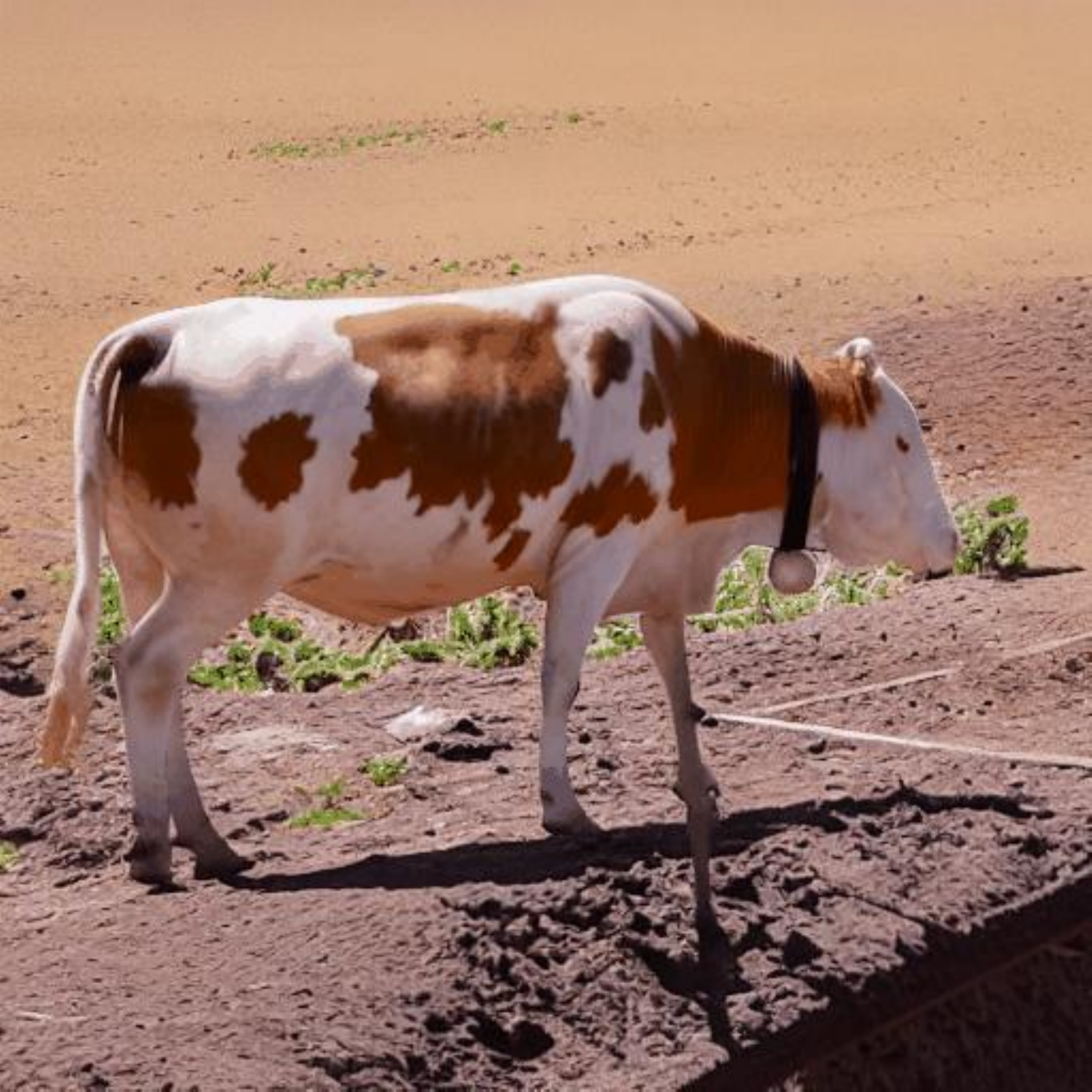}
\caption{\textbf{Qualitative Results} Additional selected samples for our model.}
\label{fig:supp_qual3}
\end{center}
\end{figure*}

\begin{figure*}
\vspace{0.8em}
\begin{center}
\makebox[0.12\textwidth]{\colorbox{pink}{\textbf{Training video}} A man is boxing}\\

\includegraphics[width=0.10\textwidth]{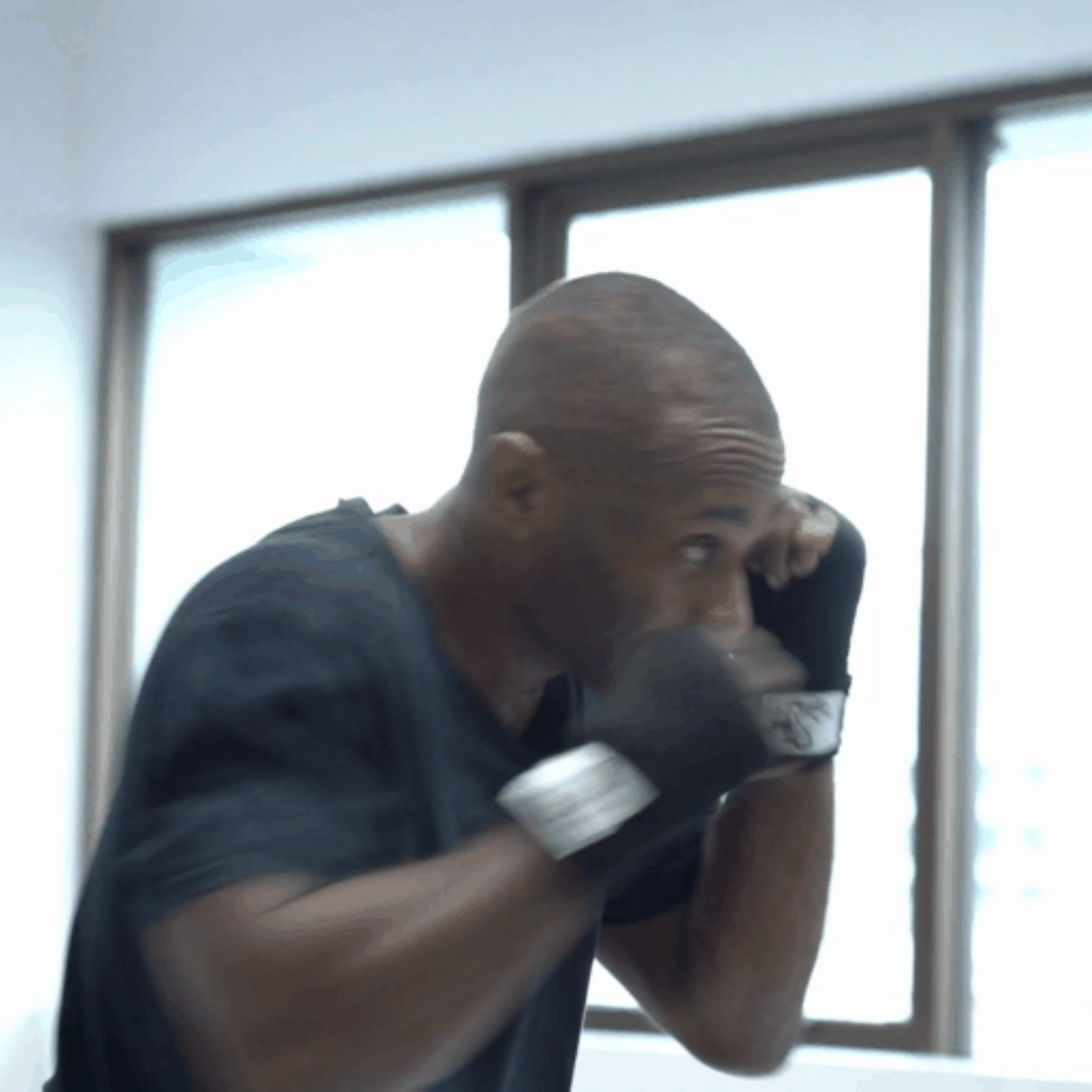}
\includegraphics[width=0.10\textwidth]{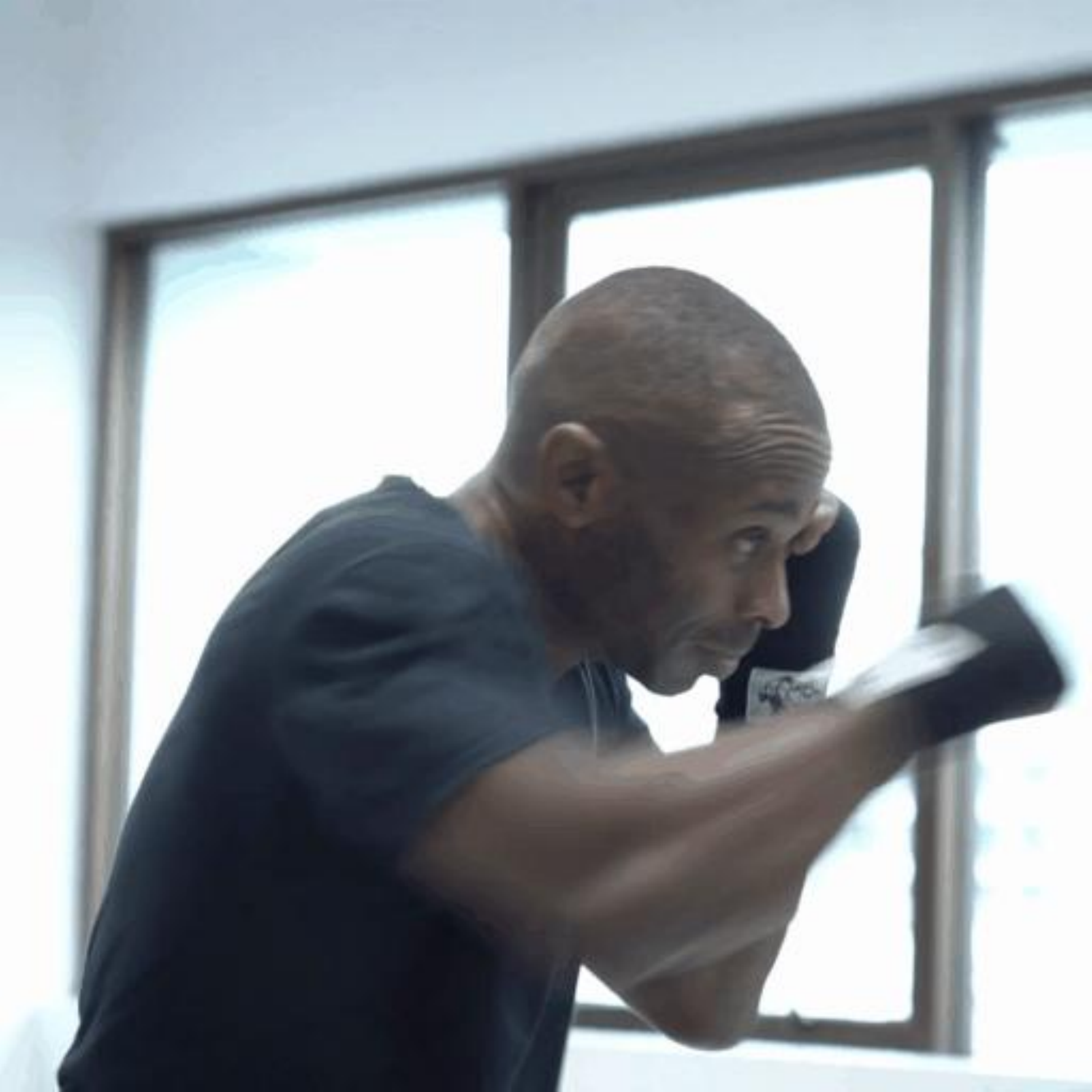}
\includegraphics[width=0.10\textwidth]{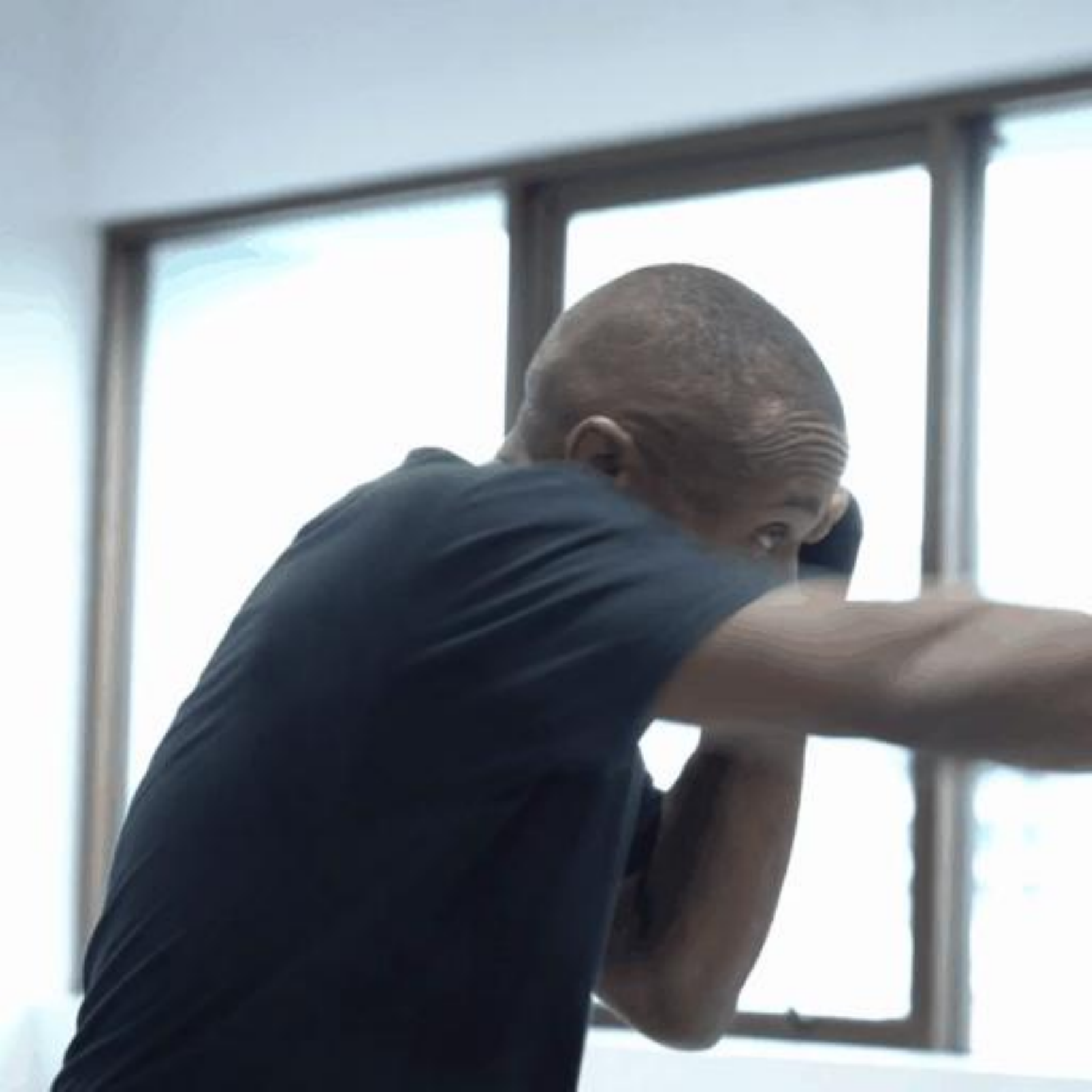}
\includegraphics[width=0.10\textwidth]{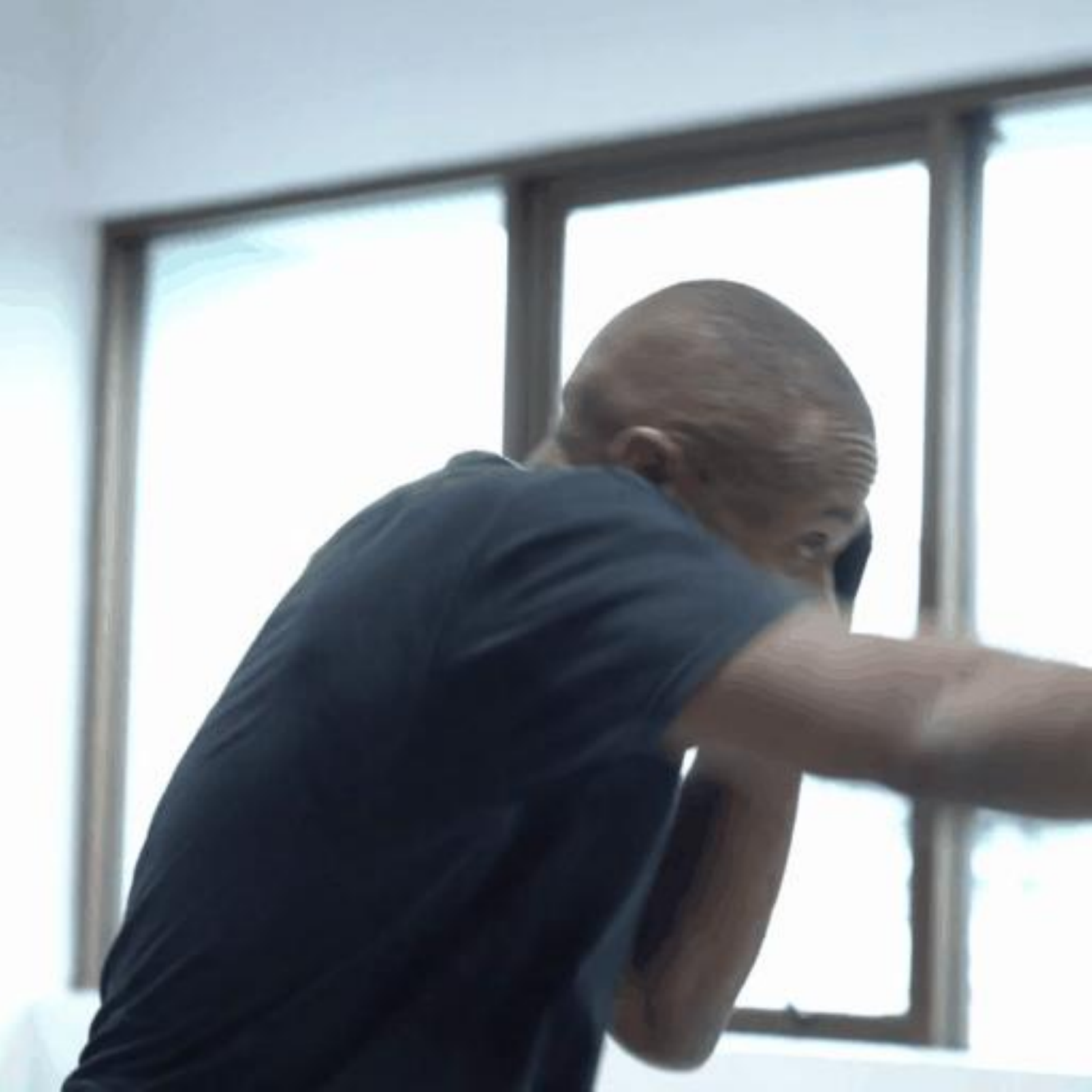}
\includegraphics[width=0.10\textwidth]{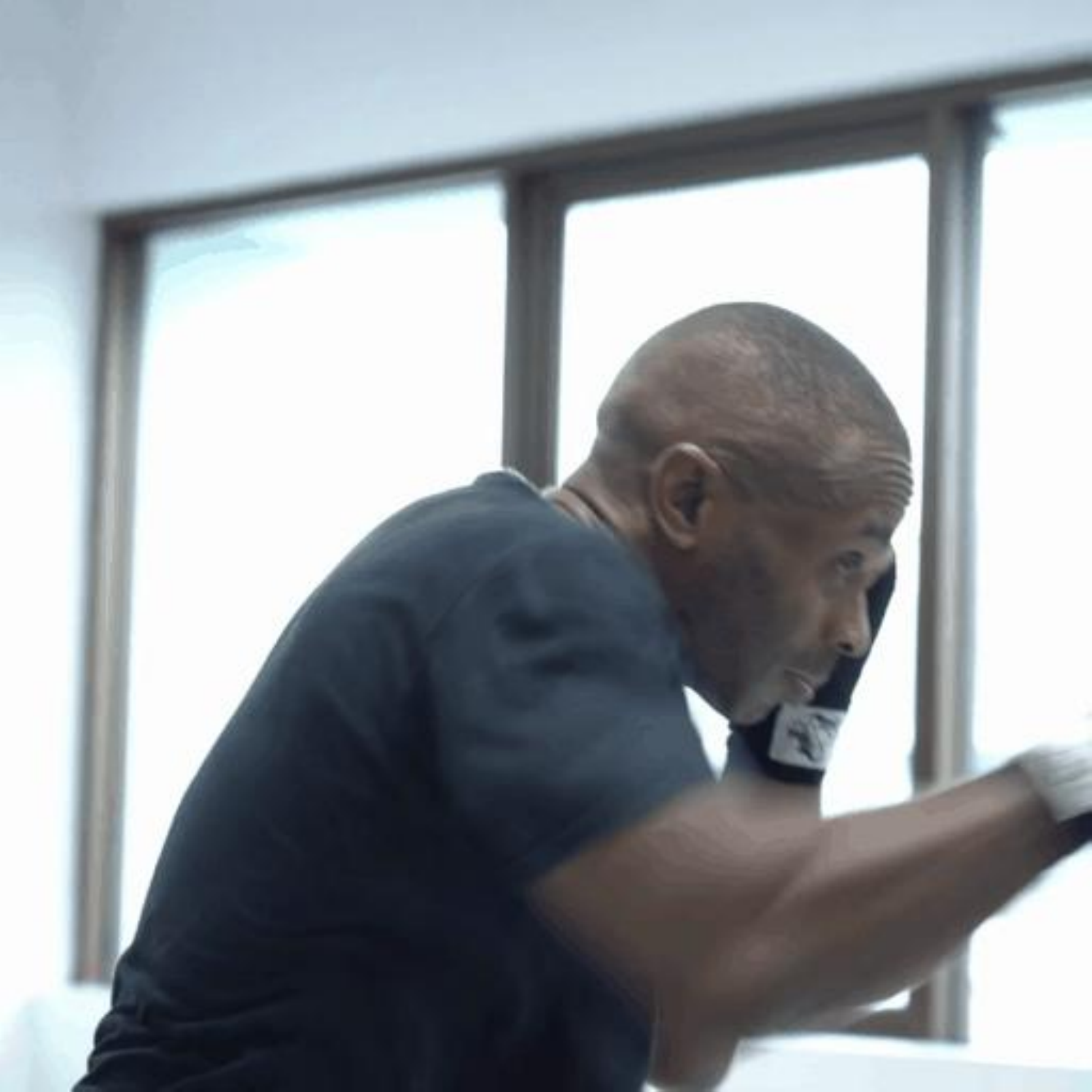}
\includegraphics[width=0.10\textwidth]{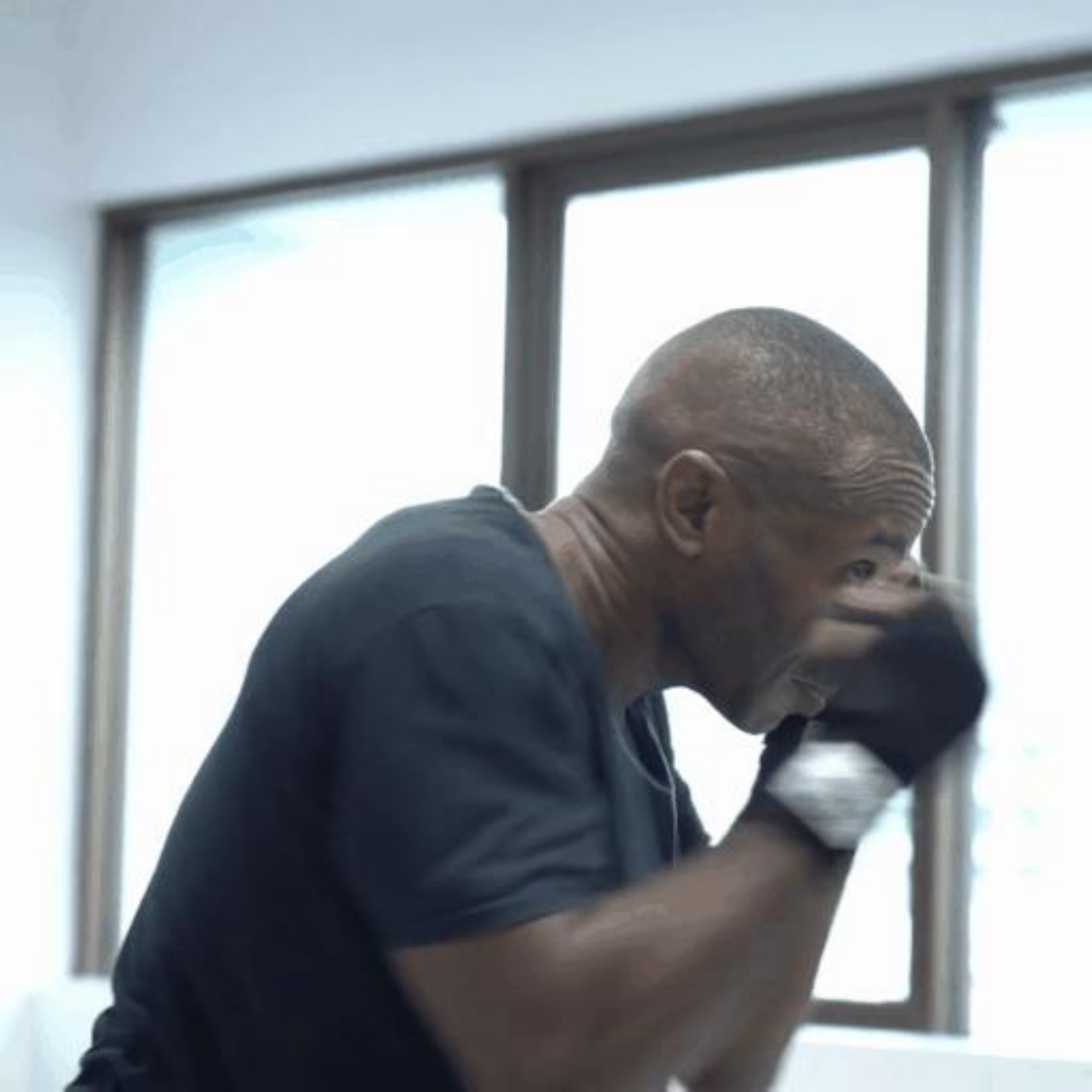}
\includegraphics[width=0.10\textwidth]{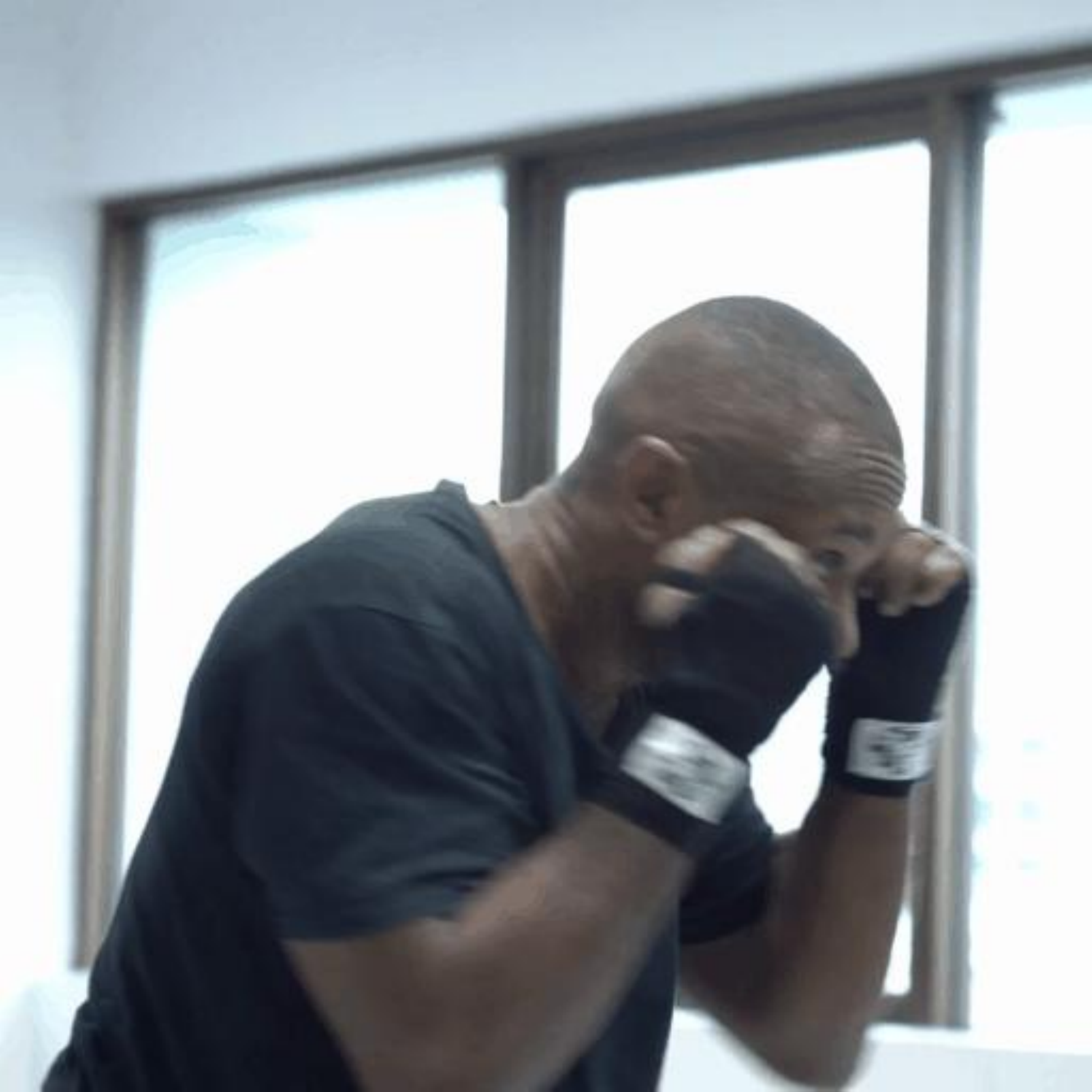}
\includegraphics[width=0.10\textwidth]{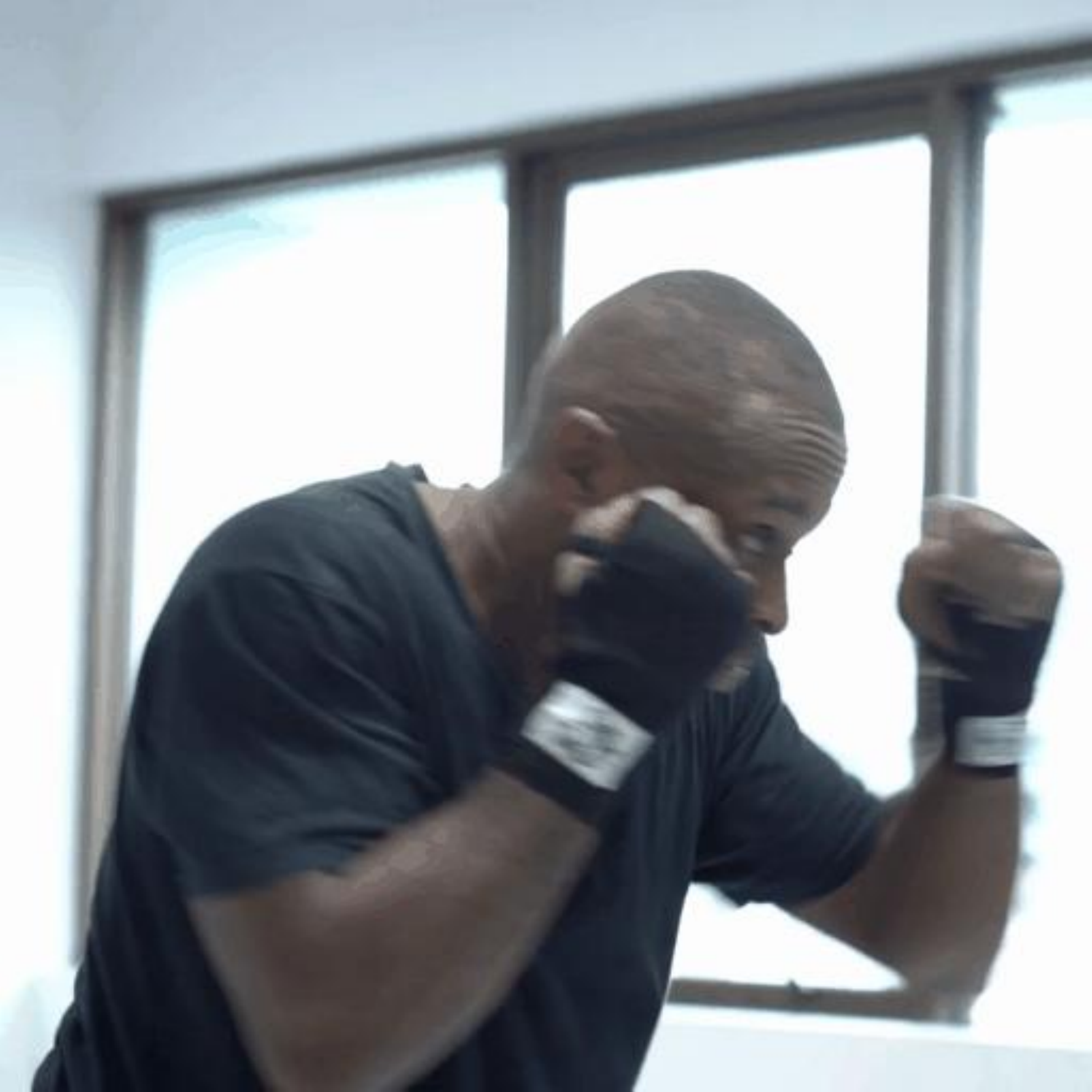}

\makebox[0.12\textwidth]{\colorbox{yellow}{\textbf{Edit-A-Video (Ours)}} A \textcolor{blue}{\textbf{Bat Man}} is boxing}\\

\includegraphics[width=0.10\textwidth]{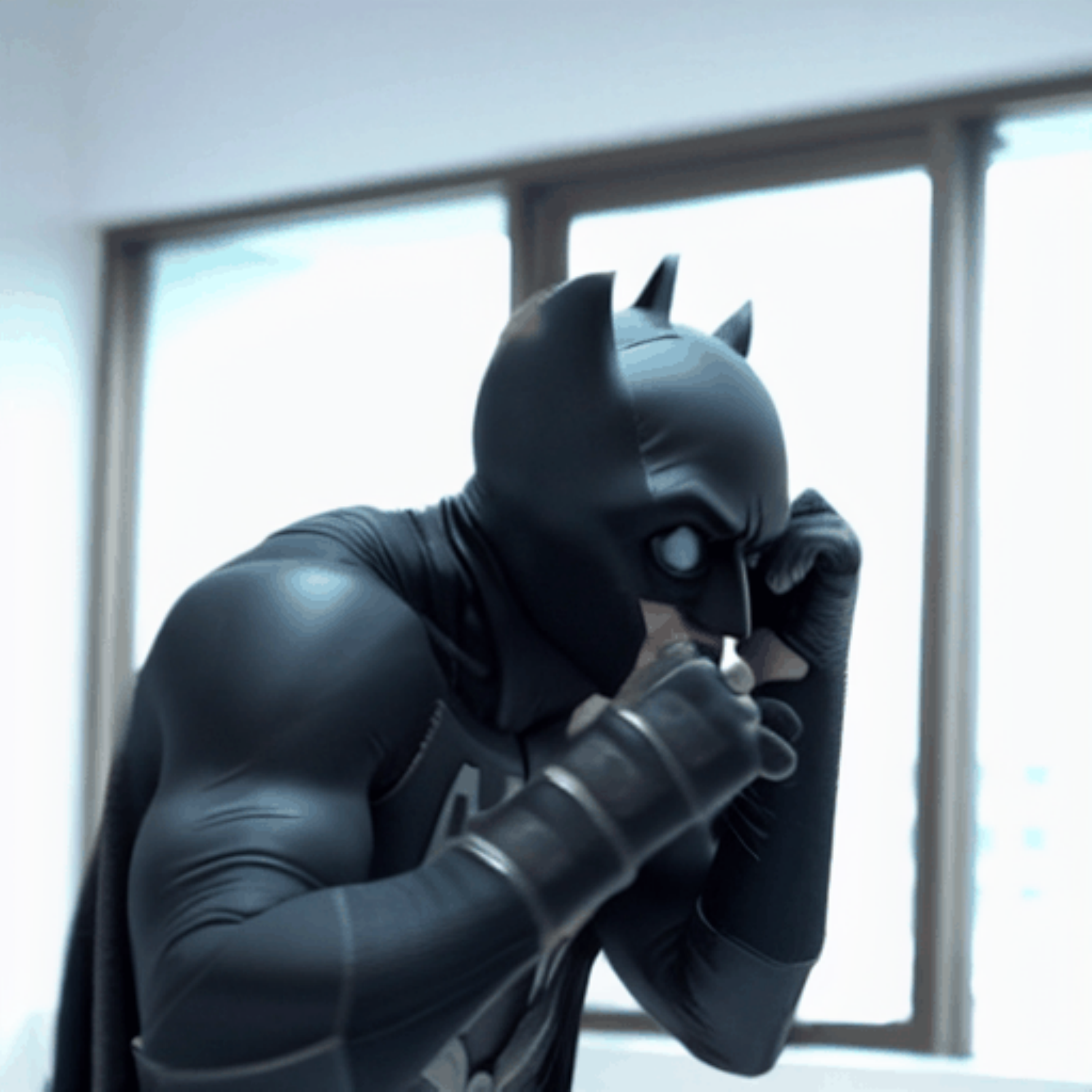}
\includegraphics[width=0.10\textwidth]{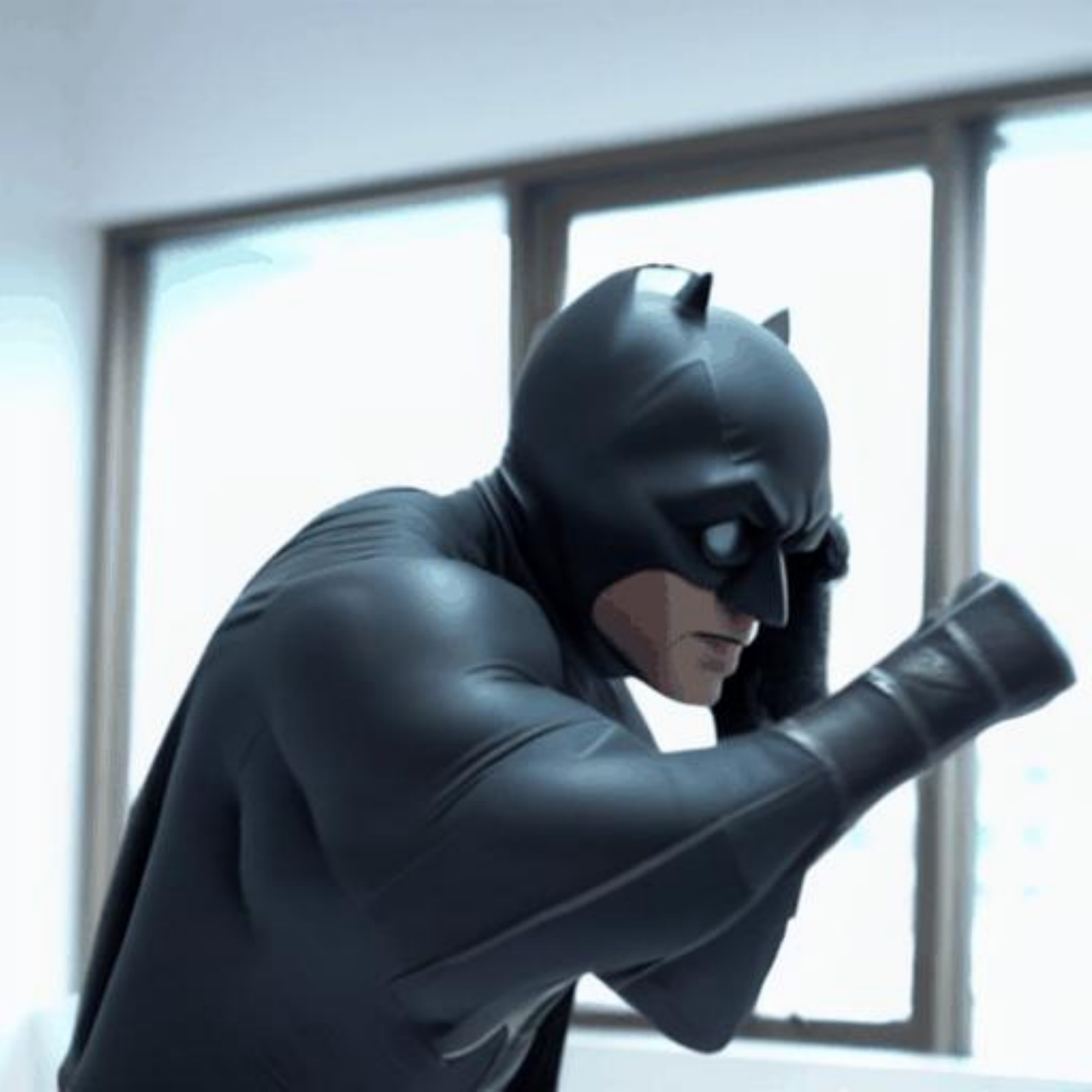}
\includegraphics[width=0.10\textwidth]{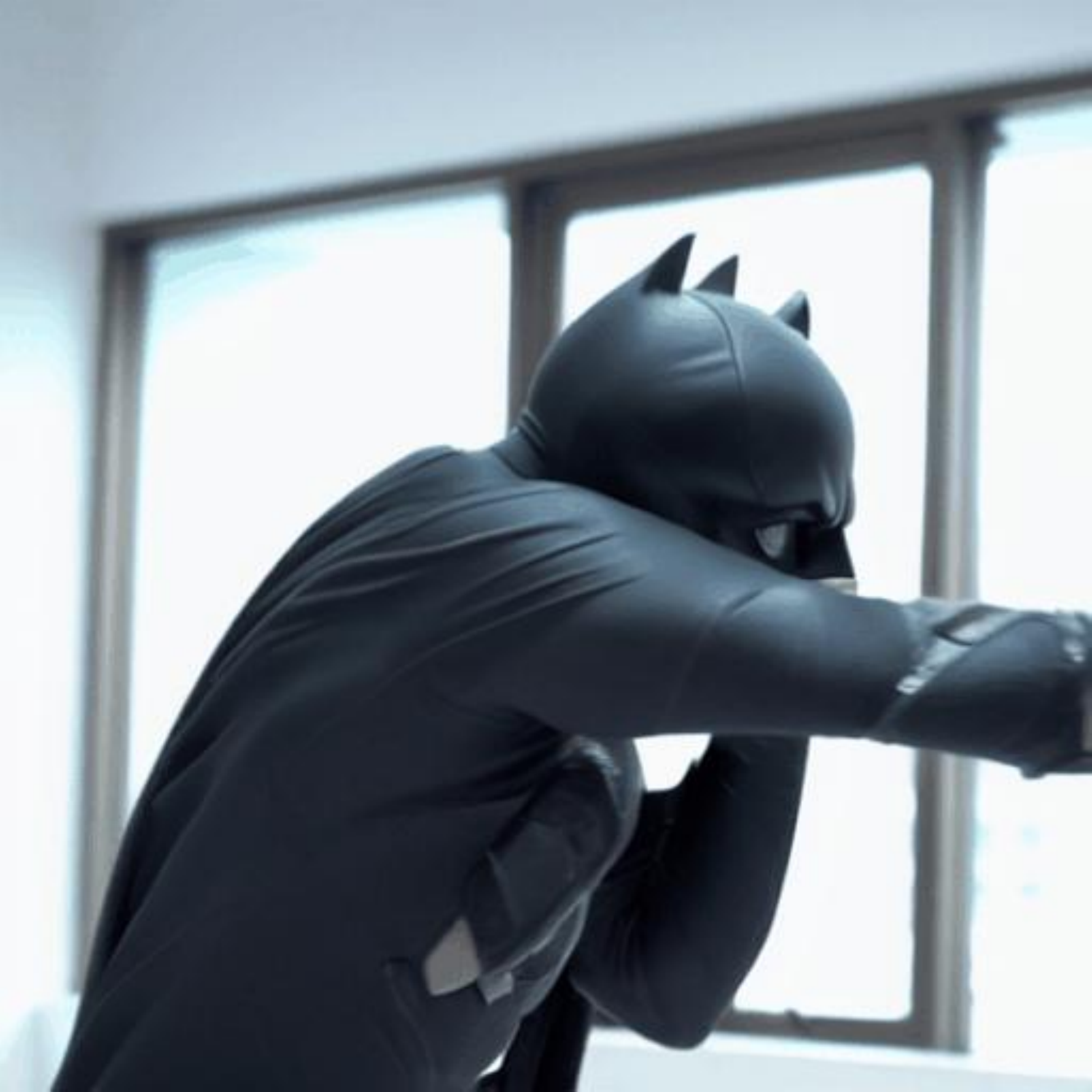}
\includegraphics[width=0.10\textwidth]{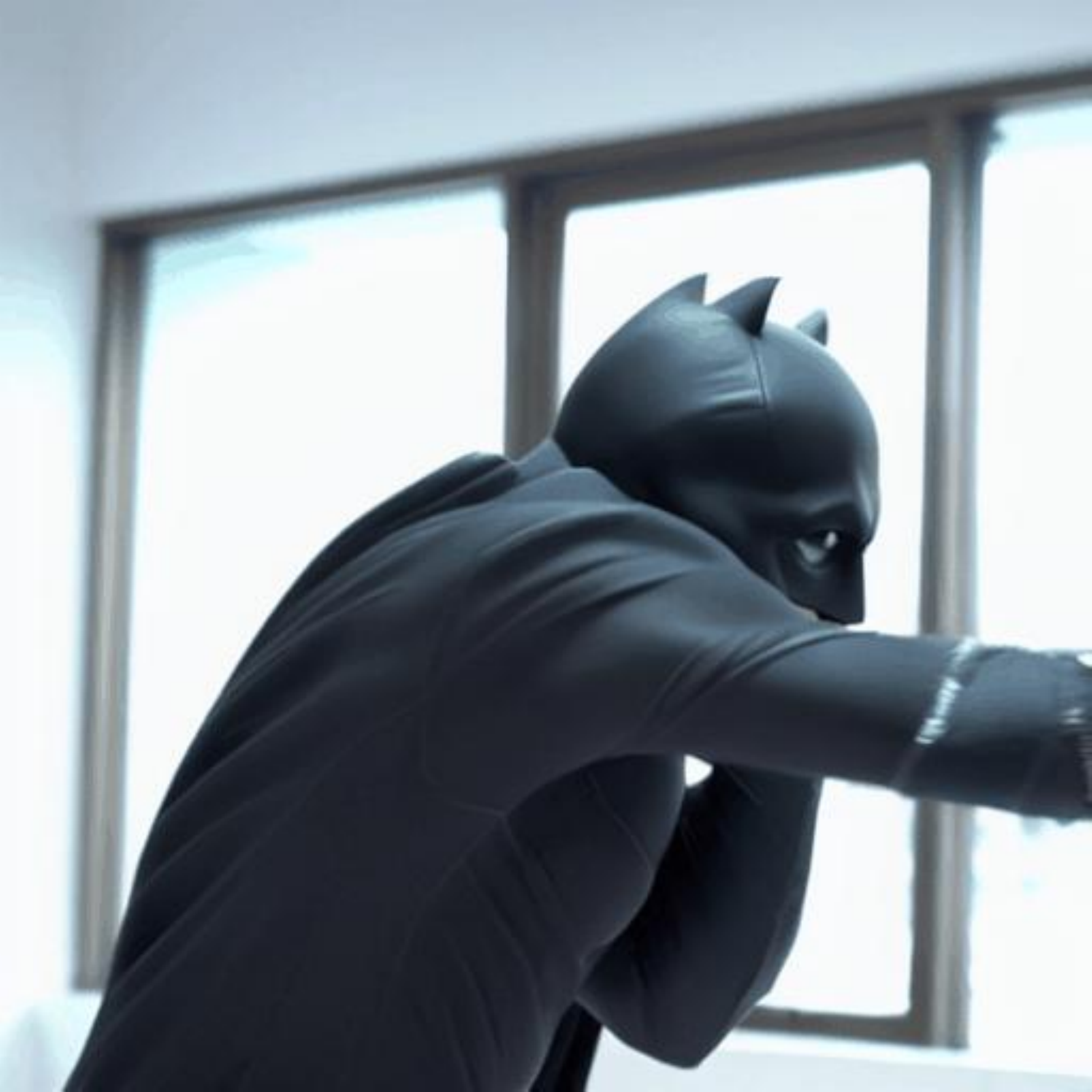}
\includegraphics[width=0.10\textwidth]{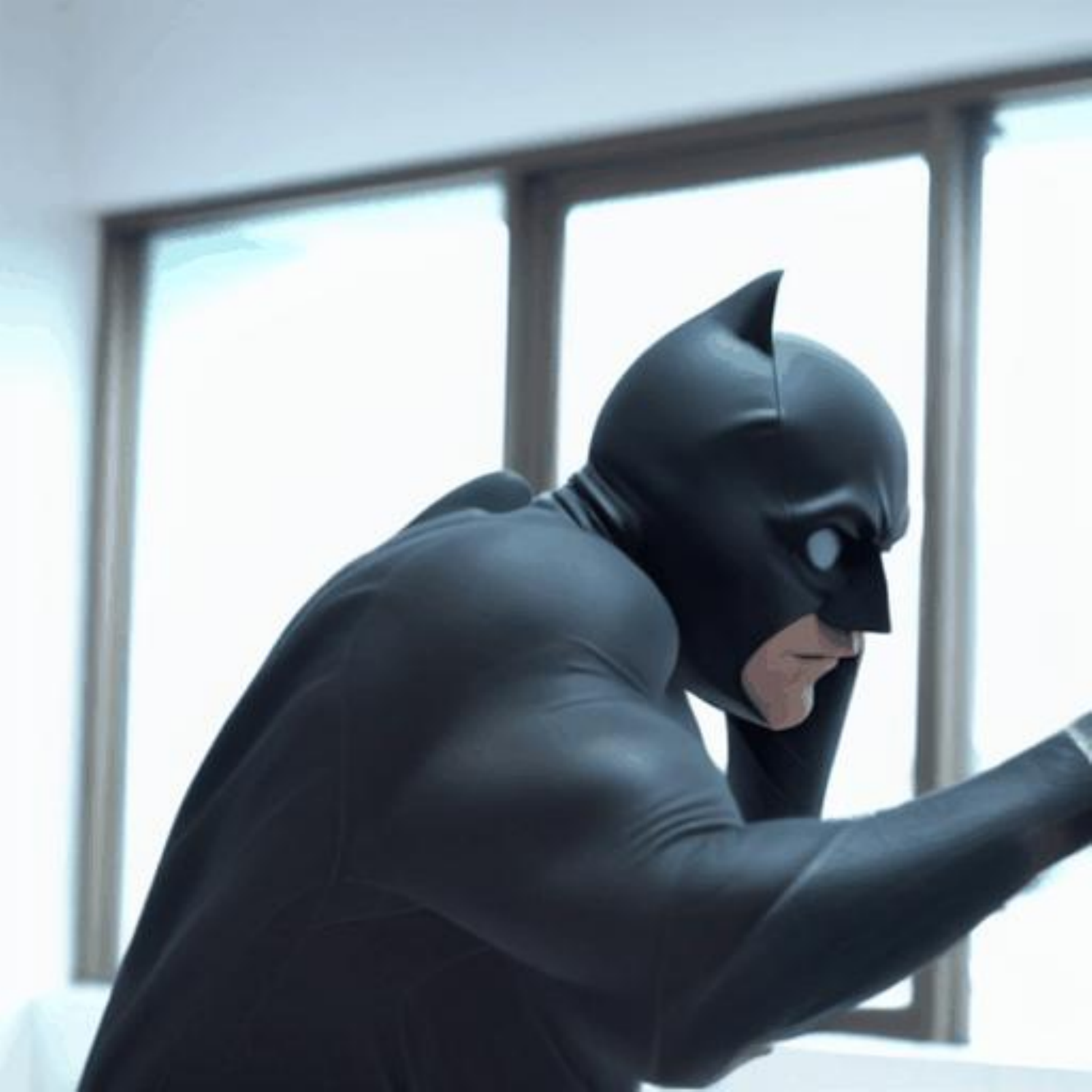}
\includegraphics[width=0.10\textwidth]{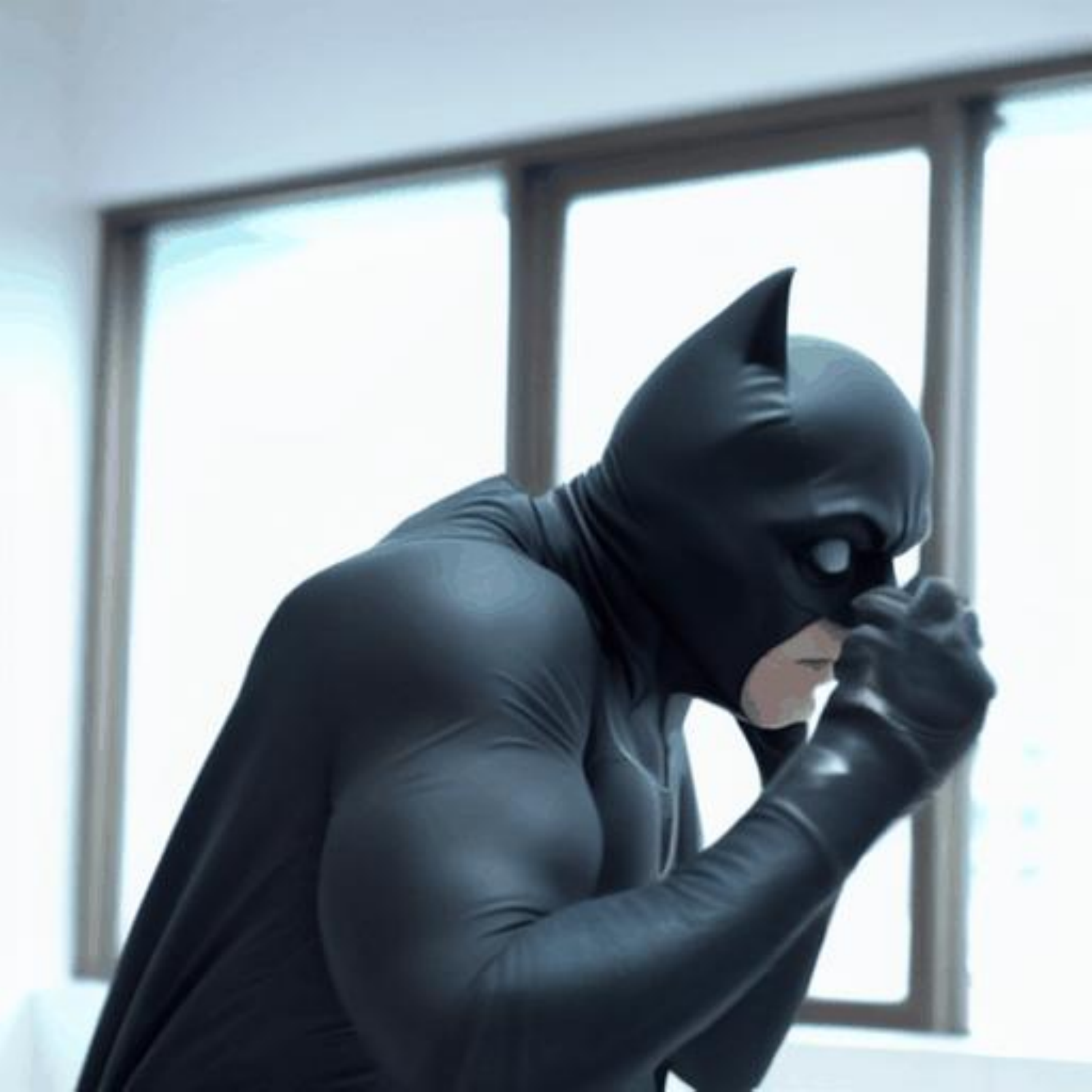}
\includegraphics[width=0.10\textwidth]{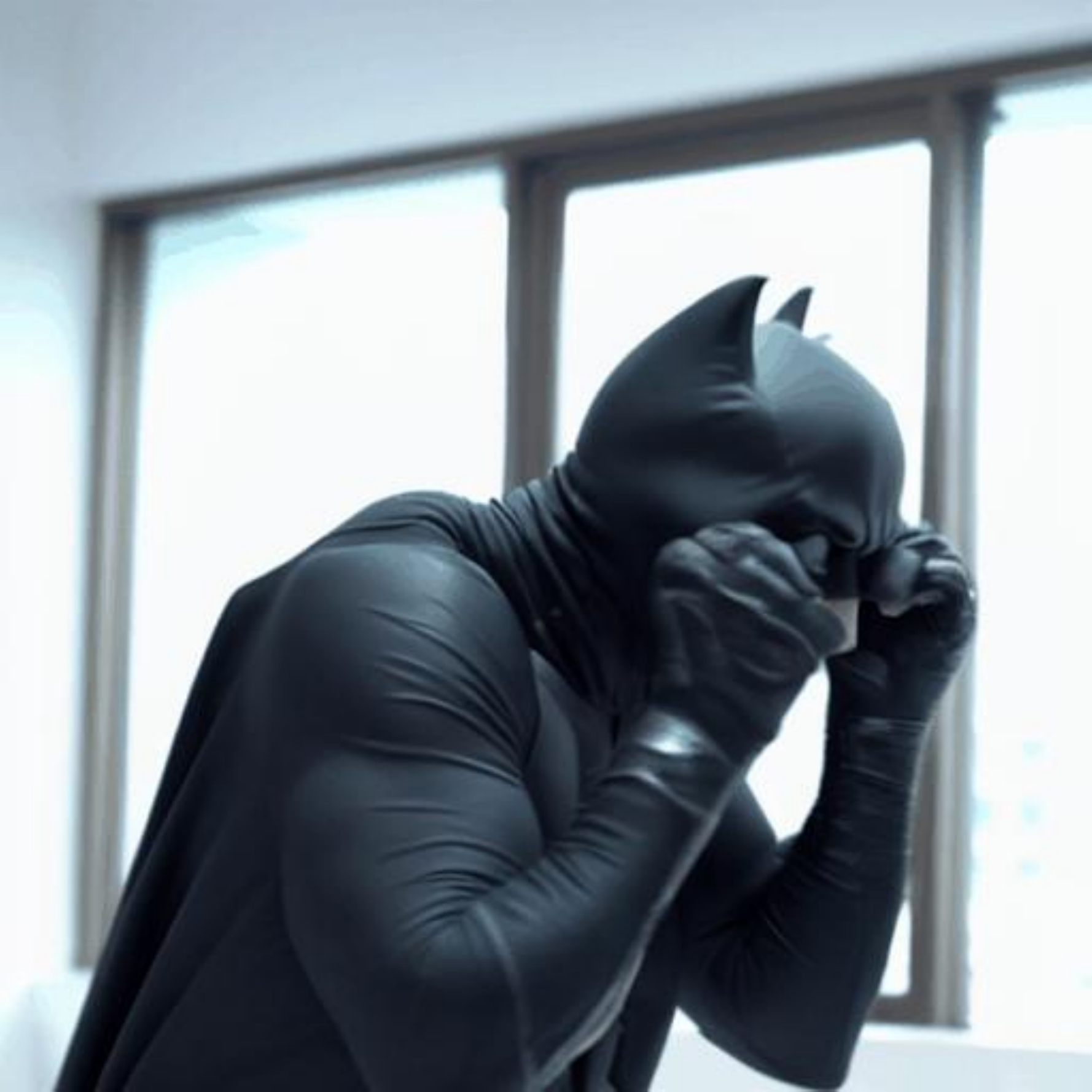}
\includegraphics[width=0.10\textwidth]{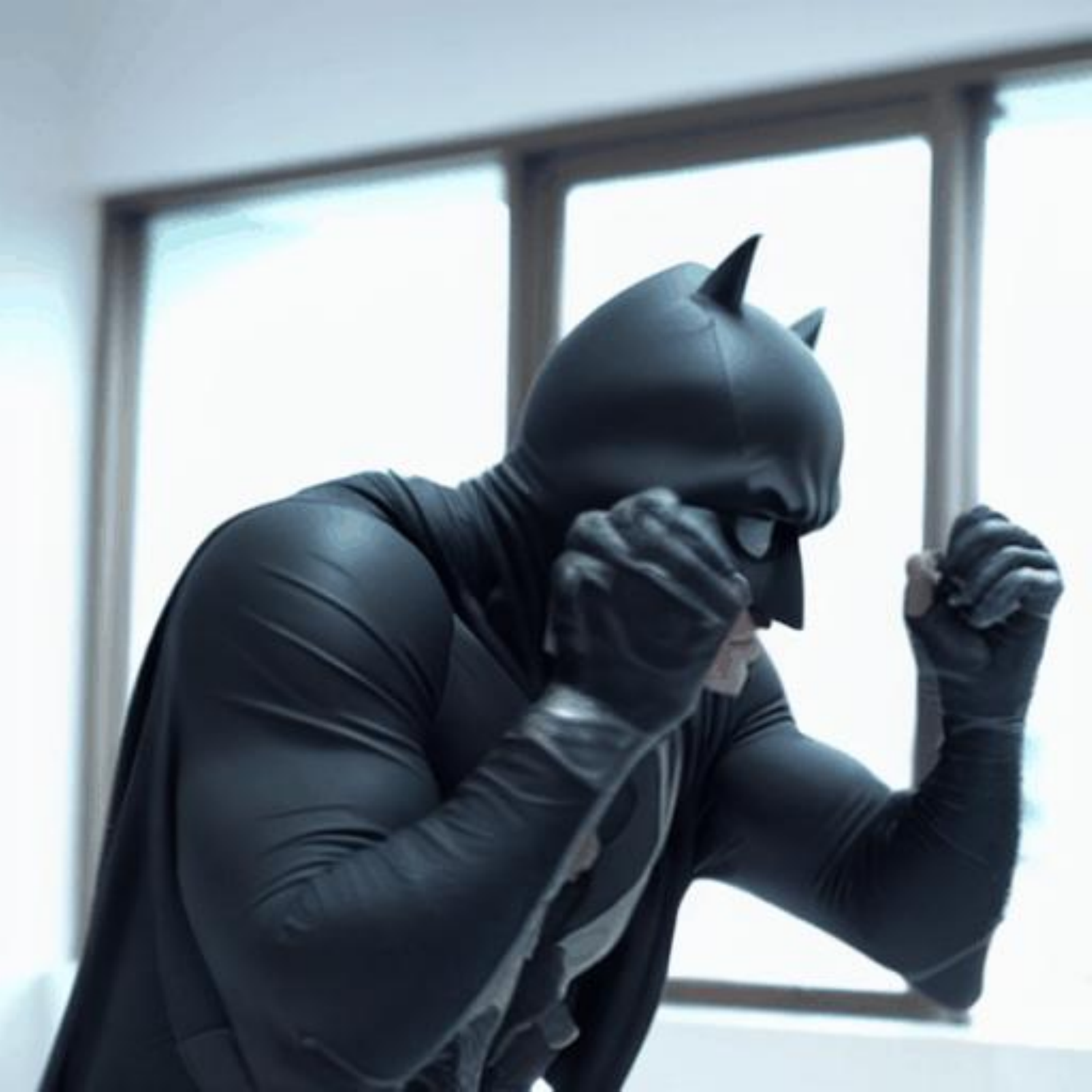}

\makebox[0.12\textwidth]{\colorbox{green}{\textbf{Tune-A-Video}} A \textcolor{blue}{\textbf{Bat Man}} is boxing}\\
\includegraphics[width=0.10\textwidth]{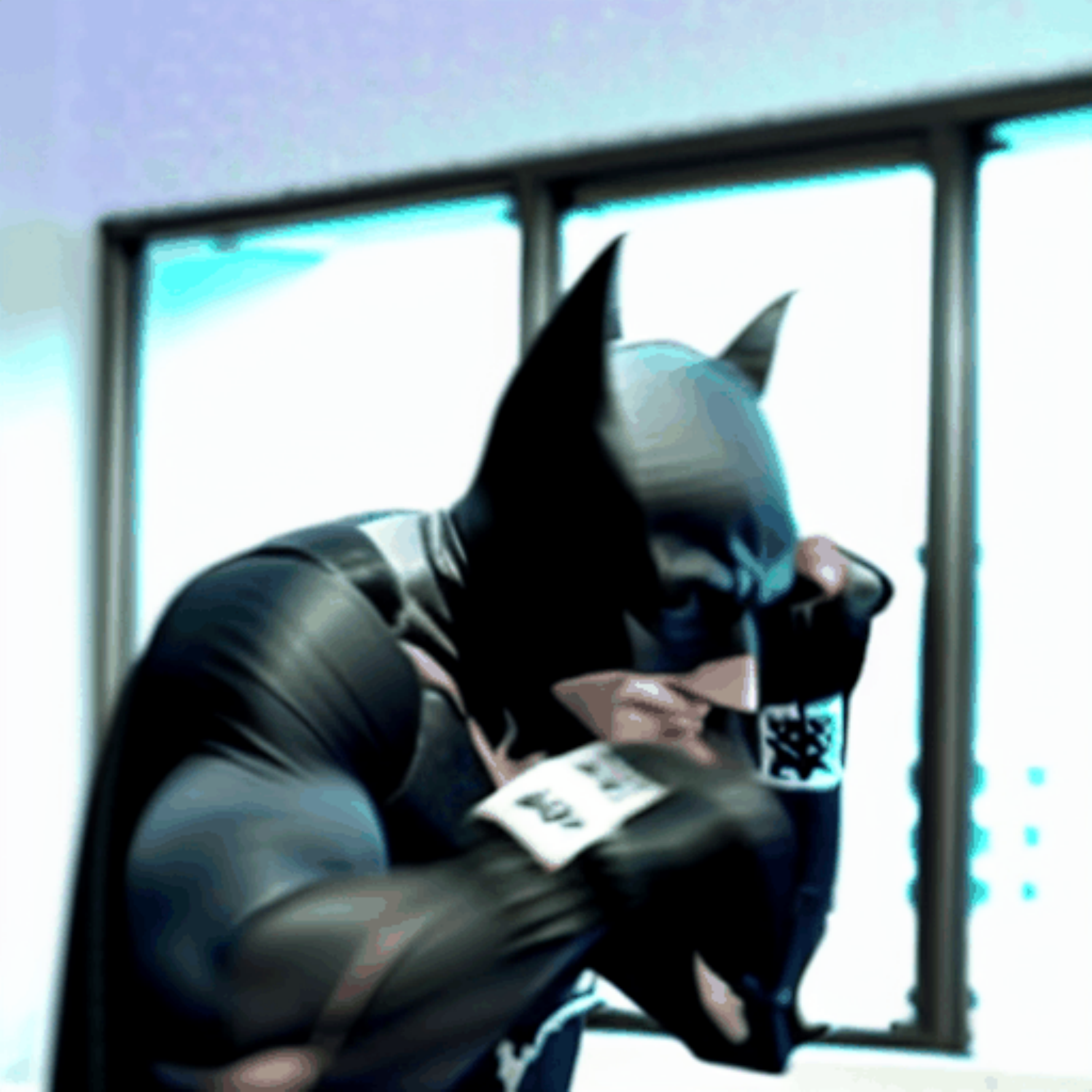}
\includegraphics[width=0.10\textwidth]{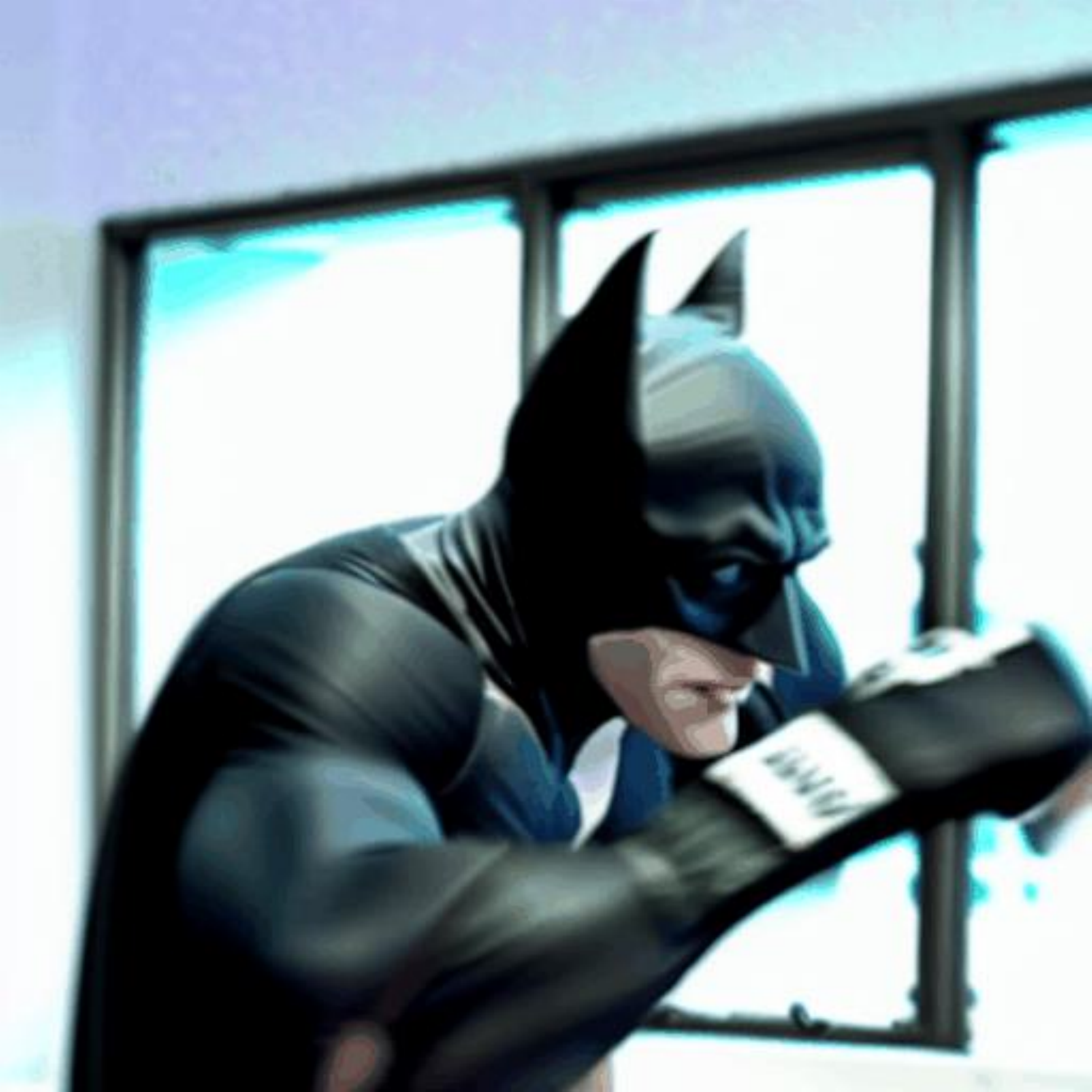}
\includegraphics[width=0.10\textwidth]{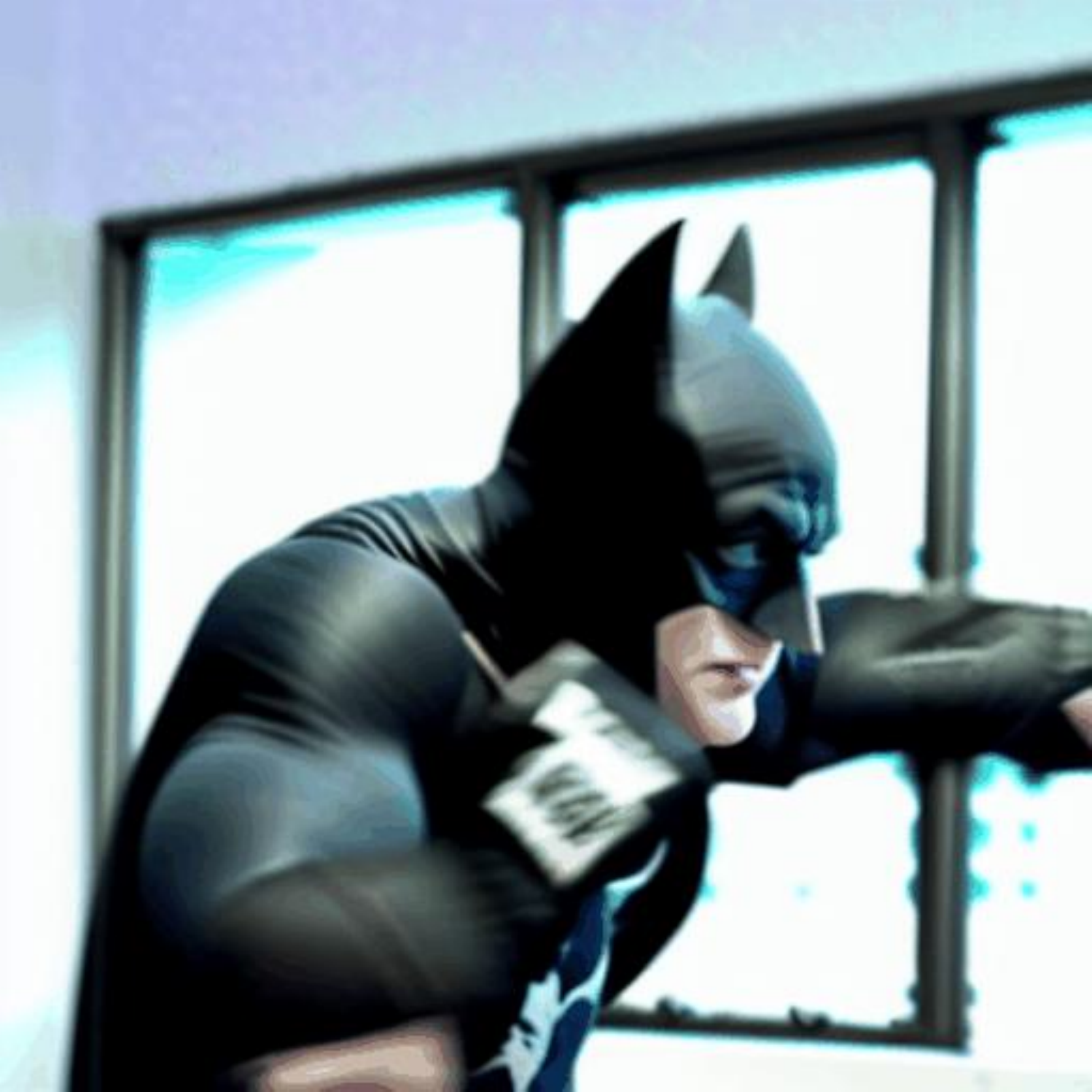}
\includegraphics[width=0.10\textwidth]{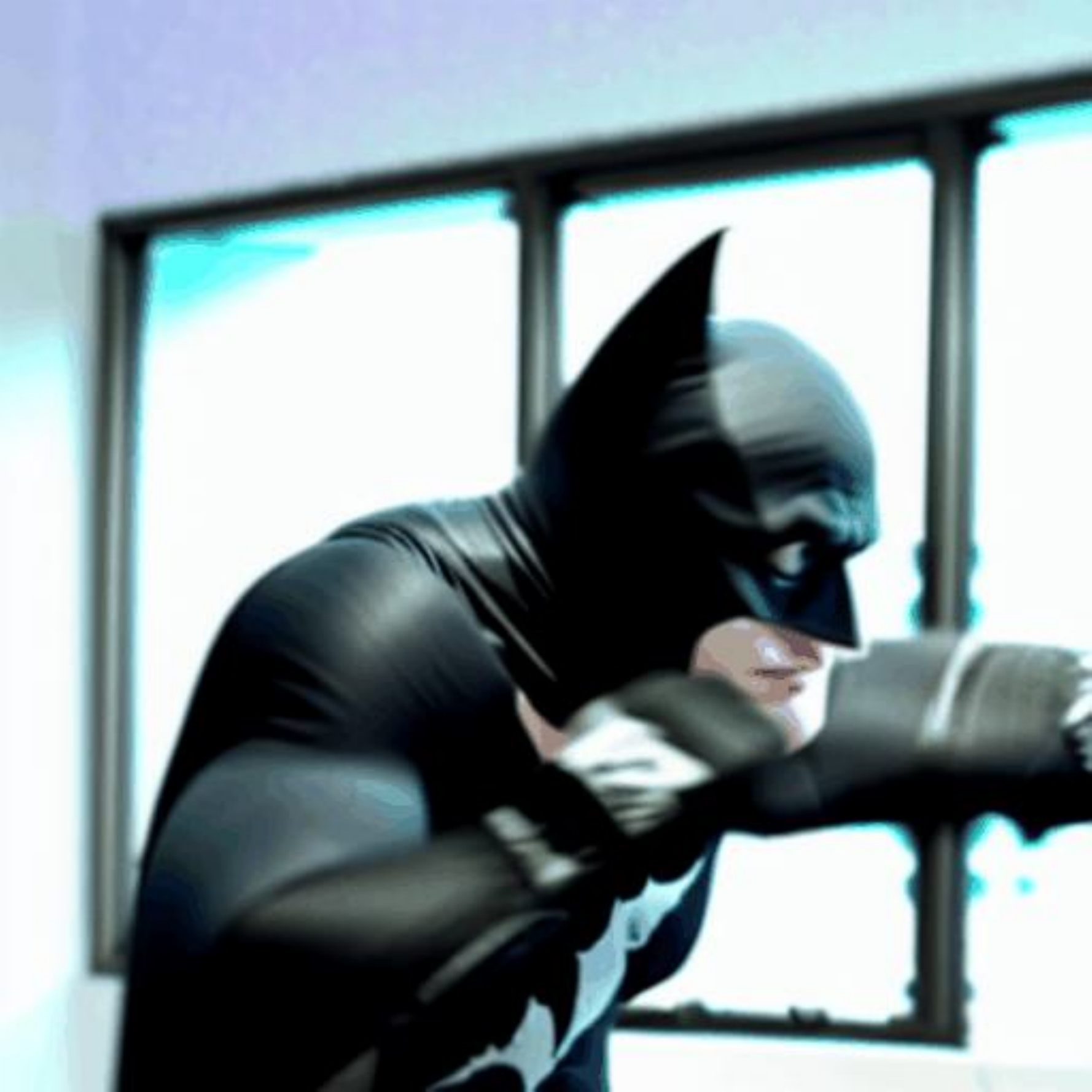}
\includegraphics[width=0.10\textwidth]{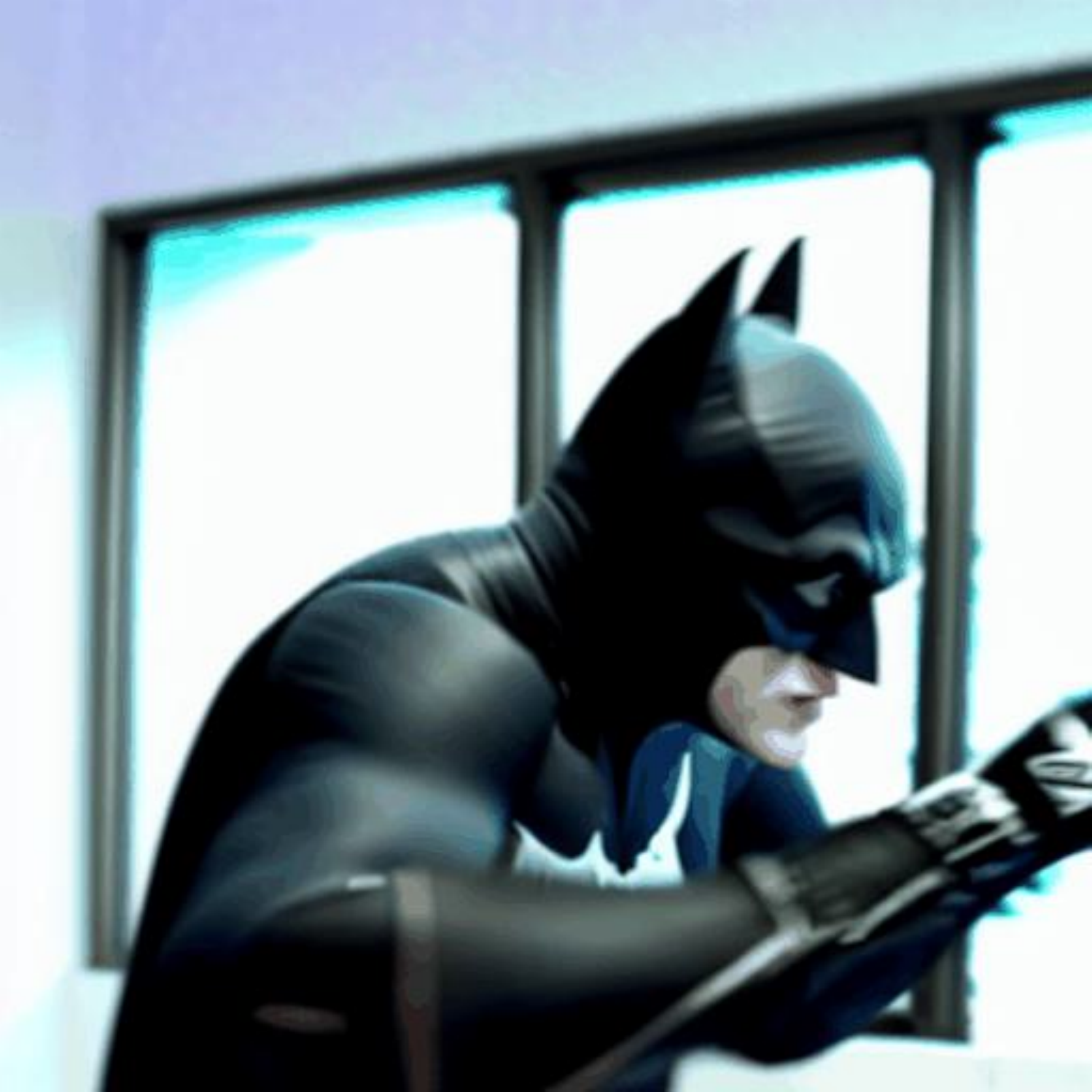}
\includegraphics[width=0.10\textwidth]{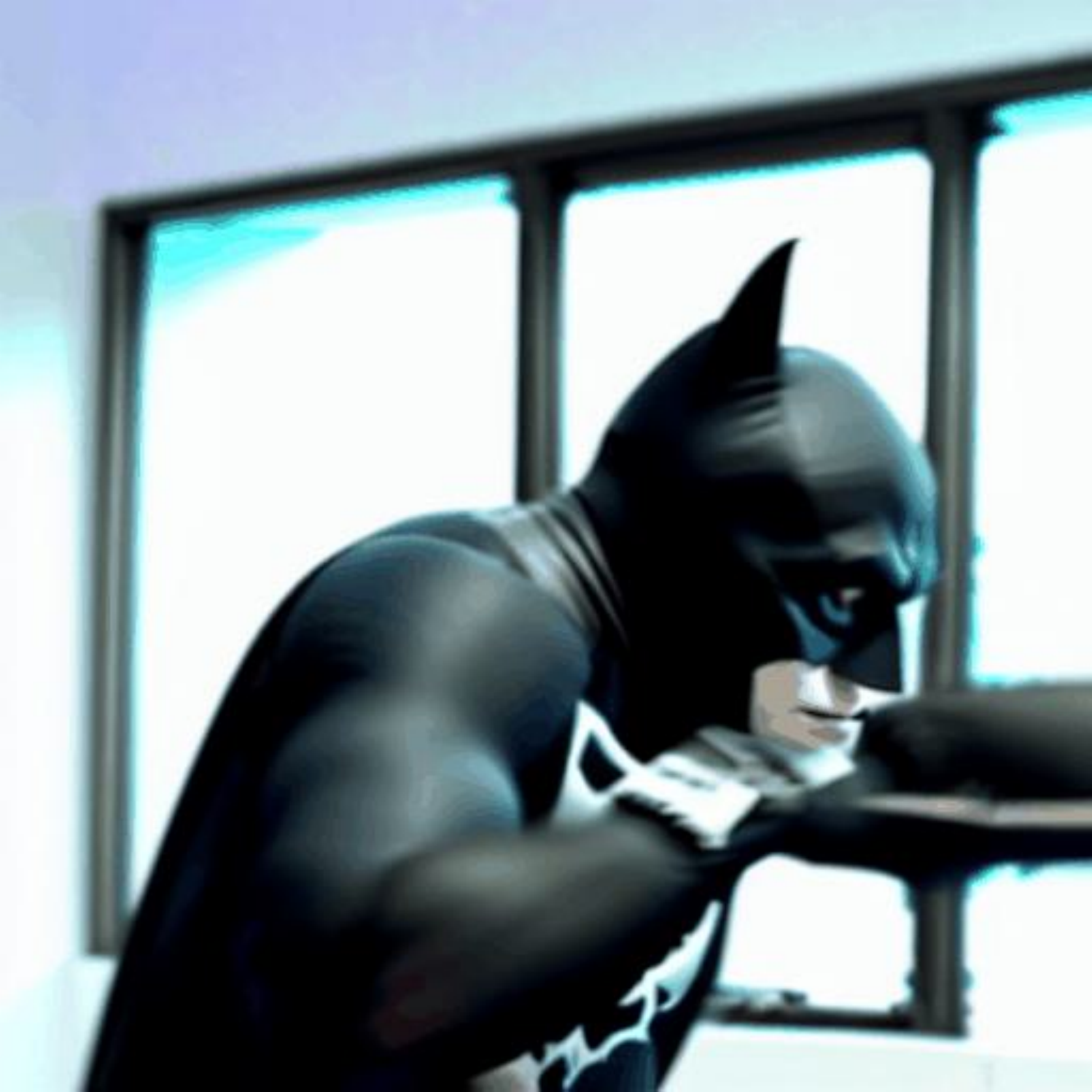}
\includegraphics[width=0.10\textwidth]{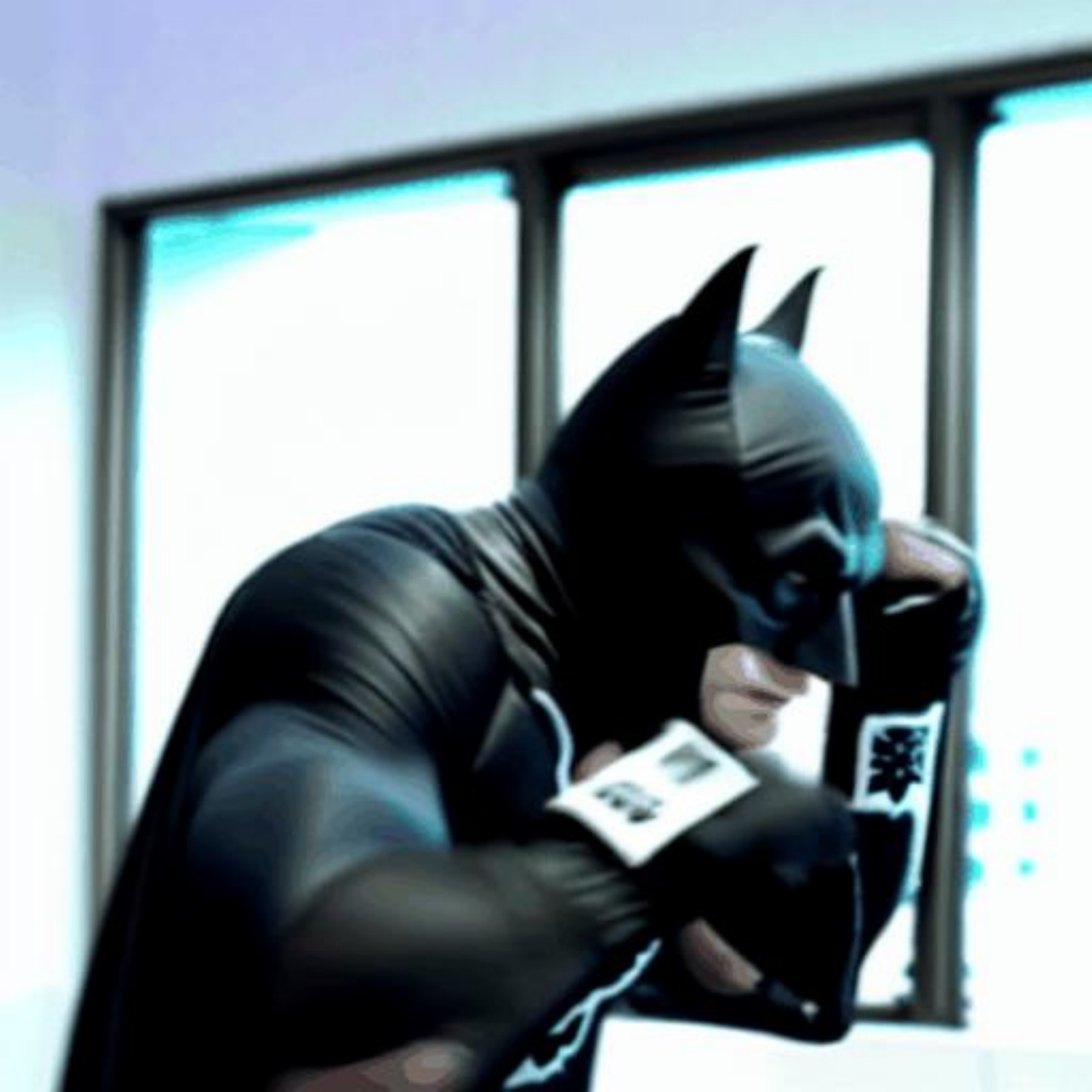}
\includegraphics[width=0.10\textwidth]{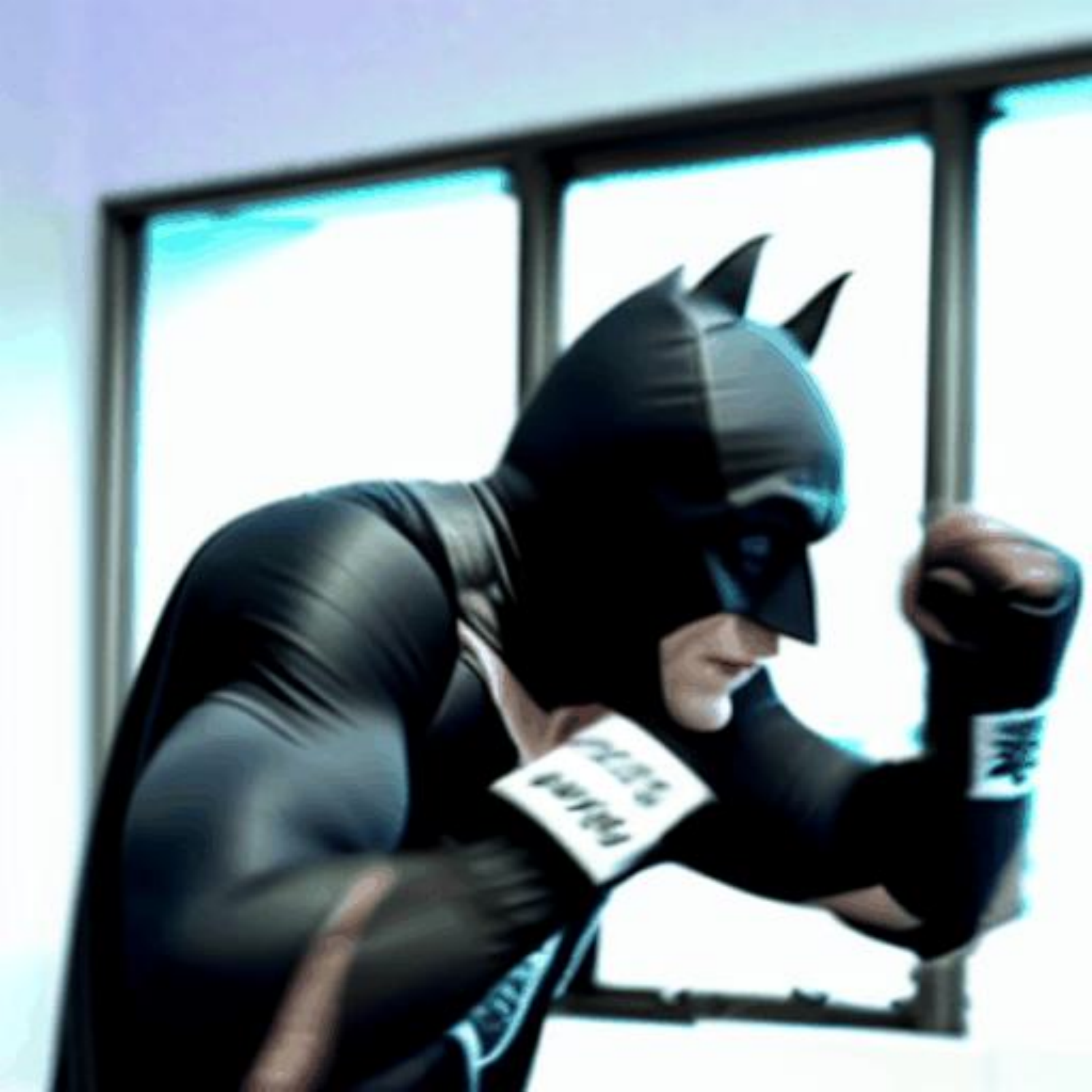}

\makebox[0.12\textwidth]{\colorbox{green}{\textbf{SDEdit}} A \textcolor{blue}{\textbf{Bat Man}} is boxing}\\
\includegraphics[width=0.10\textwidth]{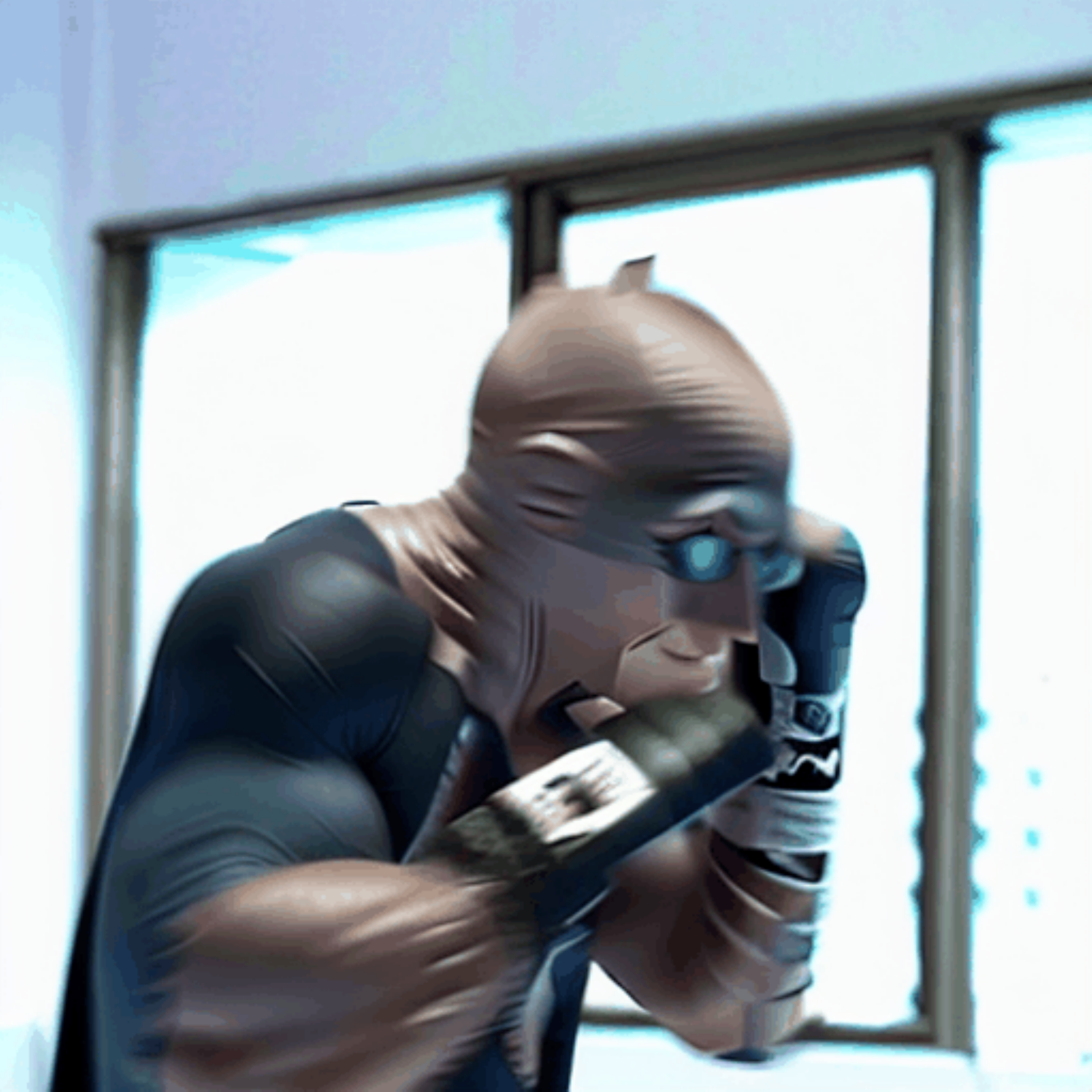}
\includegraphics[width=0.10\textwidth]{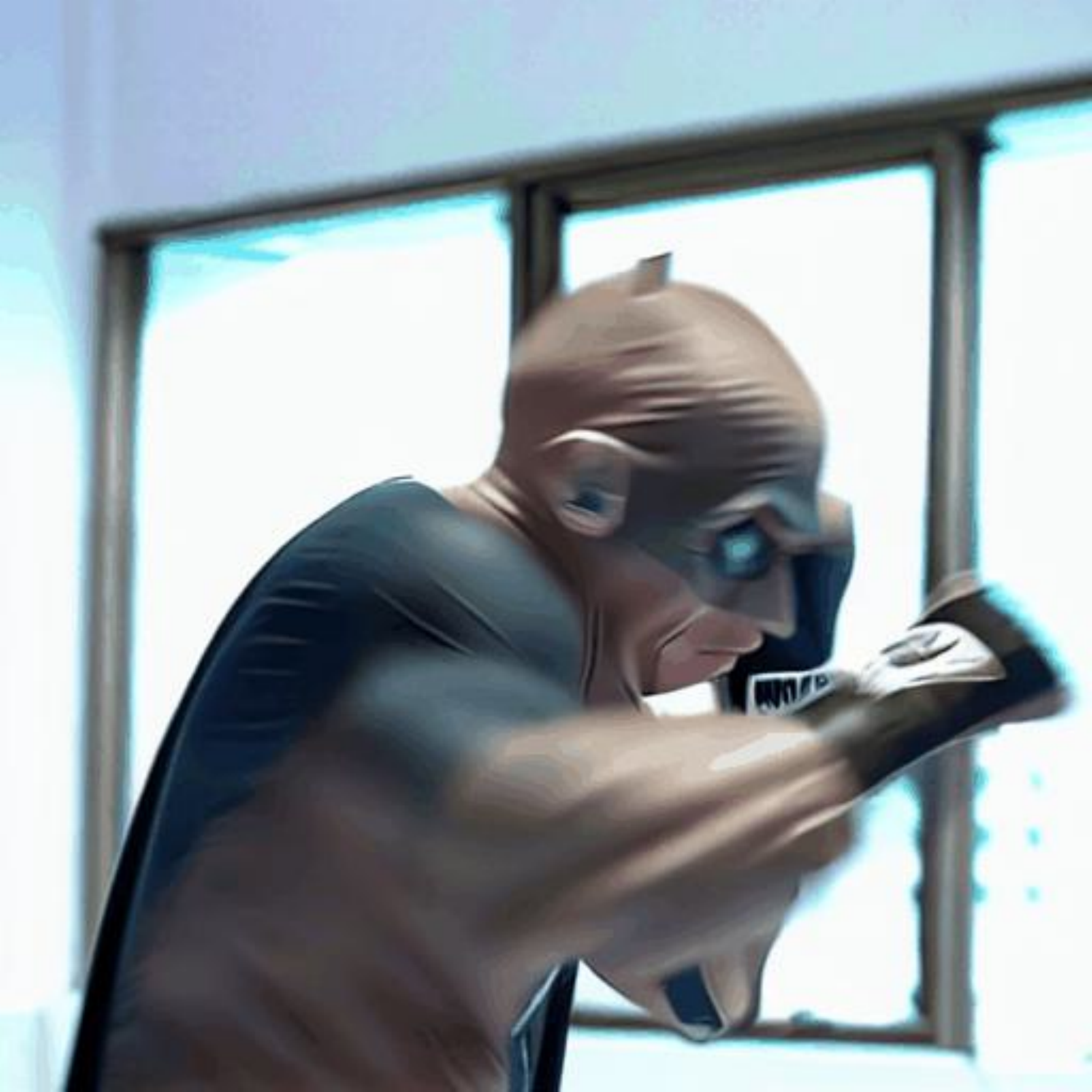}
\includegraphics[width=0.10\textwidth]{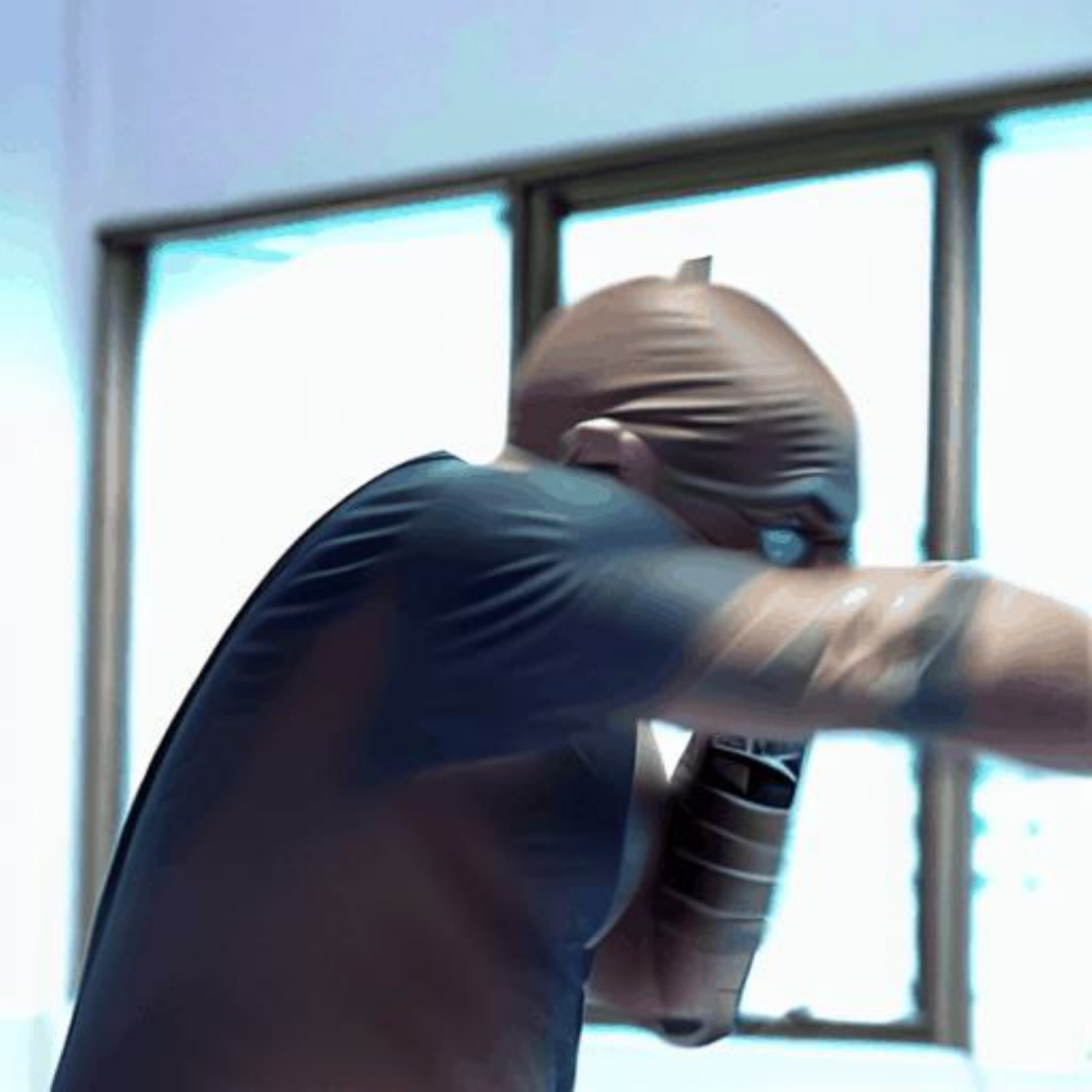}
\includegraphics[width=0.10\textwidth]{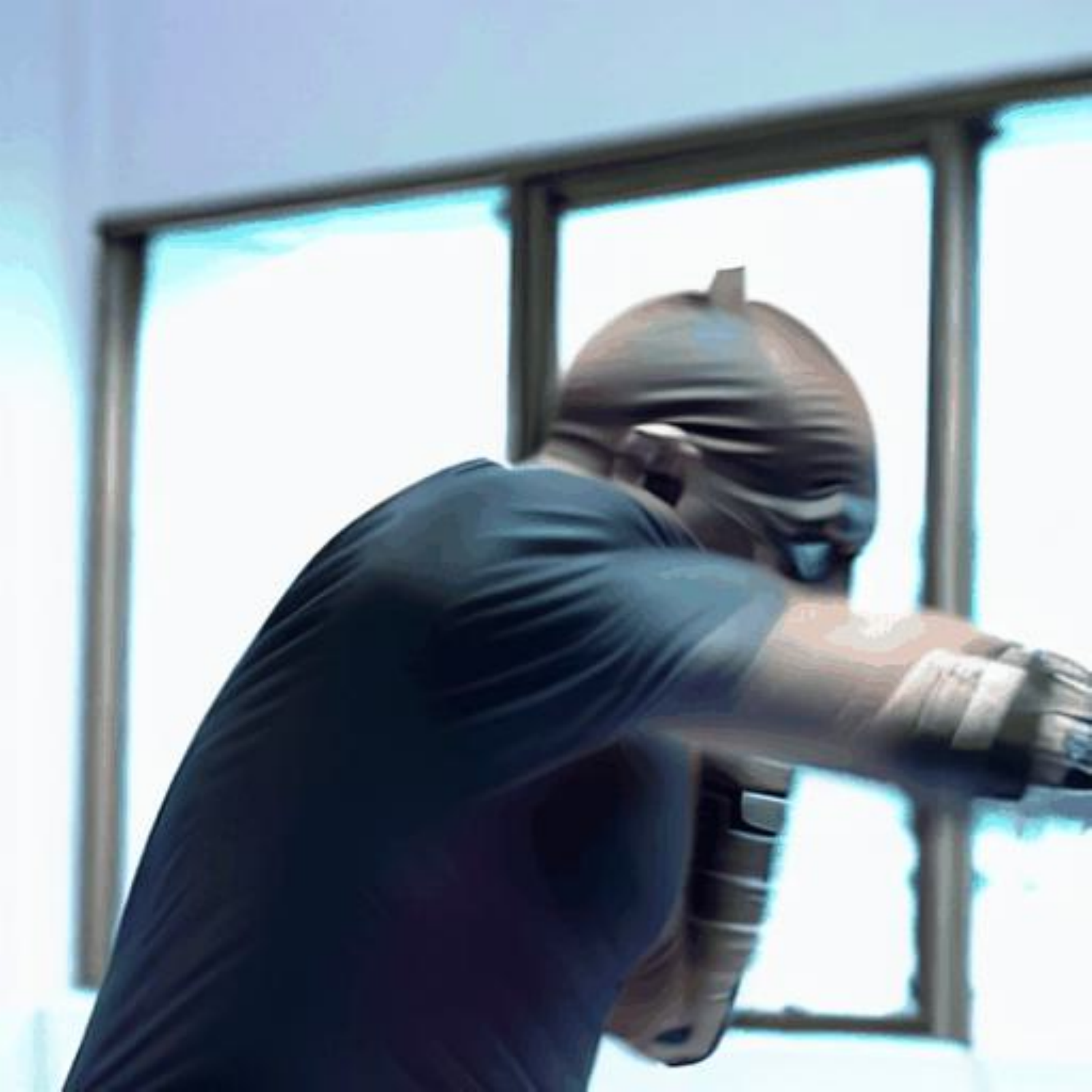}
\includegraphics[width=0.10\textwidth]{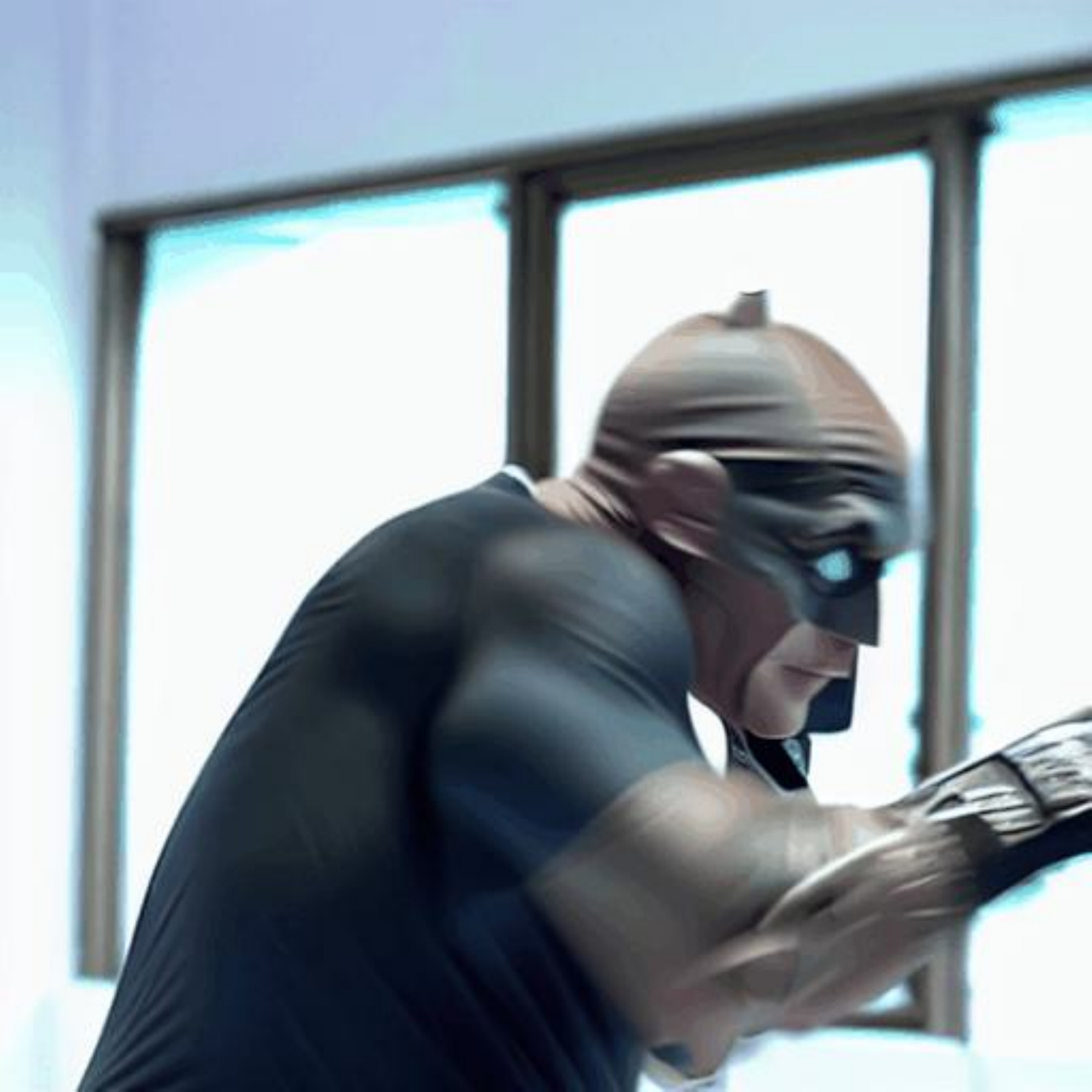}
\includegraphics[width=0.10\textwidth]{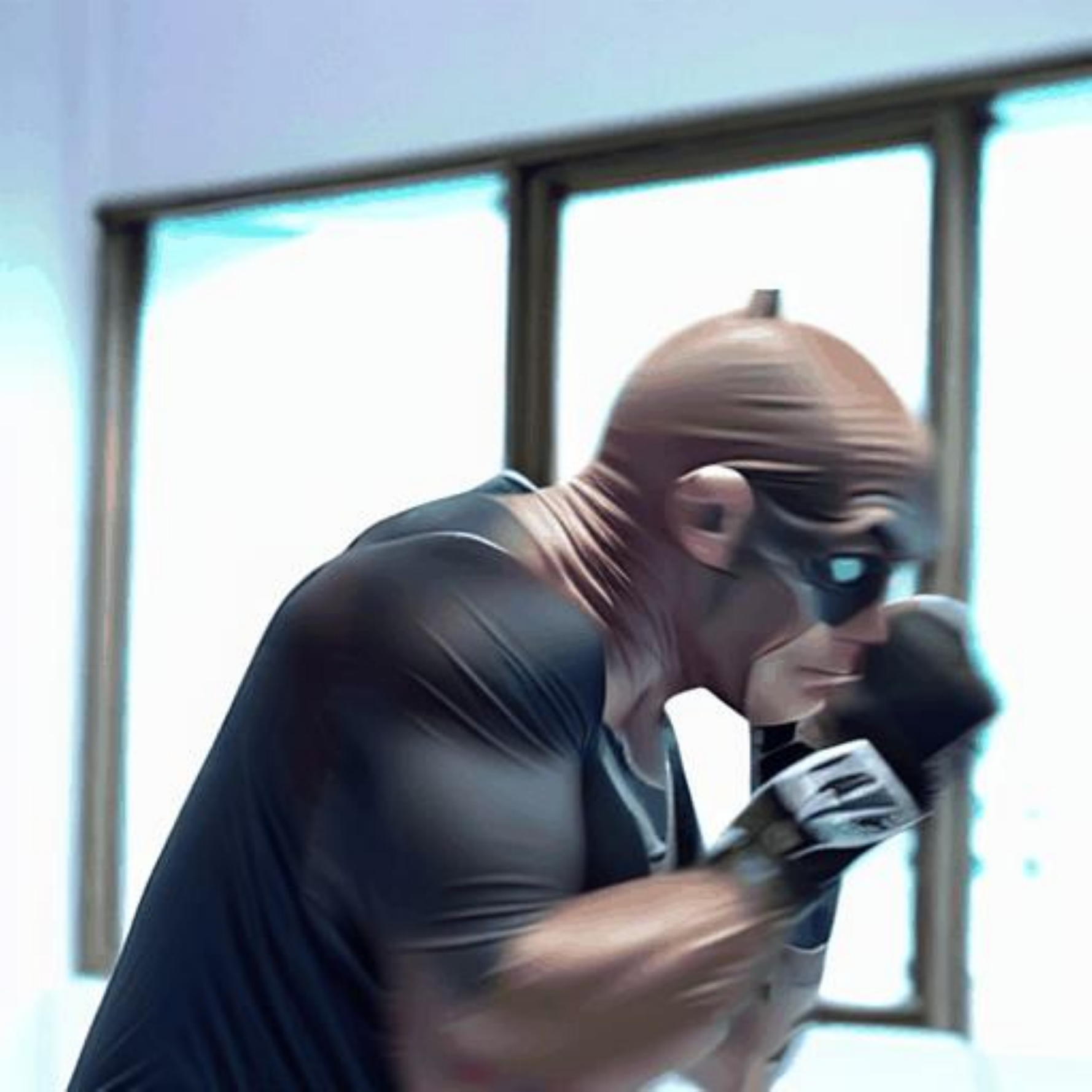}
\includegraphics[width=0.10\textwidth]{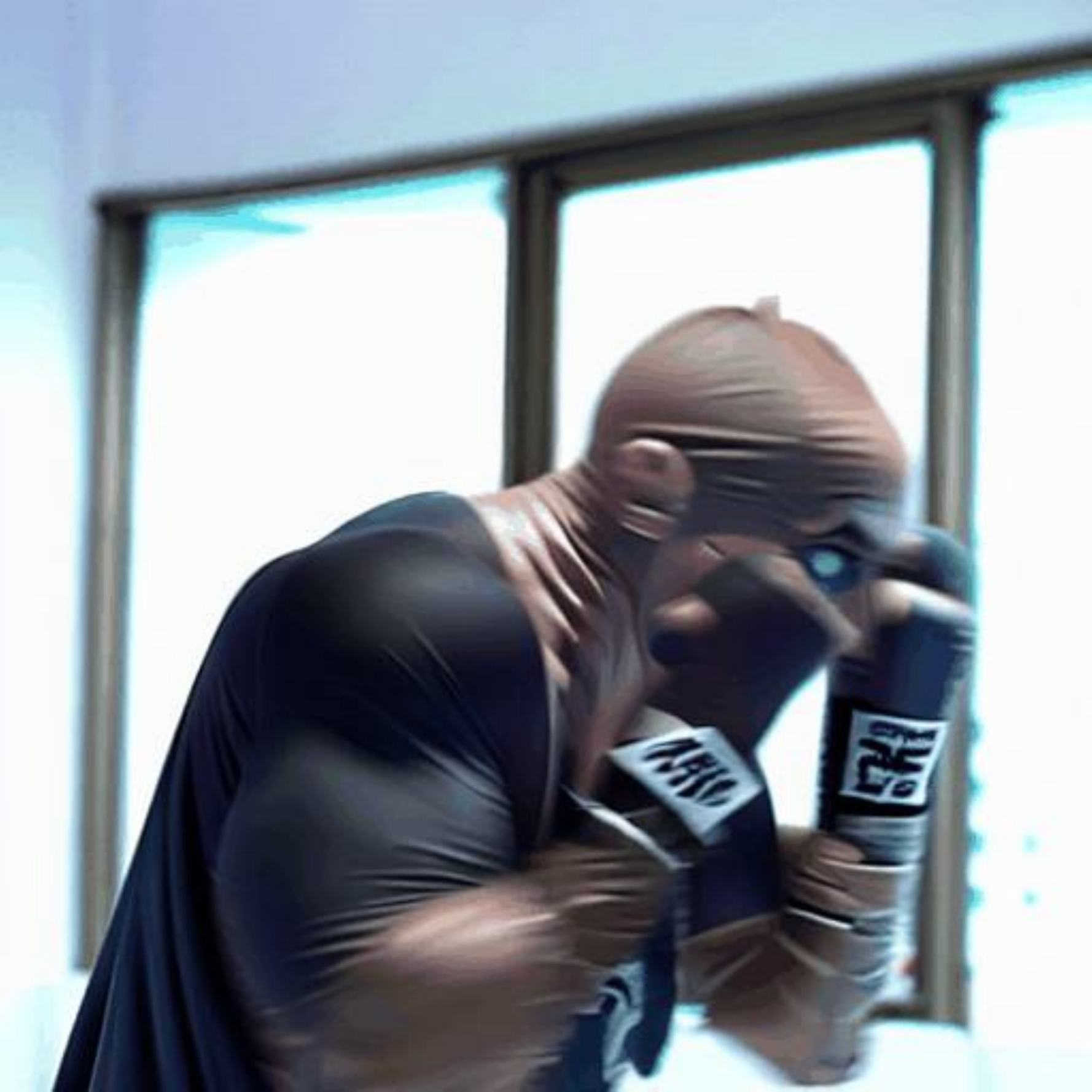}
\includegraphics[width=0.10\textwidth]{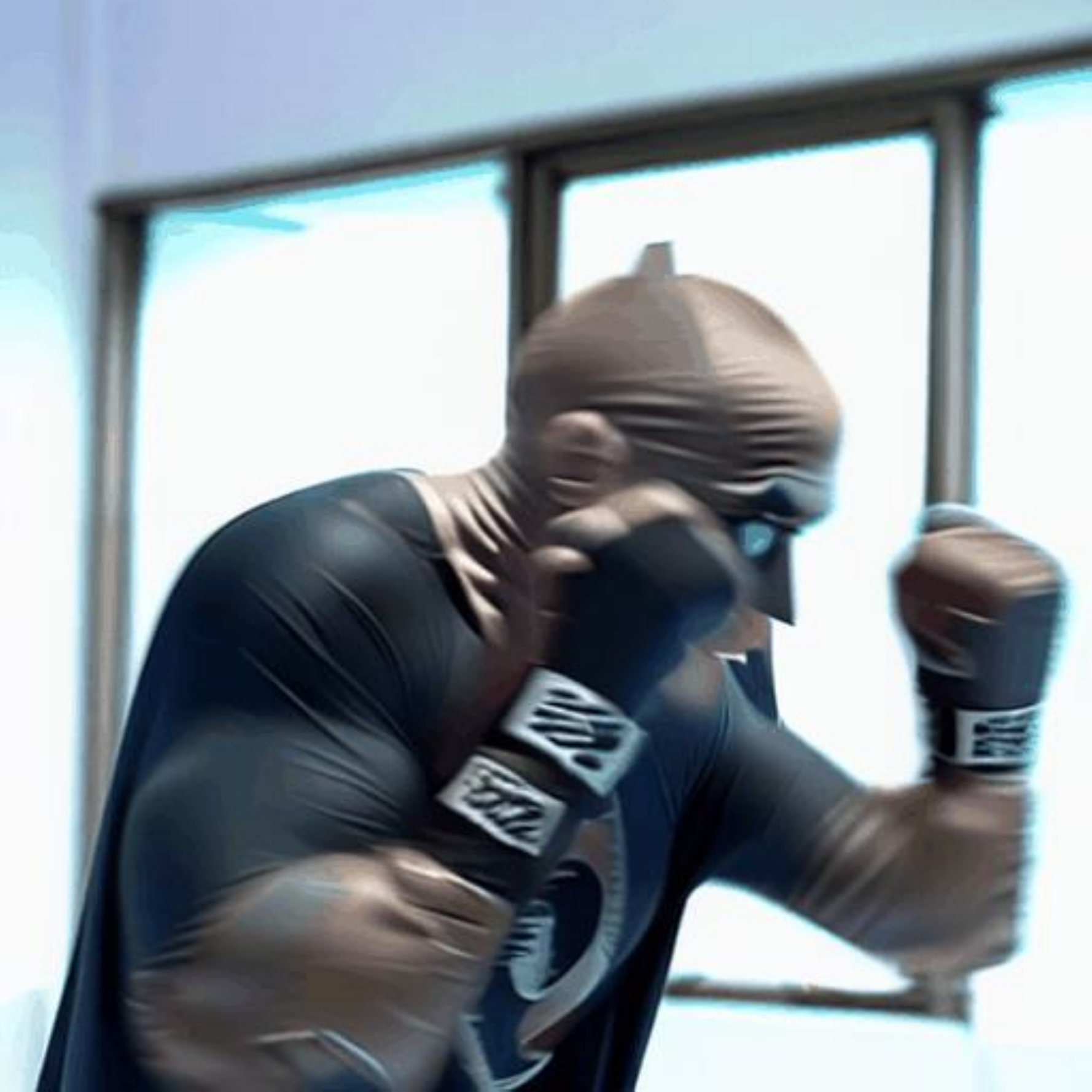}



\makebox[0.12\textwidth]{\colorbox{green}{\textbf{Video-P2P}} A \textcolor{blue}{\textbf{Bat Man}} is boxing}\\

\includegraphics[width=0.10\textwidth]{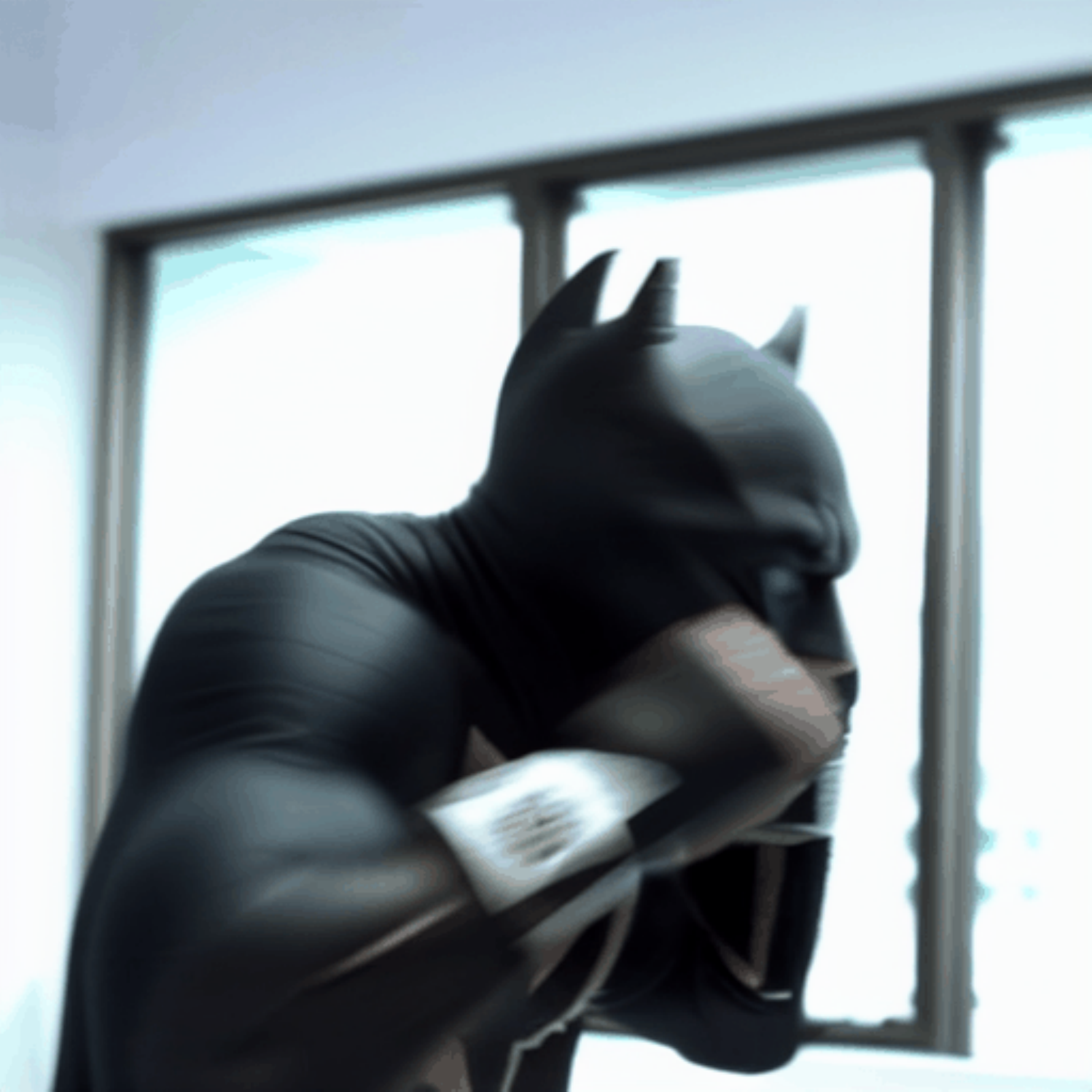}
\includegraphics[width=0.10\textwidth]{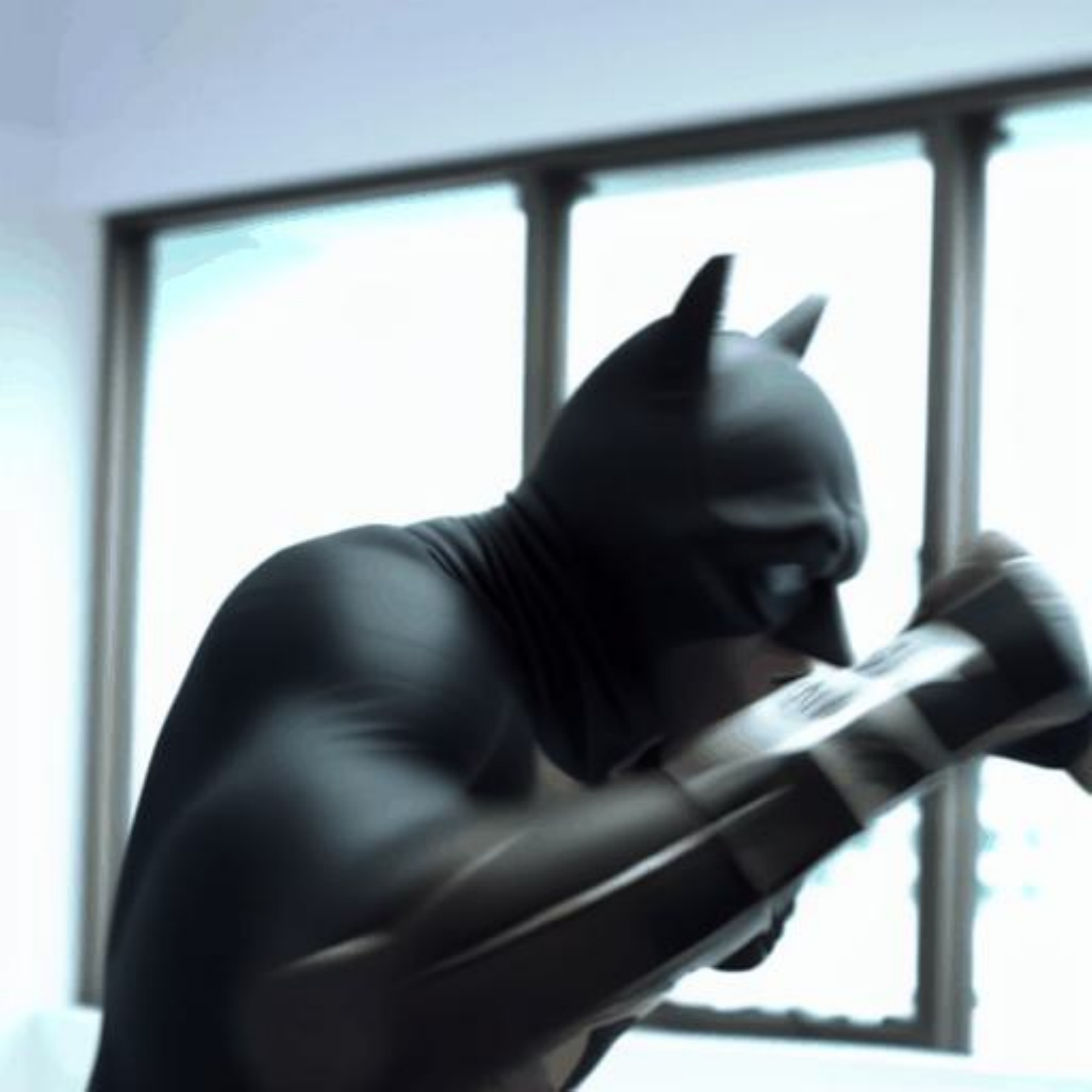}
\includegraphics[width=0.10\textwidth]{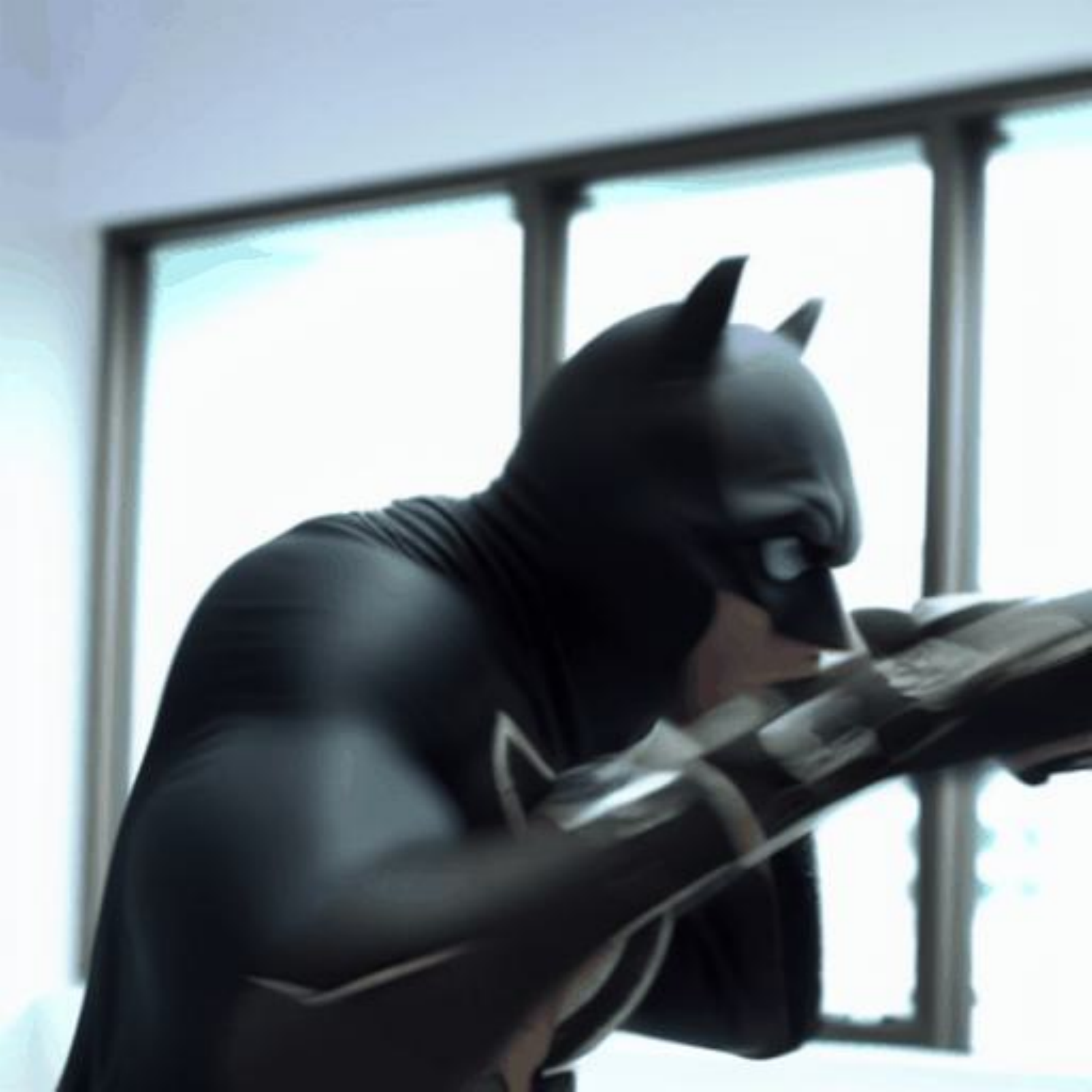}
\includegraphics[width=0.10\textwidth]{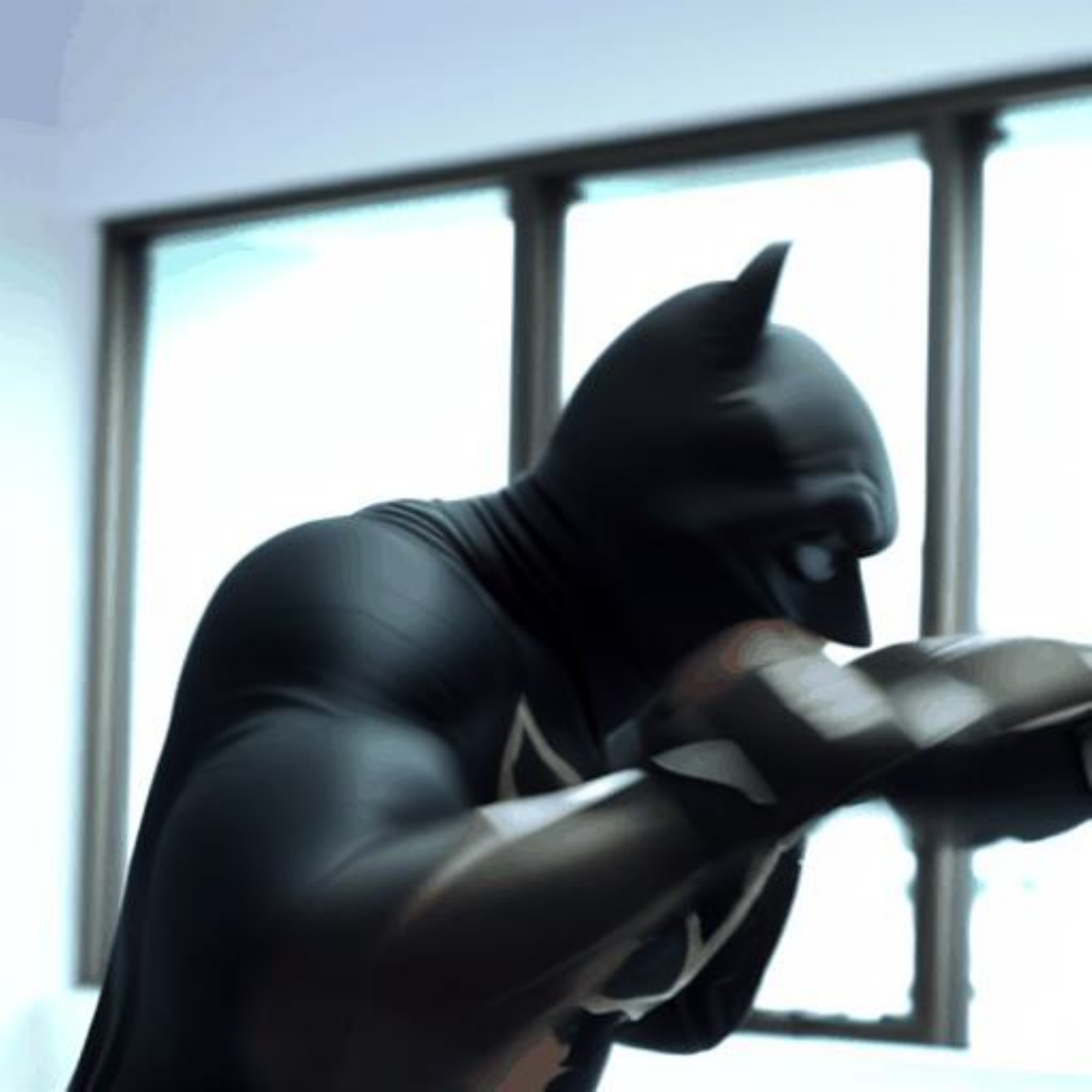}
\includegraphics[width=0.10\textwidth]{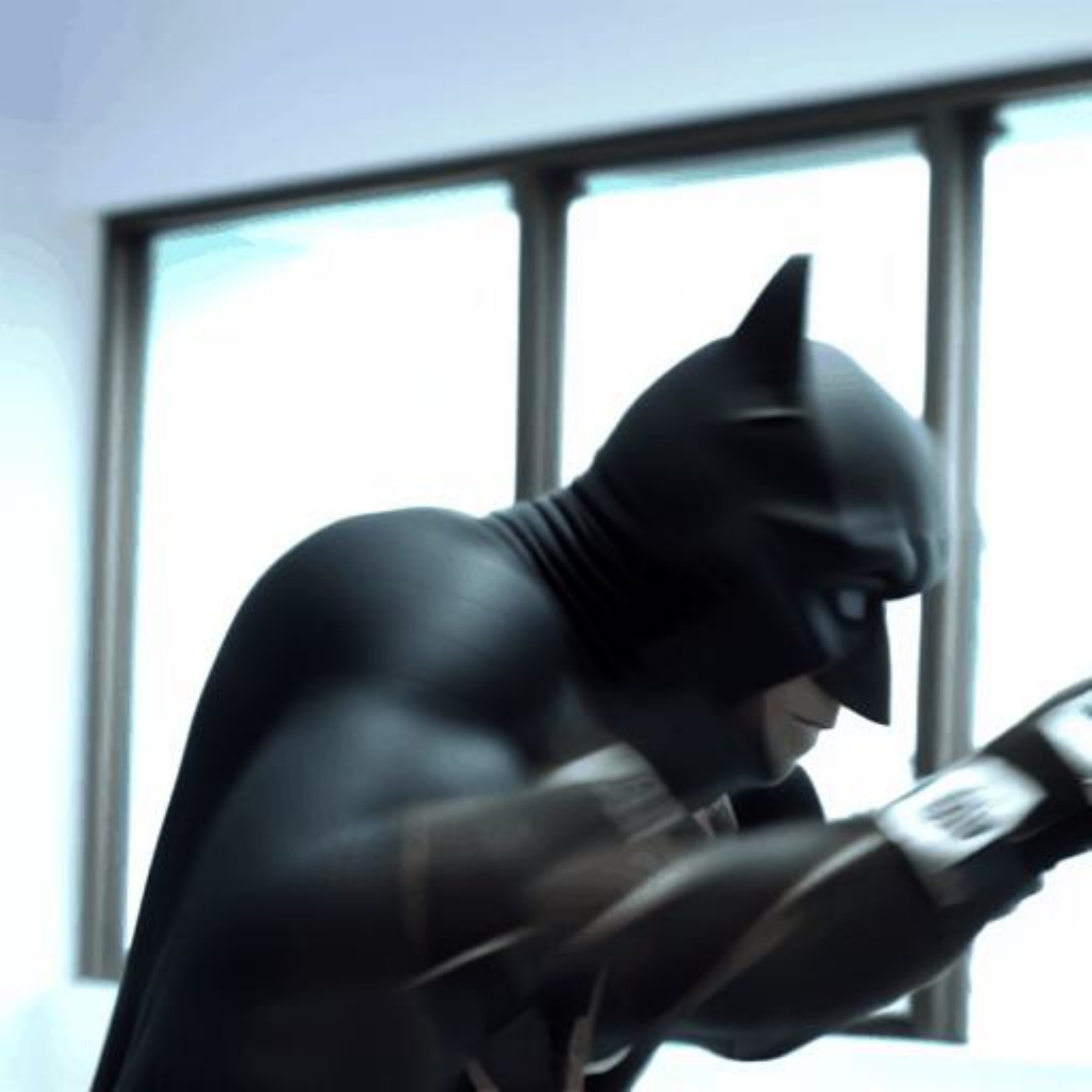}
\includegraphics[width=0.10\textwidth]{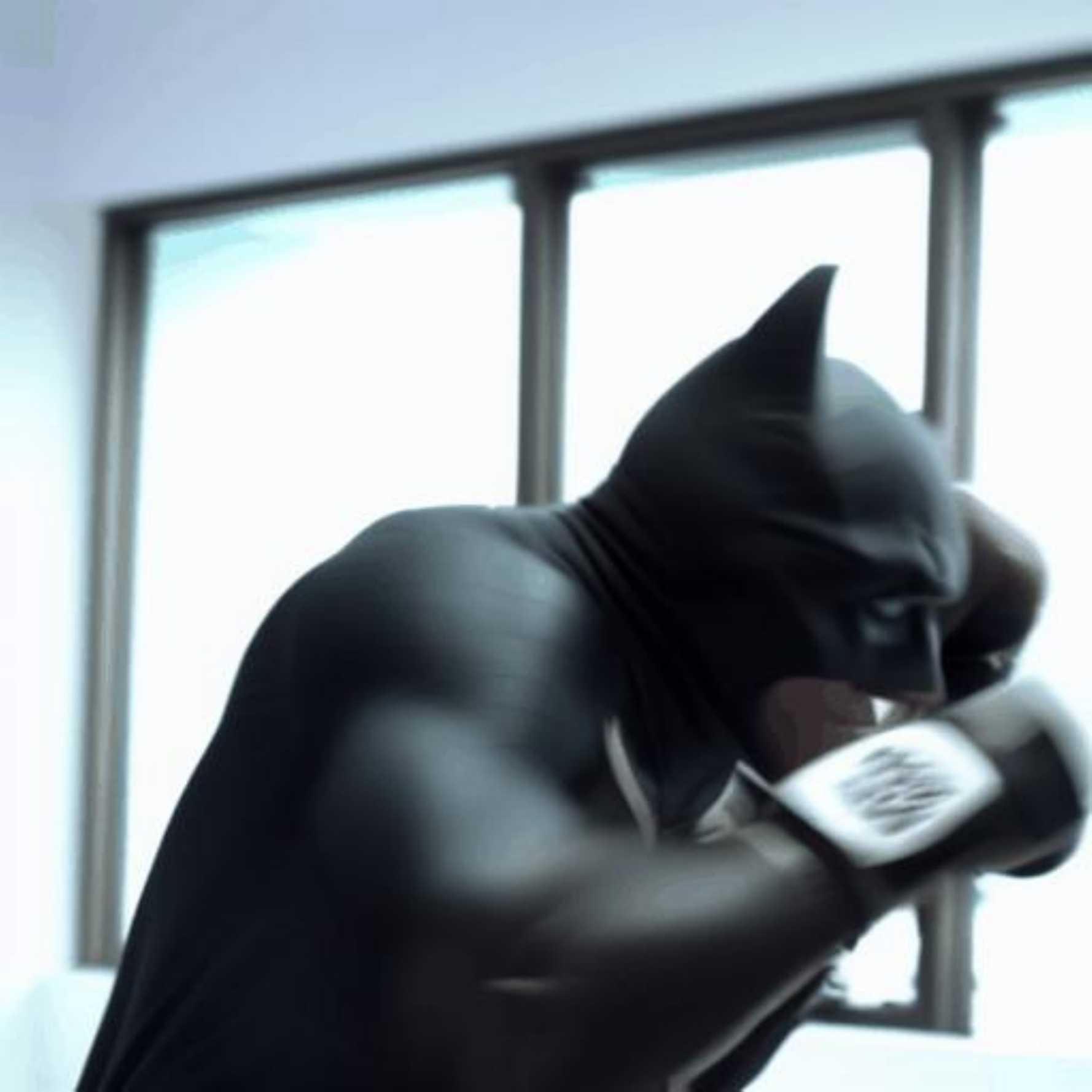}
\includegraphics[width=0.10\textwidth]{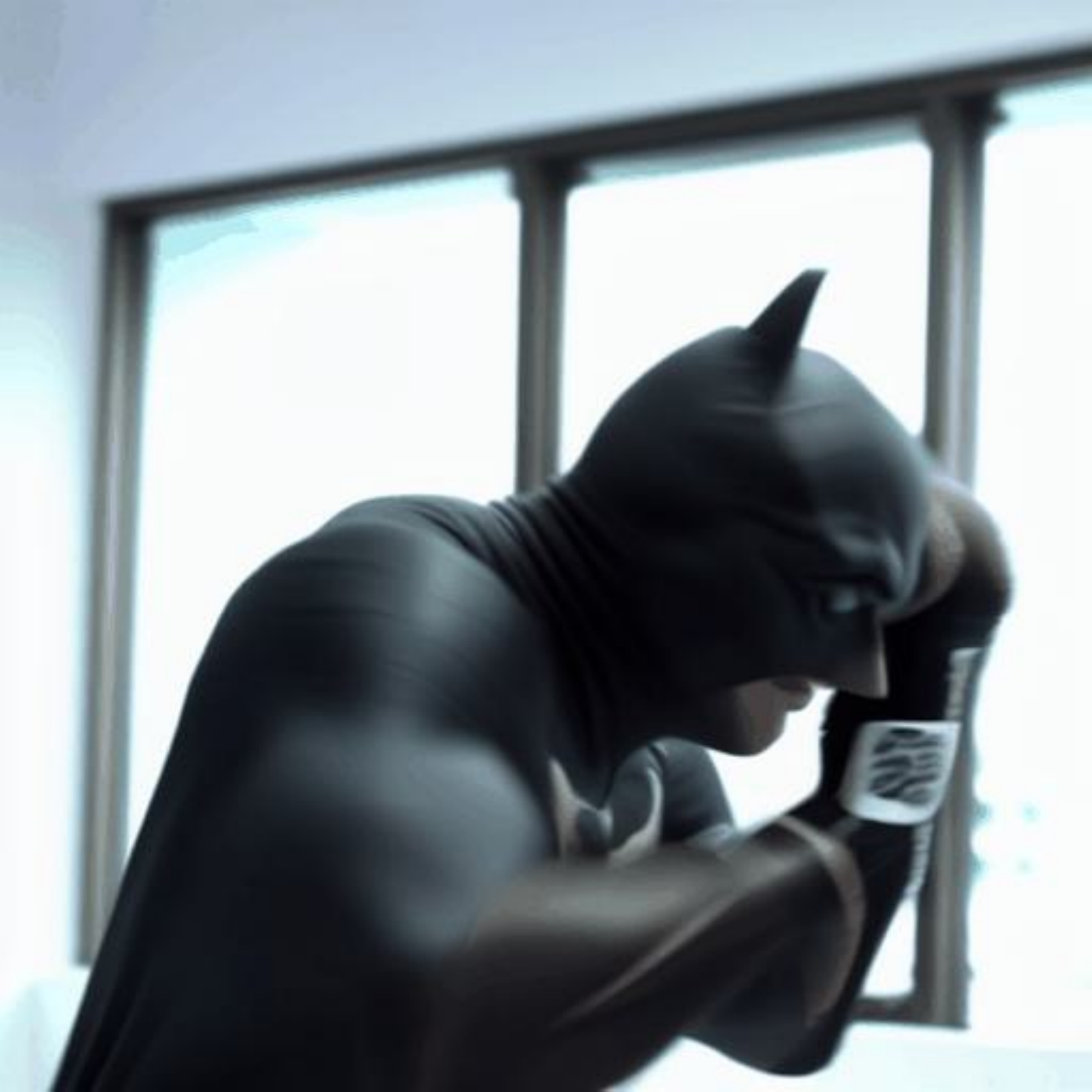}
\includegraphics[width=0.10\textwidth]{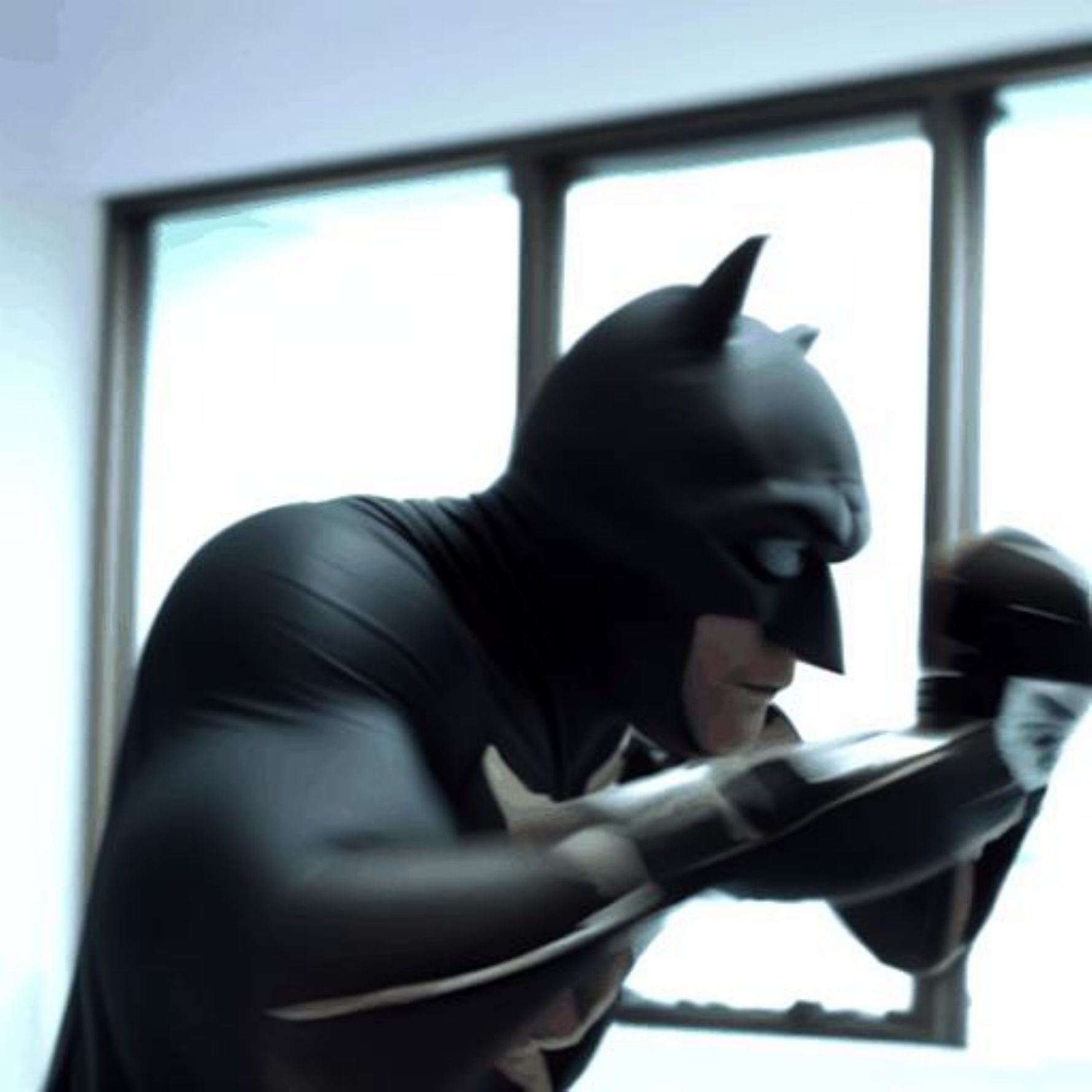}

\makebox[0.12\textwidth]{\colorbox{pink}{\textbf{Training video}} A man is riding a motorcycle}\\

\includegraphics[width=0.10\textwidth]{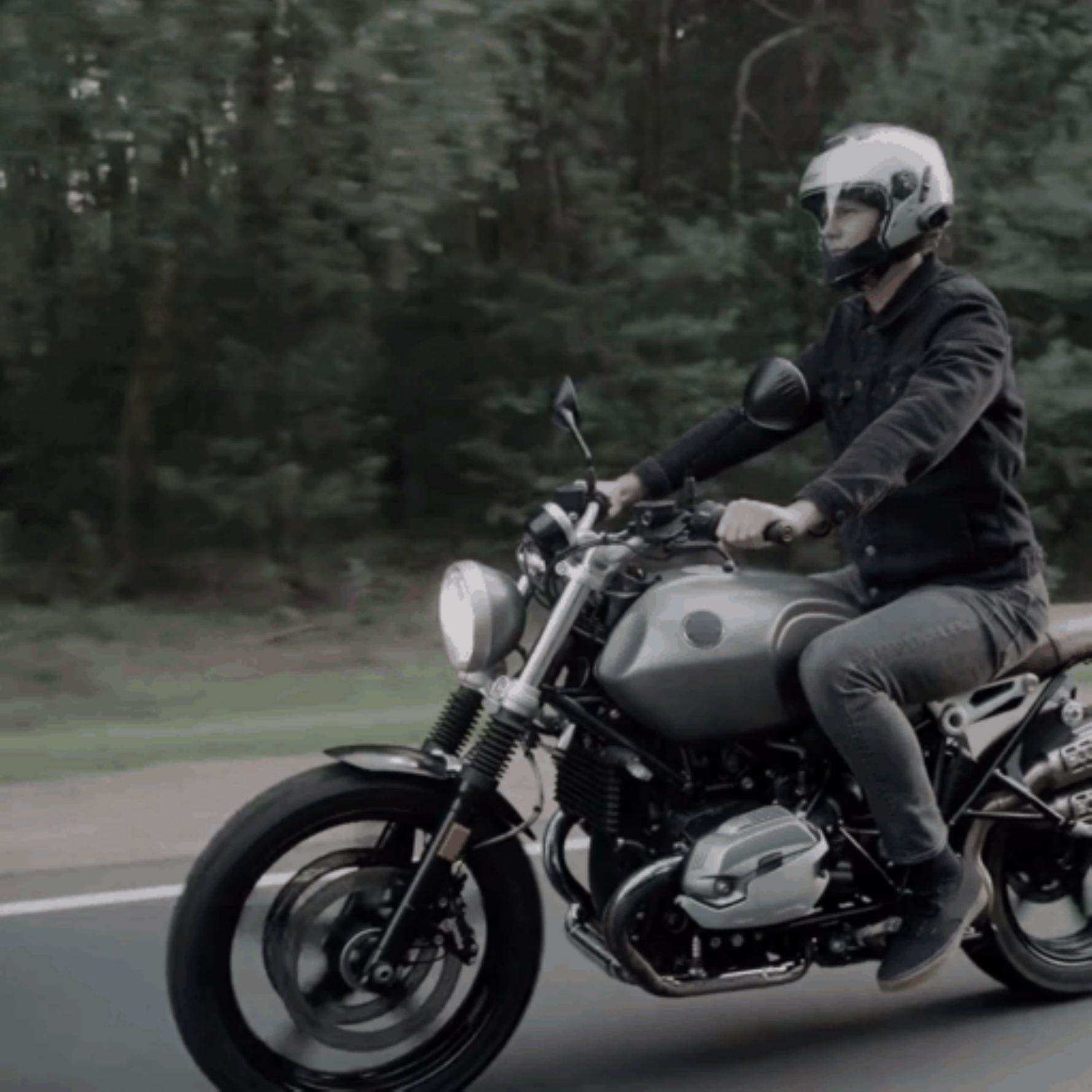}
\includegraphics[width=0.10\textwidth]{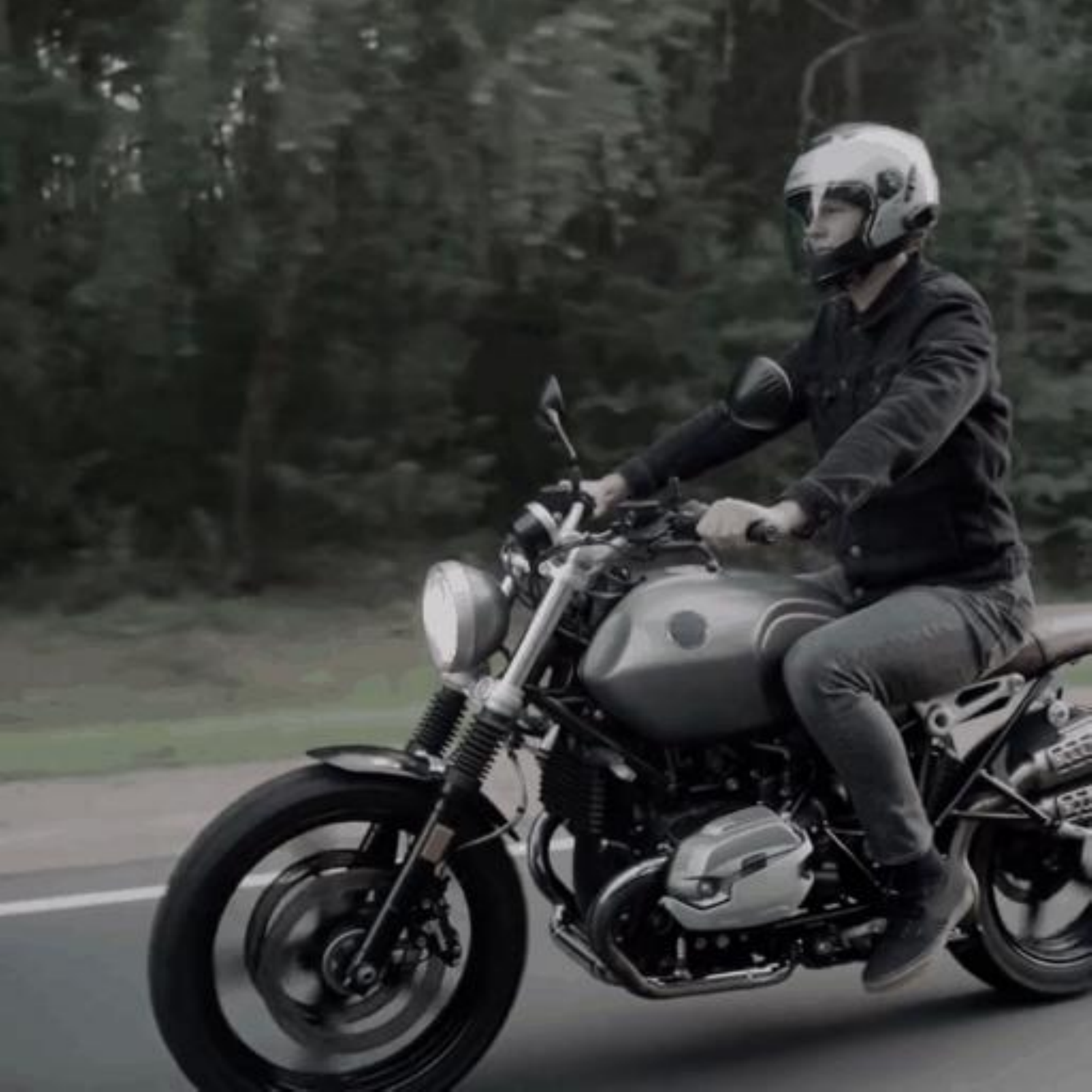}
\includegraphics[width=0.10\textwidth]{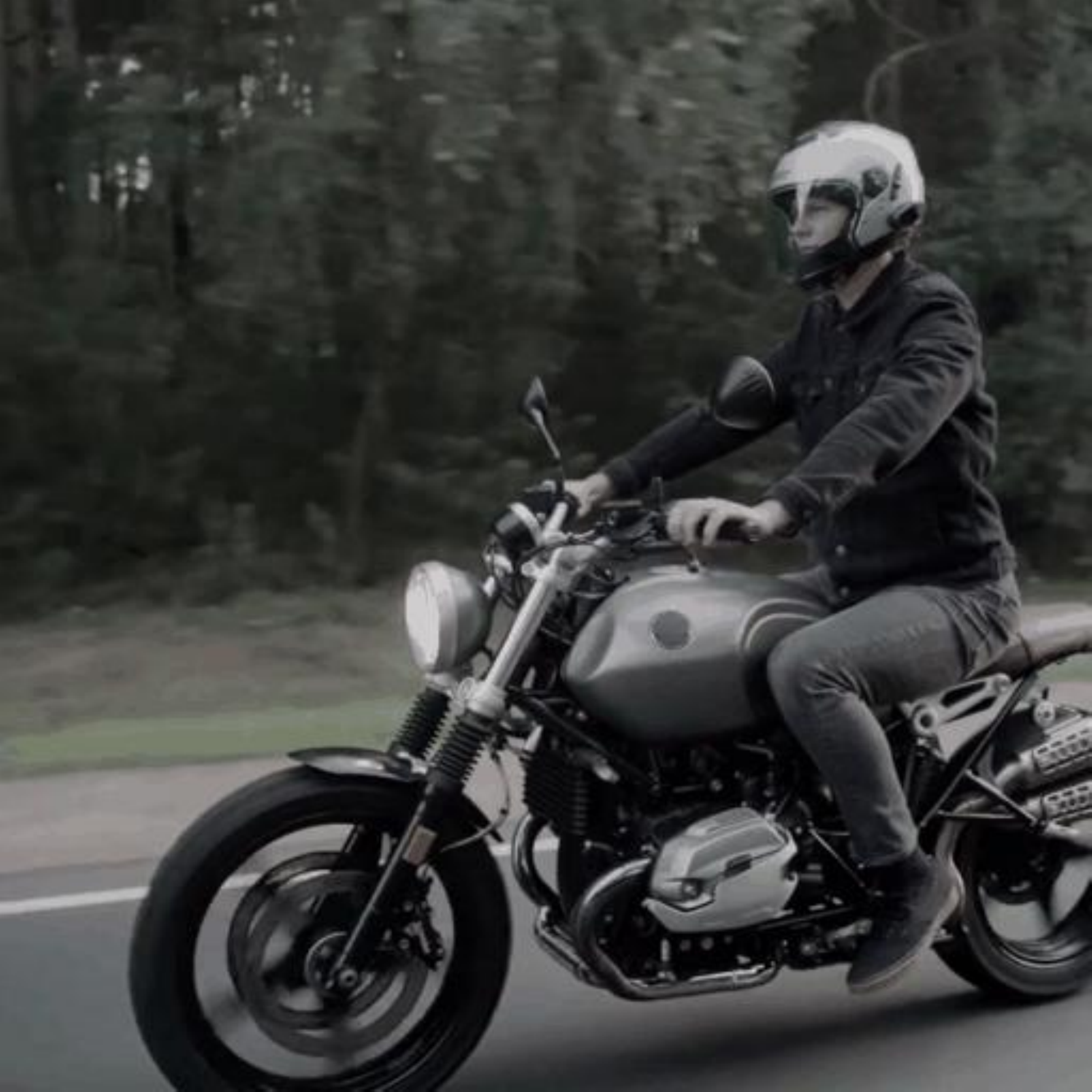}
\includegraphics[width=0.10\textwidth]{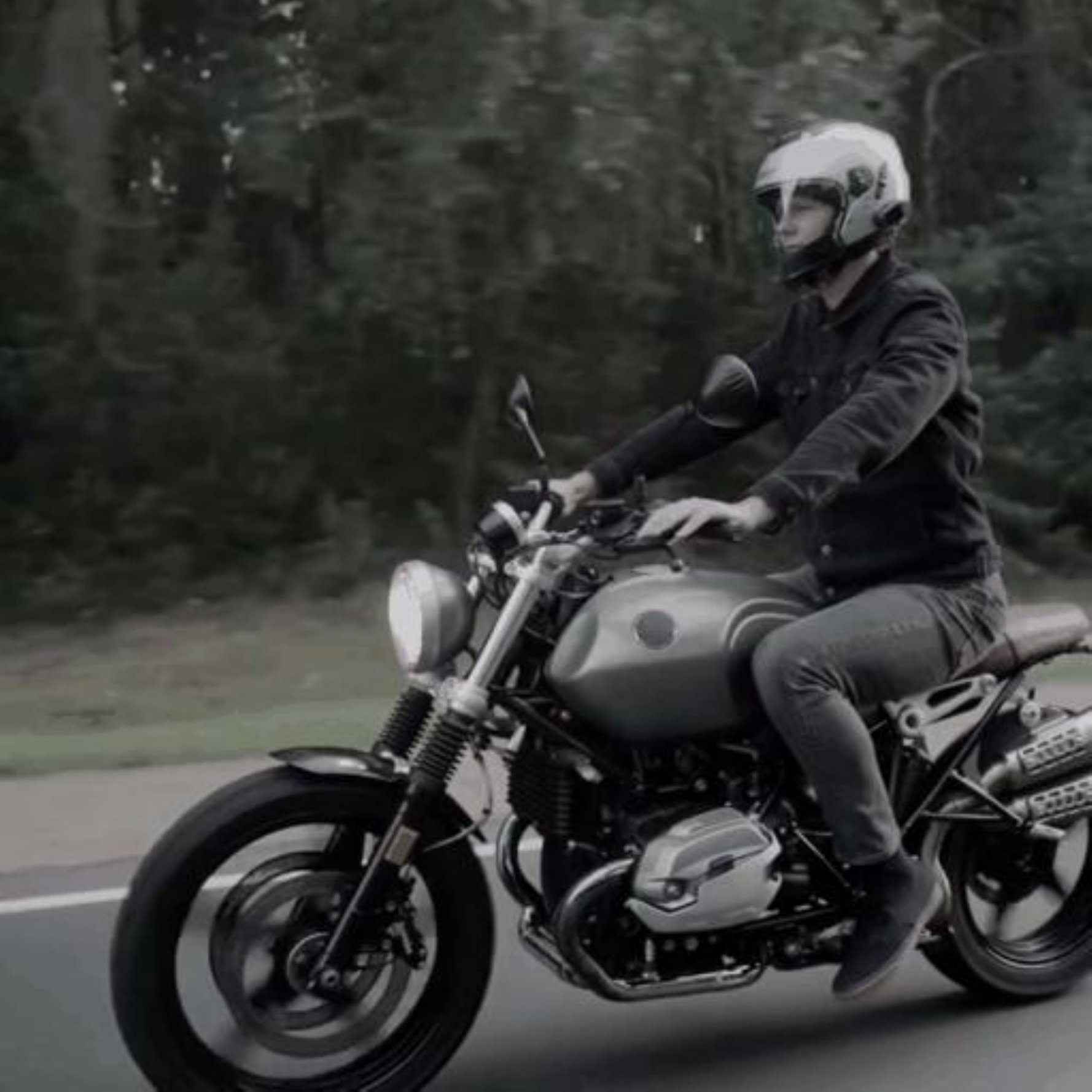}
\includegraphics[width=0.10\textwidth]{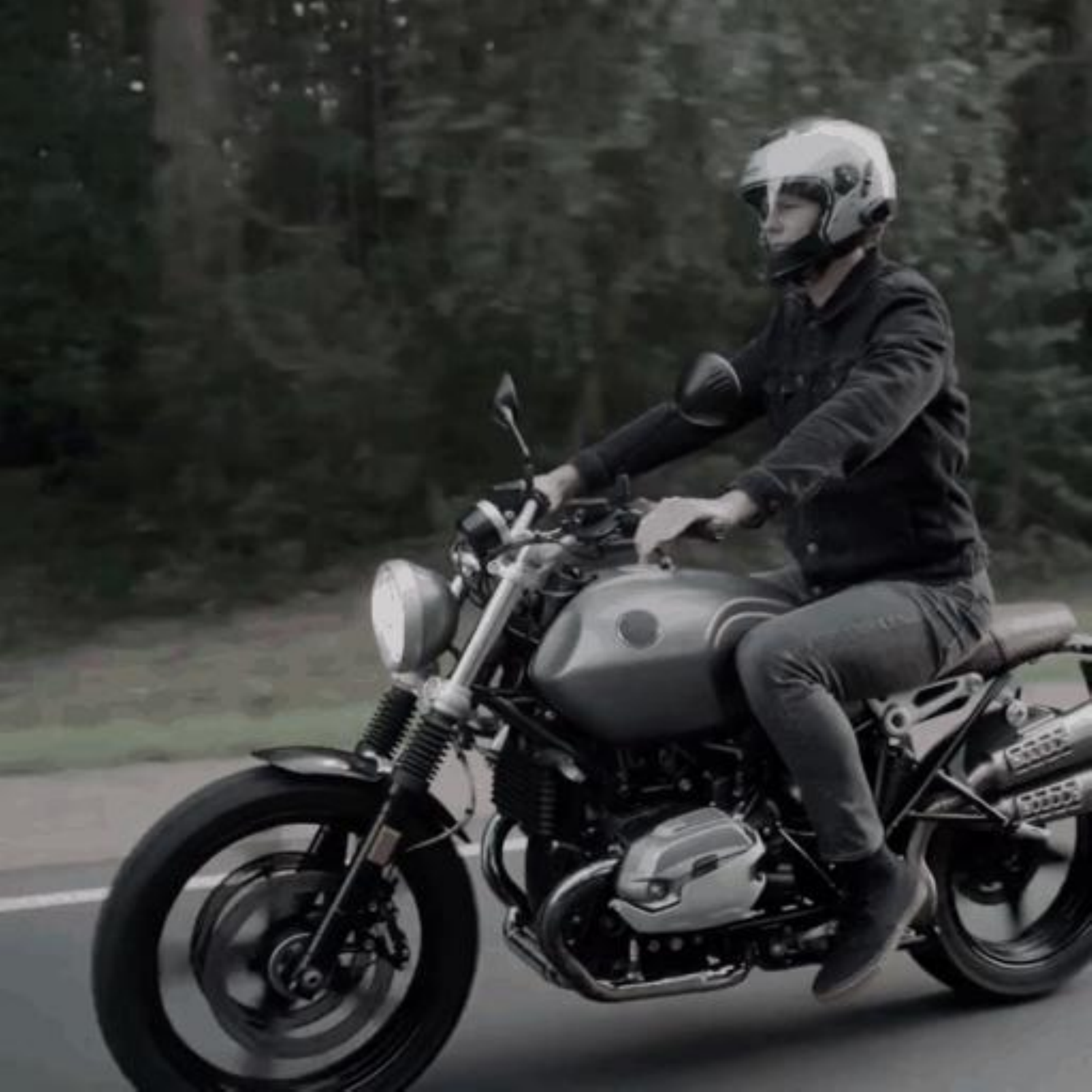}
\includegraphics[width=0.10\textwidth]{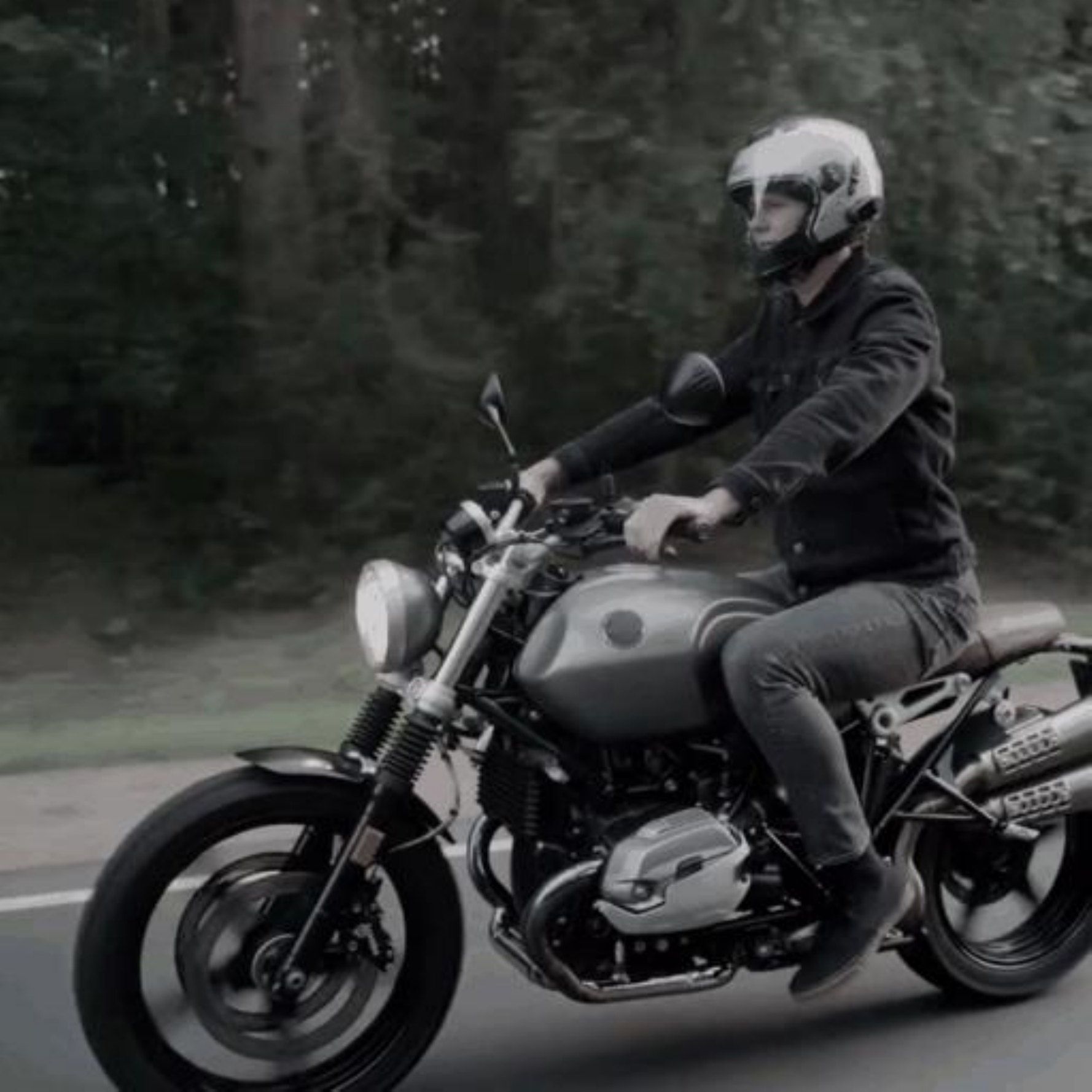}
\includegraphics[width=0.10\textwidth]{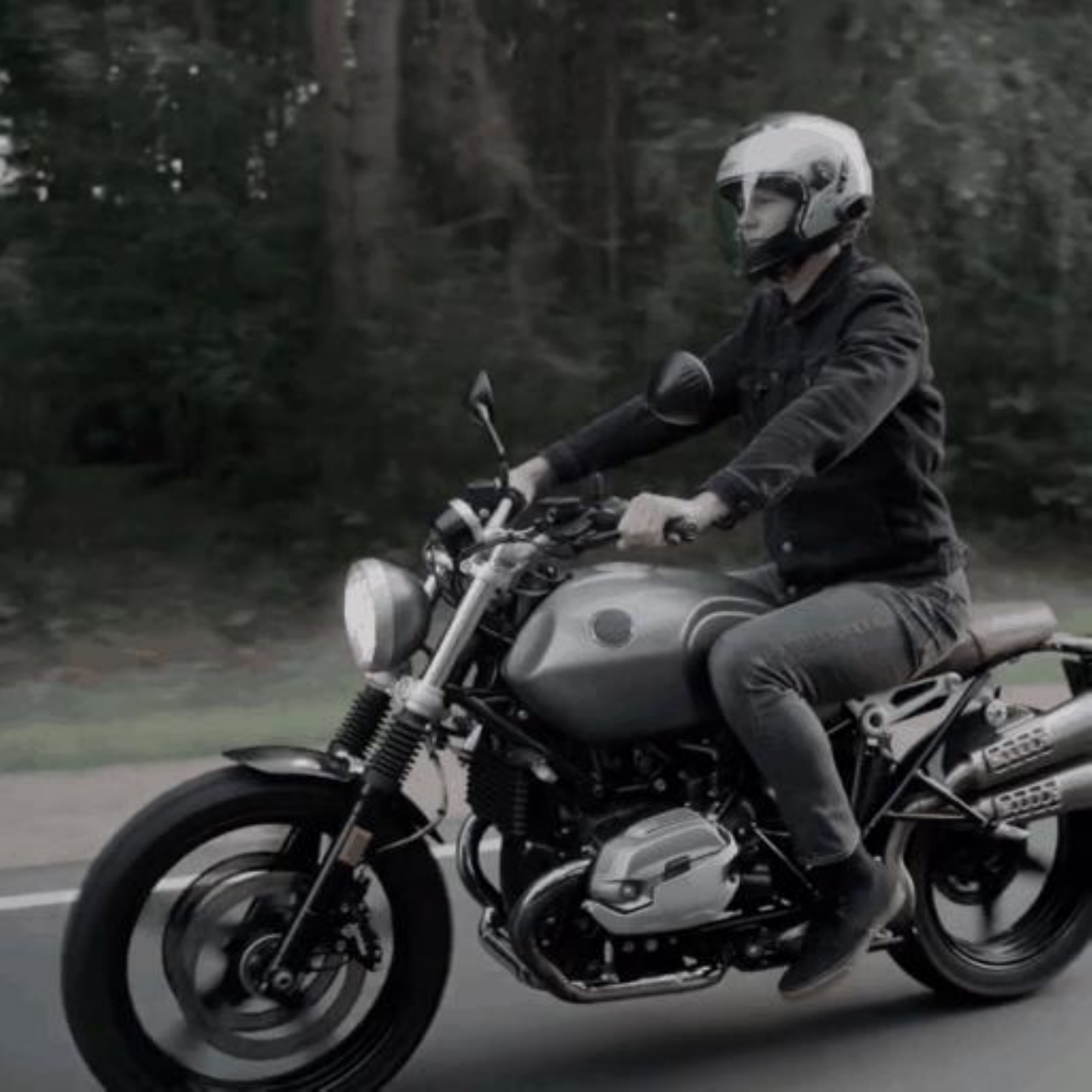}
\includegraphics[width=0.10\textwidth]{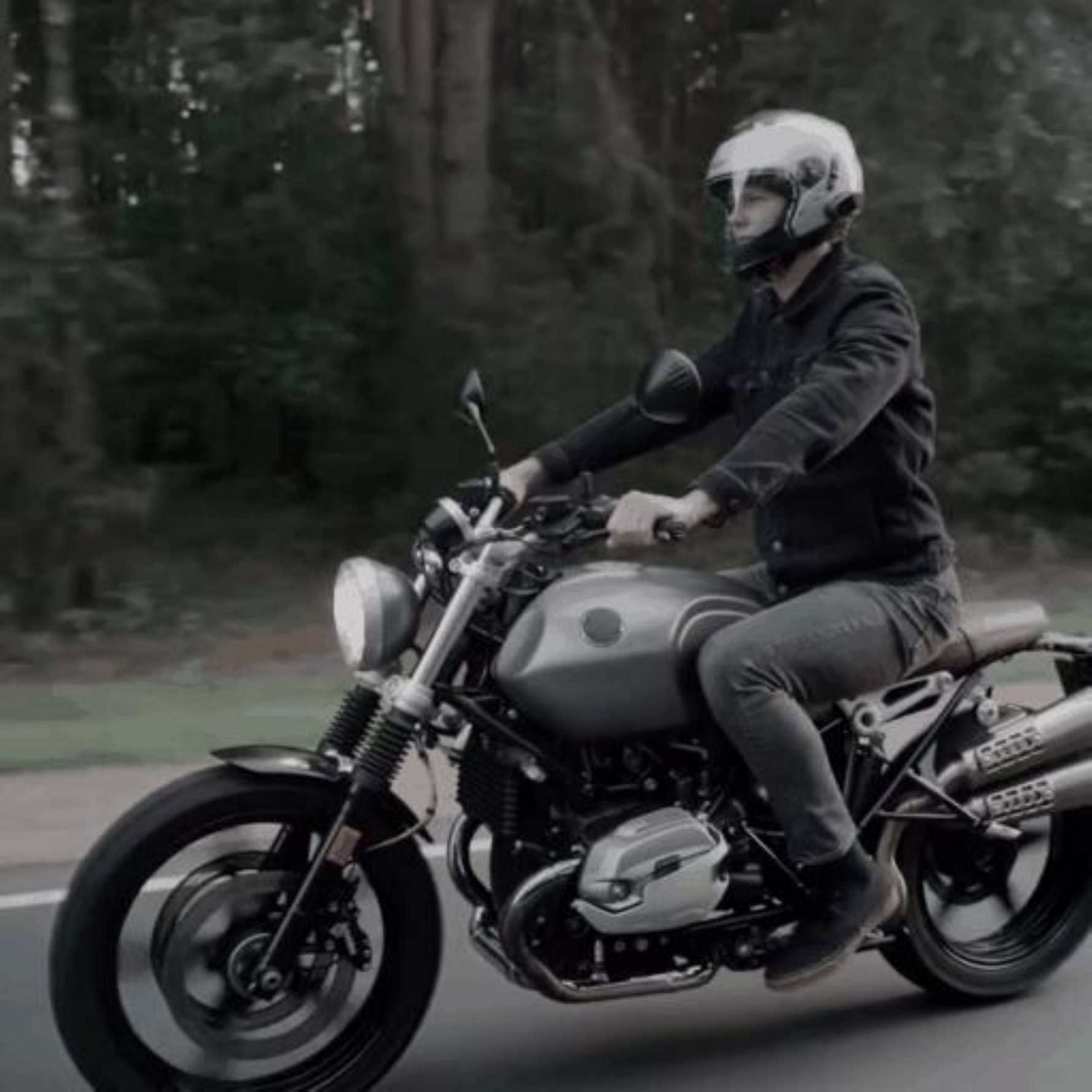}

\makebox[0.12\textwidth]{\colorbox{yellow}{\textbf{Edit-A-Video (Ours)}} A \textcolor{blue}{\textbf{gorilla}} is riding a motorcycle}\\

\includegraphics[width=0.10\textwidth]{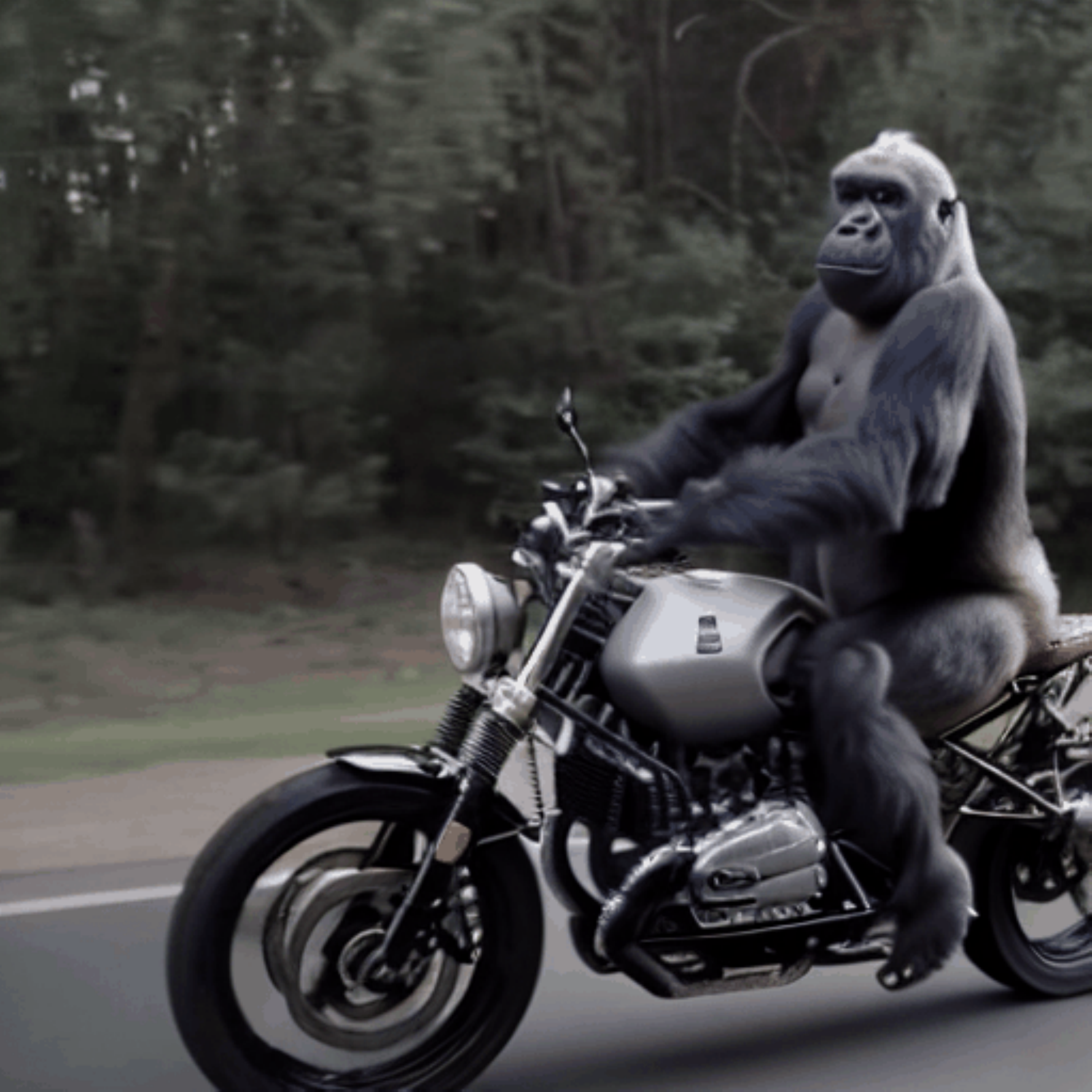}
\includegraphics[width=0.10\textwidth]{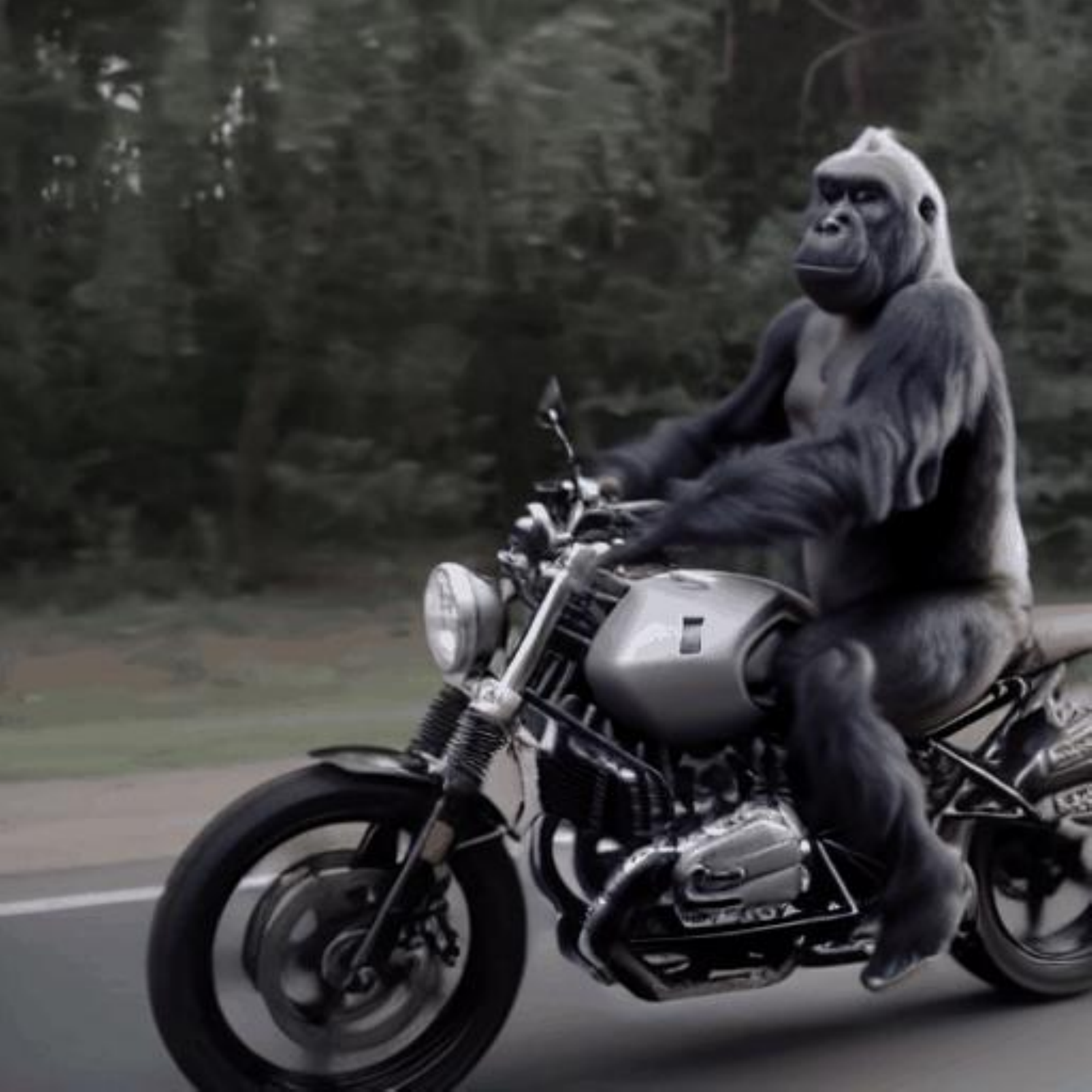}
\includegraphics[width=0.10\textwidth]{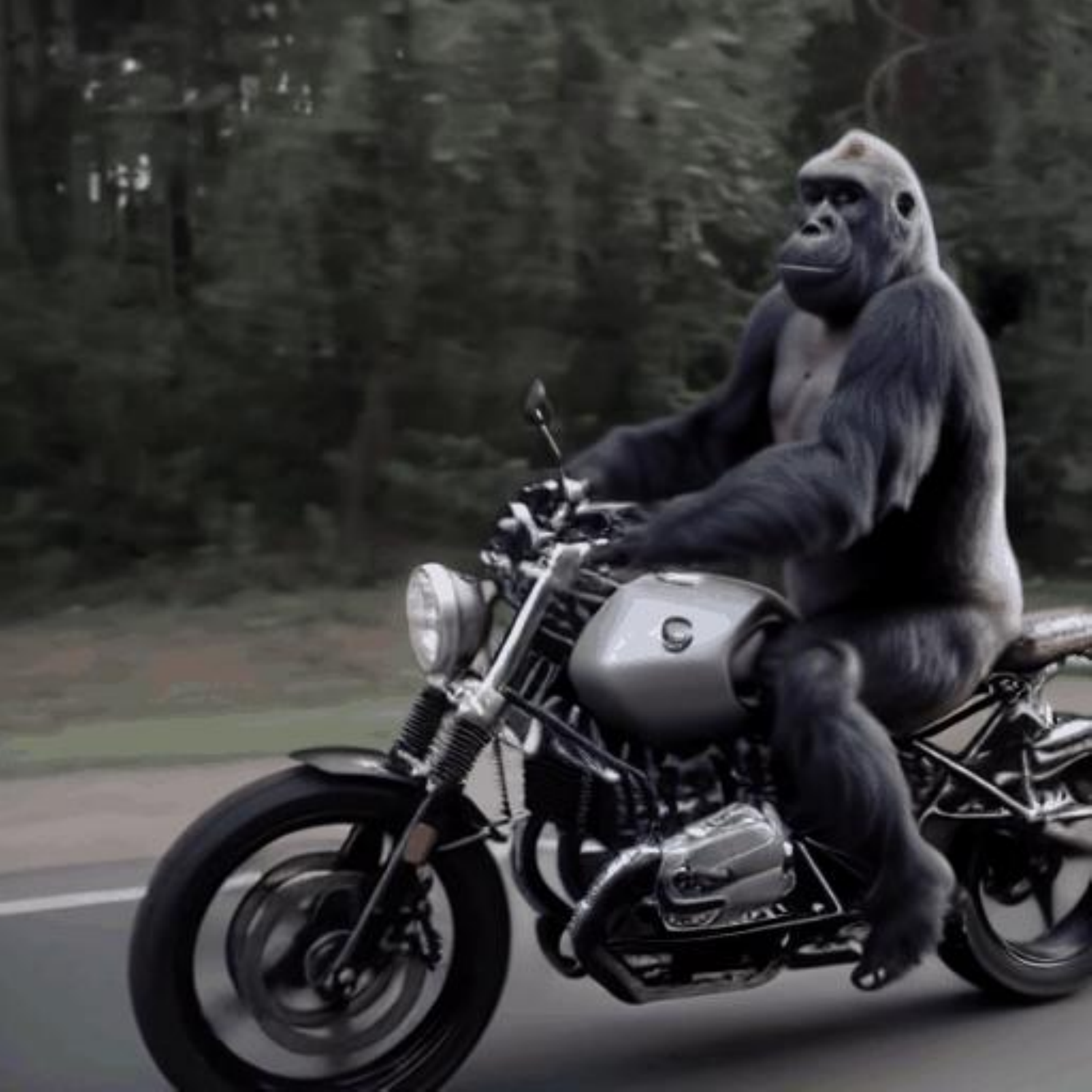}
\includegraphics[width=0.10\textwidth]{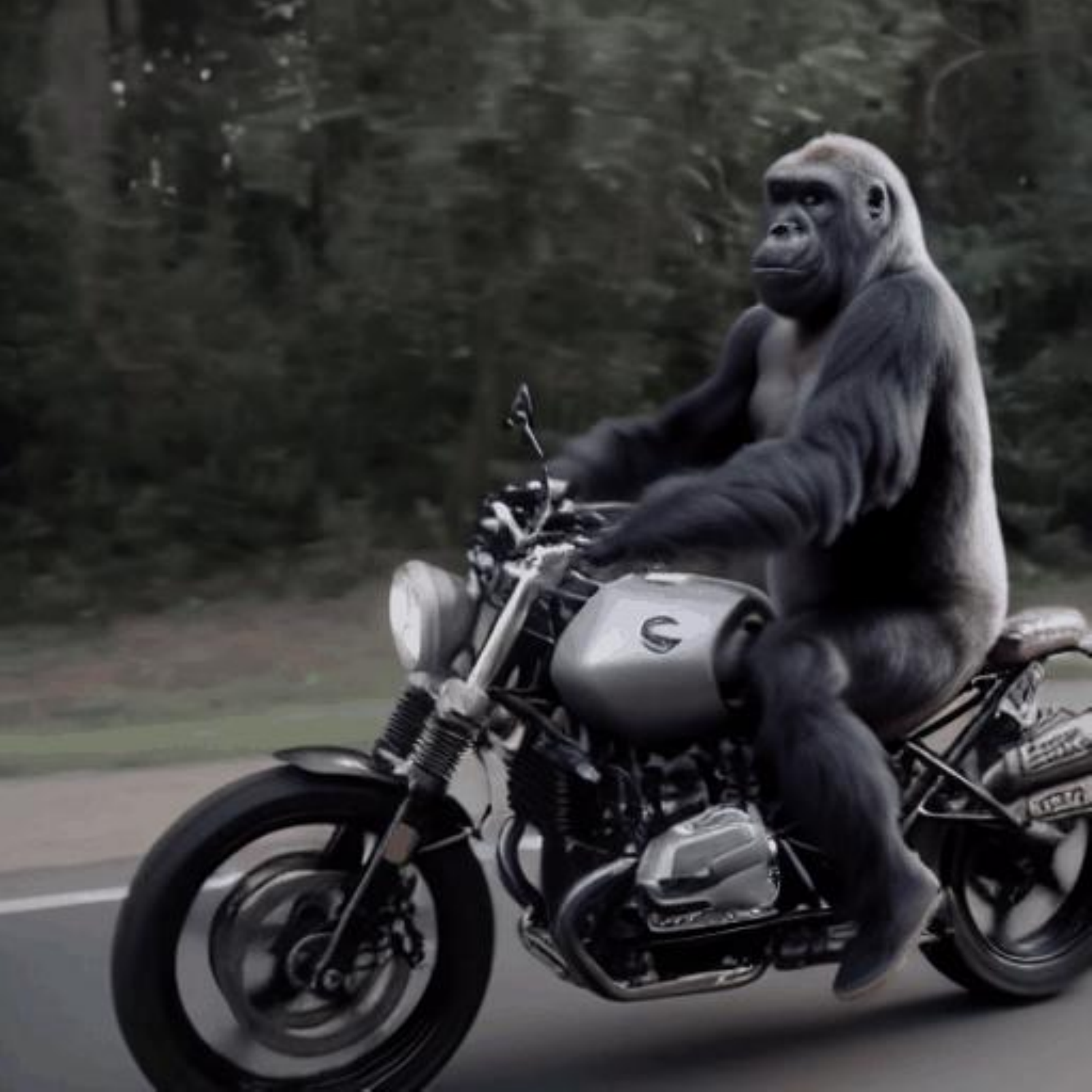}
\includegraphics[width=0.10\textwidth]{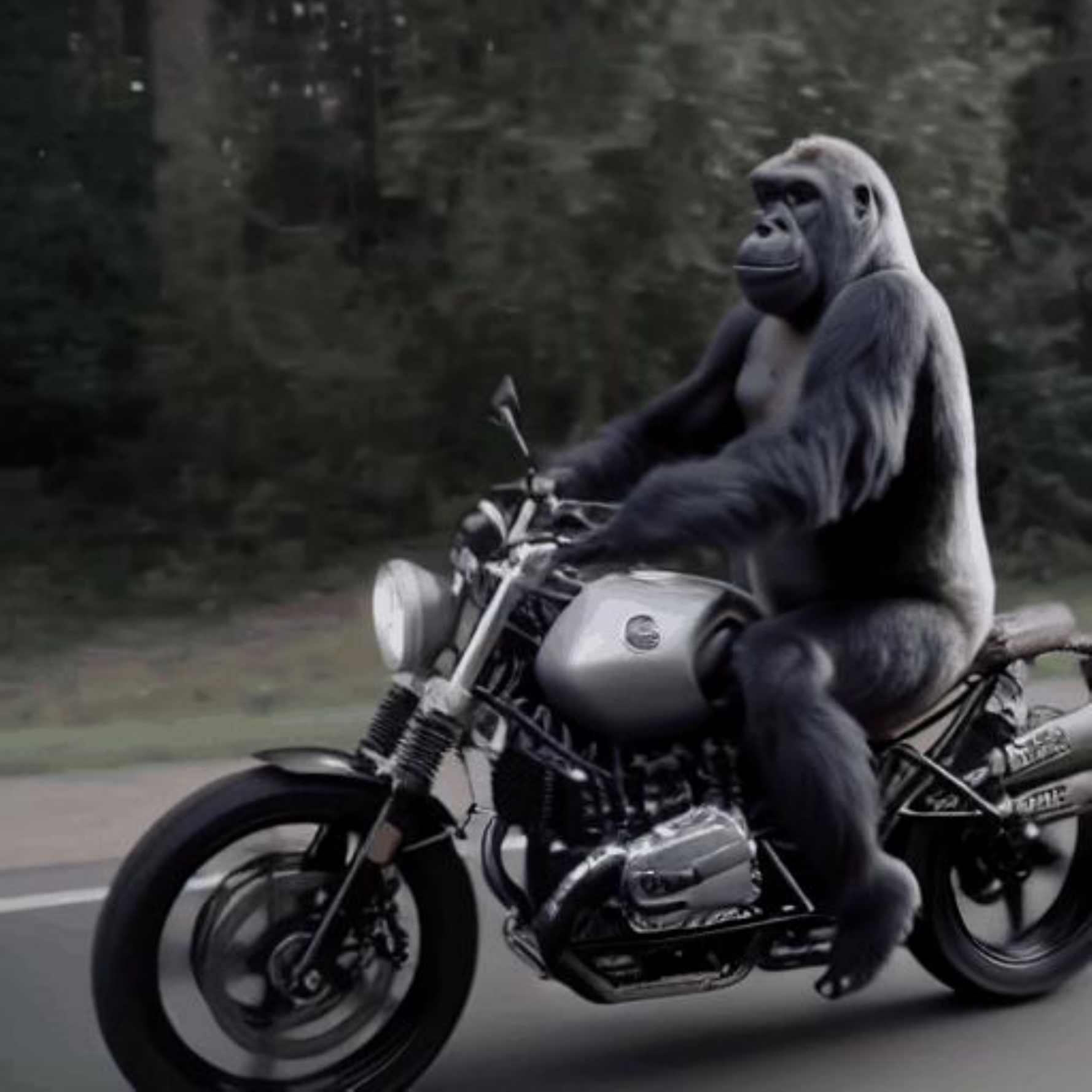}
\includegraphics[width=0.10\textwidth]{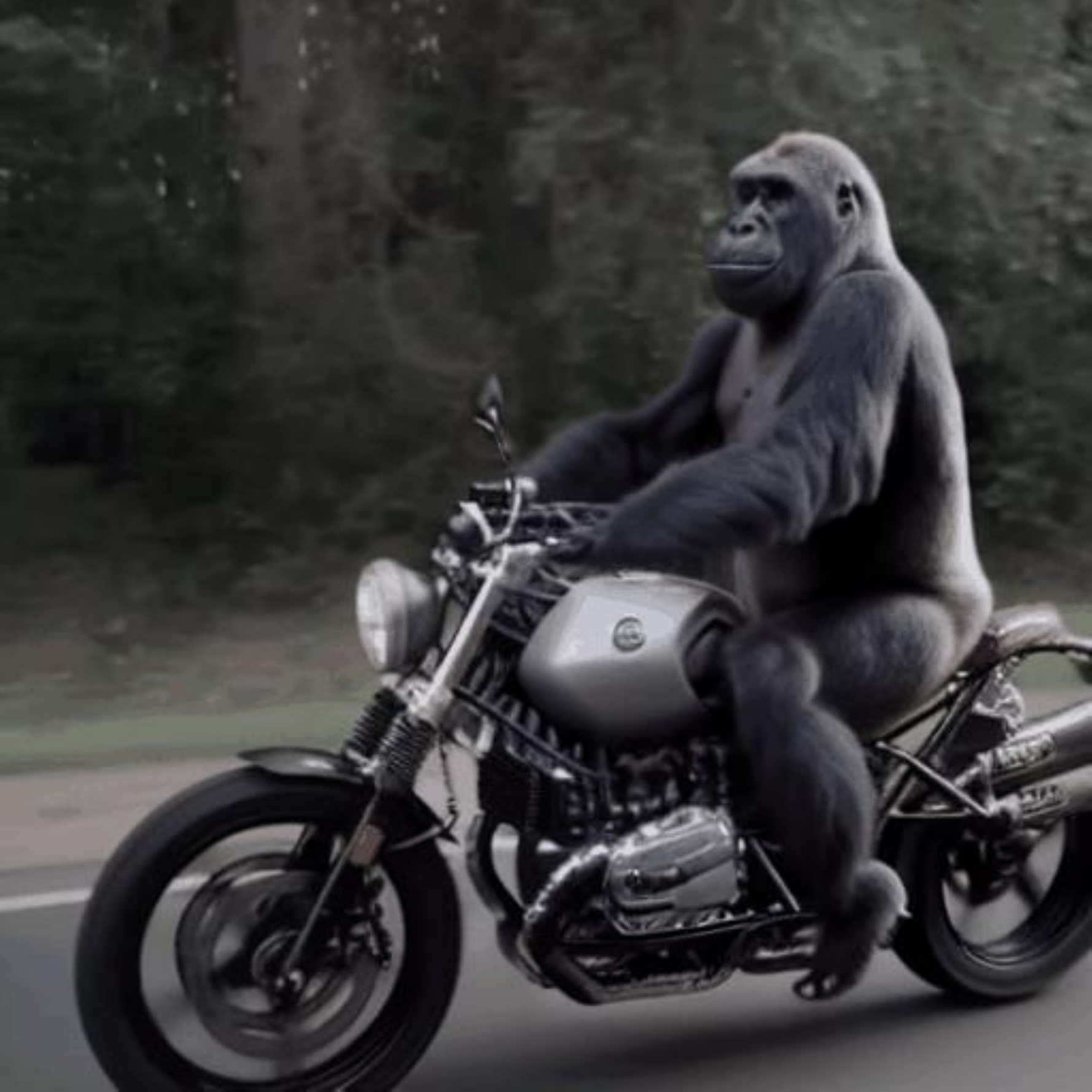}
\includegraphics[width=0.10\textwidth]{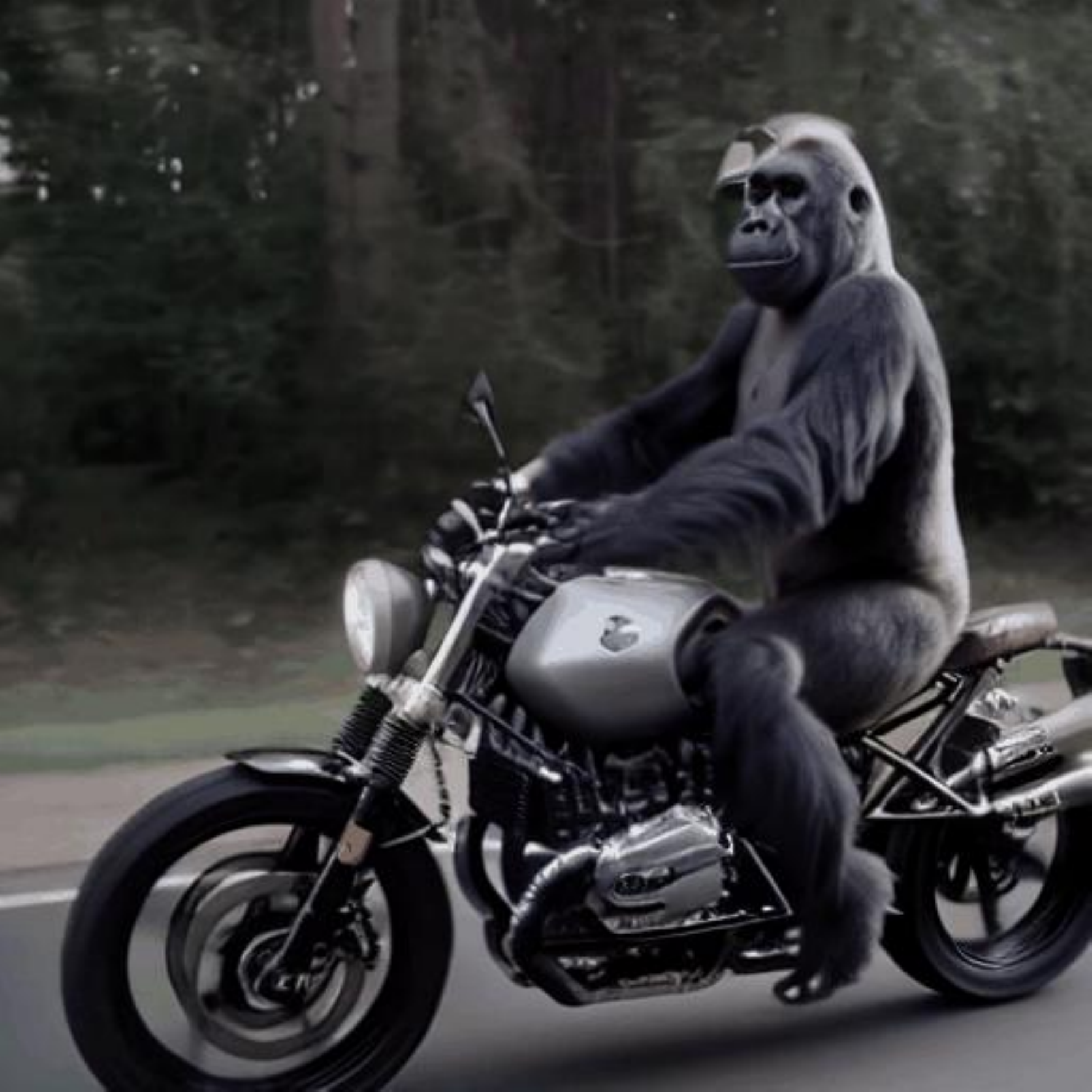}
\includegraphics[width=0.10\textwidth]{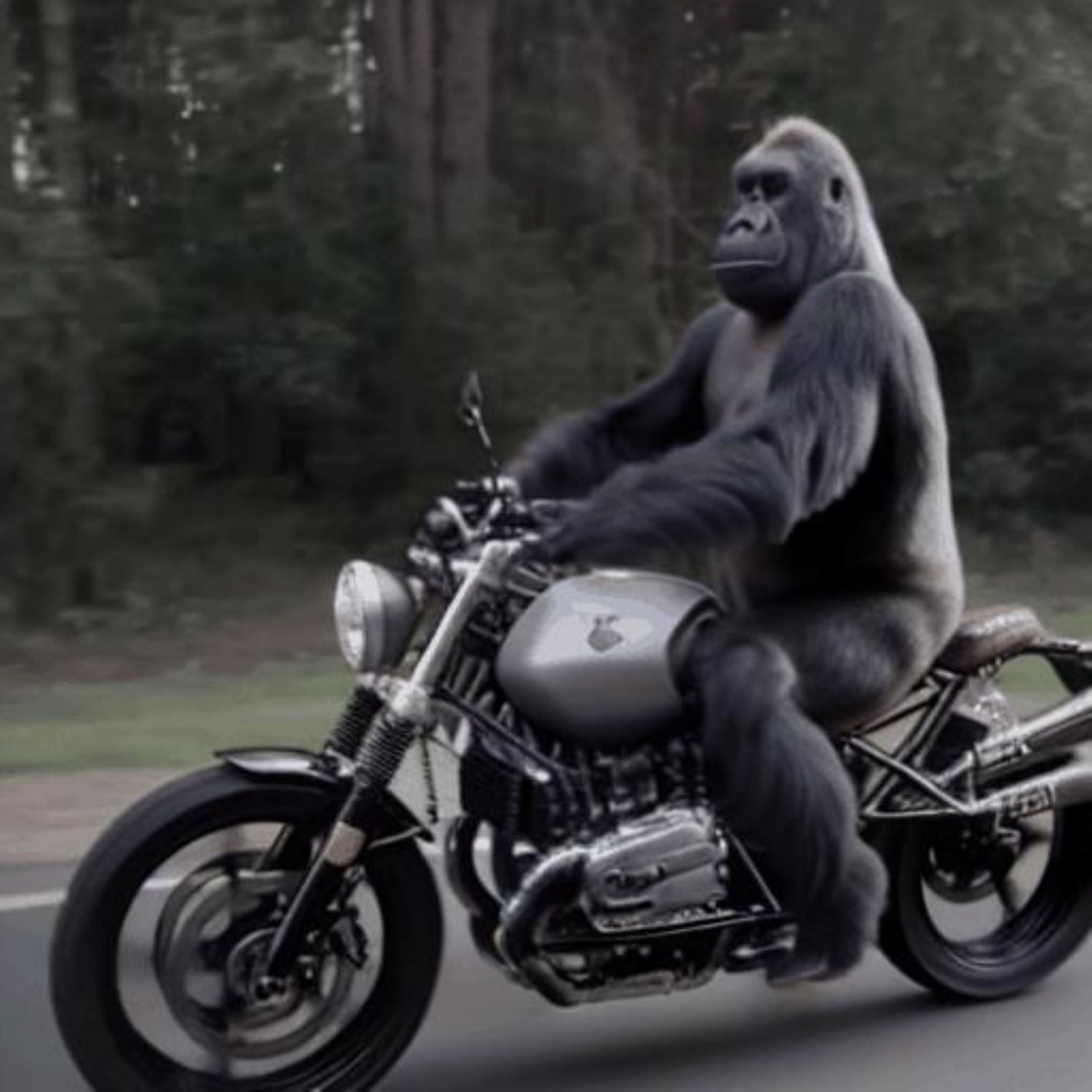}

\makebox[0.12\textwidth]{\colorbox{green}{\textbf{Tune-A-Video}} A \textcolor{blue}{\textbf{gorilla}} is riding a motorcycle}\\
\includegraphics[width=0.10\textwidth]{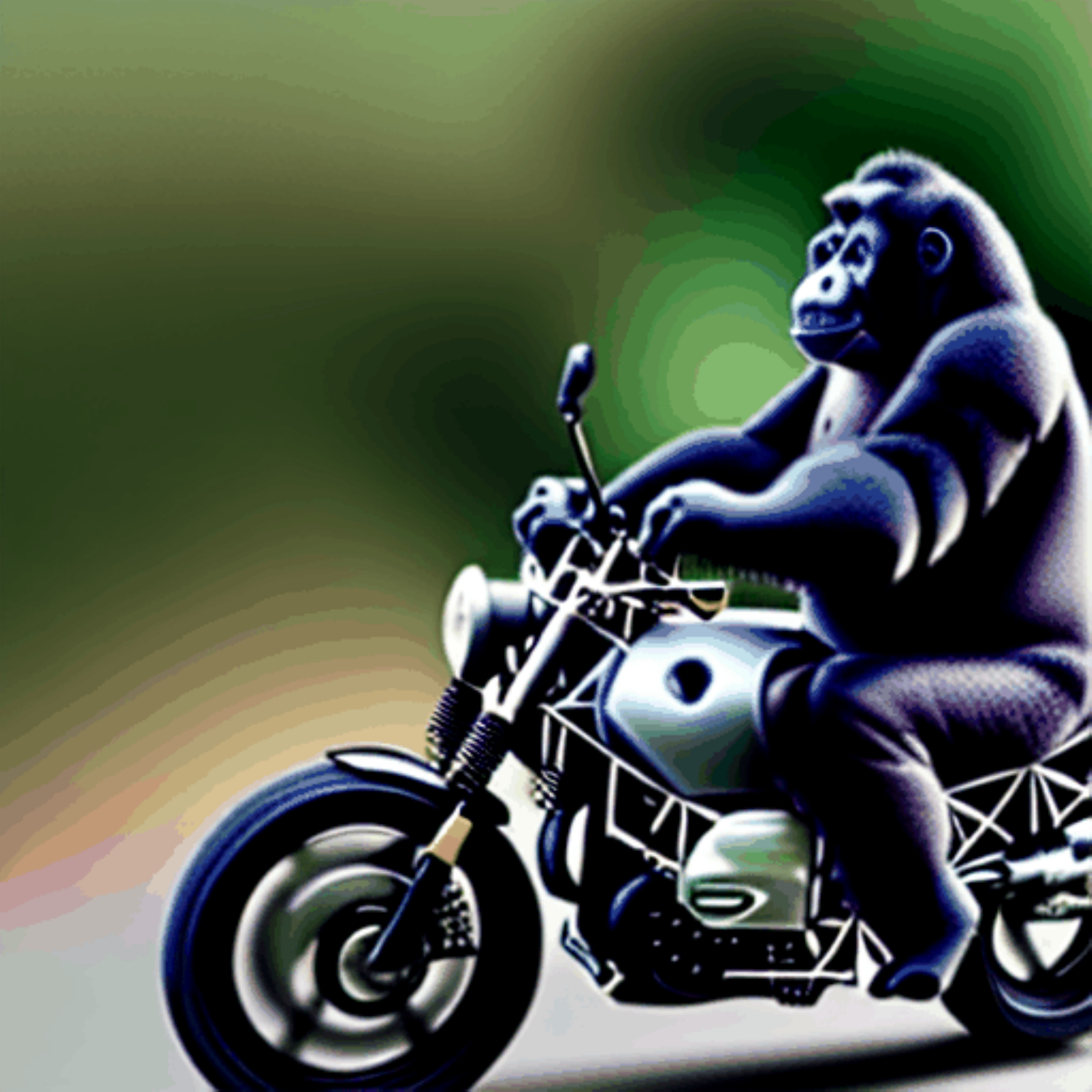}
\includegraphics[width=0.10\textwidth]{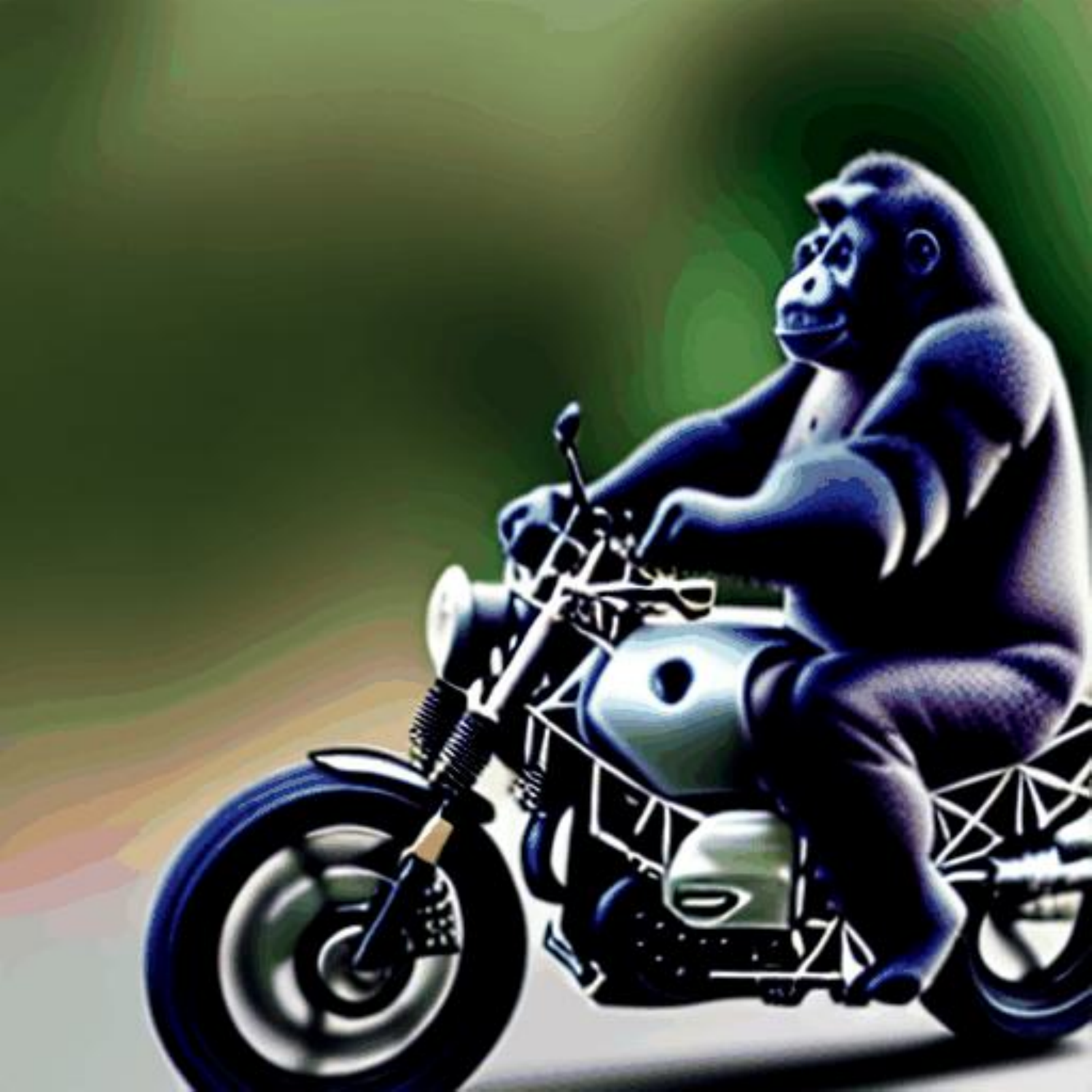}
\includegraphics[width=0.10\textwidth]{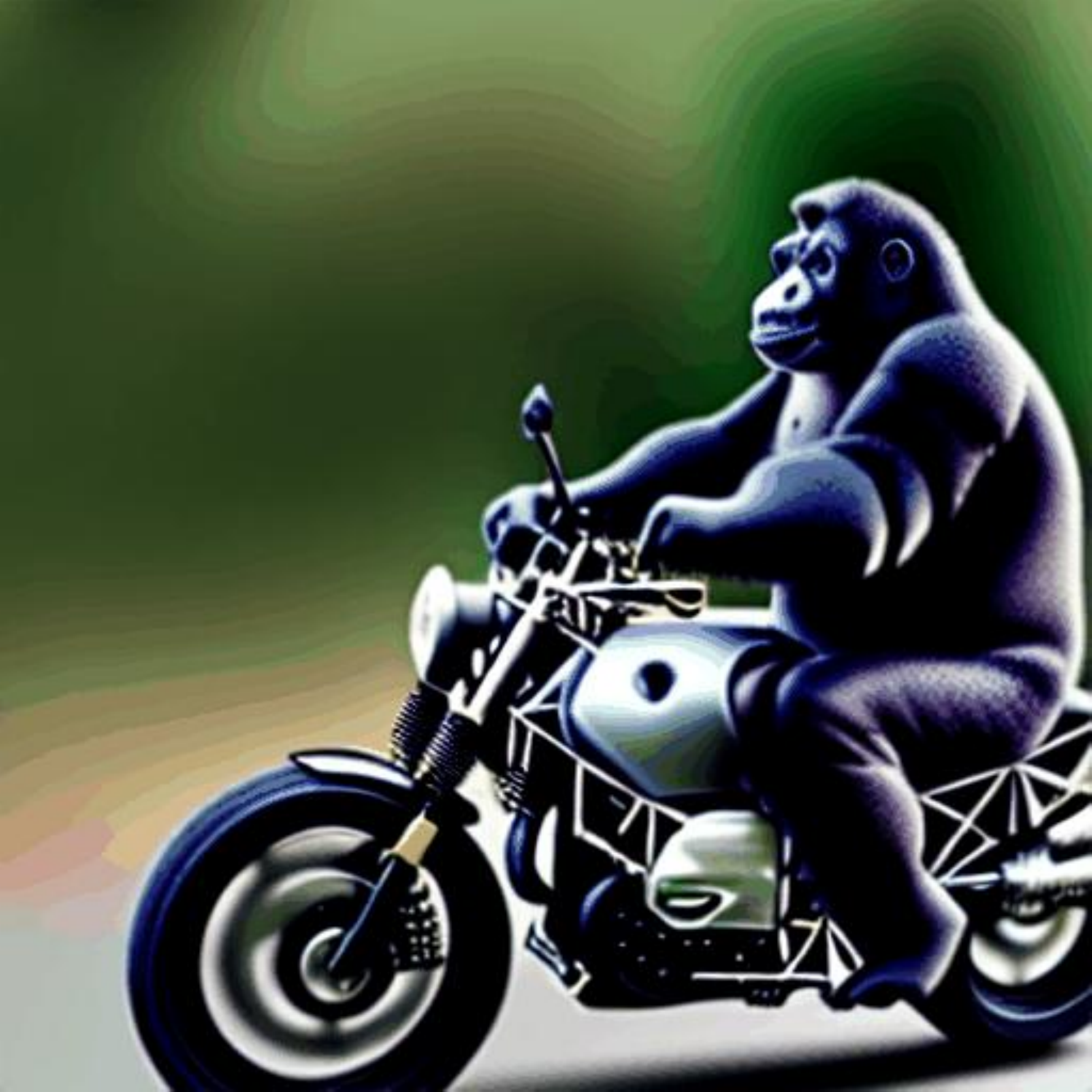}
\includegraphics[width=0.10\textwidth]{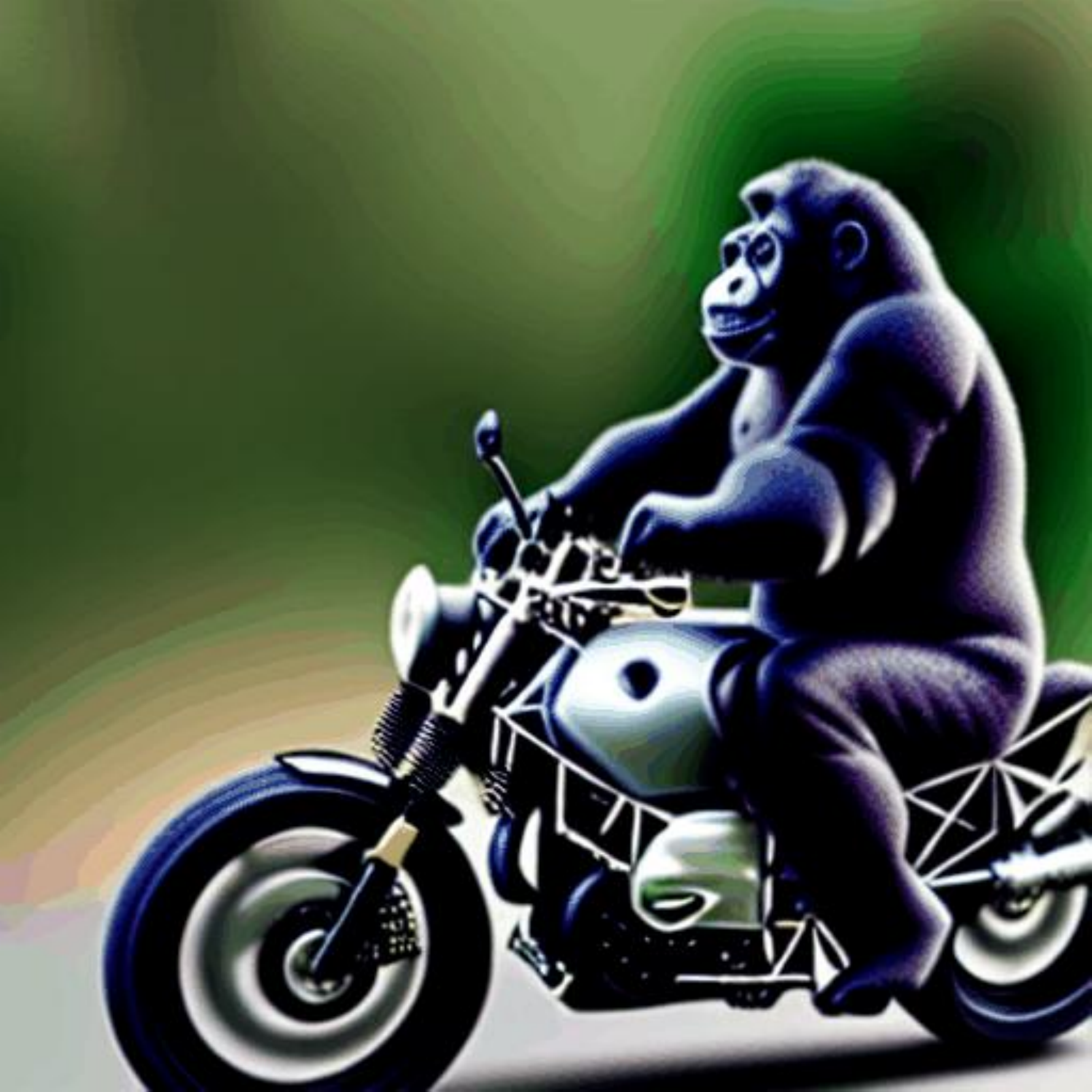}
\includegraphics[width=0.10\textwidth]{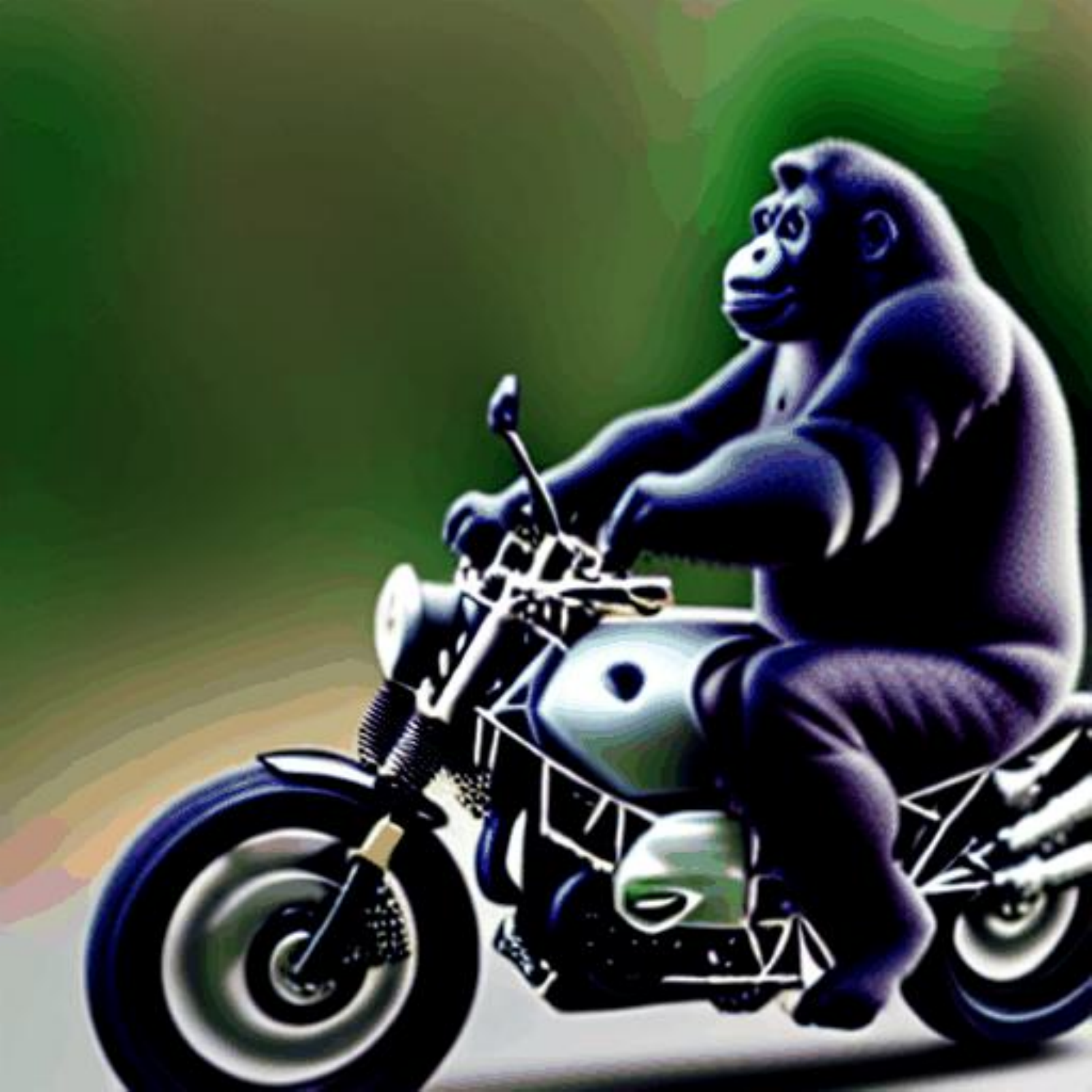}
\includegraphics[width=0.10\textwidth]{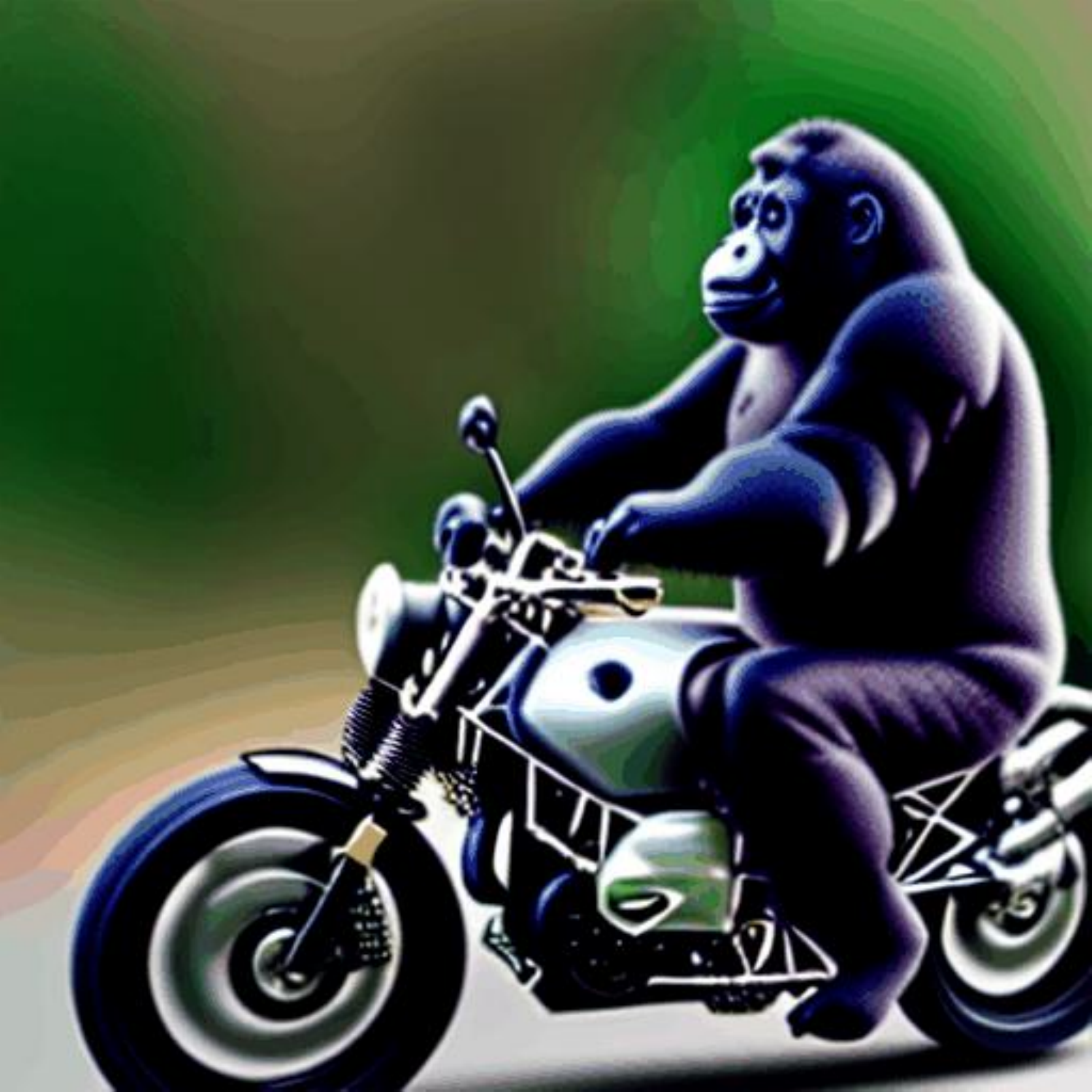}
\includegraphics[width=0.10\textwidth]{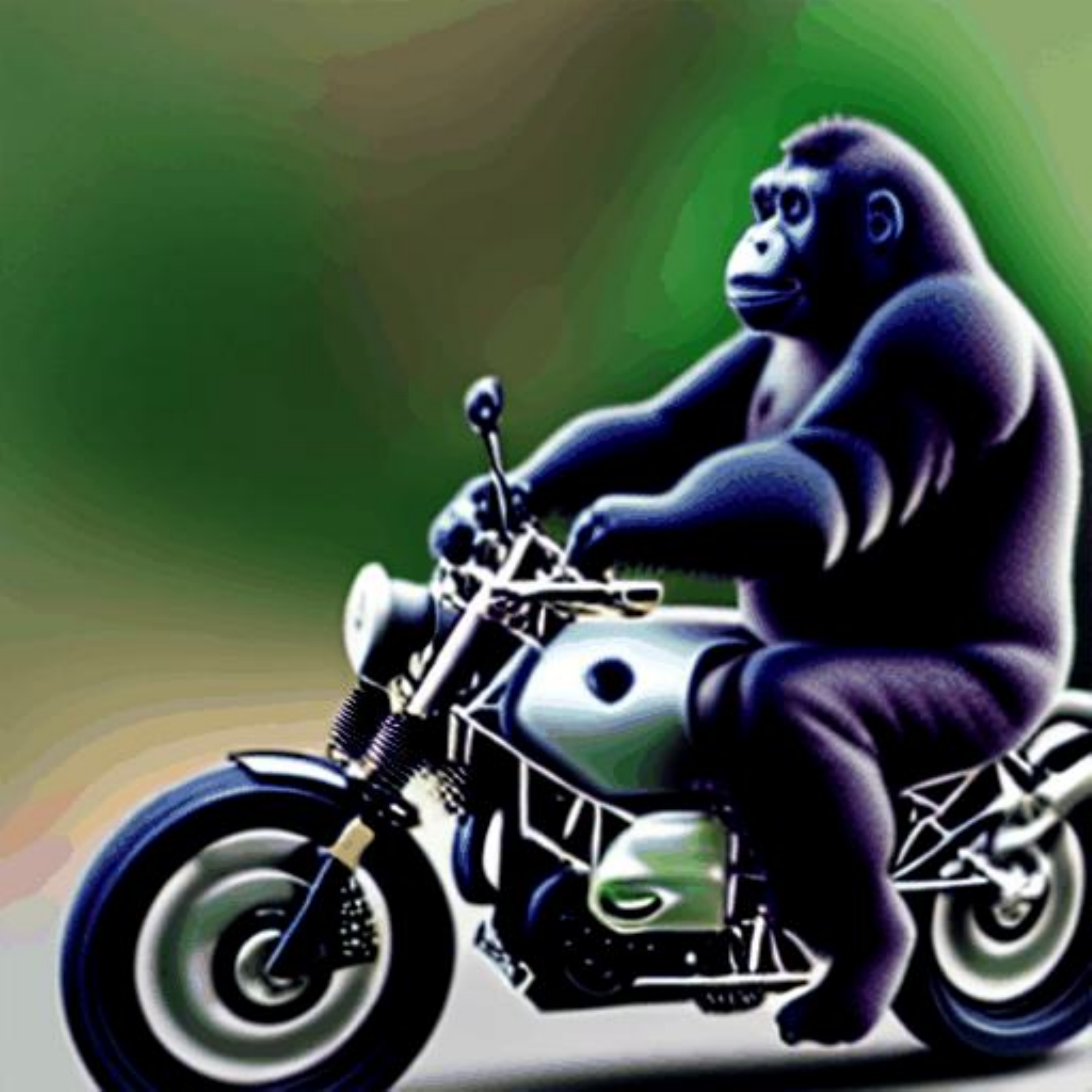}
\includegraphics[width=0.10\textwidth]{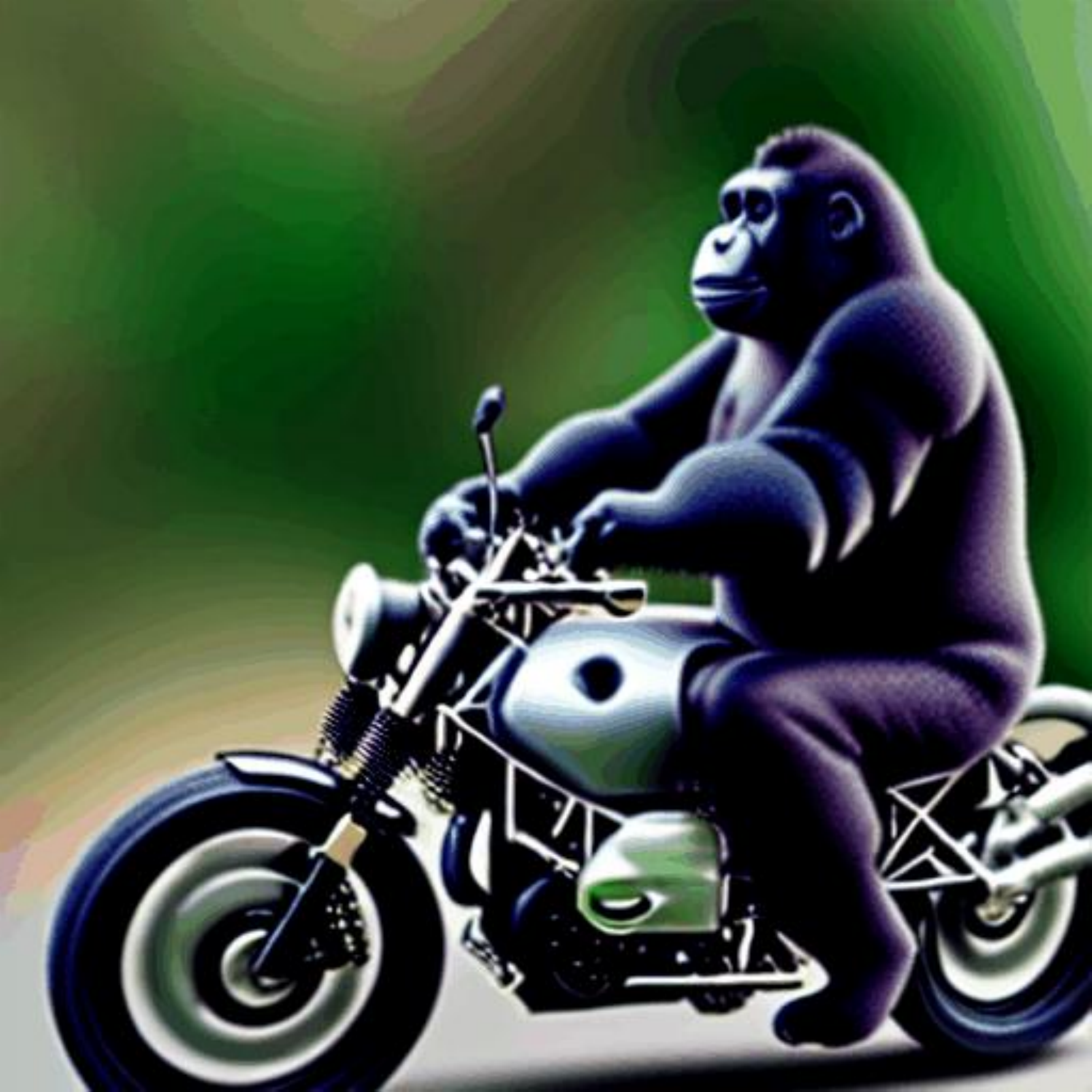}

\makebox[0.12\textwidth]{\colorbox{green}{\textbf{SDEdit}} A \textcolor{blue}{\textbf{gorilla}} is riding a motorcycle}\\
\includegraphics[width=0.10\textwidth]{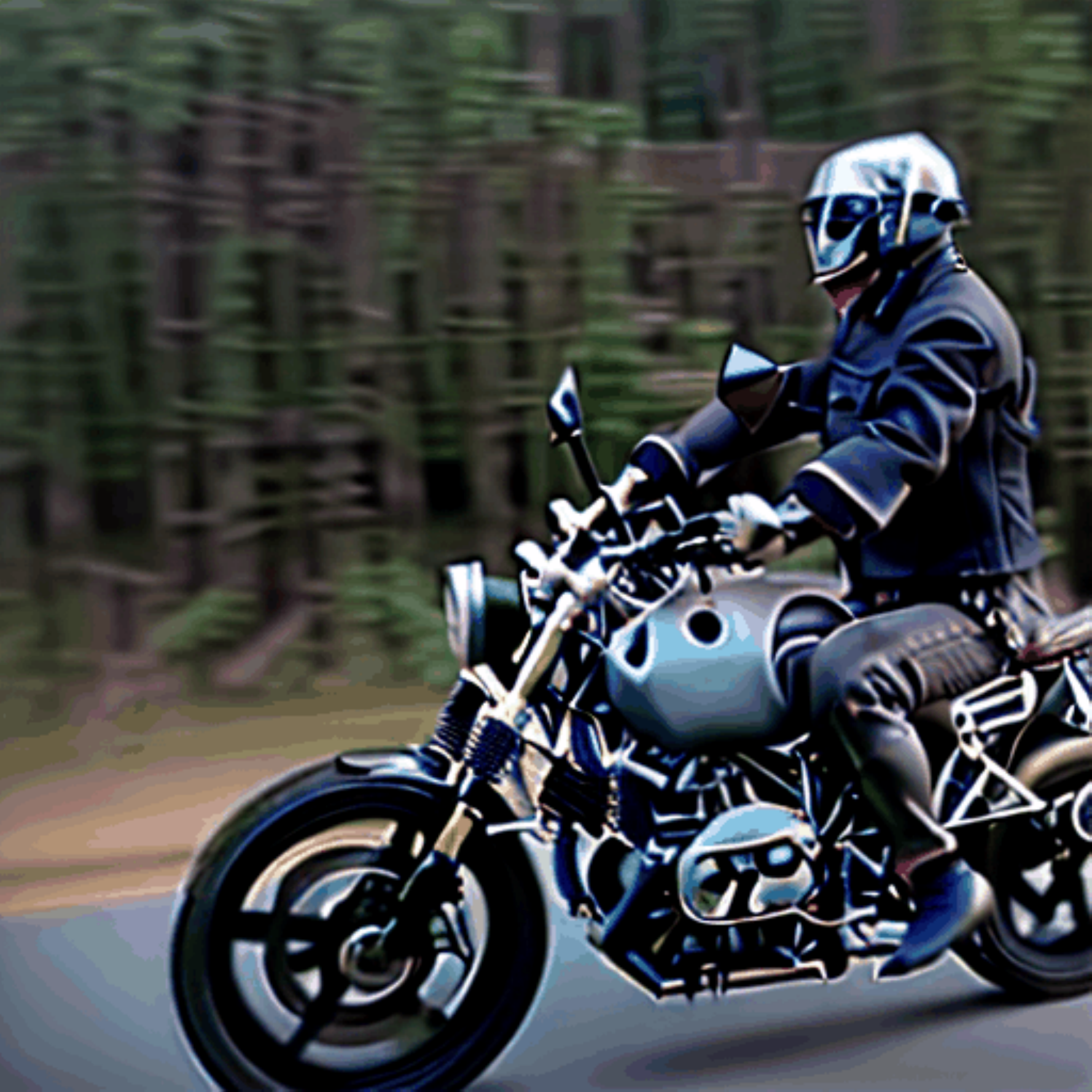}
\includegraphics[width=0.10\textwidth]{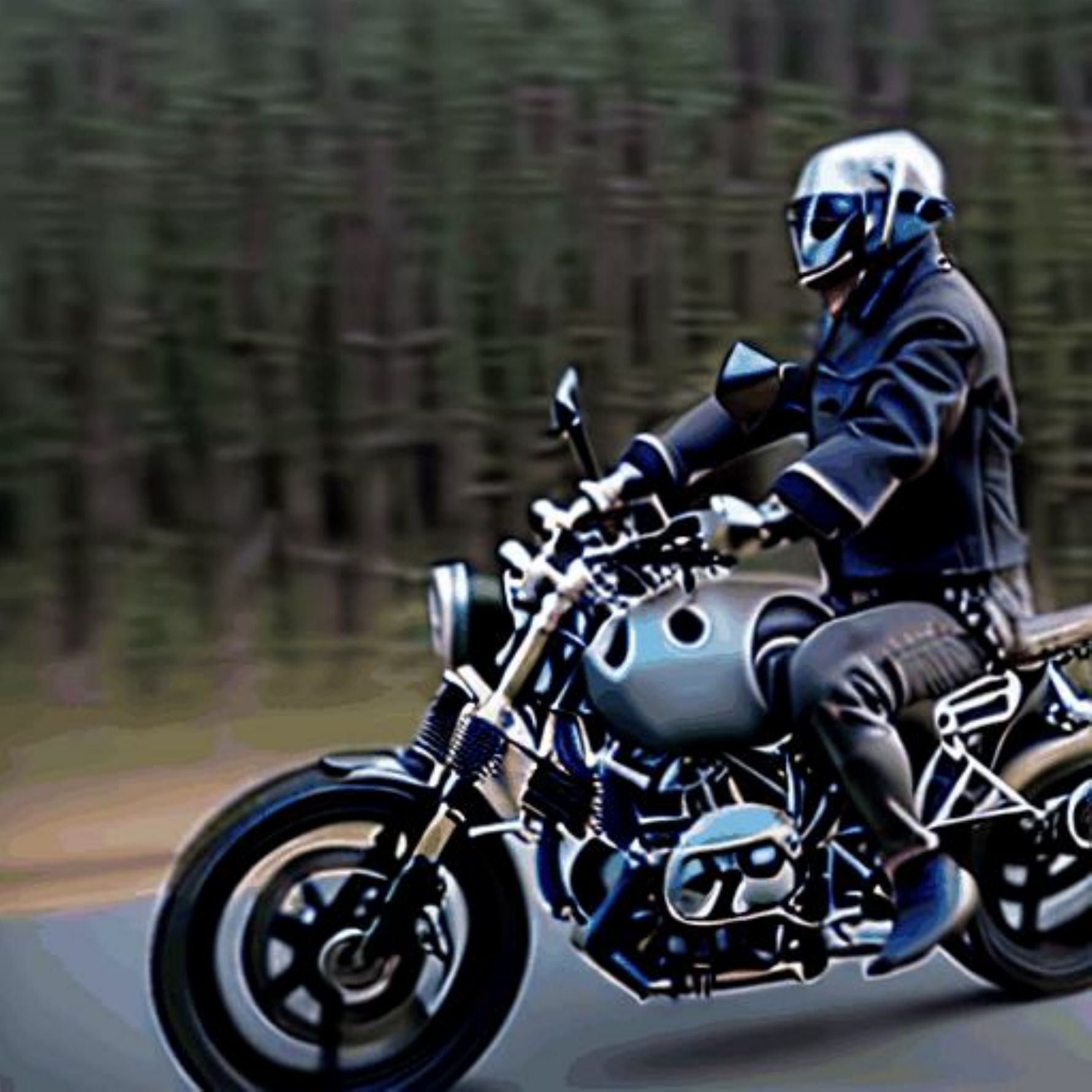}
\includegraphics[width=0.10\textwidth]{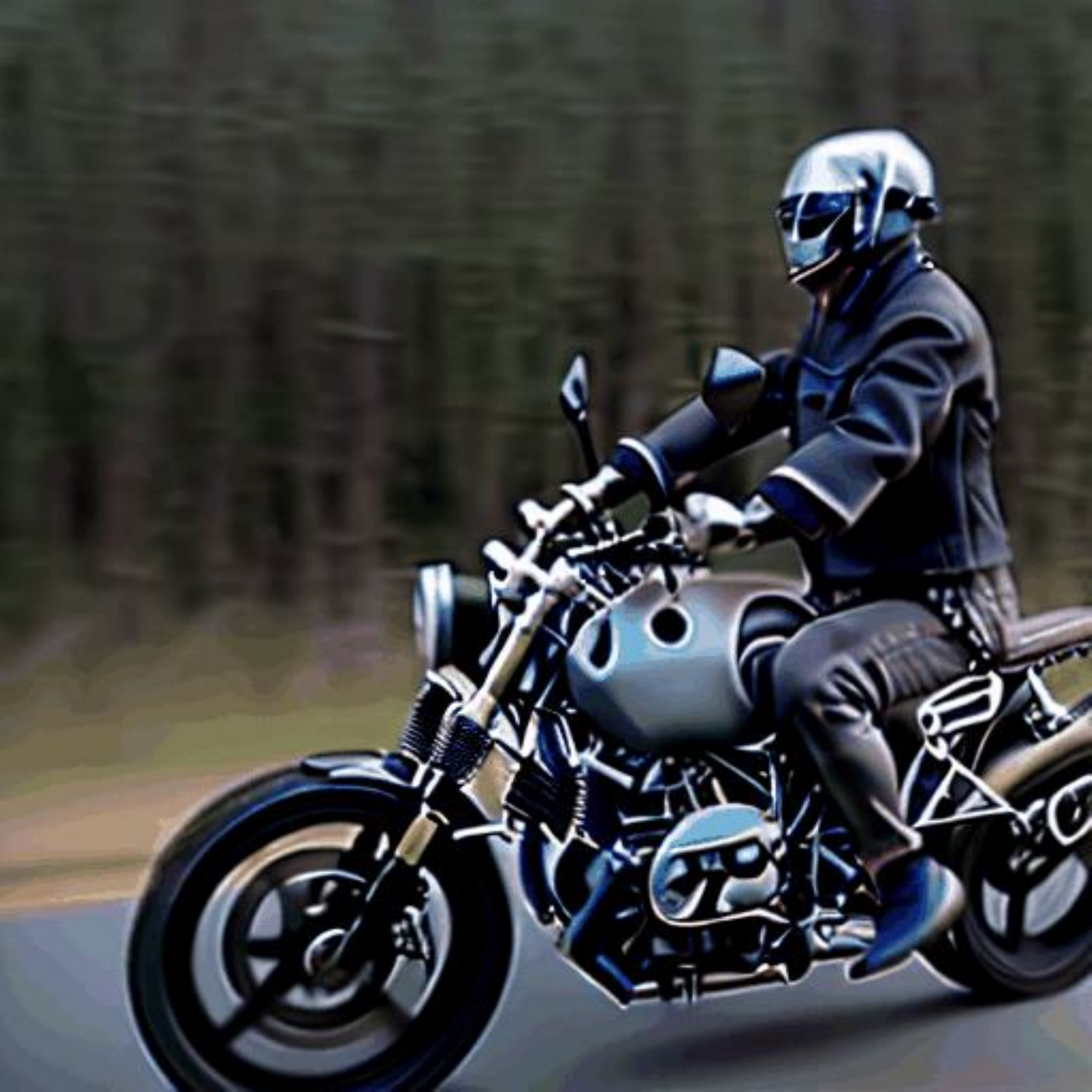}
\includegraphics[width=0.10\textwidth]{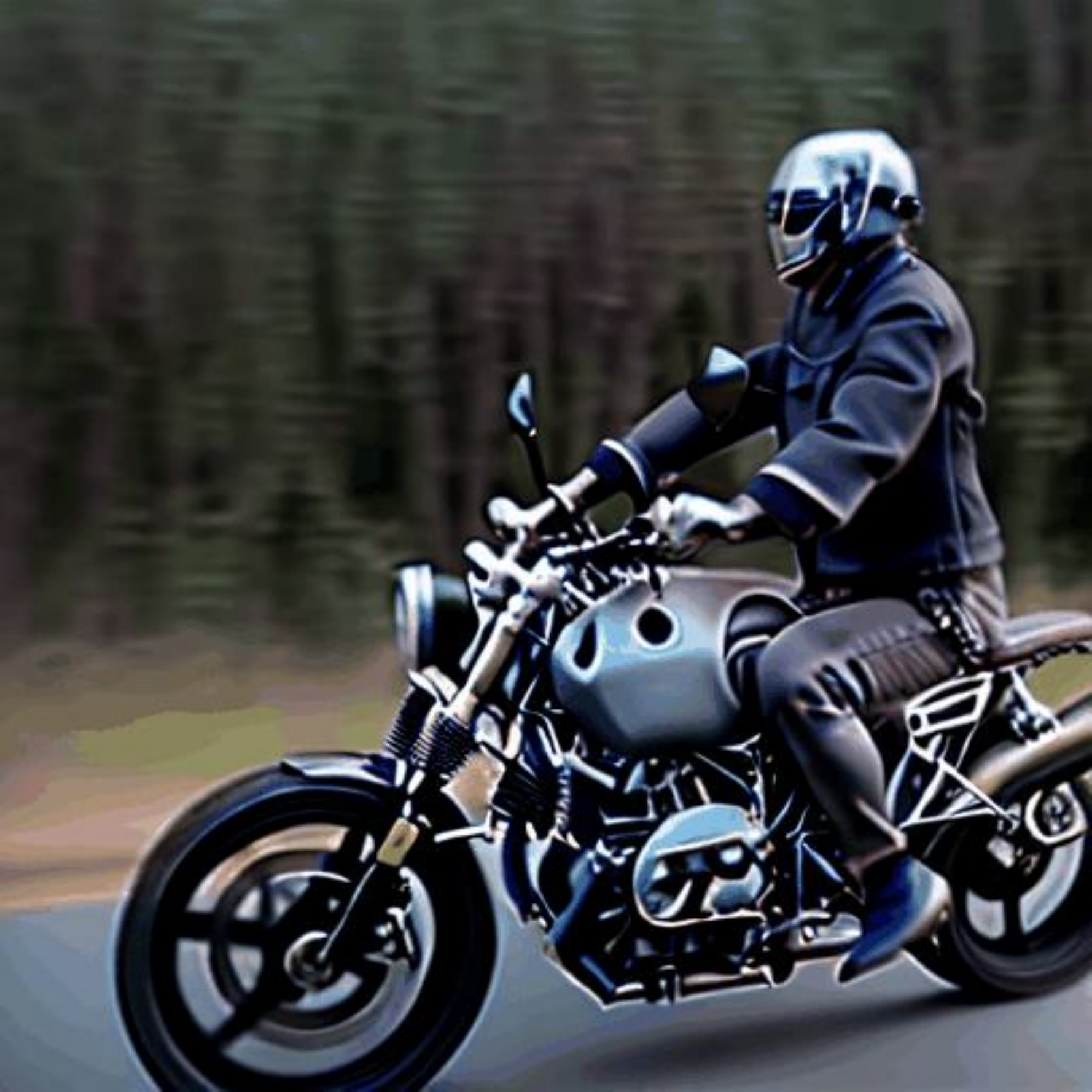}
\includegraphics[width=0.10\textwidth]{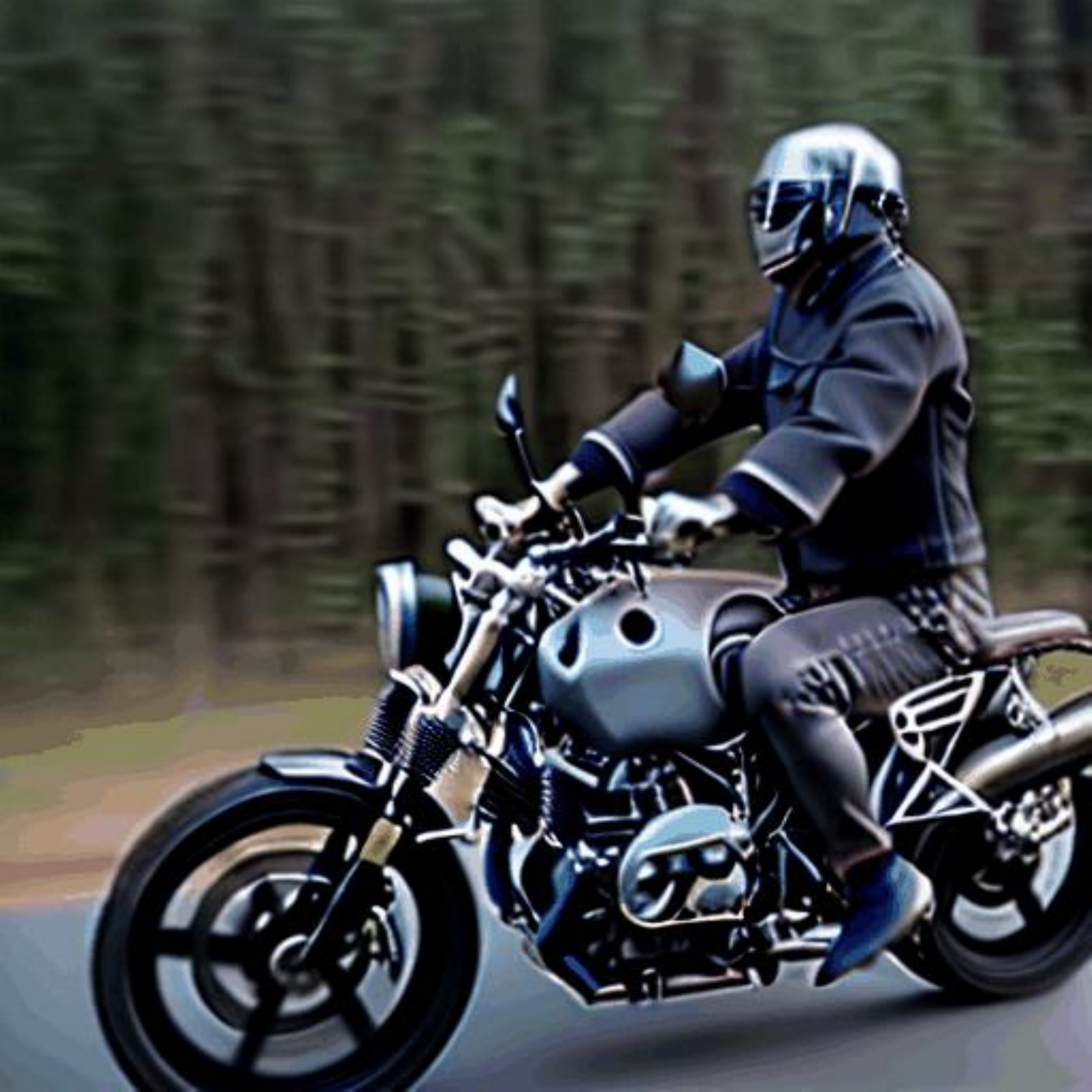}
\includegraphics[width=0.10\textwidth]{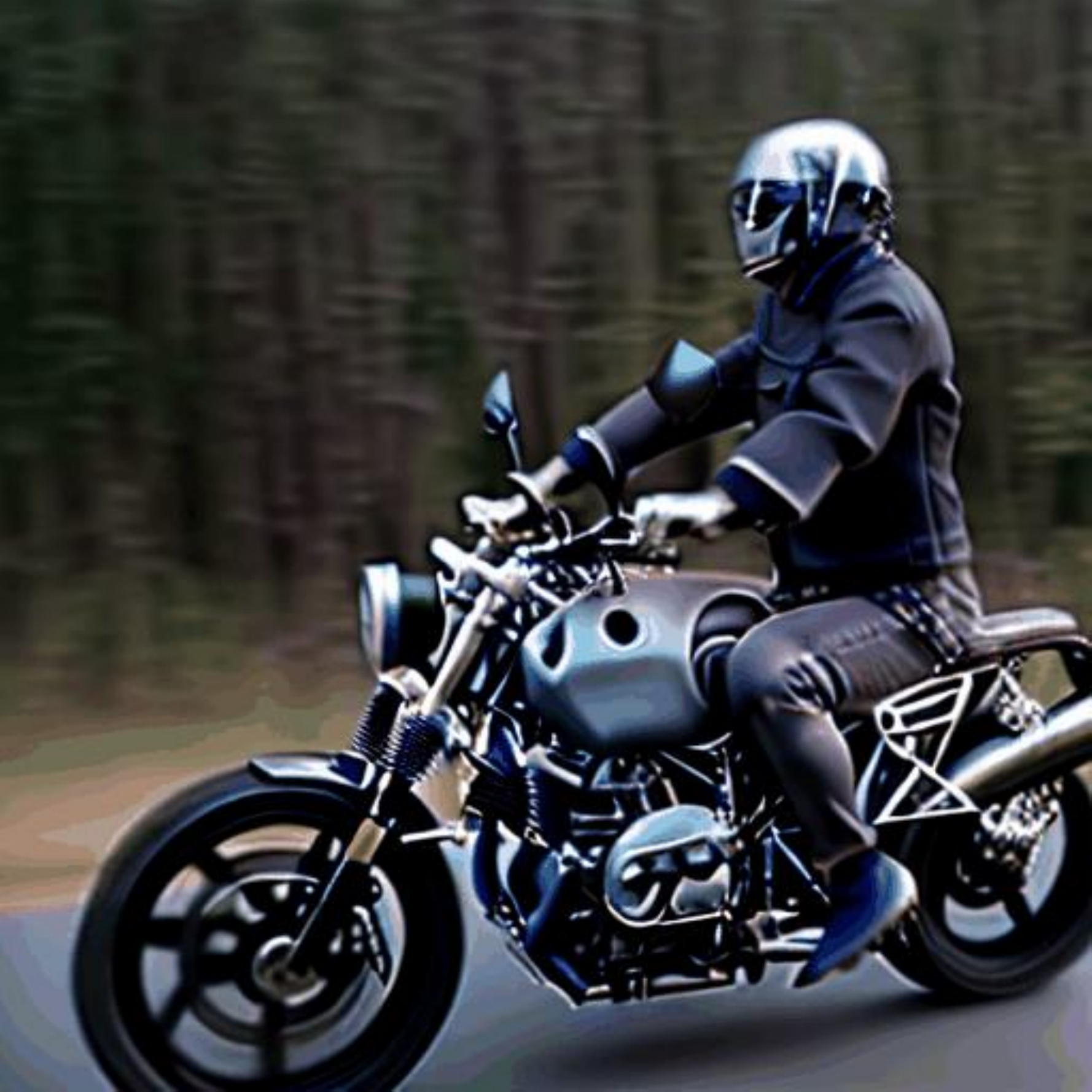}
\includegraphics[width=0.10\textwidth]{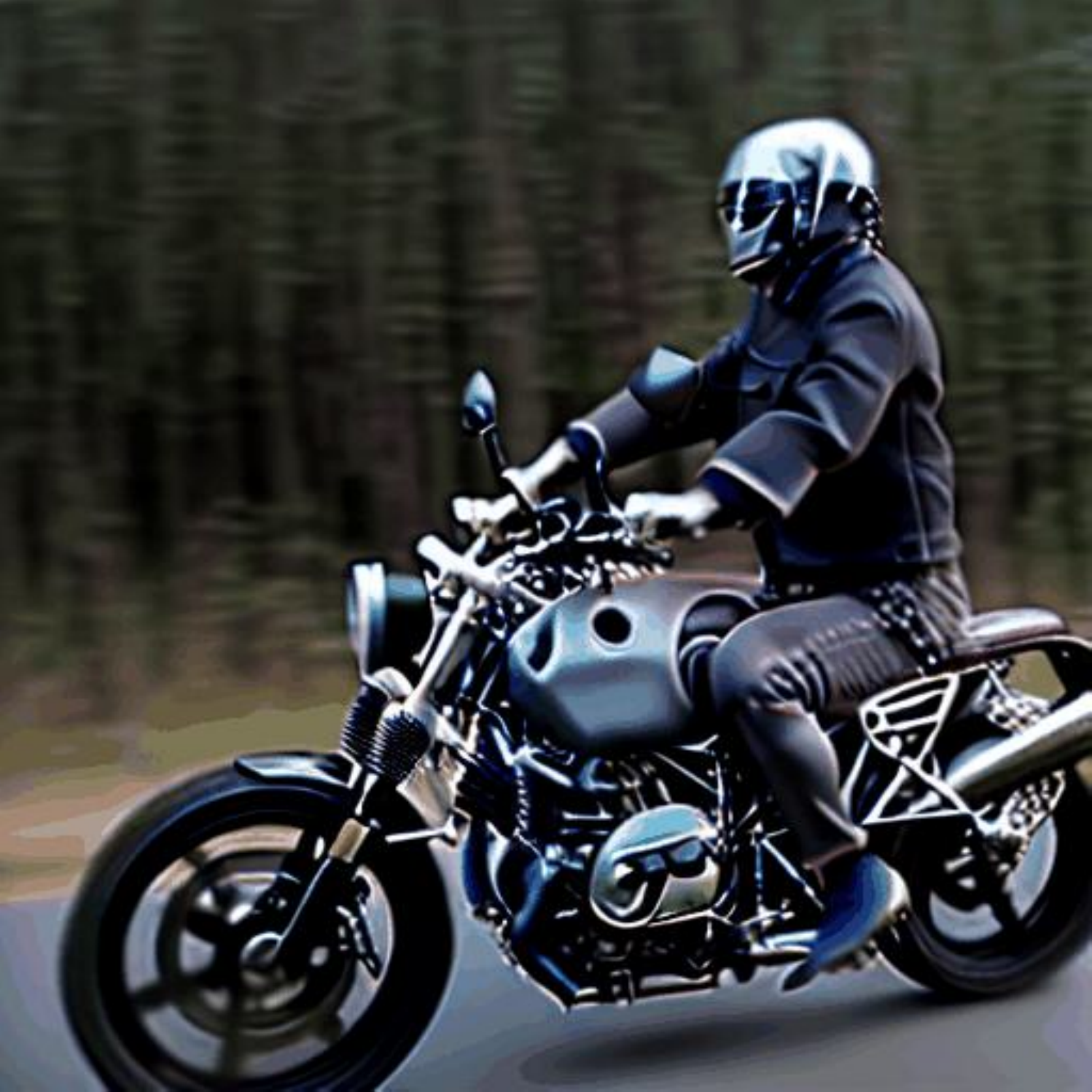}
\includegraphics[width=0.10\textwidth]{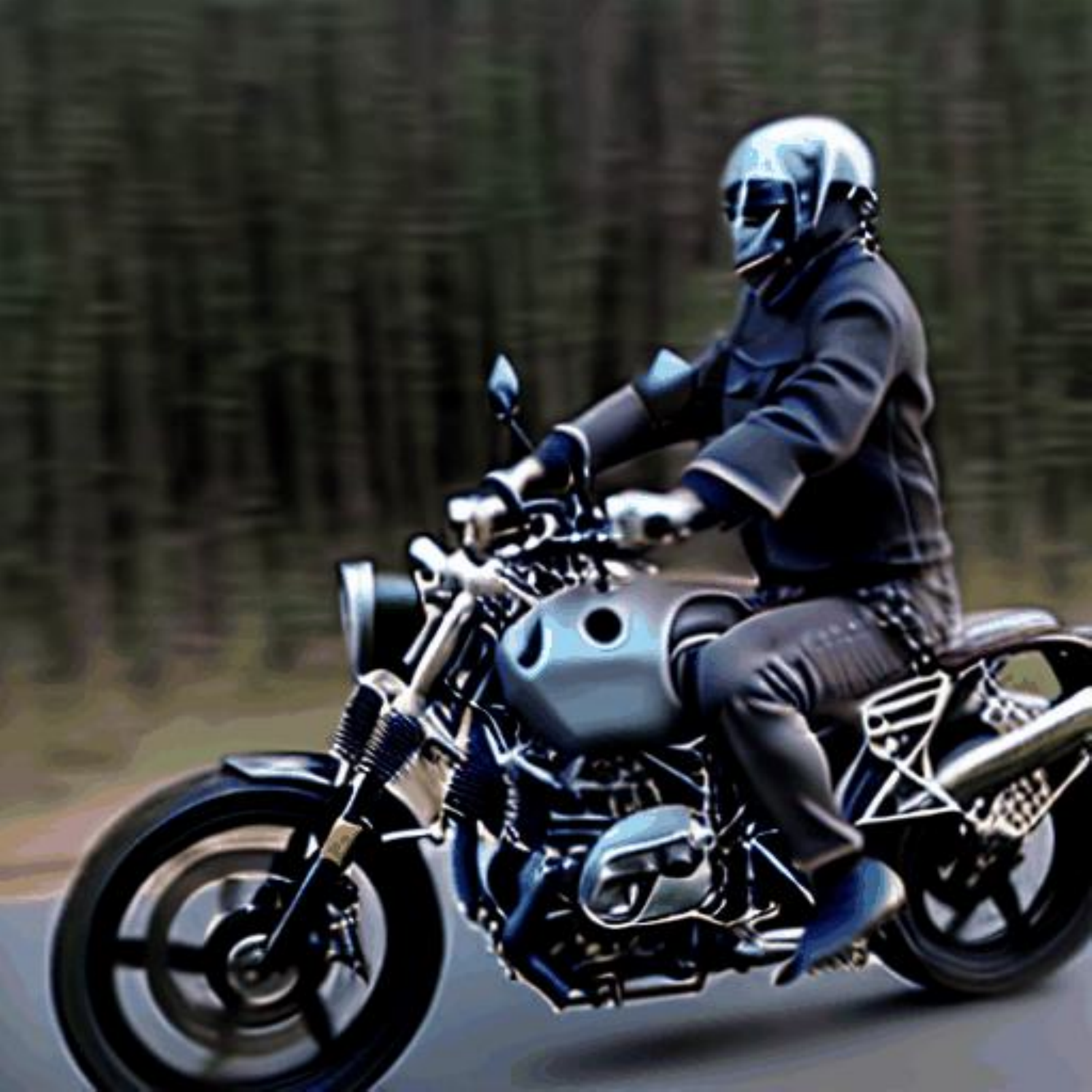}



\makebox[0.12\textwidth]{\colorbox{green}{\textbf{Video-P2P}} A \textcolor{blue}{\textbf{gorilla}} is riding a motorcycle}\\

\includegraphics[width=0.10\textwidth]{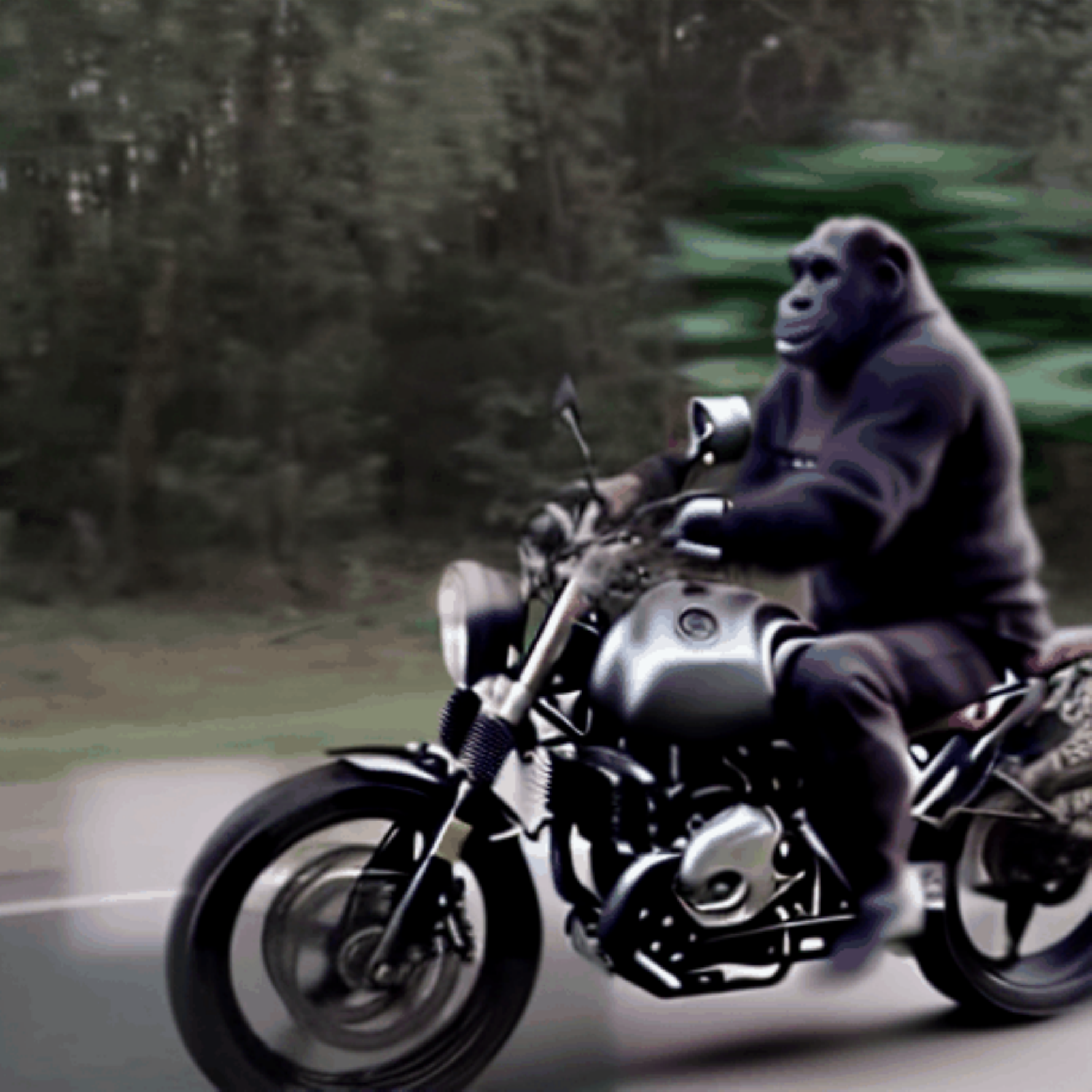}
\includegraphics[width=0.10\textwidth]{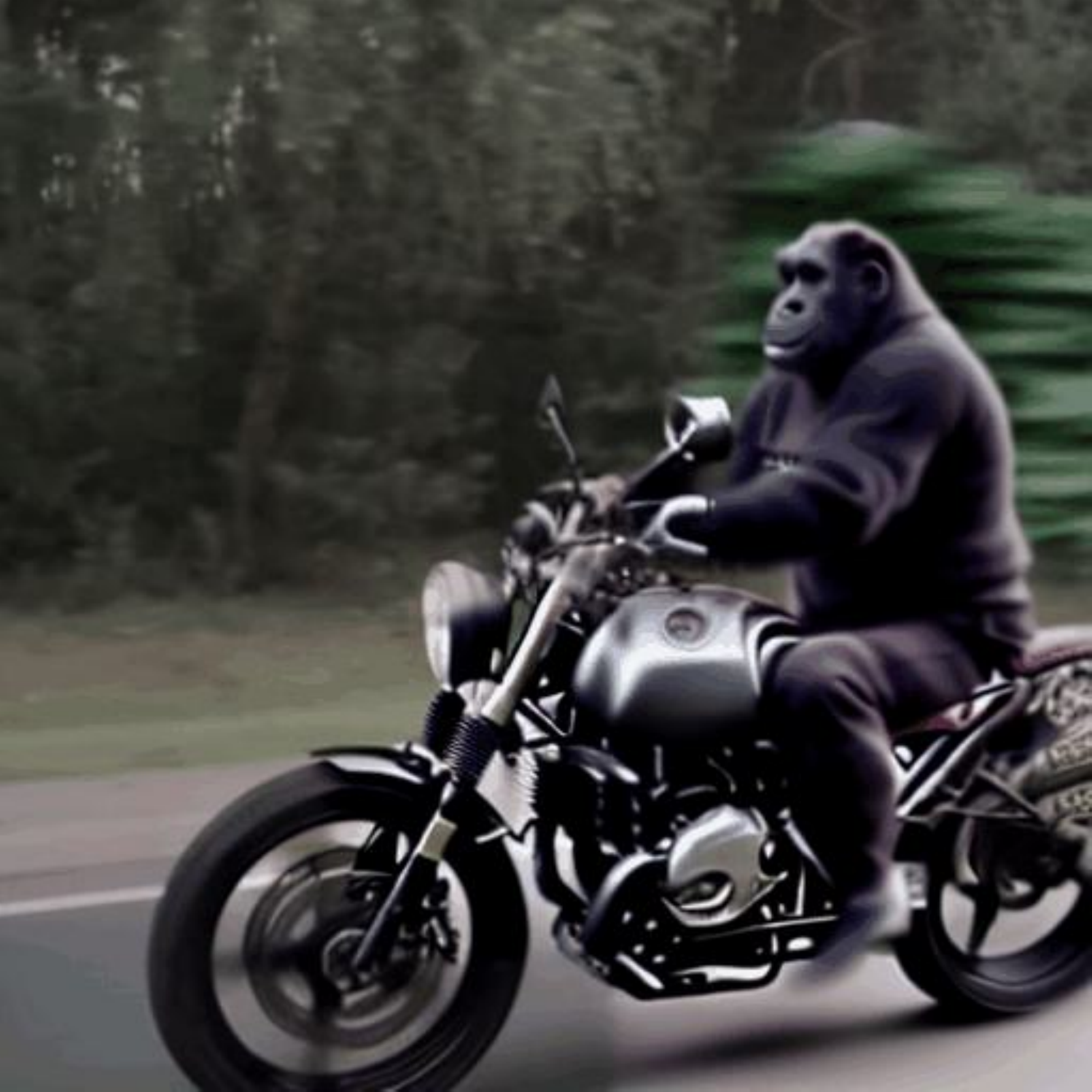}
\includegraphics[width=0.10\textwidth]{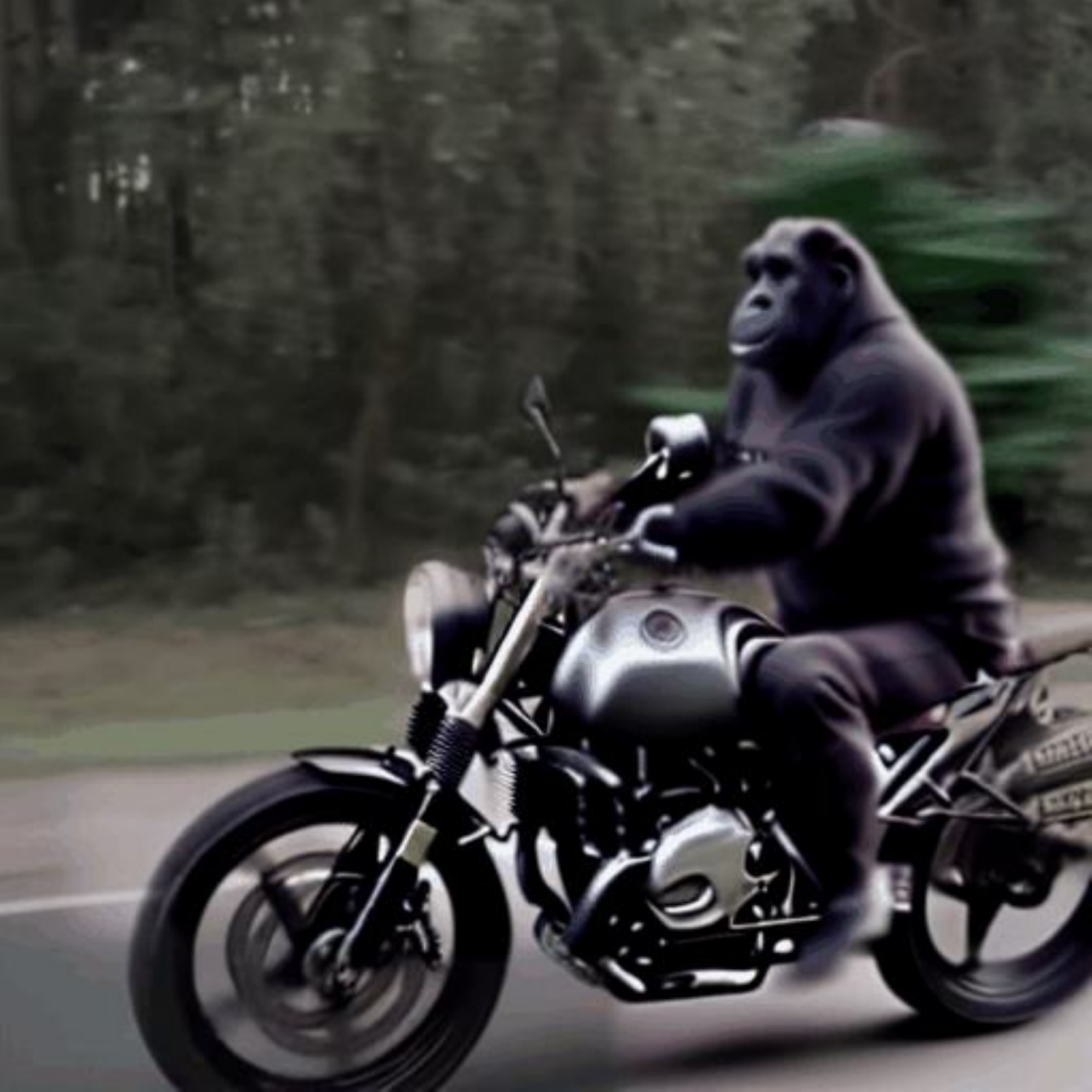}
\includegraphics[width=0.10\textwidth]{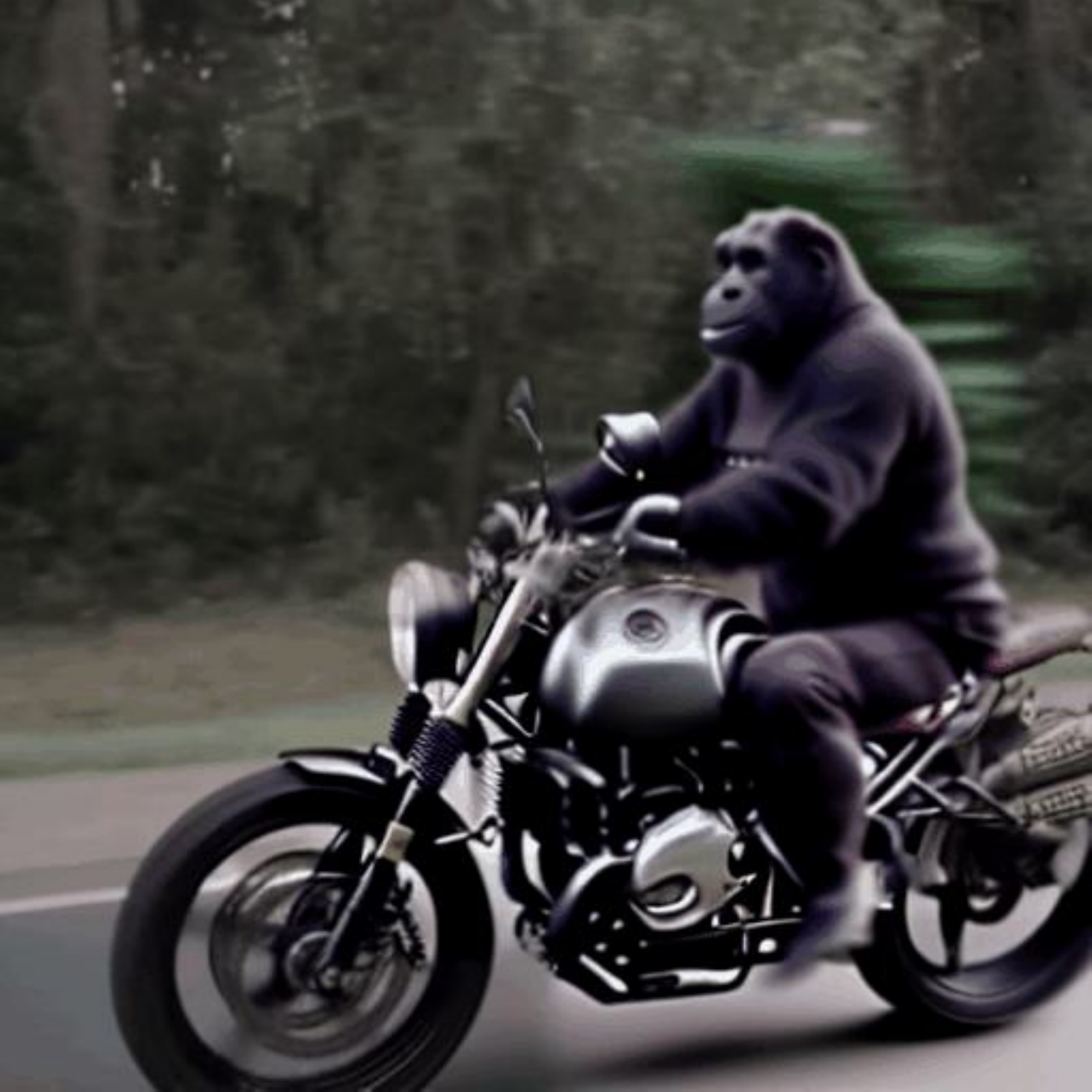}
\includegraphics[width=0.10\textwidth]{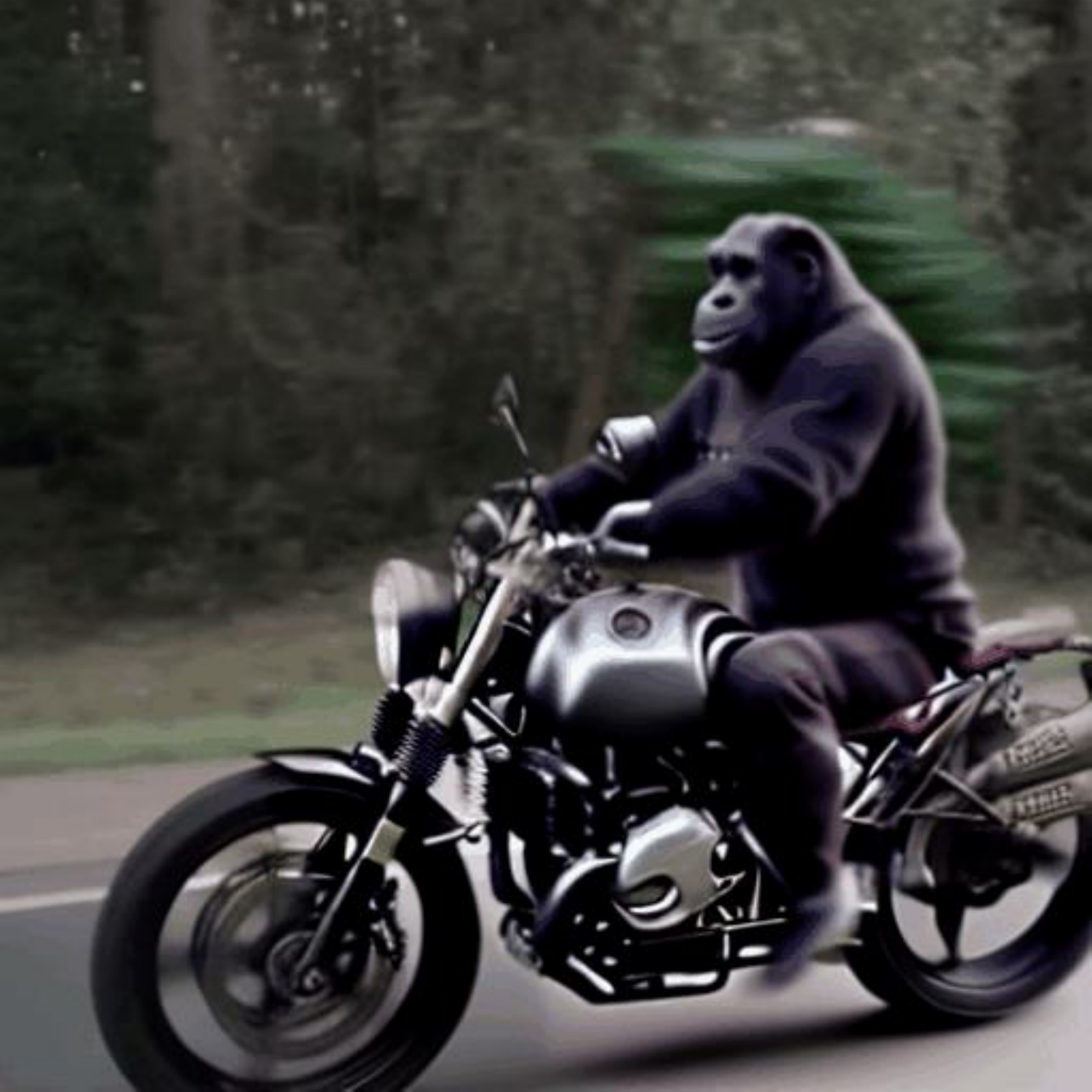}
\includegraphics[width=0.10\textwidth]{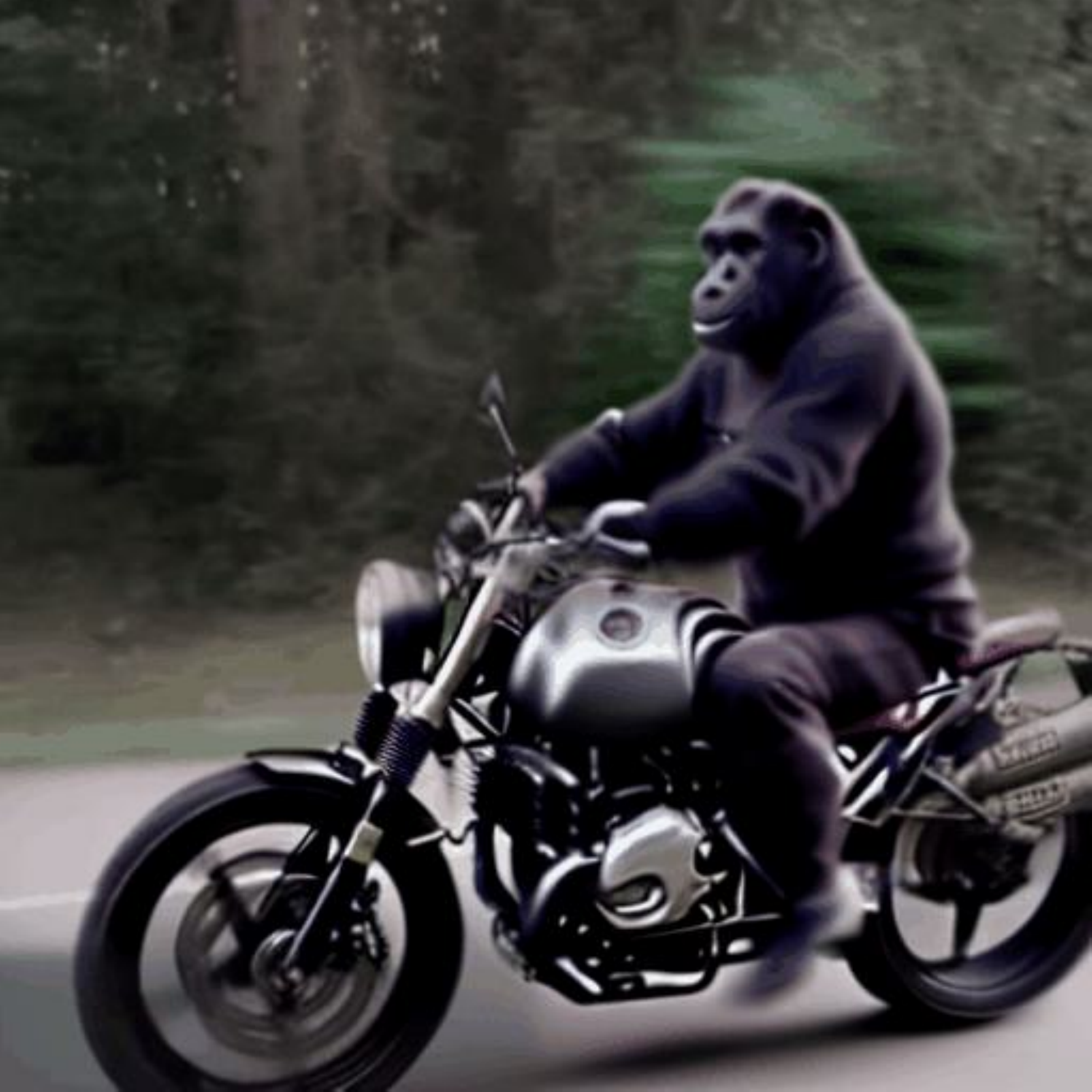}
\includegraphics[width=0.10\textwidth]{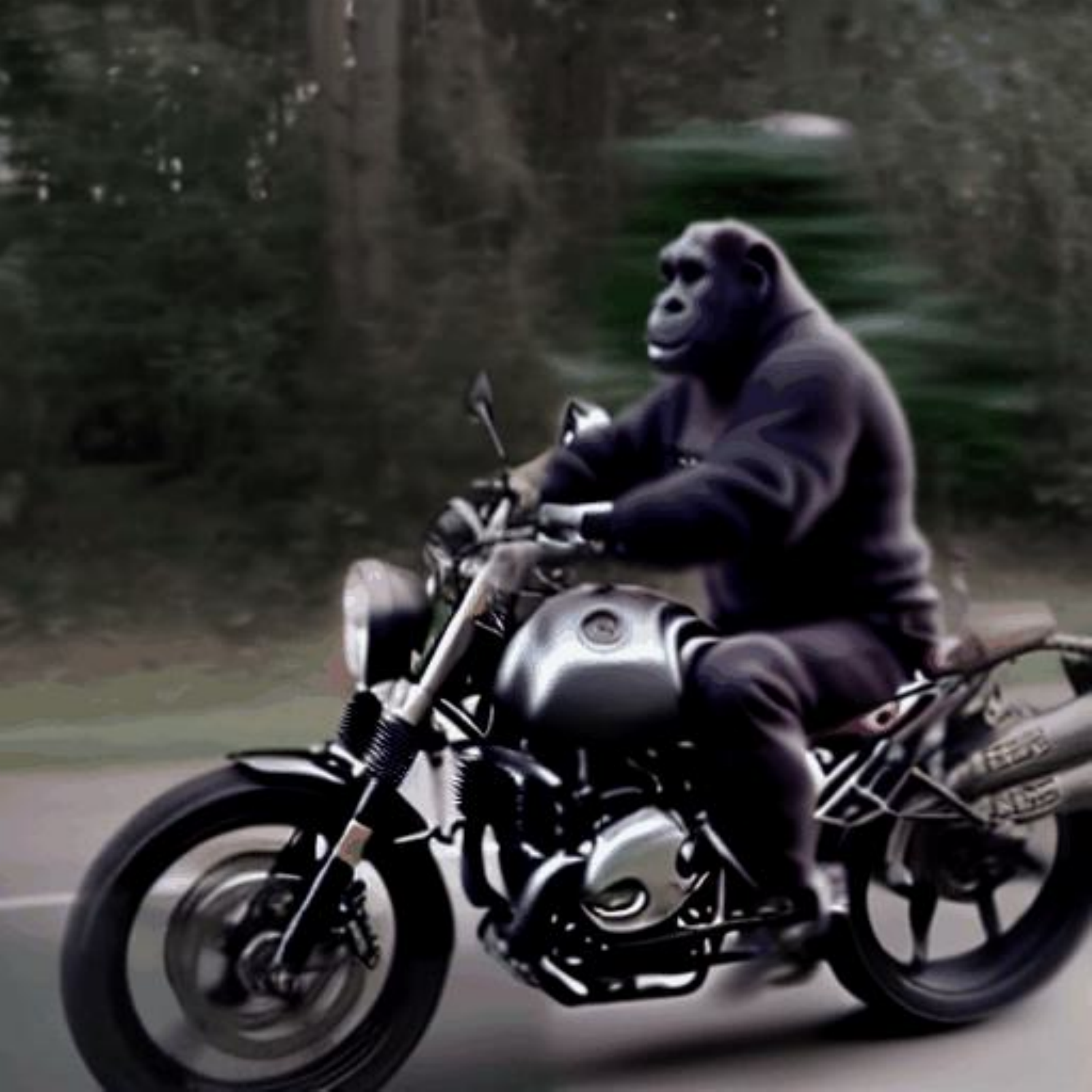}
\includegraphics[width=0.10\textwidth]{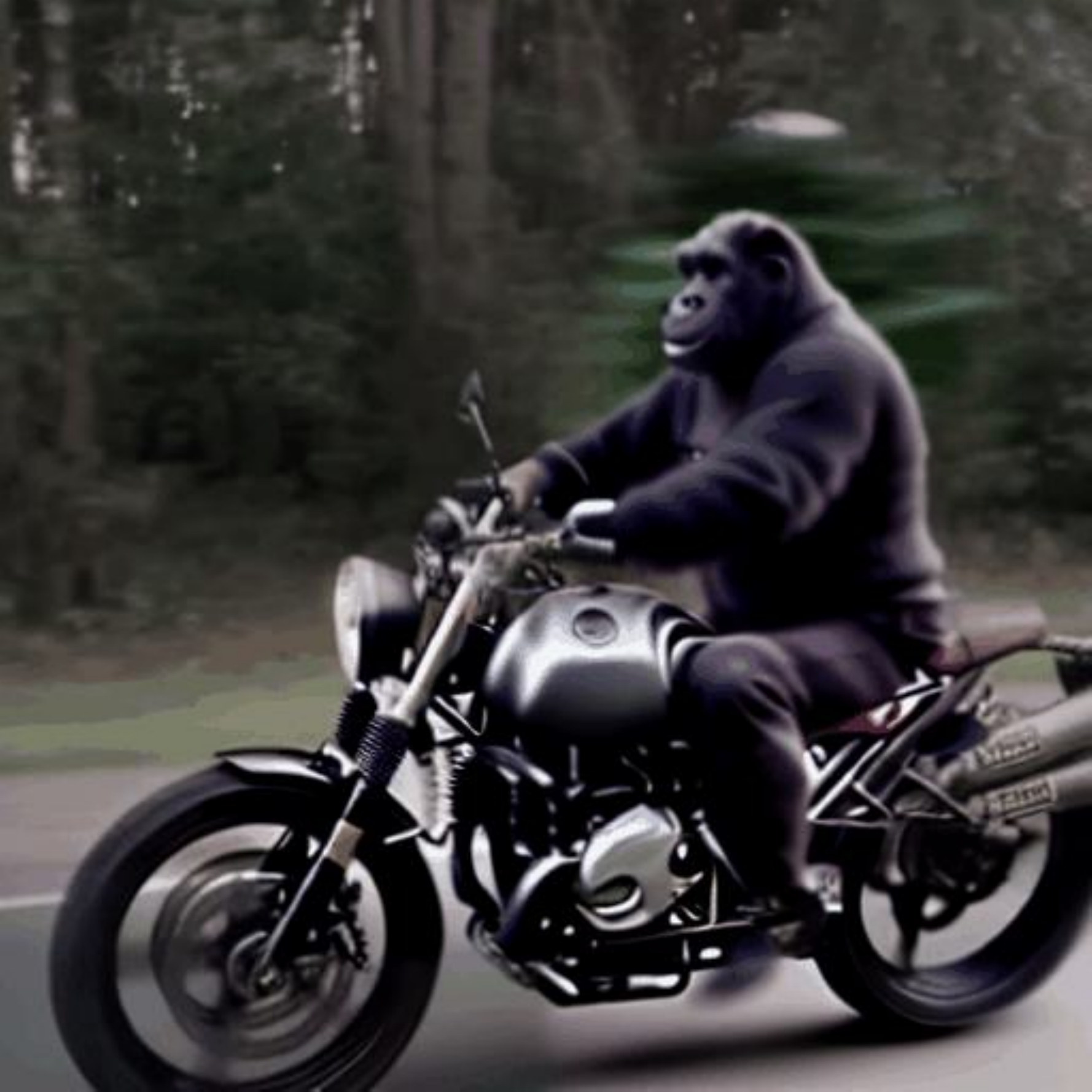}

\caption{\textbf{Baseline Comparison} Additional samples for our model and baselines.}
\label{fig:supp_comparison1}
\end{center}
\end{figure*}

\begin{figure*}[t]
\vspace{-1.8em}
\begin{center}
\makebox[0.12\textwidth]{\colorbox{pink}{\textbf{Training video}} A man is skiing }\\
\rotatebox{90}{\parbox{0.10\textwidth}{\centering ~ \\ ~ }}
\includegraphics[width=0.10\textwidth]{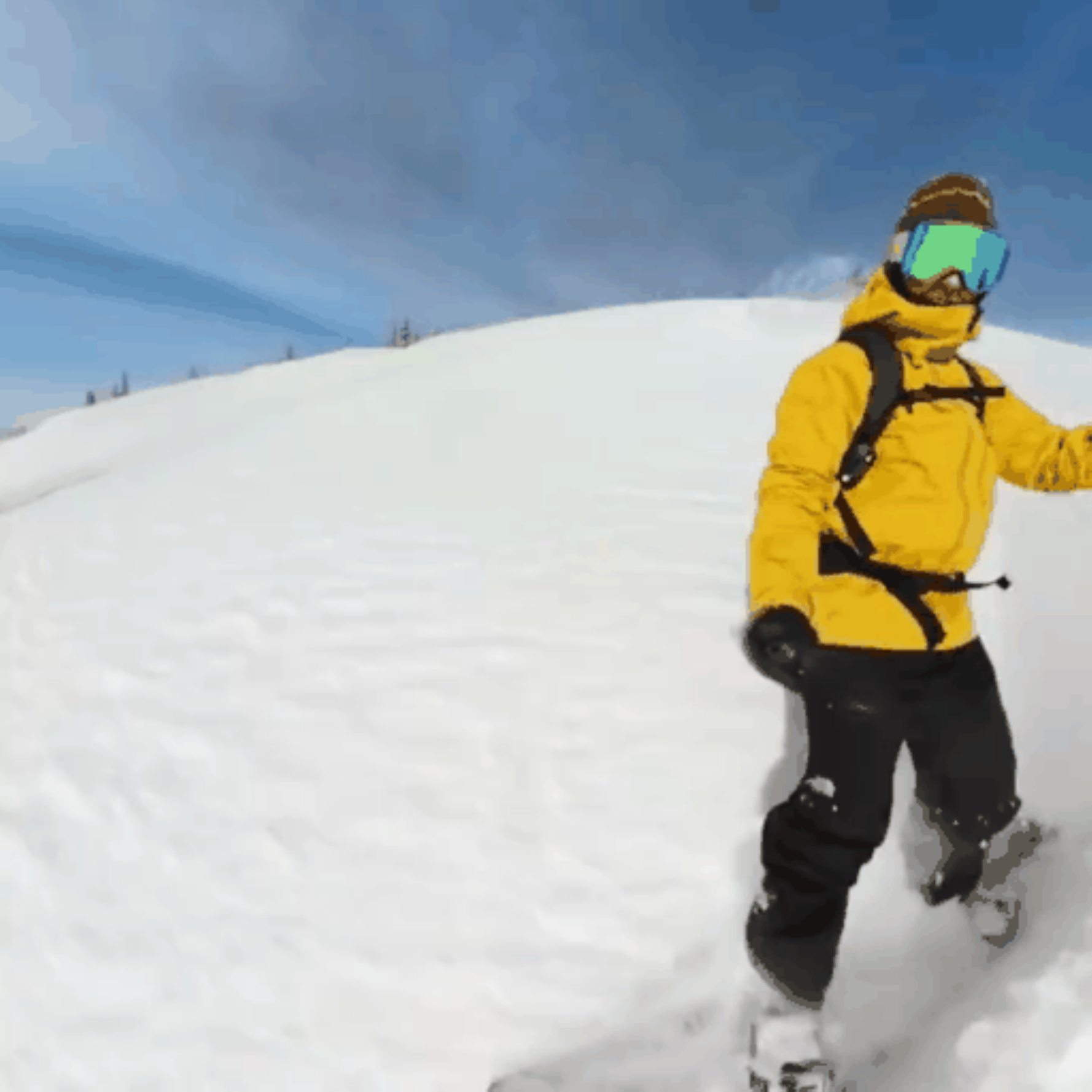}
\includegraphics[width=0.10\textwidth]{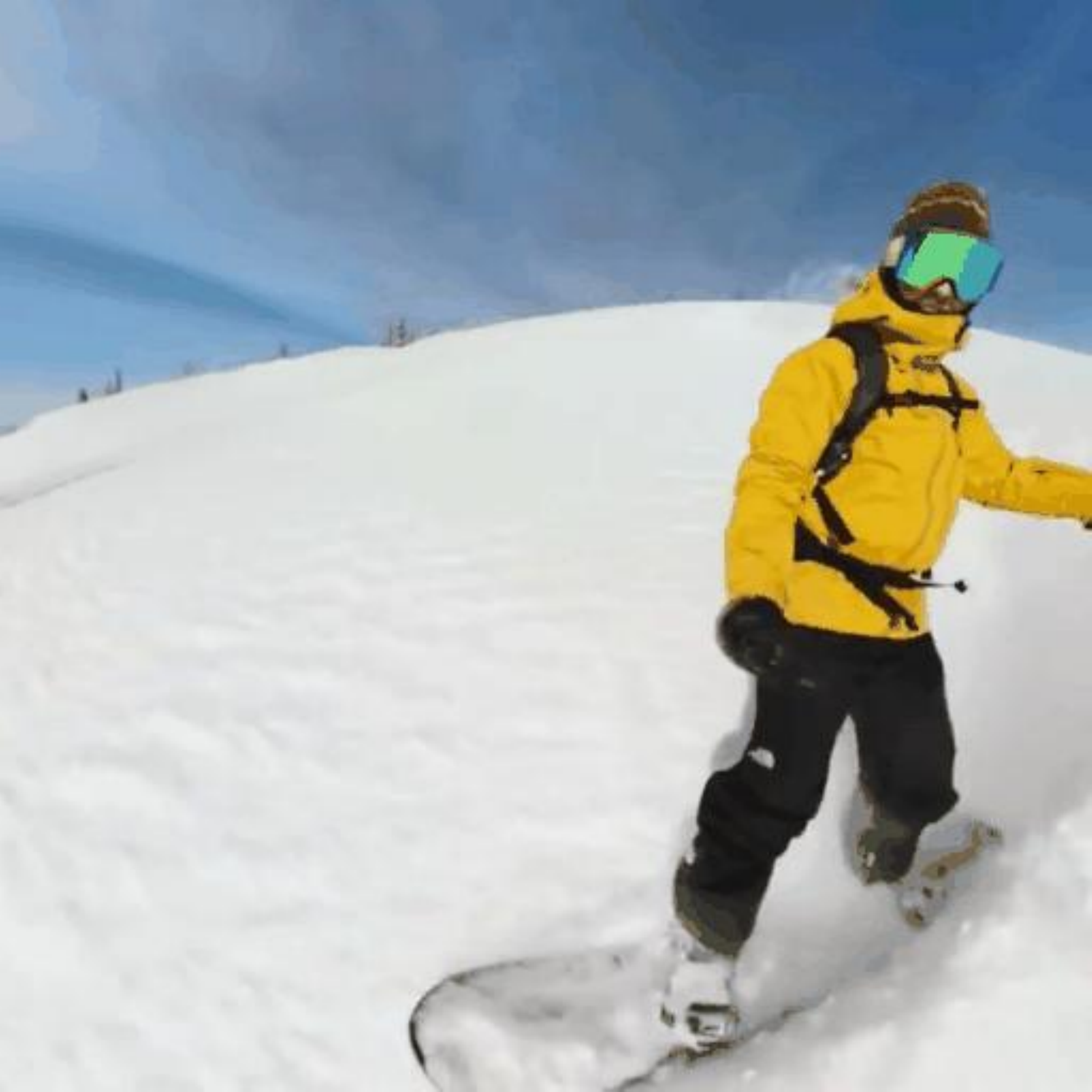}
\includegraphics[width=0.10\textwidth]{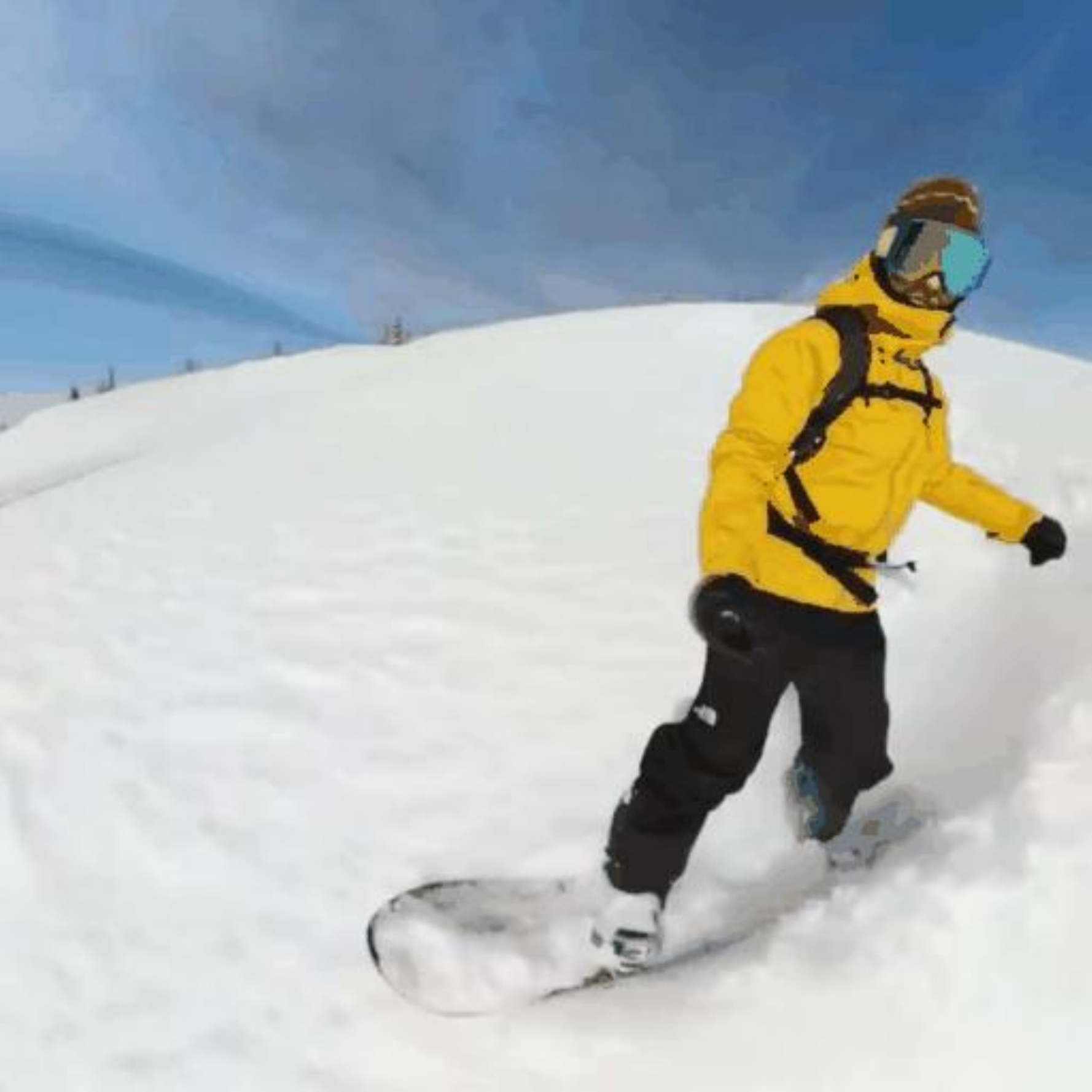}
\includegraphics[width=0.10\textwidth]{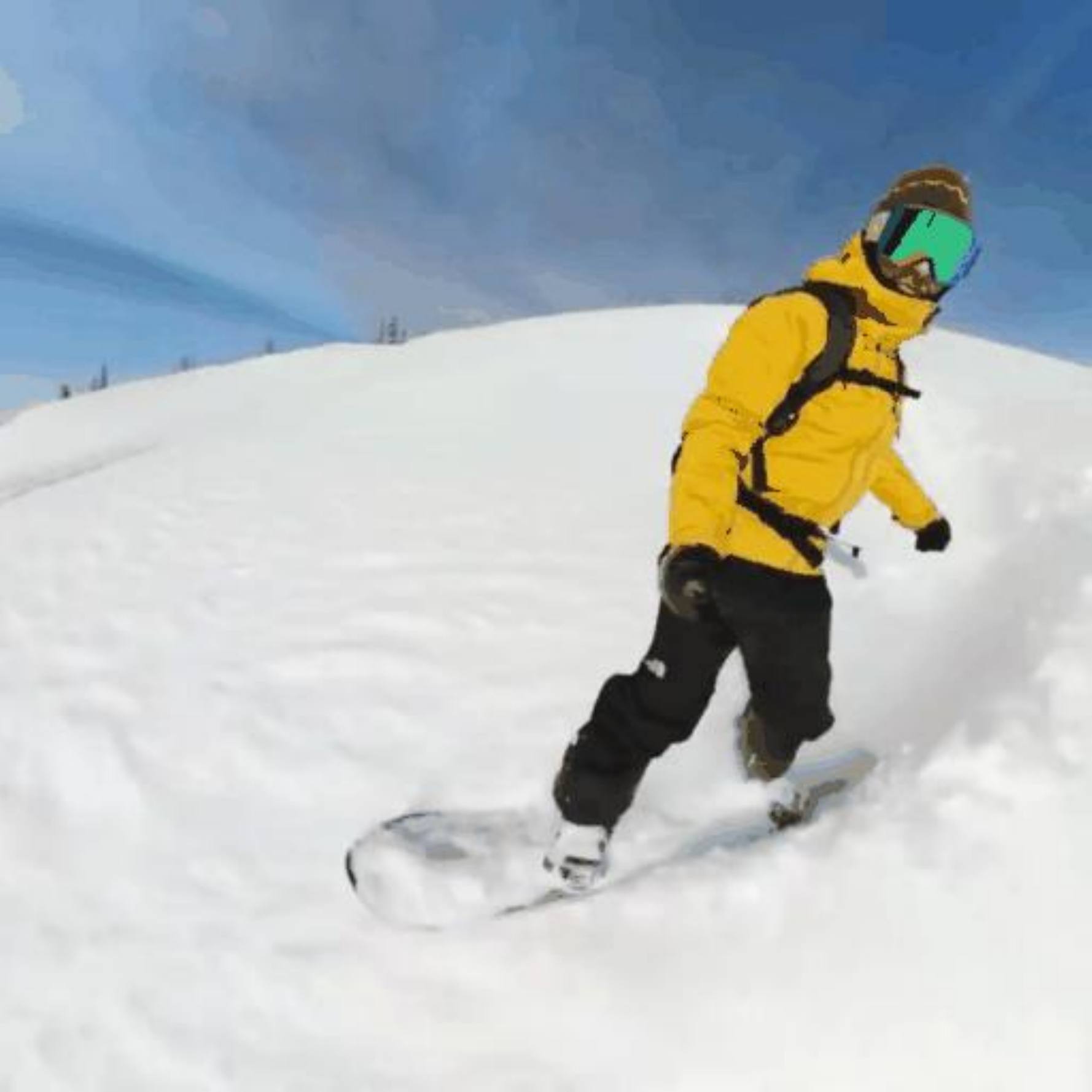}
\includegraphics[width=0.10\textwidth]{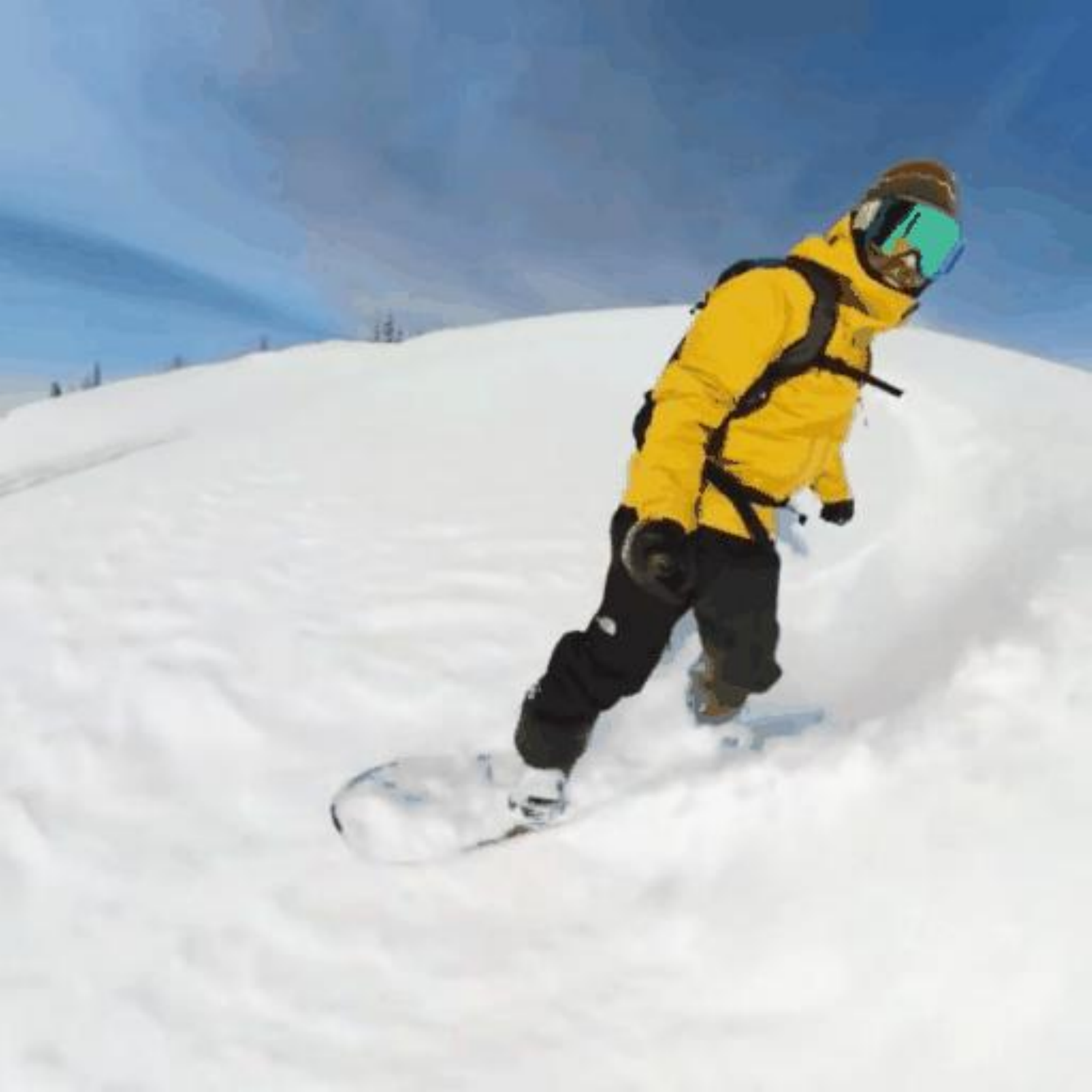}
\includegraphics[width=0.10\textwidth]{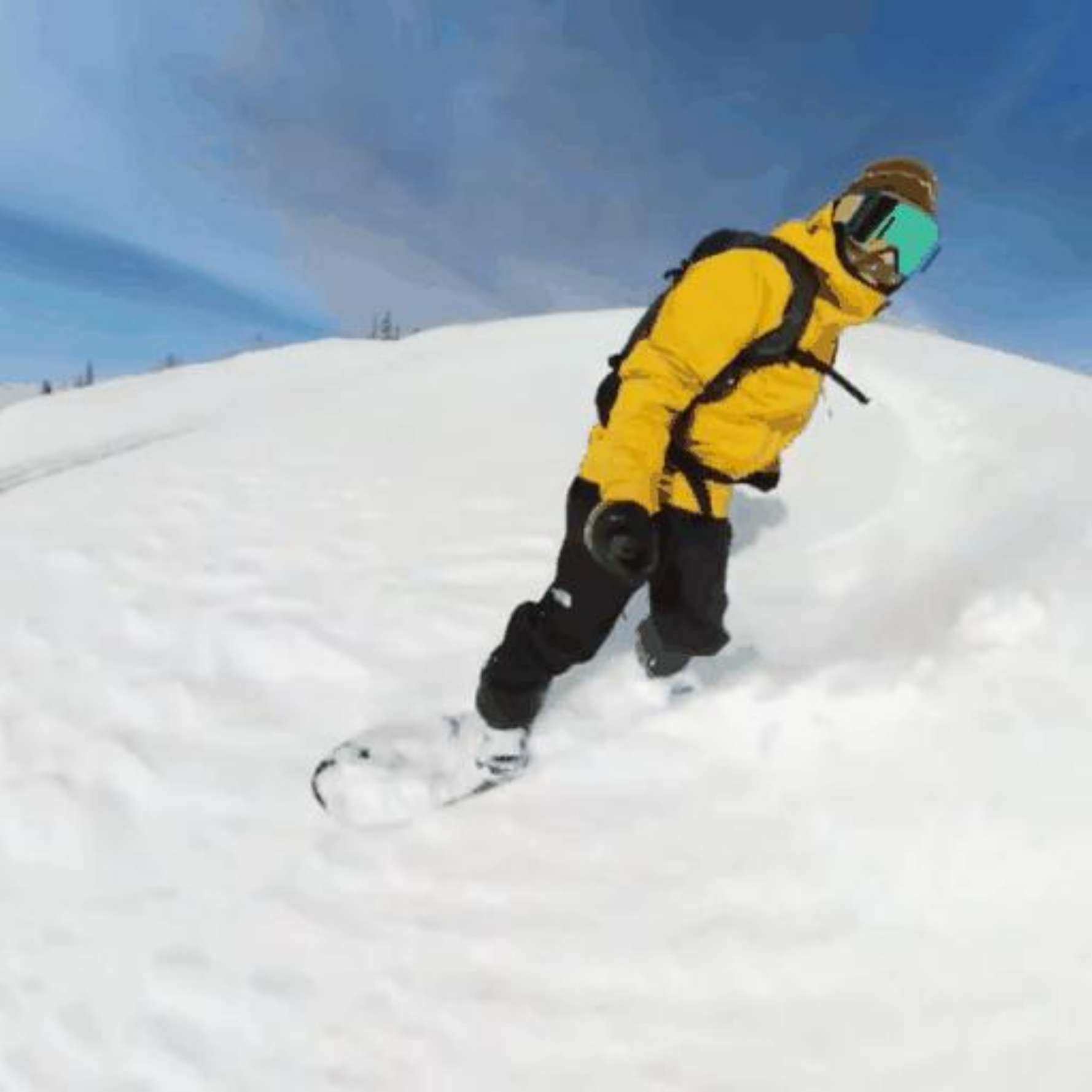}
\includegraphics[width=0.10\textwidth]{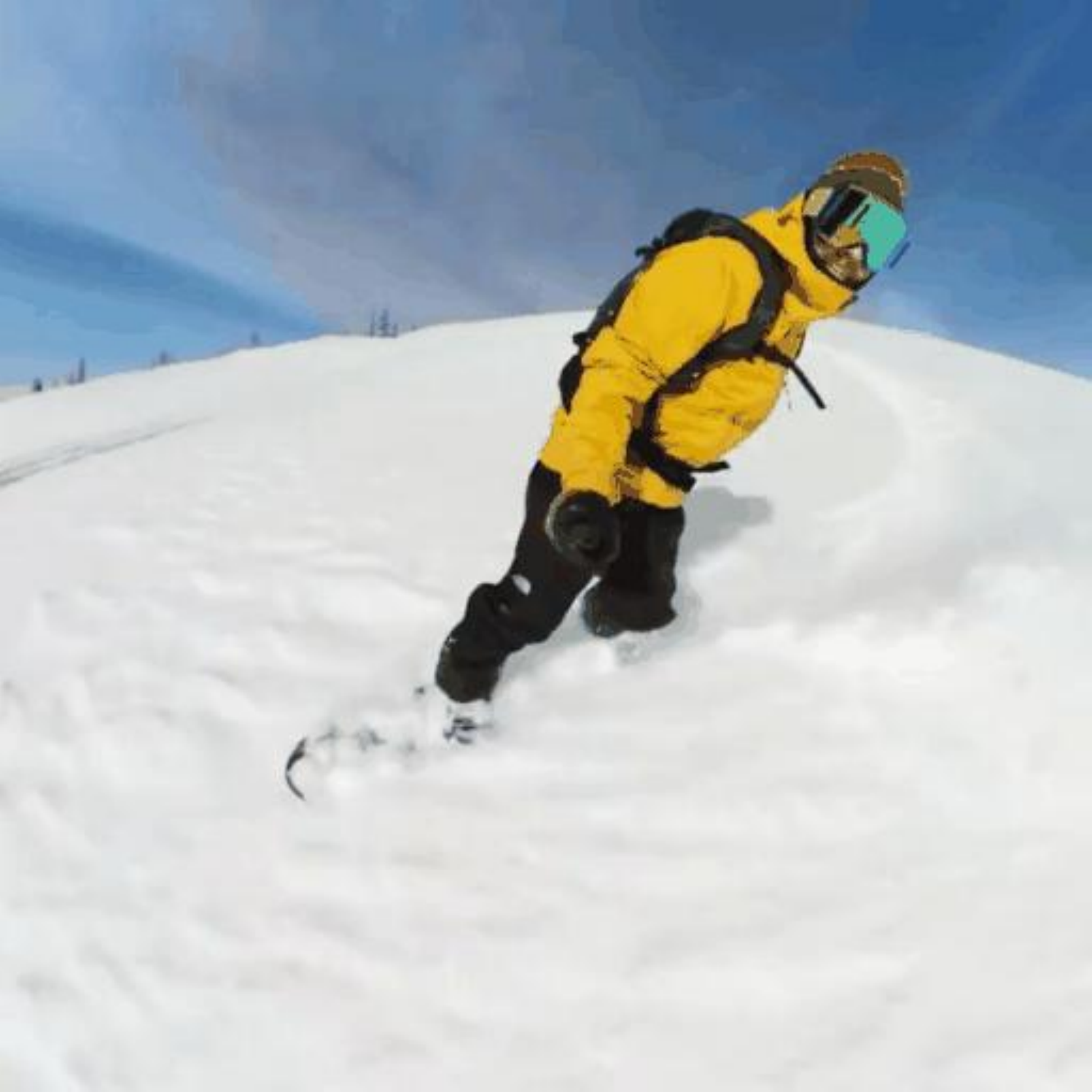}
\includegraphics[width=0.10\textwidth]{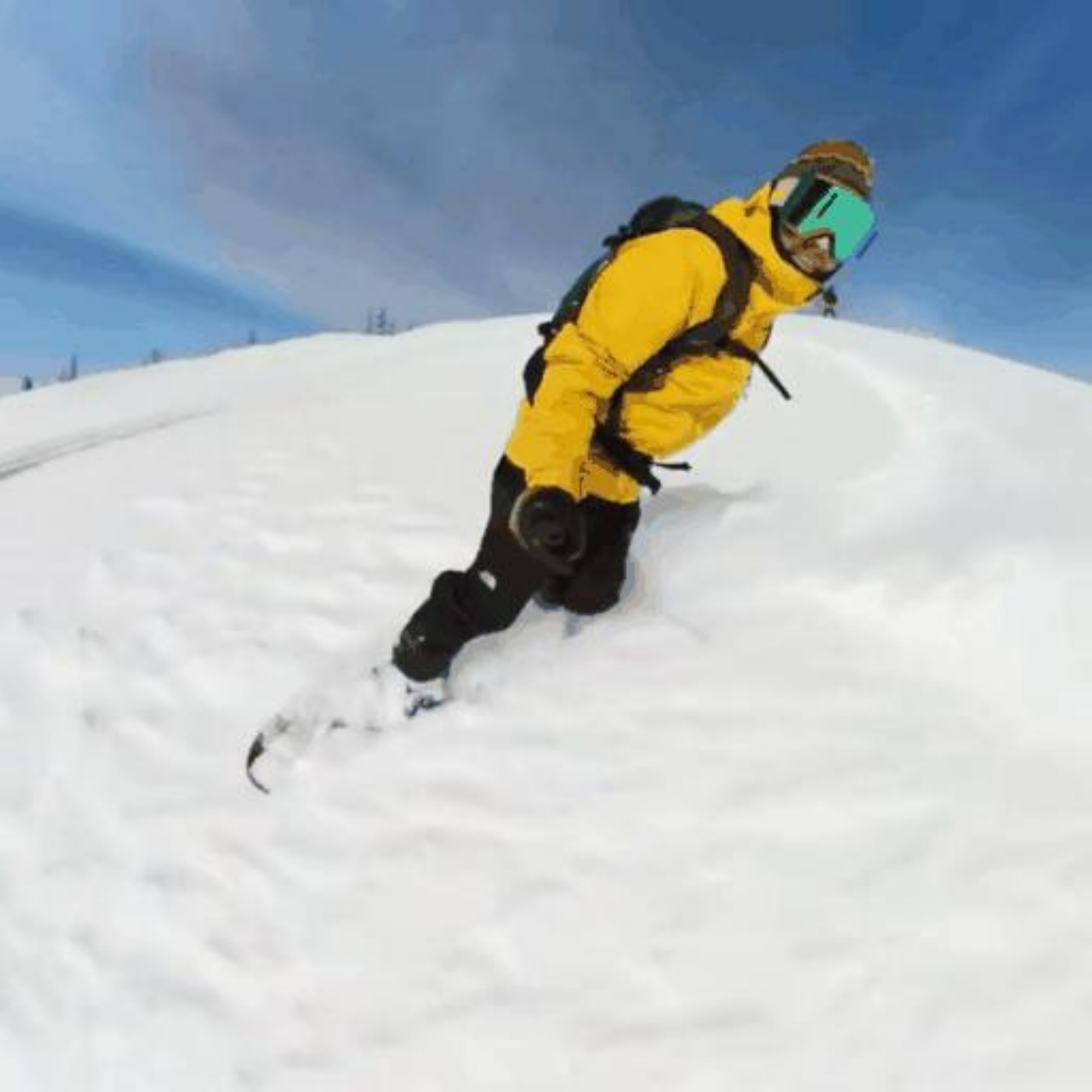}

\makebox[0.12\textwidth]{\colorbox{green}{\textbf{Cross attention}} A \textcolor{blue}{\textbf{Spider Man}} is skiing }\\

\rotatebox{90}{\parbox{0.10\textwidth}{\centering \textbf{duration \\ 0.2}}}
\includegraphics[width=0.10\textwidth]{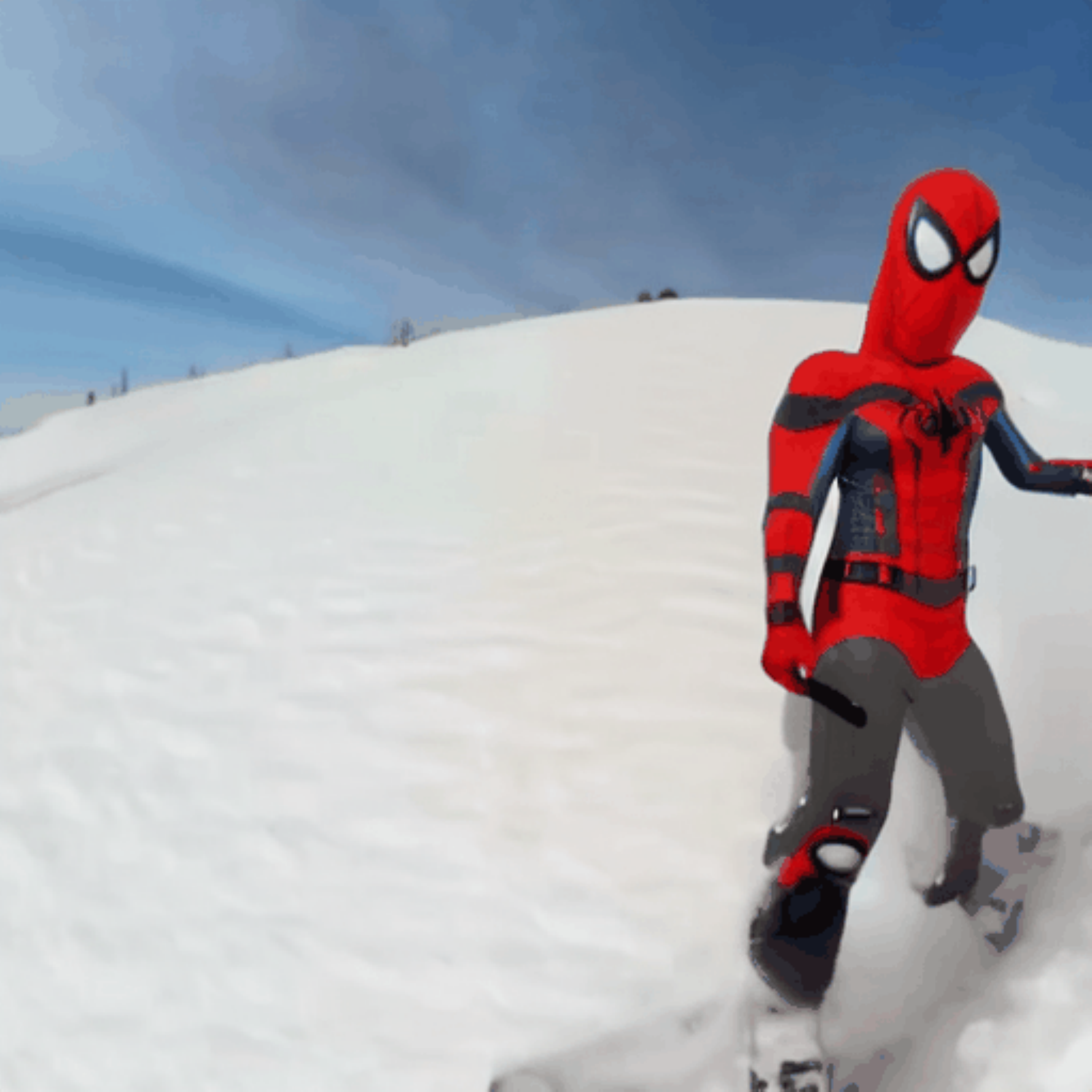}
\includegraphics[width=0.10\textwidth]{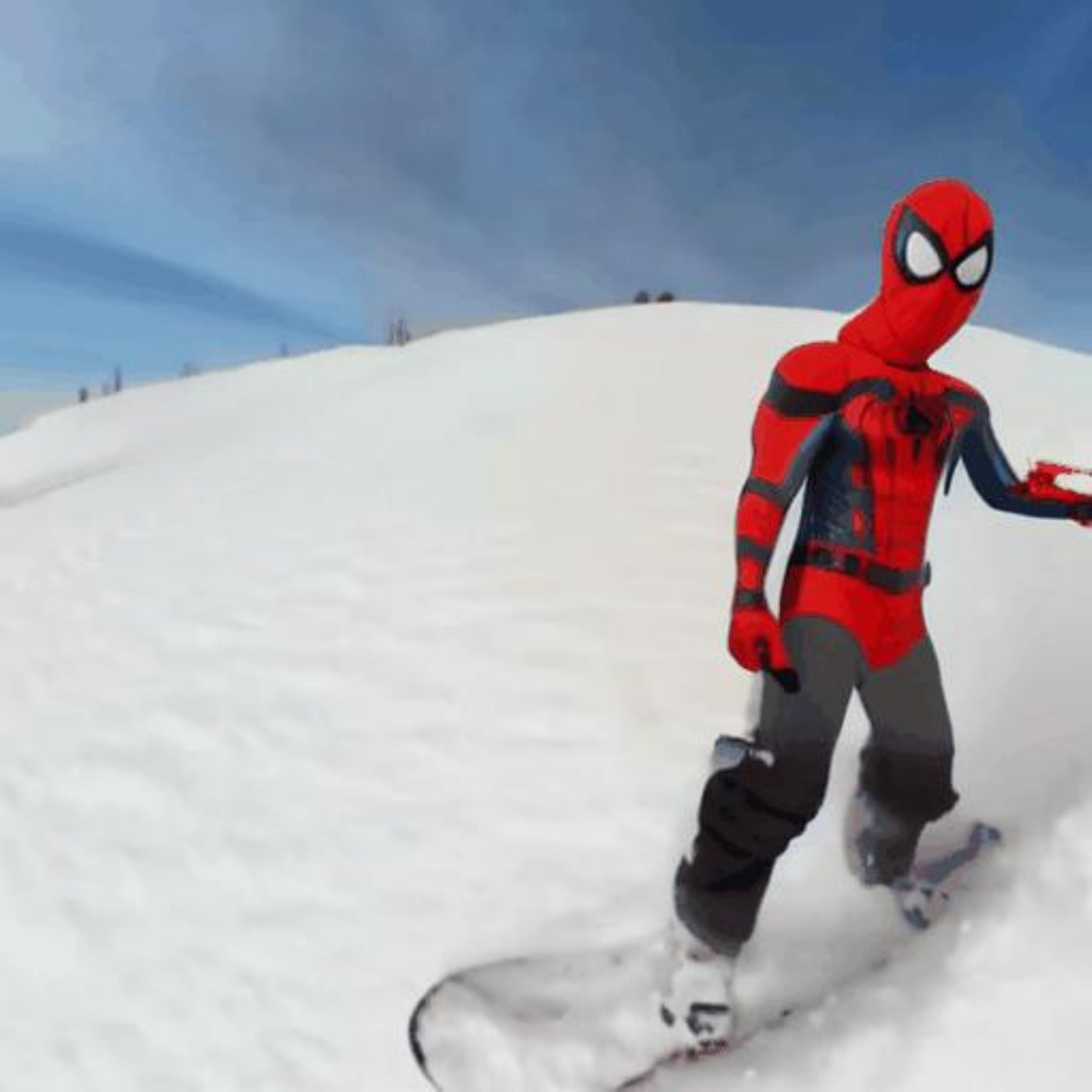}
\includegraphics[width=0.10\textwidth]{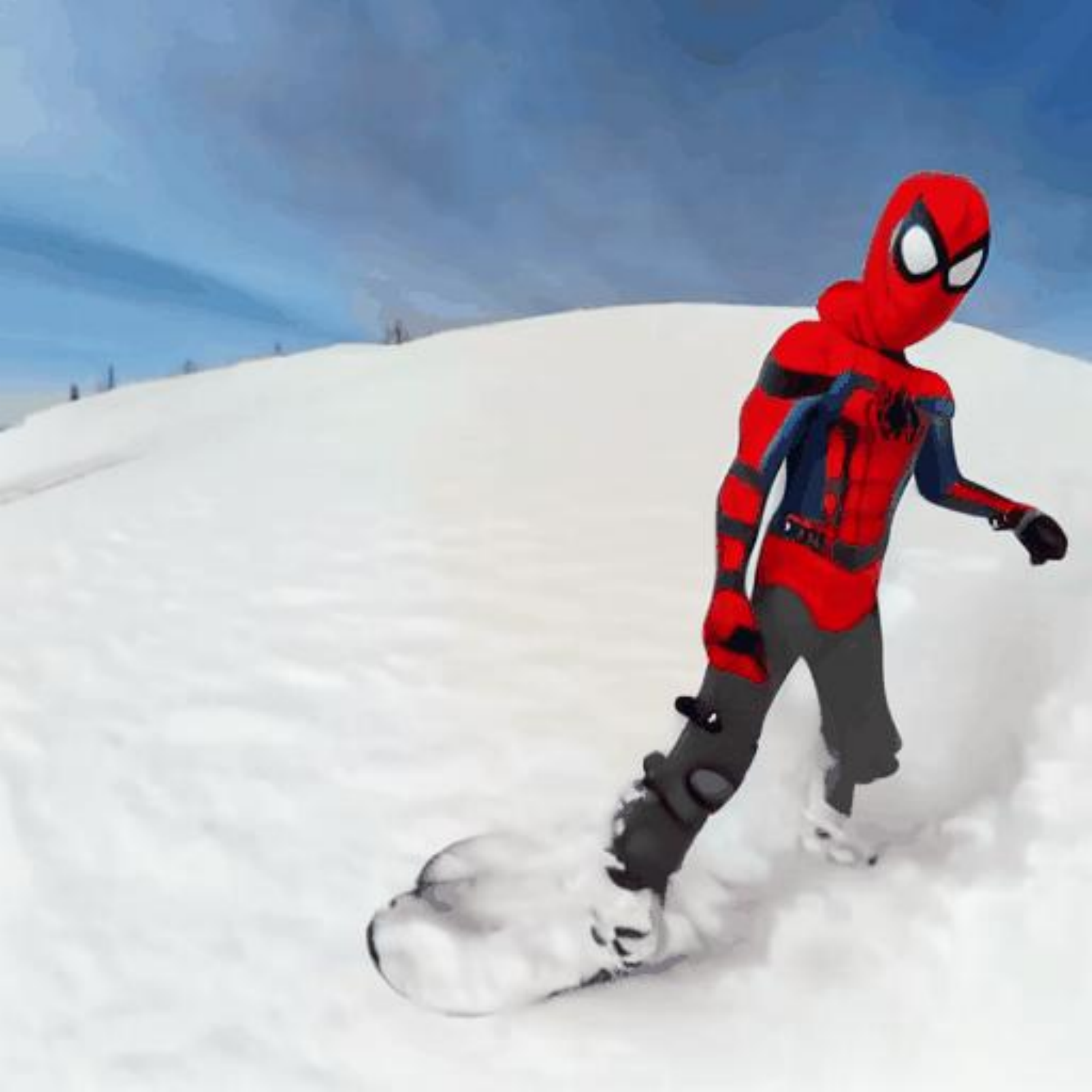}
\includegraphics[width=0.10\textwidth]{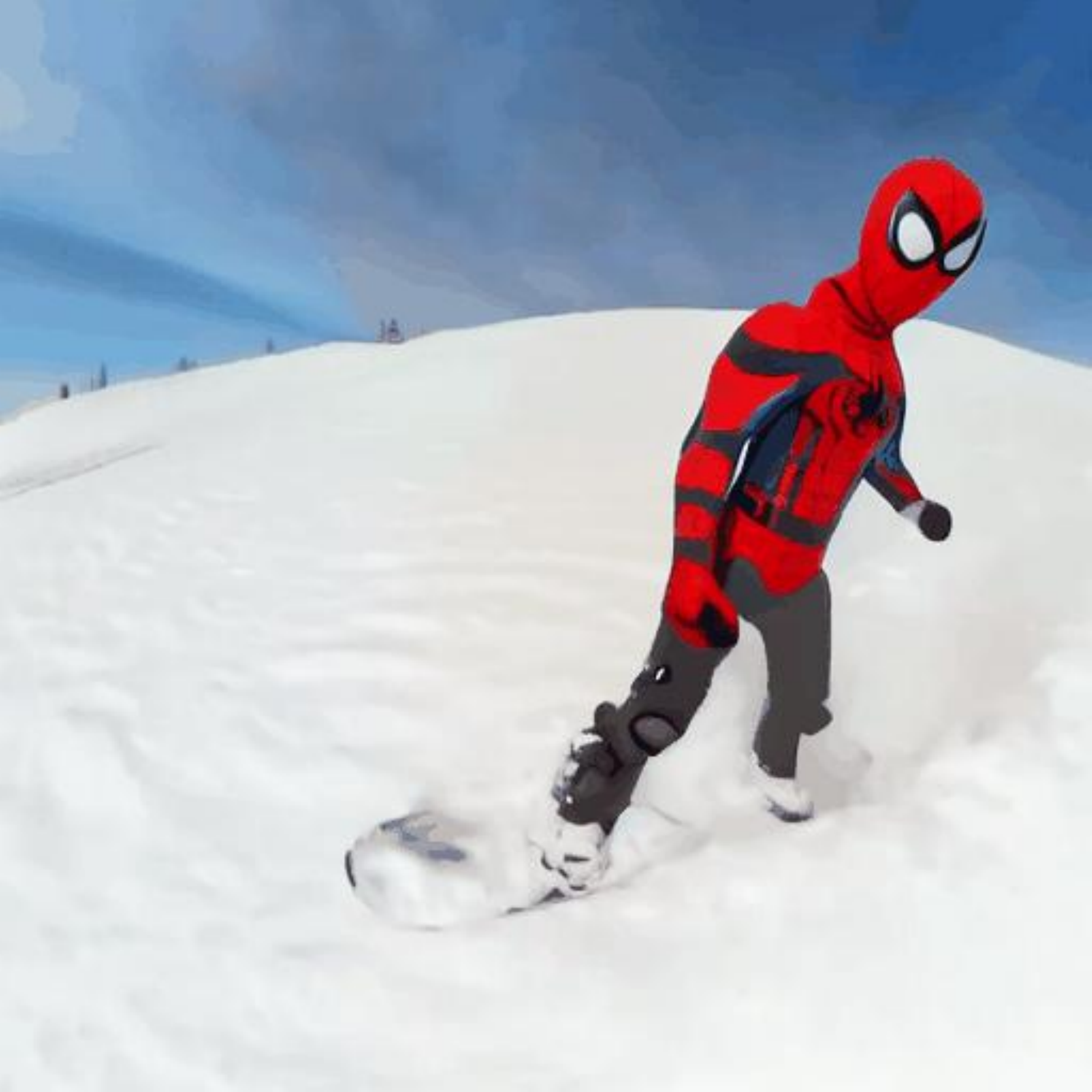}
\includegraphics[width=0.10\textwidth]{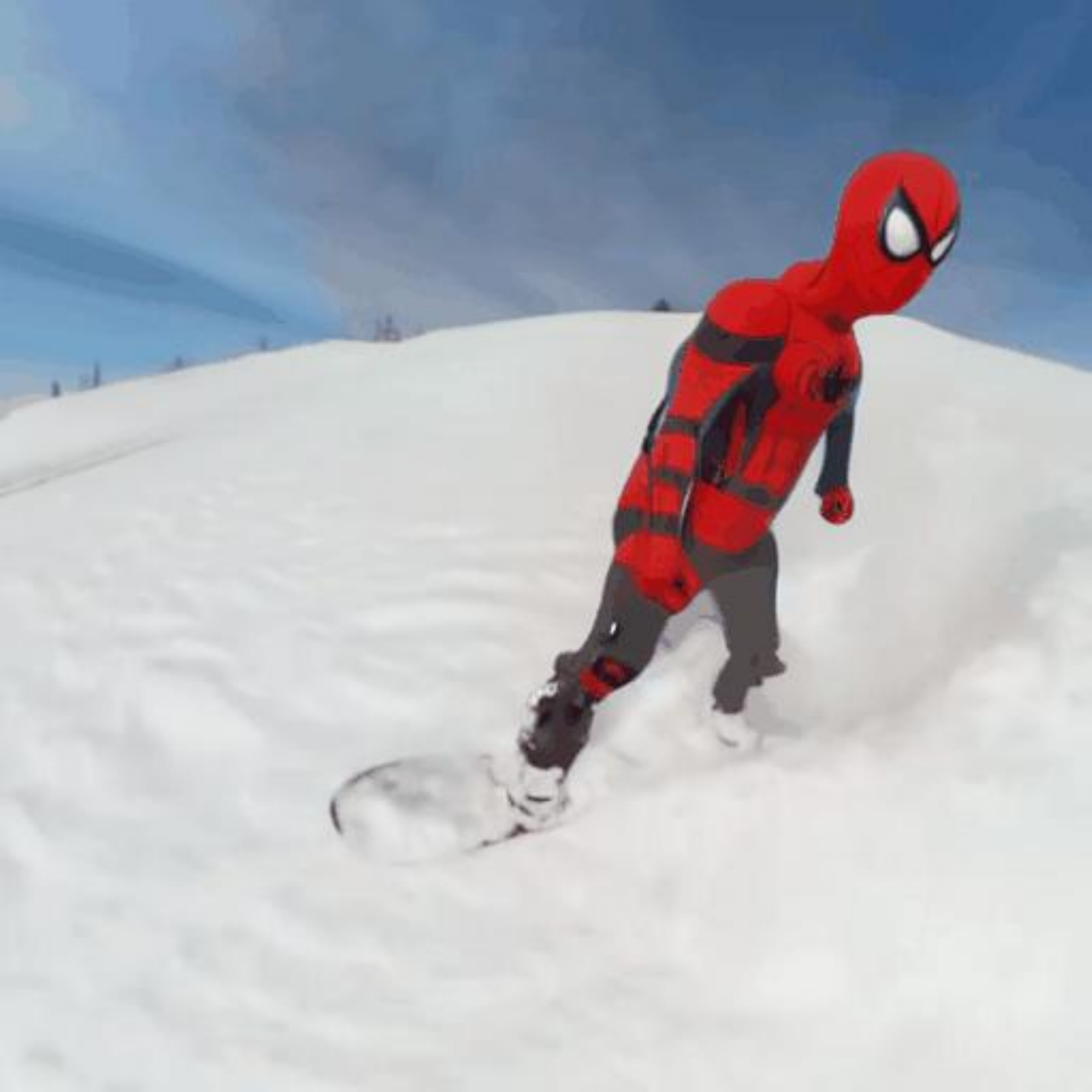}
\includegraphics[width=0.10\textwidth]{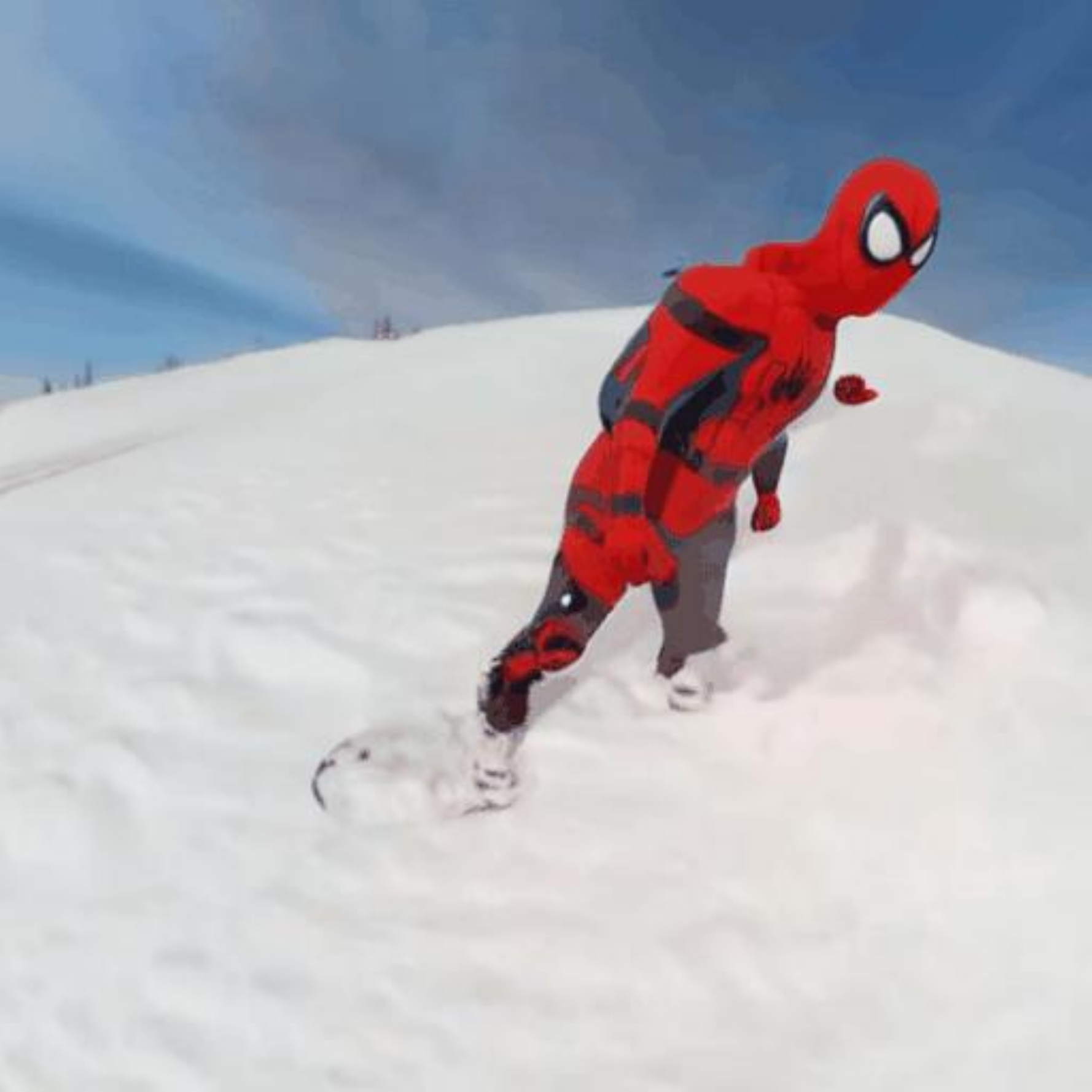}
\includegraphics[width=0.10\textwidth]{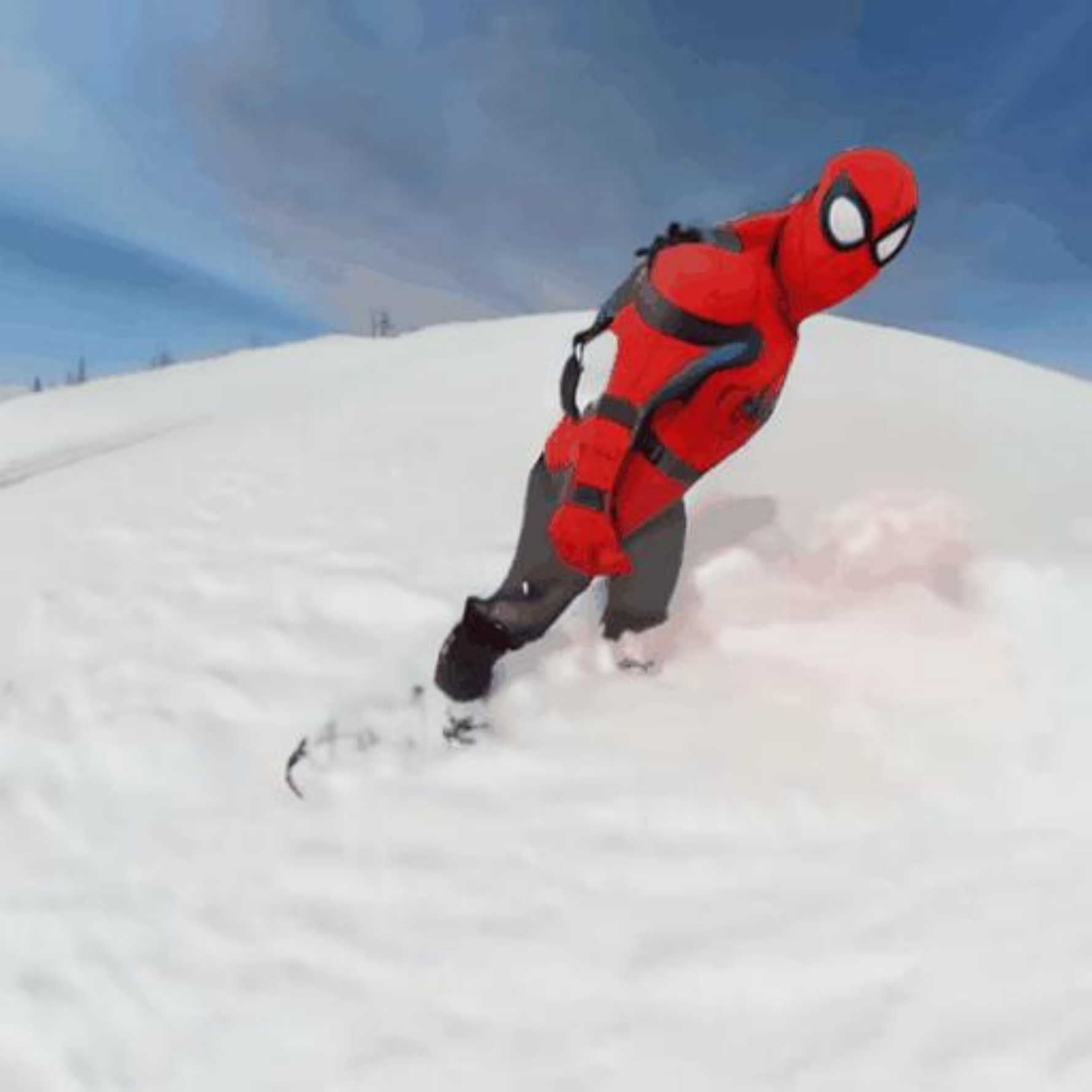}
\includegraphics[width=0.10\textwidth]{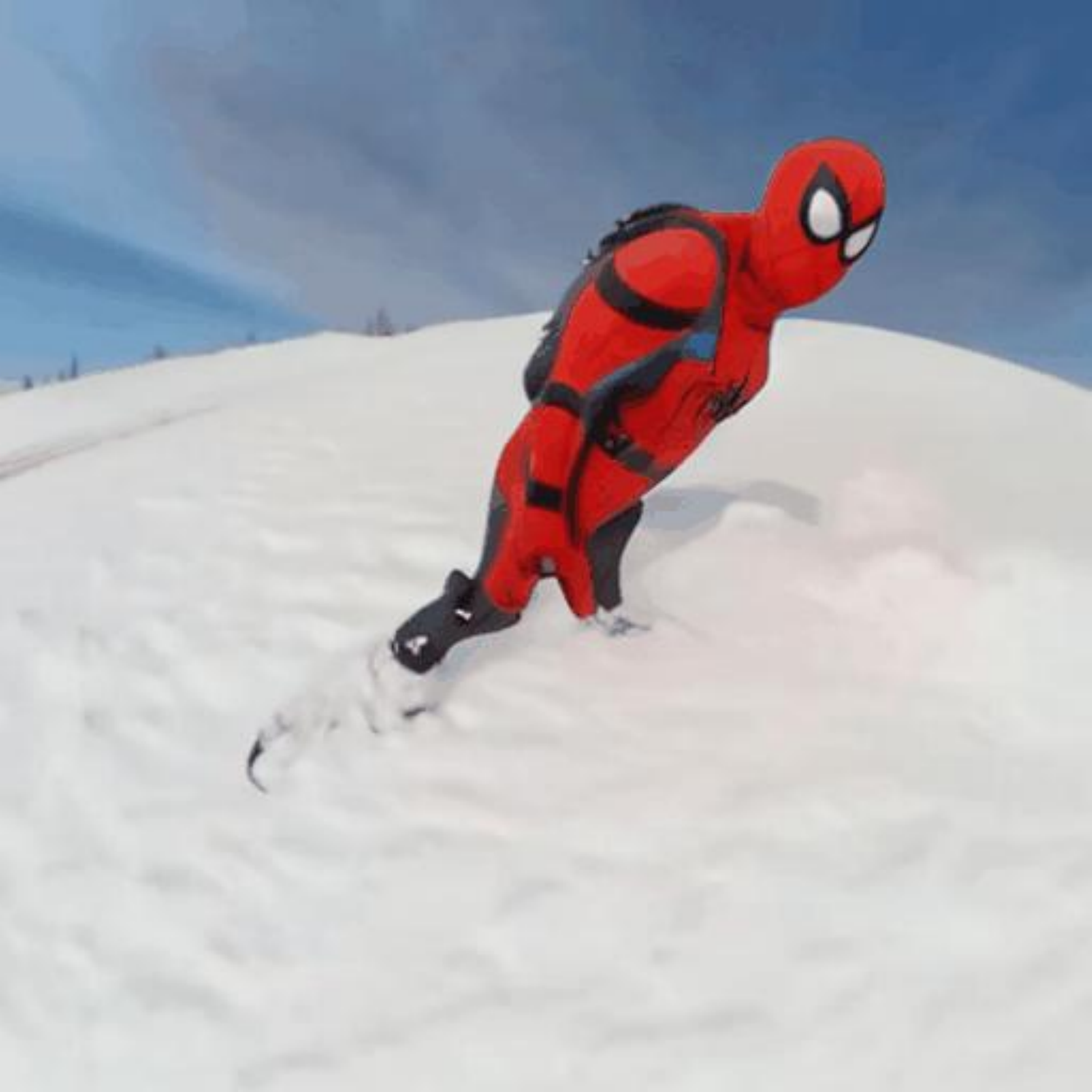}

\rotatebox{90}{\parbox{0.10\textwidth}{\centering duration \\ 0.5}}
\includegraphics[width=0.10\textwidth]{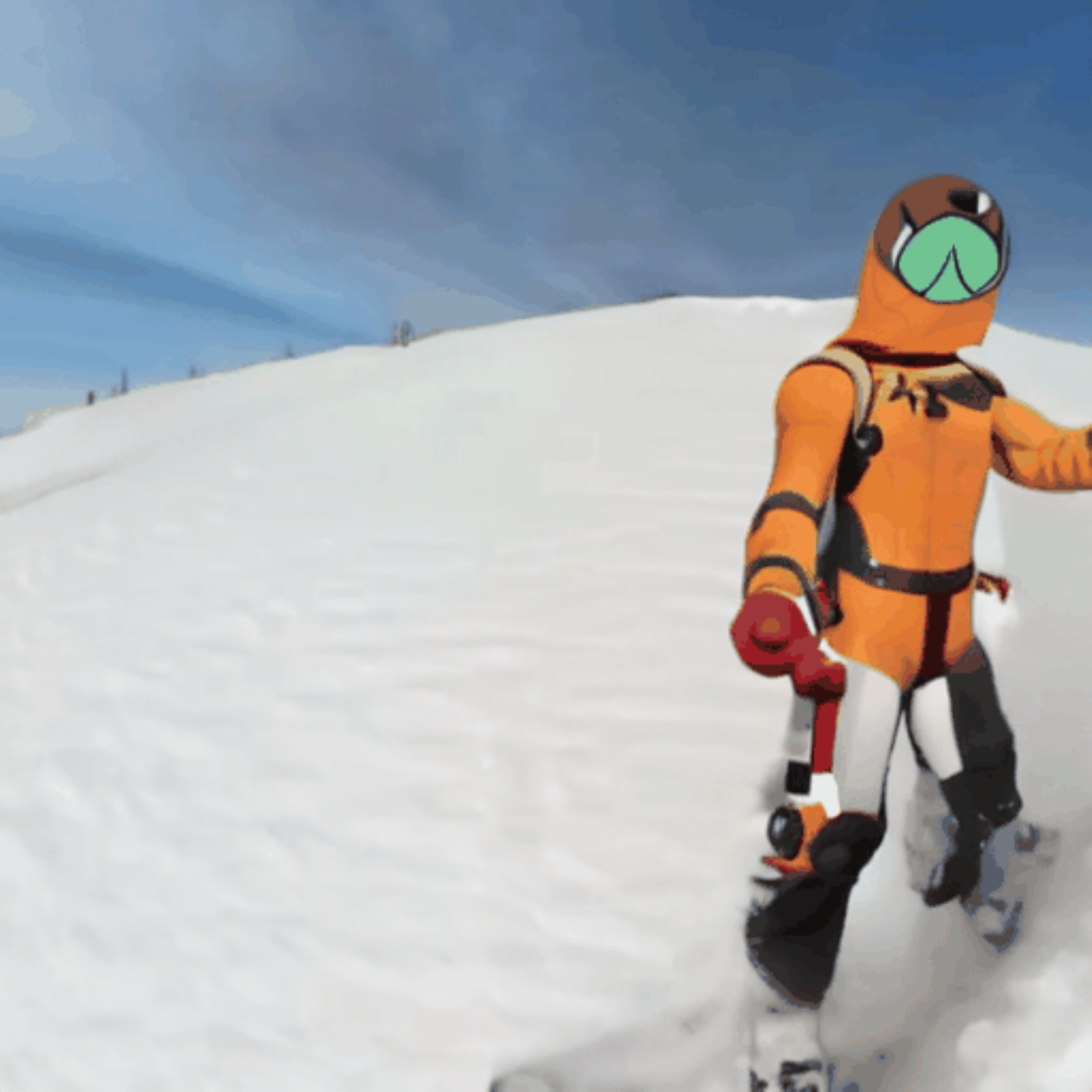}
\includegraphics[width=0.10\textwidth]{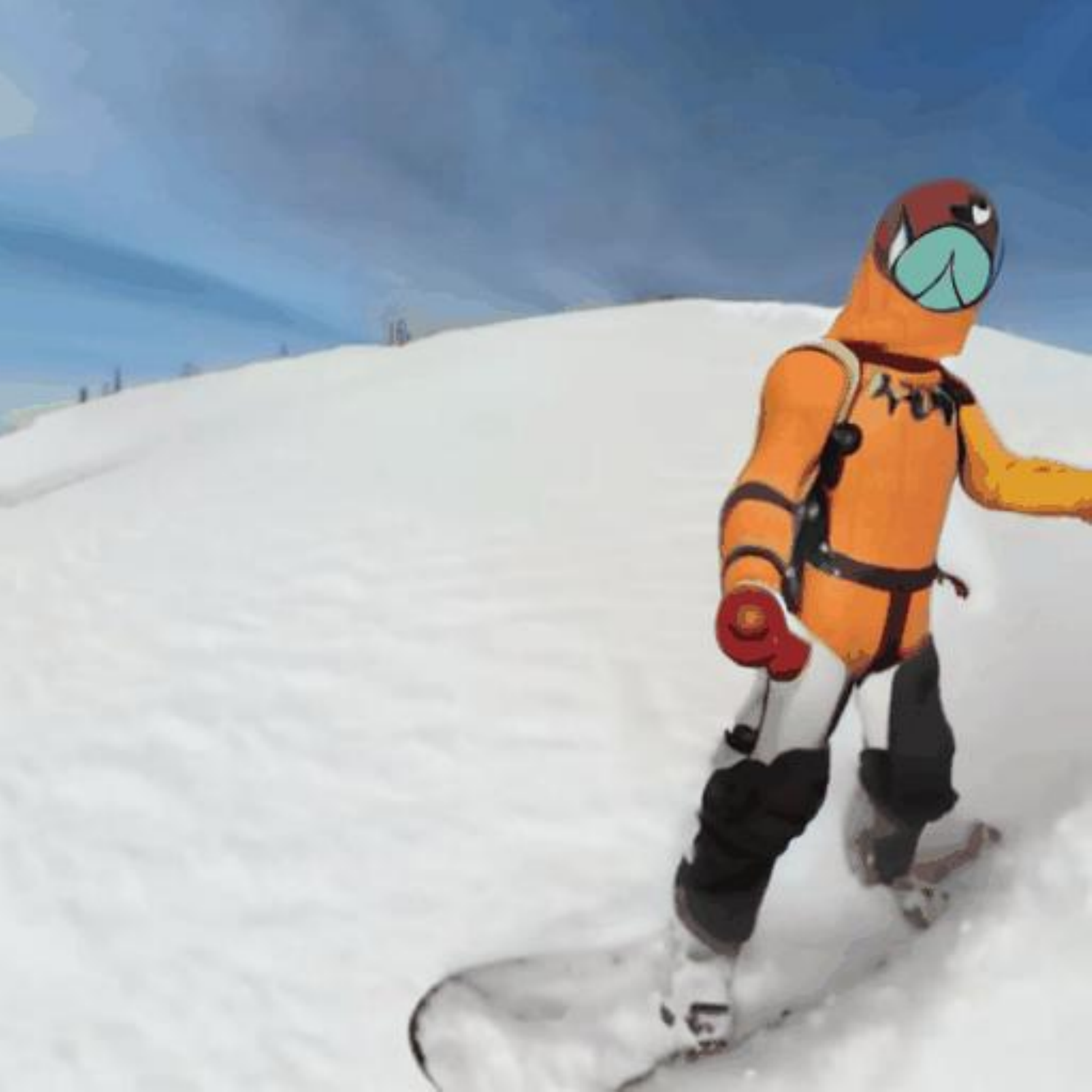}
\includegraphics[width=0.10\textwidth]{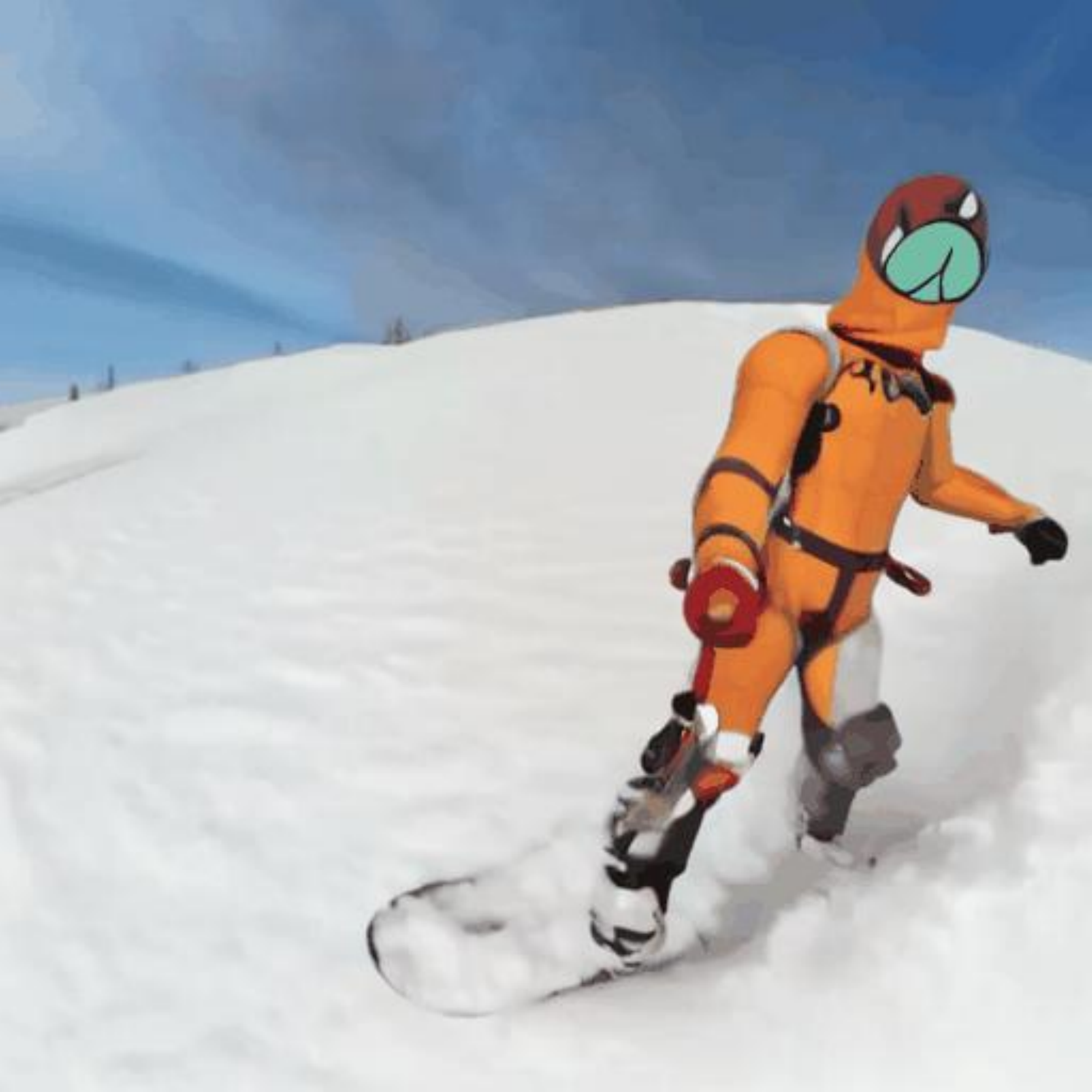}
\includegraphics[width=0.10\textwidth]{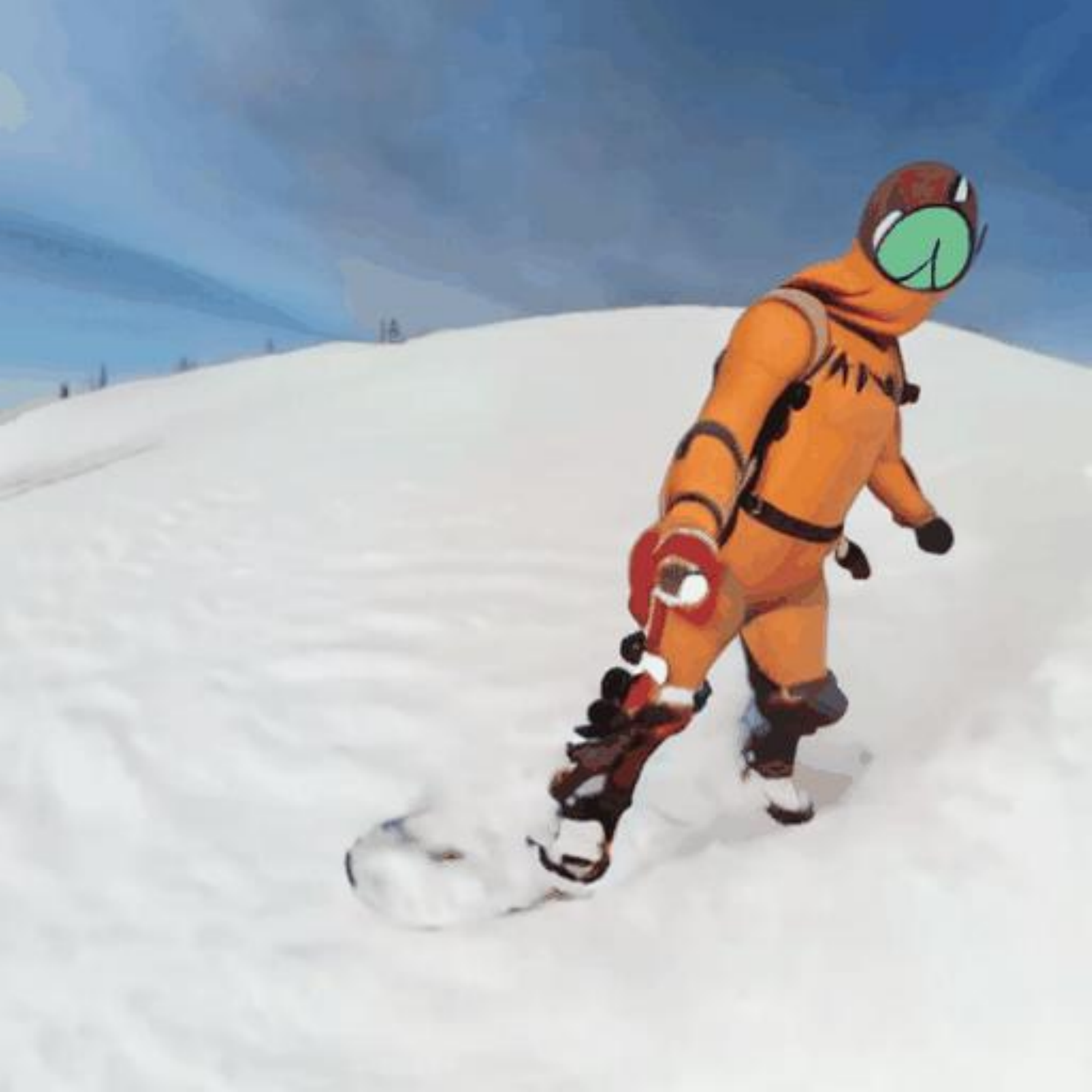}
\includegraphics[width=0.10\textwidth]{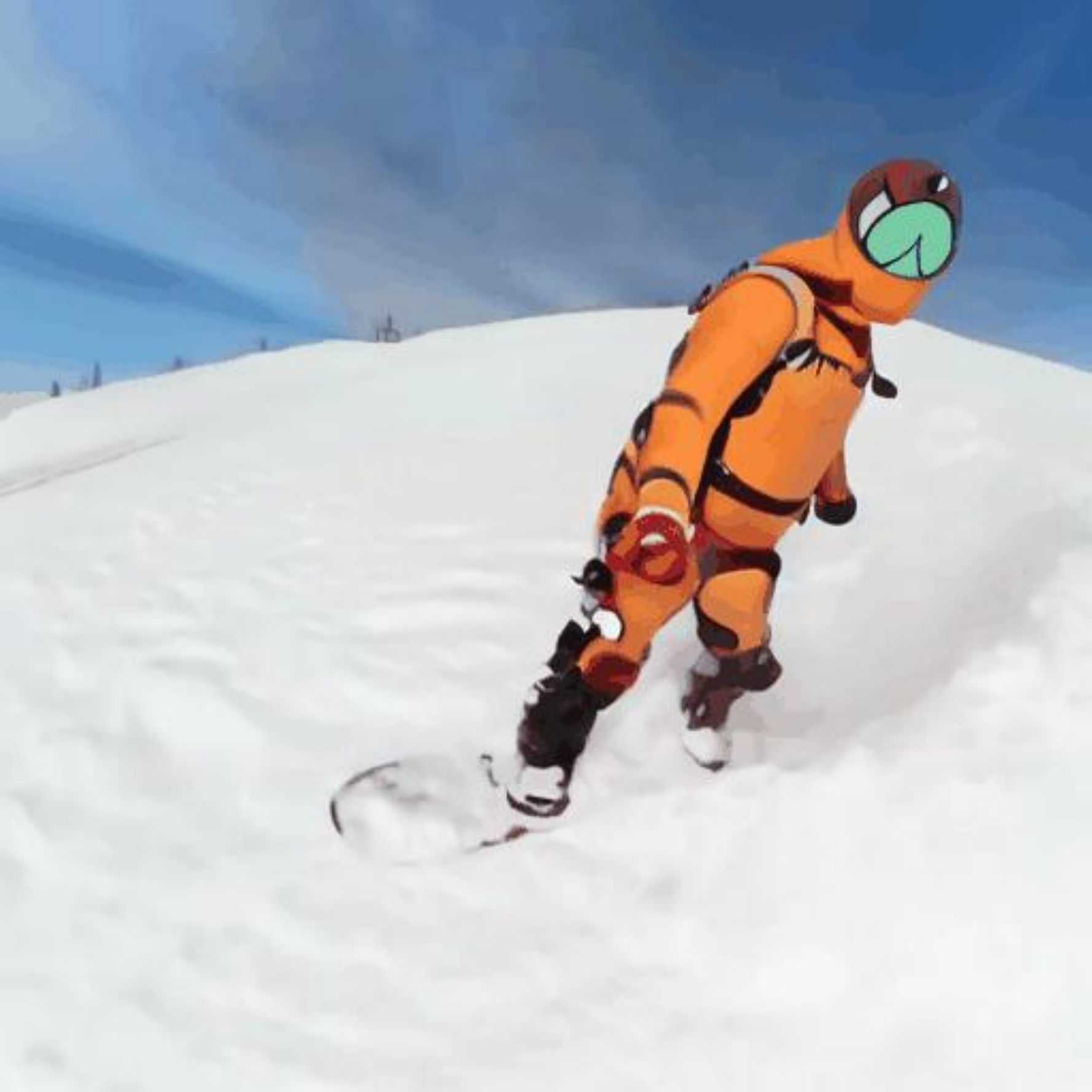}
\includegraphics[width=0.10\textwidth]{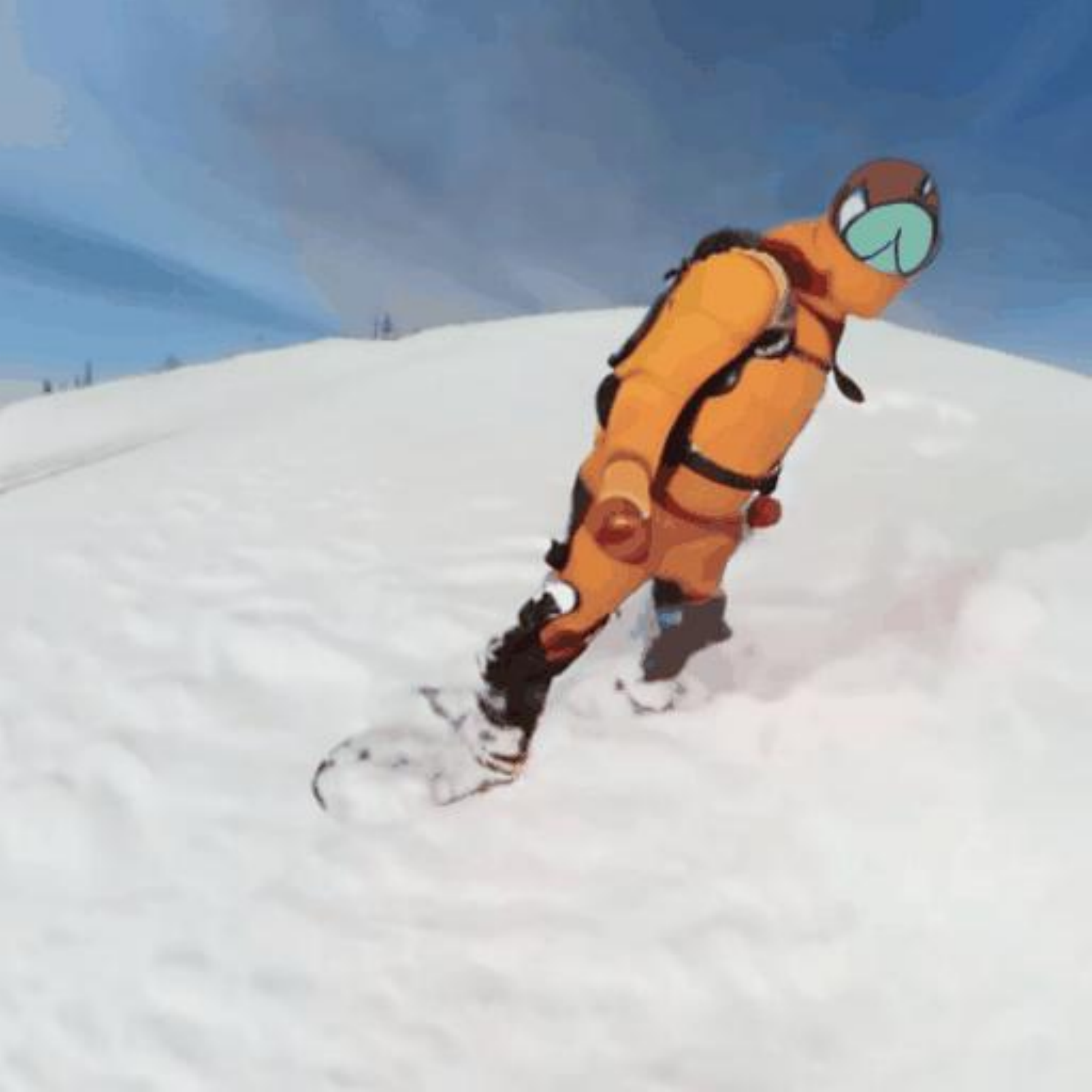}
\includegraphics[width=0.10\textwidth]{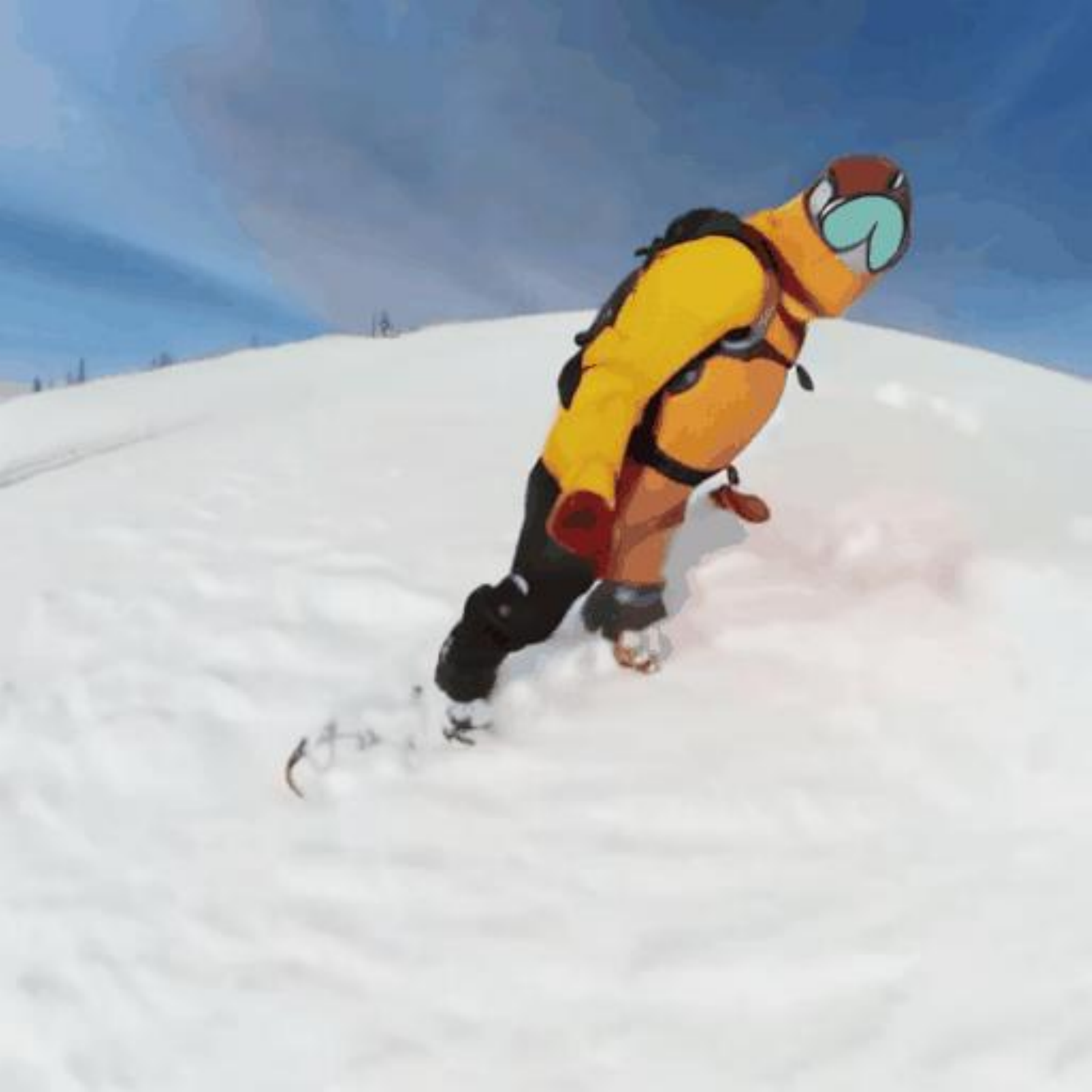}
\includegraphics[width=0.10\textwidth]{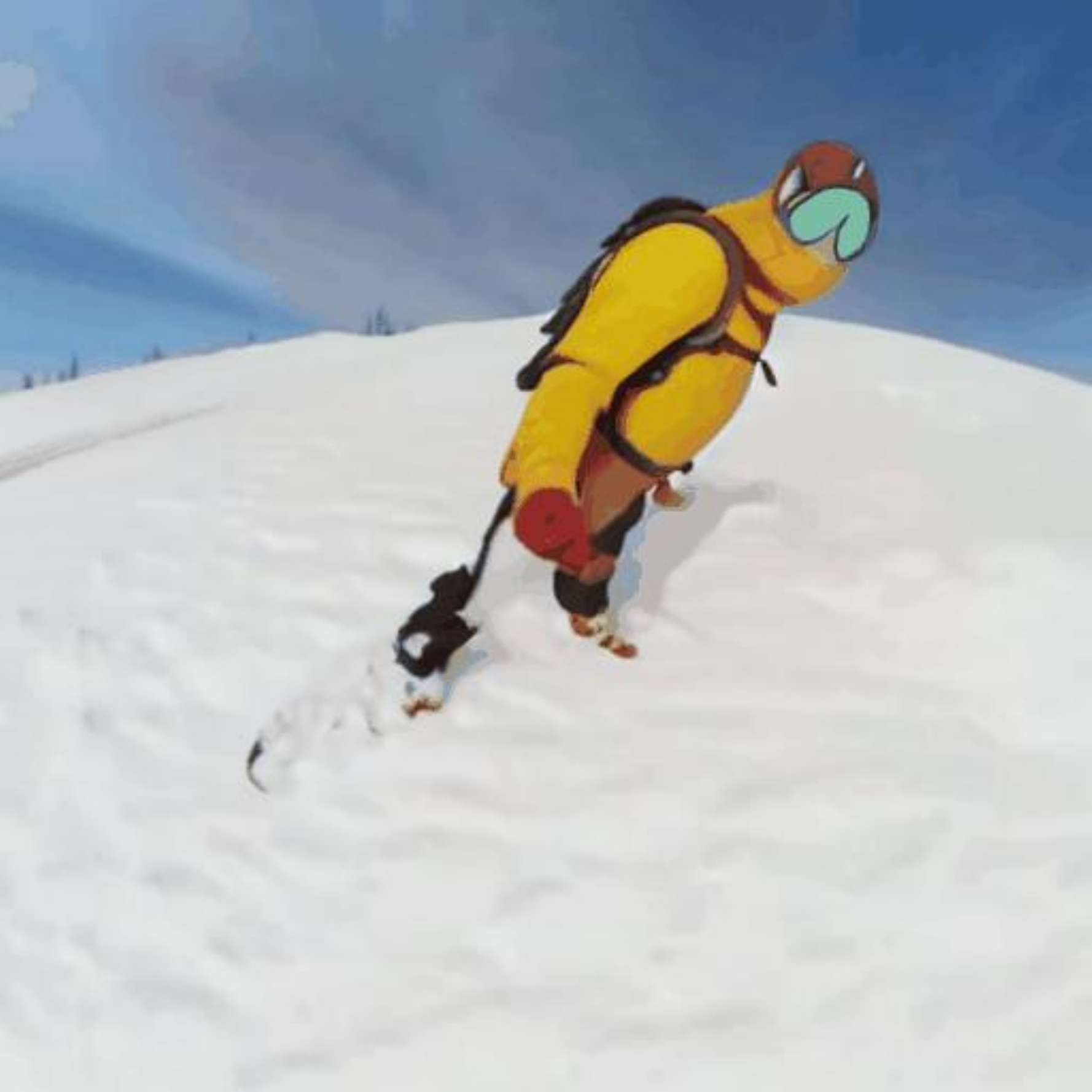}

\rotatebox{90}{\parbox{0.10\textwidth}{\centering duration \\ 0.8}}
\includegraphics[width=0.10\textwidth]{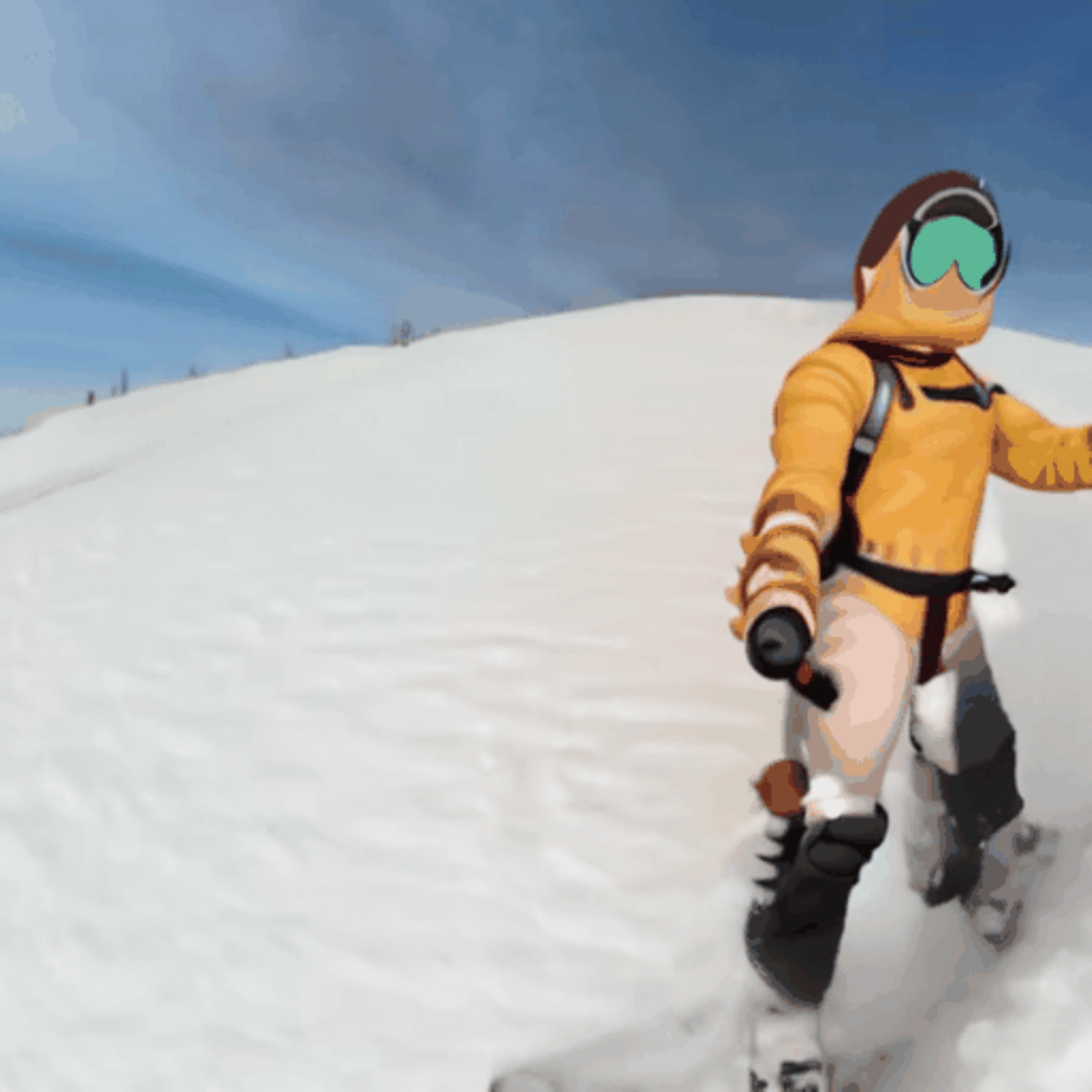}
\includegraphics[width=0.10\textwidth]{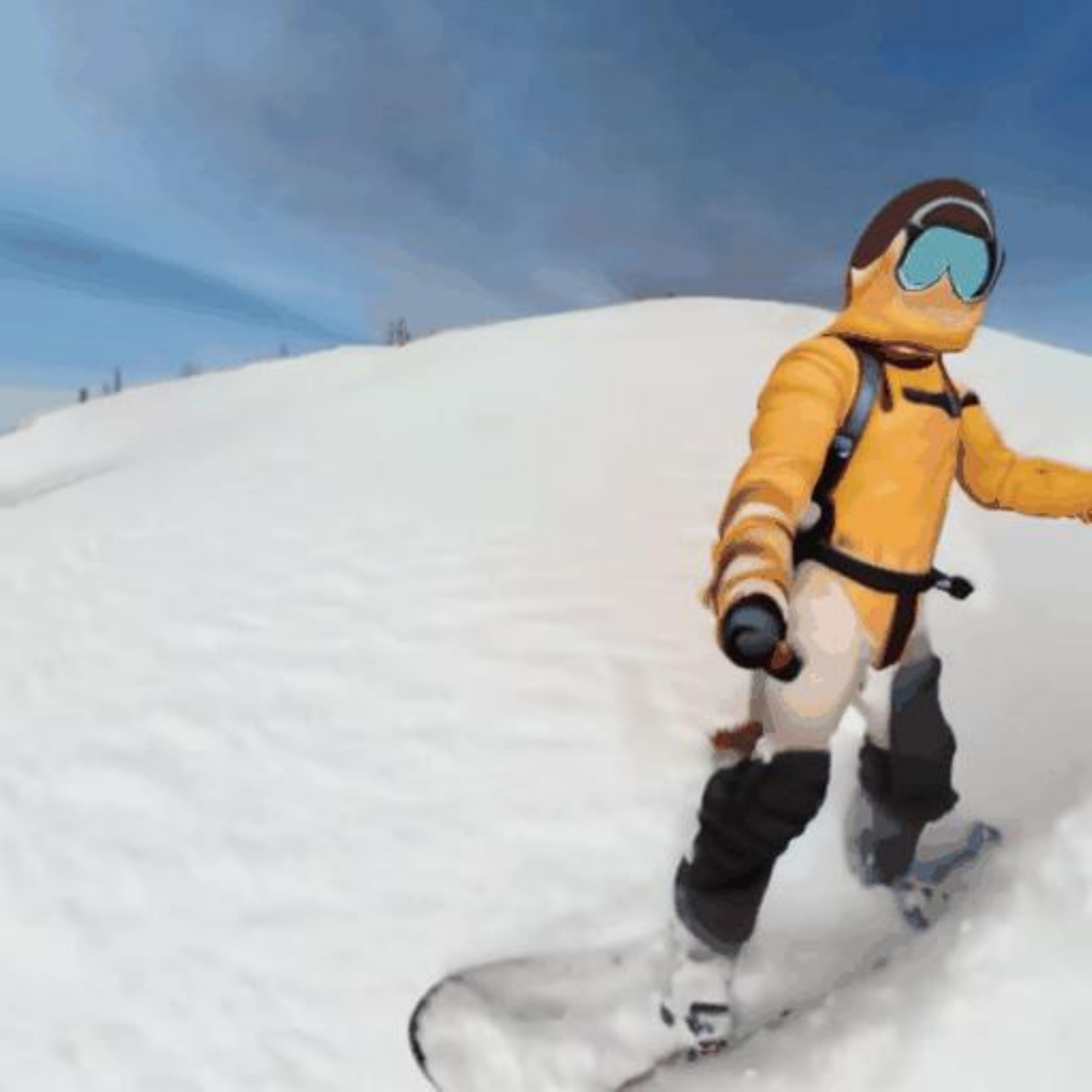}
\includegraphics[width=0.10\textwidth]{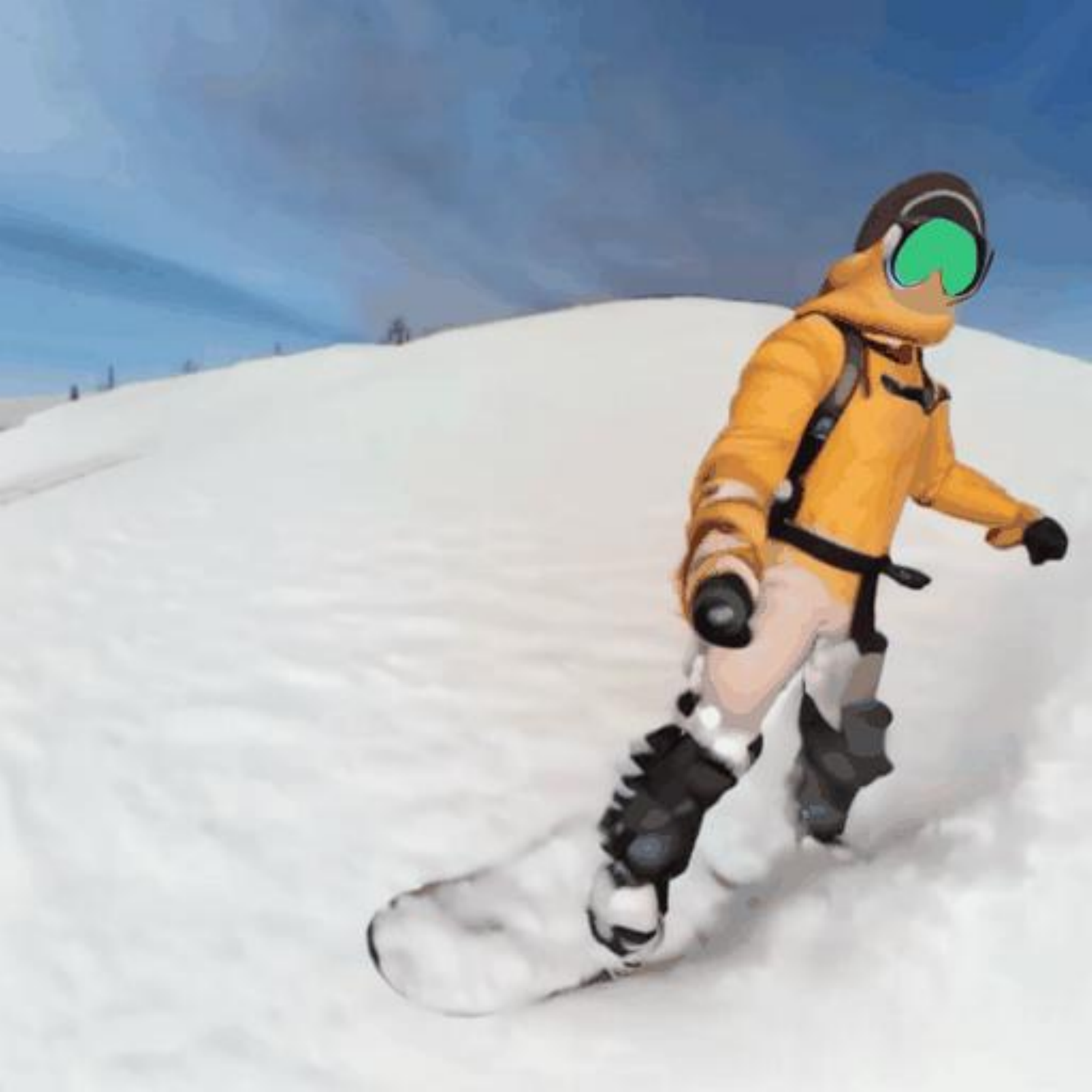}
\includegraphics[width=0.10\textwidth]{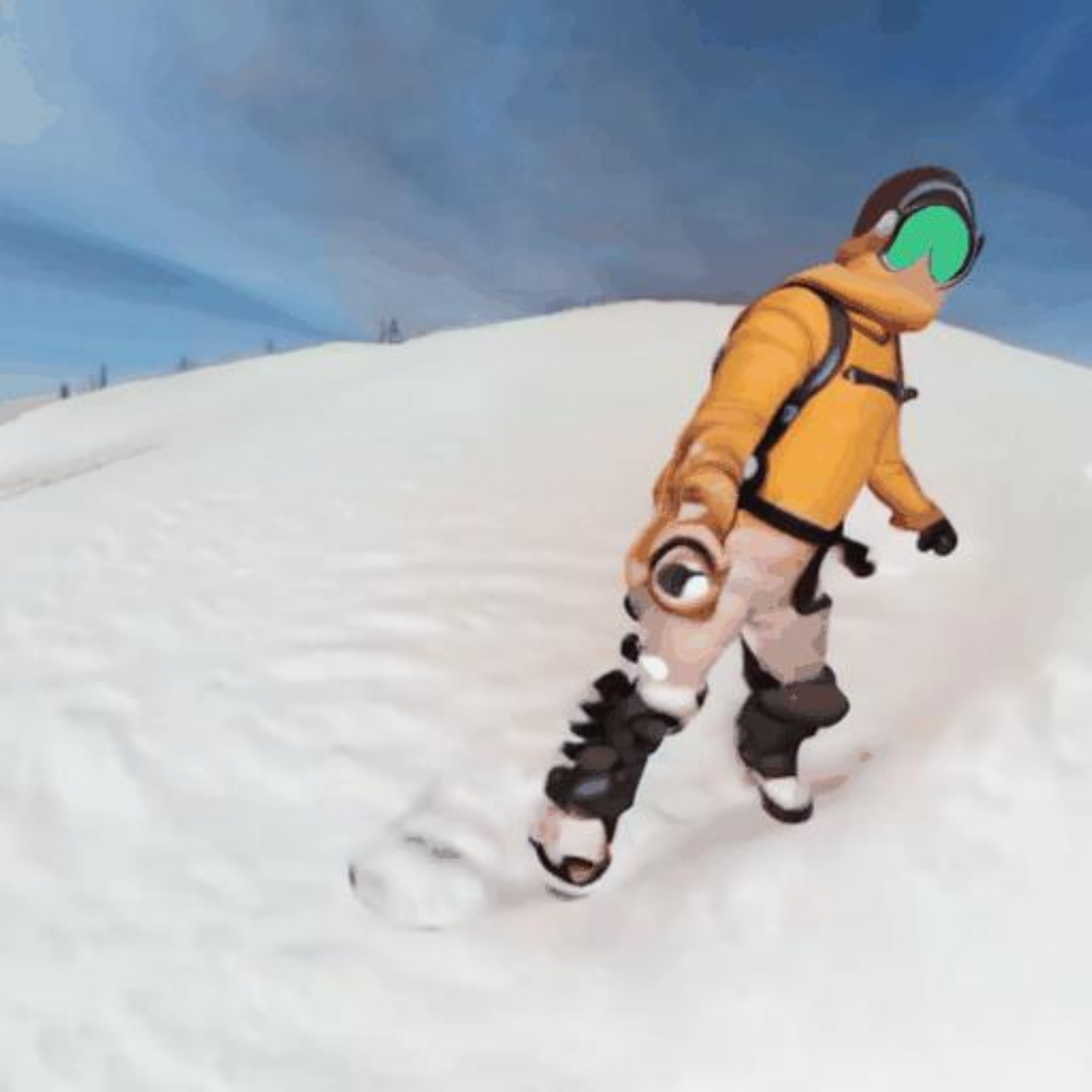}
\includegraphics[width=0.10\textwidth]{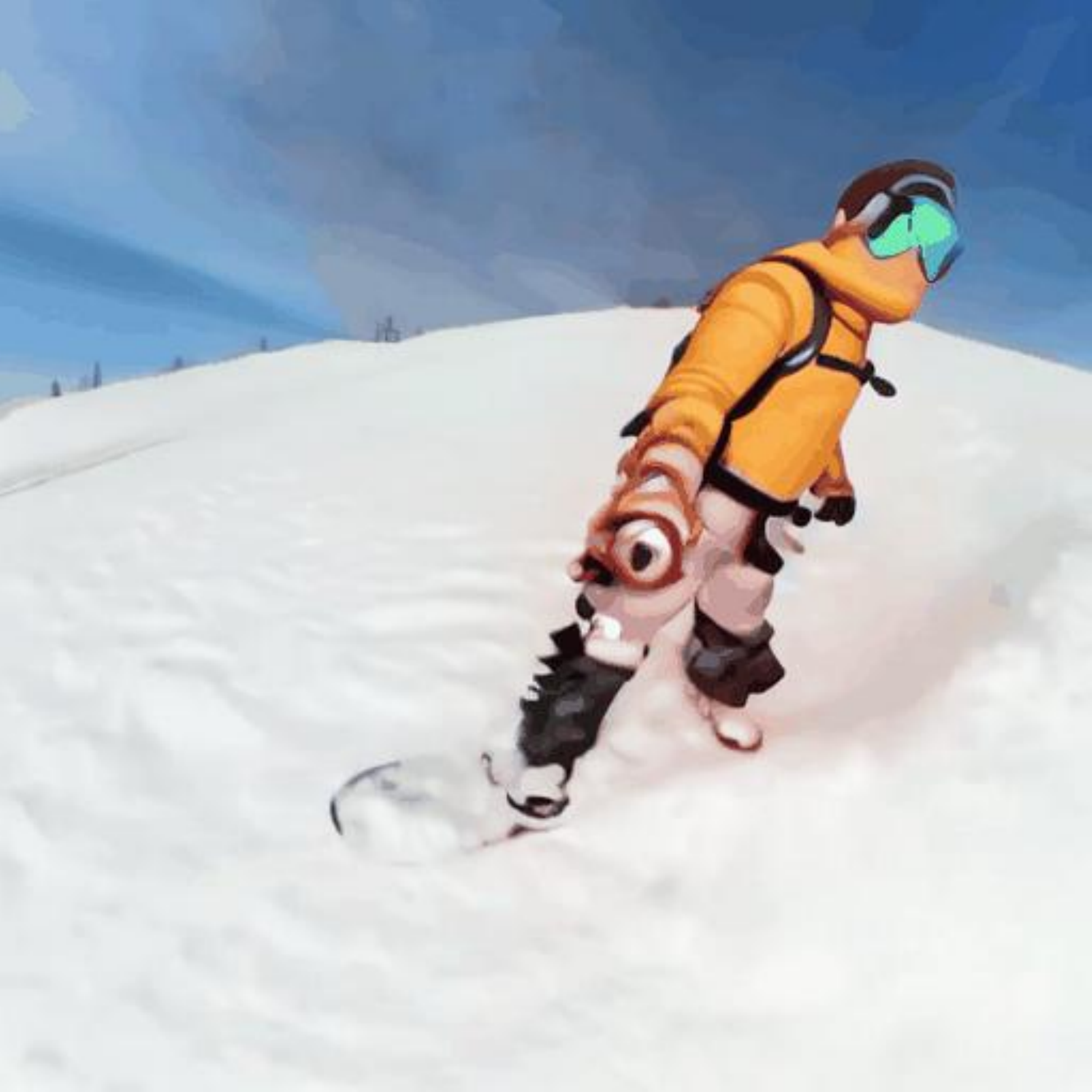}
\includegraphics[width=0.10\textwidth]{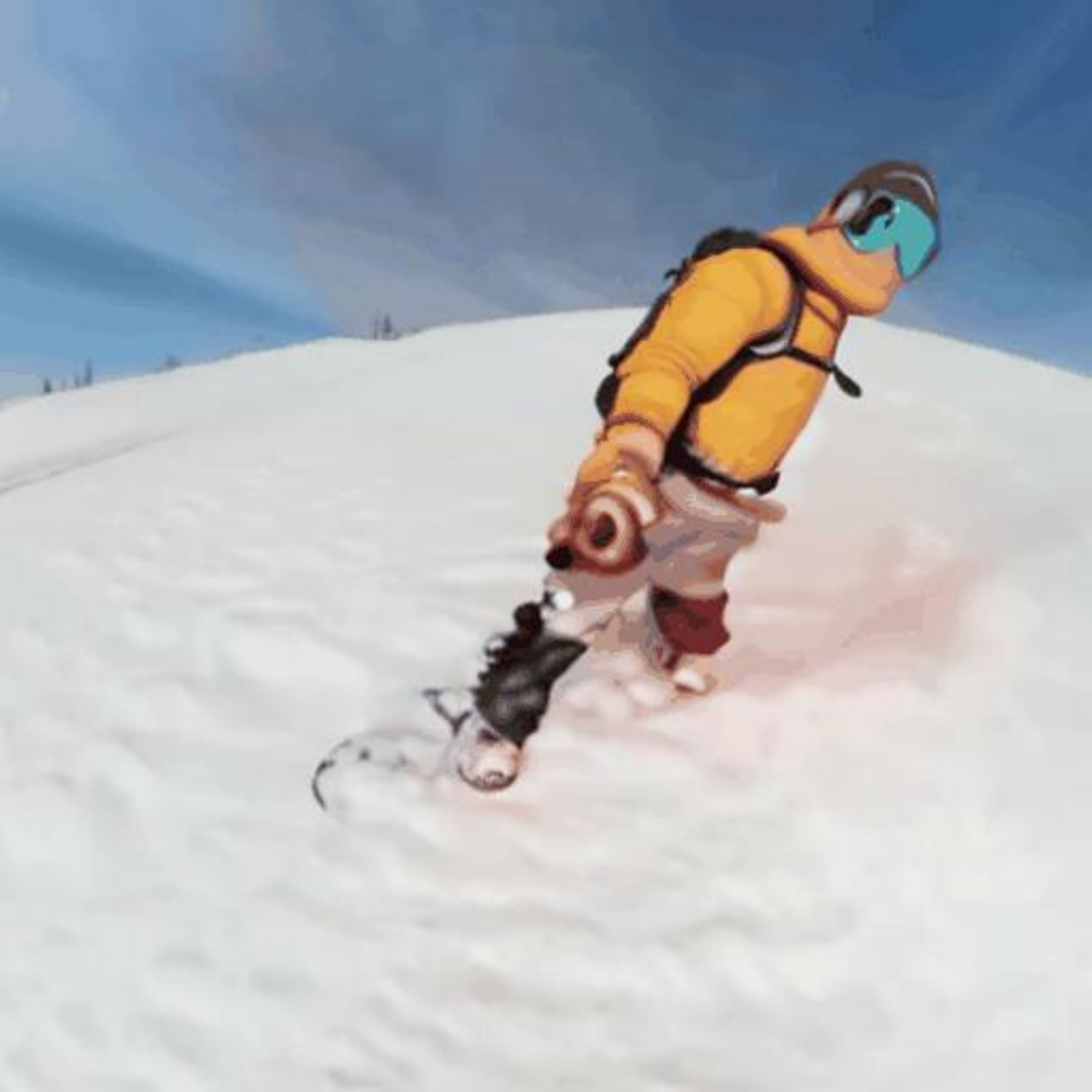}
\includegraphics[width=0.10\textwidth]{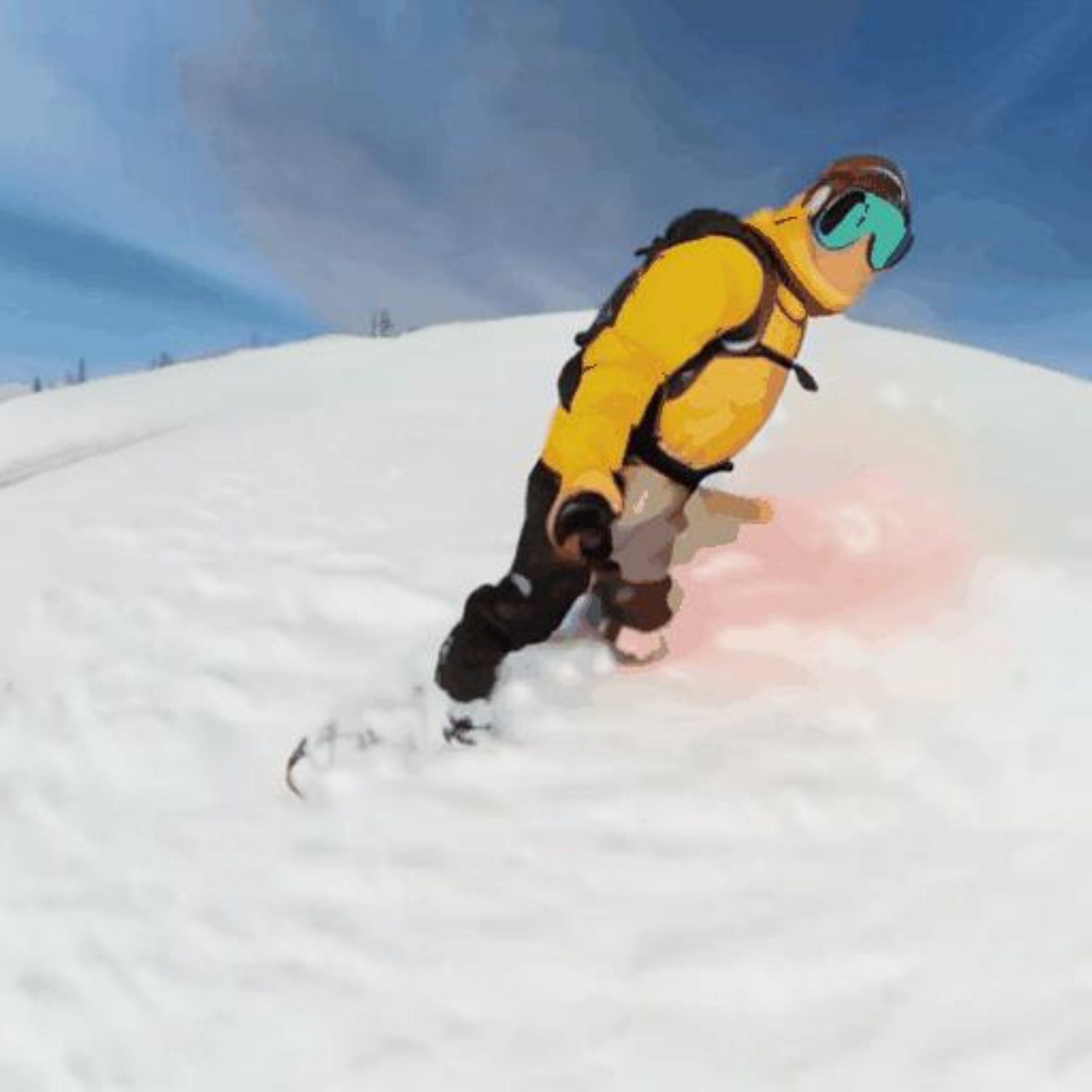}
\includegraphics[width=0.10\textwidth]{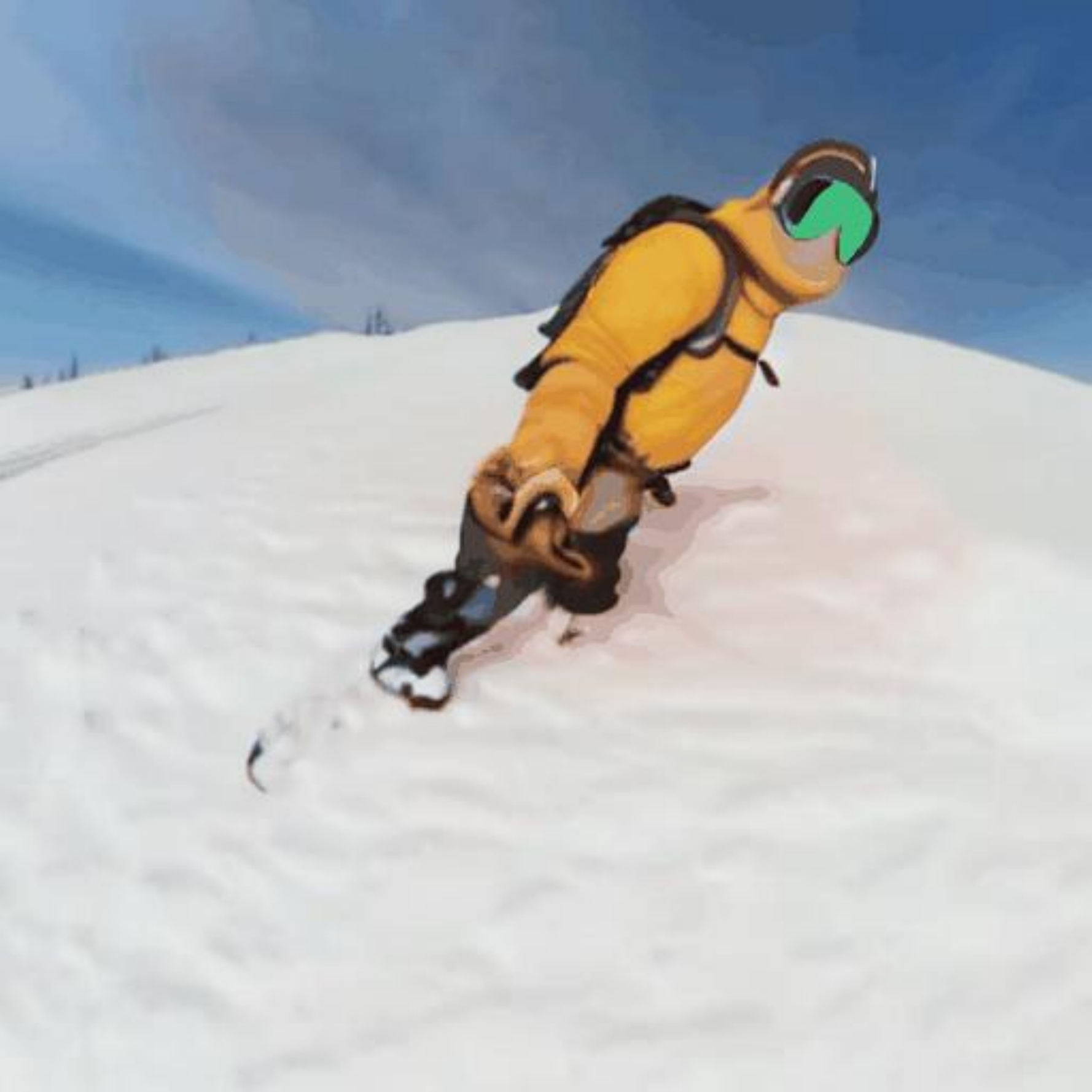}

\makebox[0.12\textwidth]{\colorbox{green}{\textbf{Temporal attention}} A \textcolor{blue}{\textbf{Spider Man}} is skiing }\\
\rotatebox{90}{\parbox{0.10\textwidth}{\centering duration \\ 0.2}}
\includegraphics[width=0.10\textwidth]{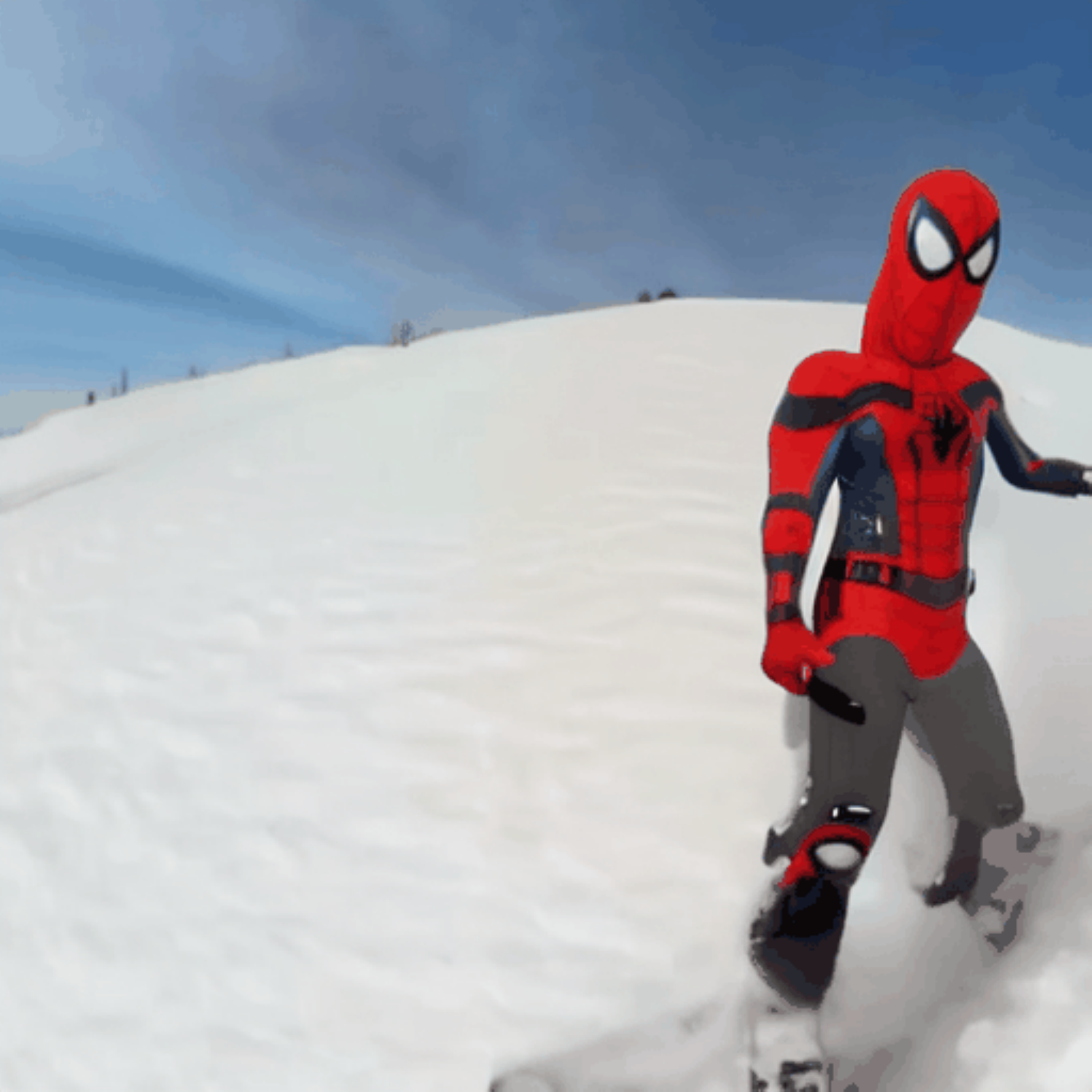}
\includegraphics[width=0.10\textwidth]{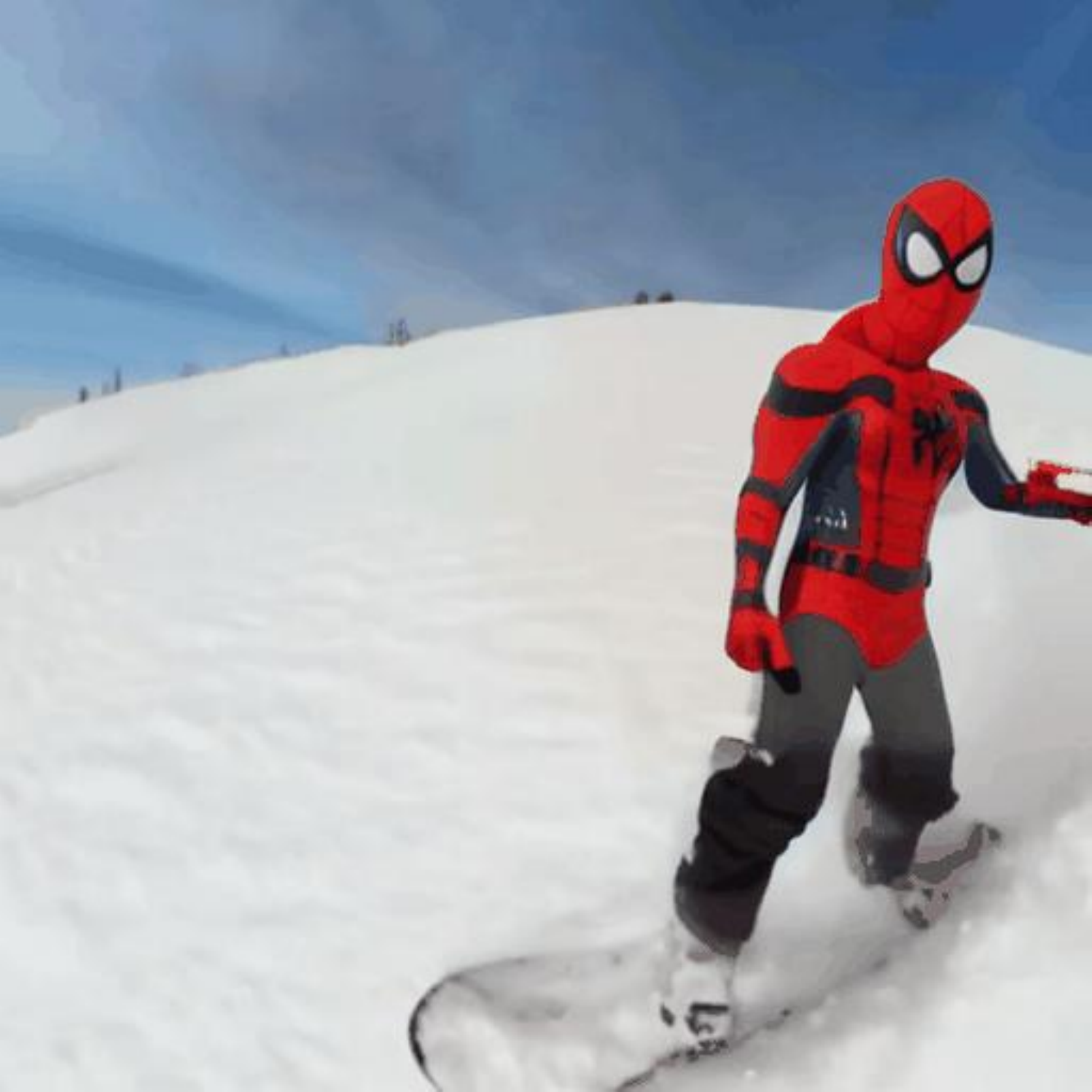}
\includegraphics[width=0.10\textwidth]{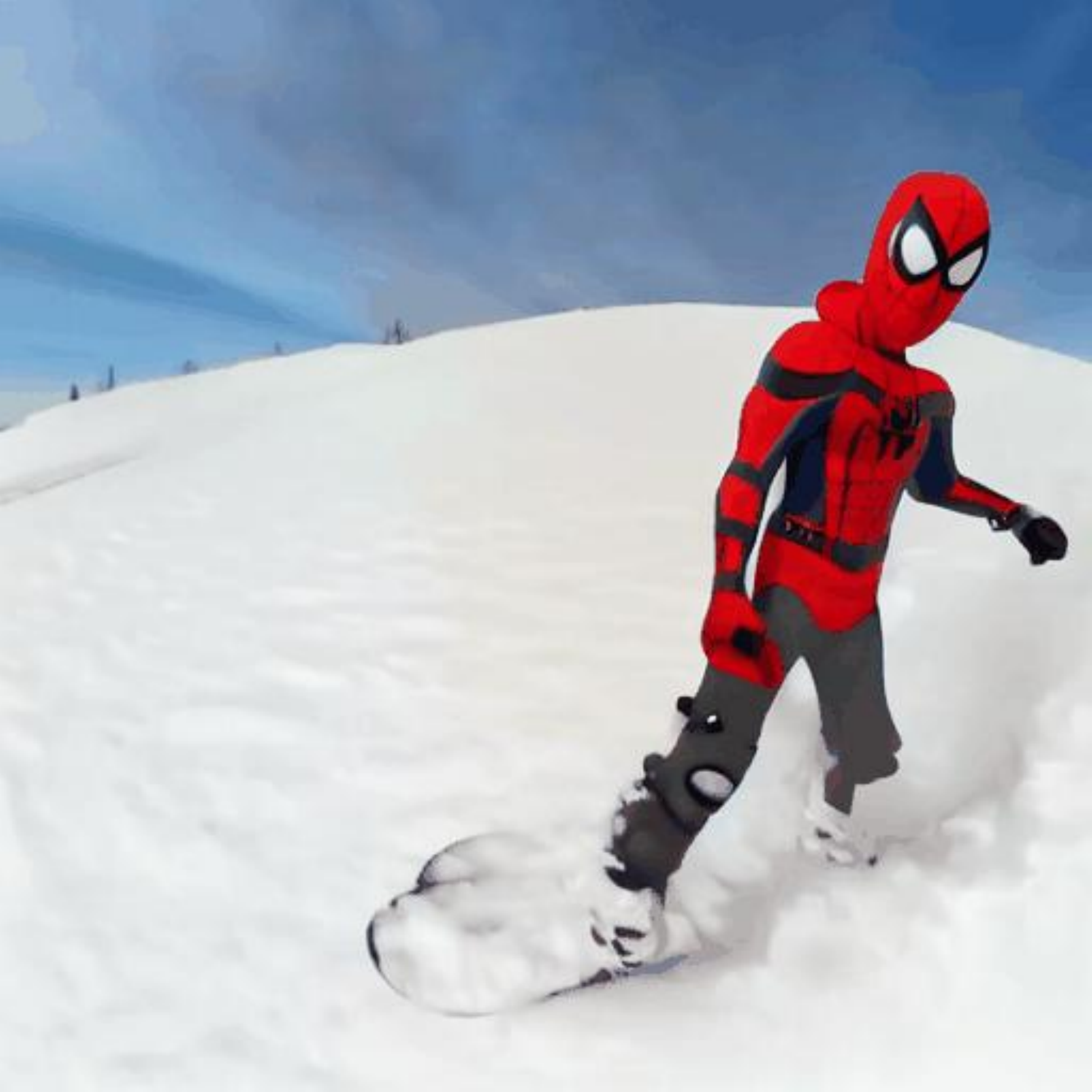}
\includegraphics[width=0.10\textwidth]{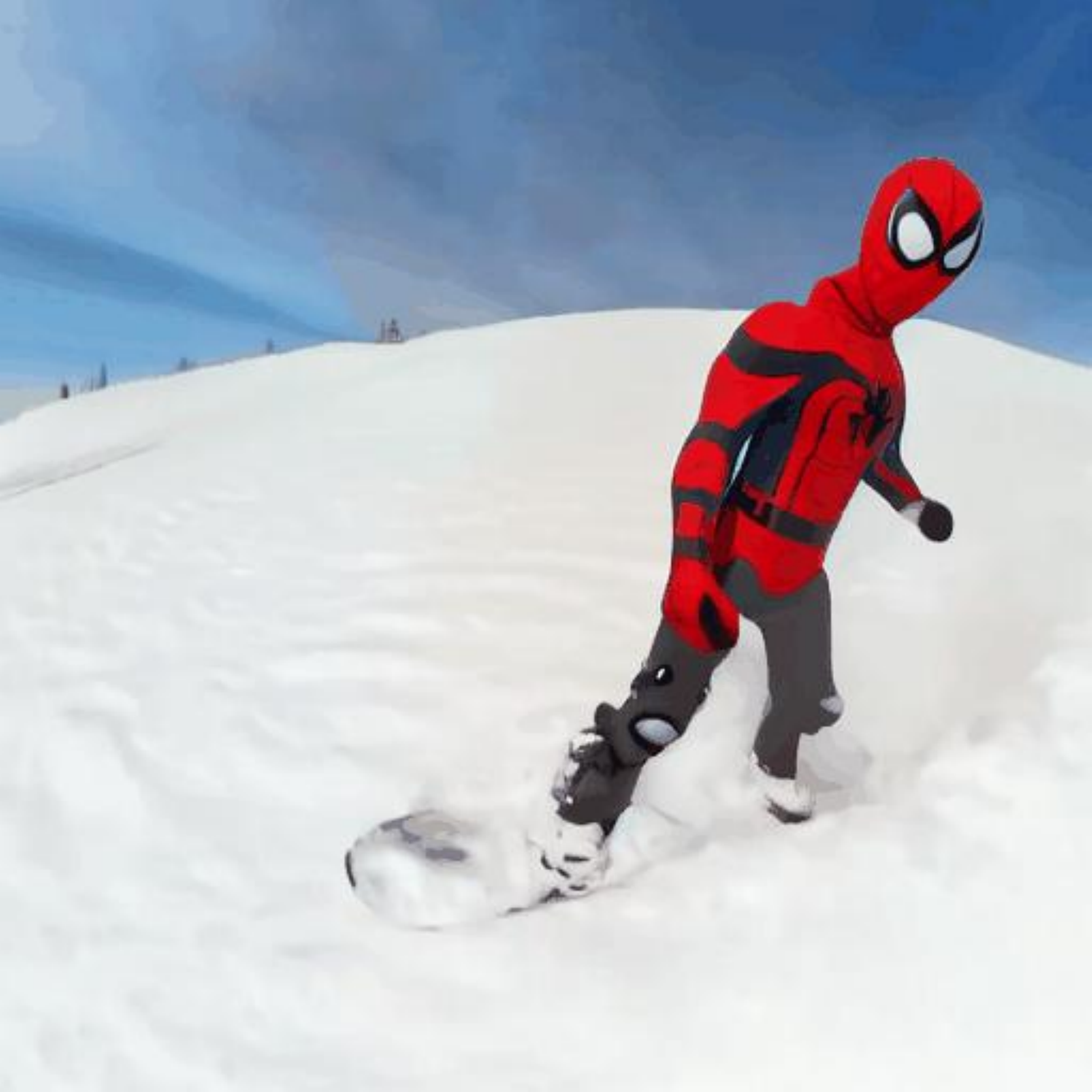}
\includegraphics[width=0.10\textwidth]{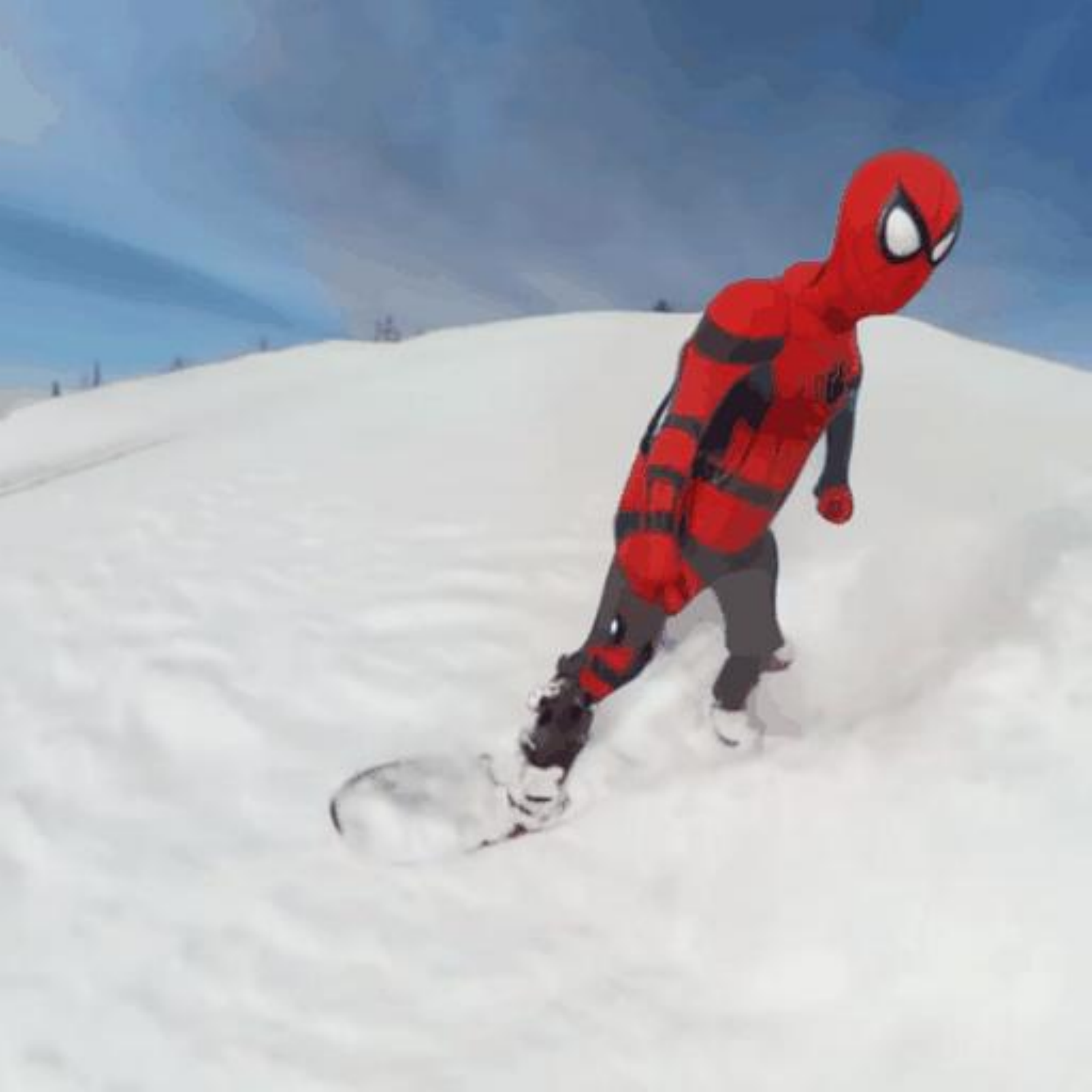}
\includegraphics[width=0.10\textwidth]{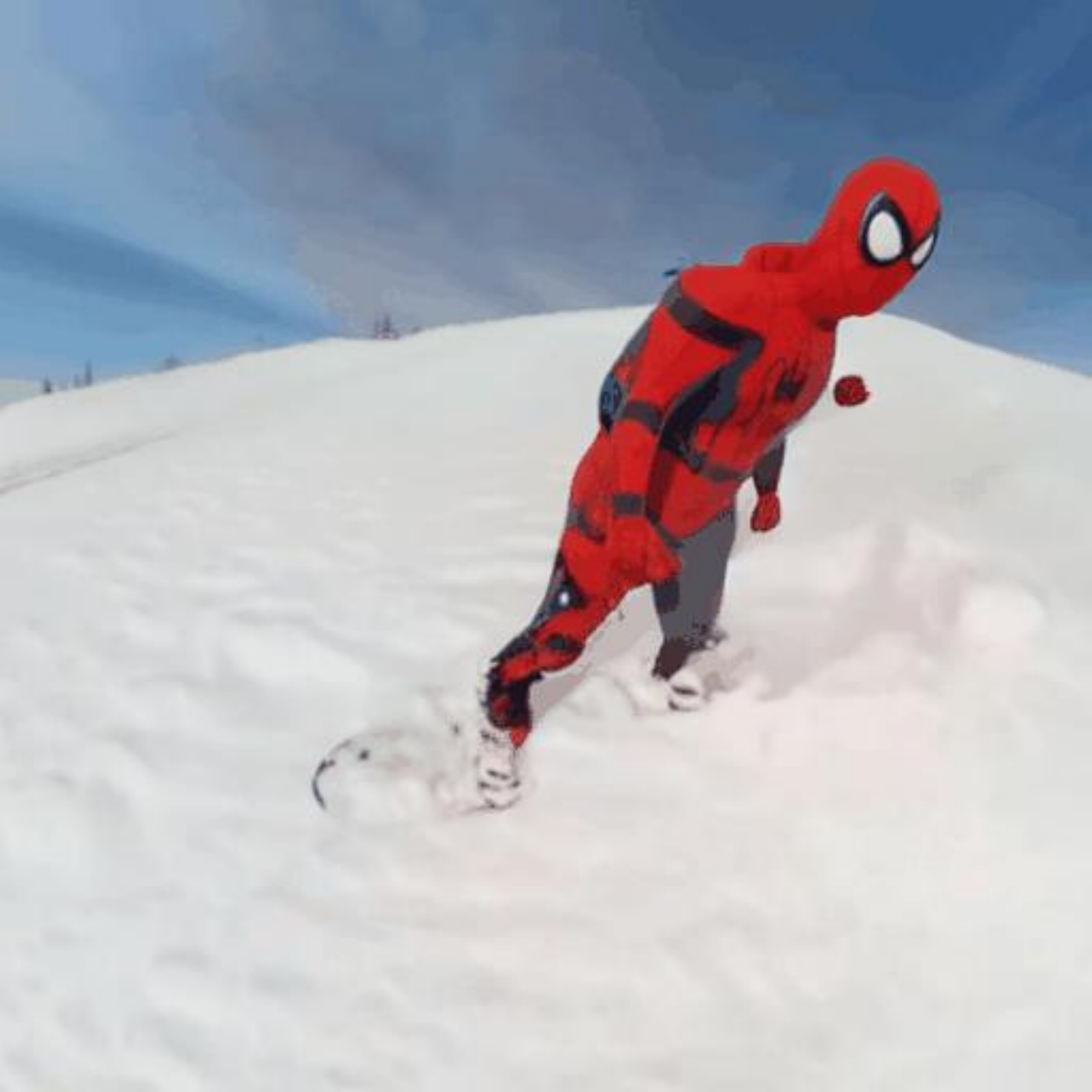}
\includegraphics[width=0.10\textwidth]{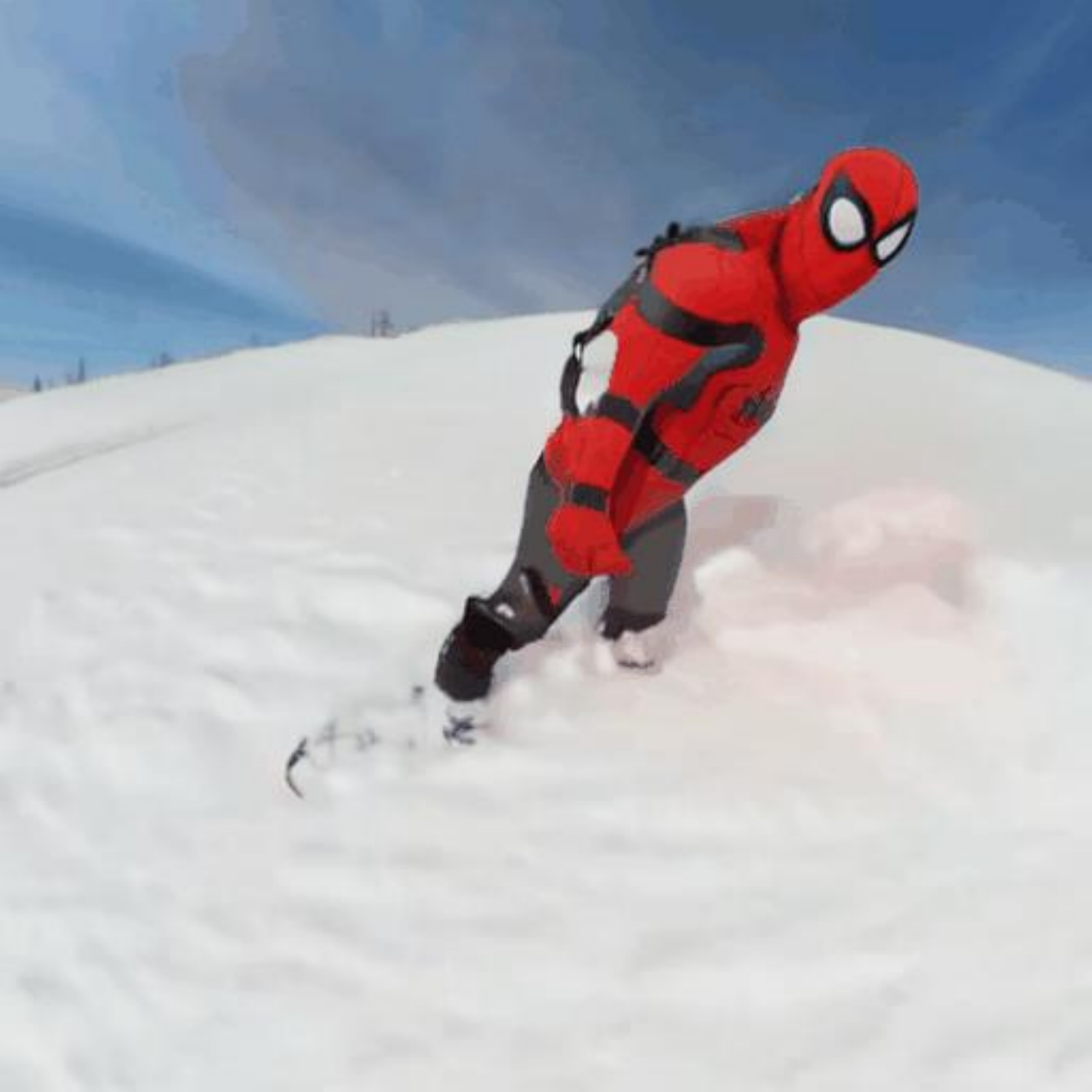}
\includegraphics[width=0.10\textwidth]{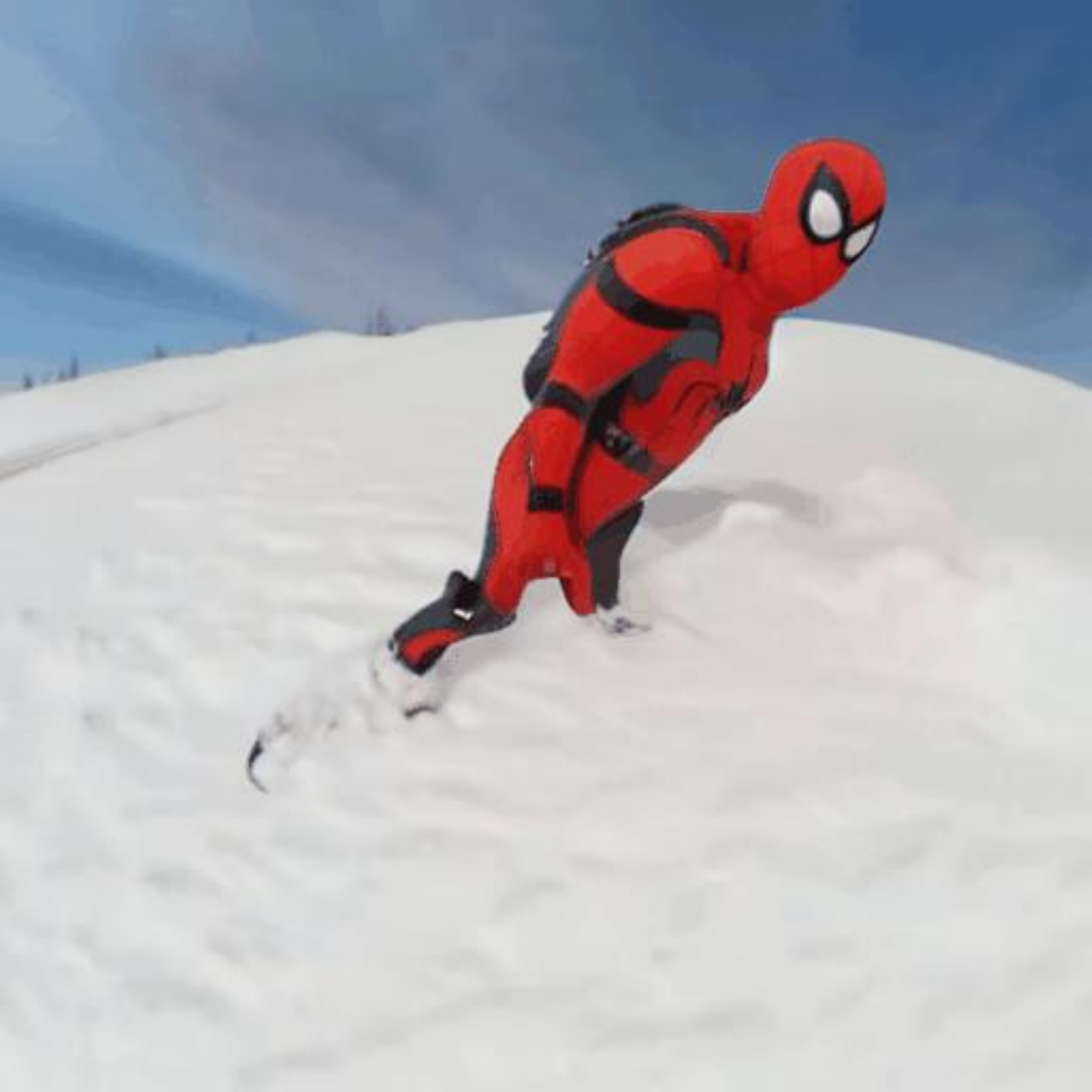}

\rotatebox{90}{\parbox{0.10\textwidth}{\centering duration \\ 0.5}}
\includegraphics[width=0.10\textwidth]{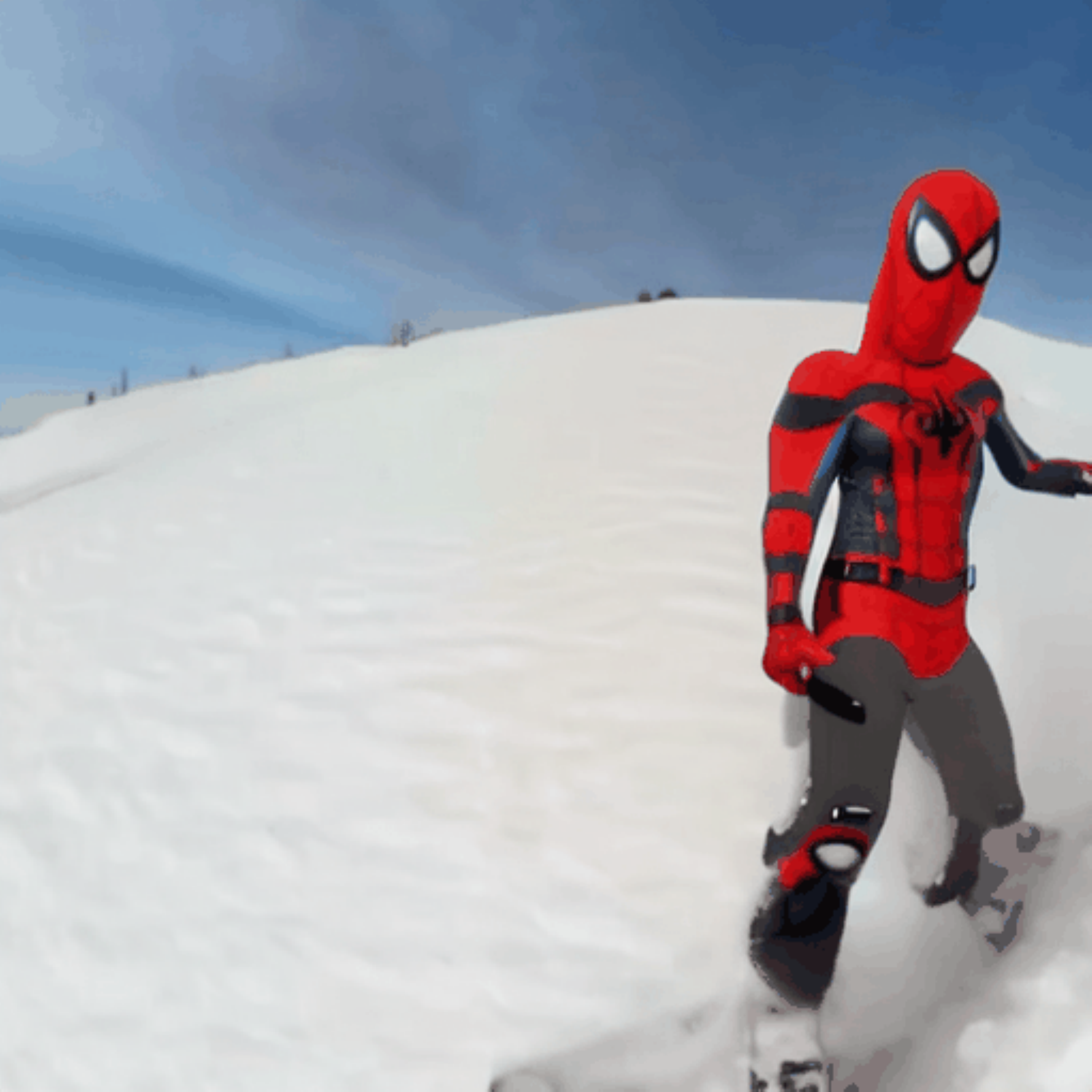}
\includegraphics[width=0.10\textwidth]{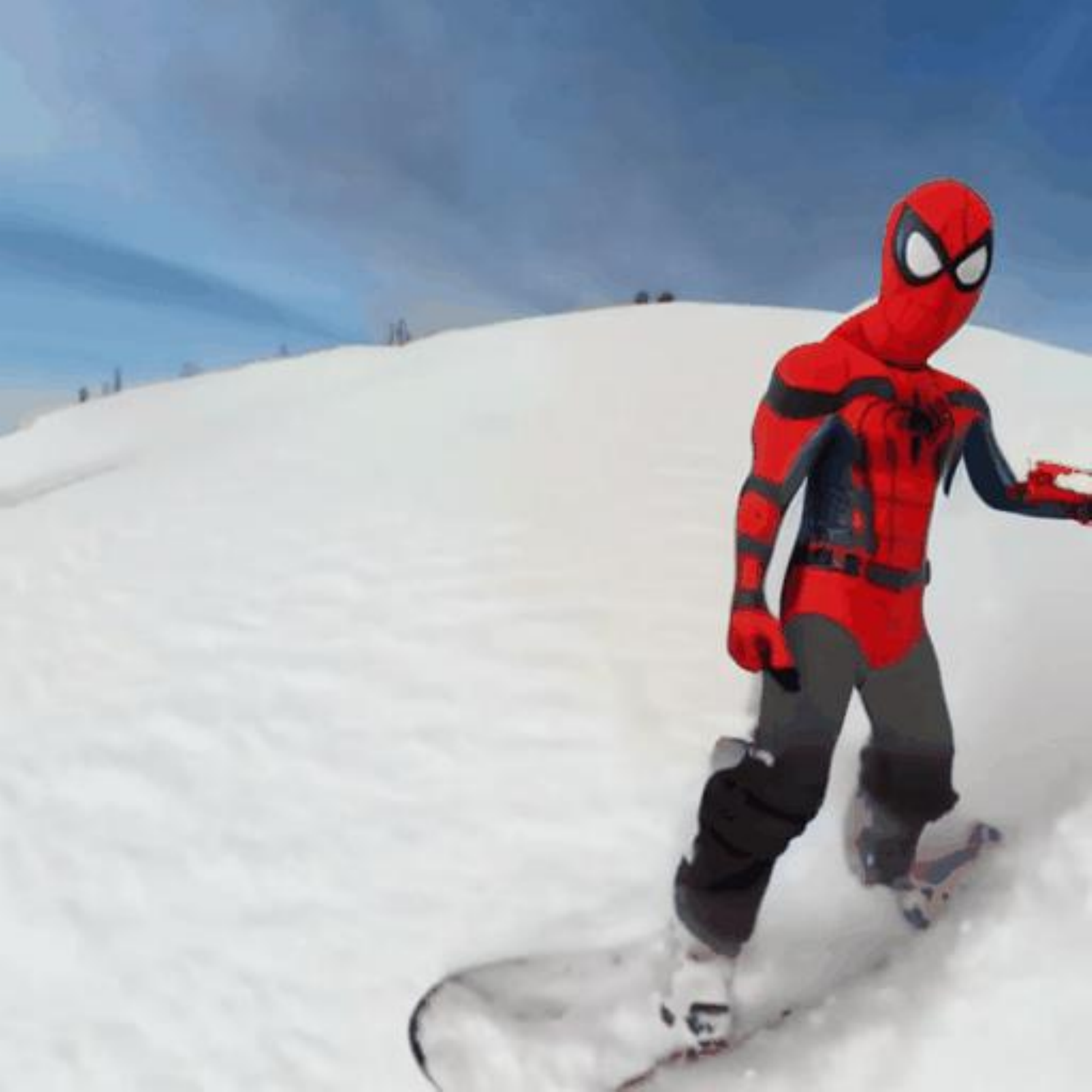}
\includegraphics[width=0.10\textwidth]{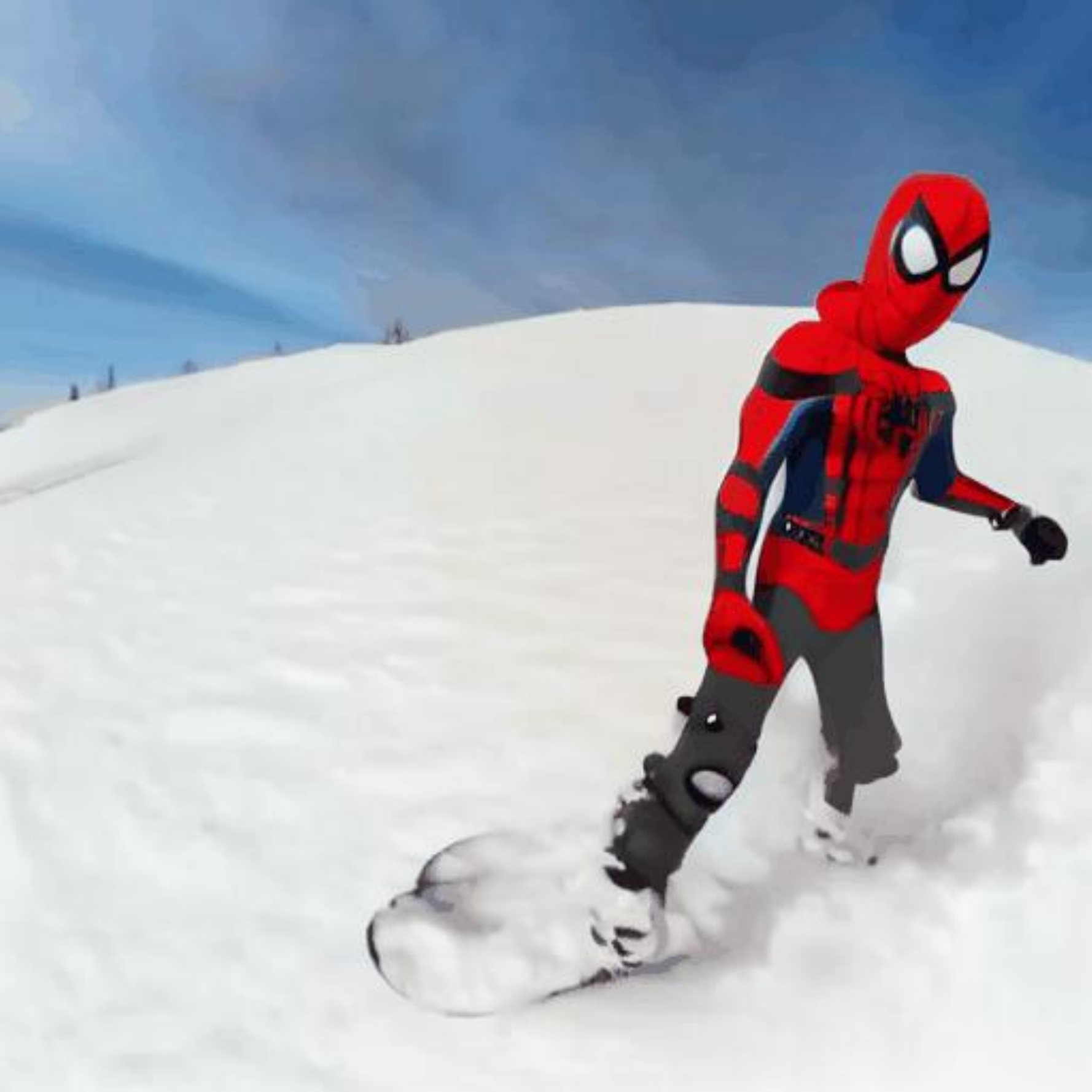}
\includegraphics[width=0.10\textwidth]{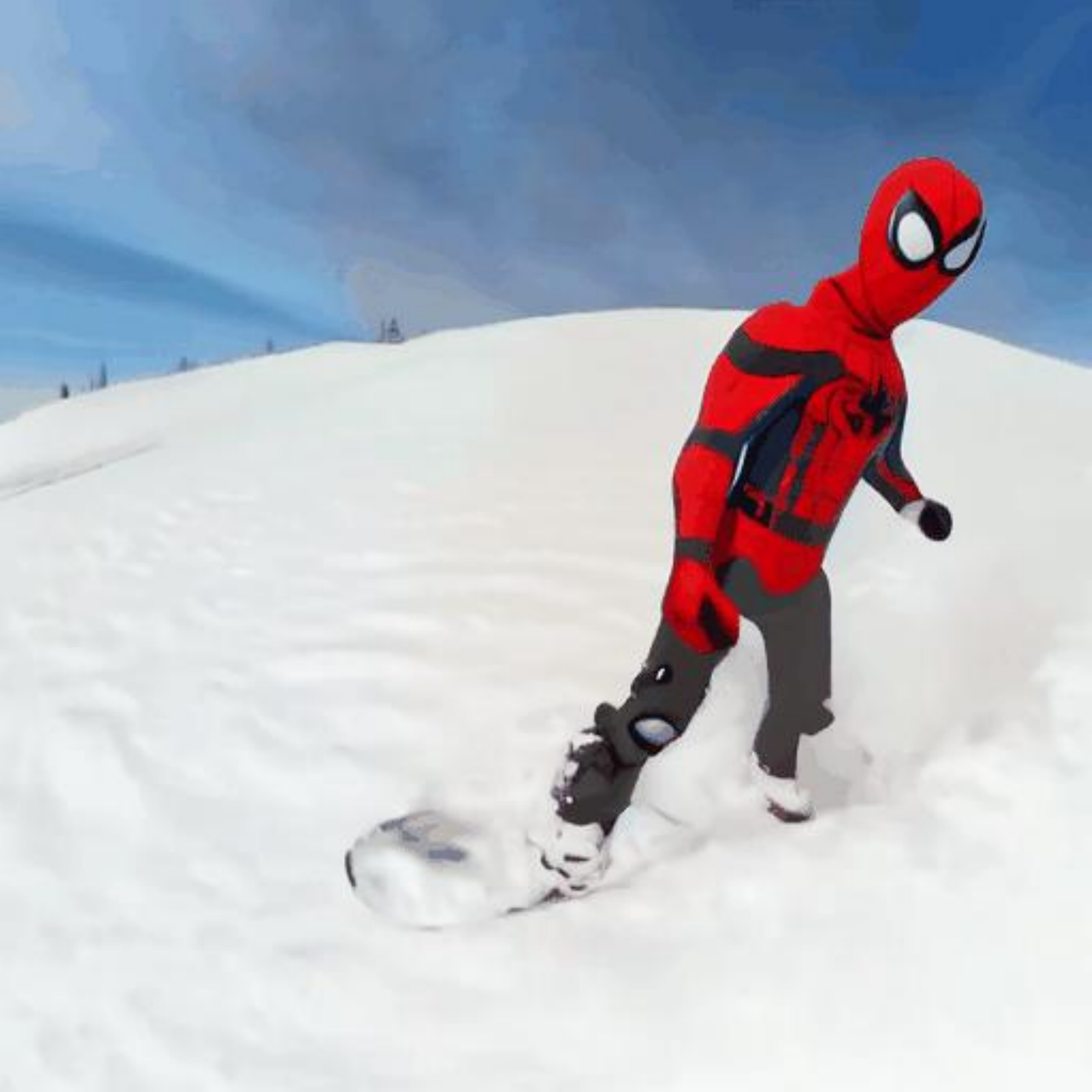}
\includegraphics[width=0.10\textwidth]{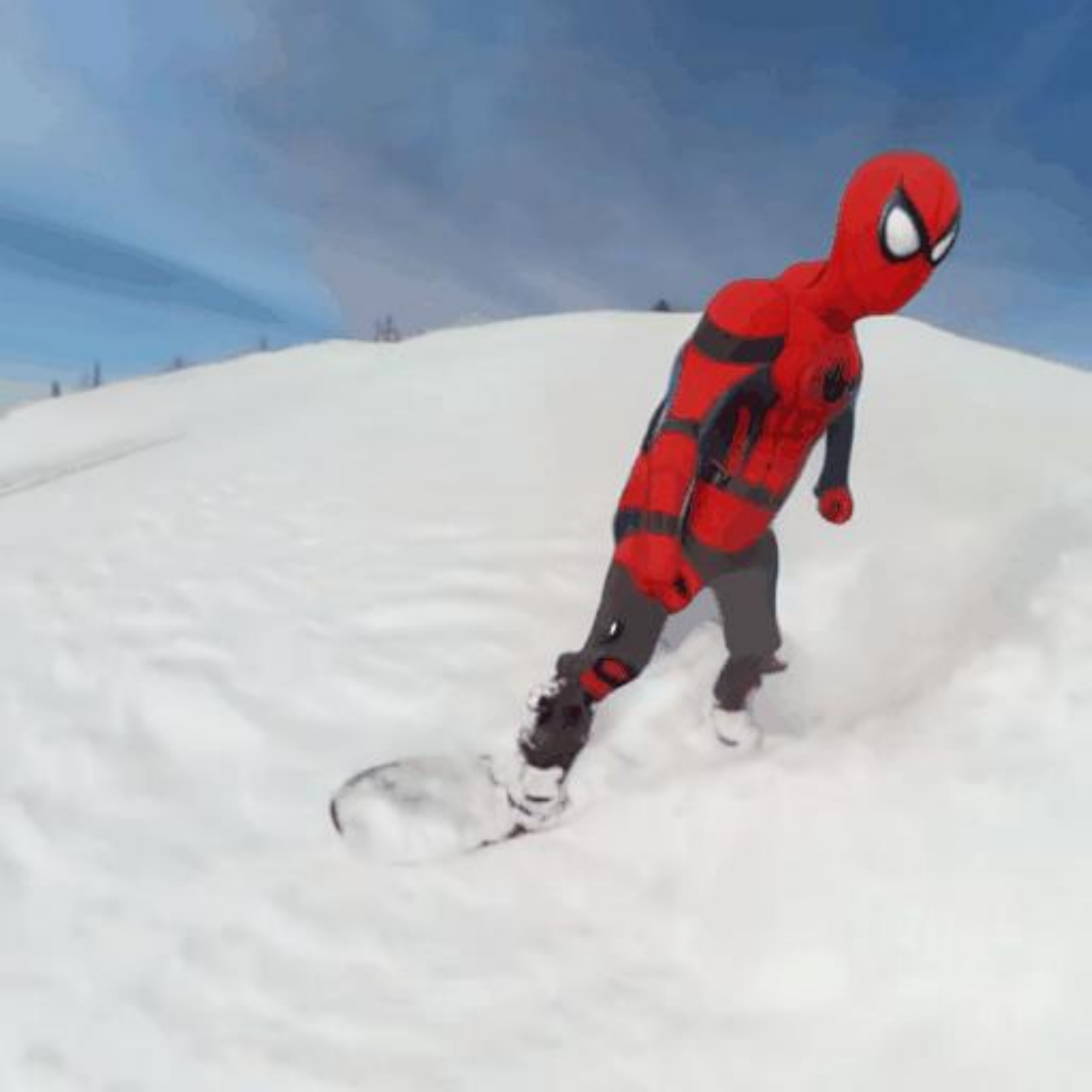}
\includegraphics[width=0.10\textwidth]{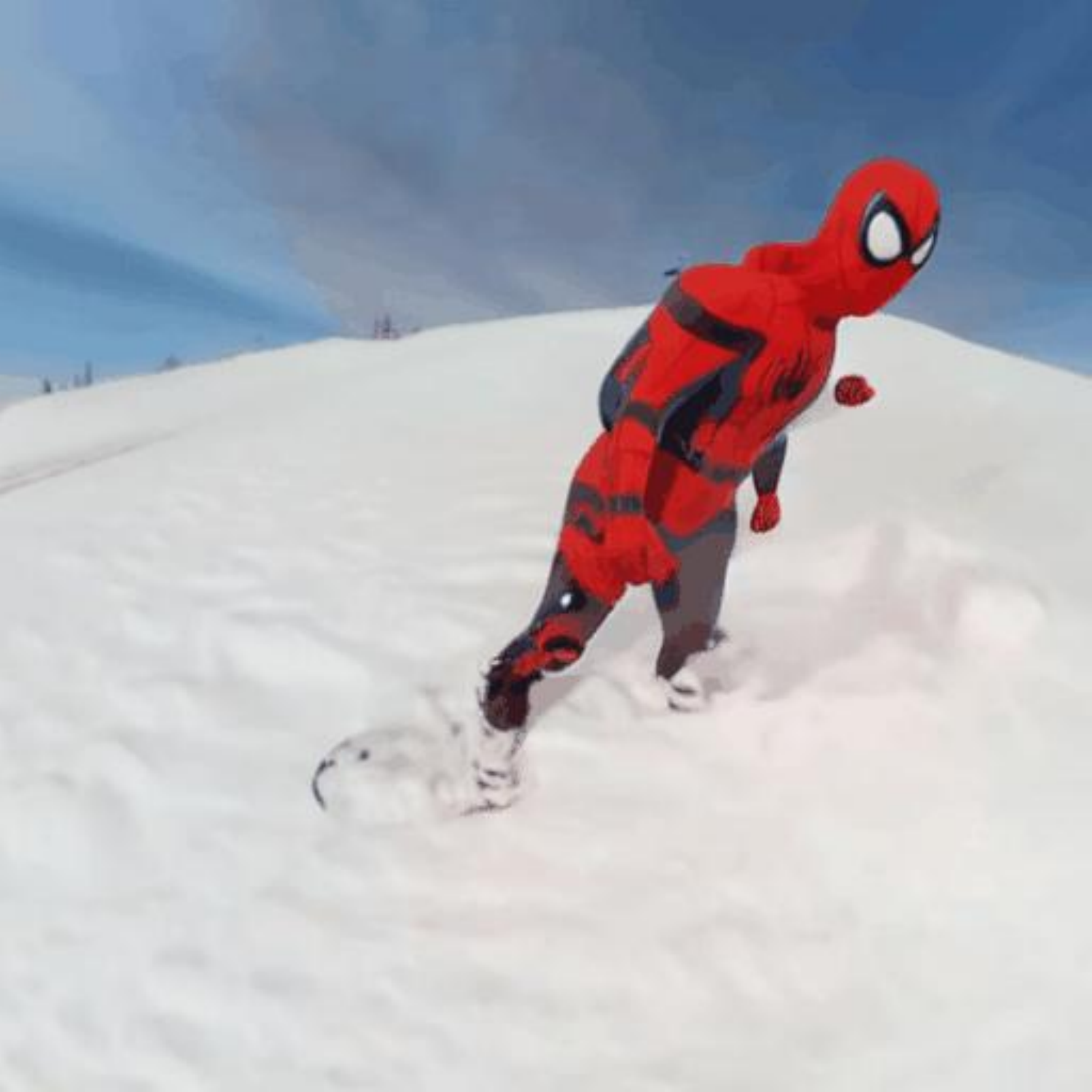}
\includegraphics[width=0.10\textwidth]{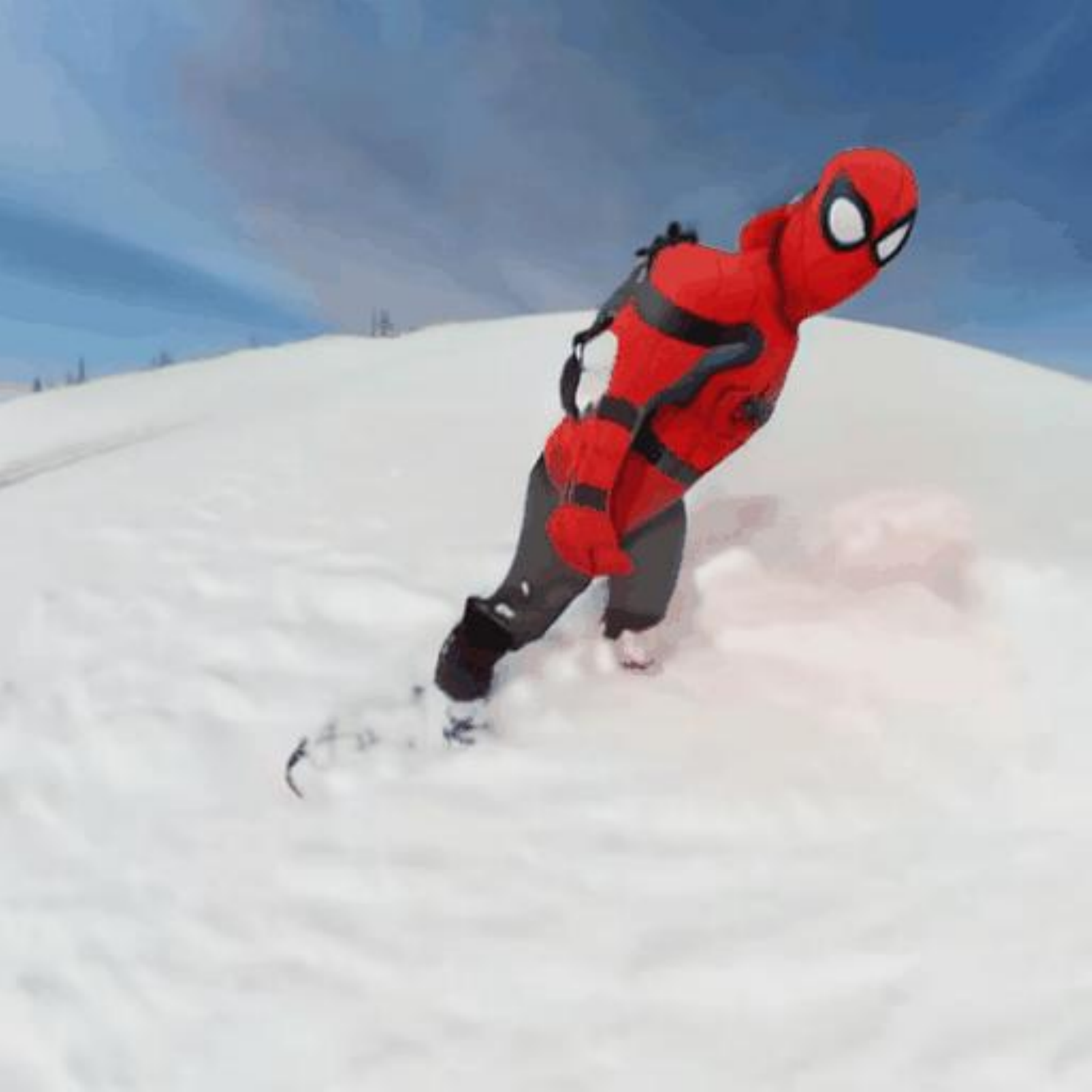}
\includegraphics[width=0.10\textwidth]{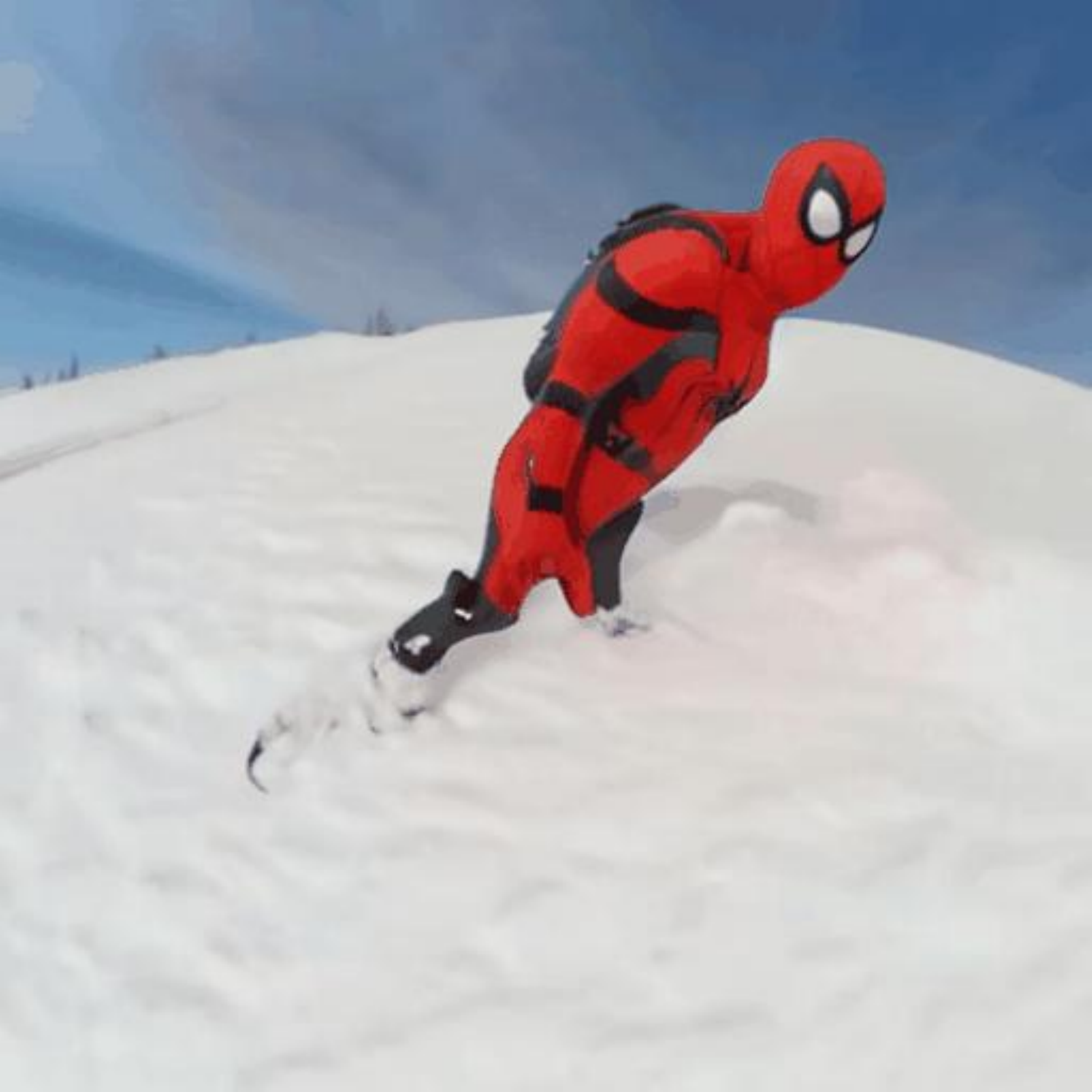}

\rotatebox{90}{\parbox{0.10\textwidth}{\centering \textbf{duration \\ 0.8}}}
\includegraphics[width=0.10\textwidth]{figures/attn_anal/self0.8_sparse0.5_cross0.2/frame_0.pdf}
\includegraphics[width=0.10\textwidth]{figures/attn_anal/self0.8_sparse0.5_cross0.2/frame_1.pdf}
\includegraphics[width=0.10\textwidth]{figures/attn_anal/self0.8_sparse0.5_cross0.2/frame_2.pdf}
\includegraphics[width=0.10\textwidth]{figures/attn_anal/self0.8_sparse0.5_cross0.2/frame_3.pdf}
\includegraphics[width=0.10\textwidth]{figures/attn_anal/self0.8_sparse0.5_cross0.2/frame_4.pdf}
\includegraphics[width=0.10\textwidth]{figures/attn_anal/self0.8_sparse0.5_cross0.2/frame_5.pdf}
\includegraphics[width=0.10\textwidth]{figures/attn_anal/self0.8_sparse0.5_cross0.2/frame_6.pdf}
\includegraphics[width=0.10\textwidth]{figures/attn_anal/self0.8_sparse0.5_cross0.2/frame_7.pdf}

\makebox[0.12\textwidth]{\colorbox{green}{\textbf{Sparse Spatio-Temporal attention}} A \textcolor{blue}{\textbf{Spider Man}} is skiing }\\
\rotatebox{90}{\parbox{0.10\textwidth}{\centering duration \\ 0.2}}
\includegraphics[width=0.10\textwidth]{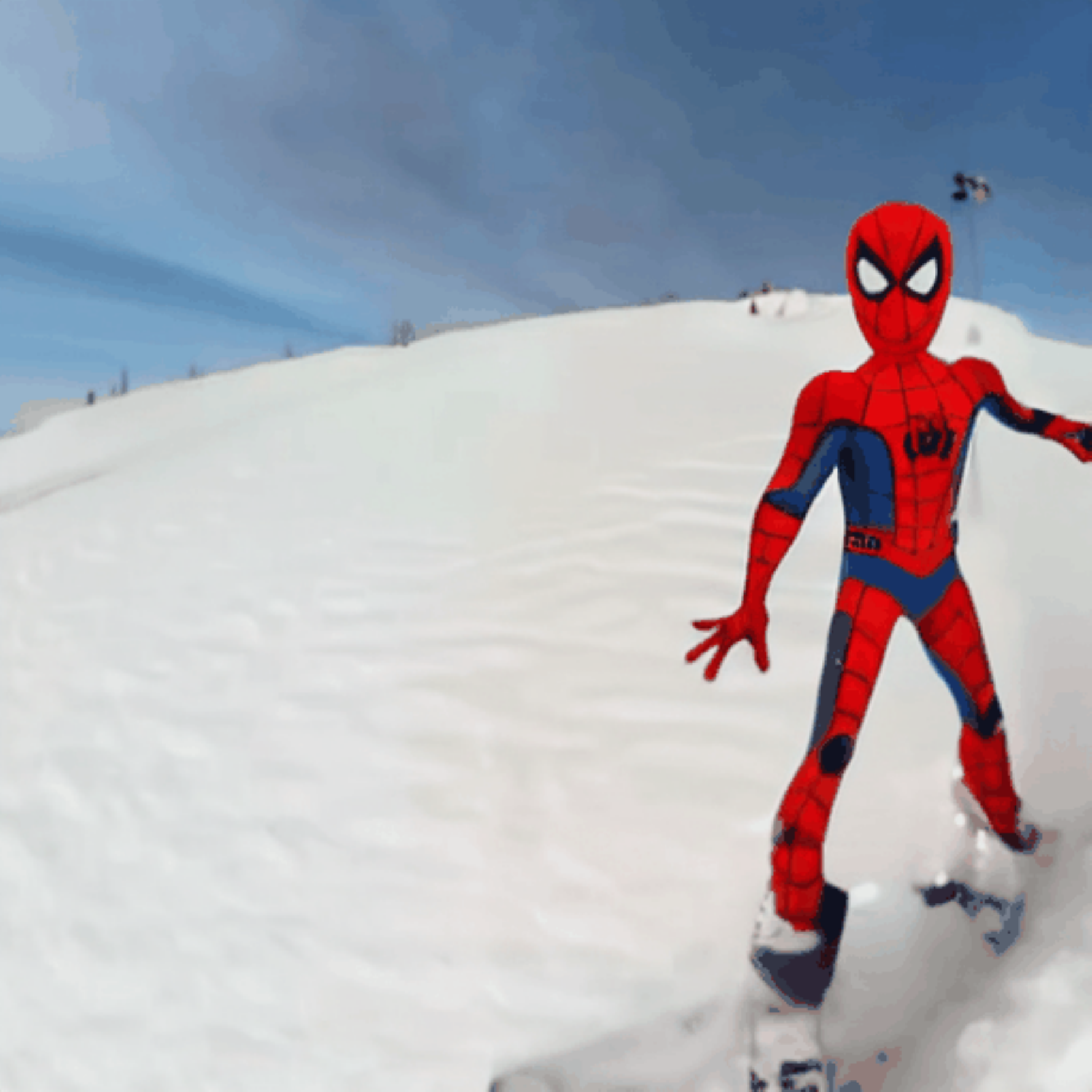}
\includegraphics[width=0.10\textwidth]{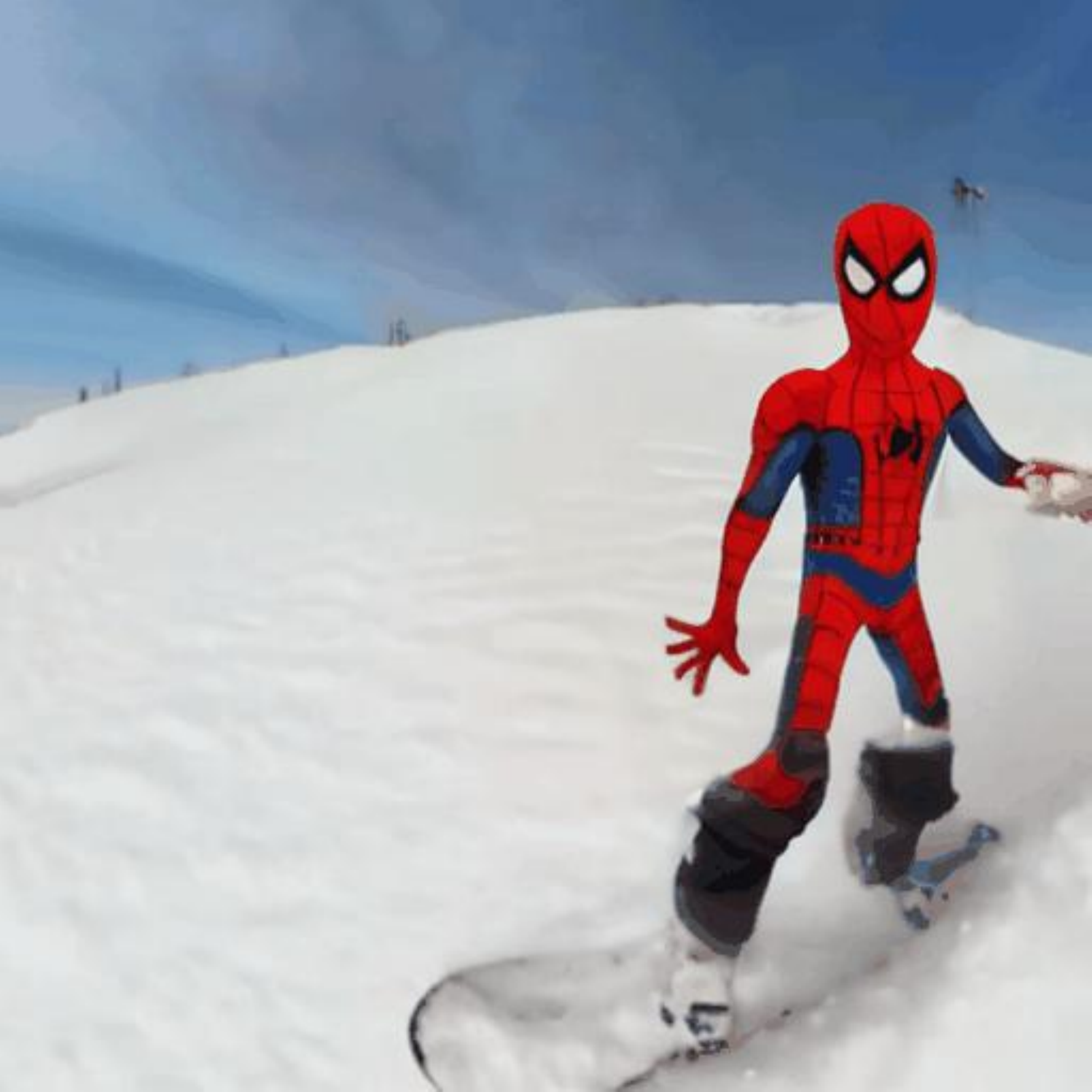}
\includegraphics[width=0.10\textwidth]{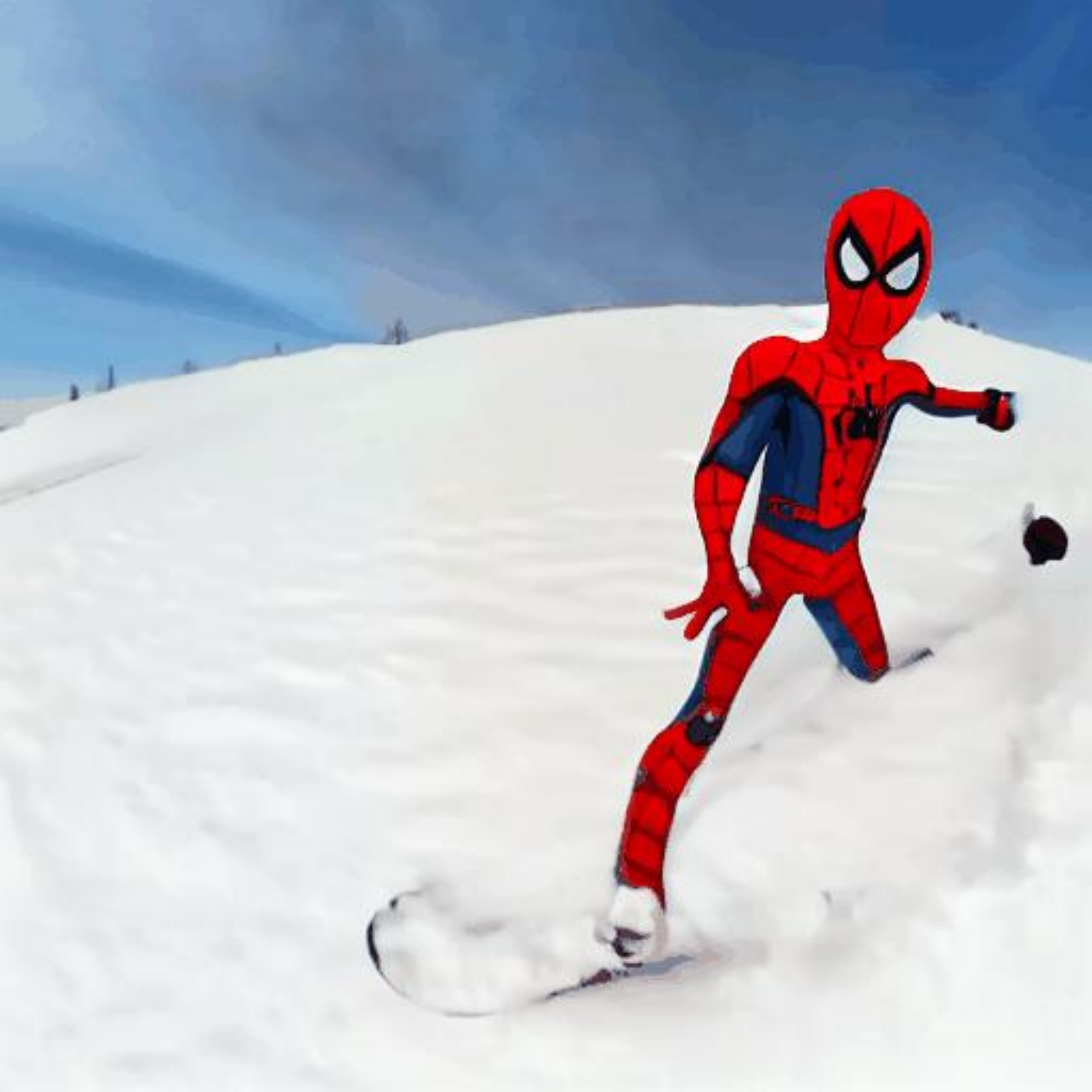}
\includegraphics[width=0.10\textwidth]{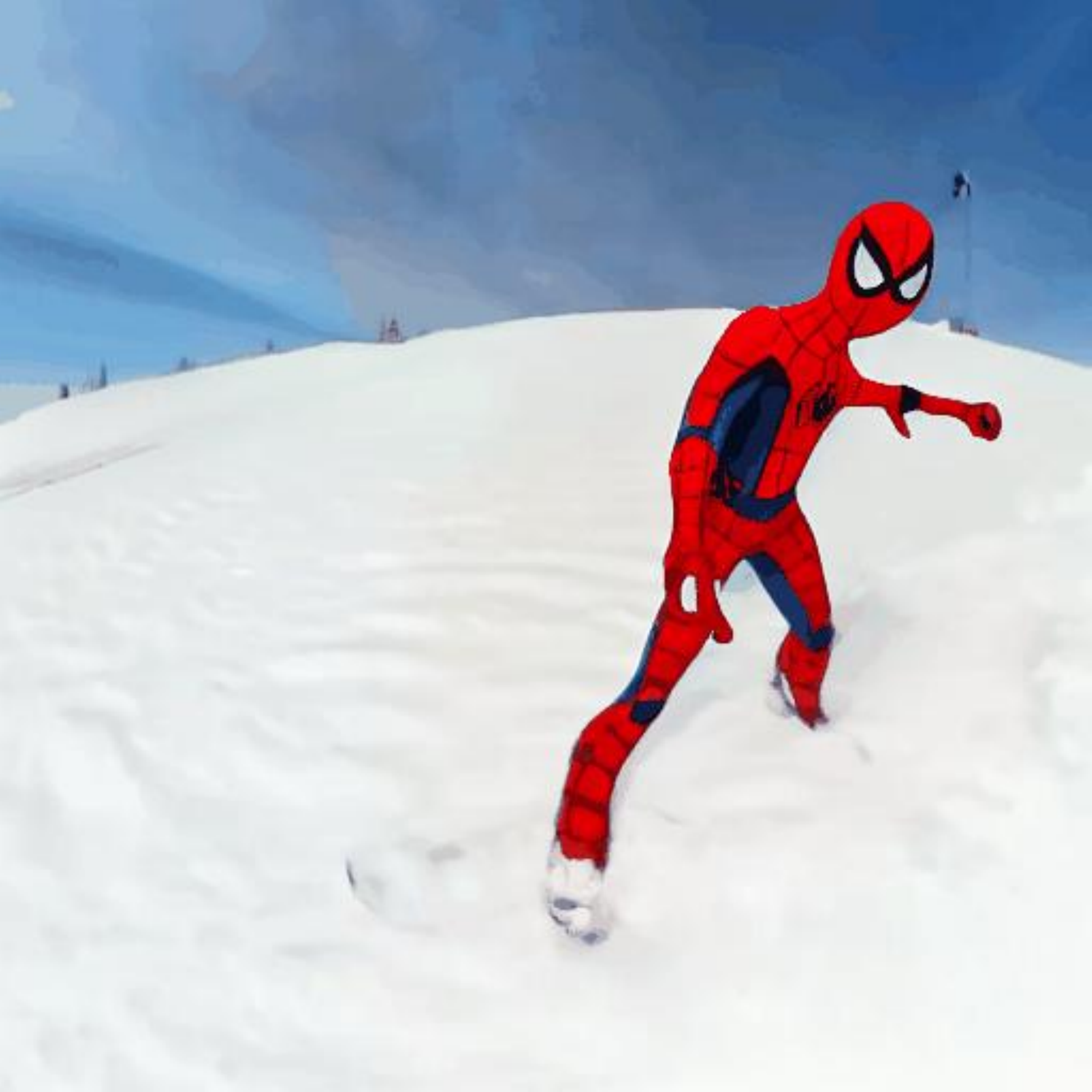}
\includegraphics[width=0.10\textwidth]{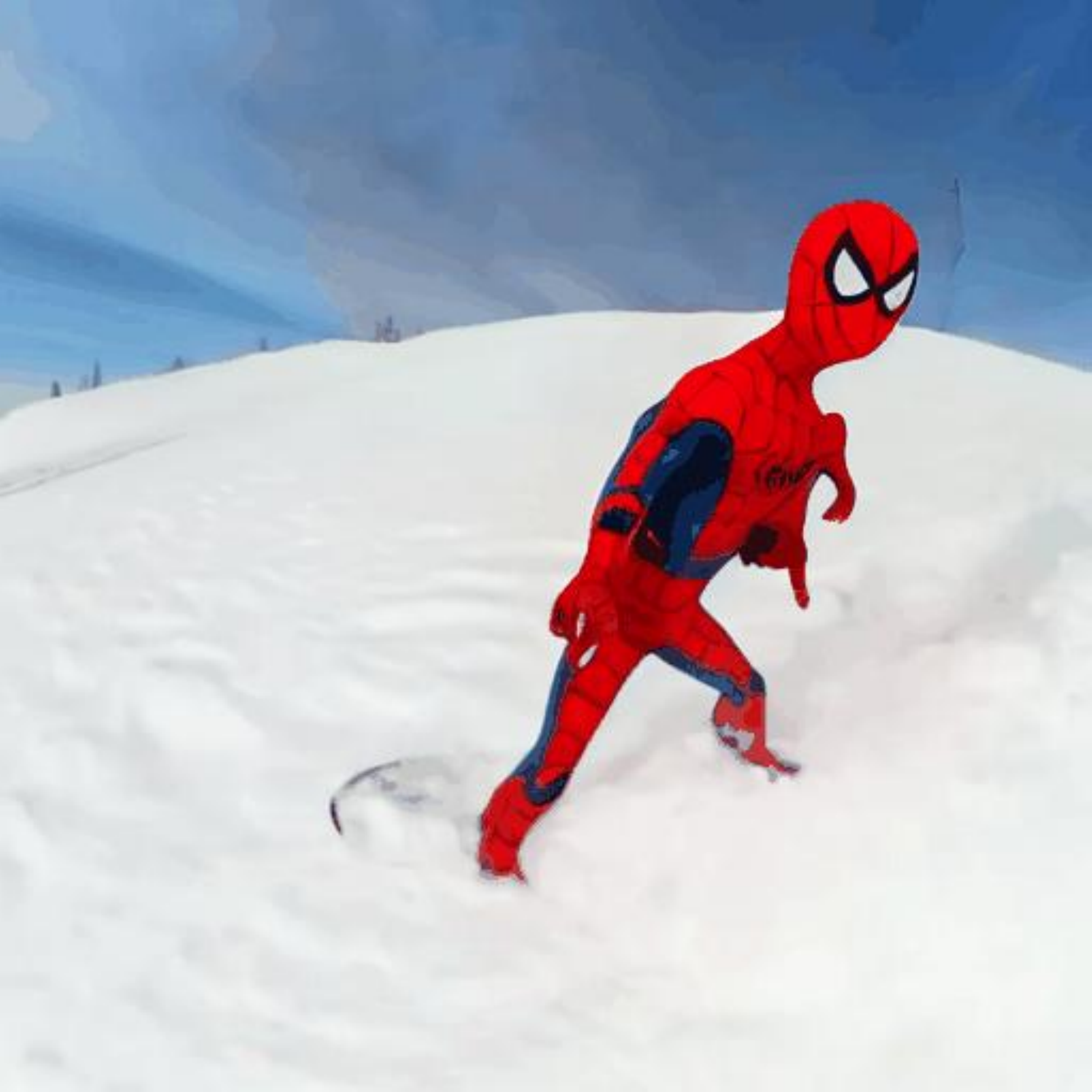}
\includegraphics[width=0.10\textwidth]{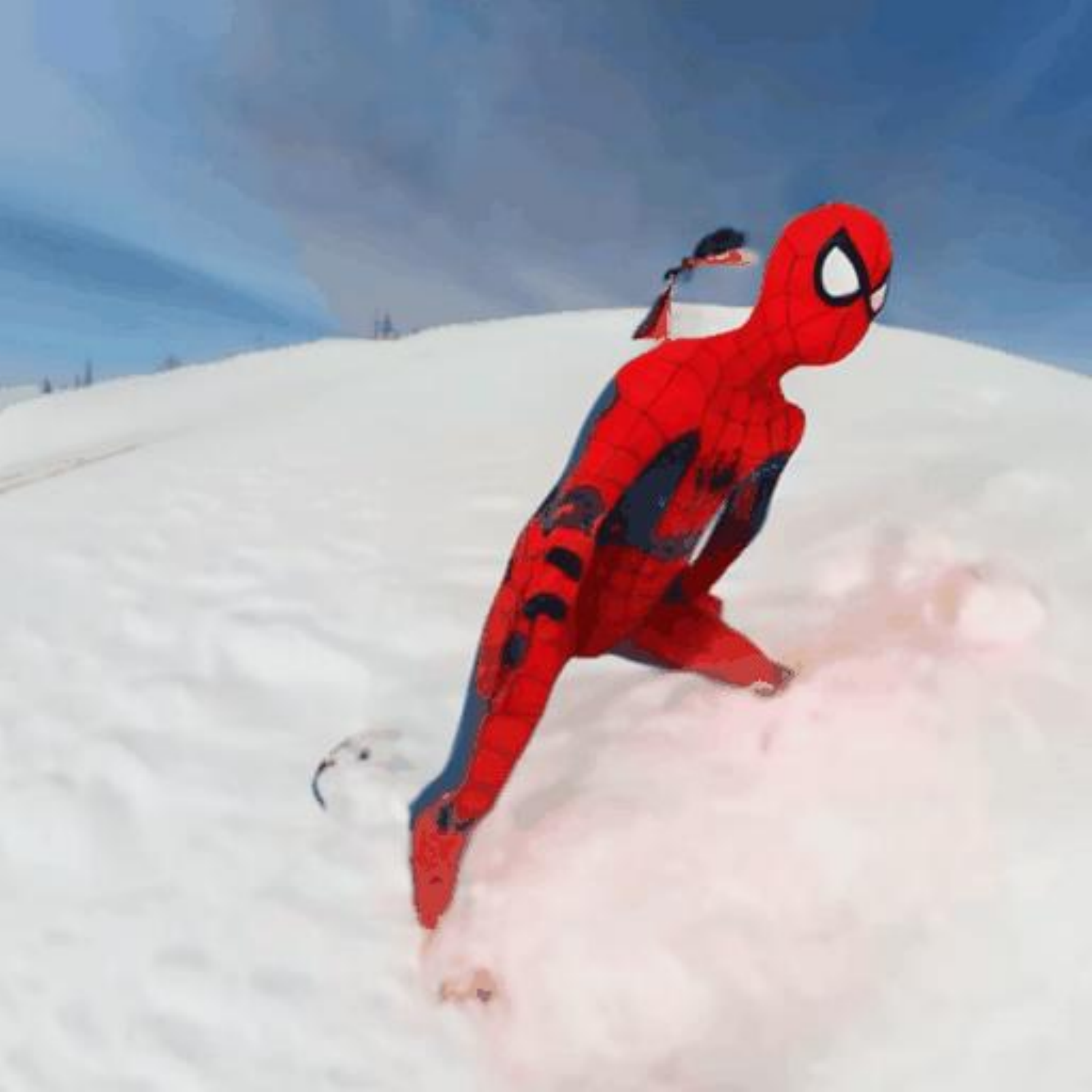}
\includegraphics[width=0.10\textwidth]{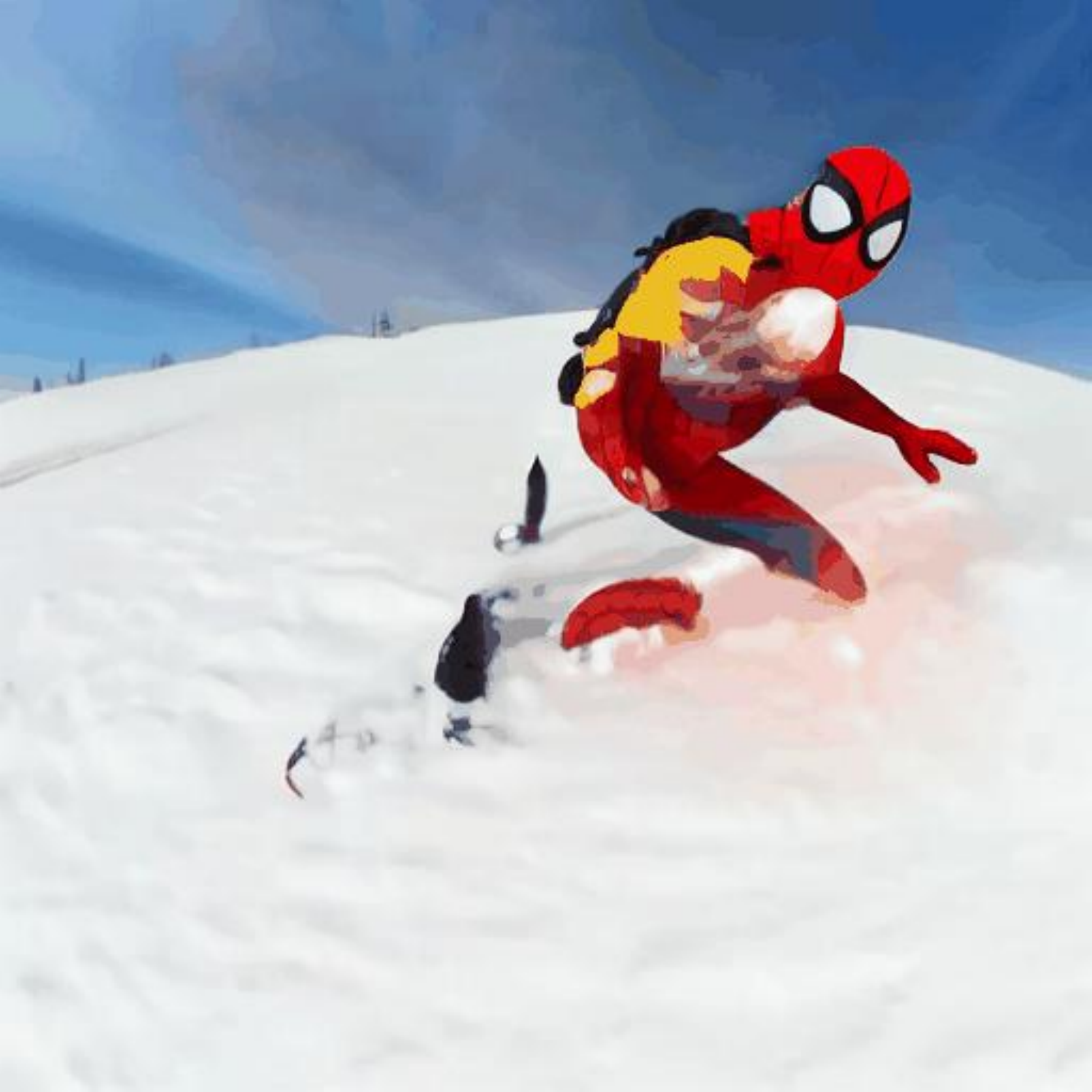}
\includegraphics[width=0.10\textwidth]{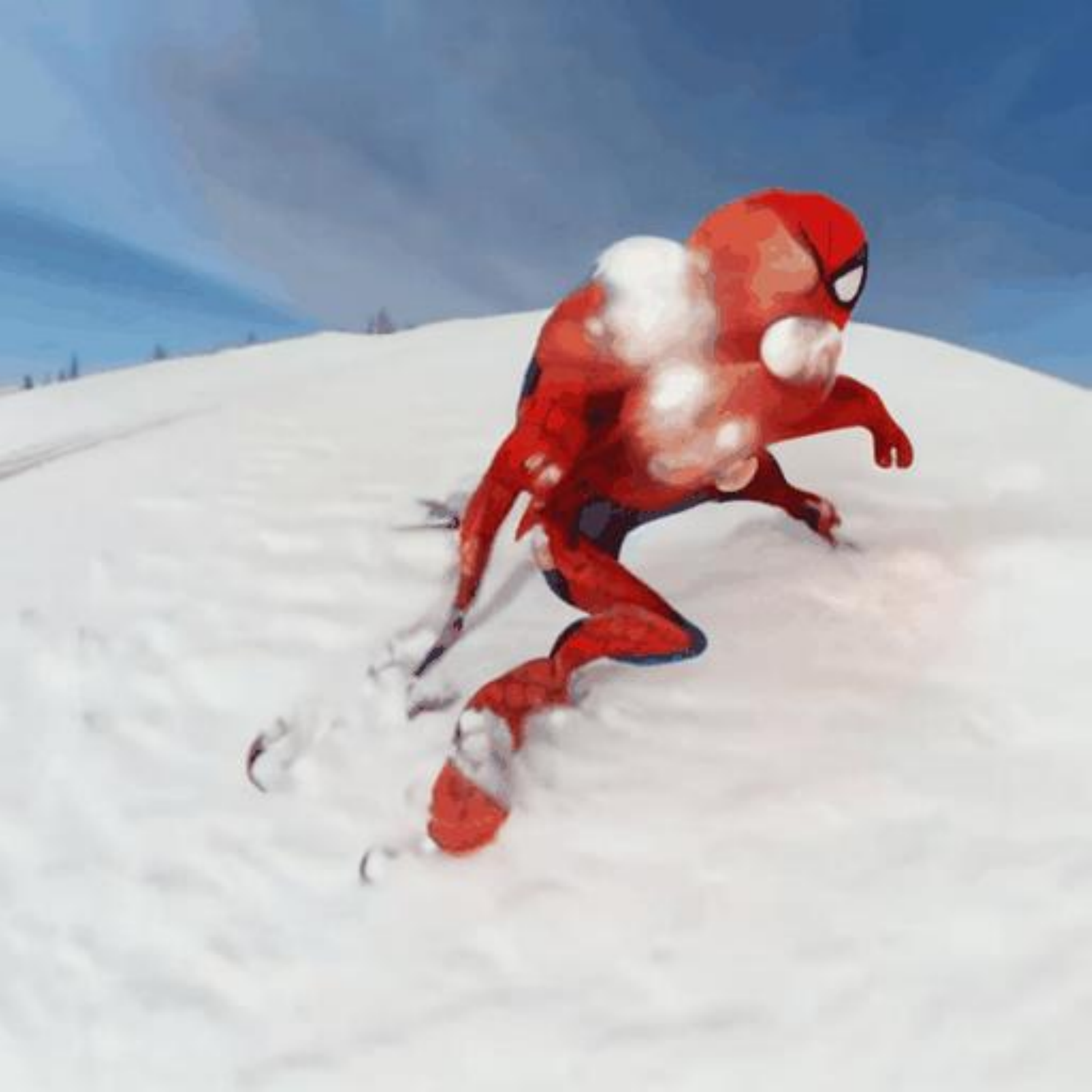}

\rotatebox{90}{\parbox{0.10\textwidth}{\centering \textbf{duration \\ 0.5}}}
\includegraphics[width=0.10\textwidth]{figures/attn_anal/self0.8_sparse0.5_cross0.2/frame_0.pdf}
\includegraphics[width=0.10\textwidth]{figures/attn_anal/self0.8_sparse0.5_cross0.2/frame_1.pdf}
\includegraphics[width=0.10\textwidth]{figures/attn_anal/self0.8_sparse0.5_cross0.2/frame_2.pdf}
\includegraphics[width=0.10\textwidth]{figures/attn_anal/self0.8_sparse0.5_cross0.2/frame_3.pdf}
\includegraphics[width=0.10\textwidth]{figures/attn_anal/self0.8_sparse0.5_cross0.2/frame_4.pdf}
\includegraphics[width=0.10\textwidth]{figures/attn_anal/self0.8_sparse0.5_cross0.2/frame_5.pdf}
\includegraphics[width=0.10\textwidth]{figures/attn_anal/self0.8_sparse0.5_cross0.2/frame_6.pdf}
\includegraphics[width=0.10\textwidth]{figures/attn_anal/self0.8_sparse0.5_cross0.2/frame_7.pdf}

\rotatebox{90}{\parbox{0.10\textwidth}{\centering duration \\ 0.8}}
\includegraphics[width=0.10\textwidth]{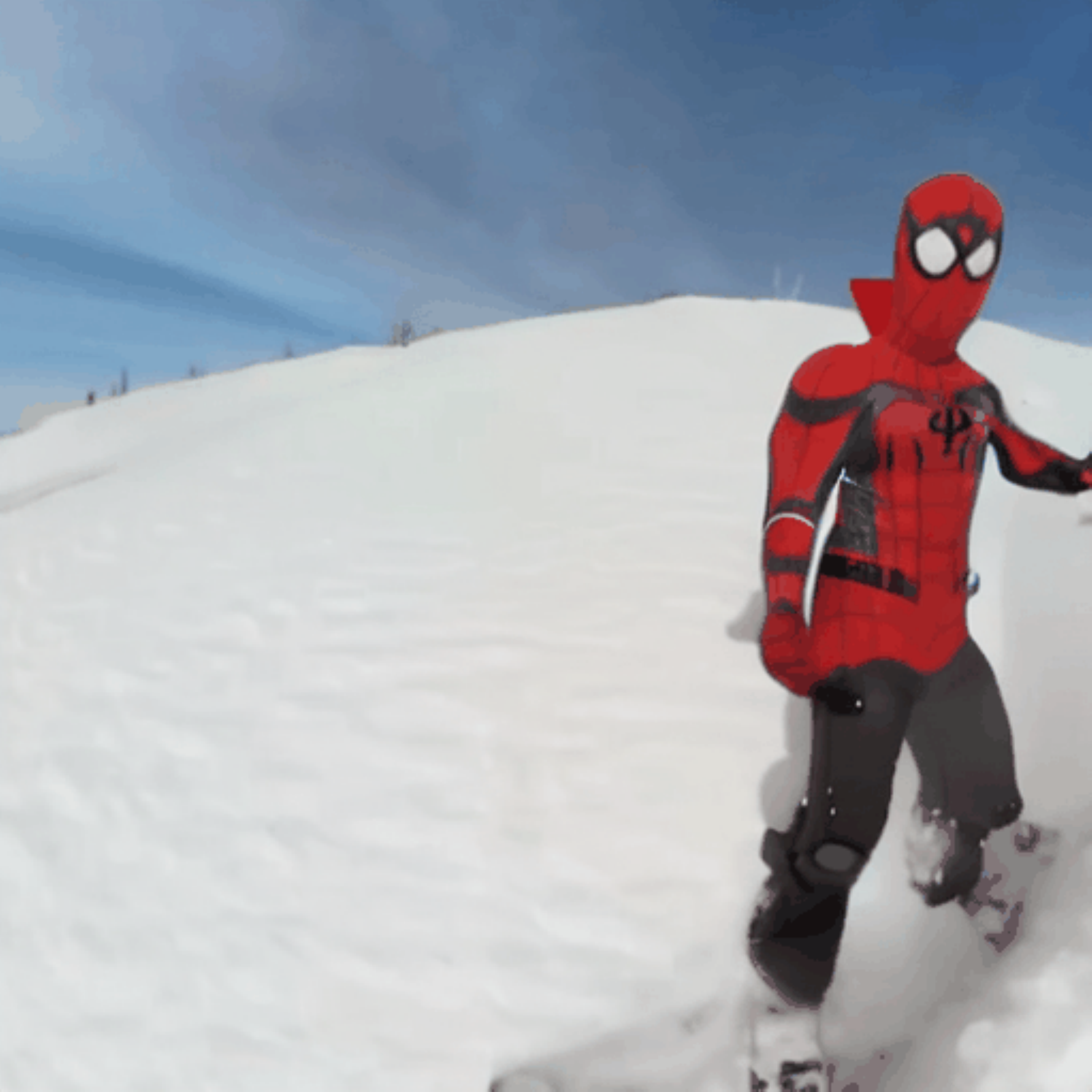}
\includegraphics[width=0.10\textwidth]{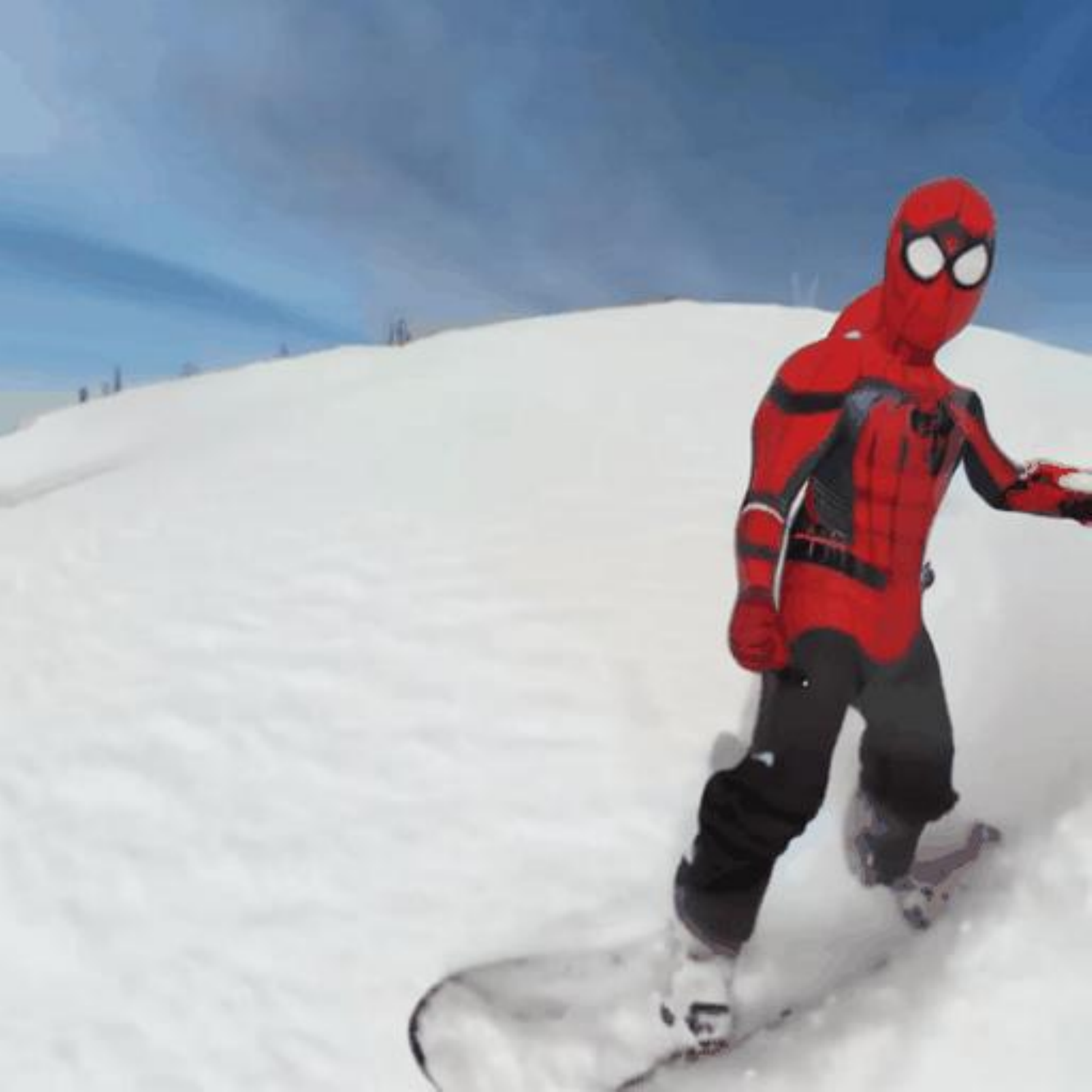}
\includegraphics[width=0.10\textwidth]{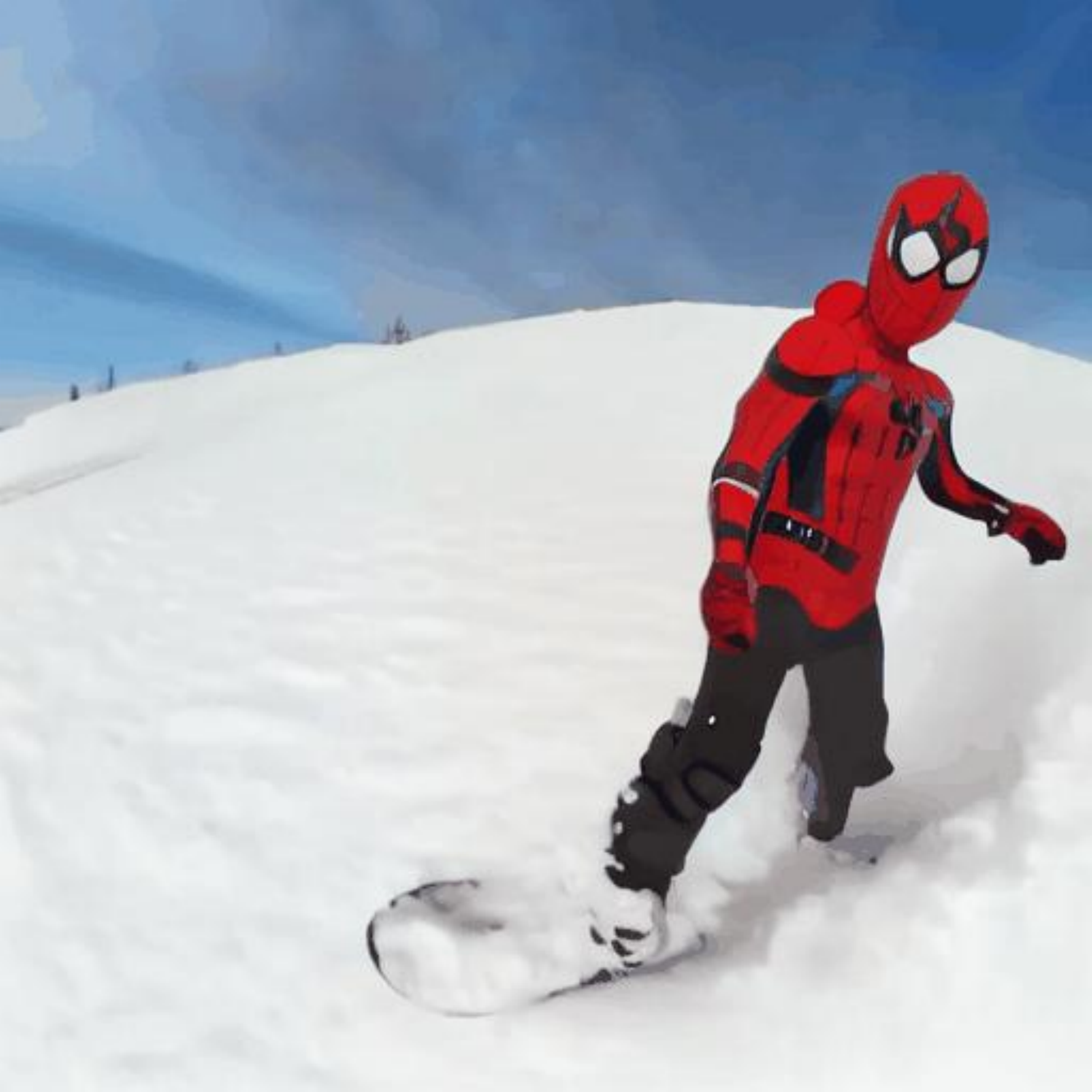}
\includegraphics[width=0.10\textwidth]{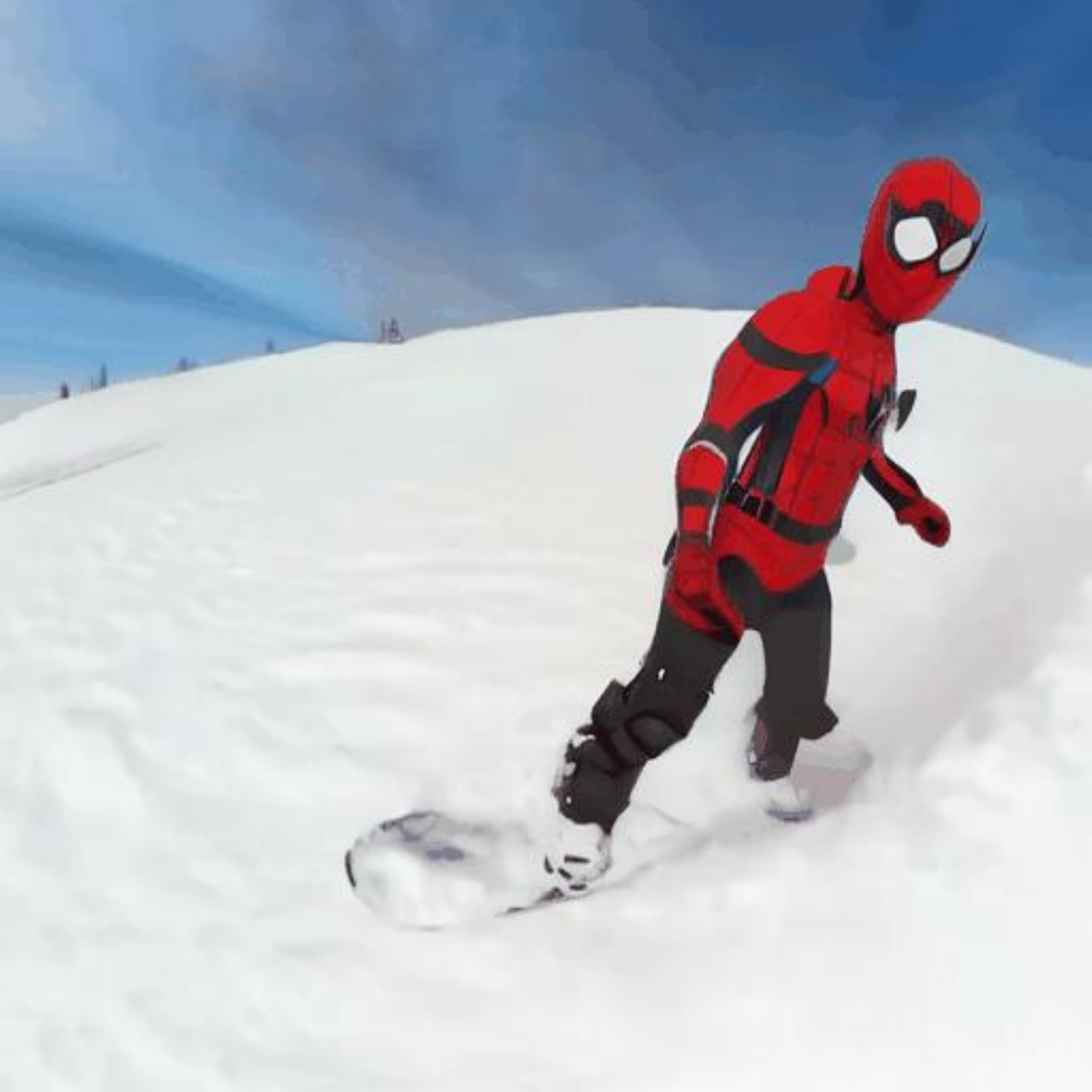}
\includegraphics[width=0.10\textwidth]{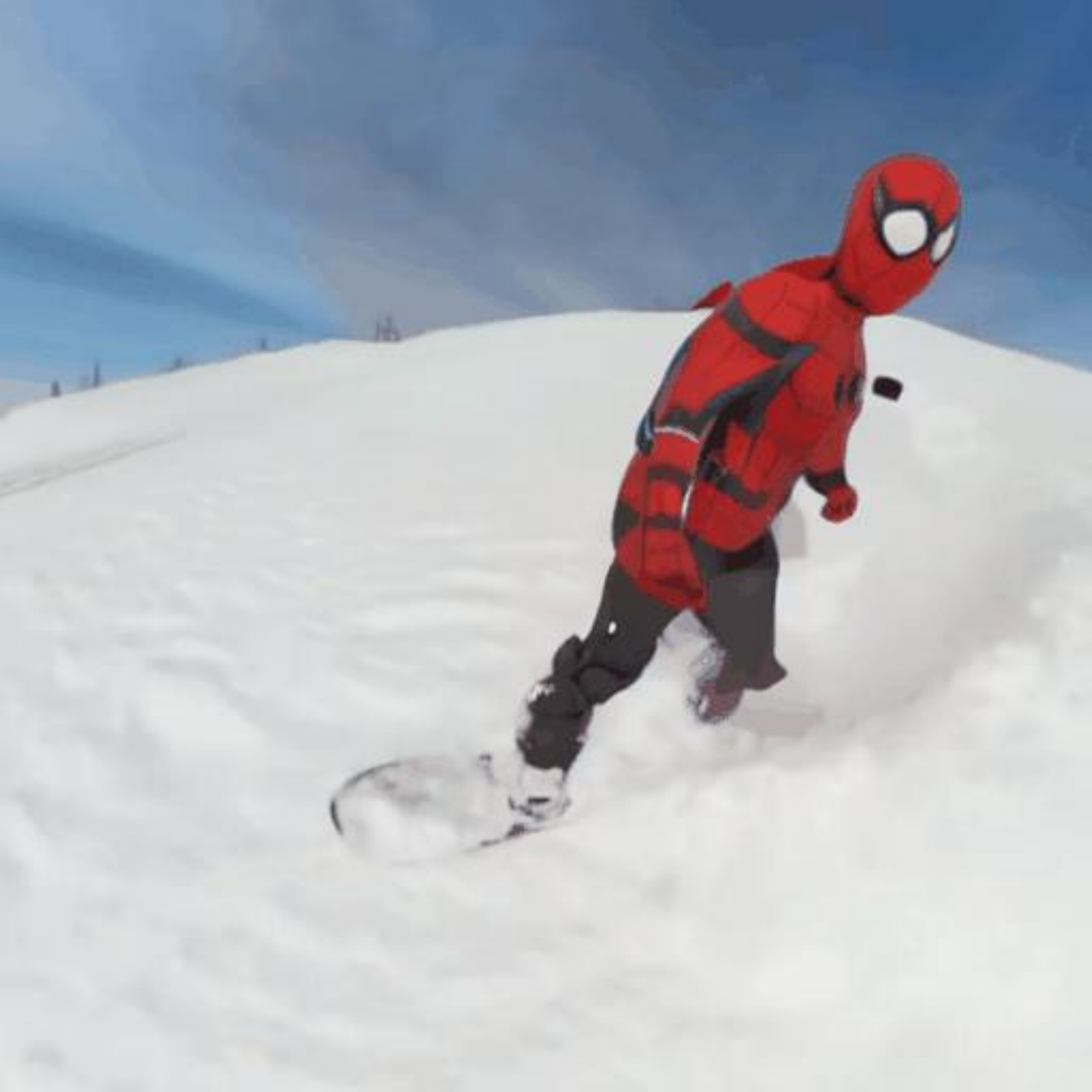}
\includegraphics[width=0.10\textwidth]{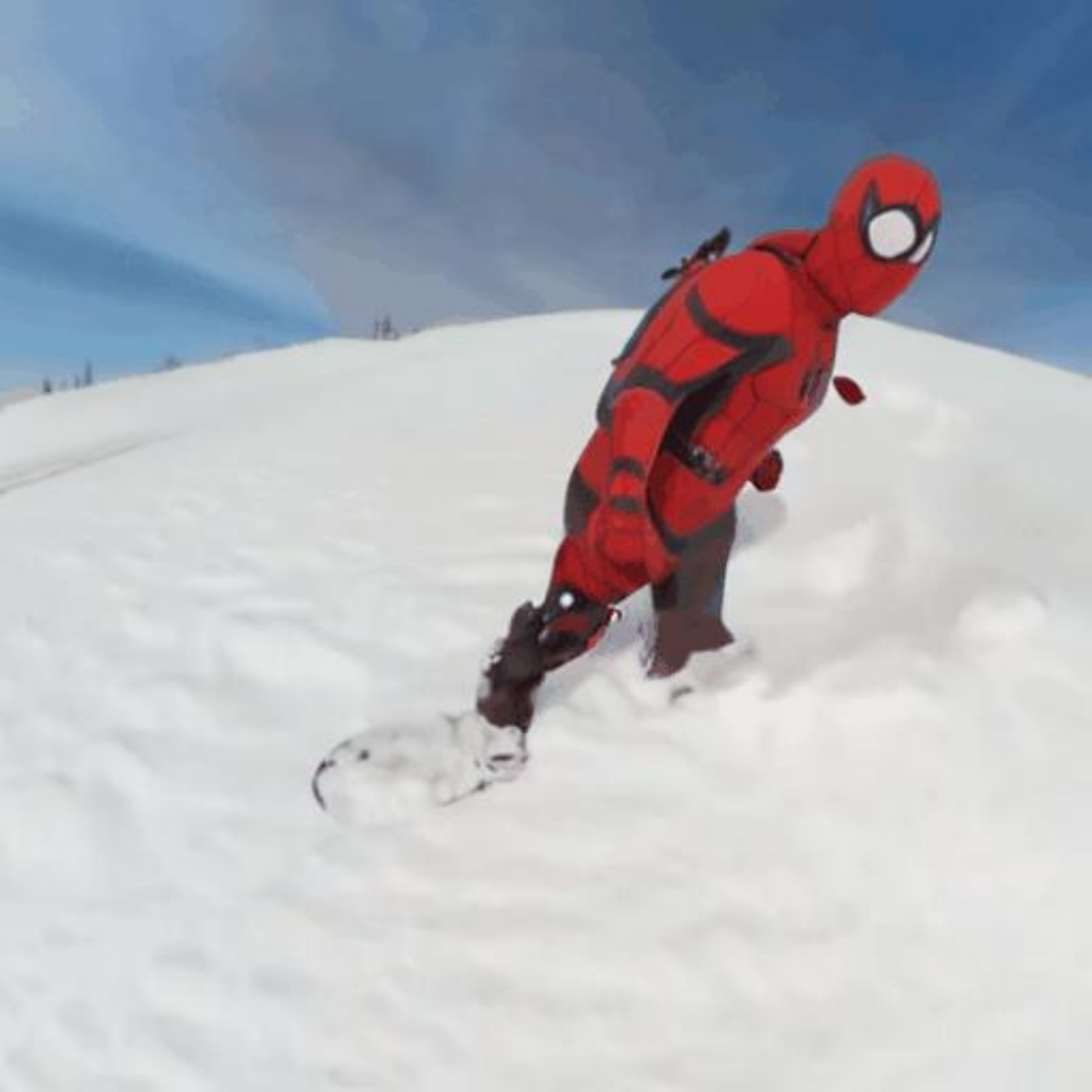}
\includegraphics[width=0.10\textwidth]{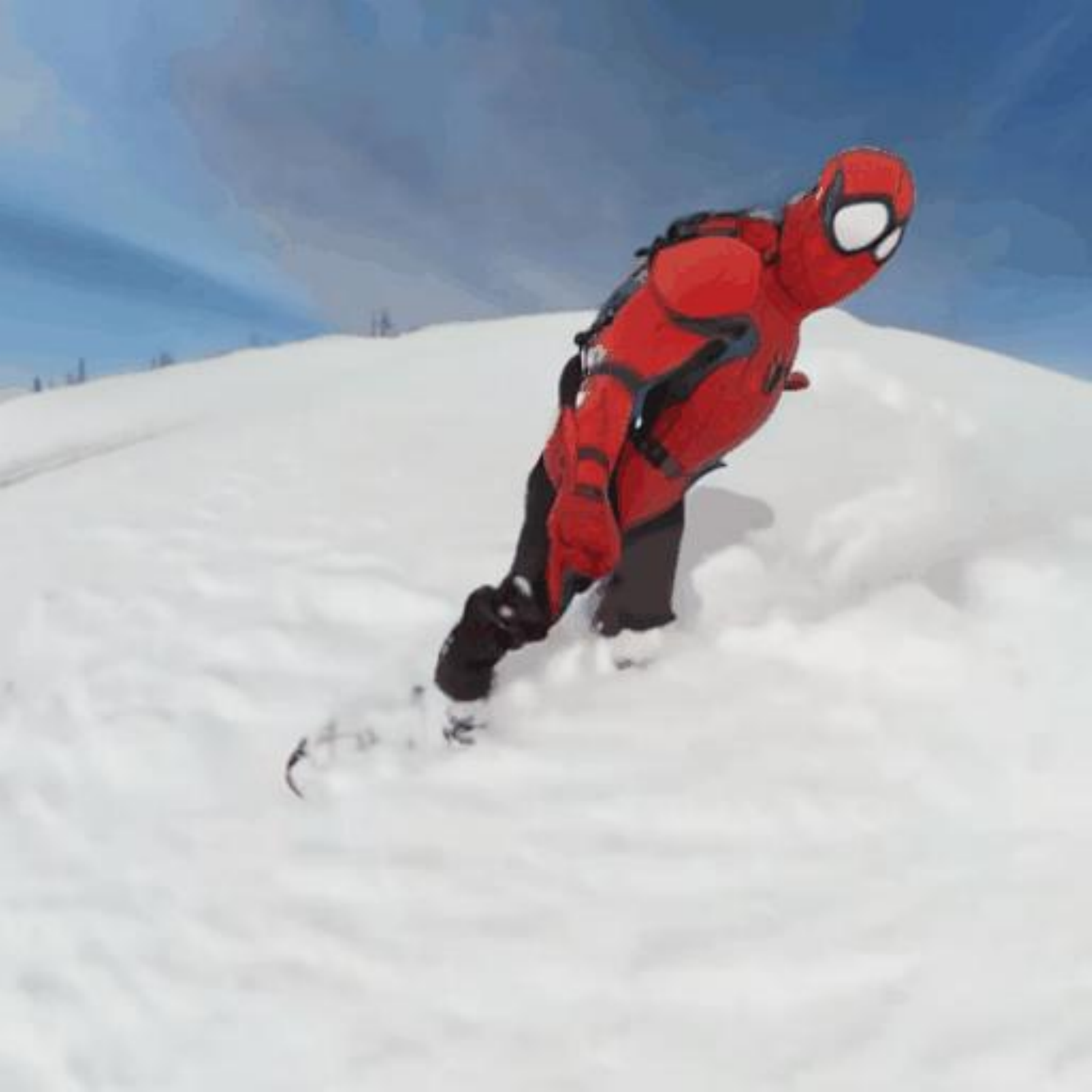}
\includegraphics[width=0.10\textwidth]{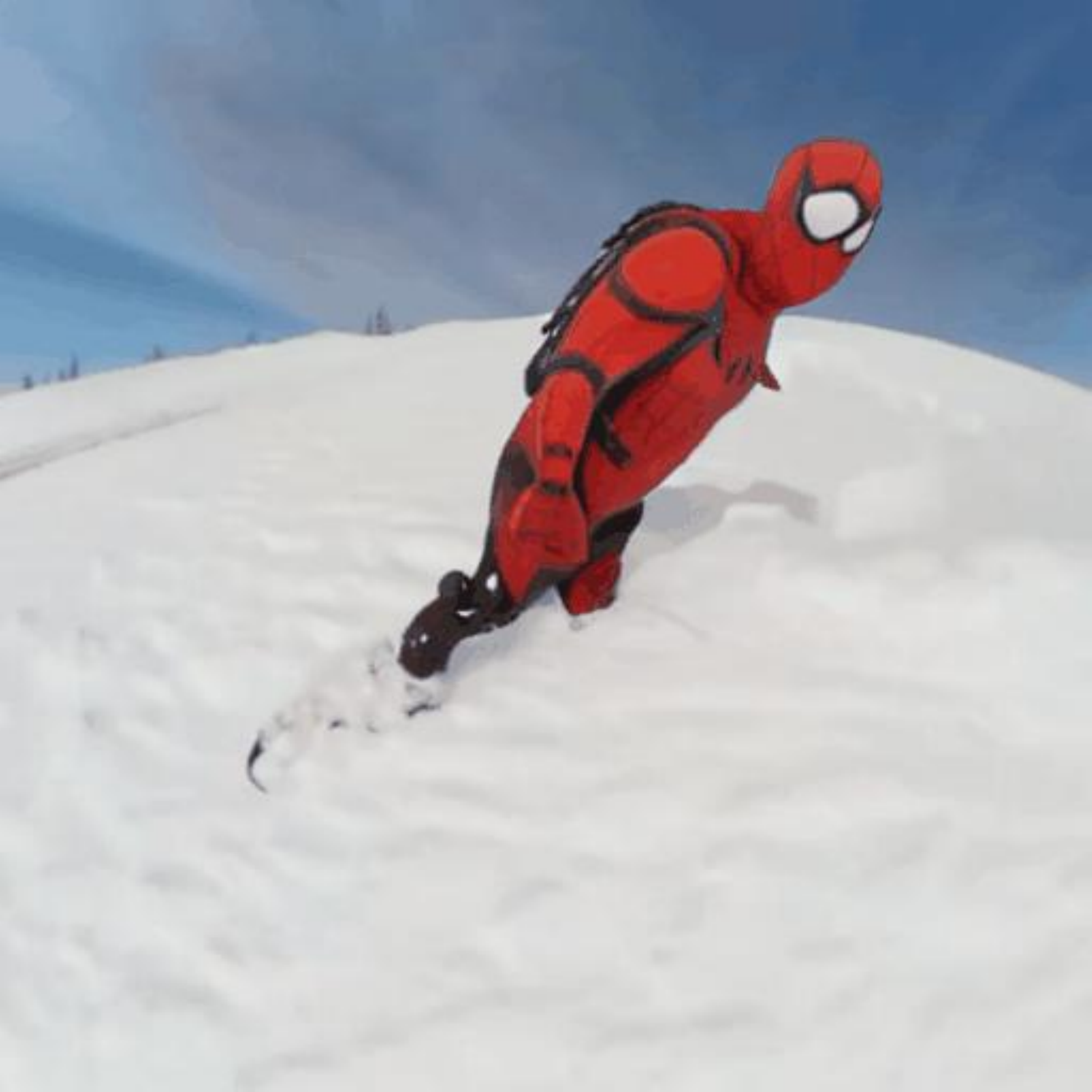}

\caption{\textbf{Attention Injection Analysis} Samples according to attention injection hyperparameters.}
\label{fig:supp_attention}
\end{center}
\vspace{-1.5em}
\end{figure*}

We show several additional samples in this section. First, various examples of our model are in Fig. \ref{fig:supp_qual1}, \ref{fig:supp_qual3}. We also compare the generated samples of our model and several baselines in Fig. \ref{fig:supp_comparison1}. 
In addition, samples according to the adjustment of various hyperparameters of attention injection are shown in Fig. \ref{fig:supp_attention}. 

\end{document}